%% file: main.tex
\documentclass[10pt,twocolumn,letterpaper]{article}

\usepackage[pagenumbers]{cvpr} %

\usepackage{times}
\usepackage{epsfig}
\usepackage{graphicx}
\usepackage{amsmath}
\usepackage{amssymb}
\usepackage{pifont}
\usepackage{array}
\usepackage{comment} 
\usepackage{capt-of}
\usepackage{booktabs}
\usepackage{tabularx}
\usepackage{soul}
\usepackage{siunitx}
\usepackage{multirow}
\usepackage{multicol}
\usepackage[dvipsnames]{xcolor}

\usepackage{appendix}

\usepackage[pagebackref,breaklinks,colorlinks]{hyperref}

\newcommand*\mc[1]{\multicolumn{2}{c}{#1}}
\newcommand{\overbar}[1]{\mkern 1.5mu\overline{\mkern-1.5mu#1\mkern-1.5mu}\mkern 1.5mu}

\usepackage[capitalize]{cleveref}
\crefname{section}{Sec.}{Secs.}
\Crefname{section}{Section}{Sections}
\Crefname{table}{Table}{Tables}
\crefname{table}{Tab.}{Tabs.}

\makeatletter
\DeclareRobustCommand\onedot{\futurelet\@let@token\@onedot}
\def\@onedot{\ifx\@let@token.\else.\null\fi\xspace}

\def\etal{\emph{et al}\onedot}

\makeatother

\begin{document}

\title{Third Time's the Charm? Image and Video Editing with StyleGAN3 \vspace{-0.25cm}}

\author{
Yuval Alaluf$^{1,*}$ \hspace{6mm}
Or Patashnik$^{1,*}$ \hspace{6mm}  
Zongze Wu$^{2}$ \hspace{6mm} 
Asif Zamir$^1$ \hspace{6mm} \\
Eli Shechtman$^3 $ \hspace{6mm}  
Dani Lischinski$^2$ \hspace{6mm}
Daniel Cohen-Or$^1$ \hspace{6mm} \\ \\
\vspace{-0.15cm}
$^1$Tel-Aviv University \hspace{10mm} $^2$Hebrew University of Jerusalem \hspace{10mm} $^3$Adobe Research
}

\maketitle

\def\thefootnote{*}\footnotetext{Denotes equal contribution.}

\vspace{-0.2cm}
\begin{abstract}
\input{abstract}
\end{abstract}

\begin{figure}
    \centering
    \vspace{-0.1cm}
    \includegraphics[width=\linewidth]{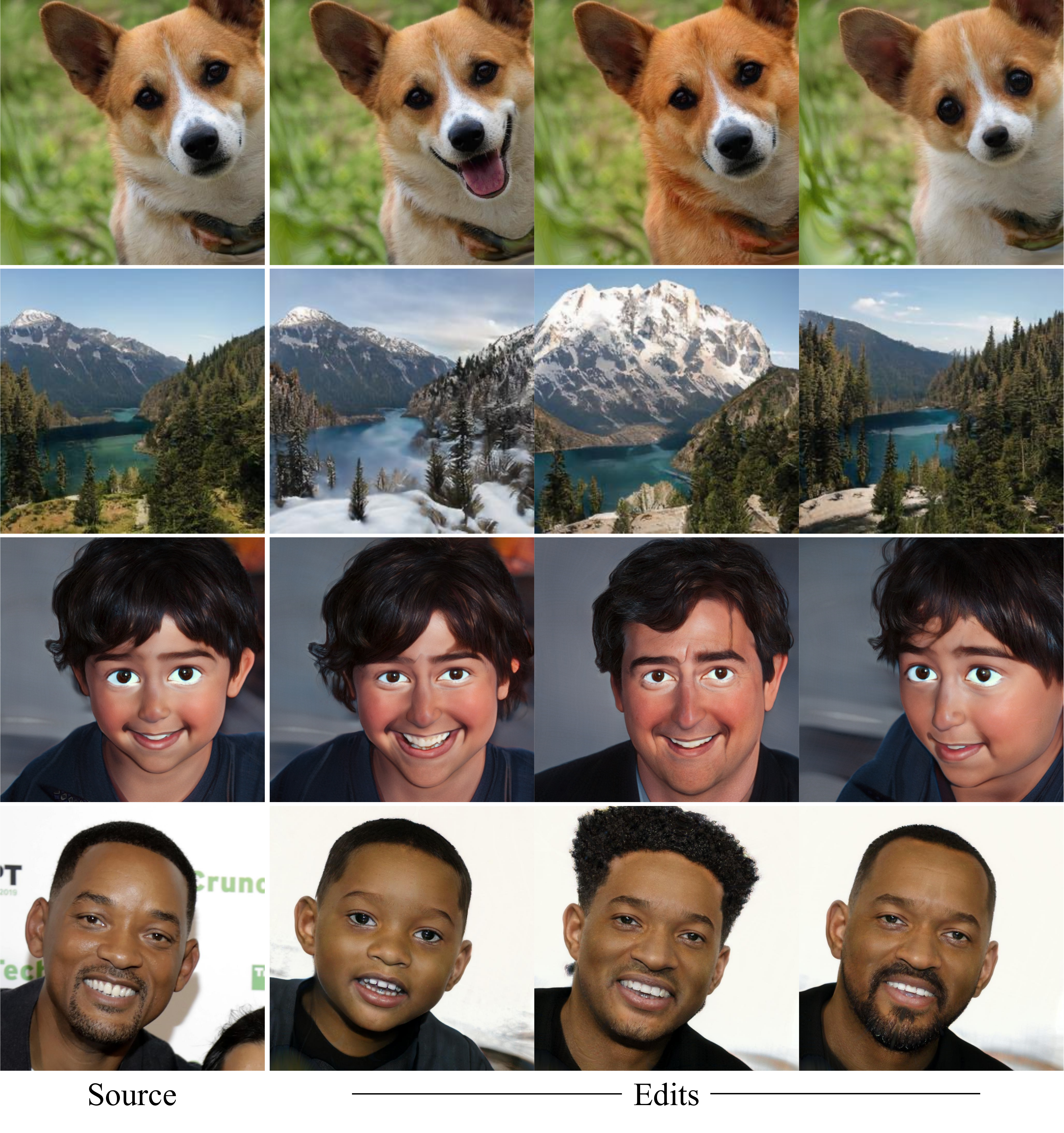}
    \vspace{-0.8cm}
    \caption{
    \textit{Image editing with StyleGAN3.} Using the recent StyleGAN3 generator, we edit unaligned input images across various domains using off-the-shelf editing techniques. Using a trained StyleGAN3 encoder, these techniques can likewise be used to edit real images and videos.
    }
    \vspace{-0.25cm}
    \label{fig:teaser}
\end{figure}

\vspace{-0.1cm}
\input{introduction}
\input{related_works}
\input{preliminaries}

\input{analysis}
\input{editing}
\input{inversion}
\input{video}
\input{conclusion}

{%
\bibliographystyle{ieee_fullname}
\bibliography{main}
}

\clearpage
\appendix
\appendixpage
\input{appendix}

\end{document}

%% file: abstract.tex
StyleGAN is arguably one of the most intriguing and well-studied generative models, demonstrating impressive performance in image generation, inversion, and manipulation. In this work, we explore the recent StyleGAN3 architecture, compare it to its predecessor, and investigate its unique advantages, as well as drawbacks. In particular, we demonstrate that while StyleGAN3 can be trained on unaligned data, one can still use aligned data for training, without hindering the ability to generate unaligned imagery. Next, our analysis of the disentanglement of the different latent spaces of StyleGAN3 indicates that the commonly used $\mathcal{W}/\mathcal{W}+$ spaces are more entangled than their StyleGAN2 counterparts, underscoring the benefits of using the \emph{StyleSpace} for fine-grained editing. Considering image inversion, we observe that existing encoder-based techniques struggle when trained on unaligned data. We therefore propose an encoding scheme trained solely on aligned data, yet can still invert unaligned images. Finally, we introduce a novel video inversion and editing workflow that leverages the capabilities of a fine-tuned StyleGAN3 generator to reduce texture sticking and expand the field of view of the edited video.
Code is available on our project page: \small{\url{https://yuval-alaluf.github.io/stylegan3-editing/}}.

%% file: introduction.tex
\section{Introduction}

In recent years, Generative Adversarial Networks (GANs)~\cite{Goodfellow2014GenerativeAN} have revolutionized image processing. Specifically, StyleGAN generators~\cite{karras2019style, karras2020analyzing, karras2020training, aliasfreeKarras2021} synthesize exceedingly realistic images, and enable editing~\cite{alaluf2021matter,patashnik2021styleclip,abdal2020styleflow,bau2021paint,Collins2020EditingIS,zhu2021barbershop,ling2021editgan}, and image-to-image translation~\cite{richardson2020encoding,park2020swapping,menon2020pulse}, particularly in well-structured domains. StyleGAN architectures are notable for their semantically-rich, disentangled, and generally well-behaved latent spaces. 

When it comes to video processing, new challenges arise. Editing should not only be disentangled and realistic, but also temporally consistent across frames. The texture-sticking phenomenon in StyleGAN1 and StyleGAN2~\cite{aliasfreeKarras2021} hinders the temporal consistency and realism of generated and manipulated videos. For example, when interpolating within the latent space, the hair and face typically do not move in unison. The recent StyleGAN3 architecture~\cite{aliasfreeKarras2021} is specifically designed to overcome such texture-sticking, and additionally offers translation and rotation equivariance. Naturally, these unique properties make StyleGAN3 better suited for video processing than previous style-based generators. However, significant changes introduced in StyleGAN3 architecture raise many questions and new challenges. Central among these is the disentanglement of its latent spaces and the ability to accurately invert and edit real images.

In this paper, we analyze StyleGAN3, aiming at understanding and exploring its capabilities and performance. 
Some of the questions we attempt to answer are: How does the disentanglement of the latent representations in StyleGAN3 compare to StyleGAN2? Do the techniques devised for identifying latent editing controls still work?
Inversion is a fundamental task required for editing \textit{real} images, and has been extensively studied in the context of StyleGAN2 \cite{abdal2019image2stylegan, abdal2020image2stylegan++, zhu2020domain, richardson2020encoding, tov2021designing, alaluf2021restyle, roich2021pivotal, alaluf2021hyperstyle}. We therefore examine how well these existing techniques can be adapted to achieve comparable performance with StyleGAN3, and, in particular, cope with the inversion of unaligned images.

In light of the translation and rotation equivariance provided by StyleGAN3, we also examine the differences between generators trained on aligned and on unaligned data. Surprisingly, we observe that both kinds of generators are comparable in terms of their ability to generate unaligned images and control their position and rotation. However, we find that using the aligned generators is preferable for tasks such as disentangled editing and inversion. We therefore leverage aligned generators in our proposed image and video inversion and editing workflows, see \cref{fig:teaser}.

Applying the insights gained in our analysis and experiments over still images, we propose a novel workflow for inverting and editing real videos using StyleGAN3. Notably, we leverage the capabilities of StyleGAN3 to reduce texture sticking and expand the field of view when working on a video with a cropped subject.

%% file: related_works.tex
\section{Related Work}

StyleGAN2~\cite{karras2020analyzing} features several semantically-rich latent spaces, which have been heavily studied and exploited in the context of image manipulation~\cite{shen2020interpreting,tewari2020stylerig,abdal2020styleflow,harkonen2020ganspace,shen2020closedform,Collins2020EditingIS} and image inversion~\cite{abdal2019image2stylegan,abdal2020image2stylegan++,semantic2019bau,guan2020collaborative,pidhorskyi2020adversarial,Zhu2020ImprovedSE}. 
In this work, we experiment with the image manipulation techniques from Shen \etal~\cite{shen2020interpreting}, Wu \etal~\cite{wu2021stylespace} and Patashnik \etal~\cite{patashnik2021styleclip} in the context of StyleGAN3~\cite{aliasfreeKarras2021}.

To manipulate real images they must first be projected into one of the StyleGAN latent spaces. 
We refer the reader to the exposition in Xia~\etal~\cite{xia2021gan} for a comprehensive review of GAN inversion and the applications it enables.
In this work, we leverage existing encoder-based inversion techniques~\cite{richardson2020encoding,tov2021designing,alaluf2021restyle} for achieving more accurate inversions of images, and in particular unaligned images, using StyleGAN3. We additionally employ existing generator tuning techniques~\cite{roich2021pivotal} to achieve higher-fidelity reconstructions of a wide range of facial expressions, which we find to be necessary for inverting and editing videos.

In contrast to editing images with StyleGAN, few works have addressed video-based attribute editing. Most notably, Yao~\etal~\cite{yao2021latent} use a pre-trained encoder and introduce a latent space transformer to achieve consistent edits of the inverted video frames. However, their method relies on facial alignment, segmentation, and Poisson blending~\cite{perez2003poisson}. We aim to leverage StyleGAN3's rotation and translation equivariance to achieve accurate and consistent video editing while reducing the overhead of previous techniques.

We refer the reader to \cref{sec:related_appendix} for additional background on StyleGAN's latent spaces, inversion techniques, and the editing capabilities it offers.

%% file: preliminaries.tex
\setlength{\abovedisplayskip}{4pt}
\setlength{\belowdisplayskip}{4pt}

\section{The StyleGAN3 Architecture}\label{sec:preliminaries}

To better understand the capabilities of StyleGAN3~\cite{aliasfreeKarras2021}, it is important to understand the overall structure and function of the different components comprising the architecture. First, as in StyleGAN~\cite{karras2019style}, a simple fully-connected mapping network translates an initial latent code $z \sim \mathcal{N}\!\left(0,1\right)^{512}$, into an intermediate code $w$ residing in a learned latent space $\mathcal{W}$. 

Compared to StyleGAN2~\cite{karras2020analyzing}, StyleGAN3's synthesis network is composed of a fixed number of convolutional layers ($16$), irrespective of the output image resolution. We denote by $(w_0, ..., w_{15})$ the set of input codes passed to these layers. In StyleGAN3, the constant $4\times4$ input tensor from StyleGAN2 is replaced by Fourier features, that can be rotated and translated using four parameters ($\sin\alpha$, $\cos\alpha$, $x$, $y$), obtained from $w_0$ via a learned affine layer. 
In the remaining layers, each $w_i$ is fed into an independently learned affine layer, which yields modulation factors used to adjust the convolutional kernel weights. 

\input{figures/user_controlled_transformations_aligned}

\input{figures/user_controlled_transformations_unaligned}

In StyleGAN2, the space spanned by the outputs of these affine layers has been referred to as the \textit{StyleSpace}~\cite{wu2021stylespace}, or $\mathcal{S}$. In this work, we similarly define the $\mathcal{S}$ space of StyleGAN3, with $9,894$ dimensions for a $1024\times1024$ generator.

\label{section:pose_correction}
Since the translation and in-plane rotation of the synthesized images are given by explicit parameters obtained from $w_0$, the result may be easily adjusted by concatenating another transformation. We parameterize this transformation using three parameters $(r, t_x, t_y)$, where $r$ is the rotation angle (in degrees), and $t_x, t_y$ are the translation parameters, and denote the resulting image by:
\begin{equation}
    y = G(w; (r, t_x, t_y)),
\end{equation}
where, by default, $t_x = t_y = 0$ and $r=0$. This transformation can be applied even in a generator trained solely on aligned data, enabling it to generate rotated and translated images, see \cref{fig:pseudo-alignment1}.

Conversely, generators trained on unaligned data may be ``coerced'' to generate roughly aligned images by setting $w_0$ to the generator's average latent code $\overline{w}$, i.e, given by $G((\overbar{w}, w_1, ..., w_{15}); (0,0,0))$. 
\cref{fig:pseudo-alignment2} demonstrates this idea on two different StyleGAN3 generators trained on unaligned FFHQ and AFHQ datasets, respectively.
Intuitively, this approximate alignment may be due to the fact that the average input pose
in the training distribution is roughly aligned and centered, combined with the fact that the translation and rotation transformations in StyleGAN3 are mainly controlled by the first layer.
It is important to note that equivariance is at the core of the StyleGAN3 design: translations or rotations in earlier layers are preserved across later layers and appear in the generated output.

%% file: figures/user_controlled_transformations_aligned.tex
\begin{figure}[tb]
	\centering
	\setlength{\tabcolsep}{1pt}	
	{\small
	\begin{tabular}{c c c c c c c}
	
		\includegraphics[width=0.24\columnwidth]{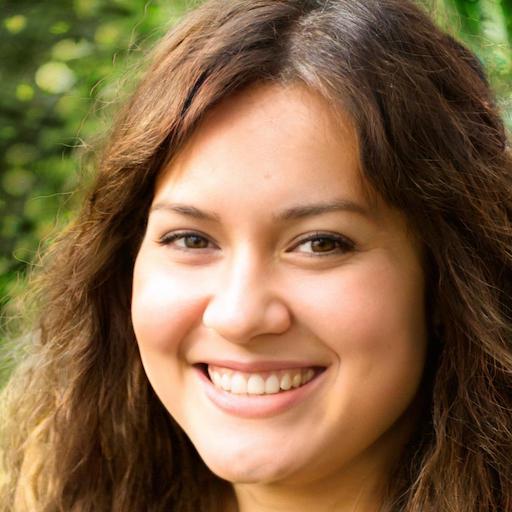} & 
		\includegraphics[width=0.24\columnwidth]{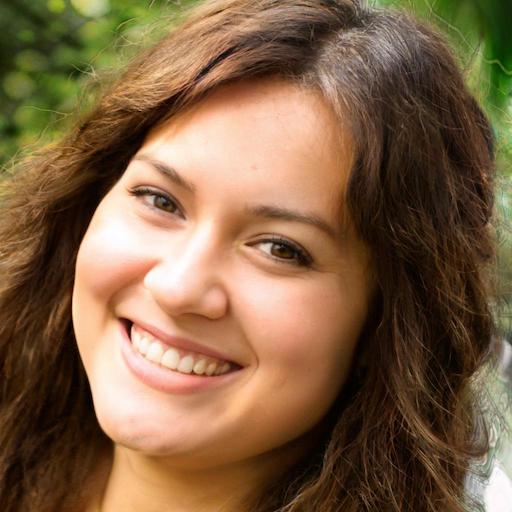} &
		\includegraphics[width=0.24\columnwidth]{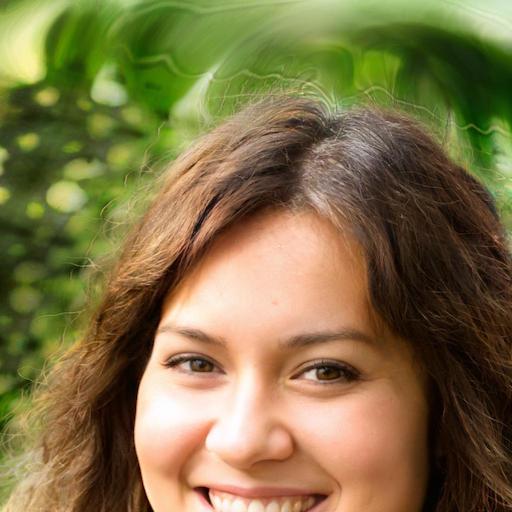} &
		\includegraphics[width=0.24\columnwidth]{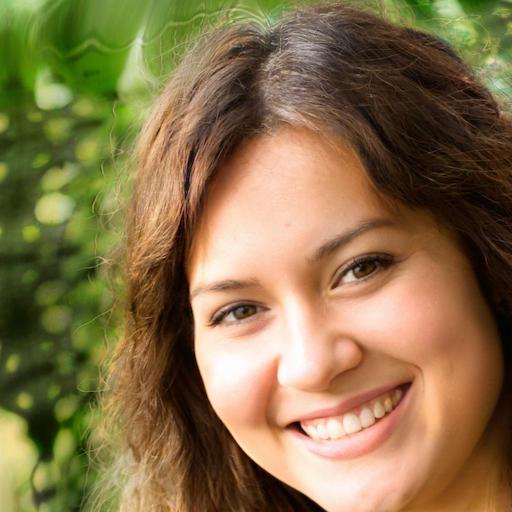} \\
	    
	    $(0^\circ,0,0)$ &	$(-20^\circ,0,0)$ & $(0^\circ,0,0.25)$ &
	    $(20^\circ,0.1,0.1)$ \\
    
	\end{tabular}
	}
	\vspace{-0.3cm}
	\caption{%
		While StyleGAN3 trained on aligned data normally generates aligned images (leftmost image), translation and in-plane rotation can be controlled by applying an explicit transformation $(r,t_x,t_y)$ over the Fourier features.
    }
	\label{fig:pseudo-alignment1}
	\vspace{-0.3cm}
\end{figure}

%% file: figures/user_controlled_transformations_unaligned.tex
\begin{figure}[tb]
	\centering
	\setlength{\tabcolsep}{1pt}	
	\begin{tabular}{c c c c c c c c}
	
    	\vspace{-0.075cm}
	
		\raisebox{0.1in}{\rotatebox{90}{\footnotesize $w_0 = w$}} &
		\raisebox{0.1in}{\rotatebox{90}{\footnotesize $(0, 0, 0)$}} &
		\includegraphics[width=0.175\columnwidth]{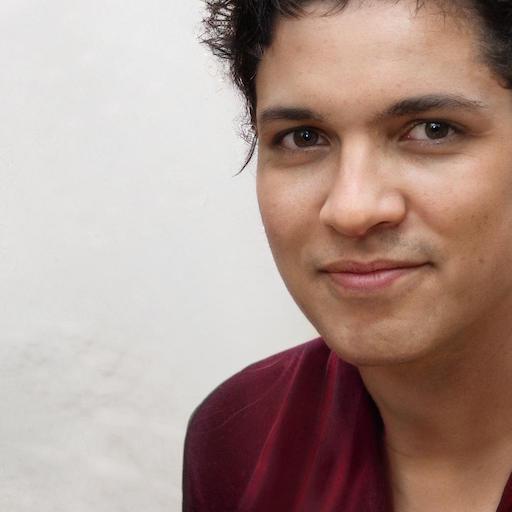} & 
		\includegraphics[width=0.175\columnwidth]{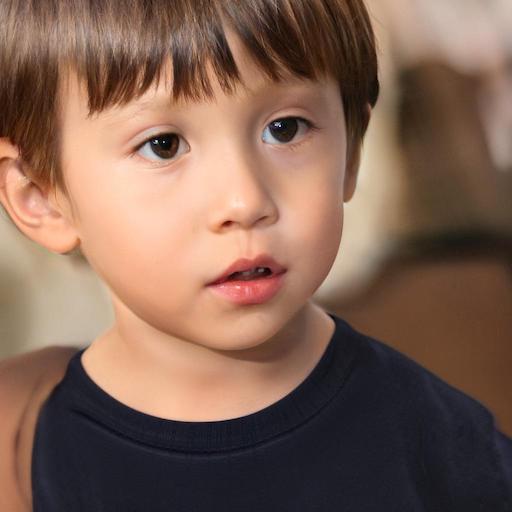} & 
		\includegraphics[width=0.175\columnwidth]{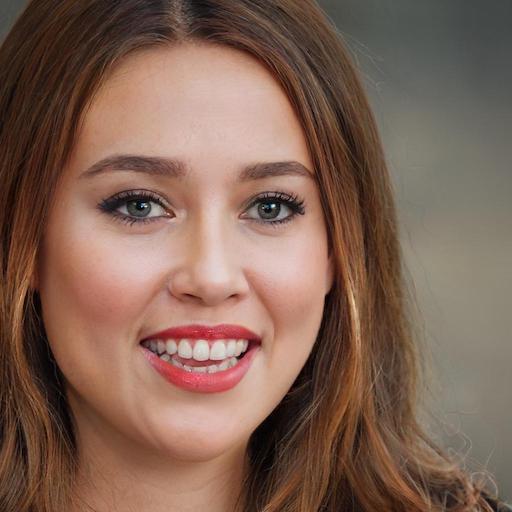} &
		\includegraphics[width=0.175\columnwidth]{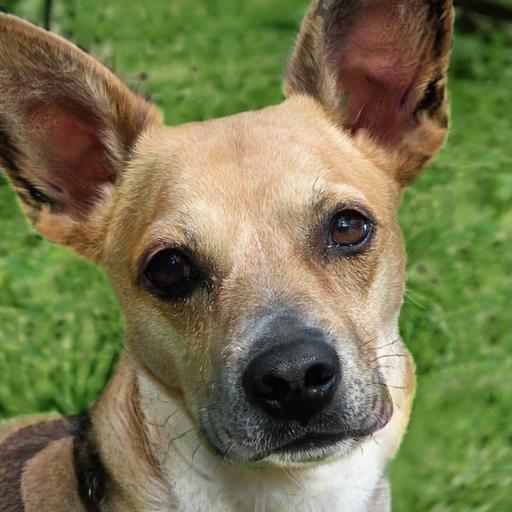} & 
		\includegraphics[width=0.175\columnwidth]{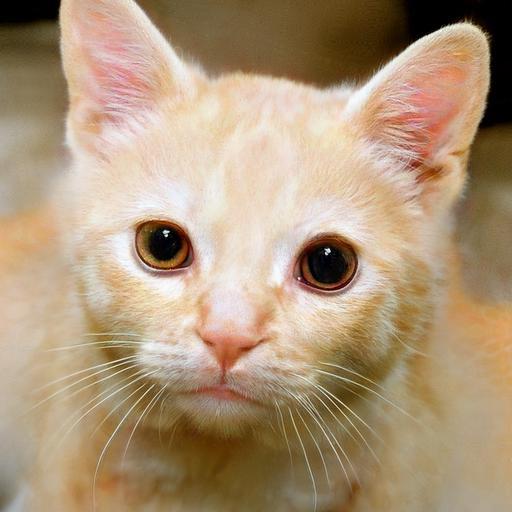} \\
		
		\raisebox{0.1in}{\rotatebox{90}{\footnotesize $w_0 = \overbar{w}$}} &
		\raisebox{0.1in}{\rotatebox{90}{\footnotesize $(0, 0, 0)$}} &
		\includegraphics[width=0.175\columnwidth]{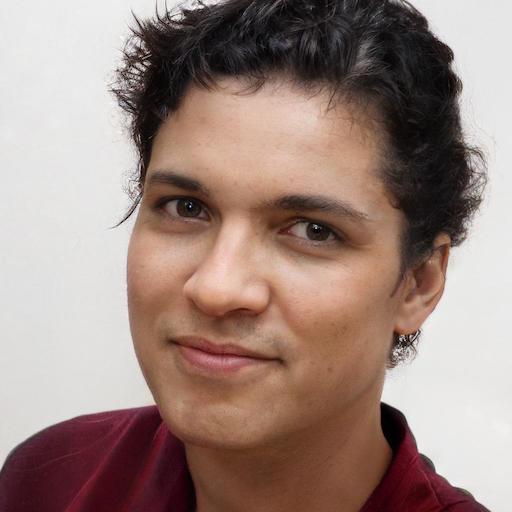} & 
		\includegraphics[width=0.175\columnwidth]{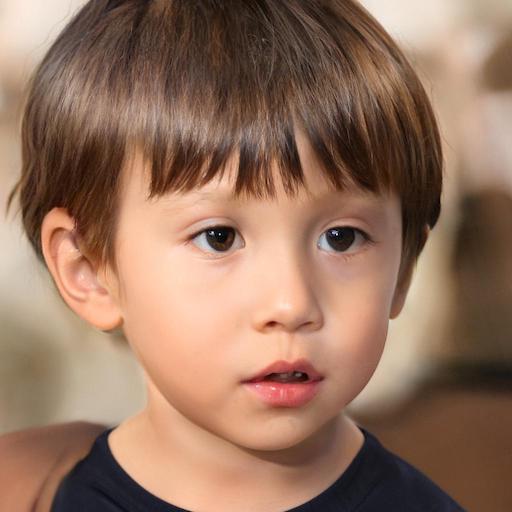} & 
		\includegraphics[width=0.175\columnwidth]{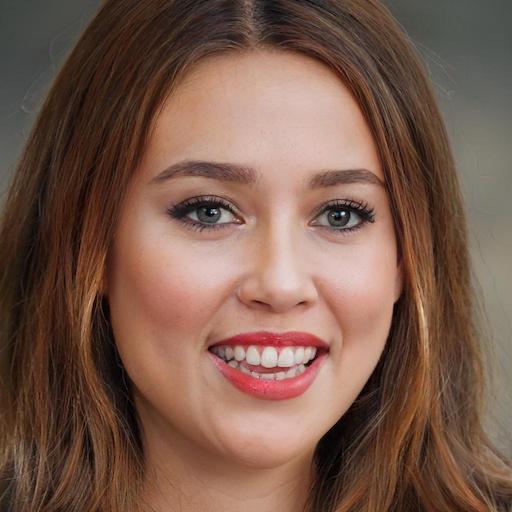} &
		\includegraphics[width=0.175\columnwidth]{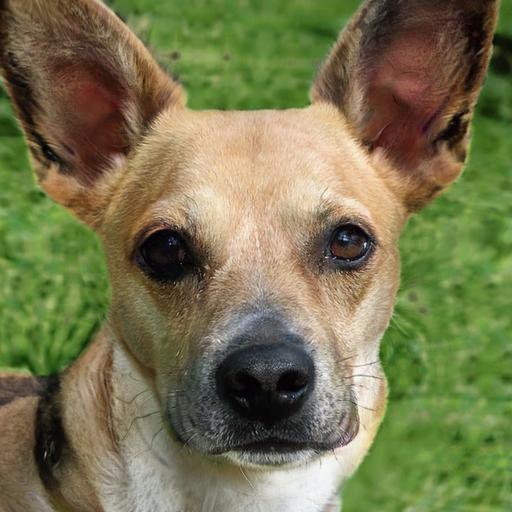} & 
		\includegraphics[width=0.175\columnwidth]{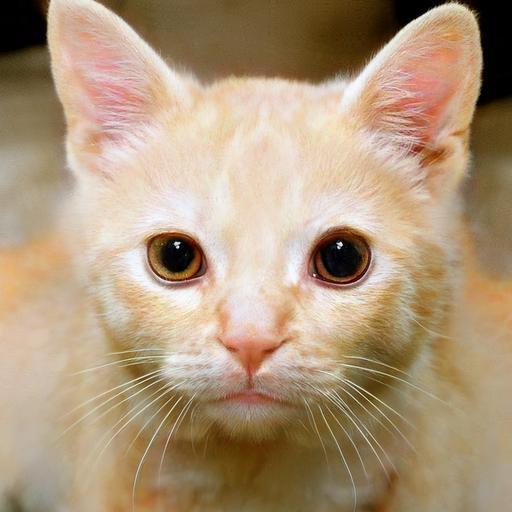} \\

    \end{tabular}
	
	\vspace{-0.4cm}
	\caption{
	Pseudo-aligning images generated by unaligned generators. 
	While these generators normally produce unaligned images (row 1), replacing $w_0$ with the average latent $\overline{w}$ yields roughly aligned images (row 2). 
    }
    \vspace{-0.375cm}
	\label{fig:pseudo-alignment2}
\end{figure}

%% file: analysis.tex
\input{figures/w0_and_w1_experiment}

\section{Analysis}\label{sec:analysis}

\subsection{Rotation Control}\label{sec:rotation_control}
As discussed above, the latent code fed into the first layer, $w_0$, controls the translation and rotation of the image content. 
However, as illustrated in \cref{fig:pseudo-alignment2}, $w_0$ affects each image in a slightly different manner. For example, the leftmost human face is slightly rotated while the face in the third column is perfectly upright. This suggests that rotation is also affected by other layers of the generator, where it is entangled with other visual attributes. To examine the extent of this phenomenon, we perform two experiments, illustrated in \cref{fig:fixing_w0_and_w1}. First, we examine a series of images $G((w^*, w_1, w^*, ..., w^*))$ that differ only in their randomly sampled $w_1$ latent entry (top row). It may be seen that altering $w_1$ affects the in-plane rotation of the face, but this change is entangled also with other attributes, such as face shape and eyes. %
In our second experiment, we generate a series of images $G((w_0, w_1, w^*, ..., w^*))$, where $w_0$ and $w_1$ are held fixed, while the remaining latent entries are set to a randomly sampled code $w^*$. It may be seen that with both $w_0$ and $w_1$ fixed, the generated images all share the same head pose. Thus, we conclude that the subsequent layers do not appear to induce any further translation or rotation, and those are determined primarily by $w_0$ and $w_1$. %

\input{tables/dci}

\subsection{Disentanglement Analysis}~\label{sec:analysis-disentanglement}
To analyze the disentanglement of the different latent spaces of StyleGAN3, we follow Wu~\etal~\cite{wu2021stylespace} and compute the DCI (disentanglement / completeness / informativeness) metrics~\cite{eastwood2018framework} of each latent space. 
To compute the above metrics we employ pre-trained attribute regressors for various attributes, as described
by Wu~\etal~\cite{wu2021stylespace}.   

Observe that accurately computing attribute scores for unaligned images is challenging given that the attribute classifiers were trained solely on aligned images. To this end, we provide the classifiers with pseudo-aligned images, generated as described in \cref{section:pose_correction}. 

We report the DCI metrics for the $\mathcal{Z}$, $\mathcal{W}$, and $\mathcal{S}$ spaces in \cref{table:dci} for both the aligned and unaligned StyleGAN3 generators trained on the FFHQ~\cite{karras2019style} dataset.  $\mathcal{S}$ achieves the highest DCI scores across both StyleGAN3 generators, as it also does for StyleGAN2. Furthermore, while gaps in D and C between $\mathcal{Z}$ and $\mathcal{W}$ are smaller in StyleGAN3 than they are in StyleGAN2, the gap between $\mathcal{W}$ and $\mathcal{S}$ is larger, suggesting that using $\mathcal{S}$ for editing may be even more beneficial in StyleGAN3.

Since most StyleGAN inversion methods invert images into the $\mathcal{W}+$ latent space, it is also beneficial to examine the DCI metrics for this extended latent space. To do so, we randomly sample a set of latent codes $w\in\mathcal{W}$ and concatenate them to form latent codes in $\mathcal{W}+$. However, we find the resulting generated images are unnatural (see \cref{sec:analysis-disentanglement}). Moreover, applying the pre-trained DCI classifiers on such images results in inaccurate attribute scores, making the computed metrics unreliable. 

%% file: figures/w0_and_w1_experiment.tex
\begin{figure}[tb]
	\centering
	\setlength{\tabcolsep}{1pt}	
	{\small
	\begin{tabular}{c c c c c}
    	\vspace{-0.075cm}
	    \includegraphics[width=\columnwidth]{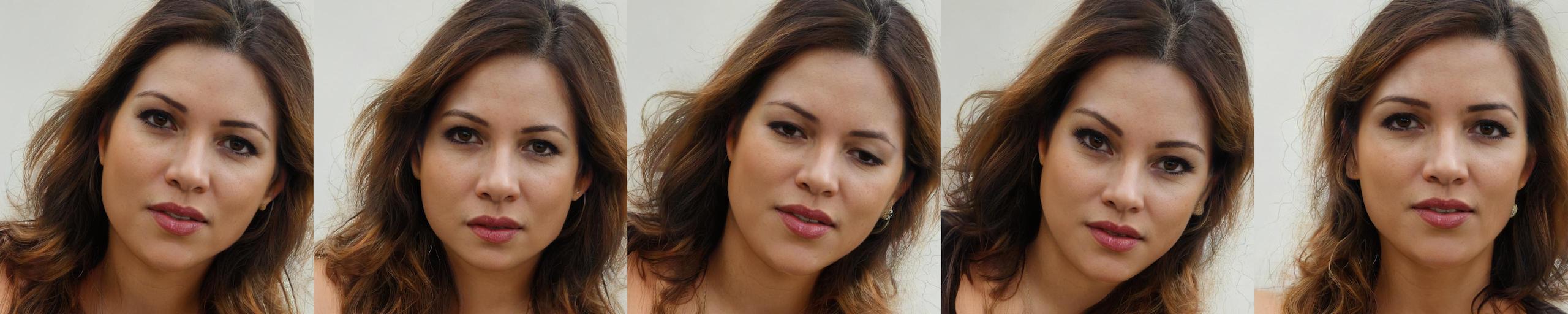} \\
	    \includegraphics[width=\columnwidth]{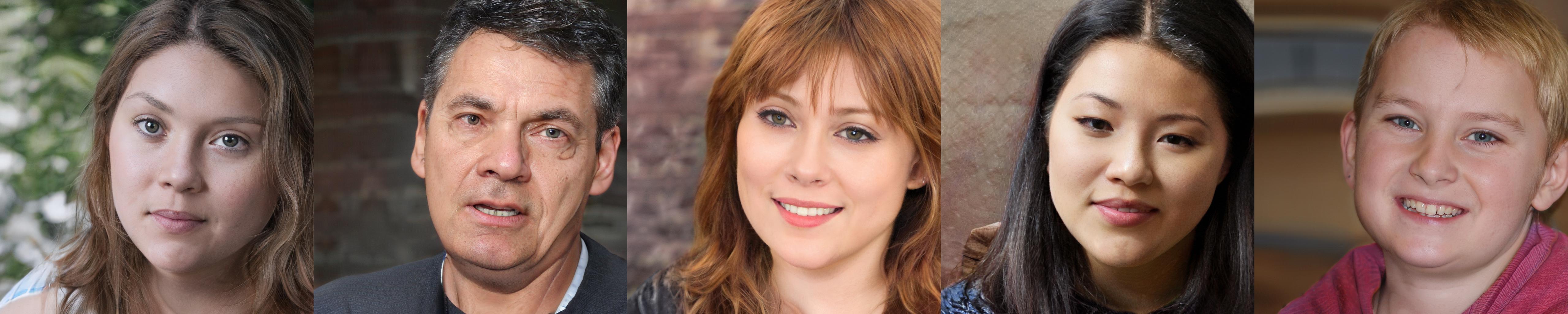} \\
    \end{tabular}
    }
	\vspace{-0.4cm}
	\caption{
	The roles and entanglement of $w_0$ and $w_1$. 
	Top row: altering only $w_1$ affects the in-plane rotation, as well as other visual aspects, implying they are entangled in $w_1$.
	Bottom row: holding $w_0$ and $w_1$ fixed while randomly sampling the remaining latent entries demonstrates that, for all practical purposes, the first two layers determine the translation and rotation.
	}
	\label{fig:fixing_w0_and_w1}
	\vspace{-0.4cm}
\end{figure}

%% file: tables/dci.tex
\begin{table}
    \begin{tabular}{l | c | c c c}
    \hline
    Generator & Space & Disent. & Compl. & Inform. \\ 
    \hline
    StyleGAN2 & $\mathcal{Z}$ & 0.31 & 0.21 & 0.72 \\
    StyleGAN2 & $\mathcal{W}$ & 0.54 & 0.57 & 0.97 \\
    StyleGAN2 & $\mathcal{S}$ & \textbf{0.75} & \textbf{0.87} & \textbf{0.99} \\ 
    \hline
    StyleGAN3 (A) & $\mathcal{Z}$ & 0.37 & 0.27 & 0.80 \\
    StyleGAN3 (A) & $\mathcal{W}$ & 0.47 & 0.43 & 0.94 \\
    StyleGAN3 (A) & $\mathcal{S}$ & \textbf{0.89}& \textbf{0.76} & \textbf{0.99} \\
    \hline
    StyleGAN3 (UA) & $\mathcal{Z}$ & 0.36 & 0.26 & 0.80 \\
    StyleGAN3 (UA) & $\mathcal{W}$ & 0.45 & 0.41 & 0.94 \\
    StyleGAN3 (UA) & $\mathcal{S}$ & \textbf{0.79} & \textbf{0.85} & \textbf{0.99} \\ 
    \bottomrule
    \end{tabular}
    \vspace{-0.25cm}
    \caption{DCI metrics for StyleGAN2 and StyleGAN3. For both StyleGAN architectures, and for aligned and unaligned datasets, the DCI scores (disentanglement / completeness / informativeness) improve consistently from the initial Gaussian noise $\mathcal{Z}$, through the intermediate space $\mathcal{W}$, and to the style parameters $\mathcal{S}$, which control channel-wise statistics ($\mathcal{S}>\mathcal{W}>\mathcal{Z}$).}
    \label{table:dci}
    \vspace{-0.15cm}
\end{table}

%% file: editing.tex
\setlength{\abovedisplayskip}{4pt}
\setlength{\belowdisplayskip}{4pt}

\section{Image Editing}\label{sec:editing}

In this section, we examine the effectiveness of various techniques for image editing with StyleGAN3, starting with the $\mathcal{W}$ and $\mathcal{W}+$ latent spaces, and proceeding to $\mathcal{S}$.

\paragraph{\textbf{Editing via Linear Latent Directions. }}
Here we use InterFaceGAN \cite{shen2020interpreting} for finding linear directions in $\mathcal{W}$ for aligned and unaligned StyleGAN3 generators.
Editing aligned images is simple and follows the approach used in StyleGAN2~\cite{karras2020analyzing}: given a randomly sampled latent code $w\in \mathcal{W}$, an editing direction $D$, and a step size $\delta$, the edited image is generated by $G_{\textit{aligned}}(w + \delta D; (0,0,0))$, where $G_{\textit{aligned}}$ is the aligned generator. 

As for unaligned images, there are two options. First, one may simply use an unaligned generator. Yet, one problem that arises in doing so is the fact that the attribute scores needed to learn these directions are obtained from classifiers pre-trained on aligned images. The scores produced by these classifiers on unaligned images may be inaccurate, resulting in poorly-learned directions in $\mathcal{W}$. To assist the pre-trained classifiers, we generate pseudo-aligned images by replacing $w_0$ with the generator's average latent code $\overline{w}$, as shown in \cref{section:pose_correction}.
The image generated by this modified latent code is then passed to the pre-trained classifier to obtain the original latent's attribute score. Yet, another problem with using the unaligned generator is that it requires learning a separate set of directions.

A second approach to mitigate the above overhead is to generate images using the aligned generator, but apply the user-defined transformations to control the rotation and placement of the generated object. Specifically, an edited unaligned image can be synthesized by $G_{\textit{aligned}}(w + \delta D; (r,t_x,t_y))$,
where $(r, t_x, t_y)$ is controlled by the user.
This gives the added benefit that the same latent directions may be used to edit both aligned and unaligned images.

\input{figures/interfacegan}

In \cref{fig:interfacegan-edits} we provide editing results obtained using the three approaches above. Notably, it is possible to achieve comparable edits on unaligned images via both the aligned and the unaligned generators. 
We also find that linear directions found in the latent space of $G_{\textit{unaligned}}$ to be generally more entangled than those found in the latent space of $G_{\textit{aligned}}$. This is most notable in the ``smile'' direction found in $G_{\textit{unaligned}}$, see row $2$. We attribute this entanglement to two factors: (1) the pseudo-aligned images may still be out-of-domain with respect to the classifier trained on aligned images, resulting in less accurate attribute scores; and (2) the linear editing directions make it more challenging to attain disentangled editing. 

Given the insight that unaligned images may be edited using a single aligned generator and the fact that the aligned generator produces higher quality images (as shown in StyleGAN3~\cite{aliasfreeKarras2021}), we focus our subsequent analysis on the aligned generator.

\input{figures/mapper_aligned}

\paragraph{\textbf{Editing via Non-Linear Latent Paths. }}
Various works have demonstrated that editing images via non-linear latent paths typically results in more faithful, disentangled edits~\cite{abdal2020styleflow,hou2020guidedstyle}. Following these works, we now explore learning non-linear latent editing paths within the $\mathcal{W}+$ latent space using the StyleCLIP mapper technique~\cite{patashnik2021styleclip}.  
As shown in ~\cref{fig:styleclip-edits}, the resulting edits are still entangled. For example, the image background typically changes across the different edits, even for local edits such as ``angry''. These results lead us to explore whether editing within the $\mathcal{S}$ space of StyleGAN3 achieves latent edits that are more disentangled than those achievable with $\mathcal{W}$ and $\mathcal{W}+$.

\input{figures/styleclip_global_joined}

\paragraph{\textbf{Editing via Latent Directions in $\mathcal{S}$. }}
Recall that our DCI analysis (\cref{table:dci}) indicates that the $\mathcal{S}$ space is more disentangled and complete than the $\mathcal{W}$ latent spaces. Here, we examine whether this finding extends to the editing quality of these spaces, particularly in terms of editing disentanglement. To this end, we find global linear editing directions in $\mathcal{S}$ using StyleCLIP \cite{patashnik2021styleclip}.  

\cref{fig:styleclip-global-directions} demonstrates that, in the domain of human faces, editing in $\mathcal{S}$ results in disentangled edits for both aligned and unaligned StyleGAN3 images.
Particularly, notice how the image backgrounds are much better preserved compared to the $\mathcal{W}$-based editing. Further, observe that the face identity is well-preserved for unrelated edits and that local edits, such as those changing hairstyle and expression, do not alter unrelated image regions (e.g., expression is consistent across the ``gender'', ``hi-top fade'', and ``tanned'' edits).
Notably, this disentanglement holds for other domains such as animal faces (AFHQv2~\cite{choi2020stargan}) and landscapes (Landscapes HQ~\cite{ALIS}). When editing animals, fur color, pose, and backgrounds are well-preserved under the various edits. Additionally, altering the landscapes preserves key contents of the original image, such as the lake (top) or road (bottom).

%% file: figures/interfacegan.tex
\begin{figure}[tb]
	\centering
	\setlength{\tabcolsep}{1.5pt}	
	
	{\small
	\begin{tabular}{c c@{} c c@{} c c@{} c}
	
		\raisebox{0.05in}{\rotatebox{90}{$G_{\textit{aligned}}$}} &
		\includegraphics[width=0.155\columnwidth]{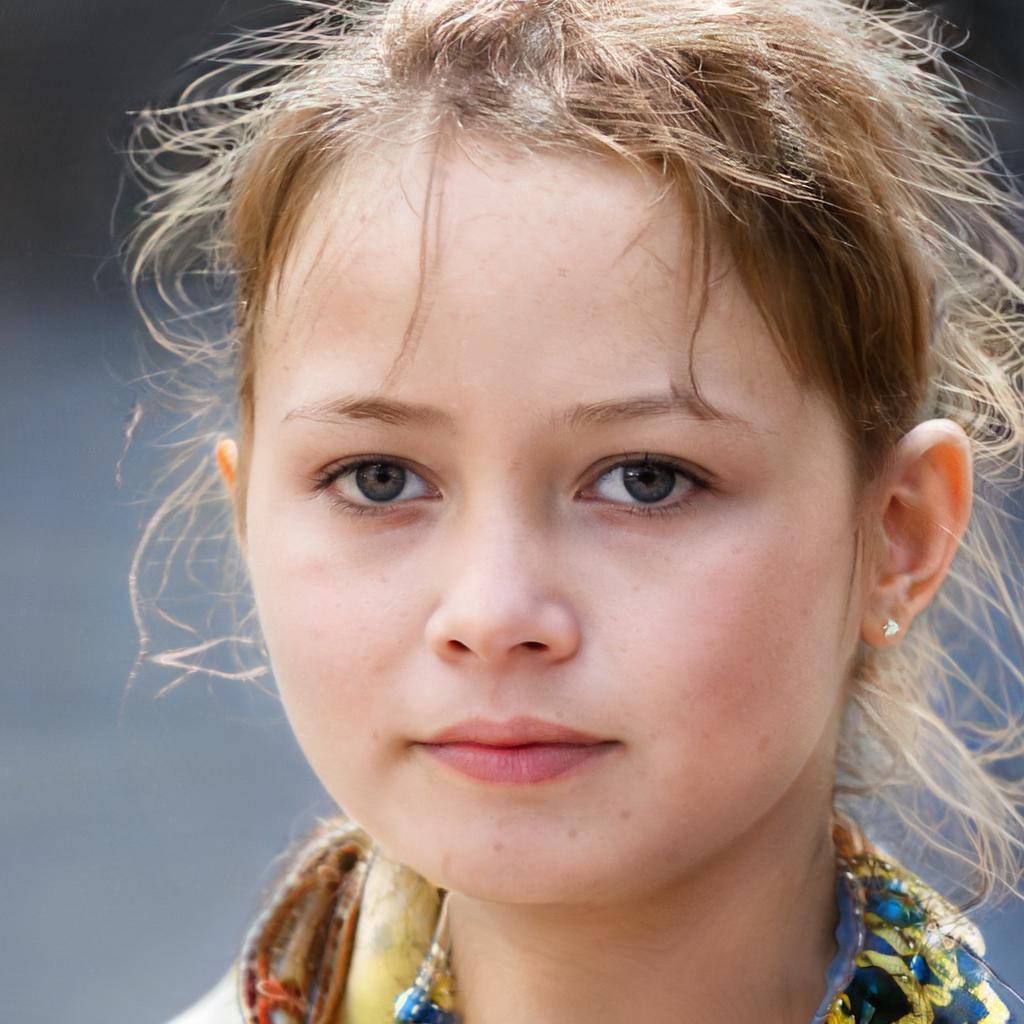} & 
		\includegraphics[width=0.155\columnwidth]{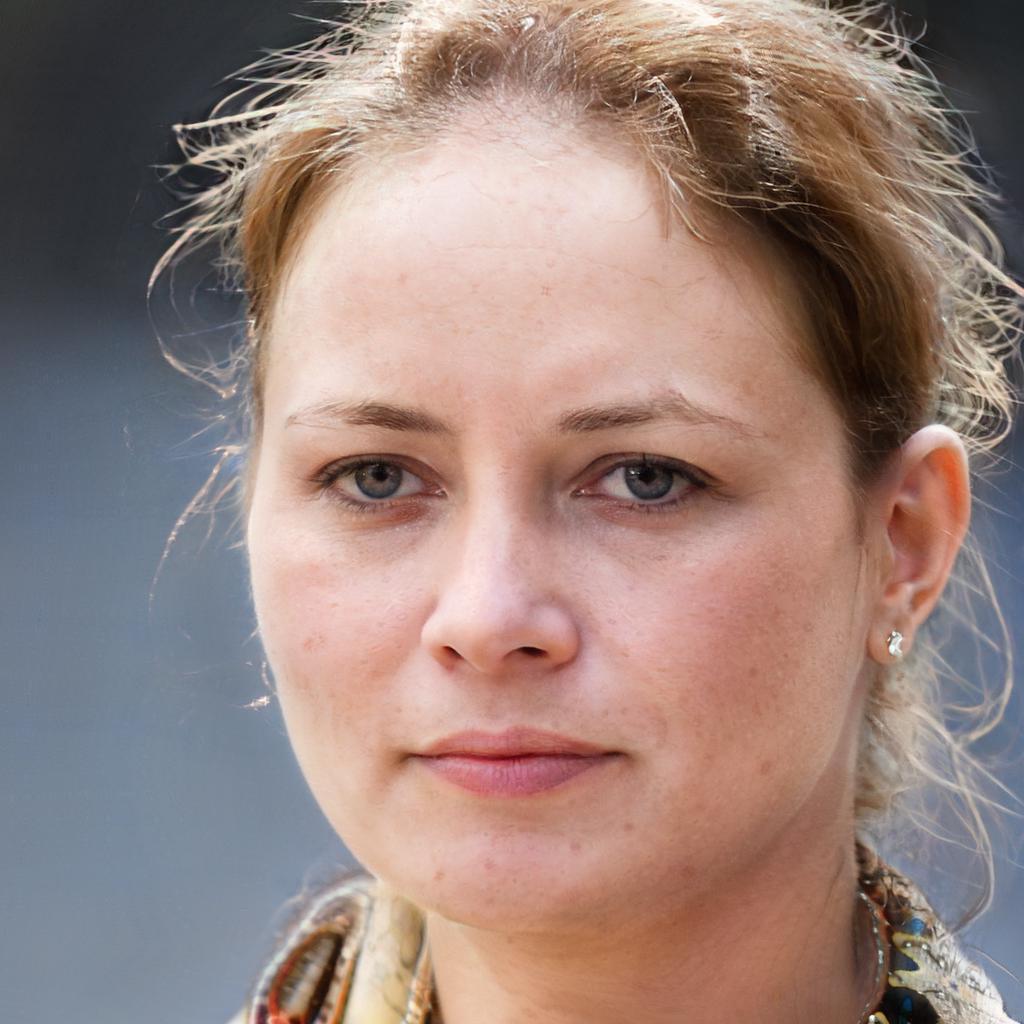} & 
		\includegraphics[width=0.155\columnwidth]{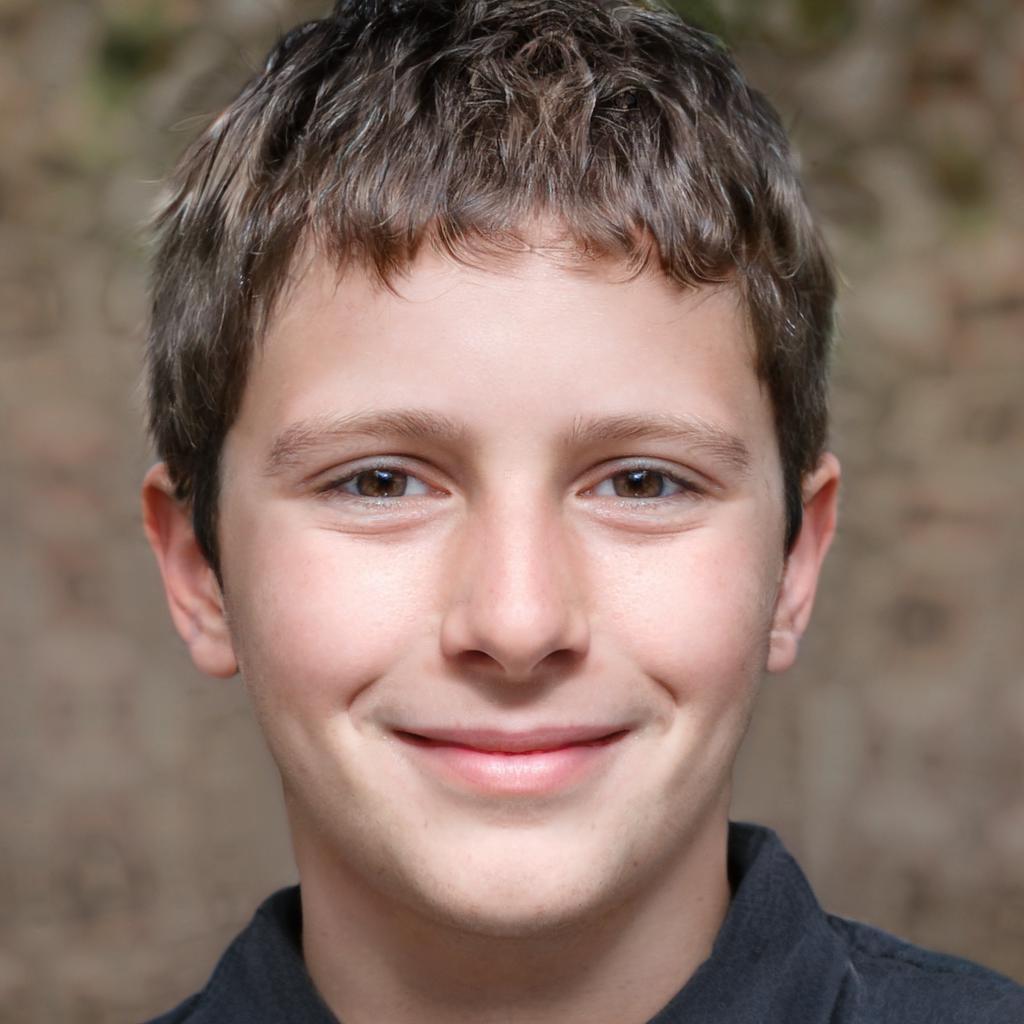} & 
		\includegraphics[width=0.155\columnwidth]{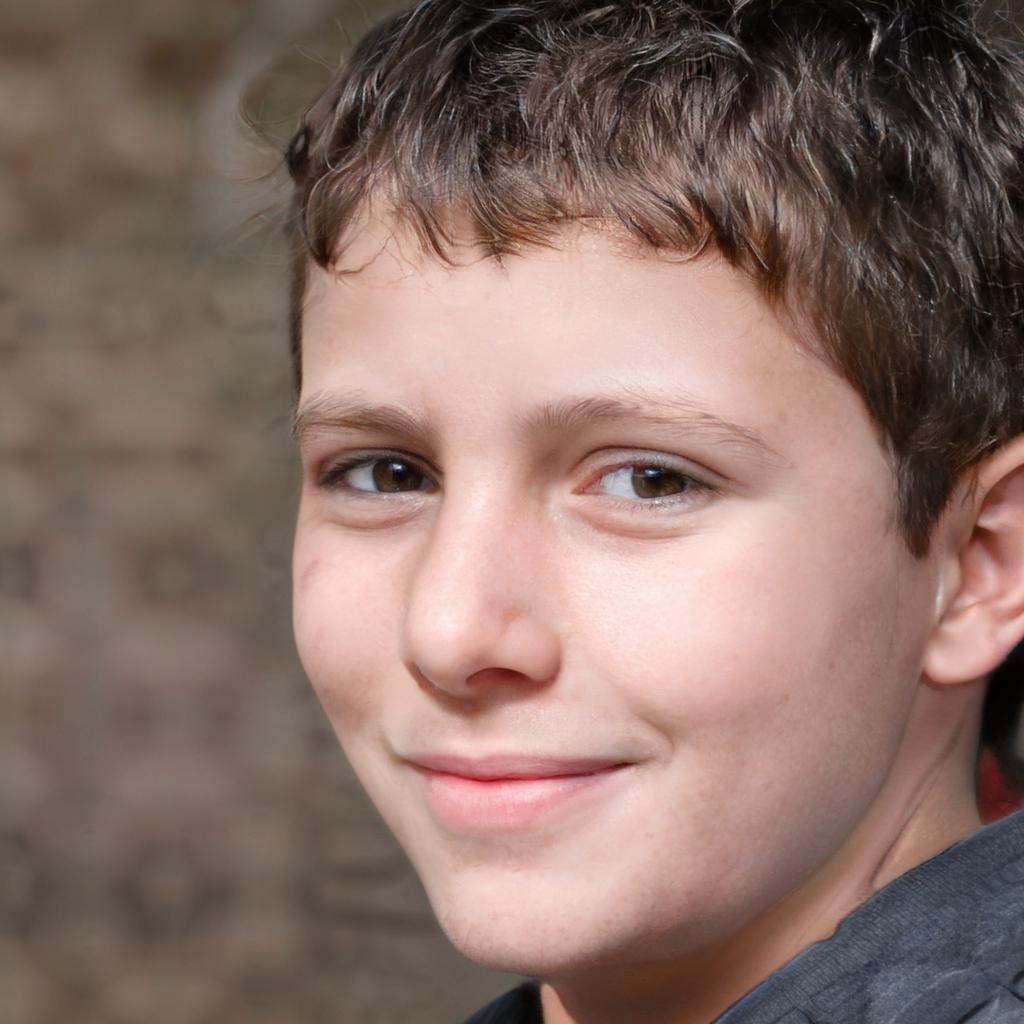} & 
		\includegraphics[width=0.155\columnwidth]{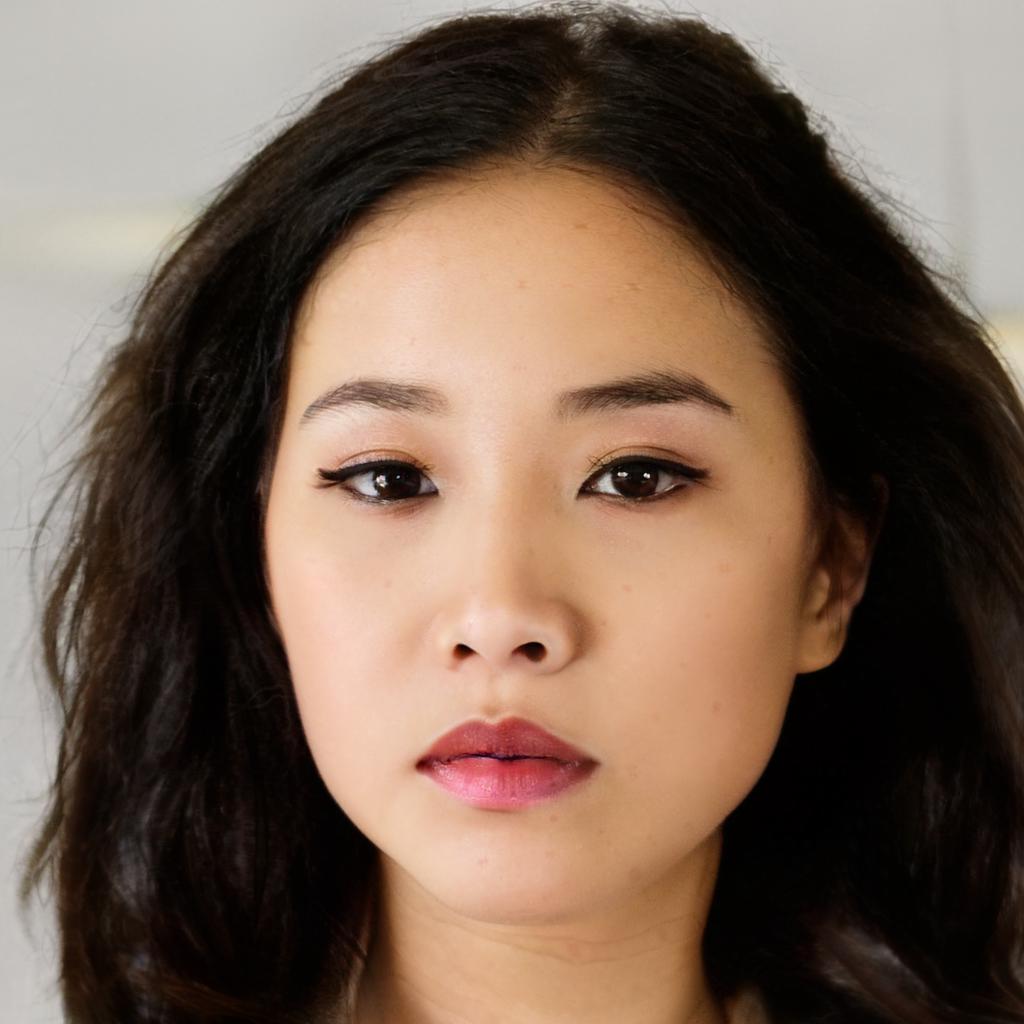} & 
		\includegraphics[width=0.155\columnwidth]{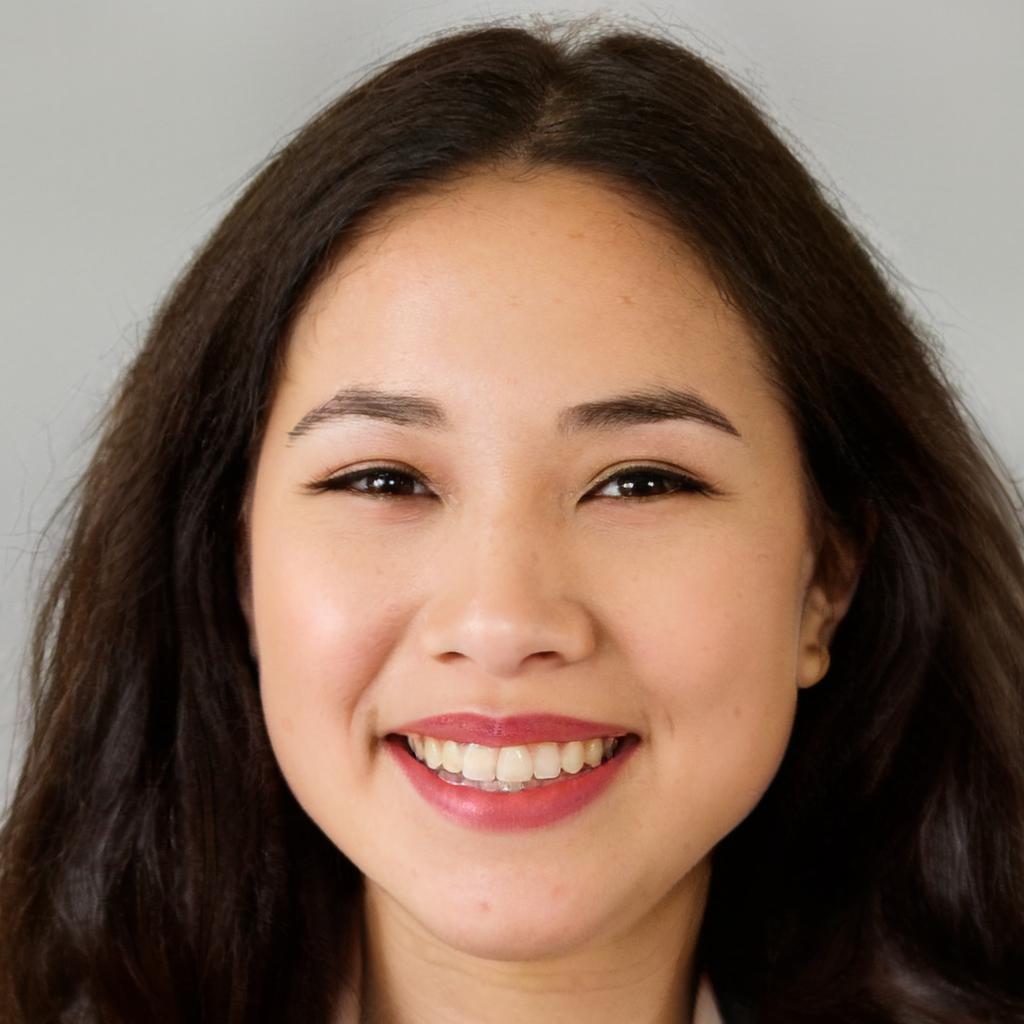} \\ 

		\raisebox{0.0in}{\rotatebox{90}{$G_{\textit{unaligned}}$}} &
		\includegraphics[width=0.155\columnwidth]{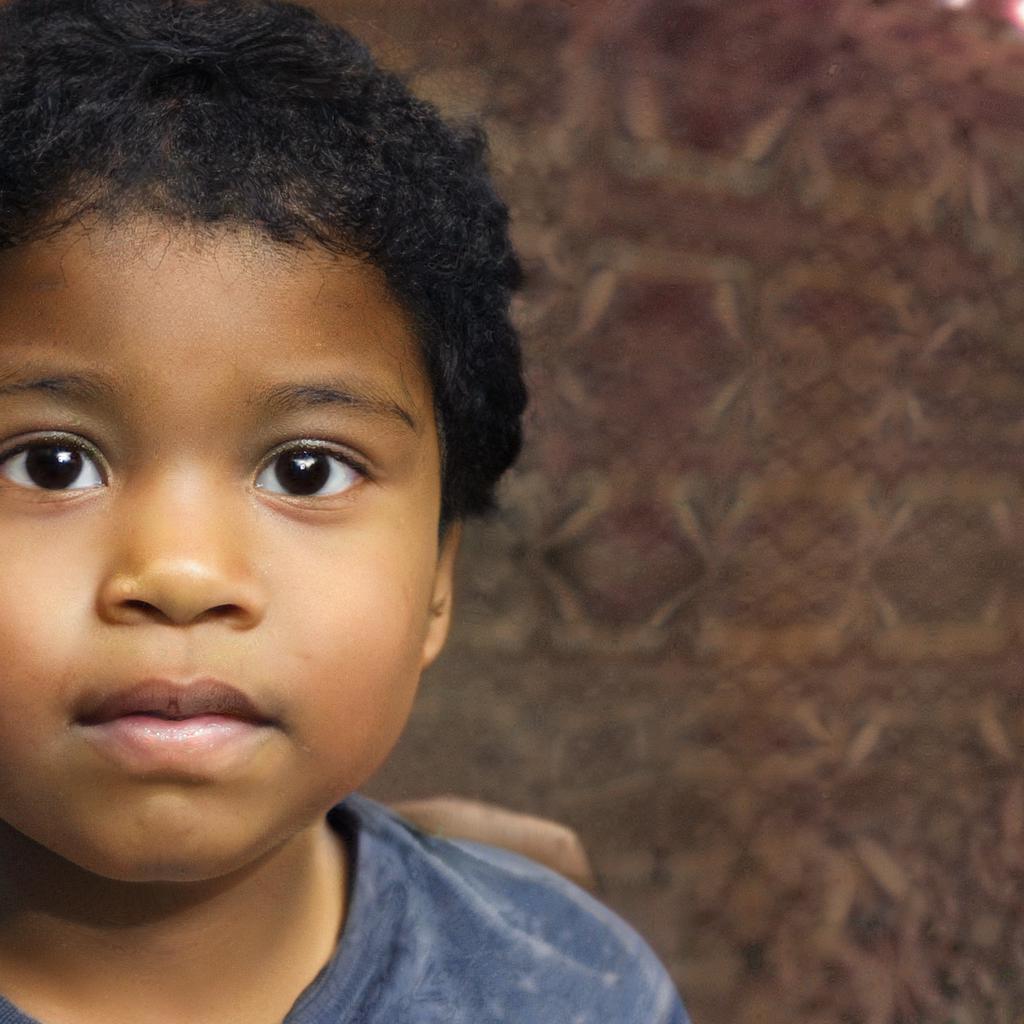} & 
		\includegraphics[width=0.155\columnwidth]{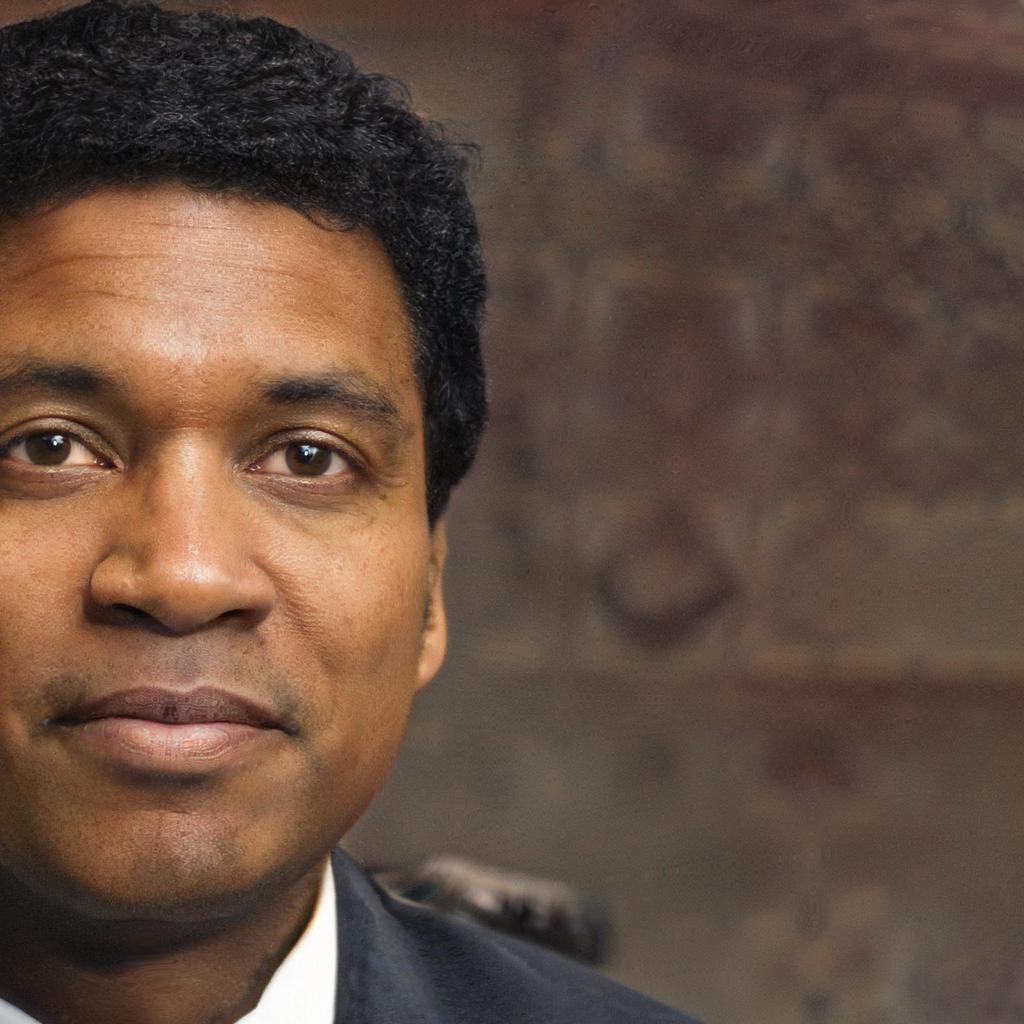} & 
		\includegraphics[width=0.155\columnwidth]{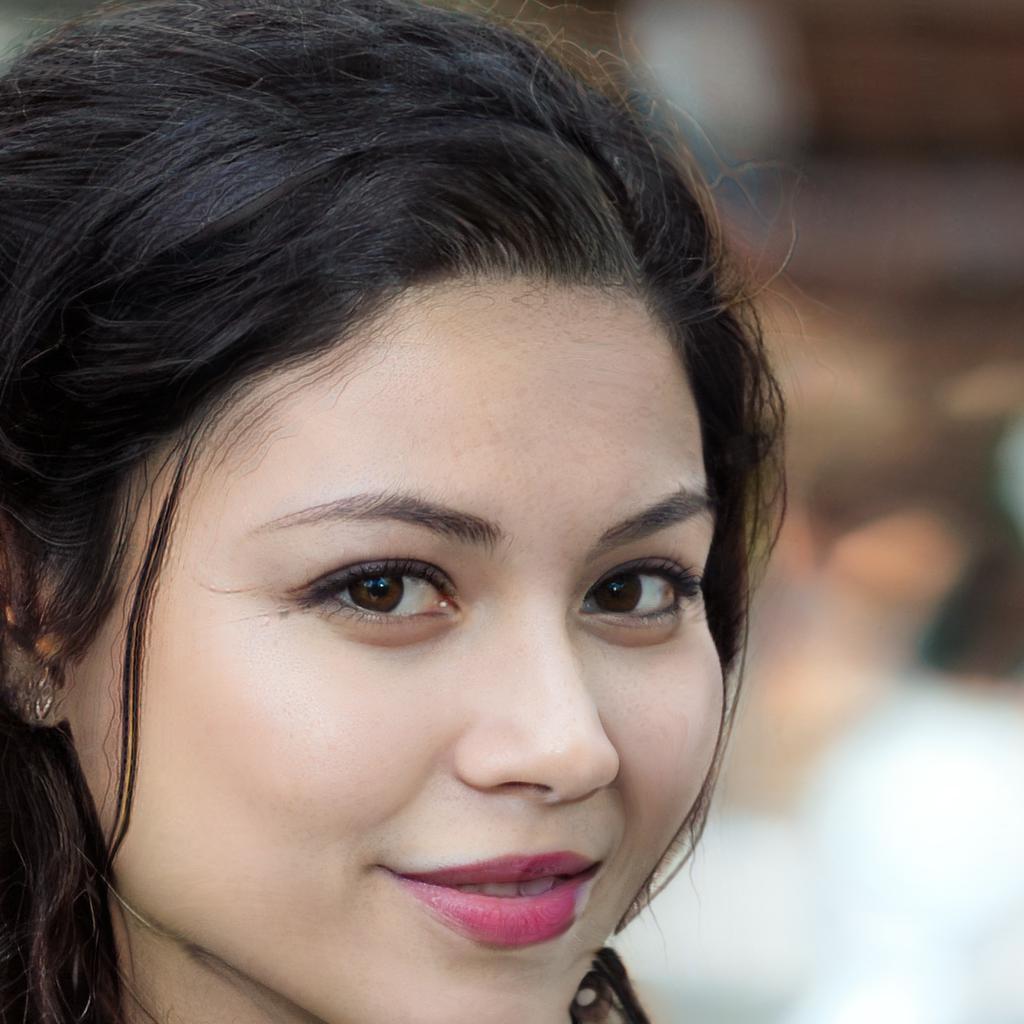} & 
		\includegraphics[width=0.155\columnwidth]{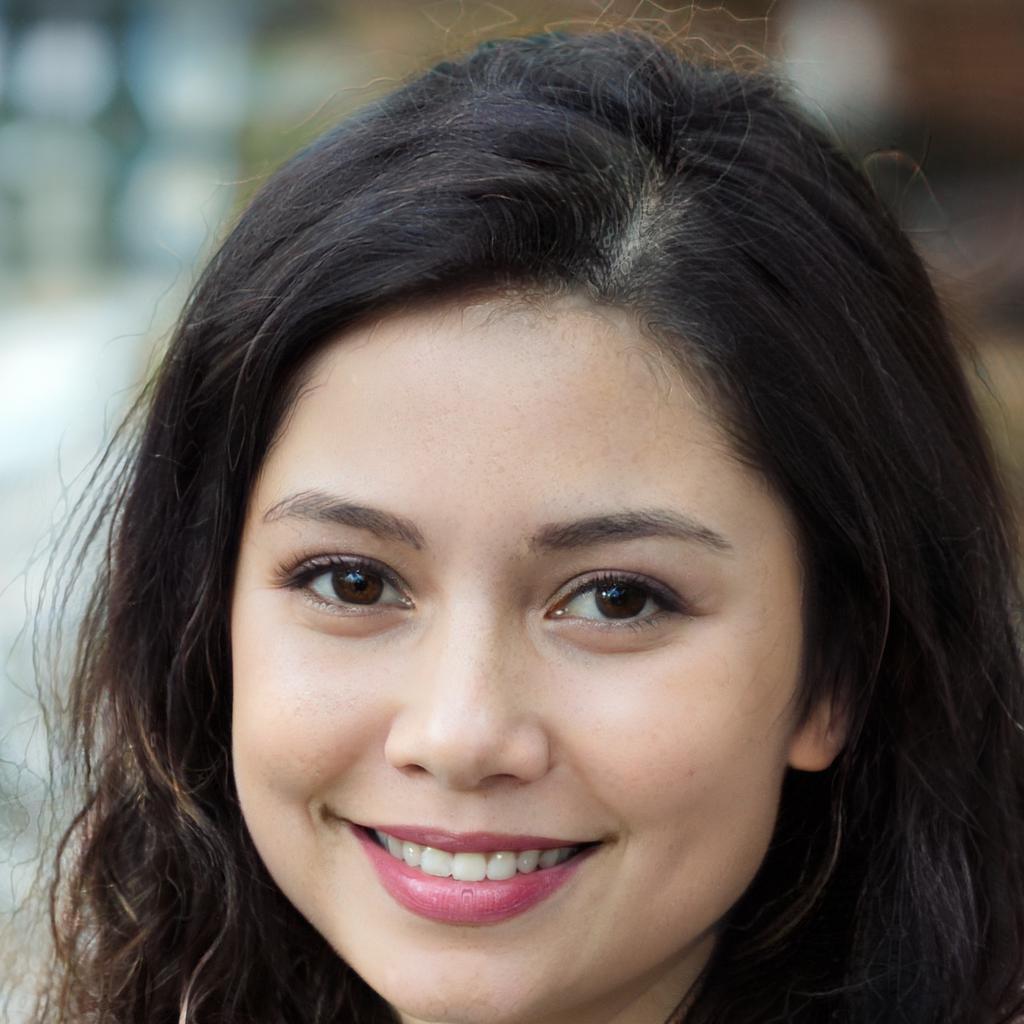} & 
		\includegraphics[width=0.155\columnwidth]{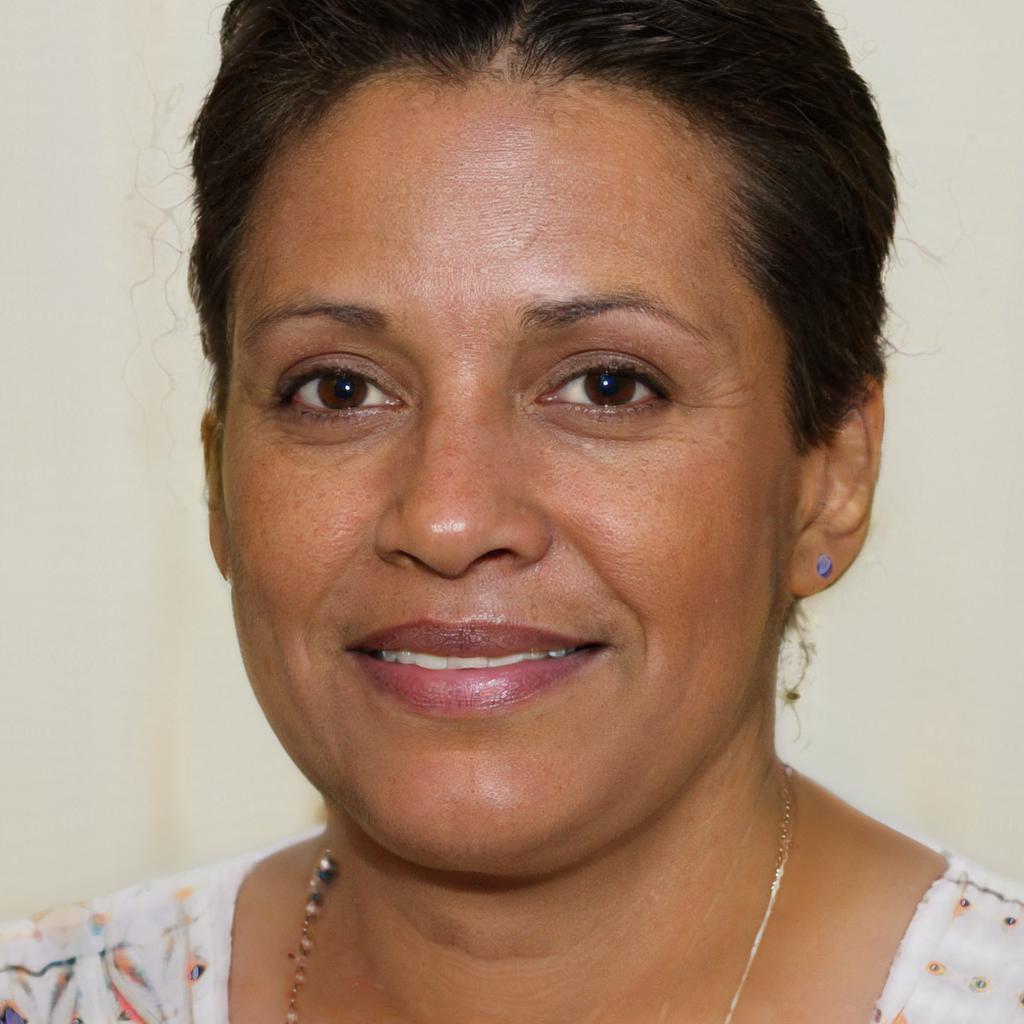} & 
		\includegraphics[width=0.155\columnwidth]{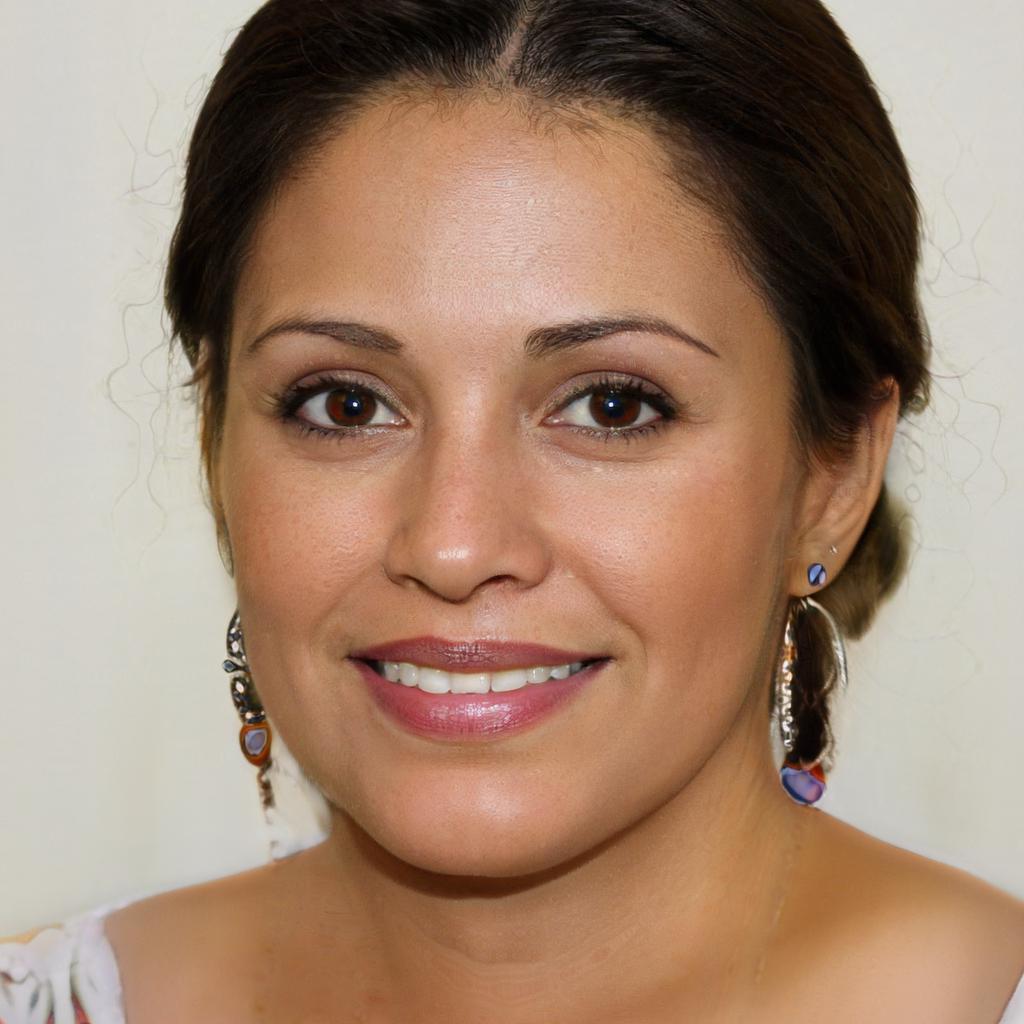} \\

		\raisebox{0.05in}{\rotatebox{90}{$G_{\textit{aligned+T}}$}} &
		\includegraphics[width=0.155\columnwidth]{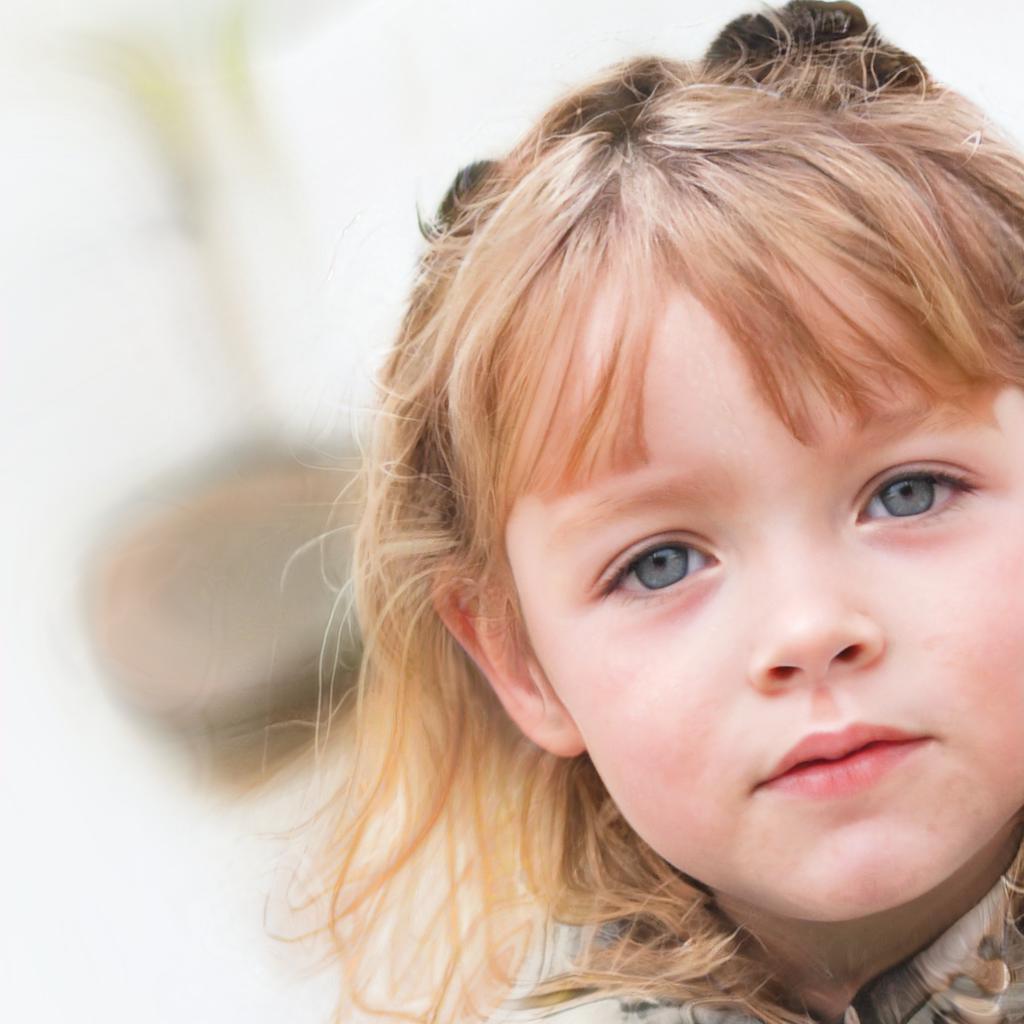} & 
		\includegraphics[width=0.155\columnwidth]{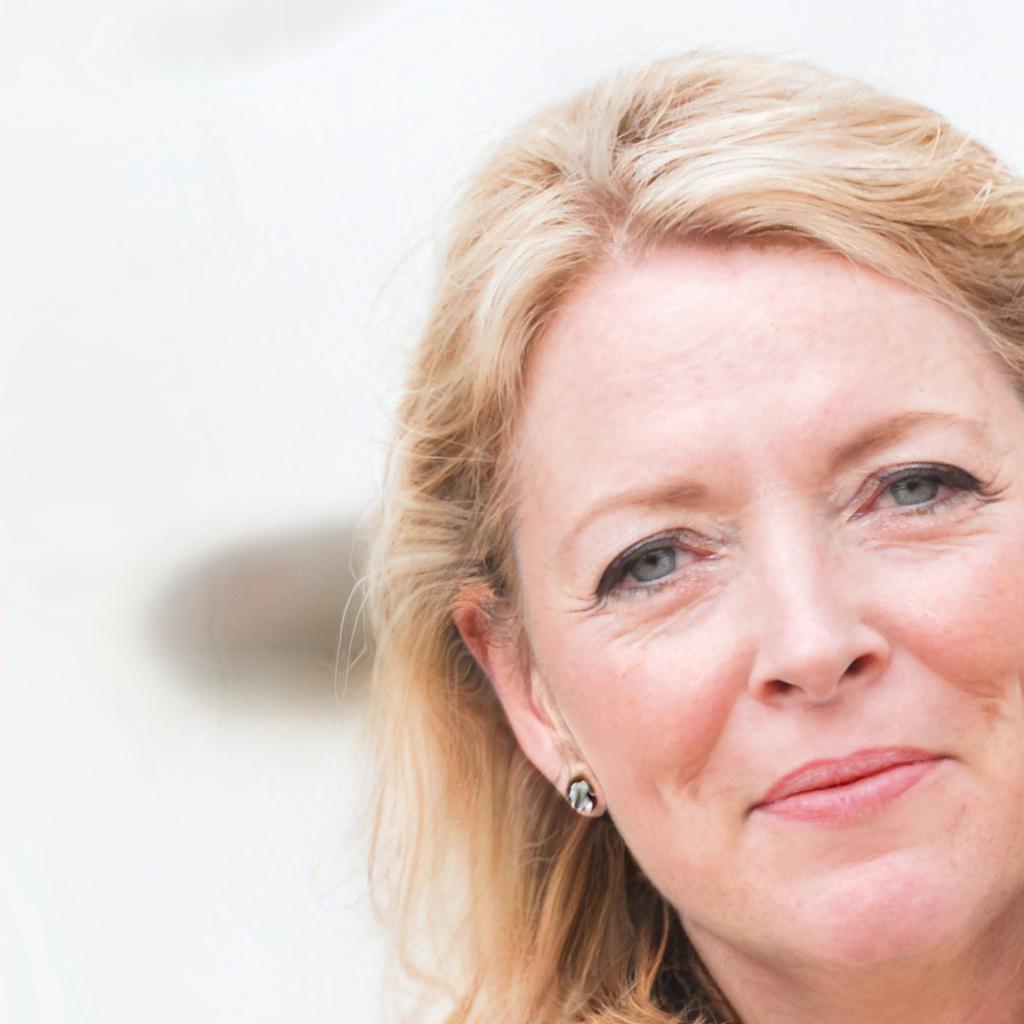} & 
		\includegraphics[width=0.155\columnwidth]{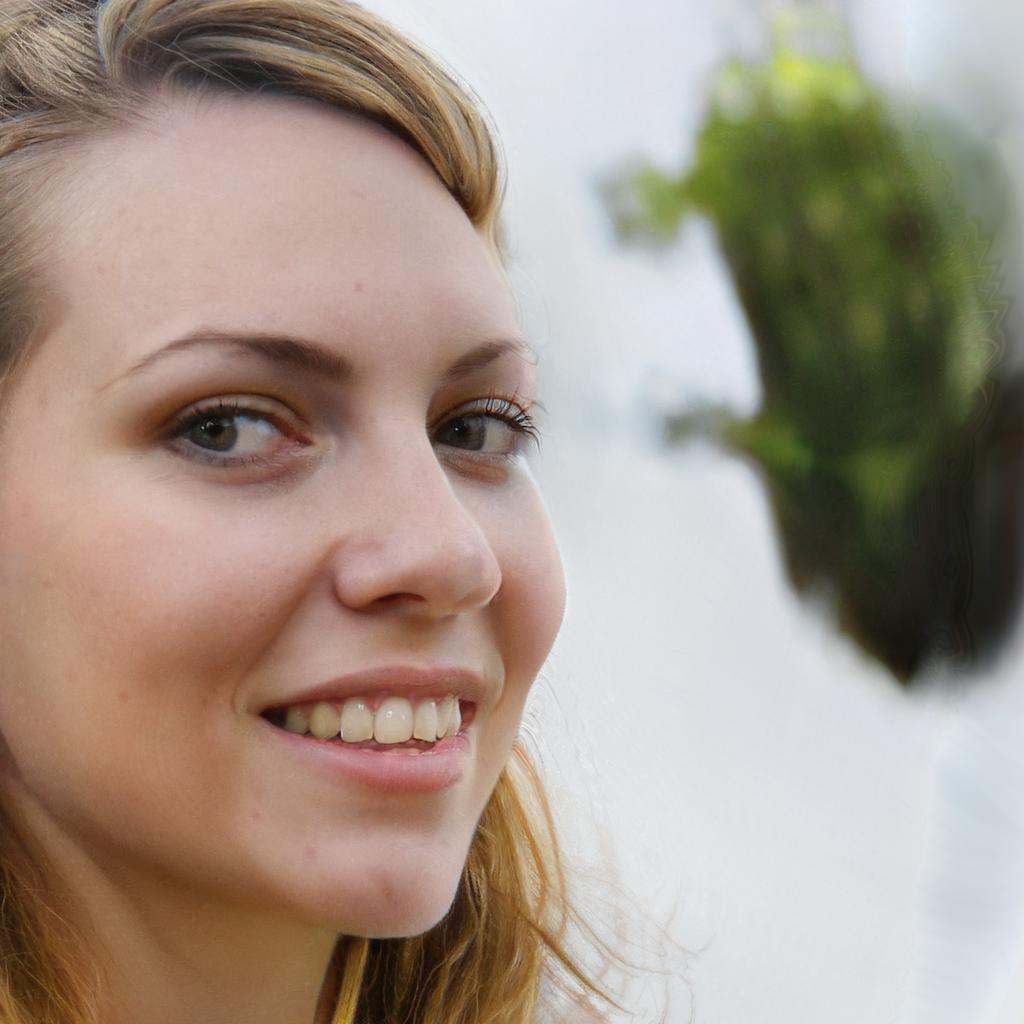} & 
		\includegraphics[width=0.155\columnwidth]{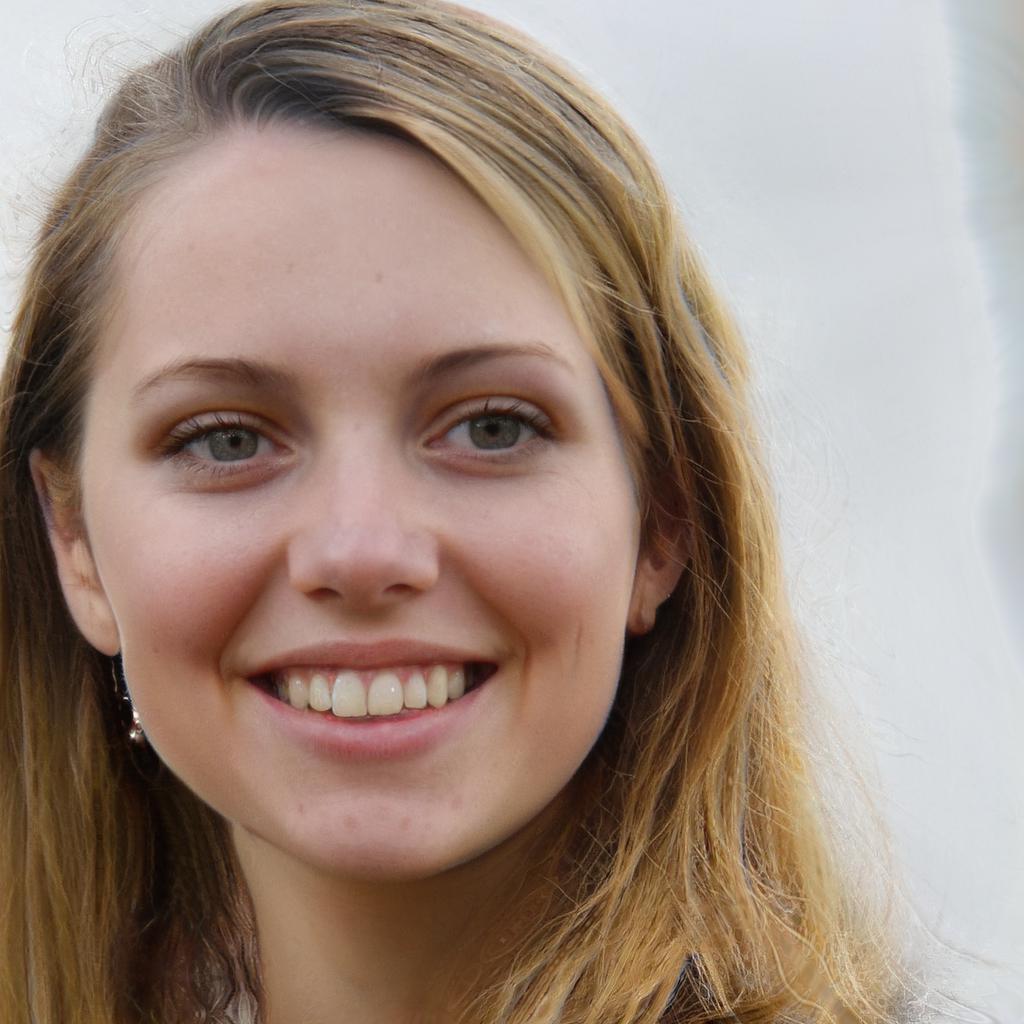} &
		\includegraphics[width=0.155\columnwidth]{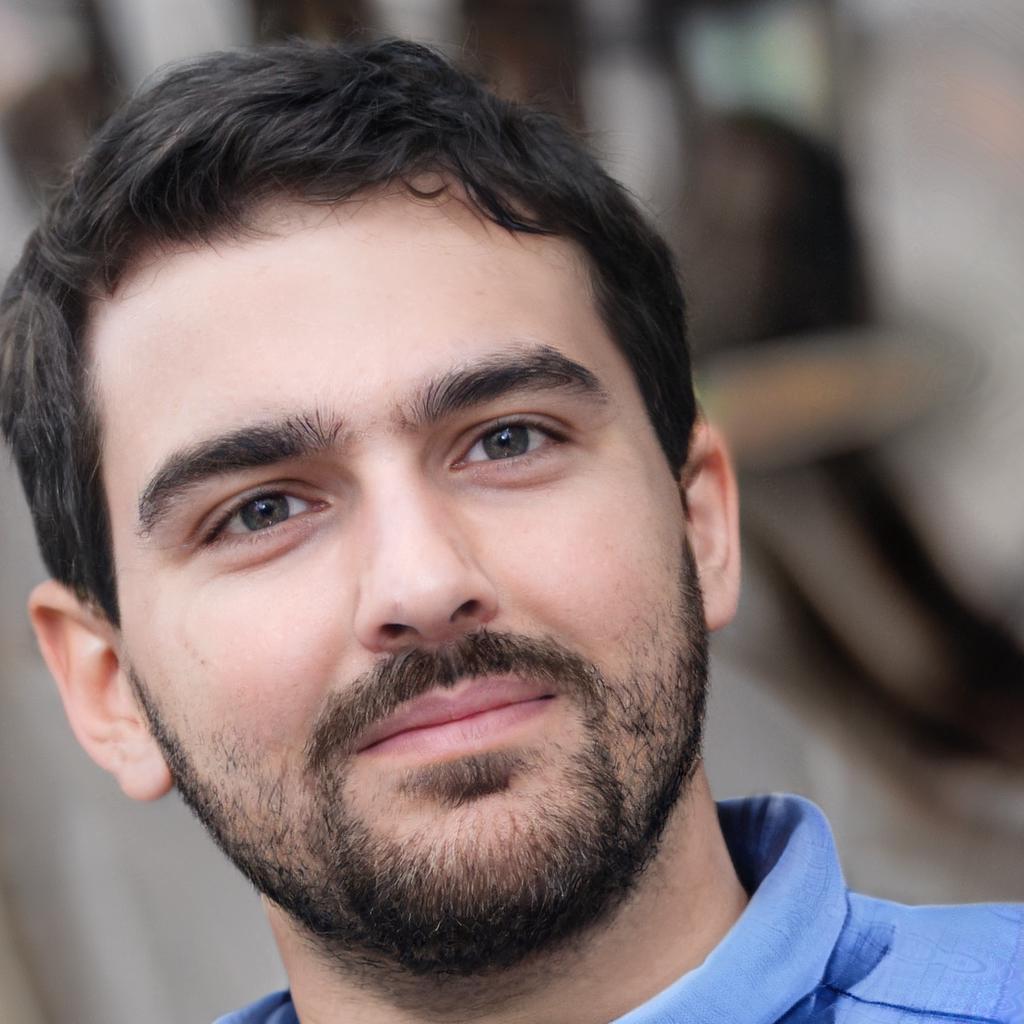} & 
		\includegraphics[width=0.155\columnwidth]{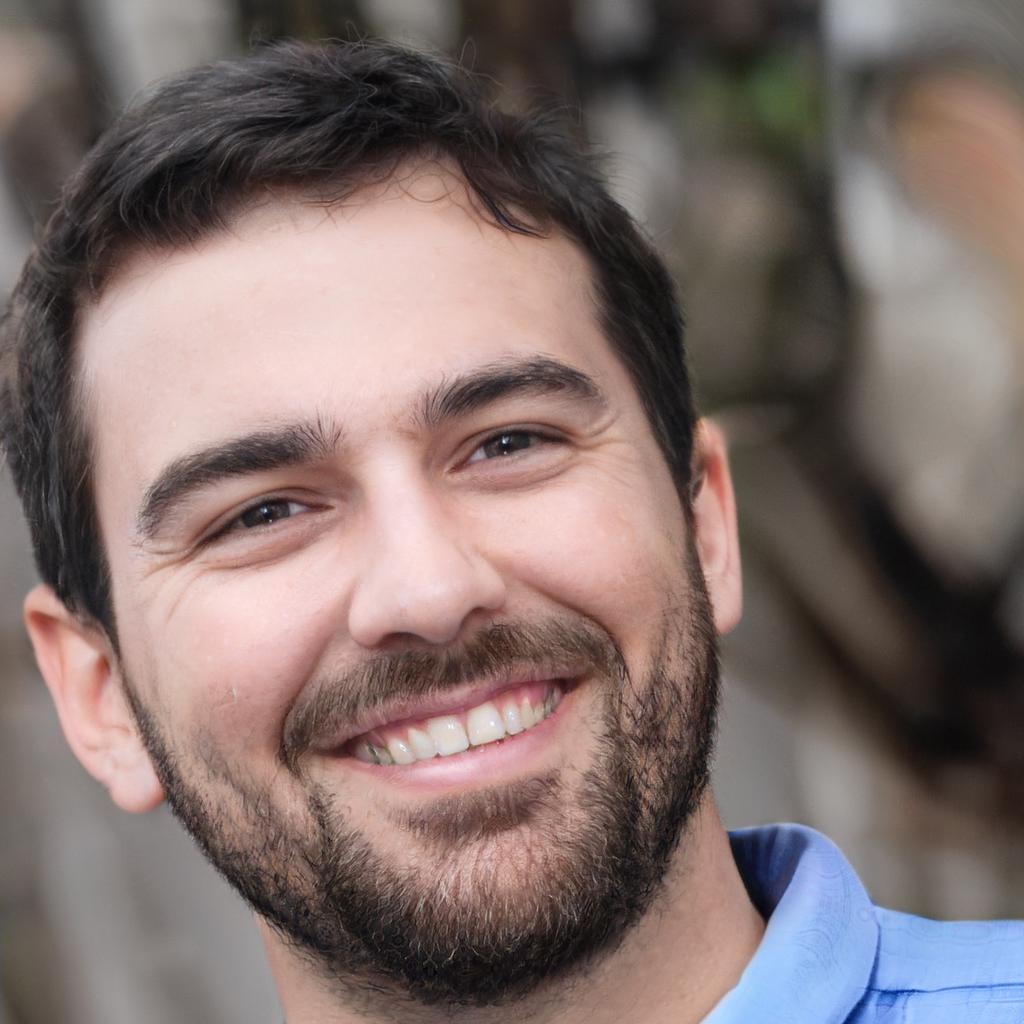} \\ 

        & \mc{Age} & \mc{Pose} & \mc{Smile}

	\end{tabular}
	}
	
	\vspace{-0.25cm}
	\caption{Linear editing in $\mathcal{W}$. Editing synthetic images using InterFaceGAN~\cite{shen2020interpreting} directions in $\mathcal{W}$. Editing unaligned images can be done either using an unaligned generator $G_{\textit{unaligned}}$, or using an aligned generator with an extra transformation $G_{\textit{aligned}+T}$.}
	\vspace{-0.2cm}
	\label{fig:interfacegan-edits}
\end{figure}

%% file: figures/mapper_aligned.tex
\begin{figure}[tb]
	\centering
	\setlength{\tabcolsep}{1pt}	
	{\small
	\begin{tabular}{c c c c c c}

		\includegraphics[width=0.155\columnwidth]{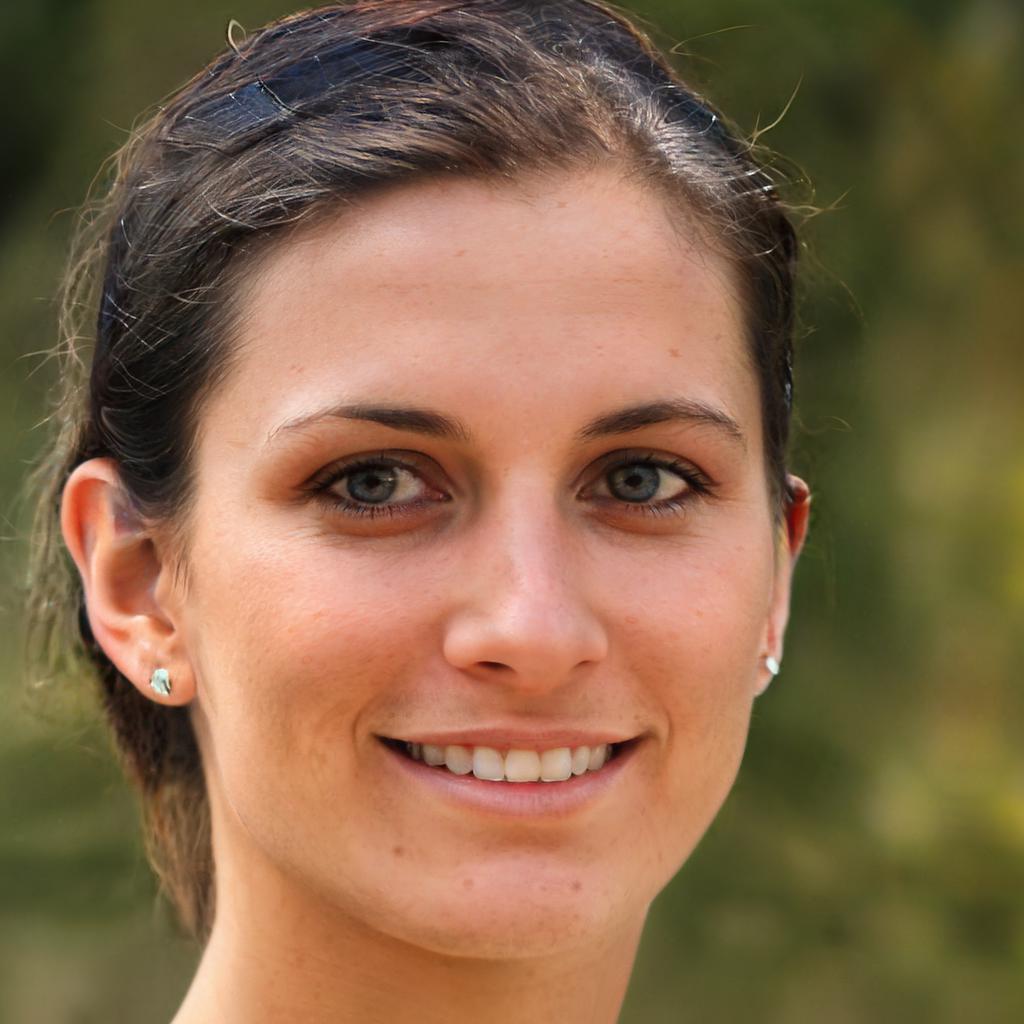} & 
		\includegraphics[width=0.155\columnwidth]{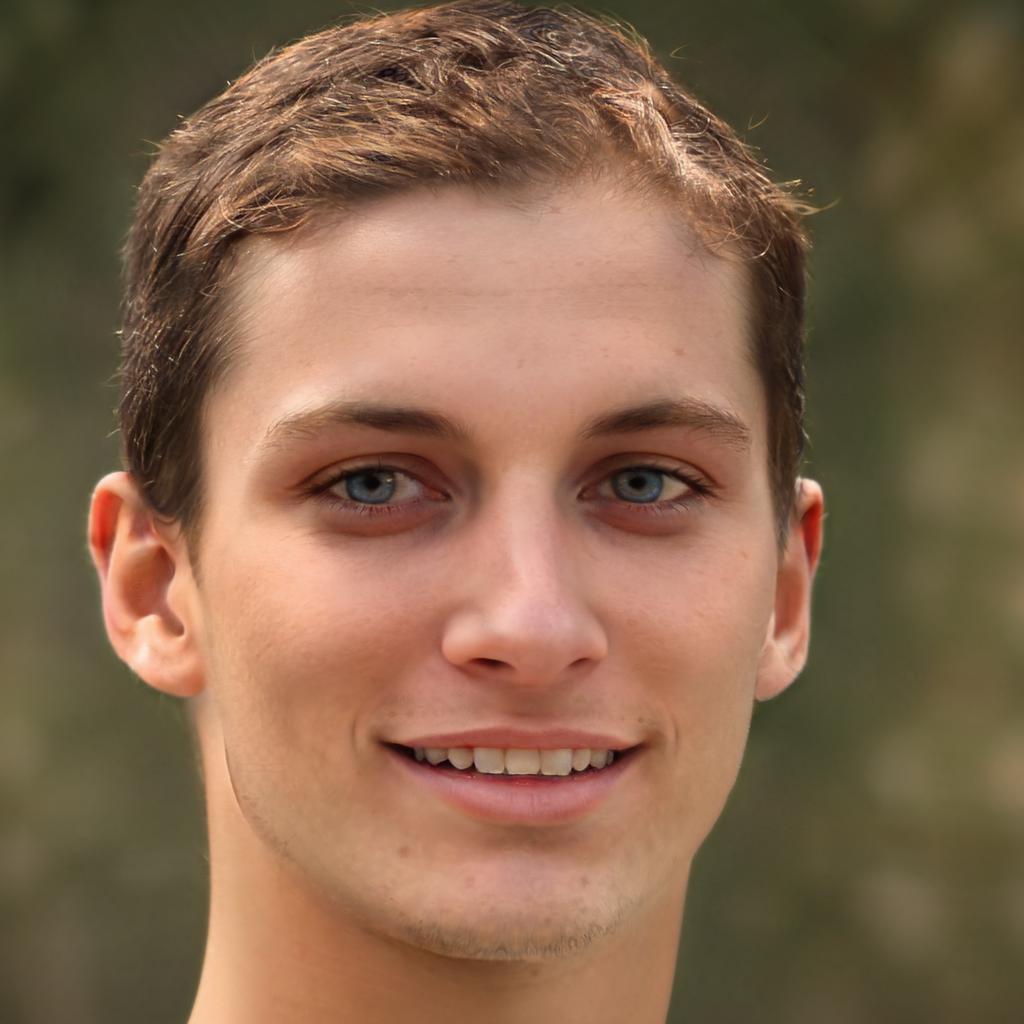} & 
		\includegraphics[width=0.155\columnwidth]{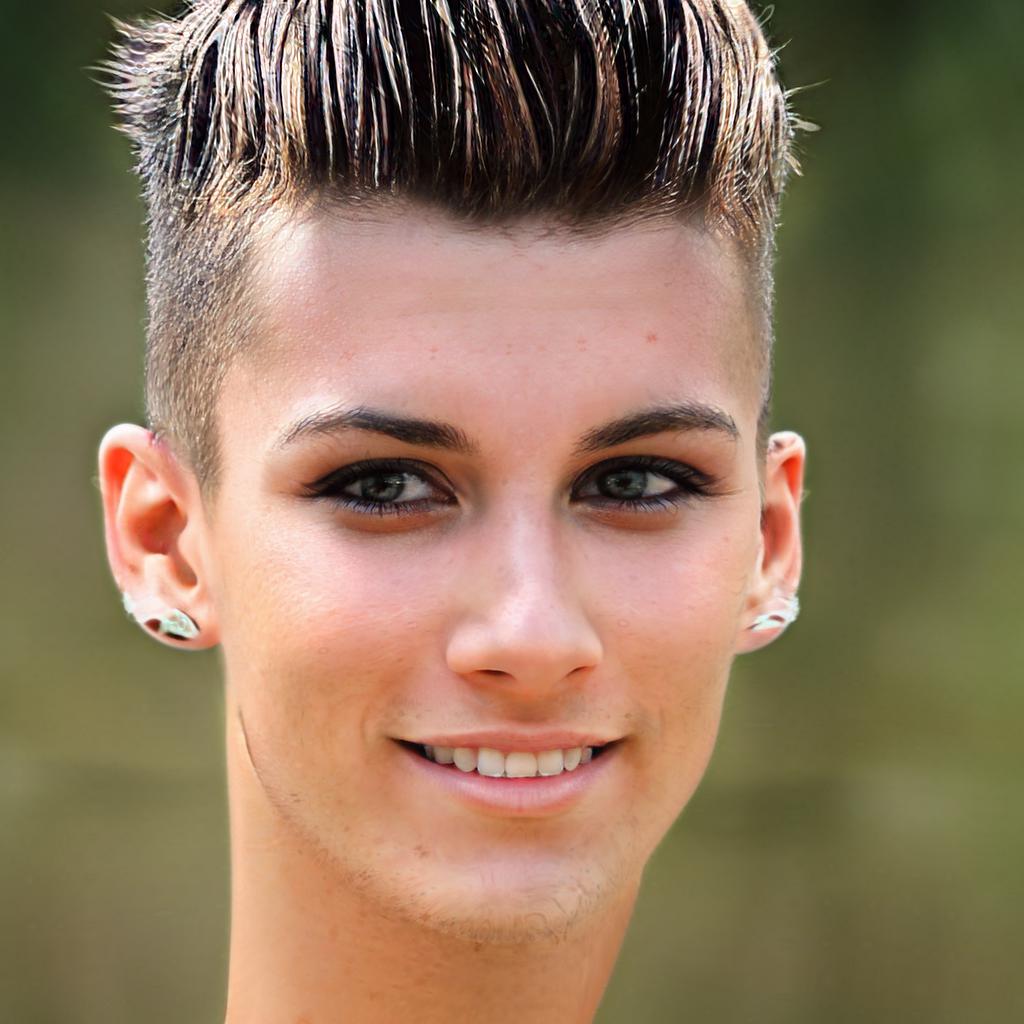} & 
		\includegraphics[width=0.155\columnwidth]{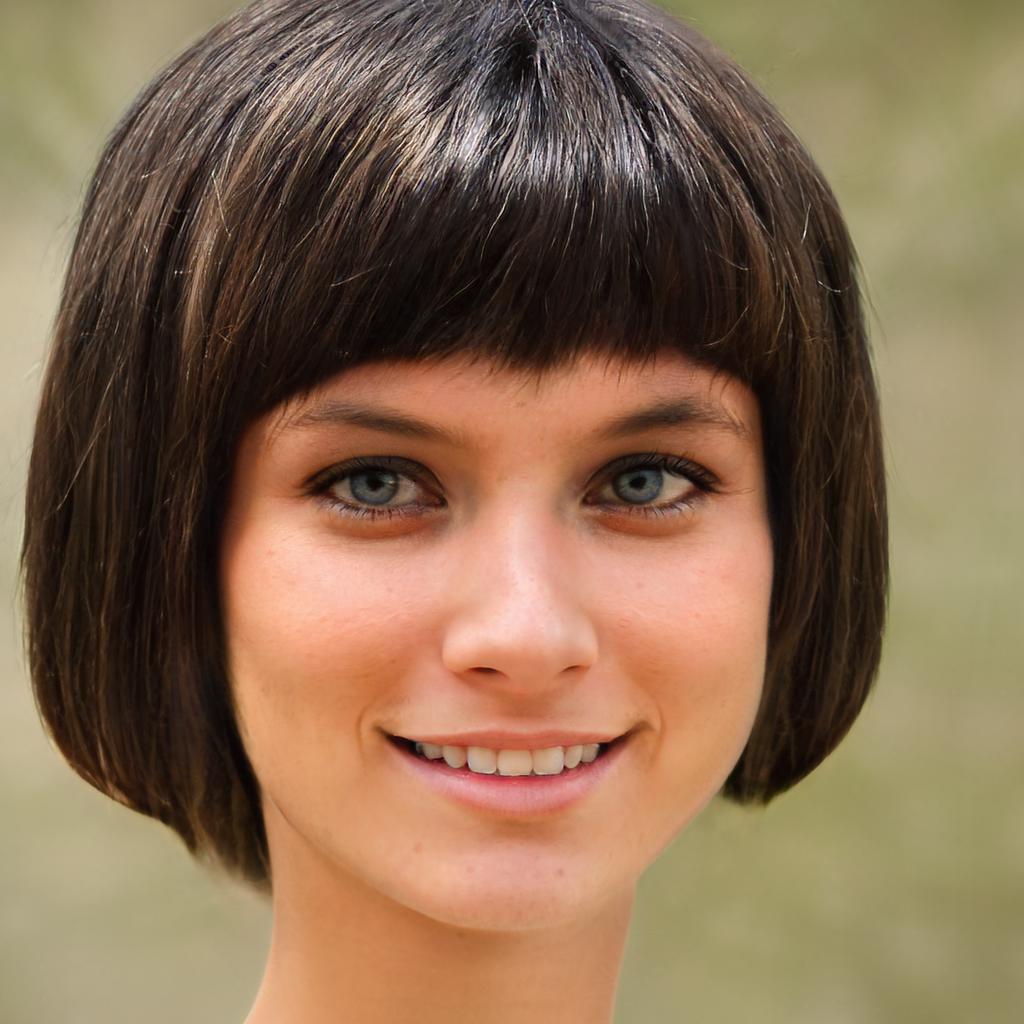} & 
		\includegraphics[width=0.155\columnwidth]{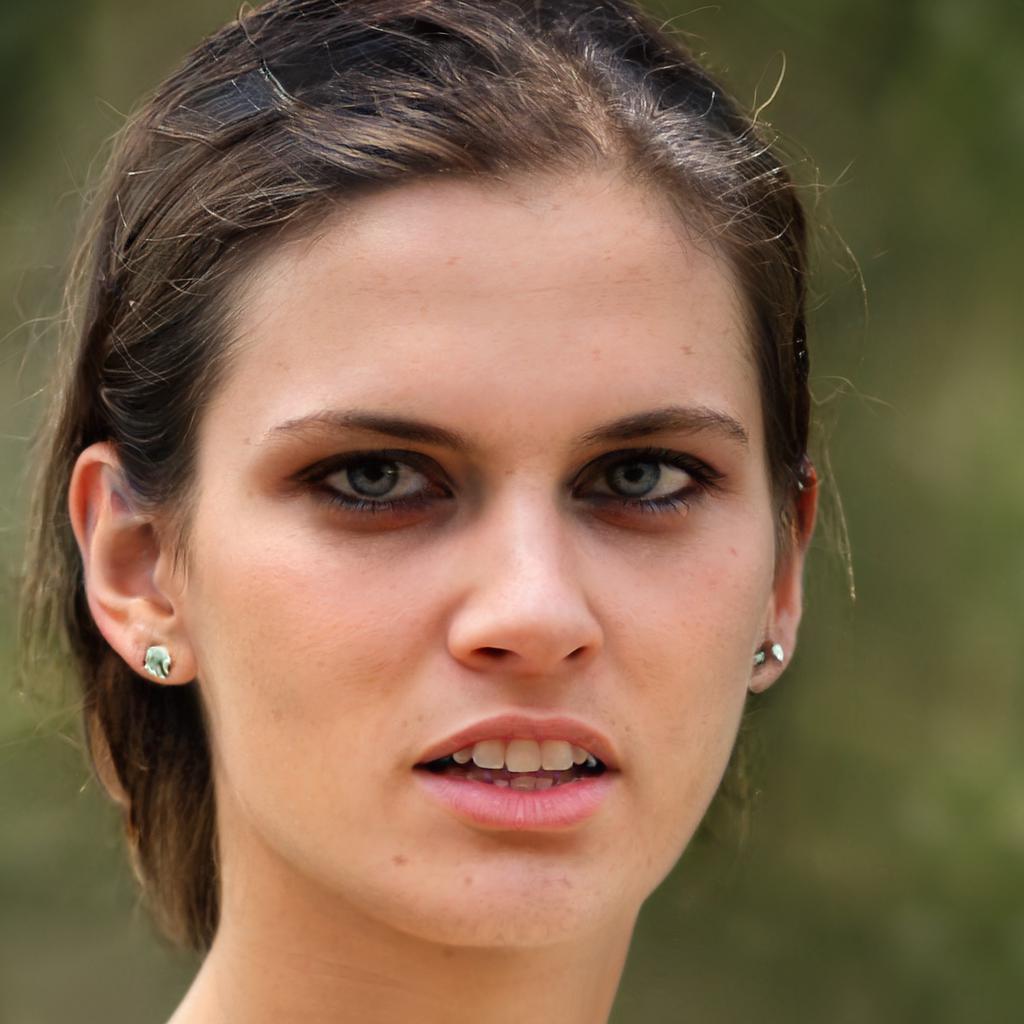} &
		\includegraphics[width=0.155\columnwidth]{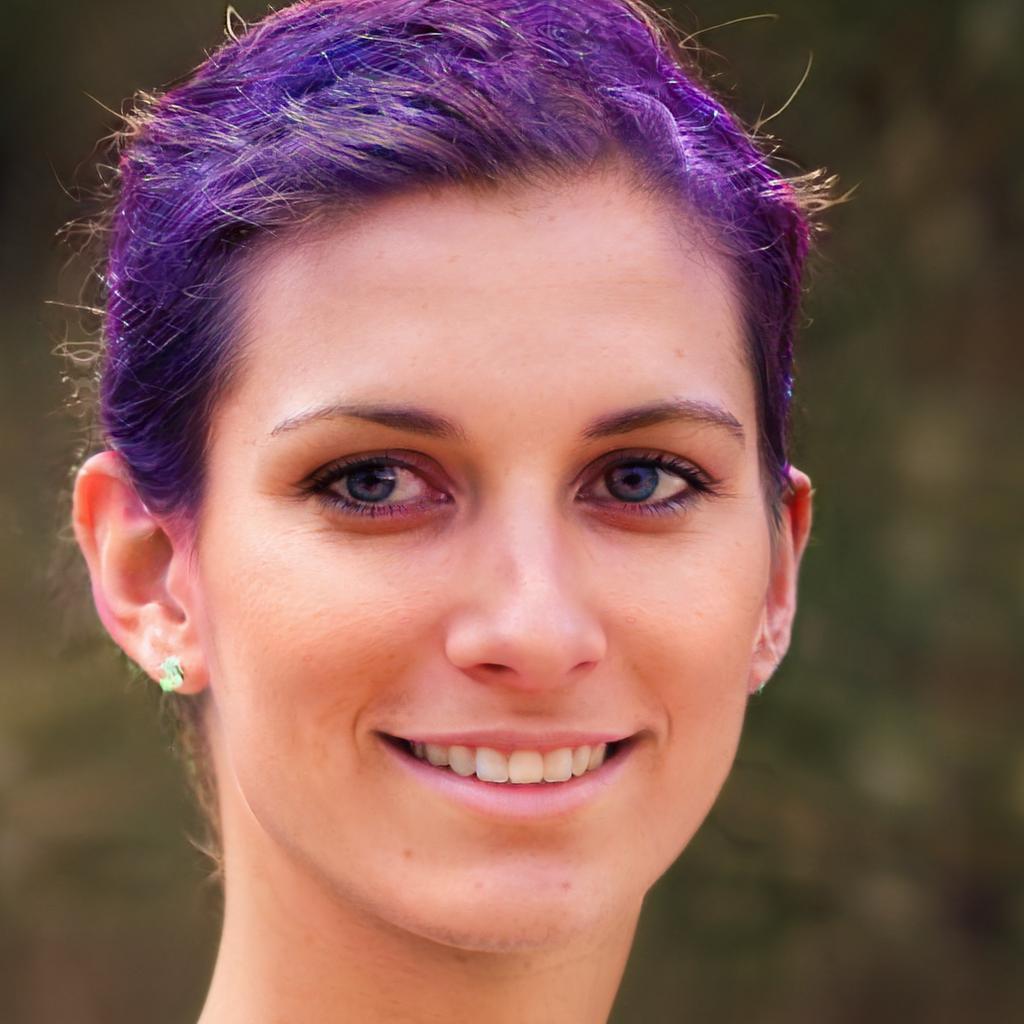}
		\\
		\includegraphics[width=0.155\columnwidth]{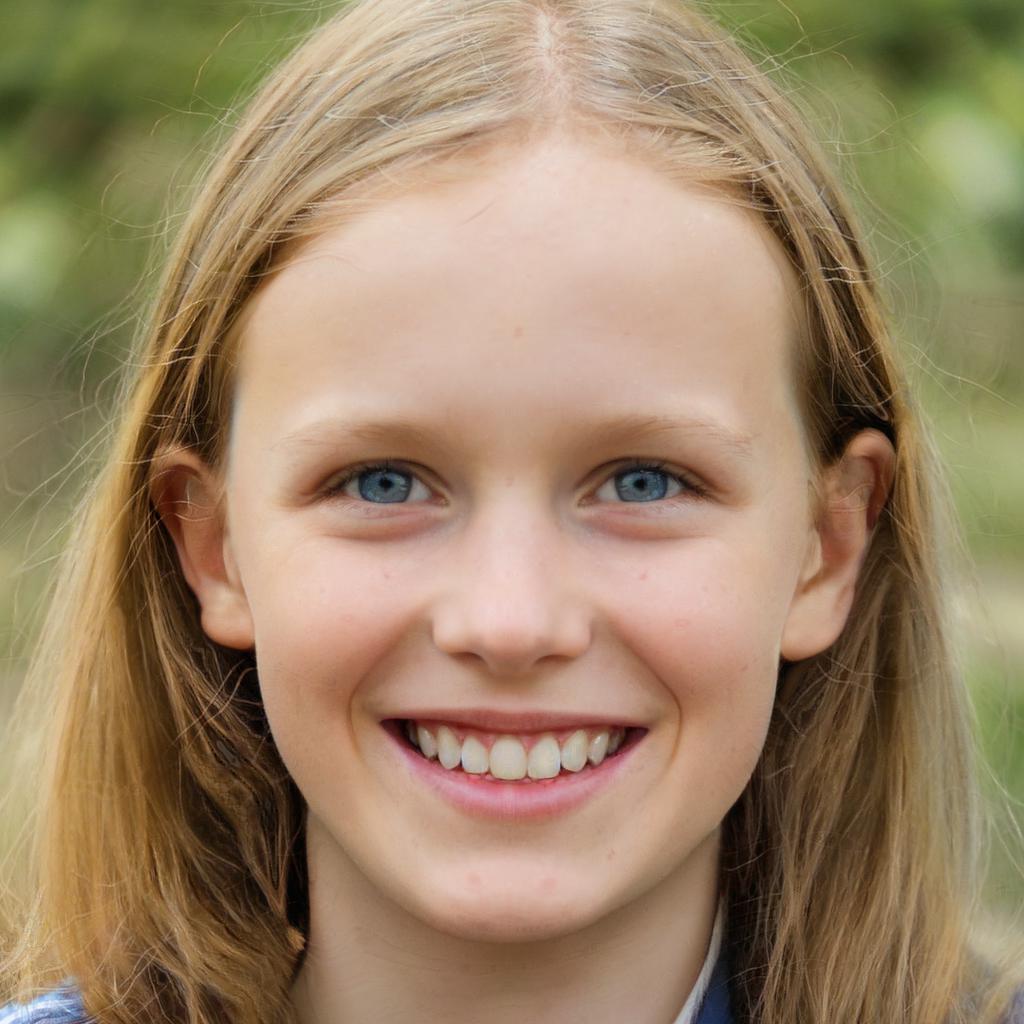} & 
		\includegraphics[width=0.155\columnwidth]{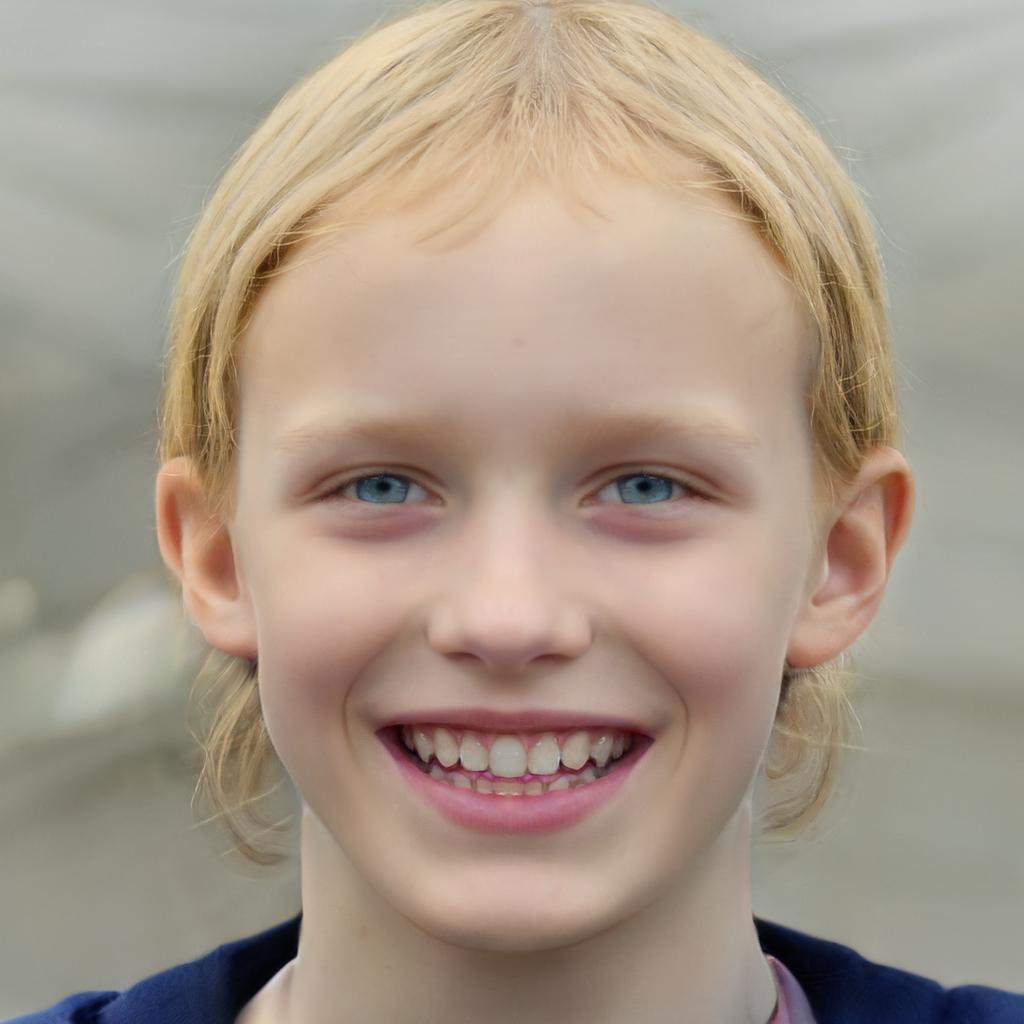} & 
		\includegraphics[width=0.155\columnwidth]{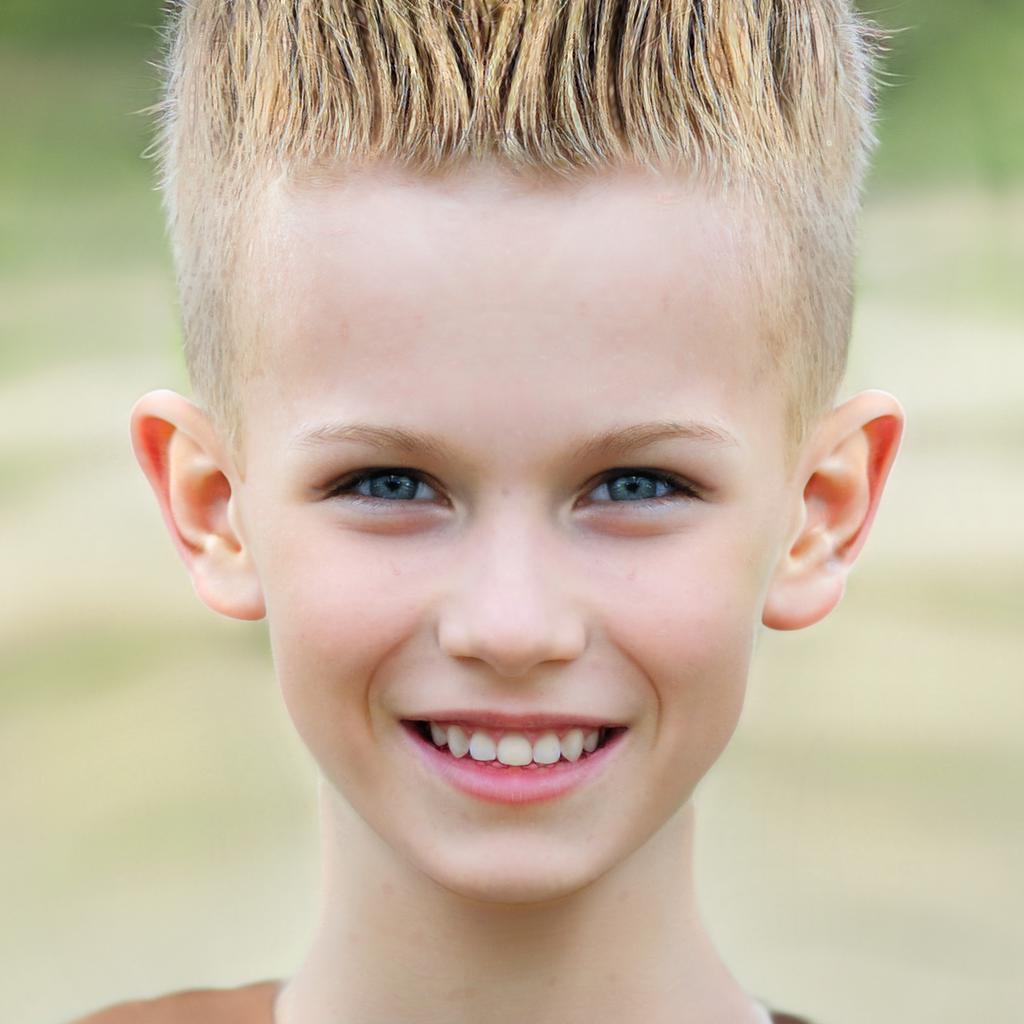} & 
		\includegraphics[width=0.155\columnwidth]{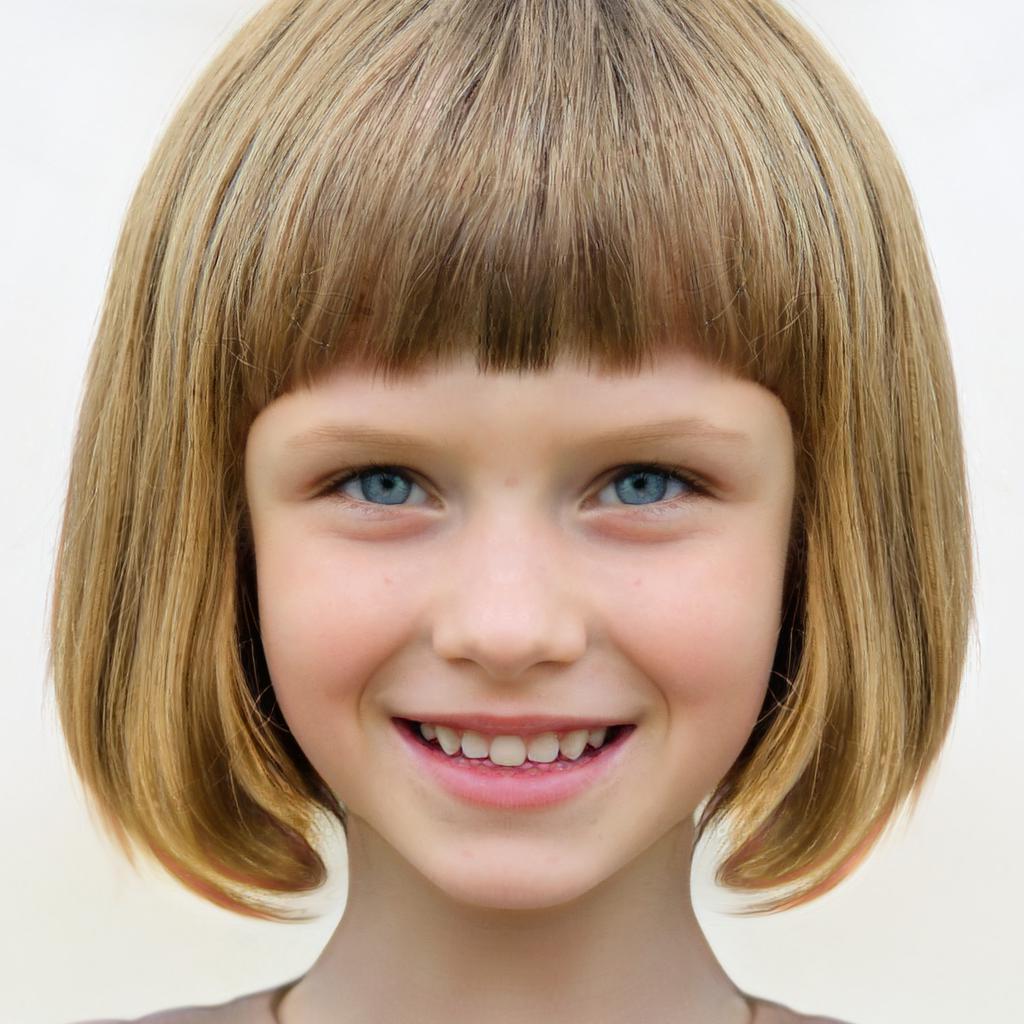} & 
		\includegraphics[width=0.155\columnwidth]{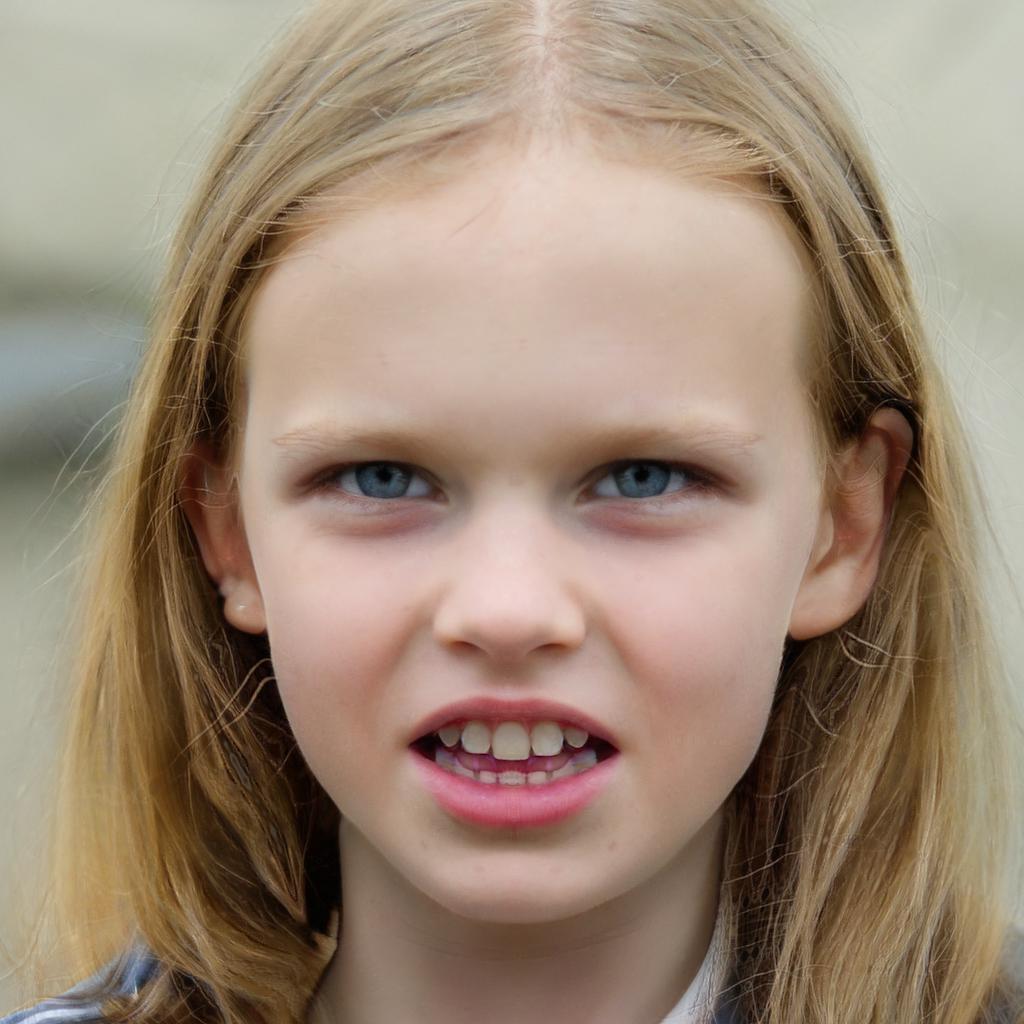} &
		\includegraphics[width=0.155\columnwidth]{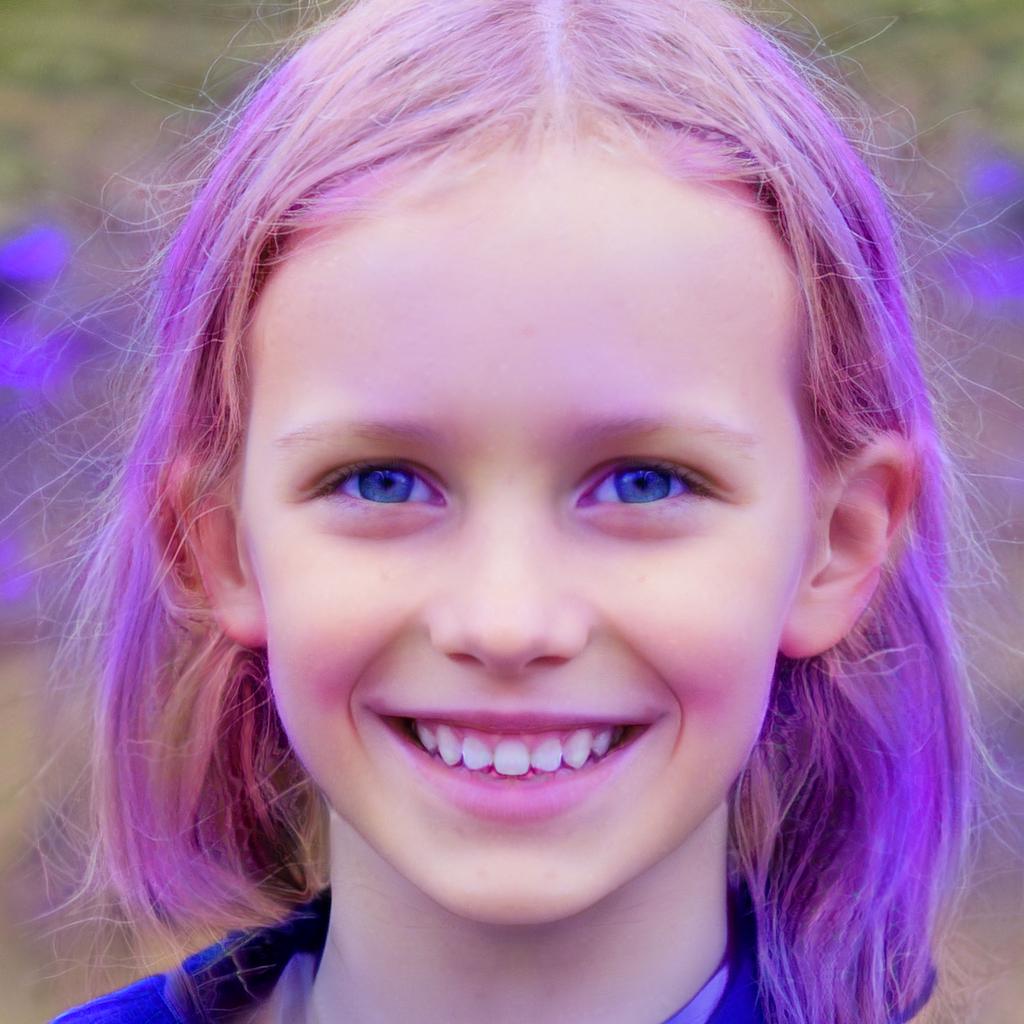}
		\\

		Source & Man & Mohawk & Bobcut & Angry & Purple Hair
	\end{tabular}
	}
	\vspace{-0.25cm}
	\caption{Non-linear editing in $\mathcal{W}+$. We edit images using the StyleCLIP mapping technique with StyleGAN3 trained on aligned faces. Even with non-linear editing paths, the edits are still entangled: local edits (e.g., expression/hairstyle) alter other attributes (e.g., background/identity).}
	\vspace{-0.25cm}
	\label{fig:styleclip-edits}
\end{figure}

%% file: figures/styleclip_global_joined.tex
\begin{figure}[tb]
	\centering
	\setlength{\tabcolsep}{1pt}
	
	{\small
	\begin{tabular}{c c c c c c}

		\includegraphics[width=0.1875\columnwidth]{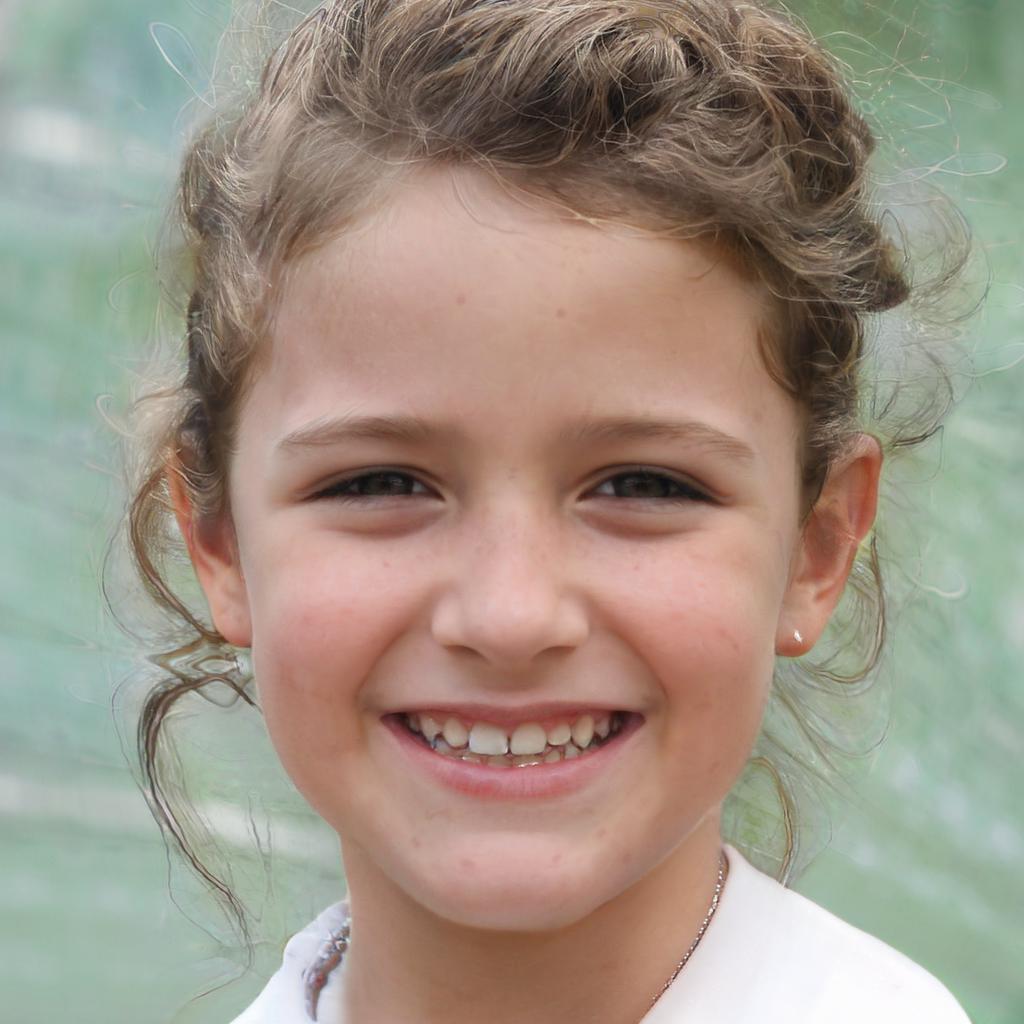} & 
        \includegraphics[width=0.1875\columnwidth]{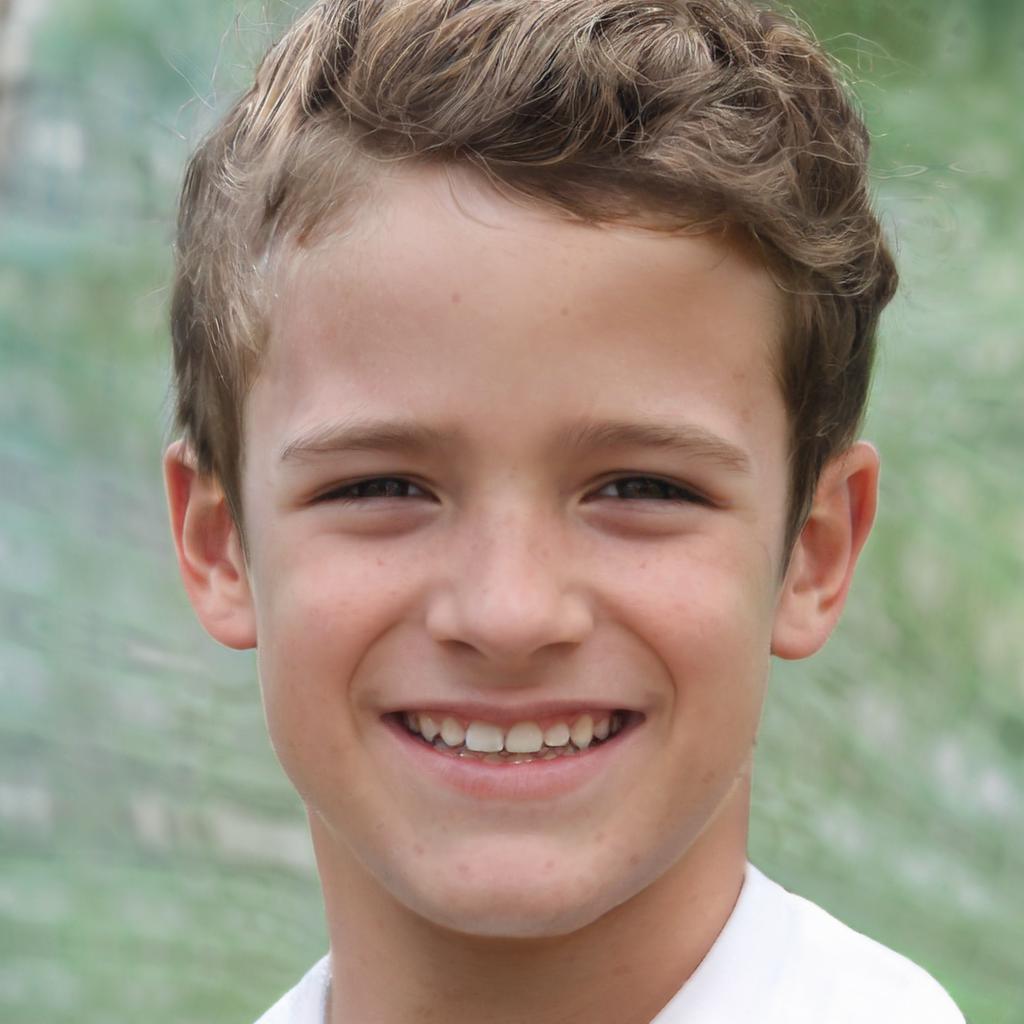} & 
        \includegraphics[width=0.1875\columnwidth]{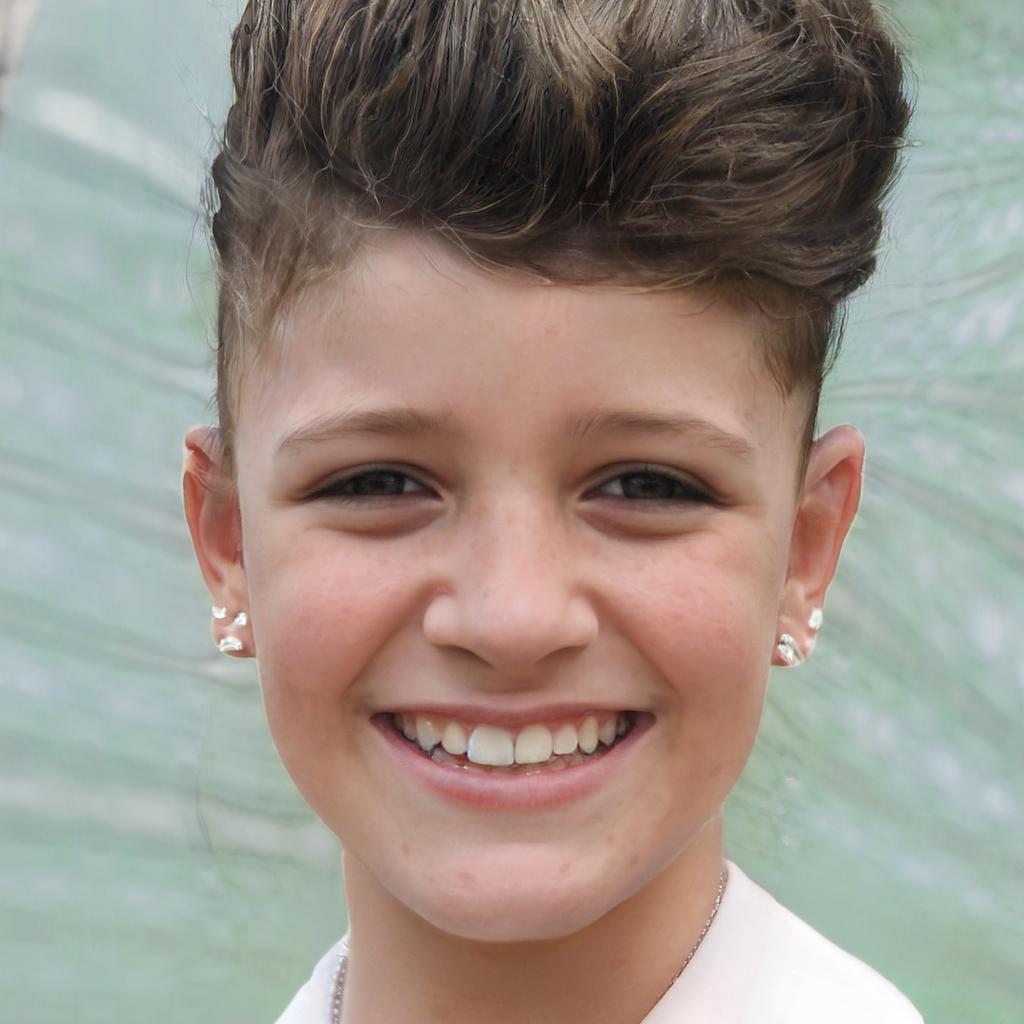} & 
        \includegraphics[width=0.1875\columnwidth]{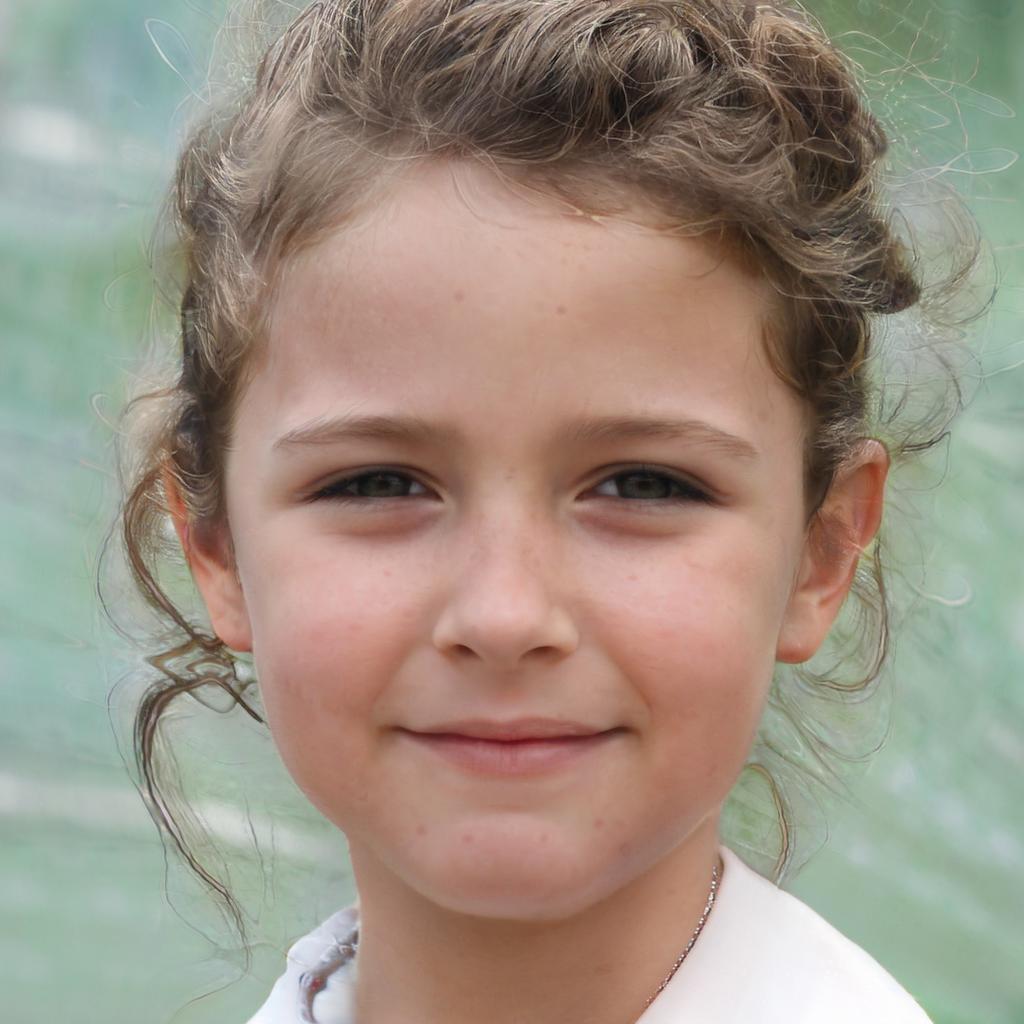} & 
        \includegraphics[width=0.1875\columnwidth]{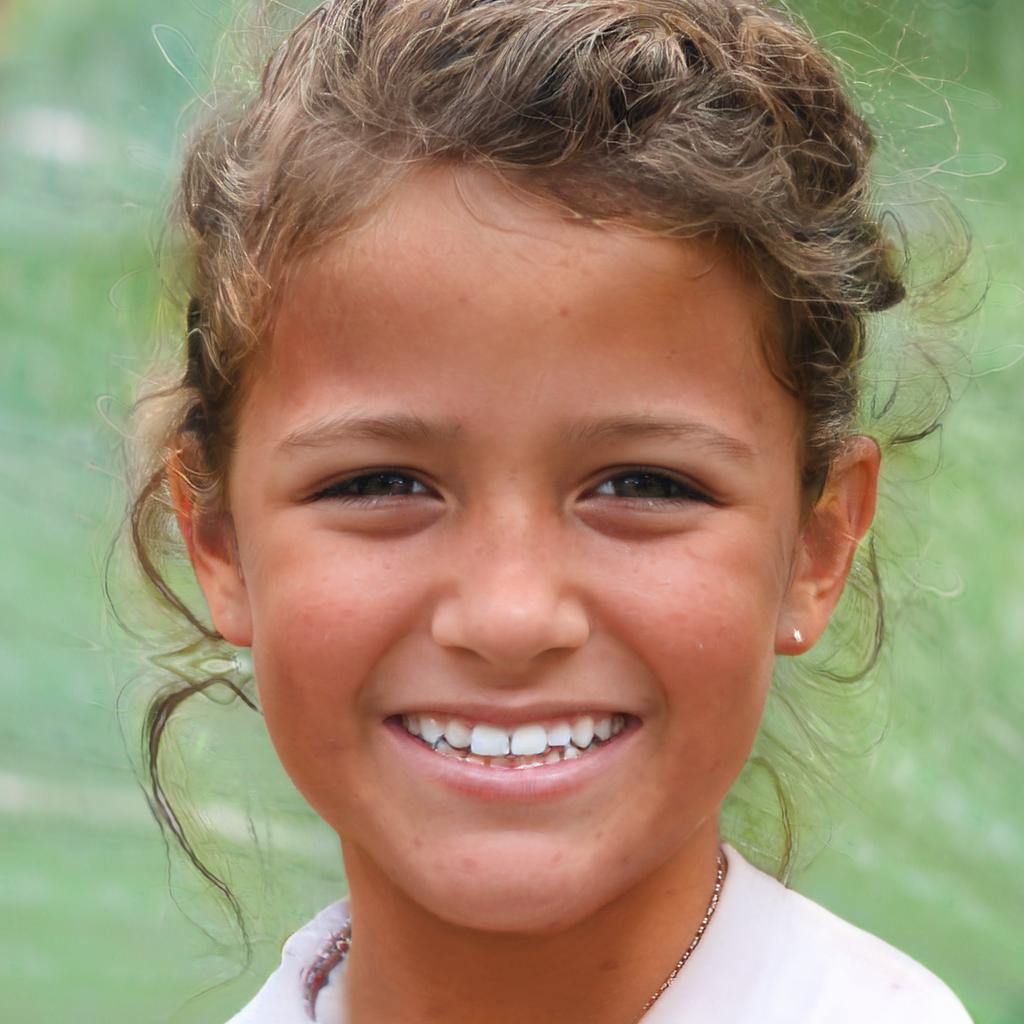} 
        \\
        
		\includegraphics[width=0.1875\columnwidth]{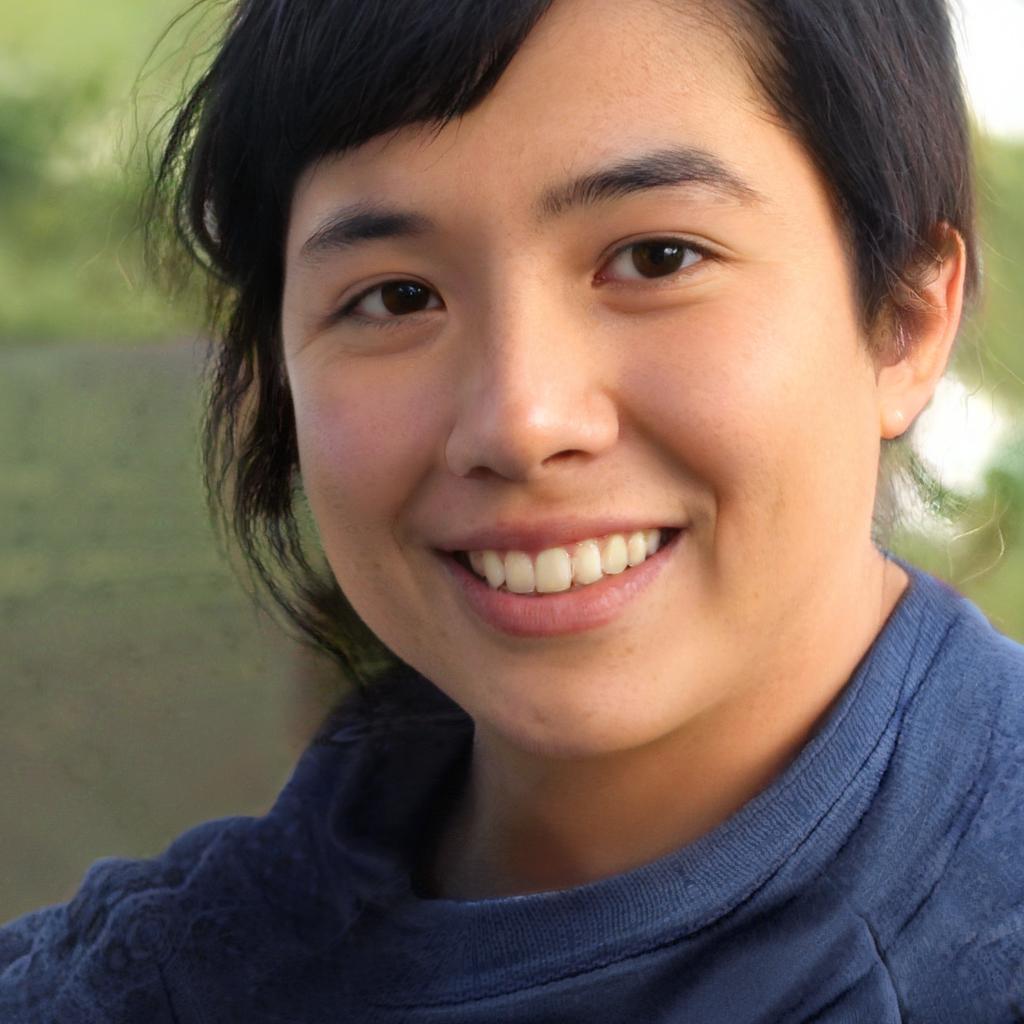} & 
        \includegraphics[width=0.1875\columnwidth]{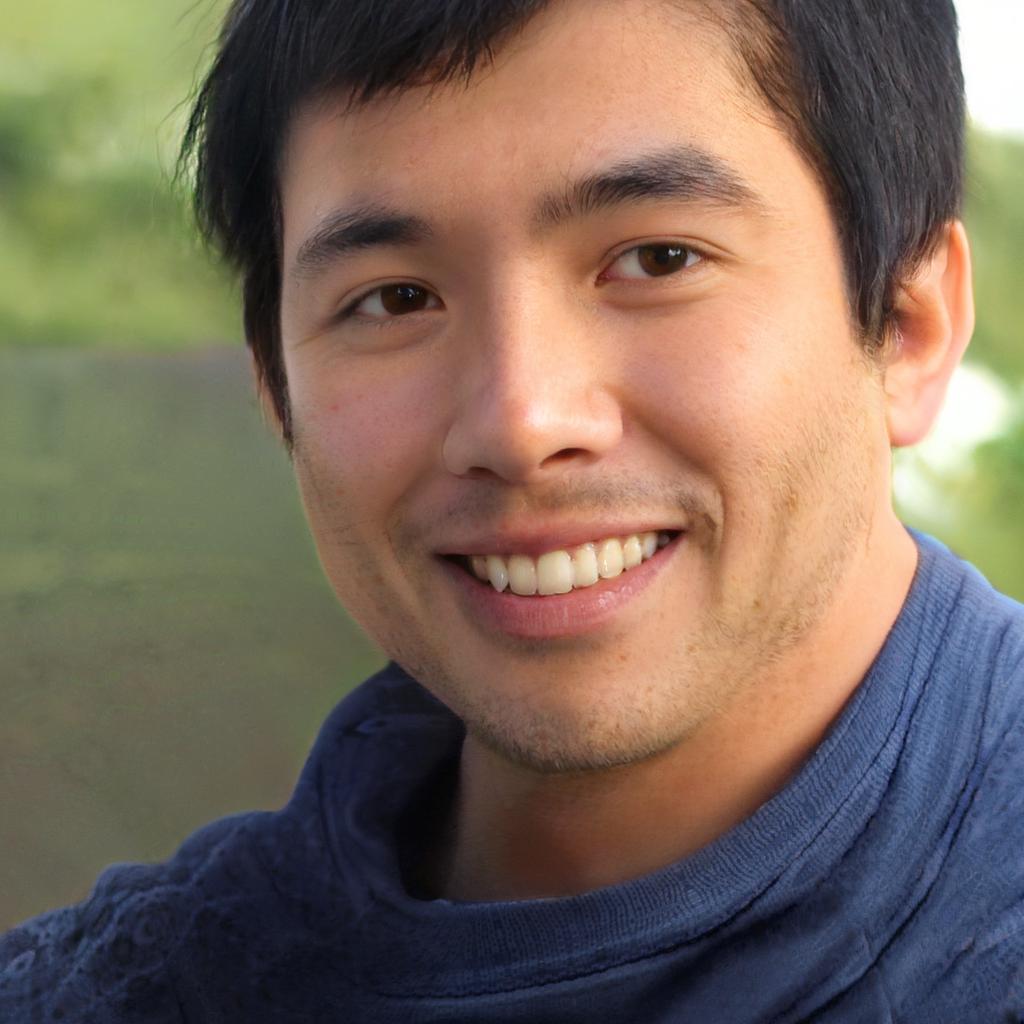} & 
        \includegraphics[width=0.1875\columnwidth]{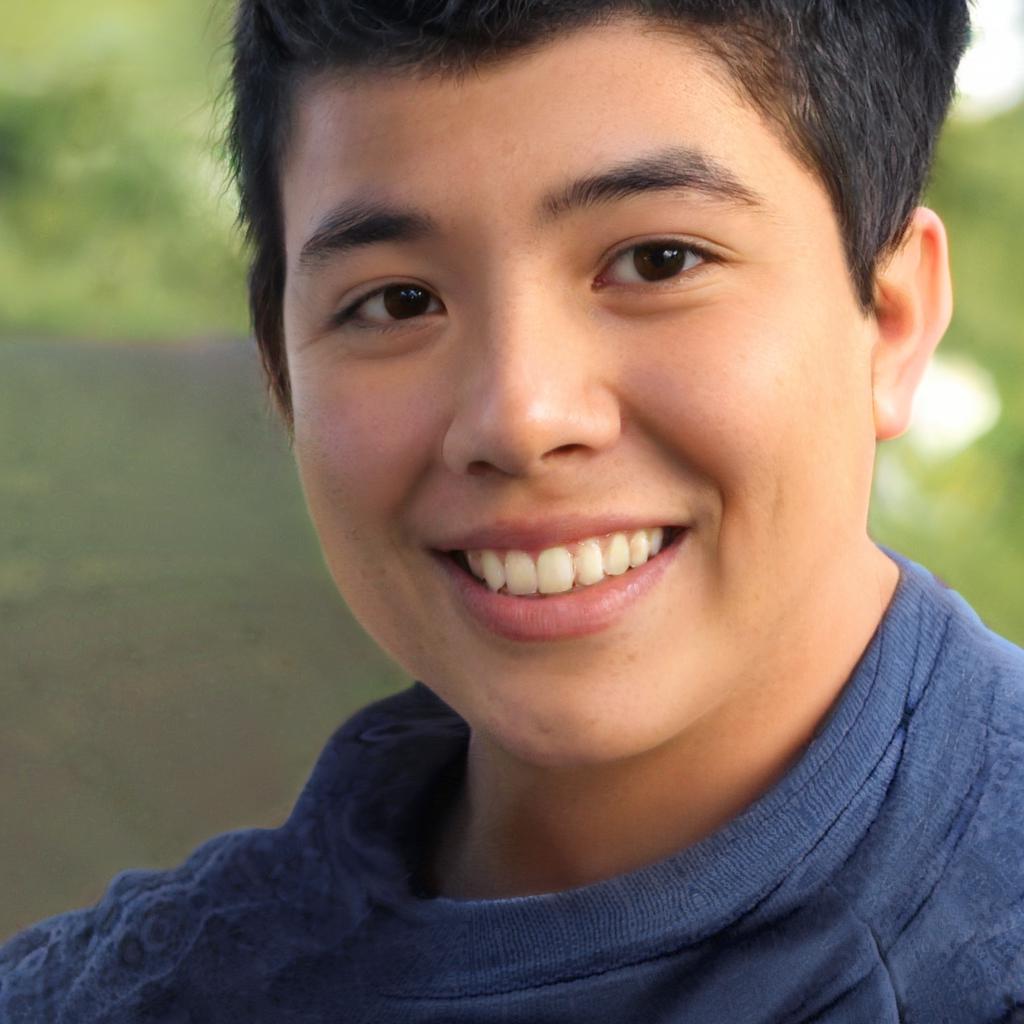} & 
        \includegraphics[width=0.1875\columnwidth]{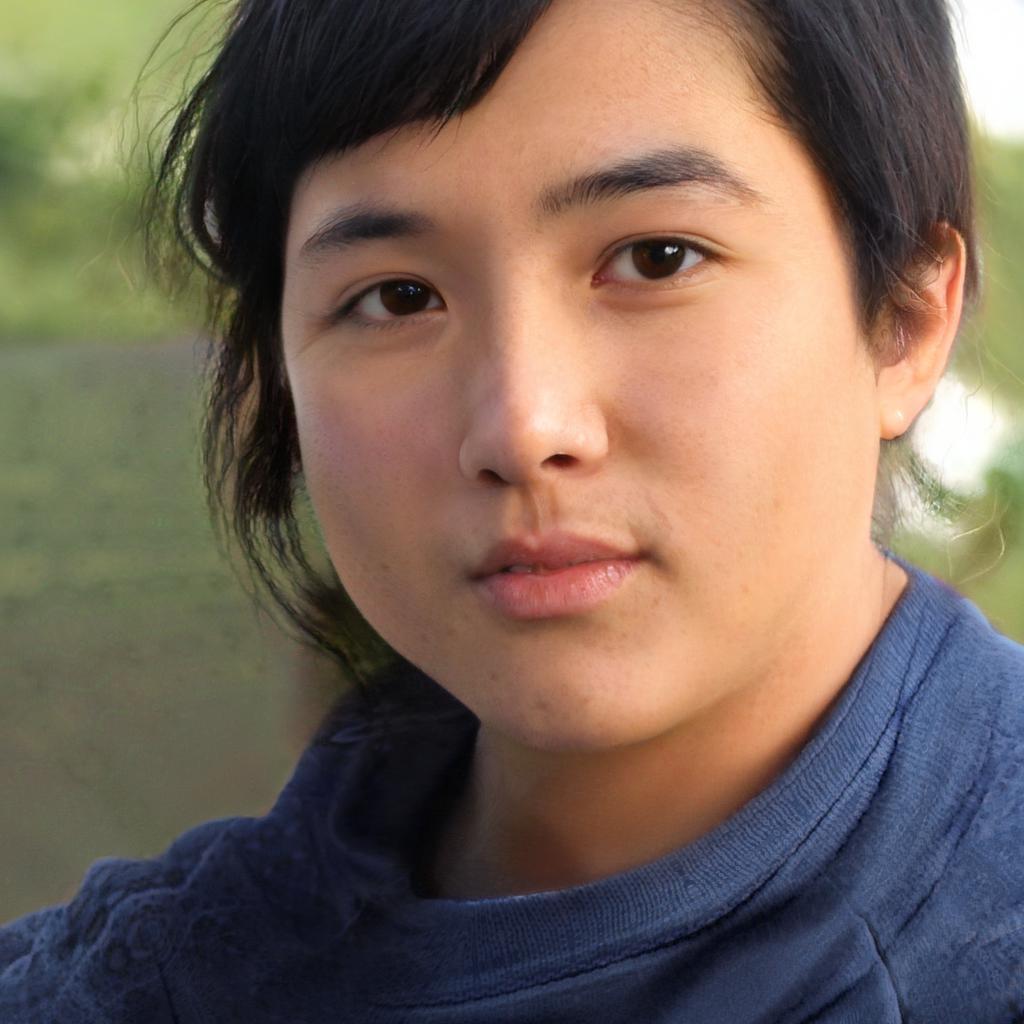} & 
        \includegraphics[width=0.1875\columnwidth]{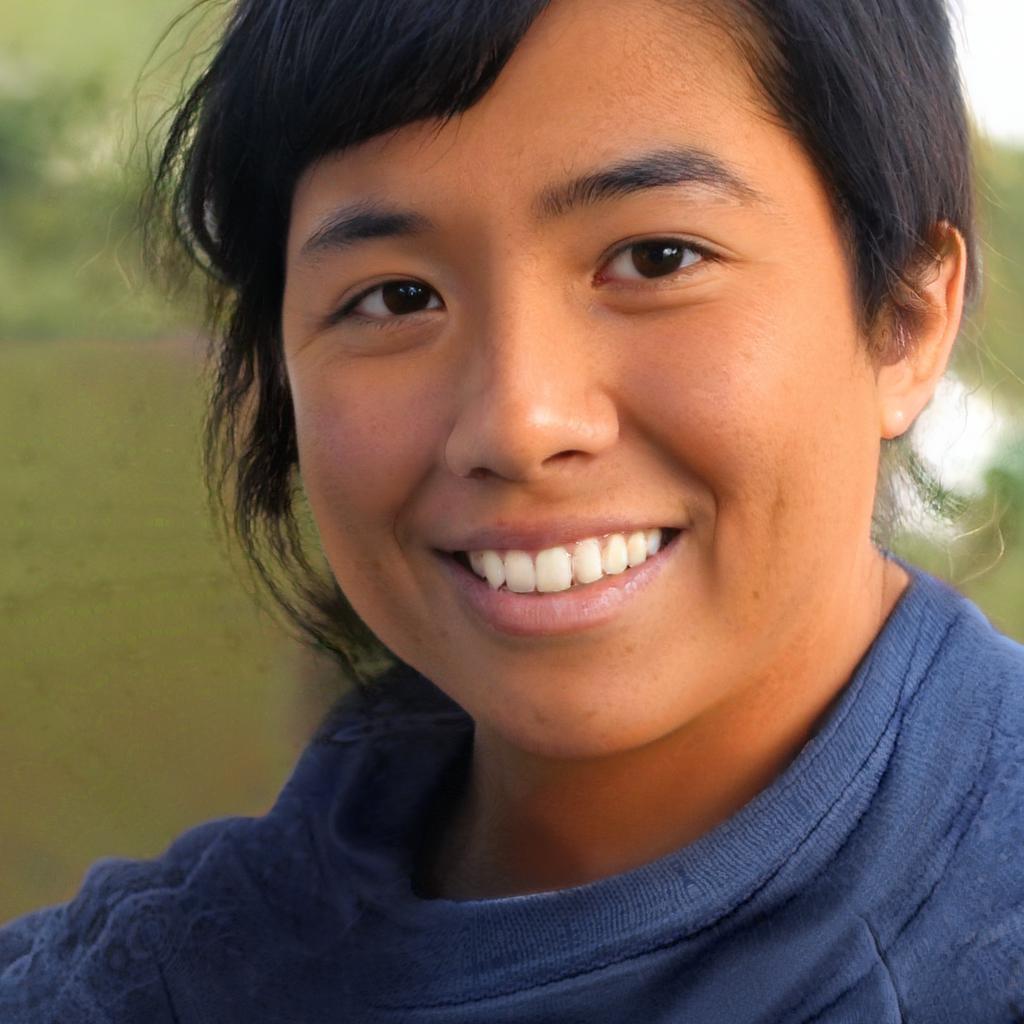} \\

		 Source & Gender & Hi-top Fade & Expression & Tanned
	\end{tabular}
	}
	
	{\small
	\begin{tabular}{c c c c c}

        \includegraphics[width=0.1875\columnwidth]{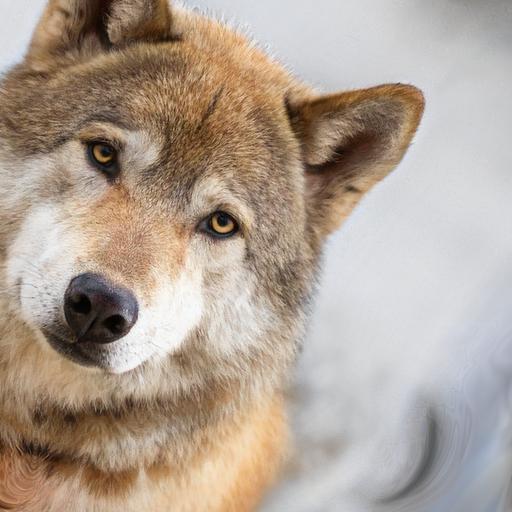} & 
        \includegraphics[width=0.1875\columnwidth]{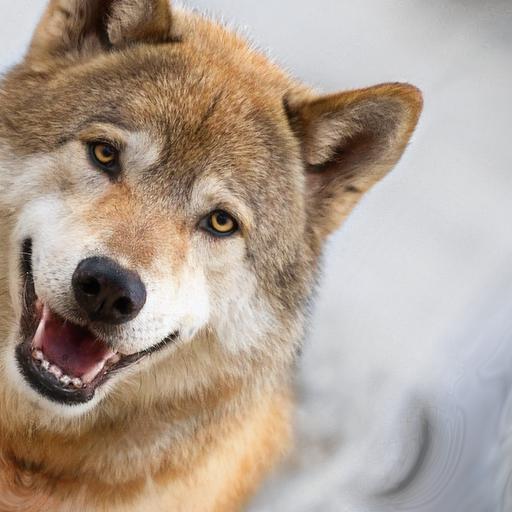} &
        \includegraphics[width=0.1875\columnwidth]{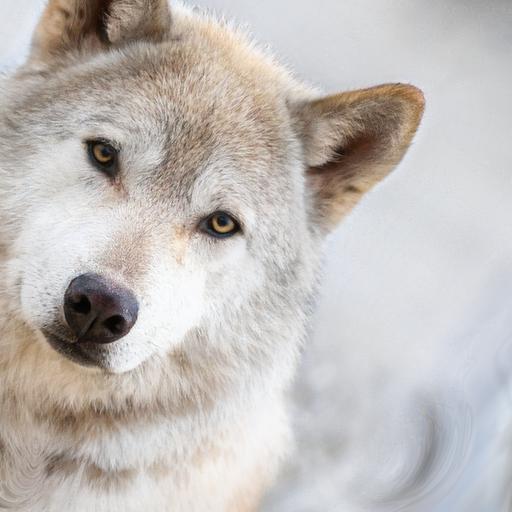} &
        \includegraphics[width=0.1875\columnwidth]{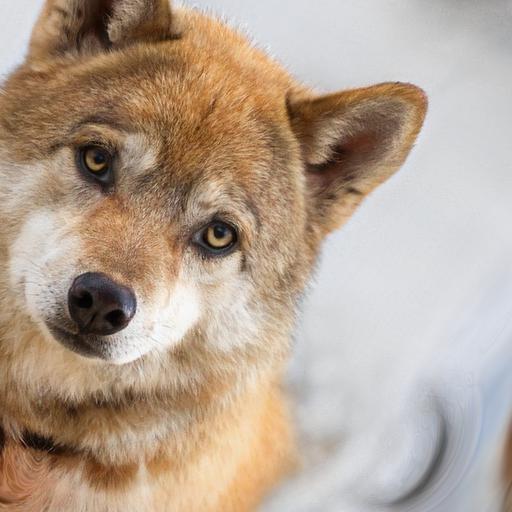} &
        \includegraphics[width=0.1875\columnwidth]{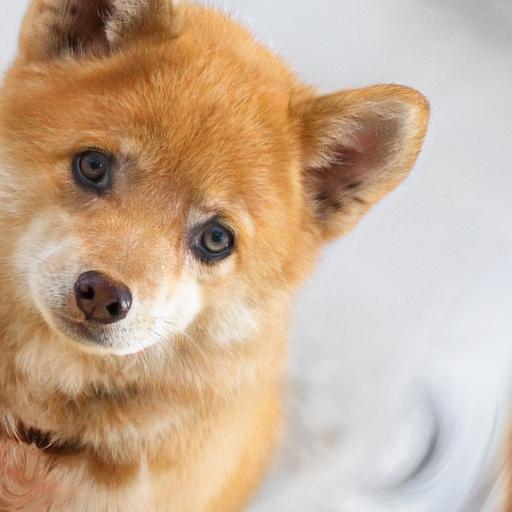}
        \\

        \includegraphics[width=0.1875\columnwidth]{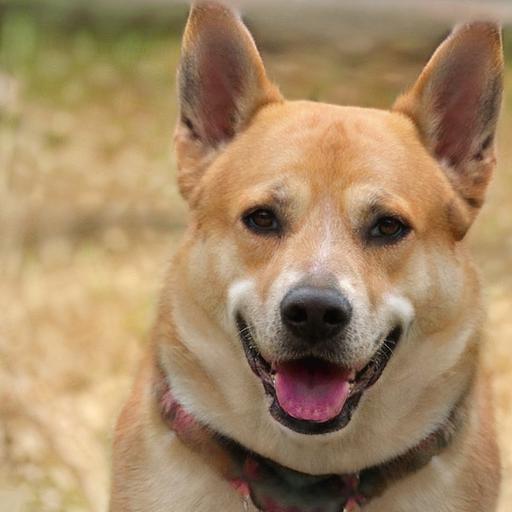} &
        \includegraphics[width=0.1875\columnwidth]{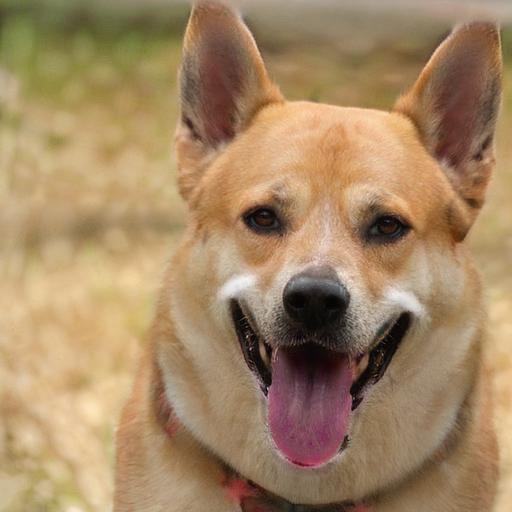} &
        \includegraphics[width=0.1875\columnwidth]{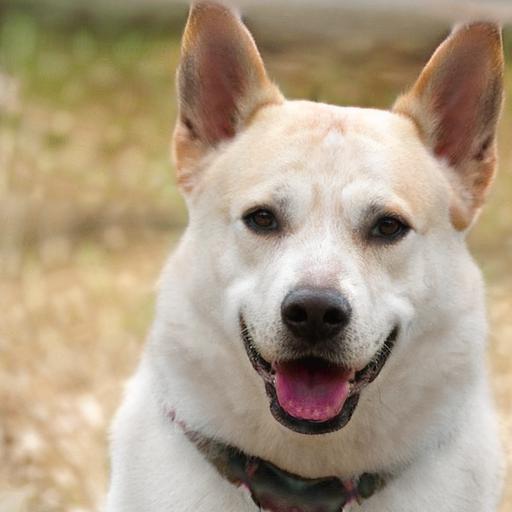} &
        \includegraphics[width=0.1875\columnwidth]{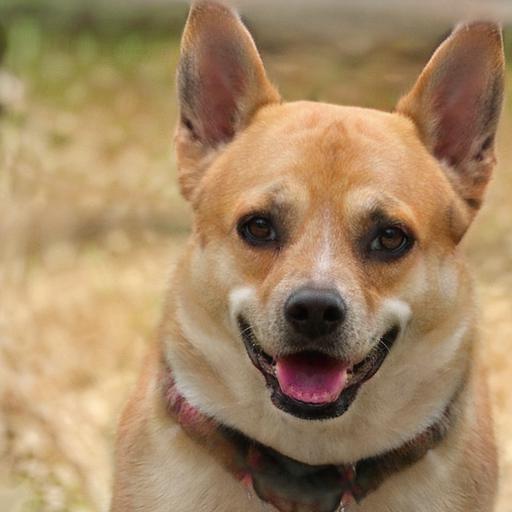} &
        \includegraphics[width=0.1875\columnwidth]{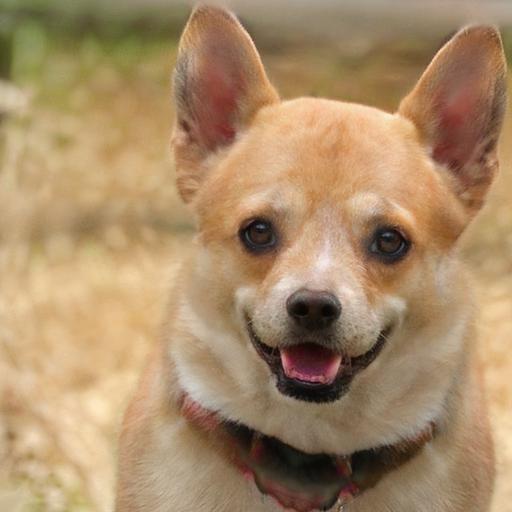} \\
        
		Source & Happy & White & Big Eyes & Baby
	\end{tabular}
	}
	
	{\small
	\begin{tabular}{c c c c c}
	
        \includegraphics[width=0.1875\columnwidth]{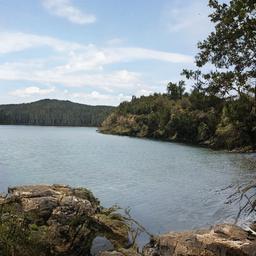} & 
        \includegraphics[width=0.1875\columnwidth]{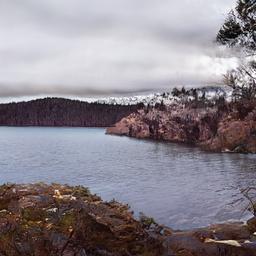} & 
        \includegraphics[width=0.1875\columnwidth]{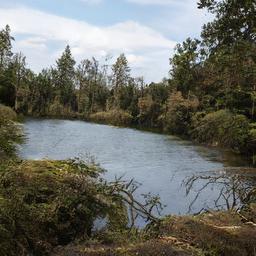} & 
        \includegraphics[width=0.1875\columnwidth]{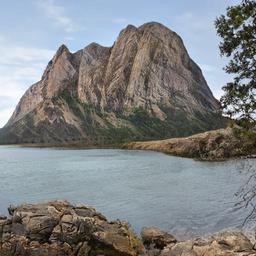} &
        \includegraphics[width=0.1875\columnwidth]{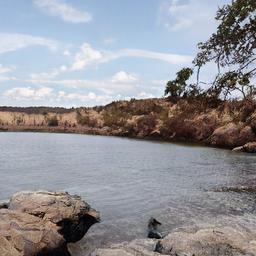} \\

        \includegraphics[width=0.1875\columnwidth]{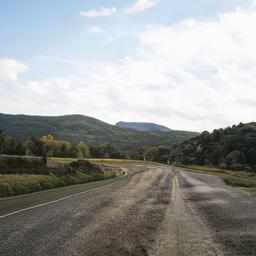} & 
        \includegraphics[width=0.1875\columnwidth]{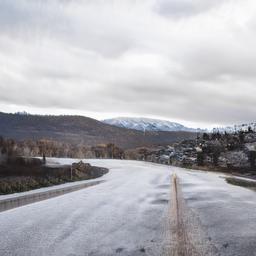} & 
        \includegraphics[width=0.1875\columnwidth]{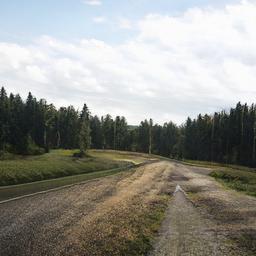} & 
        \includegraphics[width=0.1875\columnwidth]{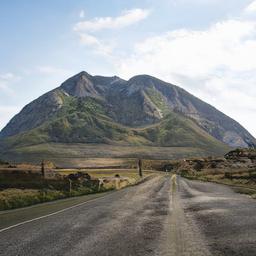} &
        \includegraphics[width=0.1875\columnwidth]{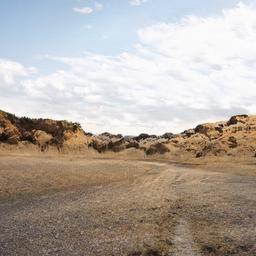}
        \\
		Source & Winter & Forest & Mountain & Desert
	\end{tabular}
    }
	
	\vspace{-0.3cm}
	\caption{Editing in $\mathcal{S}$. We edit synthetic images using the StyleCLIP~\cite{patashnik2021styleclip} global directions technique using StyleGAN3 generators trained on the FFHQ~\cite{karras2019style}, AFHQv2~\cite{choi2020stargan,aliasfreeKarras2021}, and Landscapes HQ~\cite{ALIS} datasets.
	}
	\vspace{-0.35cm}
	\label{fig:styleclip-global-directions}
\end{figure}

%% file: inversion.tex
\setlength{\abovedisplayskip}{3pt}
\setlength{\belowdisplayskip}{3pt}

\section{StyleGAN3 Inversion}~\label{sec:inversion}
\vspace{-0.15cm}

In this section, we address the task of inverting a pre-trained StyleGAN3 generator $G$. In other words, given a target image $x$, we seek a latent code $\hat{w}$ that optimally reconstructs it:
\begin{align}~\label{eq:inversion}
	\hat{w} = \underset{w}{\arg\min} \; \mathcal{L} \left ( x, G(w; (r, t_x, t_y) \right ),
\end{align}
where $\mathcal{L}$ is the $L_2$ or LPIPS~\cite{zhang2018perceptual} reconstruction loss. 

Motivated by the goal of employing StyleGAN3 for editing real videos, solving the inversion task via a learned encoder (as opposed to latent vector optimization) may assist in achieving better temporal consistency due to its natural smoothness and bias for learning lower frequency representations~\cite{rahaman2019spectral,tzaban2022stitch}. More formally, we seek to train an encoder $E$ over a large set of images $\{x_i\}_{i=1}^N$ for minimizing the objective: 
\begin{align}~\label{eq:encoding_objective}
\sum_{i=1}^N \; \mathcal{L} ( x_i, G ( E ( x_i ) ) ),
\end{align}
where $E(x)$ encodes an input image $x$ into a latent code $w$. 

As discussed in \cref{sec:preliminaries}, having obtained the latent code $w = E(x)$, an additional transformation may be passed to the generator to control the translation and rotation of the reconstructed image $y = G(w; (r, t_x, t_y))$. Finally, some latent manipulation $f$ may also be applied over this latent code to obtain an edited image
\begin{equation}
    y_{\textit{edit}} = G(f(w); (r, t_x, t_y)).
\end{equation}

\vspace{-0.2cm}
\subsection{Designing the Encoder Network}
To enable the encoding and editing of aligned and unaligned images (such as those found in a video sequence), our inversion scheme must support the generation of both input types. 
A natural first attempt at doing so is to design an encoder trained on both types of images, paired with an unaligned generator. Specifically, one can employ the training schemes of existing StyleGAN2 encoders~\cite{richardson2020encoding,tov2021designing,alaluf2021restyle} to minimize the objective given in \cref{eq:encoding_objective} for unaligned inputs. Yet, we find that such a training scheme struggles in capturing the high variability of the unaligned facial images, resulting in poor reconstructions, see \cref{sec:ablation_study} for an ablation study of such a design.

\paragraph{\textbf{Encoding Unaligned Images. }}
We instead choose to leverage an aligned StyleGAN3 generator and design an encoder trained solely on \emph{aligned} images. As previously shown, this scheme can then be used for editing and synthesizing both aligned and unaligned imagery. 
In this formulation, the encoder no longer needs to correctly capture the highly-variable placement and pose of the unaligned input images. This in turn simplifies the encoder's training objective, allowing it to instead focus on faithfully capturing the input identity and other image features.

Given the encoder trained to reconstruct aligned images, we are left with the question of how to extend this encoding scheme to support the encoding and editing of unaligned images at inference time. Assume we have a given unaligned image $x_{\textit{unaligned}}$. We begin by using an off-the-shelf facial detector~\cite{dlib09} to detect and align the image, resulting in an aligned version of the input, denoted by $x_{\textit{aligned}}$.
We then predict the translation $(t_x,t_y)$ and rotation $r$ between $x_{\textit{aligned}}$ and $x_{\textit{unaligned}}$ by detecting and aligning the eyes in the two images. We refer the reader to \cref{sec:computing_transforms} for details on computing these parameters. 

Finally, the inversion and resulting reconstruction of the unaligned input $x_{\textit{unaligned}}$ are given by:
\begin{align*}
    w_{\textit{aligned}} &= E(x_{\textit{aligned}}) \\
    y_{\textit{unaligned}} &= G(w_{\textit{aligned}}; (r,t_x,t_y)).
\end{align*}
Observe that while the encoder receives the aligned image $x_{\textit{aligned}}$, the reconstruction is able to capture the placement and rotation of the unaligned input through the use of the extracted transformation $(r,t_x,t_y)$. 
As such, our inversion scheme, although trained solely on aligned images, is able to faithfully encode \textit{both} aligned and unaligned images by leveraging the unique design of StyleGAN3.

\vspace{-0.2cm}
\paragraph{\textbf{Additional Details. }}
In practice, we employ the pSp~\cite{richardson2020encoding} and e4e~\cite{tov2021designing} encoders for performing the inversion task. We additionally follow the ReStyle iterative refinement scheme from Alaluf~\etal~\cite{alaluf2021restyle} to gradually refine the predicted inversion via small number of forward passes (e.g., $3$) through the encoder. Please see \cref{sec:encoder_details} for additional details.

\input{figures/inversion_comparison}

\subsection{Inverting Images into StyleGAN3} 
We now compare our inversion scheme introduced above to the $\text{ReStyle}_{pSp}$ and $\text{ReStyle}_{e4e}$ encoders used for inverting StyleGAN2. 

\paragraph{\textbf{Qualitative Evaluation. }}
As shown in \cref{fig:inversions}, our StyleGAN3 encoders attain visually comparable results to their StyleGAN2 counterparts. Observe that with StyleGAN3 we are able to faithfully reproduce the input position, even when given aligned inputs, by using our landmark-based predicted transformations.

\paragraph{\textbf{Quantitative Evaluation. }}
In \cref{tb:quantitative_inversion} we provide a quantitative comparison between encoder-based inversion techniques for both StyleGAN2 and StyleGAN3 generators on the human facial domain. Since StyleGAN2 is limited to encoding aligned images, we perform our evaluation on the CelebA-HQ~\cite{karras2017progressive,liu2015deep} test set.
In addition to the inference time required by each inversion technique, we report the $L_2$ distance, the LPIPS~\cite{zhang2018perceptual} distance, identity similarity~\cite{huang2020curricularface}, and MS-SSIM~\cite{wang2003multiscale} score between the reconstructions and their sources. 

Our StyleGAN3 encoders reach a slightly worse performance compared to the StyleGAN2 encoders. We believe the higher difficulty in inverting StyleGAN3 is in part due to its less well-behaved $\mathcal{W}+$ latent space. This is also supported by our experiment in \cref{sec:analysis-disentanglement}, where we observed that the quality of the generated images in StyleGAN3 quickly deteriorates as we move away from the $\mathcal{W}$ space. We believe this quick collapse of the latent space contributes to the challenge of training inversion encoders for StyleGAN3.

\input{tables/inversion}

\paragraph{\textbf{Editability via Latent Space Manipulation. }}
We now turn to evaluating the \emph{editability} of our ReStyle encoders for StyleGAN3. As illustrated in \cref{fig:editing_reals}, our $\text{Restyle}_{e4e}$ encoder achieves realistic and meaningful edits while  preserving the input identity. This is in contrast to $\text{Restyle}_{pSp}$, which despite achieving high-quality reconstructions, yields visibly less editable inversions. 
Notably, observe that in StyleGAN3, the gap in editing quality achieved by $\text{ReStyle}_{e4e}$ compared to $\text{ReStyle}_{pSp}$ is much larger compared to in StyleGAN2. This is most evident in artifacts along the hair in the second and last rows and hints at the increased importance of inverting into well-behaved latent regions compared to in StyleGAN2.

\input{figures/editing_real_images}

%% file: figures/inversion_comparison.tex
\begin{figure}[tb]
	\centering
	\setlength{\tabcolsep}{1pt}	
	{\footnotesize
	\begin{tabular}{c c c c c c}

        \includegraphics[width=0.195\columnwidth]{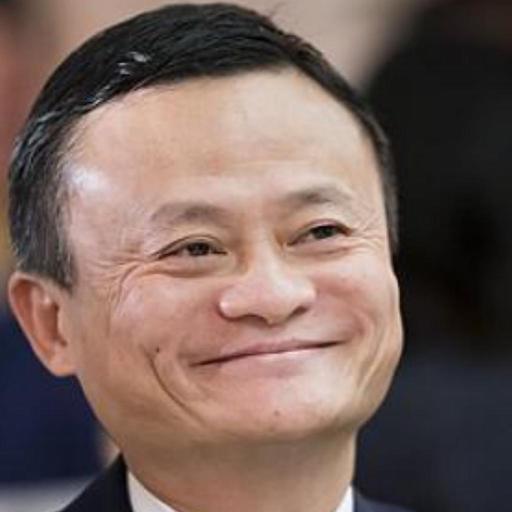} & 
        \includegraphics[width=0.195\columnwidth]{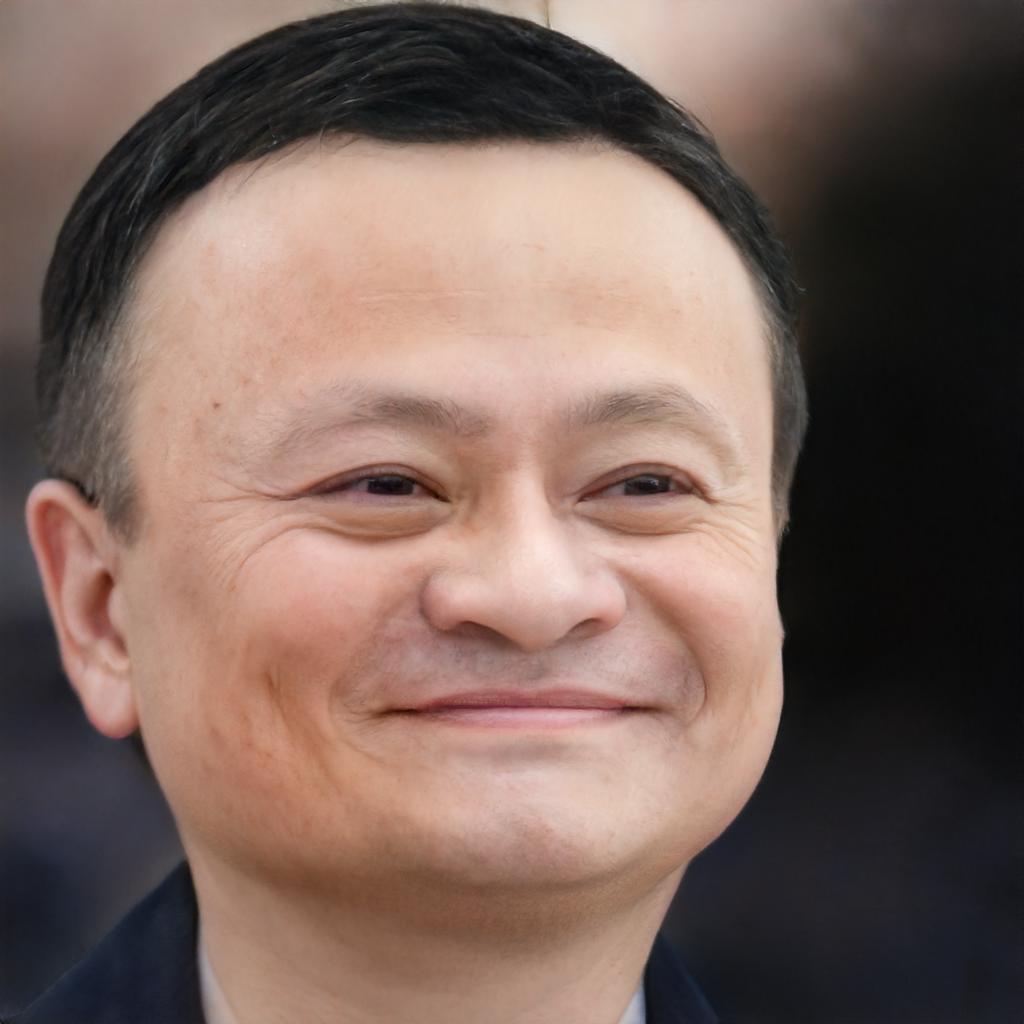} & 
        \includegraphics[width=0.195\columnwidth]{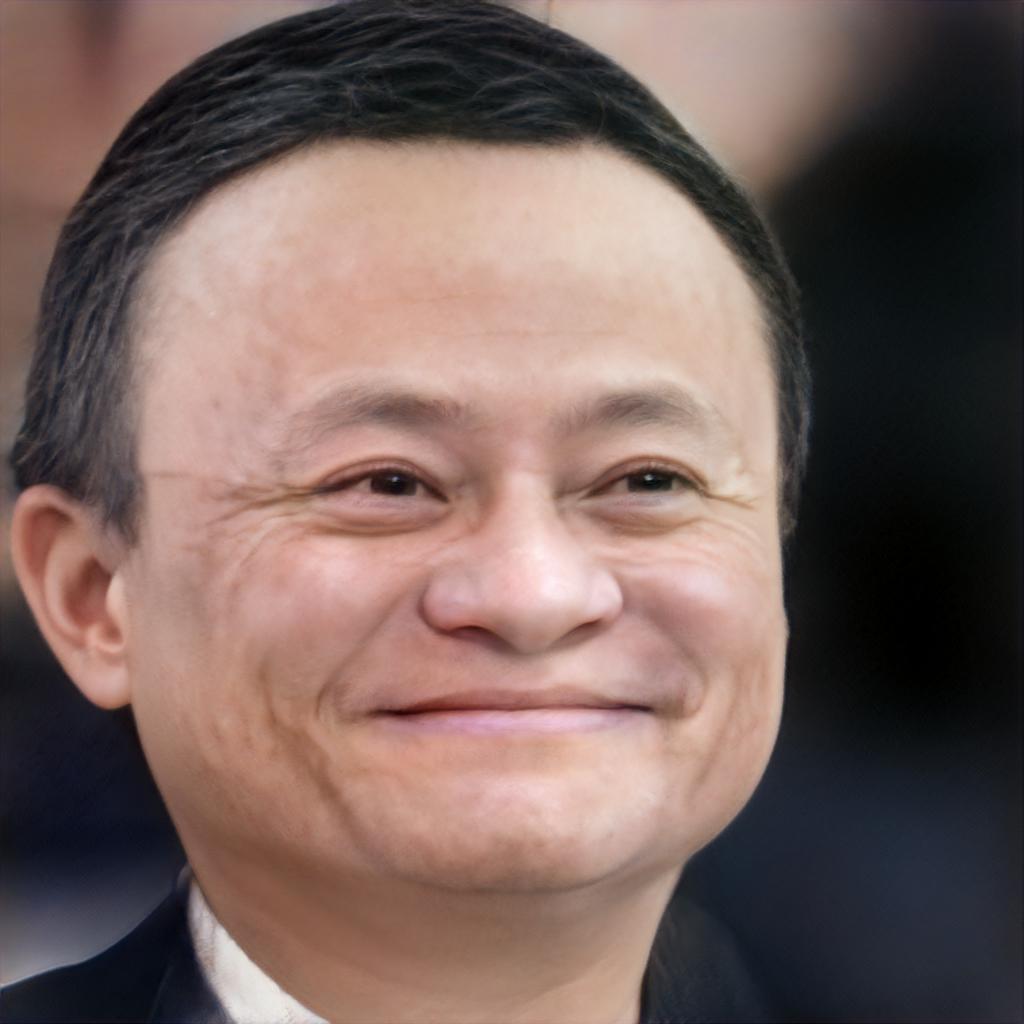} & 
        \includegraphics[width=0.195\columnwidth]{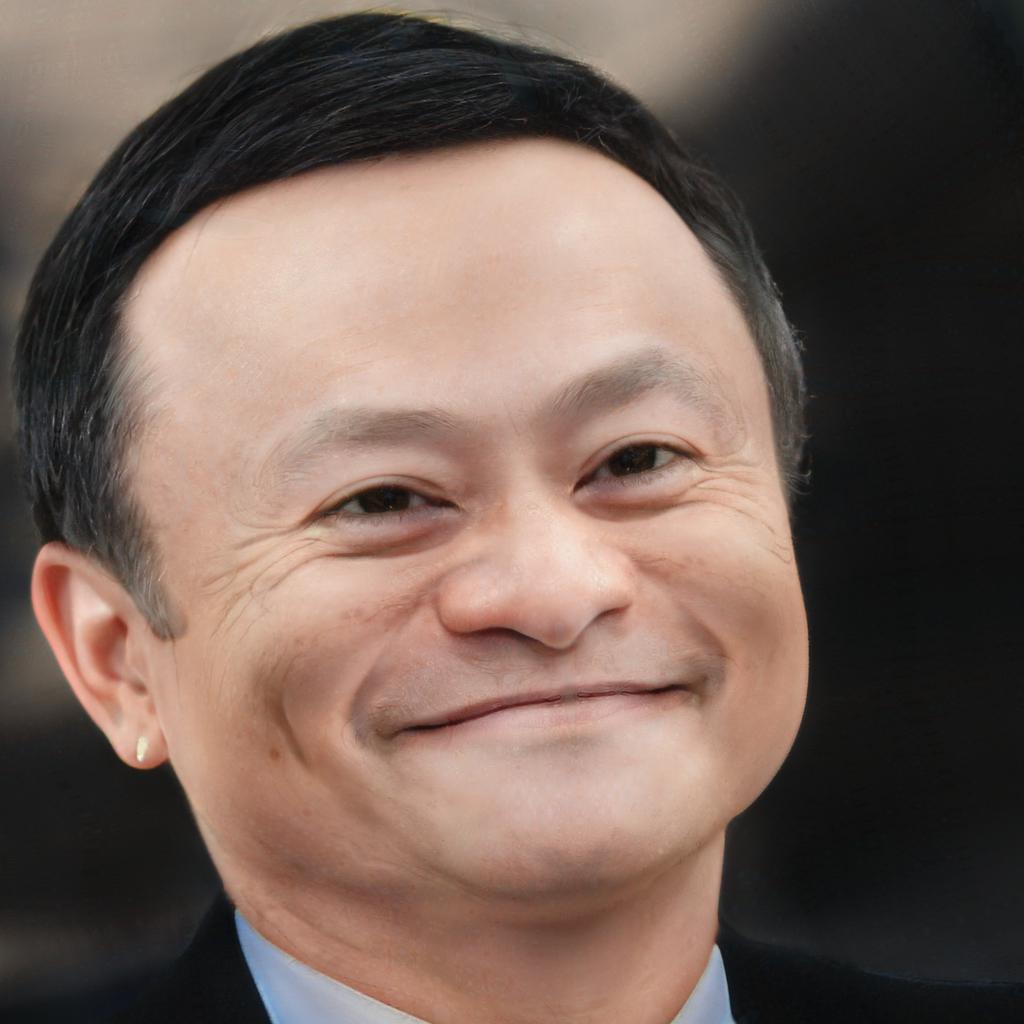} & 
        \includegraphics[width=0.195\columnwidth]{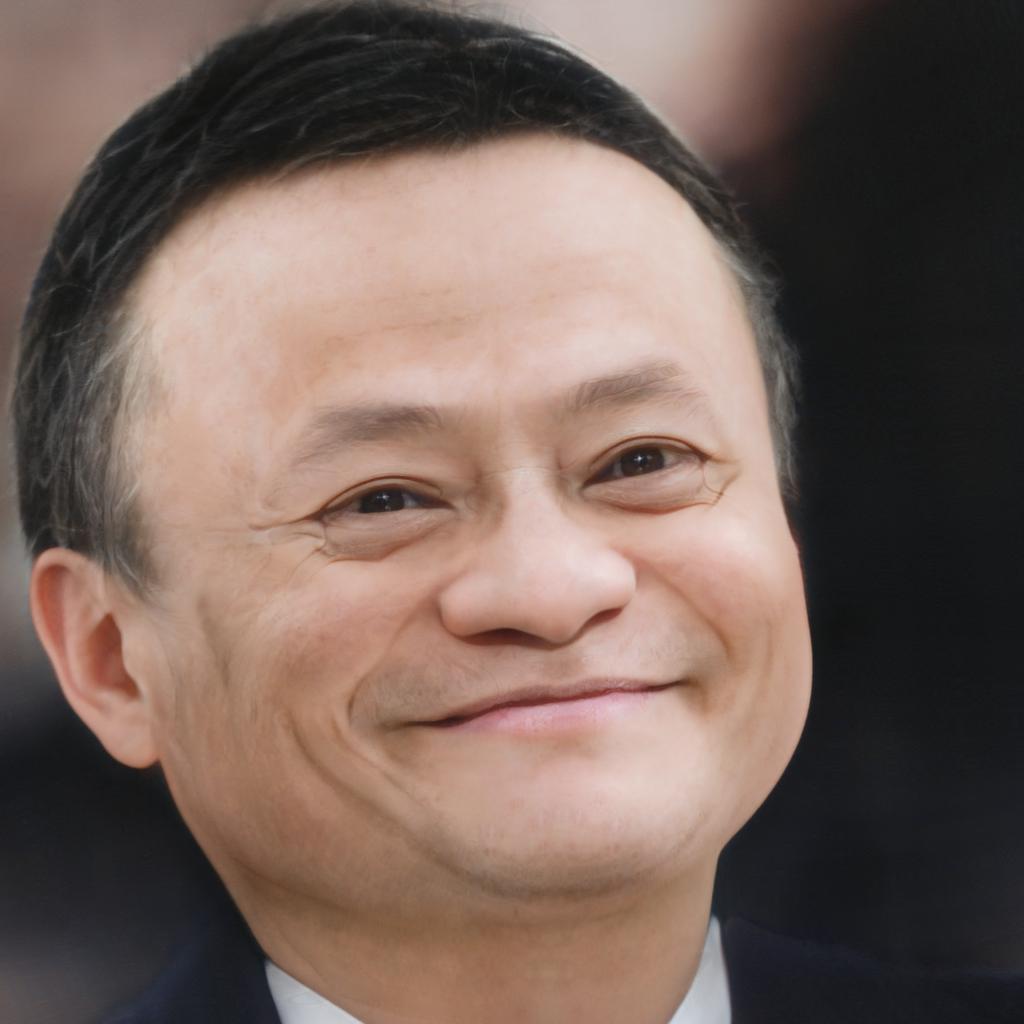} \\

        \includegraphics[width=0.195\columnwidth]{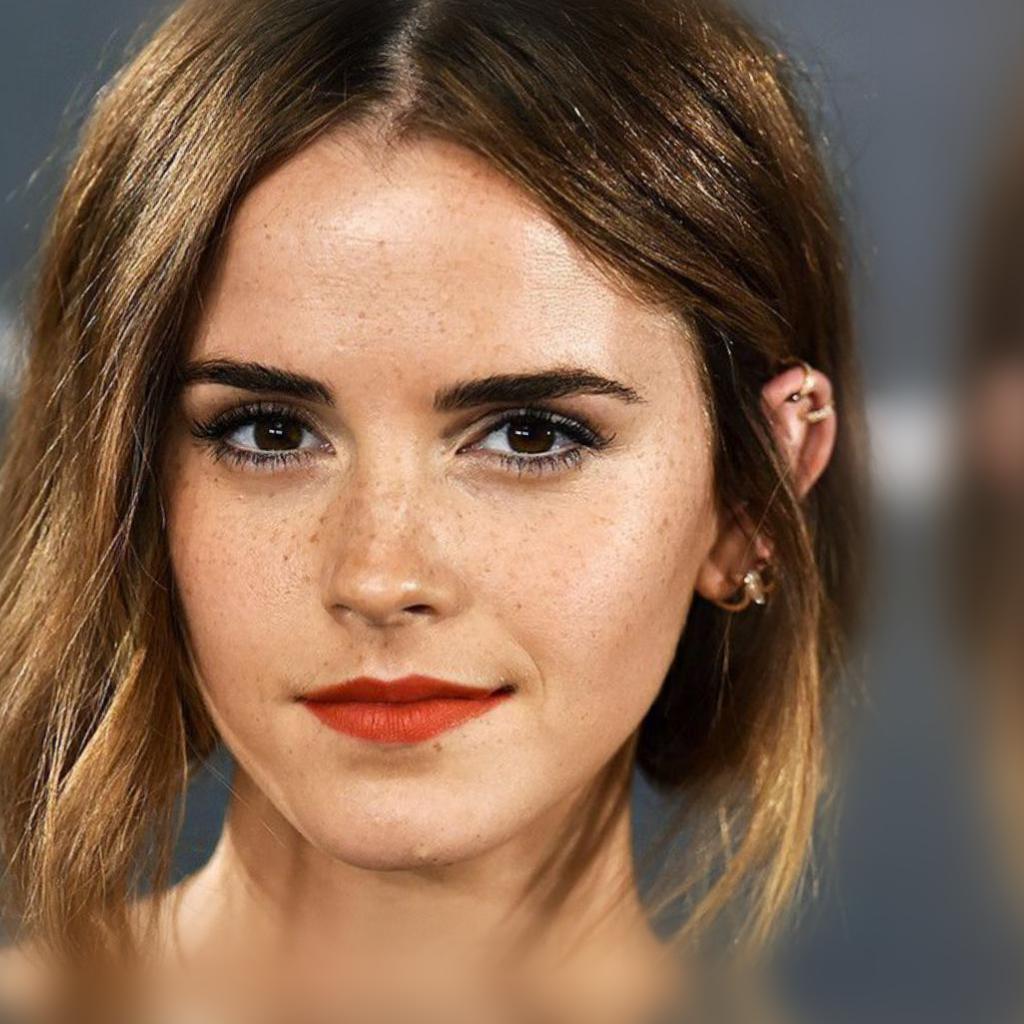} & 
        \includegraphics[width=0.195\columnwidth]{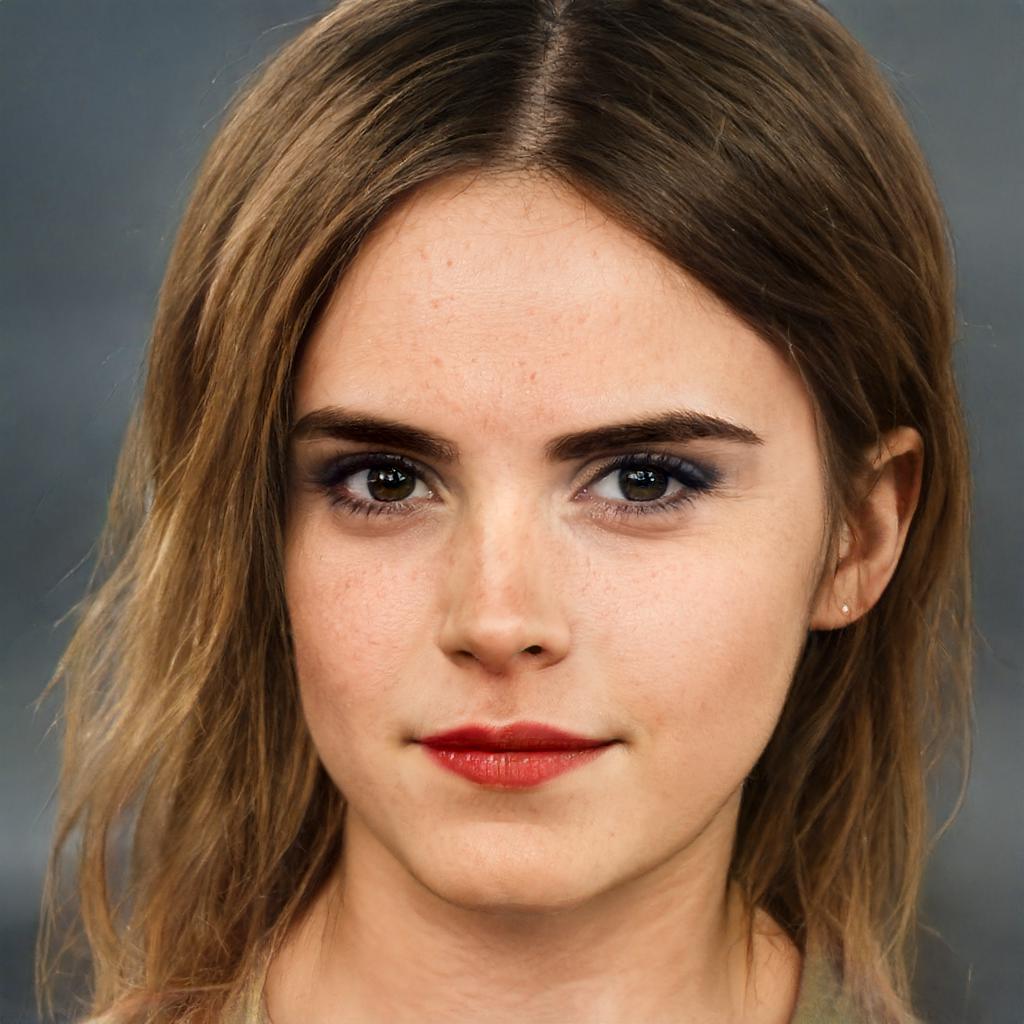} & 
        \includegraphics[width=0.195\columnwidth]{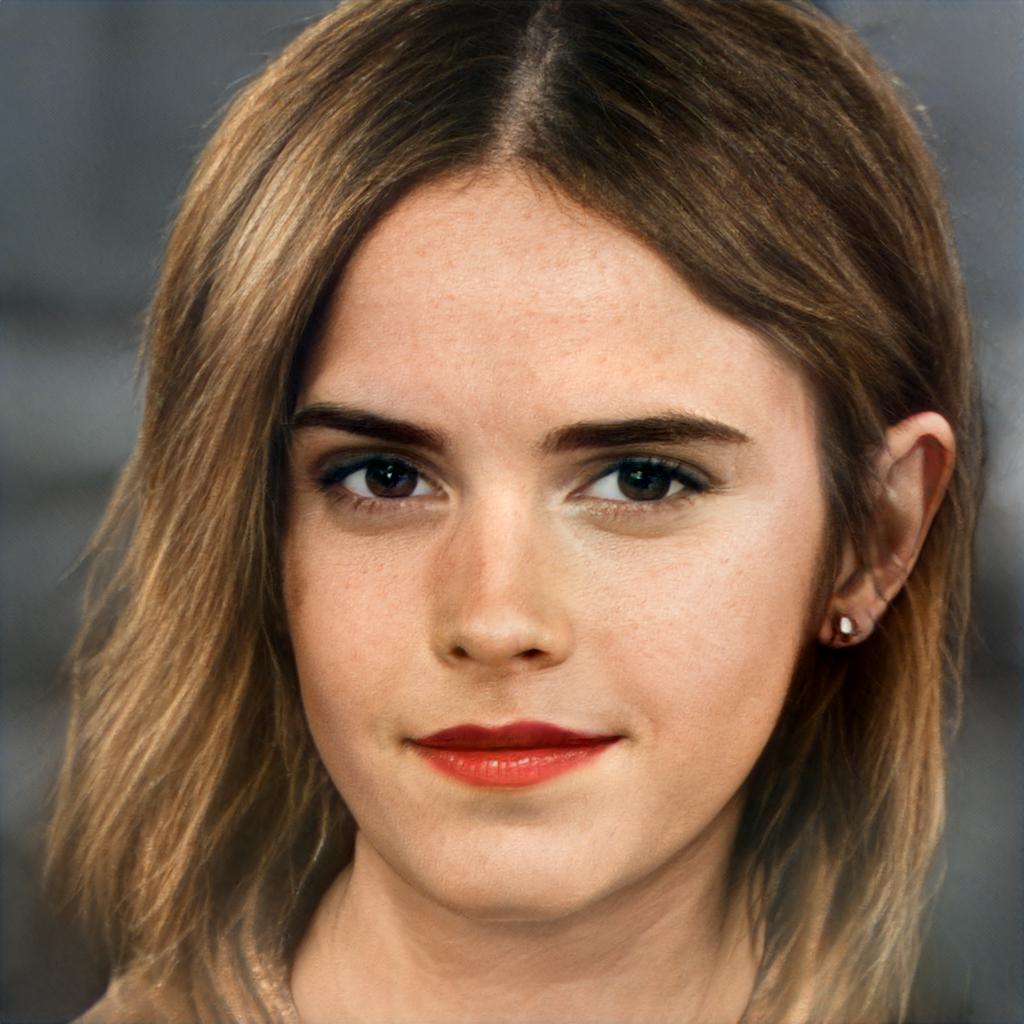} & 
        \includegraphics[width=0.195\columnwidth]{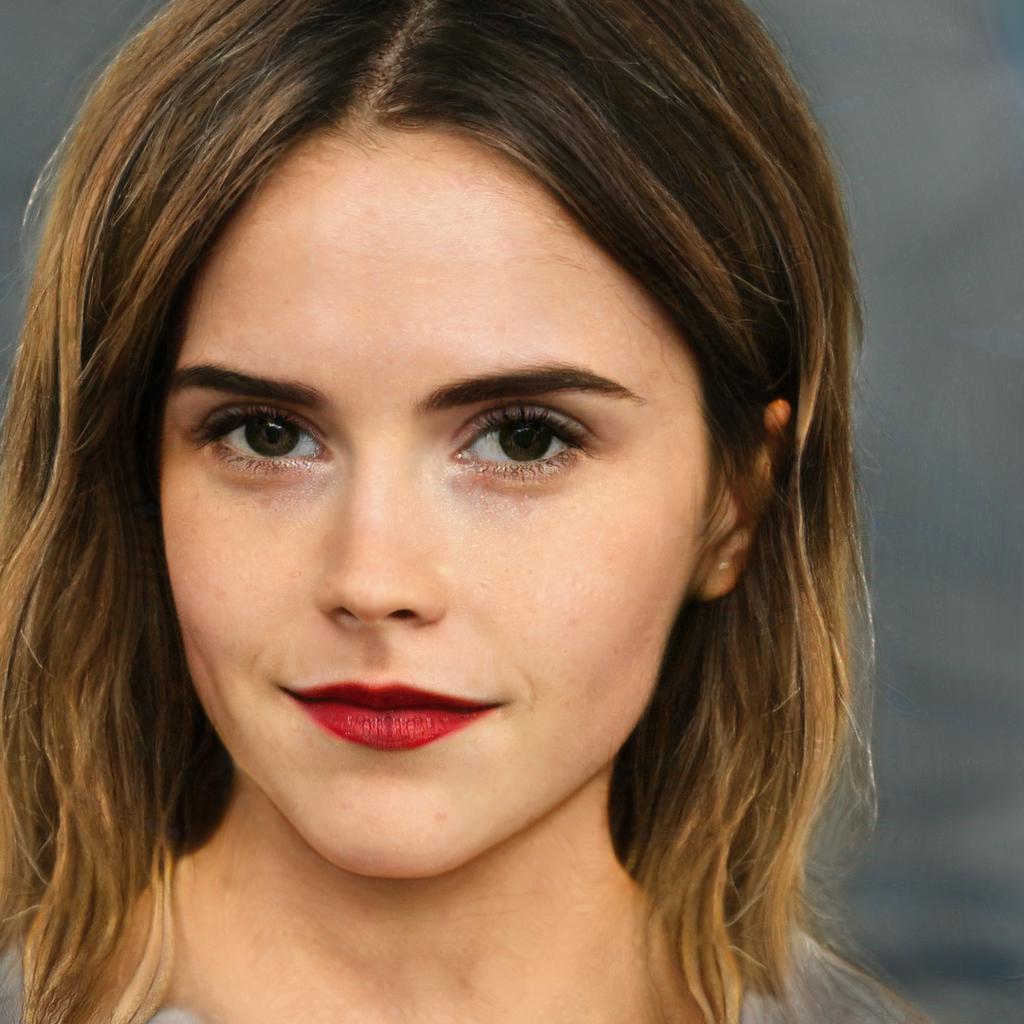} & 
        \includegraphics[width=0.195\columnwidth]{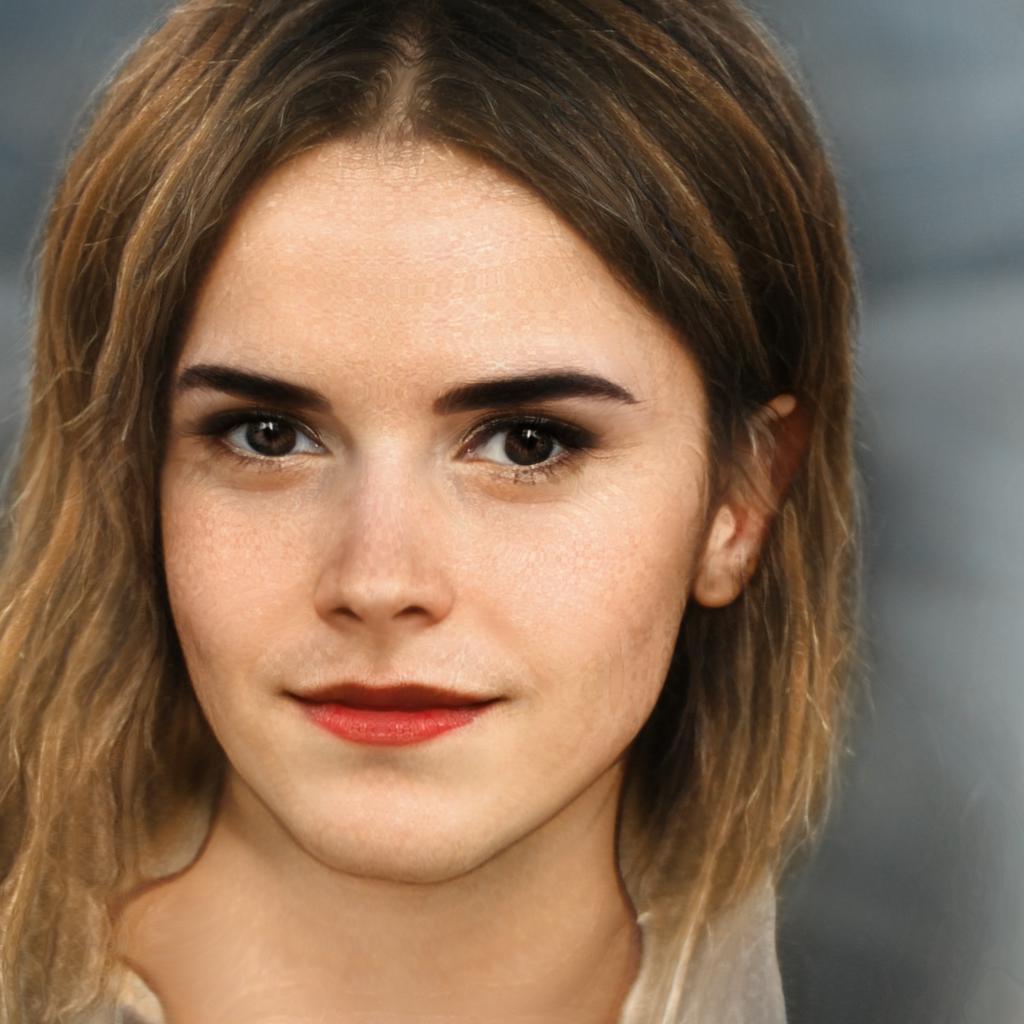} \\
        
        \includegraphics[width=0.195\columnwidth]{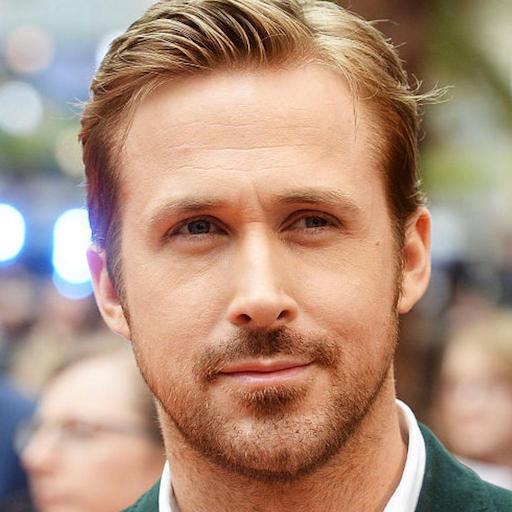} & 
        \includegraphics[width=0.195\columnwidth]{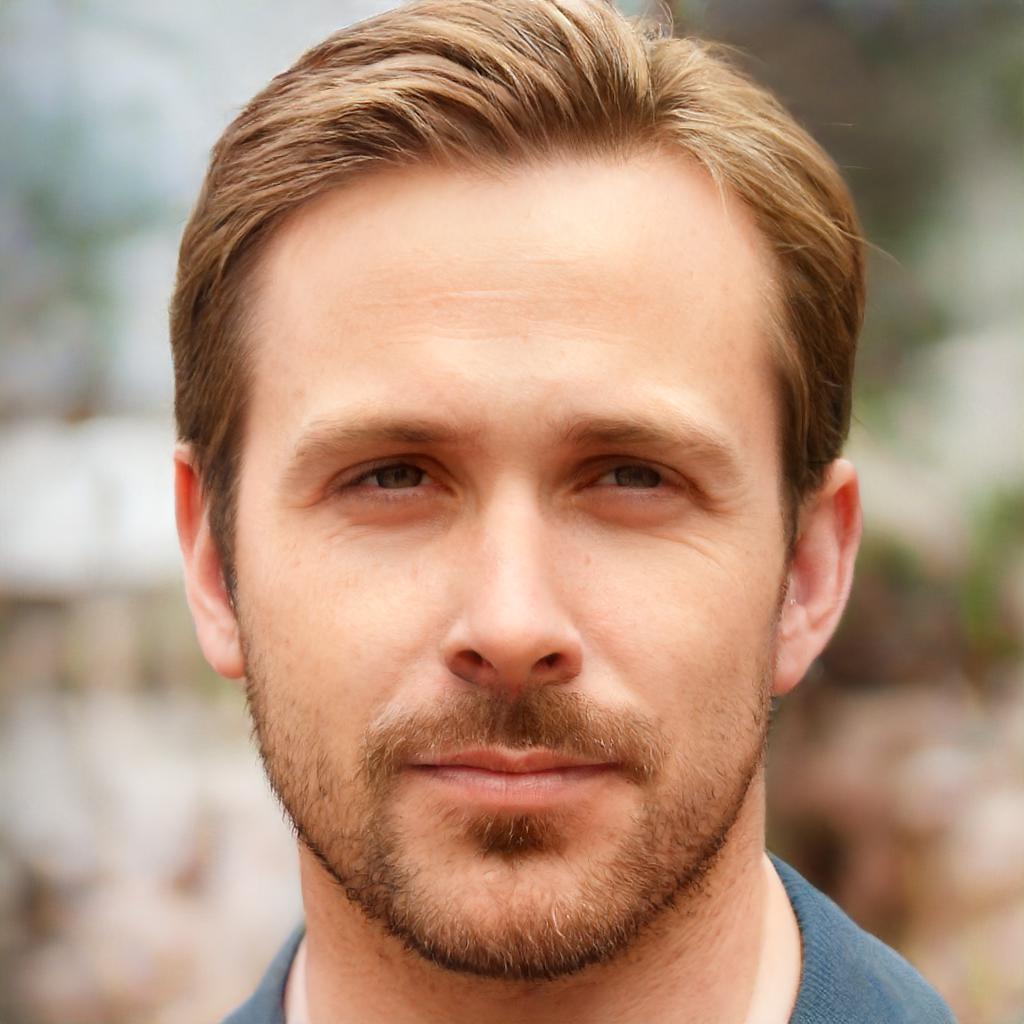} & 
        \includegraphics[width=0.195\columnwidth]{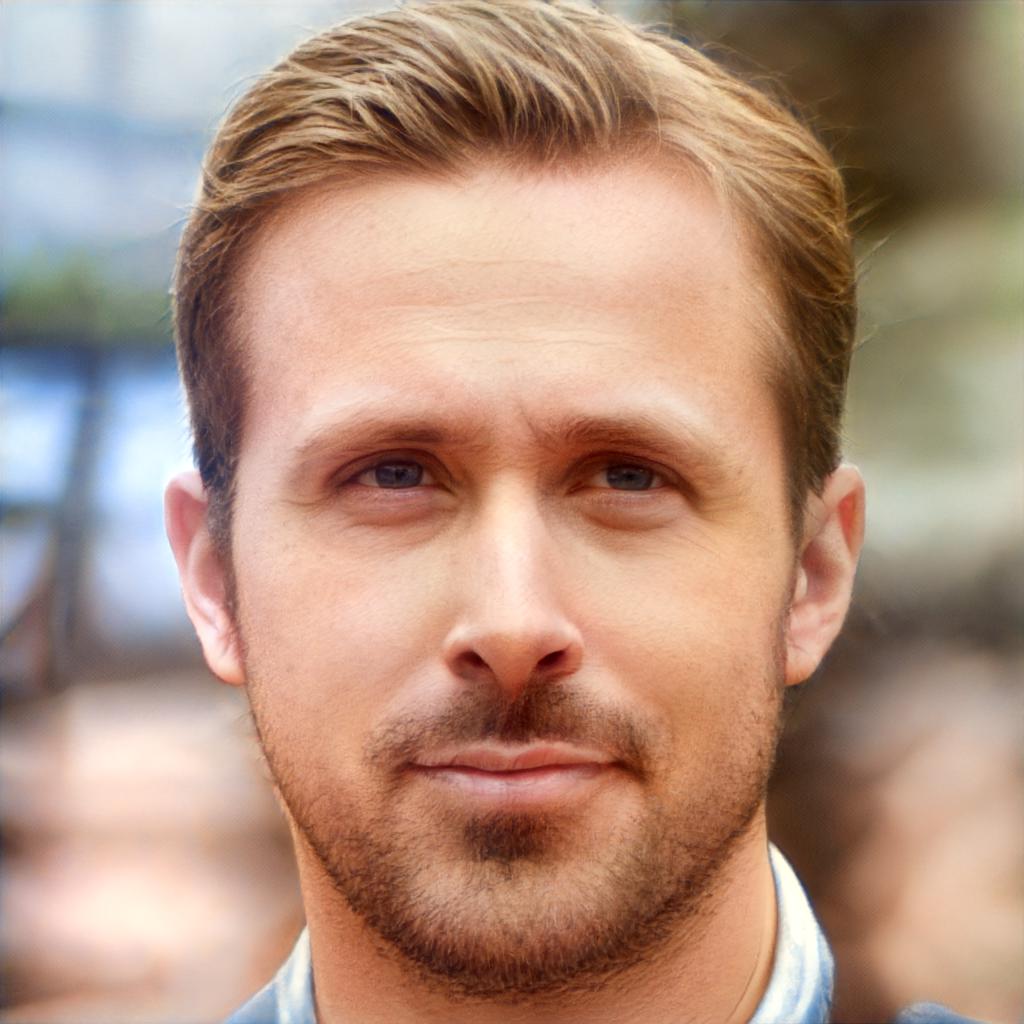} & 
        \includegraphics[width=0.195\columnwidth]{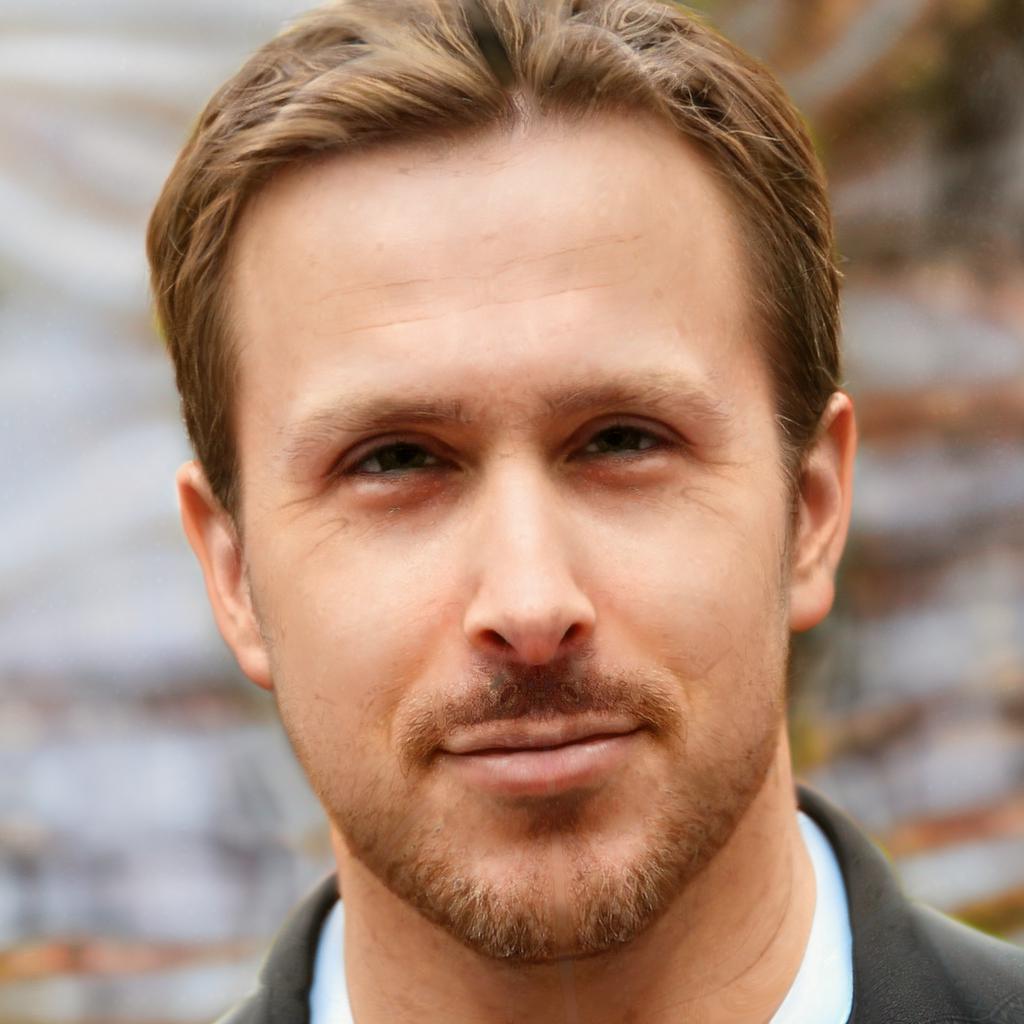} & 
        \includegraphics[width=0.195\columnwidth]{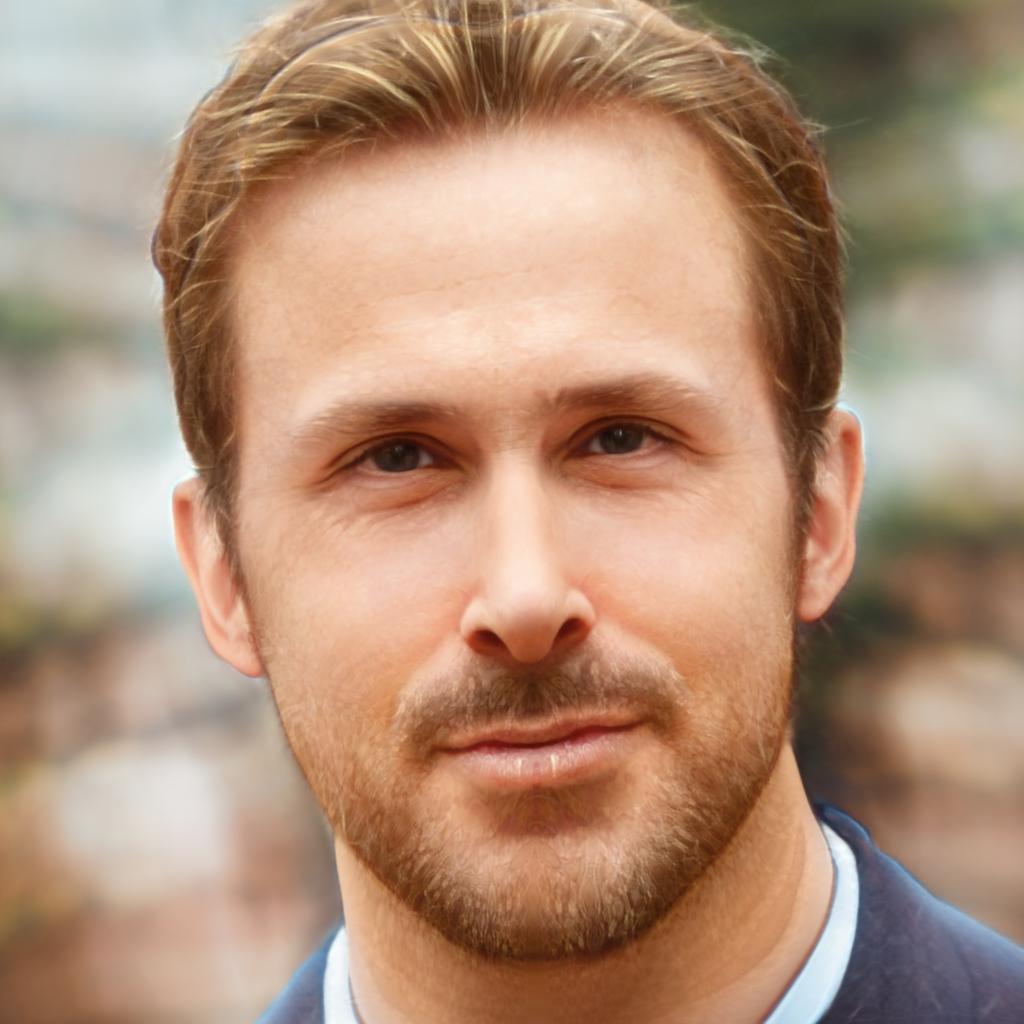} \\

        \begin{tabular}{c@{}c@{}}Source \\ Image\end{tabular} &
        \begin{tabular}{c@{}c@{}}StyleGAN2 \\ $\text{ReStyle}_{e4e}$\end{tabular} &
        \begin{tabular}{c@{}c@{}}StyleGAN2 \\ $\text{ReStyle}_{pSp}$\end{tabular} &
        \begin{tabular}{c@{}c@{}}StyleGAN3 \\ $\text{ReStyle}_{e4e}$\end{tabular} &
        \begin{tabular}{c@{}c@{}}StyleGAN3 \\ $\text{ReStyle}_{pSp}$\end{tabular} \\
	
        \end{tabular}
	}
	
        \vspace{-0.15cm}
	\caption{
	Reconstruction quality comparison between encoders trained for inverting StyleGAN2 and StyleGAN3 generators. 
	}
	\vspace{-0.35cm}
	\label{fig:inversions}
\end{figure}

%% file: tables/inversion.tex
\begin{table}
    \small
    \setlength{\tabcolsep}{2.65pt}
    \centering
    { \small
    \begin{tabular}{l@{\,\, }| c c c c c}
        \toprule
        Method & $\uparrow$ ID & $\uparrow$ MS-SSIM & $\downarrow$ LPIPS & $\downarrow$ $L_2$ & Time (s) \\
        \midrule
        SG2 $\text{ReStyle}_{pSp}$ & 
        \multicolumn{1}{c}{$0.66$} & 
        \multicolumn{1}{c}{$0.79$} &
        \multicolumn{1}{c}{$0.13$} & 
        \multicolumn{1}{c}{$0.03$} &
        \multicolumn{1}{c}{$0.37$} \\
        SG2 $\text{ReStyle}_{e4e}$ & 
        \multicolumn{1}{c}{$0.52$} & 
        \multicolumn{1}{c}{$0.74$} &
        \multicolumn{1}{c}{$0.19$} & 
        \multicolumn{1}{c}{$0.04$} &
        \multicolumn{1}{c}{$0.37$} \\
        \midrule
        SG3 $\text{ReStyle}_{pSp}$& 
        \multicolumn{1}{c}{$0.60$} & 
        \multicolumn{1}{c}{$0.77$} &
        \multicolumn{1}{c}{$0.17$} & 
        \multicolumn{1}{c}{$0.03$} &
        \multicolumn{1}{c}{$0.52$} \\
        SG3 $\text{ReStyle}_{e4e}$ & 
        \multicolumn{1}{c}{$0.49$} & 
        \multicolumn{1}{c}{$0.70$} &
        \multicolumn{1}{c}{$0.22$} & 
        \multicolumn{1}{c}{$0.06$} &
        \multicolumn{1}{c}{$0.52$} \\
        \midrule

    \end{tabular}
    }
    \vspace{-0.3cm}
    \caption{Quantitative reconstruction results on the human facial domain measured over the CelebA-HQ~\cite{karras2017progressive,liu2015deep} test set. For StyleGAN3, we use a generator trained on aligned images.} 
    \label{tb:quantitative_inversion}
    \vspace{-0.4cm}
\end{table}

%% file: figures/editing_real_images.tex
\begin{figure}[tb]
	\centering
	\setlength{\tabcolsep}{1pt}	
	{\footnotesize
	\begin{tabular}{c c c c c c c}

		\raisebox{0.0in}{\rotatebox{90}{$+$ Brown Hair}} &
        \includegraphics[width=0.185\columnwidth]{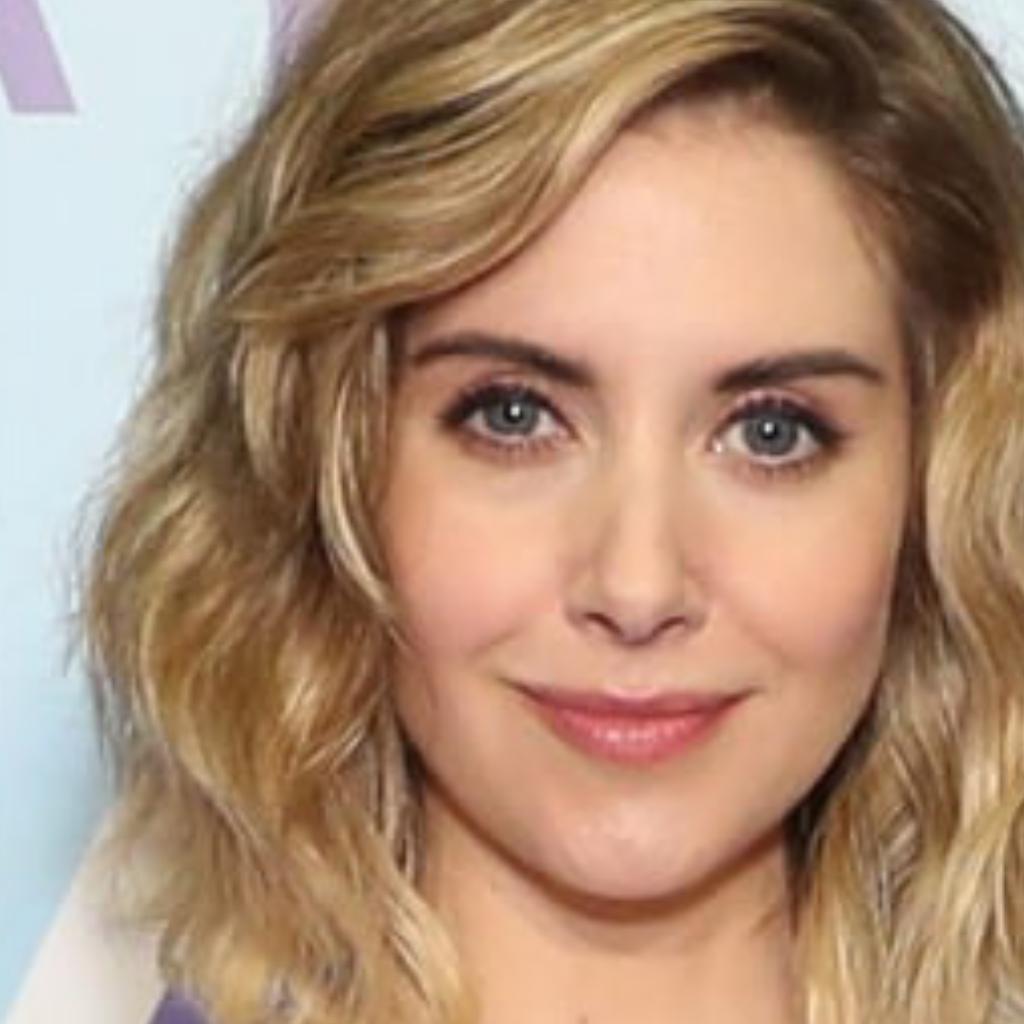} & 
        \includegraphics[width=0.185\columnwidth]{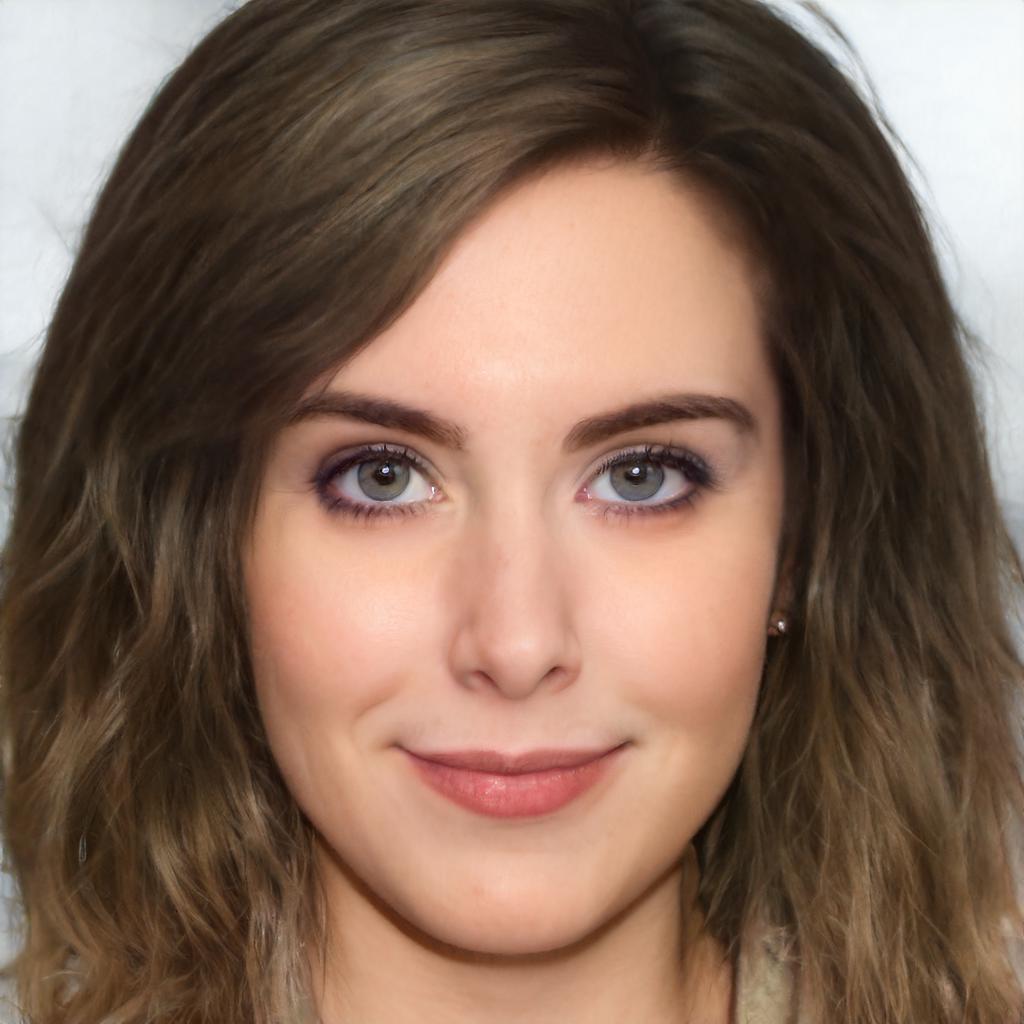} & 
        \includegraphics[width=0.185\columnwidth]{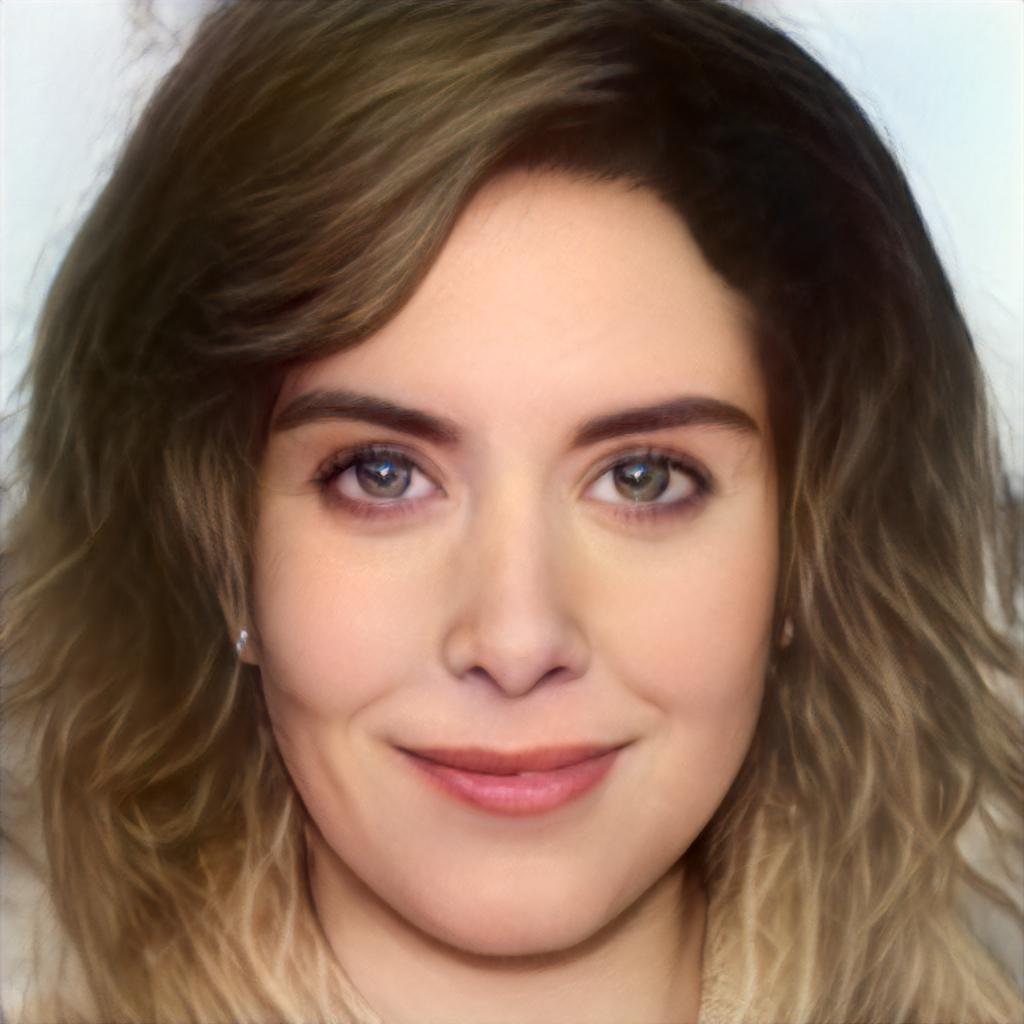} & 
        \includegraphics[width=0.185\columnwidth]{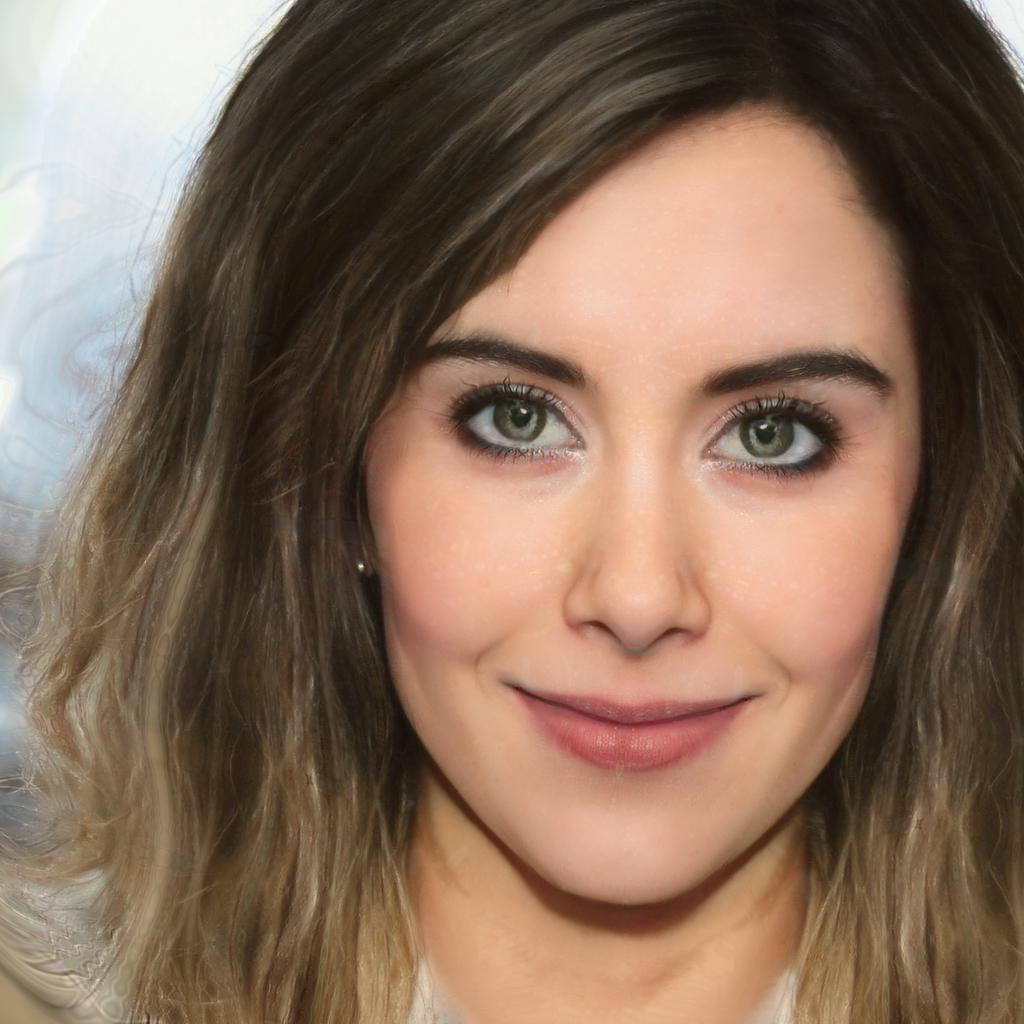} & 
        \includegraphics[width=0.185\columnwidth]{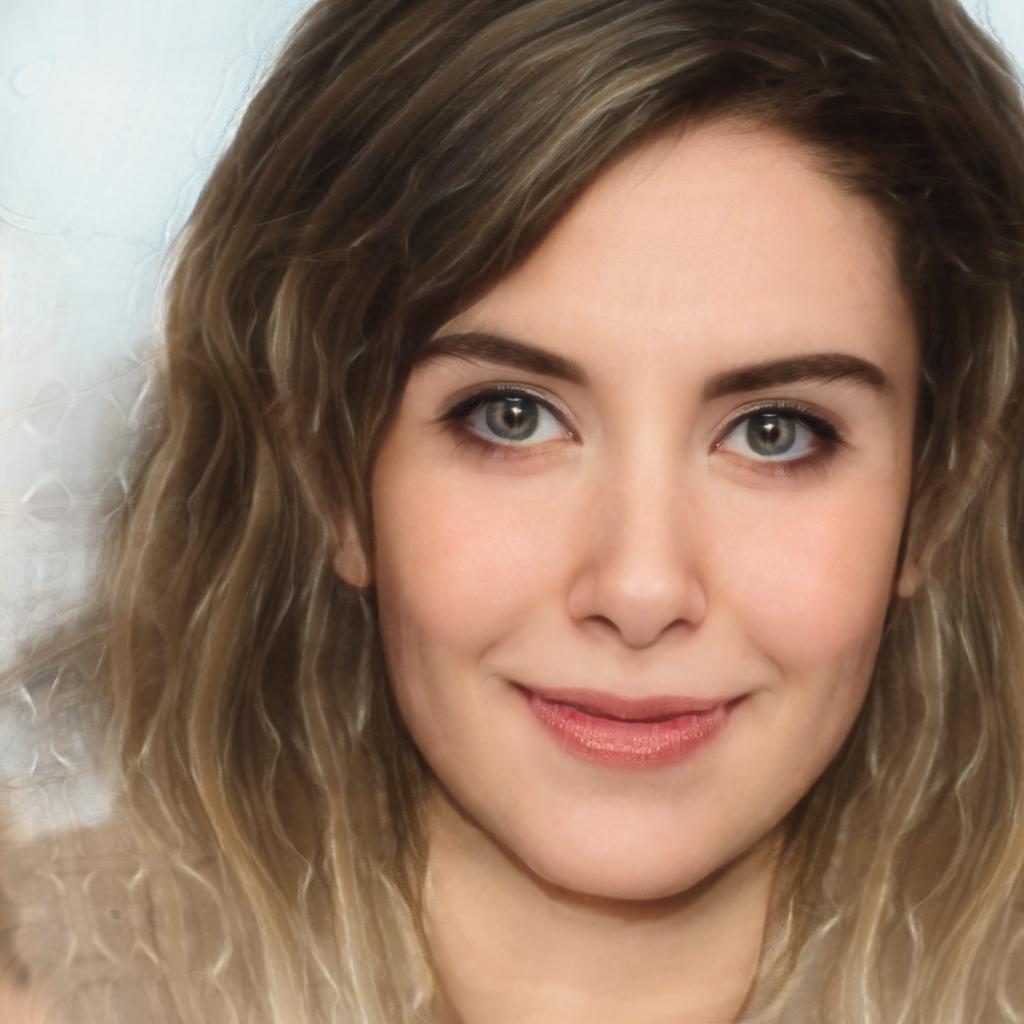} \\

		\raisebox{0.1in}{\rotatebox{90}{$-$ Age}} &
        \includegraphics[width=0.185\columnwidth]{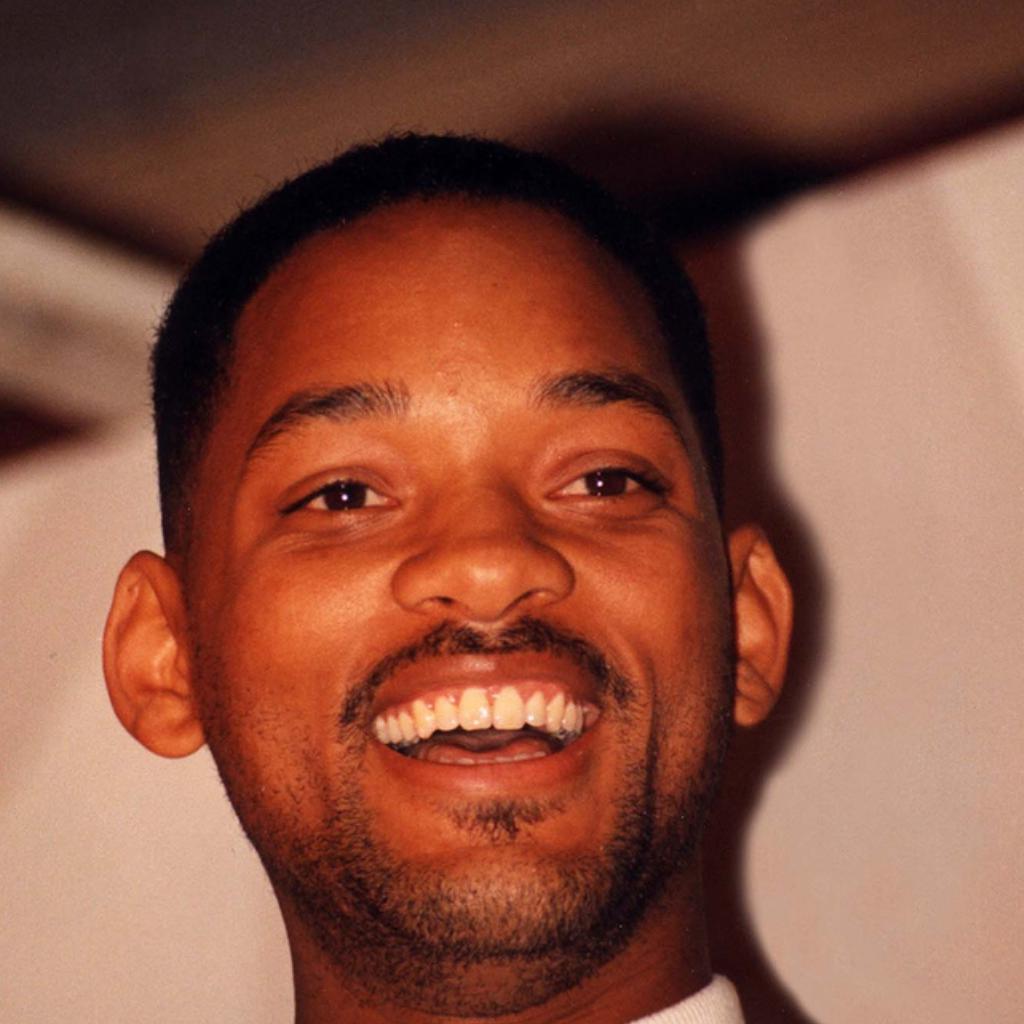} & 
        \includegraphics[width=0.185\columnwidth]{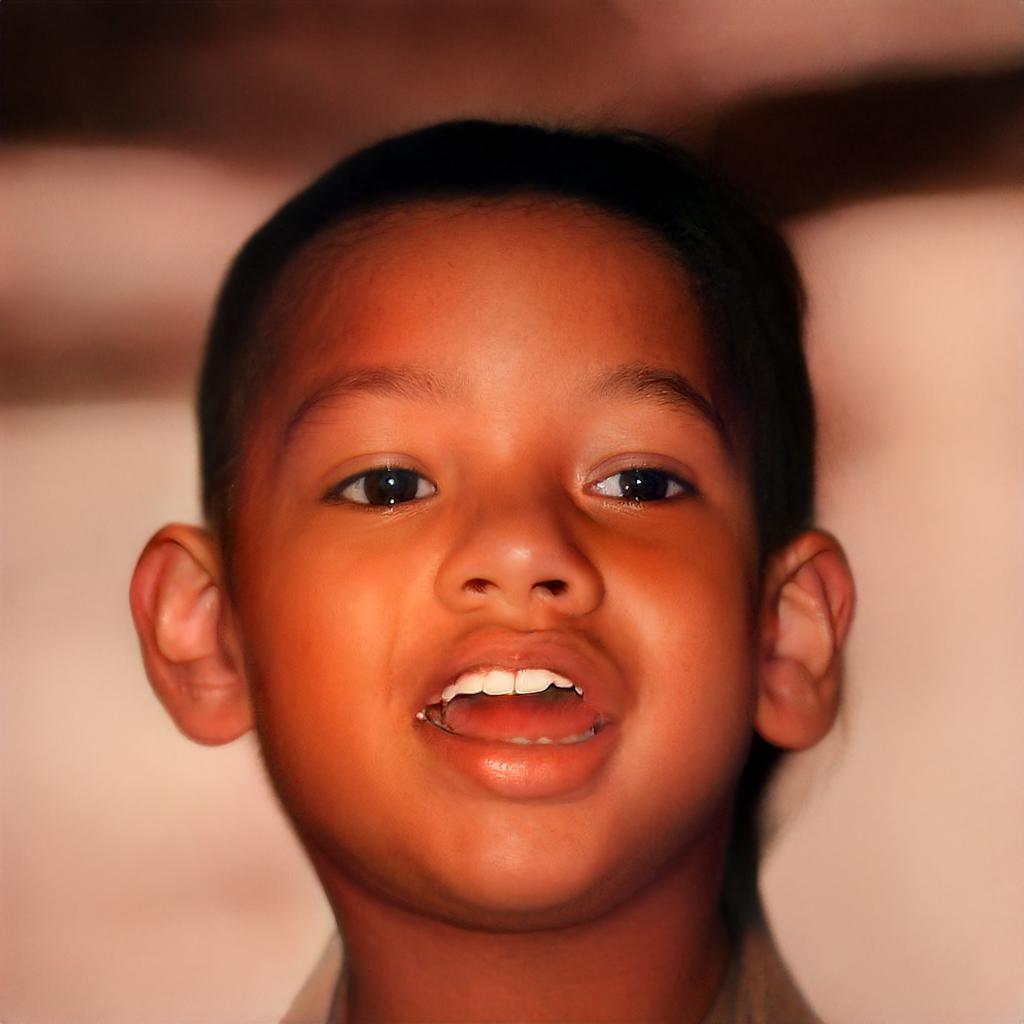} & 
        \includegraphics[width=0.185\columnwidth]{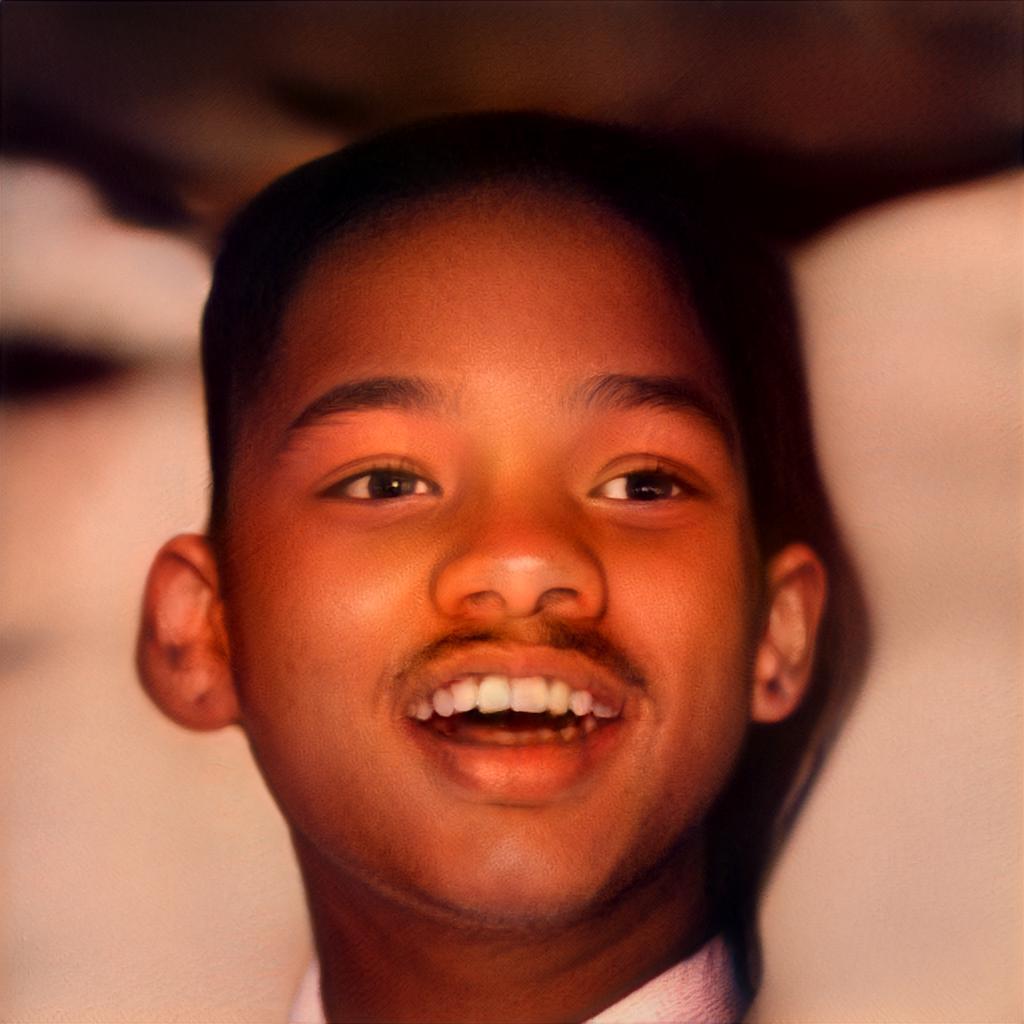} & 
        \includegraphics[width=0.185\columnwidth]{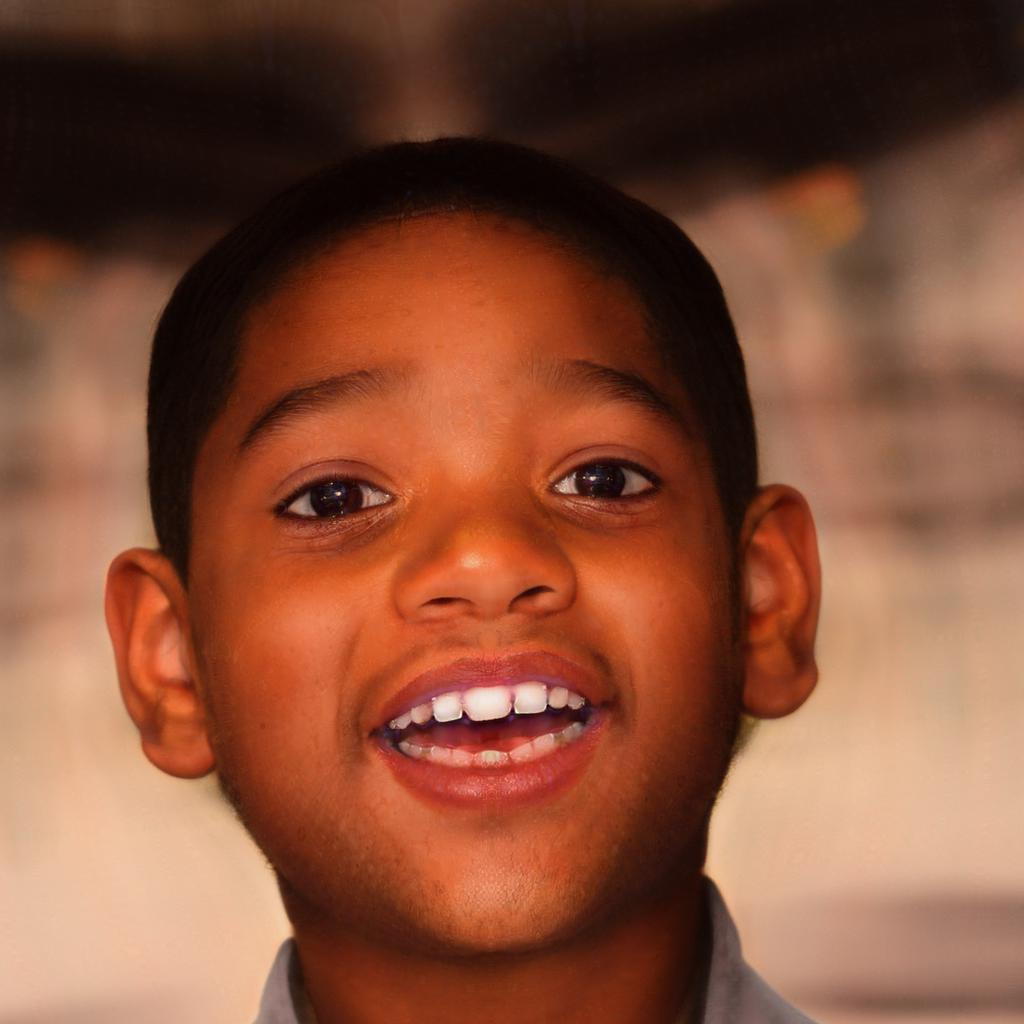} & 
        \includegraphics[width=0.185\columnwidth]{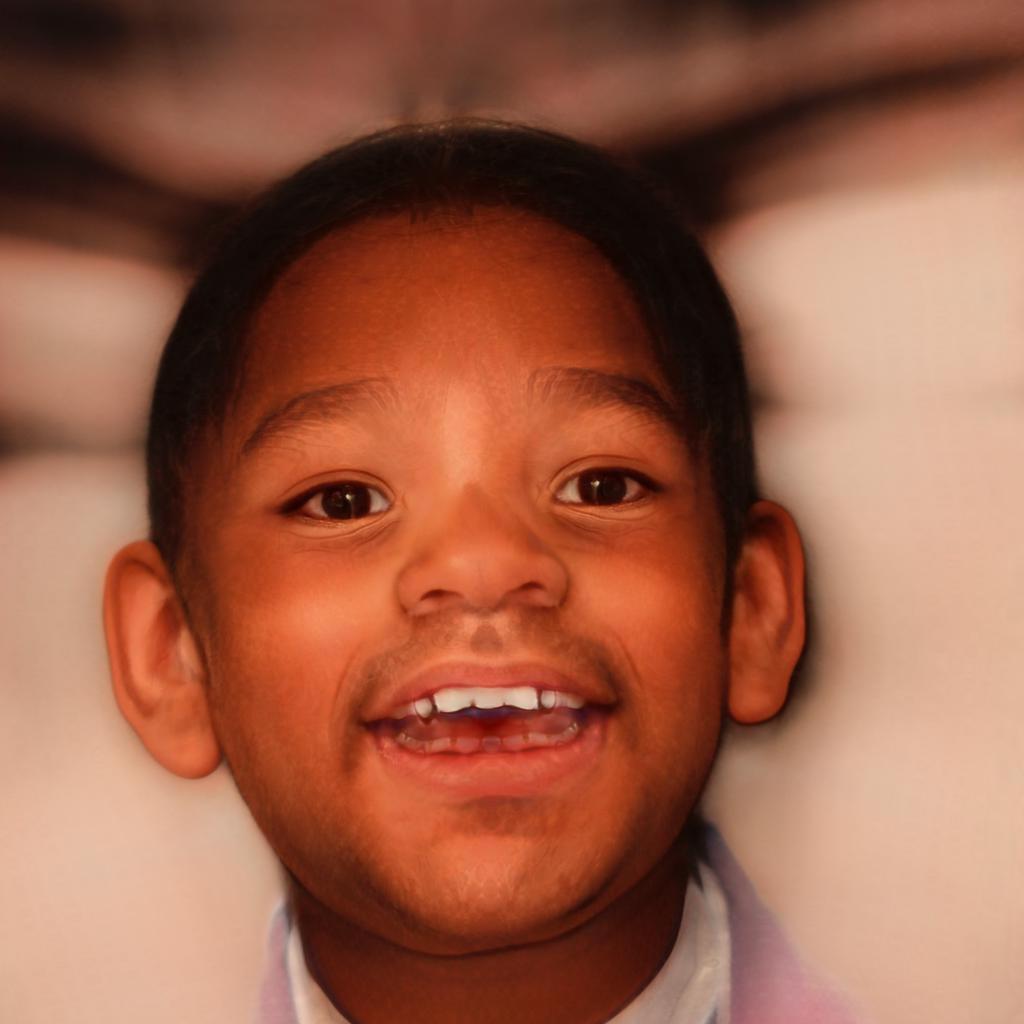} \\

		\raisebox{0.075in}{\rotatebox{90}{$-$ Smile}} &
        \includegraphics[width=0.185\columnwidth]{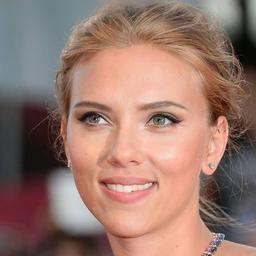} & 
        \includegraphics[width=0.185\columnwidth]{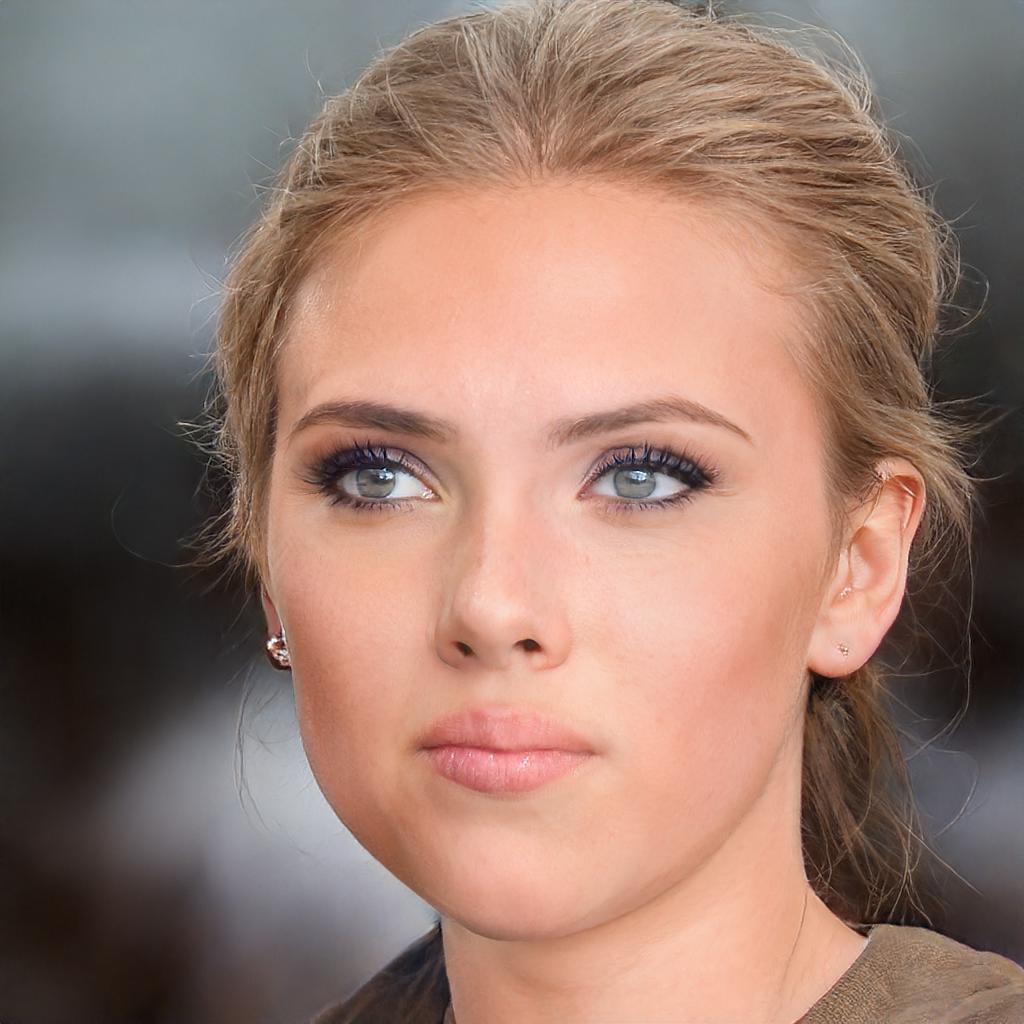} & 
        \includegraphics[width=0.185\columnwidth]{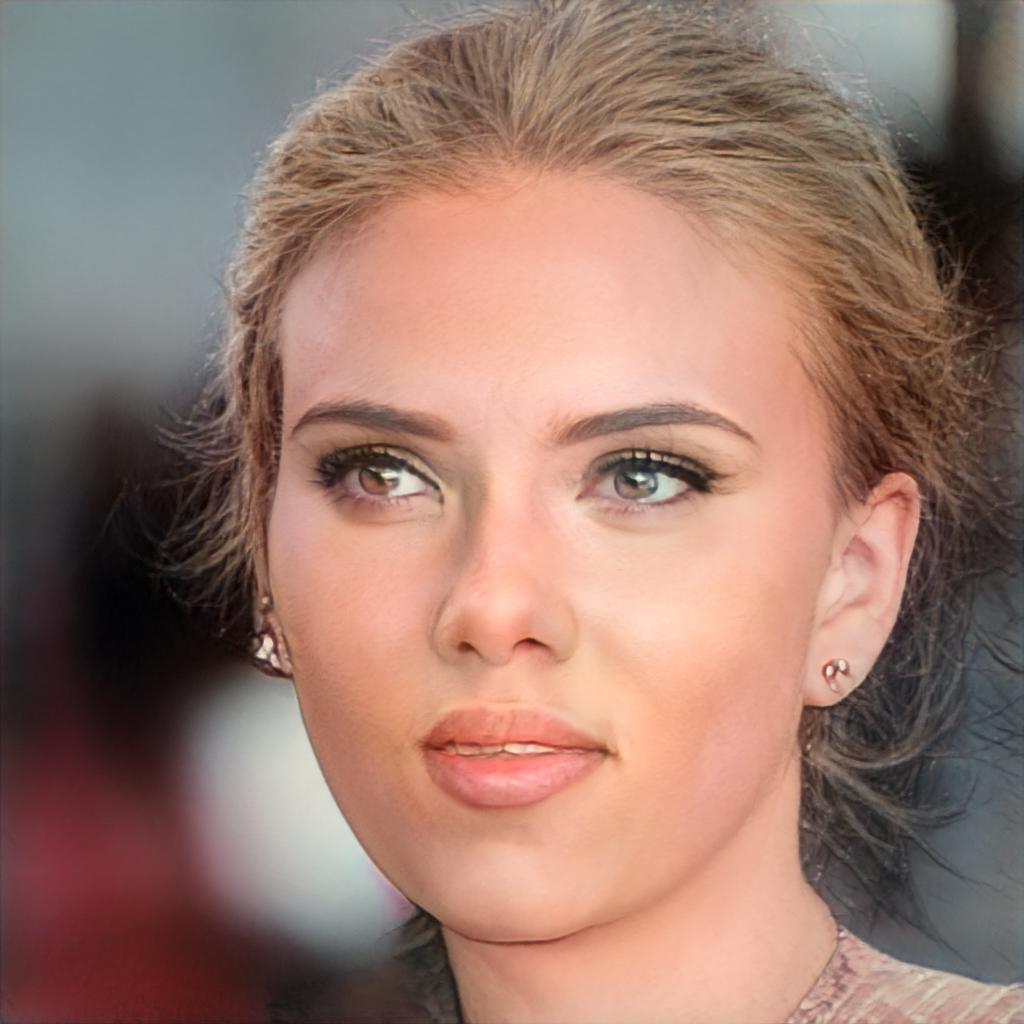} & 
        \includegraphics[width=0.185\columnwidth]{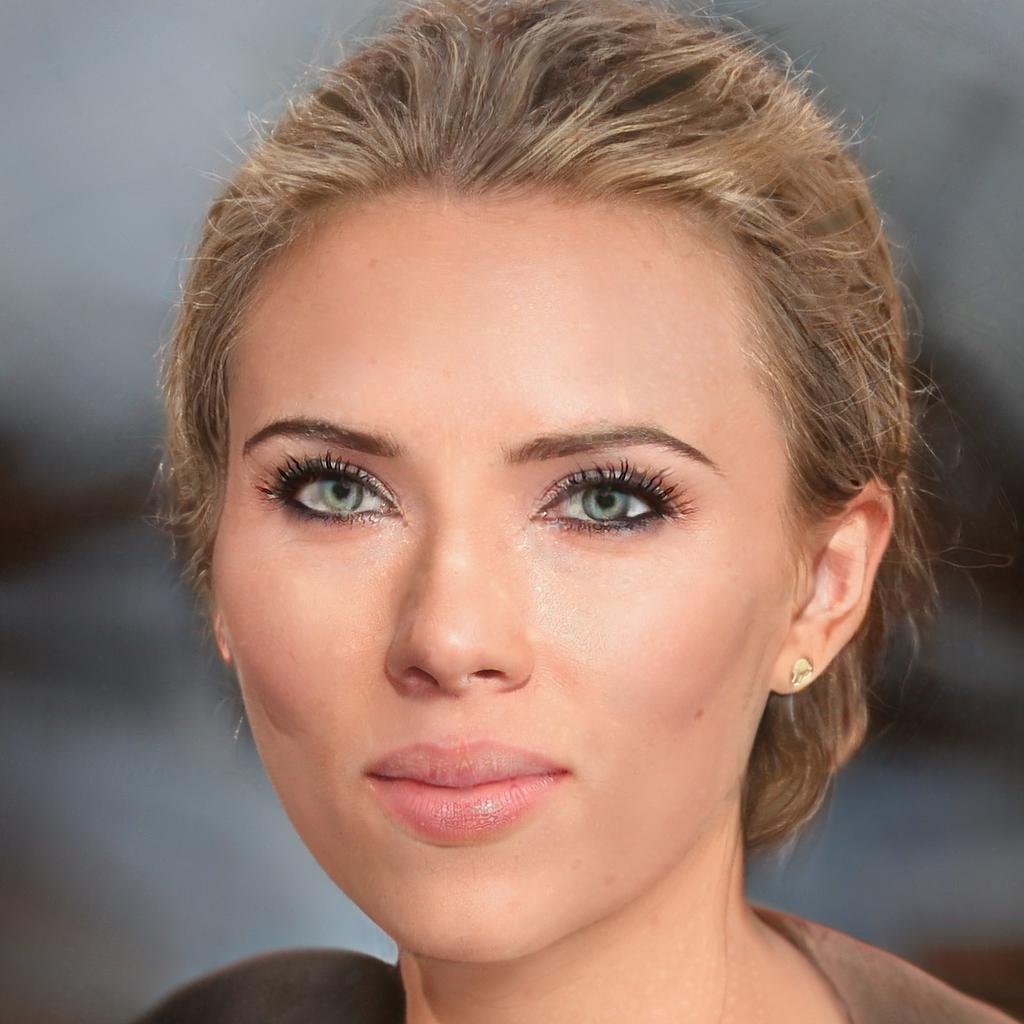} & 
        \includegraphics[width=0.185\columnwidth]{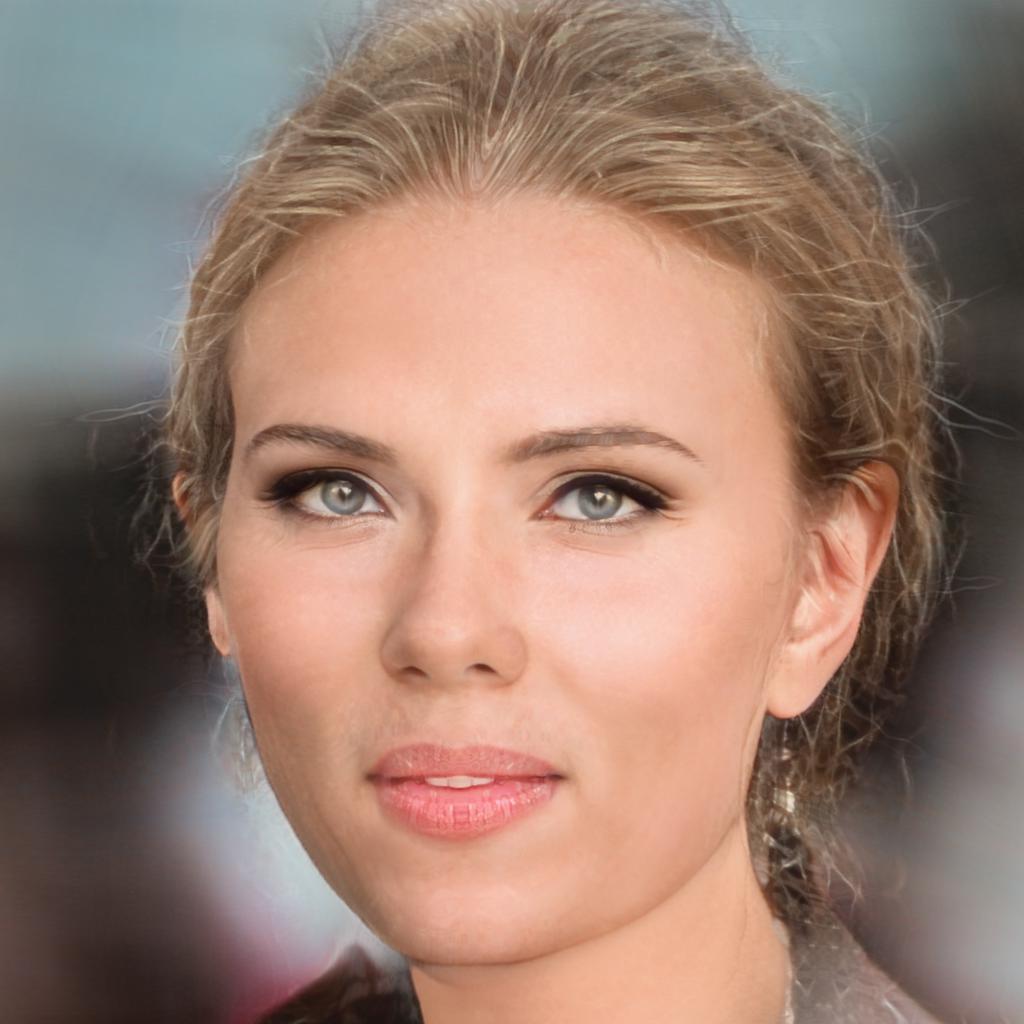} \\
        
		\raisebox{0.0in}{\rotatebox{90}{$+$ Short Hair}} &
        \includegraphics[width=0.185\columnwidth]{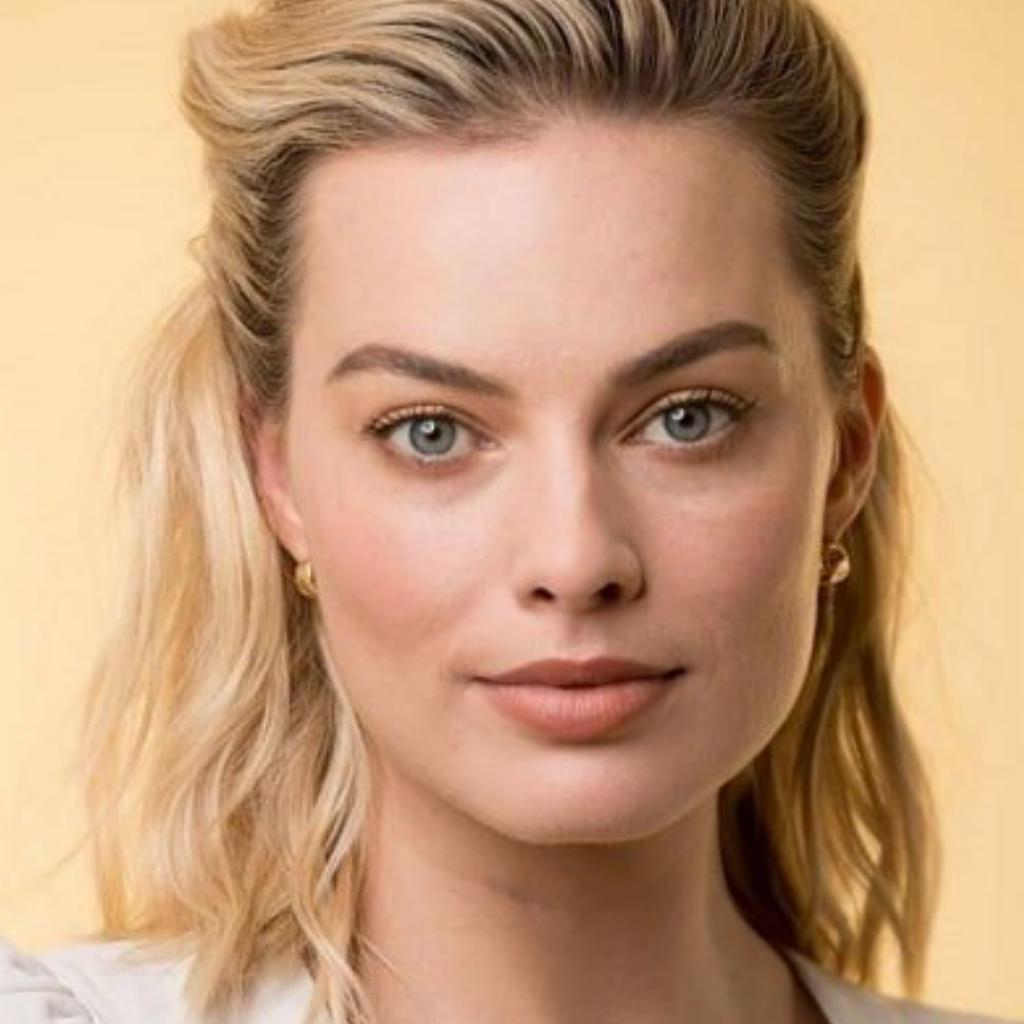} & 
        \includegraphics[width=0.185\columnwidth]{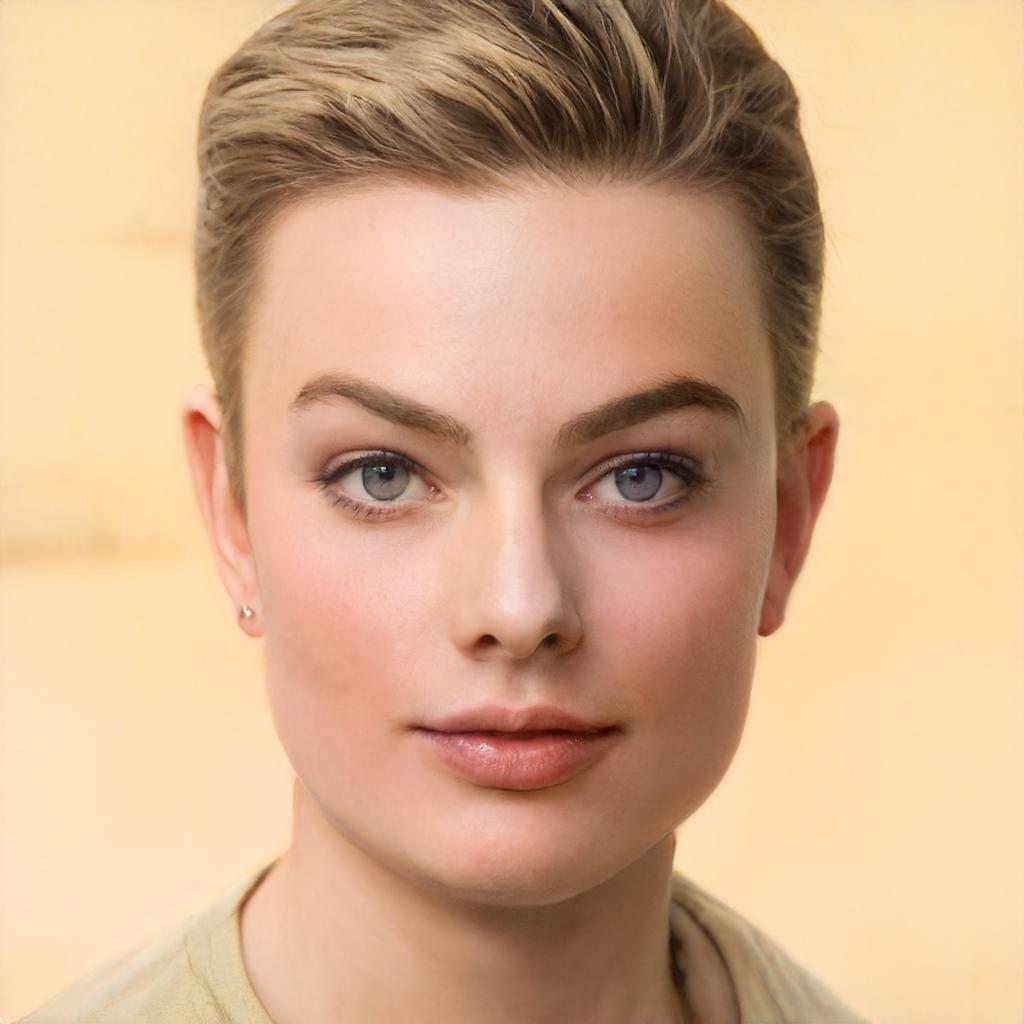} & 
        \includegraphics[width=0.185\columnwidth]{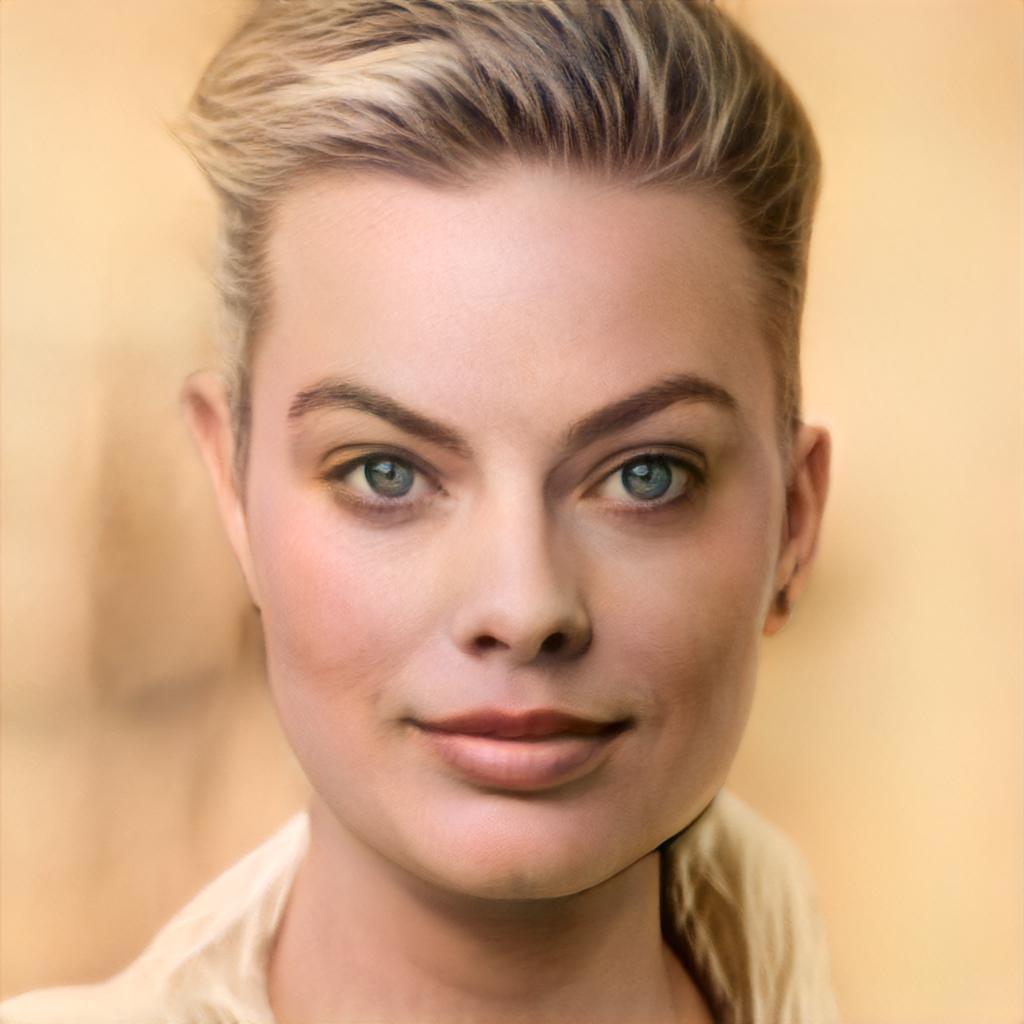} & 
        \includegraphics[width=0.185\columnwidth]{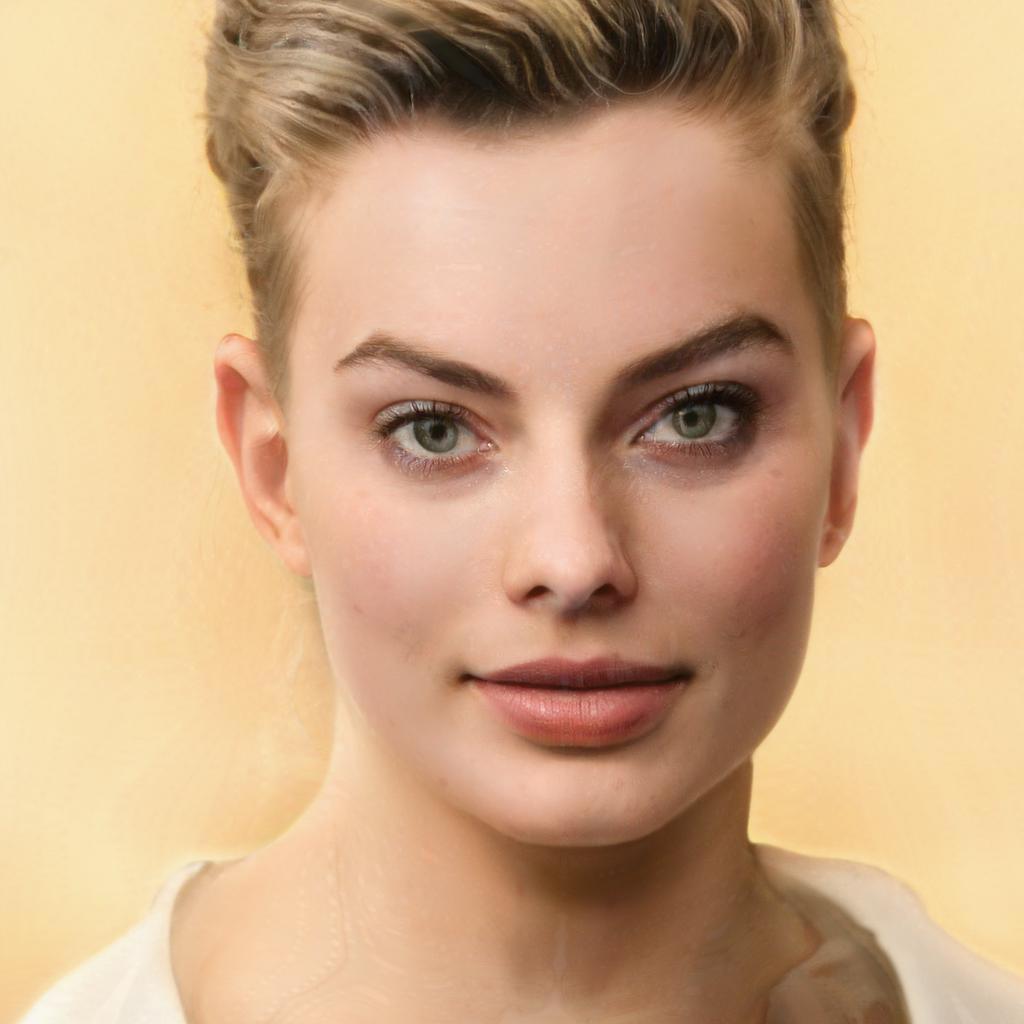} & 
        \includegraphics[width=0.185\columnwidth]{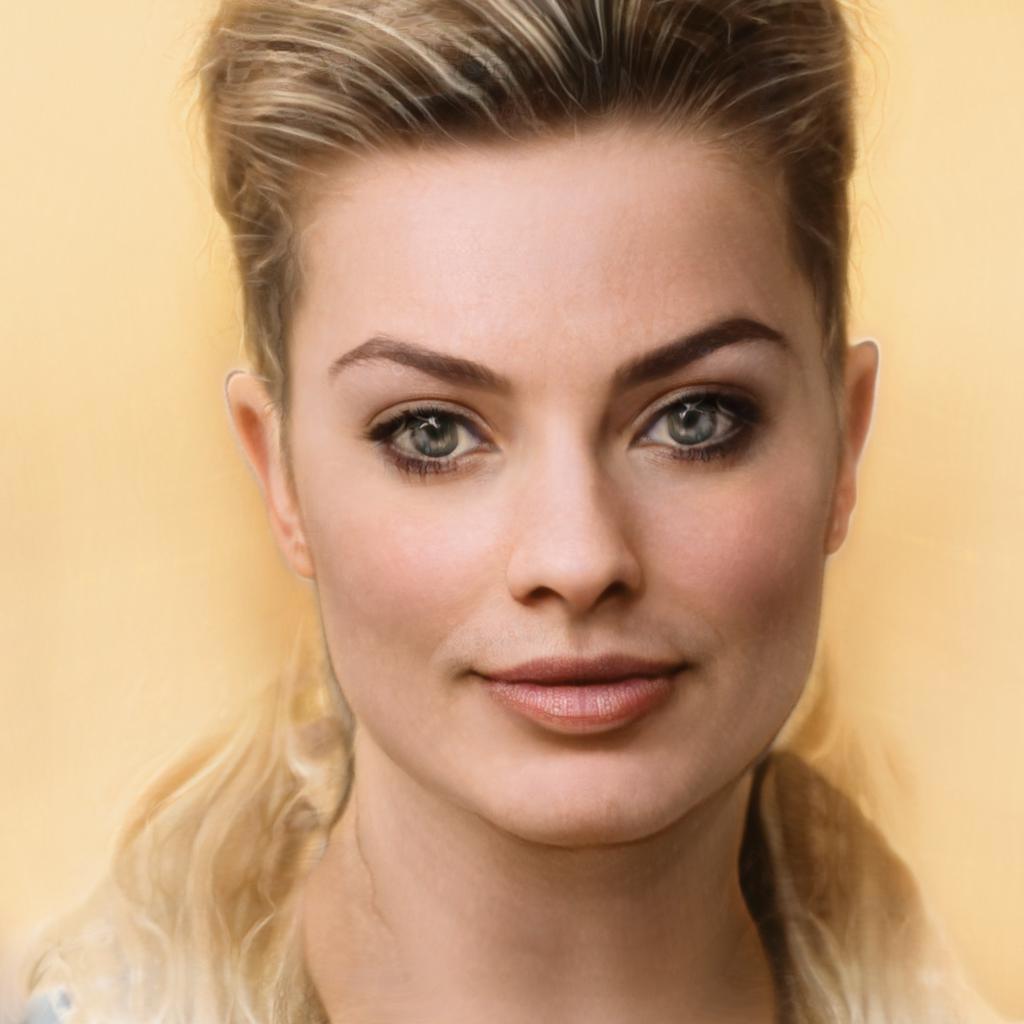} \\

		&
        \begin{tabular}{c@{}c@{}}Source \\ Image\end{tabular} &
        \begin{tabular}{c@{}c@{}}StyleGAN2 \\ $\text{ReStyle}_{e4e}$\end{tabular} &
        \begin{tabular}{c@{}c@{}}StyleGAN2 \\ $\text{ReStyle}_{pSp}$\end{tabular} &
        \begin{tabular}{c@{}c@{}}StyleGAN3 \\ $\text{ReStyle}_{e4e}$\end{tabular} &
        \begin{tabular}{c@{}c@{}}StyleGAN3 \\ $\text{ReStyle}_{pSp}$\end{tabular} \\
		
	\end{tabular}
	}
	\vspace{-0.1cm}
	\caption{
	Editing comparison. We perform various edits~\cite{patashnik2021styleclip,shen2020interpreting} over latent codes obtained by each inversion method.
	}
	\vspace{-0.2cm}
	\label{fig:editing_reals}
\end{figure}

%% file: video.tex
\input{figures/video_overview}

\vspace{0.5cm}
\section{Inverting and Editing Videos}~\label{sec:video}
We now extend our inversion method to encoding and editing videos. This extension introduces two central challenges. First, the reconstructed and edited video frames should be \textit{temporally consistent}, which may be difficult to attain when inverting each frame independently. Second, individual frames of a human face video often feature more challenging facial expressions than those found in still image training sets, (e.g., closed eyes or a mouth open mid-speech).
These challenges must be addressed, regardless of the architecture. Using StyleGAN3 is appealing because it reduces texture sticking and inherently handles varying face positions and rotations.
Additionally, as we demonstrate below, StyleGAN3 may be leveraged to increase the field of view, resulting in wide-view reconstructed and edited videos, rather than close-ups of an individual.
This even enables the faithful edit of attributes that partially ``spill out'' of the input frame. 
Below, we describe our end-to-end video encoding and editing pipeline, summarized in \cref{fig:video_overview}.

\vspace{-0.1cm}
\paragraph{\textbf{Video Preprocessing. }}
Given an input video, we begin by cropping each video frame to be compatible with the input head size expected by StyleGAN3. 
For achieving a stable video that looks as if it was captured from a non-moving camera, we crop a fixed bounding box across all video frames (as illustrated by the red bounding box in \cref{fig:video_overview}). We denote the resulting cropped images by $\{x_{i,\textit{unaligned}}\}_{i=1}^N$. 
As with still images, to invert each frame using our trained encoder, we additionally align each frame (green box in \cref{fig:video_overview}), yielding images $\{x_{i,\textit{aligned}}\}_{i=1}^N$. We also compute the transformation $(r_i, t_{x,i}, t_{y,i})$ between each $(x_{i,\textit{aligned}},x_{i,\textit{unaligned}})$ pair. 

\vspace{-0.05cm}
\paragraph{\textbf{Initial Video Encoding. }}
We use our trained $\text{ReStyle}_{e4e}$ encoder $E$ to obtain the initial frame inversions $w_i = E(x_{i,\textit{aligned}})$, whose unaligned reconstructions are given by,
\begin{equation*}
    y_i = G(w_i; (r_i, t_{x,i}, t_{y,i})).
\end{equation*}
We can additionally apply some manipulation $f$ to obtain an edited version of the input frame:
\begin{equation*}
    y_{i,\textit{edit}} = G(f(w_{i}); (r_i, t_{x,i}, t_{y,i})).
\end{equation*}

\vspace{-0.05cm}
\paragraph{\textbf{Latent Vector Smoothing. }}
Inverting each frame independently may result in inconsistencies between successive reconstructed frames. This may be caused by the pre-processing alignment, the encoder network itself, or by the manipulation applied on the inverted latent codes. To mitigate temporal discontinuities, we temporally smooth the inverted, edited latent codes $f(\!w_i)$ and the predicted transformation matrix $T_i$ applied on the Fourier features and derived by $(r_i, t_{x,i}, t_{y,i})$, using a weighted moving average: 
\begin{align*}
    w_{i,\textit{smooth}} &= \textstyle \sum_{j=i-2}^{i+2} \mu_j f(w_j) \\
    T_{i,\textit{smooth}} &= \textstyle \sum_{j=i-2}^{i+2} \mu_j T_j
\end{align*}
where 
\begin{equation*}
    [\mu_{i-2}, \mu_{i-1}, \mu_{i}, \mu_{i+1}, \mu_{i+2}] = \frac{1}{3}[0.25, 0.75, 1, 0.75, 0.5].
\end{equation*}
We find that this smoothing operation improves temporal coherence without harming the reconstruction quality. 

\vspace{-0.05cm}
\paragraph{\textbf{Pivotal Tuning for Improved Reconstructions. }}
To further improve the frame reconstructions, we adopt the pivotal tuning inversion (PTI) method~\cite{roich2021pivotal}.
Specifically, the initial inversions are used for fine-tuning the weights of the StyleGAN3 generator to achieve better reconstructions of the input frames~\cite{tzaban2022stitch}.
Note, while the encoder network is trained to reconstruct aligned images, we perform the PTI fine-tuning using the original \textit{unaligned} images. That is, when performing PTI, losses are computed between the $x_{i,\textit{unaligned}}$ images and their refined reconstructions given by:
\begin{equation*}
    y_{i,\textrm{PTI}} = G_{\textrm{PTI}}(w_{i}; (r_i, t_{x,i}, t_{y,i})),
\end{equation*}
where $G_{\textrm{PTI}}$ is the PTI-modified generator. 

\input{figures/video_reconstructions_and_edits}

\vspace{-0.05cm}
\paragraph{\textbf{Bringing It All Together. }}
Having obtained the smoothed edited latent codes, their corresponding smoothed transformations, and the fine-tuned generator, we can now generate the unified edited video. Formally, the final edited $i$-th frame is given by: 
\begin{equation}
    y_{i,\textit{final}} = G_{\textrm{PTI}}(w_{i,\textit{smooth}}; T_{i,\textit{smooth}}).
\end{equation}
We provide reconstruction and editing results in \cref{fig:video_results} and in \cref{sec:additional_results}. In addition, by training StyleGAN-NADA~\cite{gal2021stylegannada} on $G_{\textrm{PTI}}$ for a given video and text prompt, we can generate edited videos in various styles (e.g., a cartoon video of Obama). We refer the reader to \cref{sec:video_details,sec:additional_results} for additional details and results.

\input{figures/wide_dwight}

\paragraph{\textbf{Expanding the Field of View. }}
We now describe how we can expand the field of view (FOV) of the video reconstruction. Denote by $\Delta$ the desired expansion (illustrated in the original frame of \cref{fig:video_overview}). 
To expand the FOV, we construct a transformation matrix $T_{\Delta}$. For example, for a vertical expansion of the frame, we define a transform matrix corresponding to a vertical shift derived from the parameters $(0,0,\Delta)$. 
For each input frame, we then generate two images: 
\begin{align*}
    y &= G_{\textrm{PTI}}(w_{i,smooth};T_{i,smooth}) \\
    y_{\textit{shift}} &= G_{\textrm{PTI}}(w_{i,smooth};T_{\Delta} \cdot T_{i,smooth}). 
\end{align*}    
Finally, we adjoin to $y$ the added non-overlapping parts from $y_{\textit{shift}}$, obtaining the wider output frame. Results of such an expansion are shown in \cref{fig:wide_results} where we demonstrate the ability to reconstruct the entirety of the individual's head.
Notice, images generated by StyleGAN2 are aligned, and as such, attributes that we wish to edit may overflow outside the frame boundary. In addition, since the images are aligned, we must project the edited frame back to the original context. Doing so, we may obtain a mismatch between regions within the generated image boundary (which were edited) and those outside the boundary (which were untouched). In StyleGAN3, however, this expansion allows editing the desired attribute in its entirety, resulting in a full, coherent edit. 

%% file: figures/video_overview.tex
\vspace{-0.1cm}
\begin{figure*}
    \centering
    \setlength{\belowcaptionskip}{-5pt}
    \includegraphics[width=\linewidth]{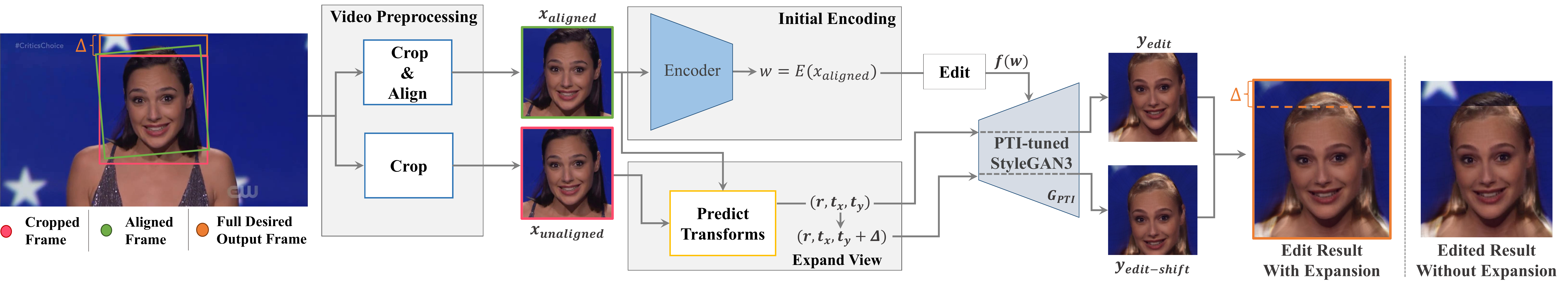}
    \vspace{-0.35cm}
    \caption{
    An overview of our video editing workflow. The target region containing the face is cropped (red frame) and aligned (green frame) using an off-the-shelf keypoints detector~\cite{dlib09}. Note that the top of the head is cropped out.
	The aligned frame is inverted by our encoder to a latent code $w = E(x_{\textit{aligned}})$, and undergoes some latent manipulation $f$ (e.g., hair color), yielding an edited code $f\!(\!w)$.
	In parallel, we compute the transformation $(r,t_x,t_y)$ between $x_{\textit{aligned}}$ and $x_{\textit{unaligned}}$. 
	Given $f\!(\!w)$ and $(r, t_x, t_y)$, the edited frame is generated by $y_{\textit{edit}} = G_{PTI}(f\!(\!w); (r, t_x, t_y))$.  
	To fix the over-cropping, we generate an additional \emph{shifted} image by translating the Fourier features by some value $\Delta$ (marked in orange). The two generated images are then merged into an extended frame comprising the entire edited head shown on the right.
	In the rightmost image, we show the edited result obtained \textit{without} expansion. Observe that the edited result is not uniform along the entirety of the hair region when pasted back to the original context.
    }
    \vspace{-0.1cm}
    \label{fig:video_overview}
\end{figure*}

%% file: figures/video_reconstructions_and_edits.tex
\begin{figure}[tb]
	\centering
	\setlength{\tabcolsep}{1pt}	
	{\footnotesize
	\begin{tabular}{c c c c c c c c}

		\raisebox{0.05in}{\rotatebox{90}{Original}} &
        \includegraphics[width=0.15\columnwidth]{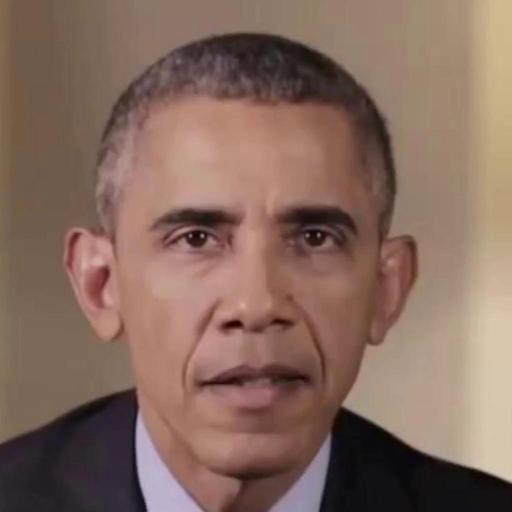} & 
        \includegraphics[width=0.15\columnwidth]{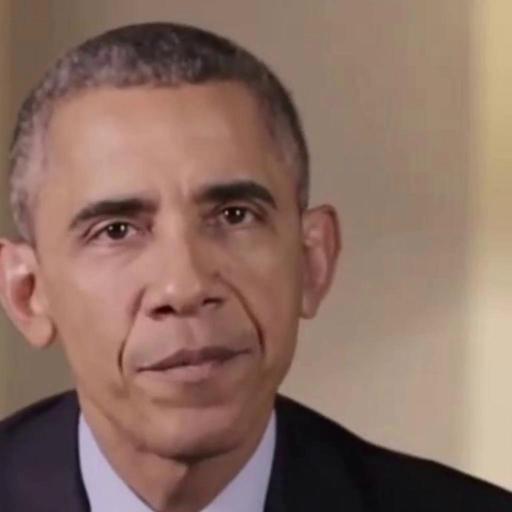} & 
        \includegraphics[width=0.15\columnwidth]{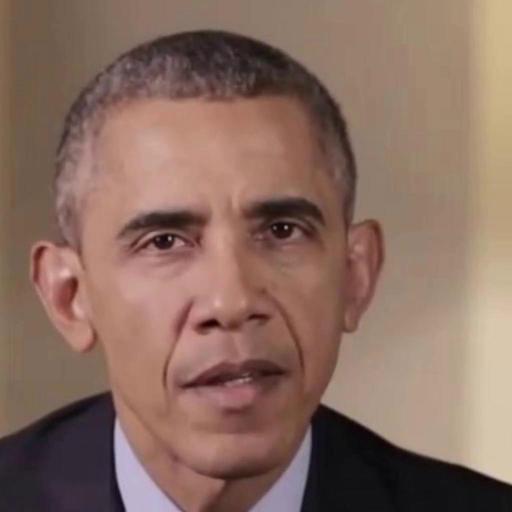} & 
        \includegraphics[width=0.15\columnwidth]{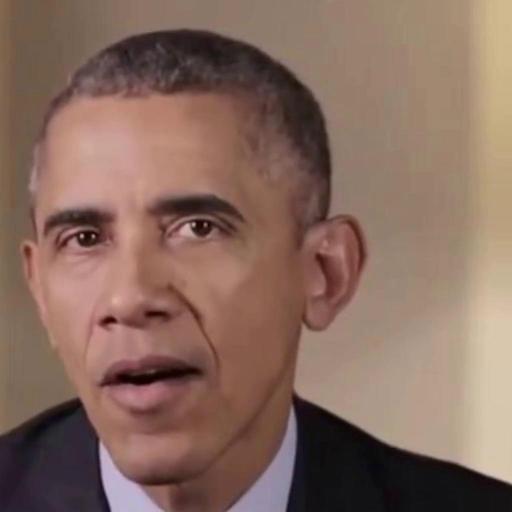} & 
        \includegraphics[width=0.15\columnwidth]{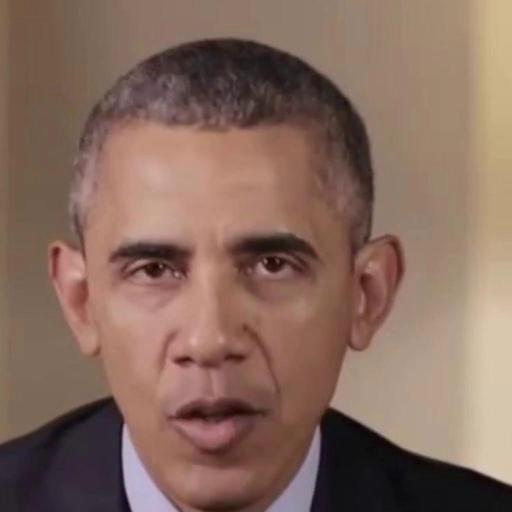} & 
        \includegraphics[width=0.15\columnwidth]{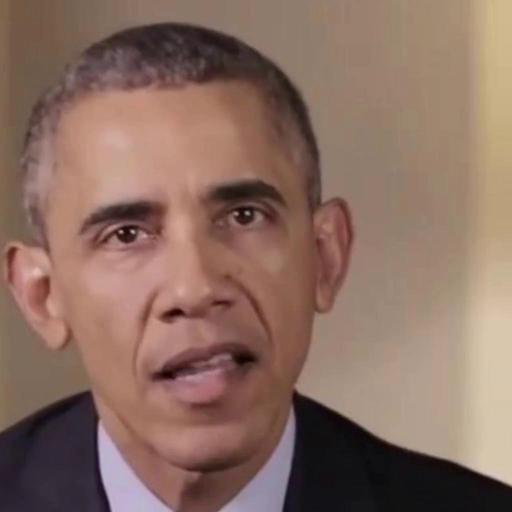} \\

		\raisebox{0.075in}{\rotatebox{90}{Recon.}} &
        \includegraphics[width=0.15\columnwidth]{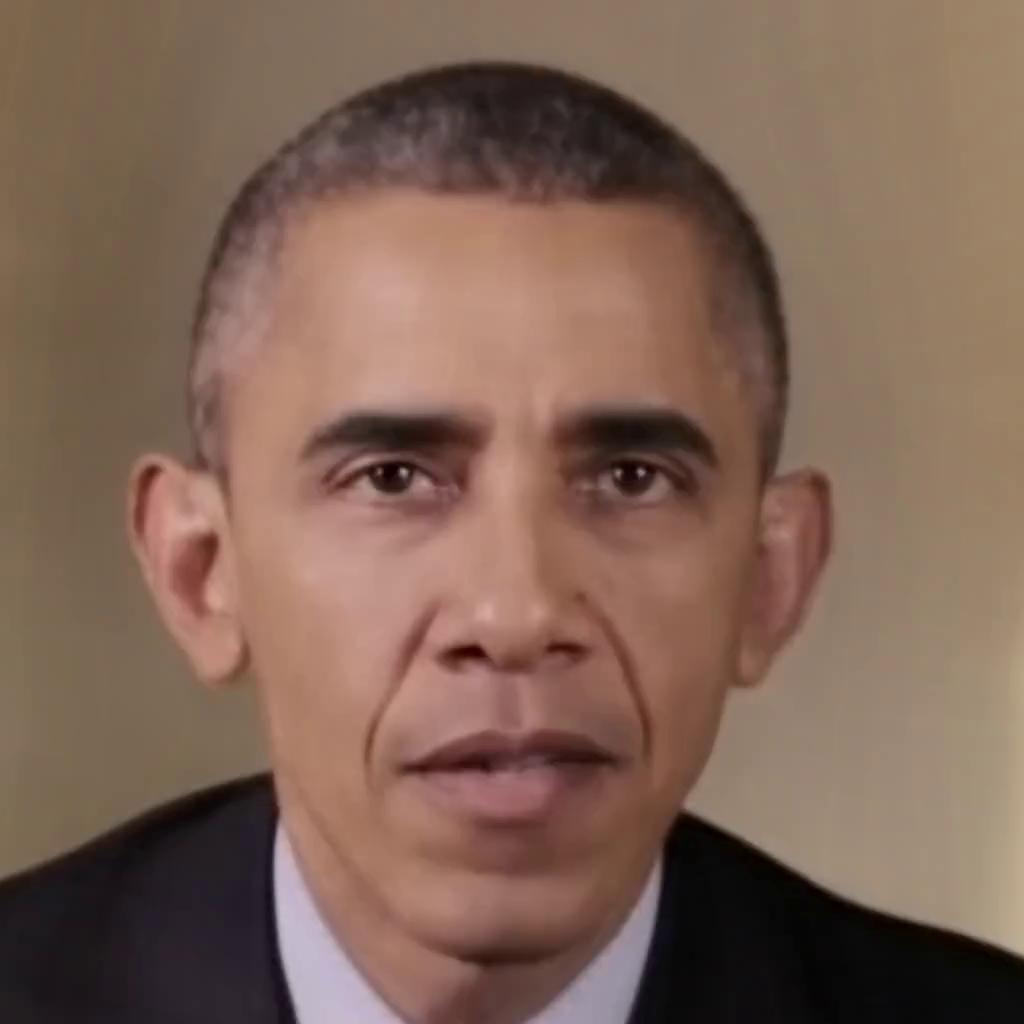} & 
        \includegraphics[width=0.15\columnwidth]{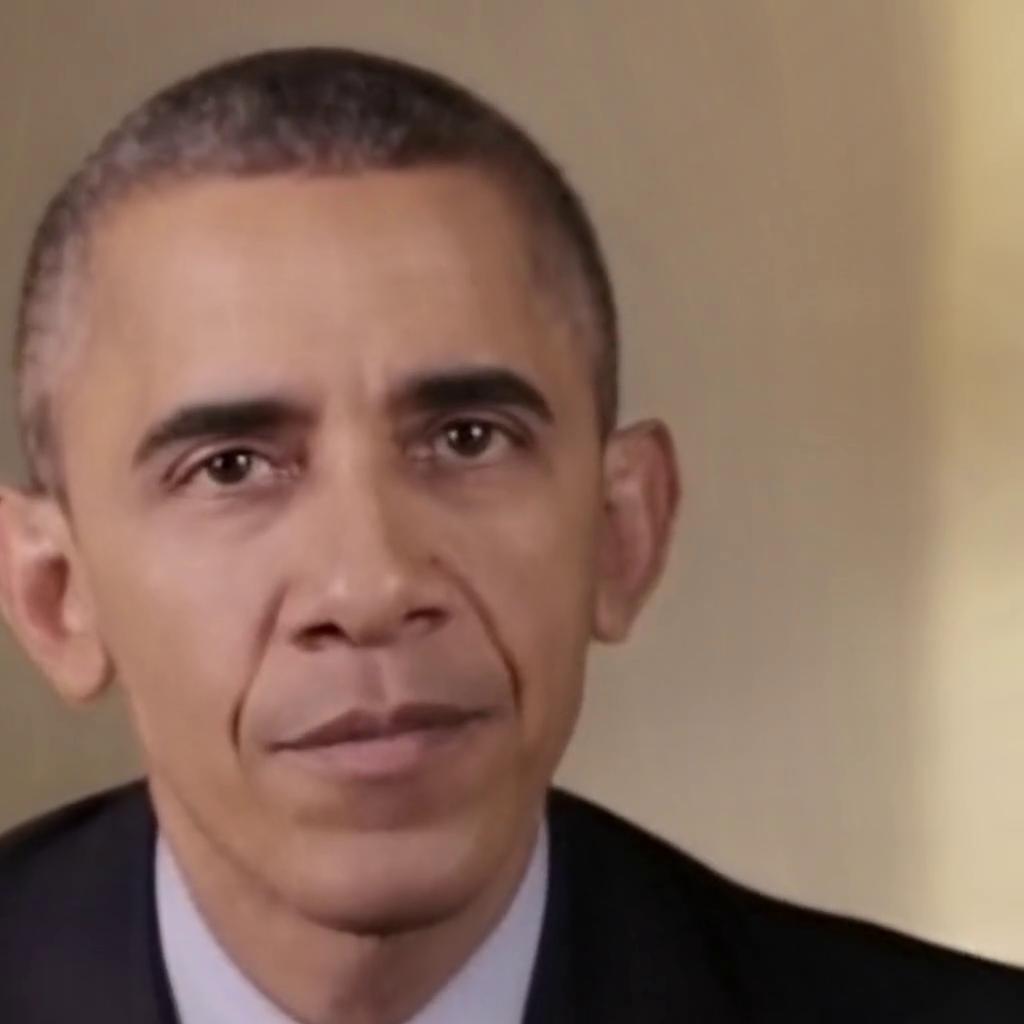} & 
        \includegraphics[width=0.15\columnwidth]{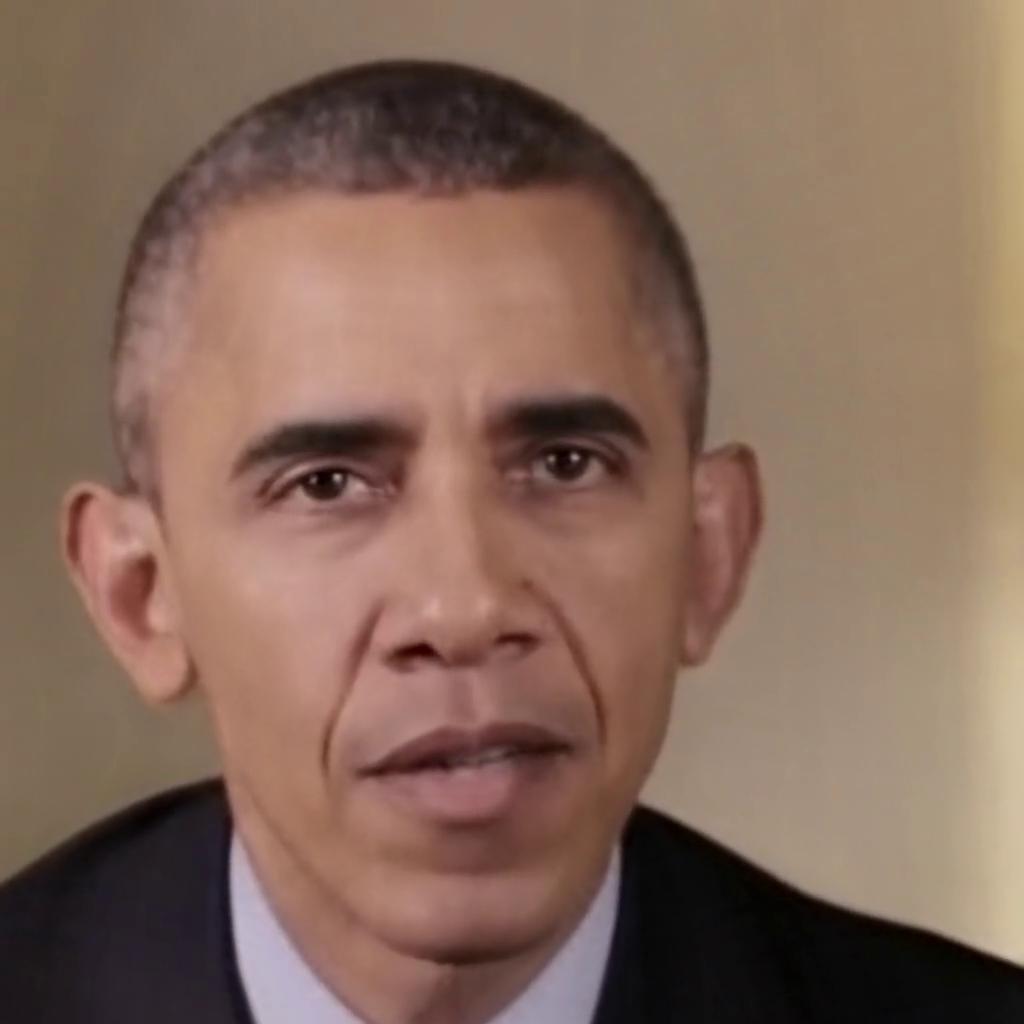} & 
        \includegraphics[width=0.15\columnwidth]{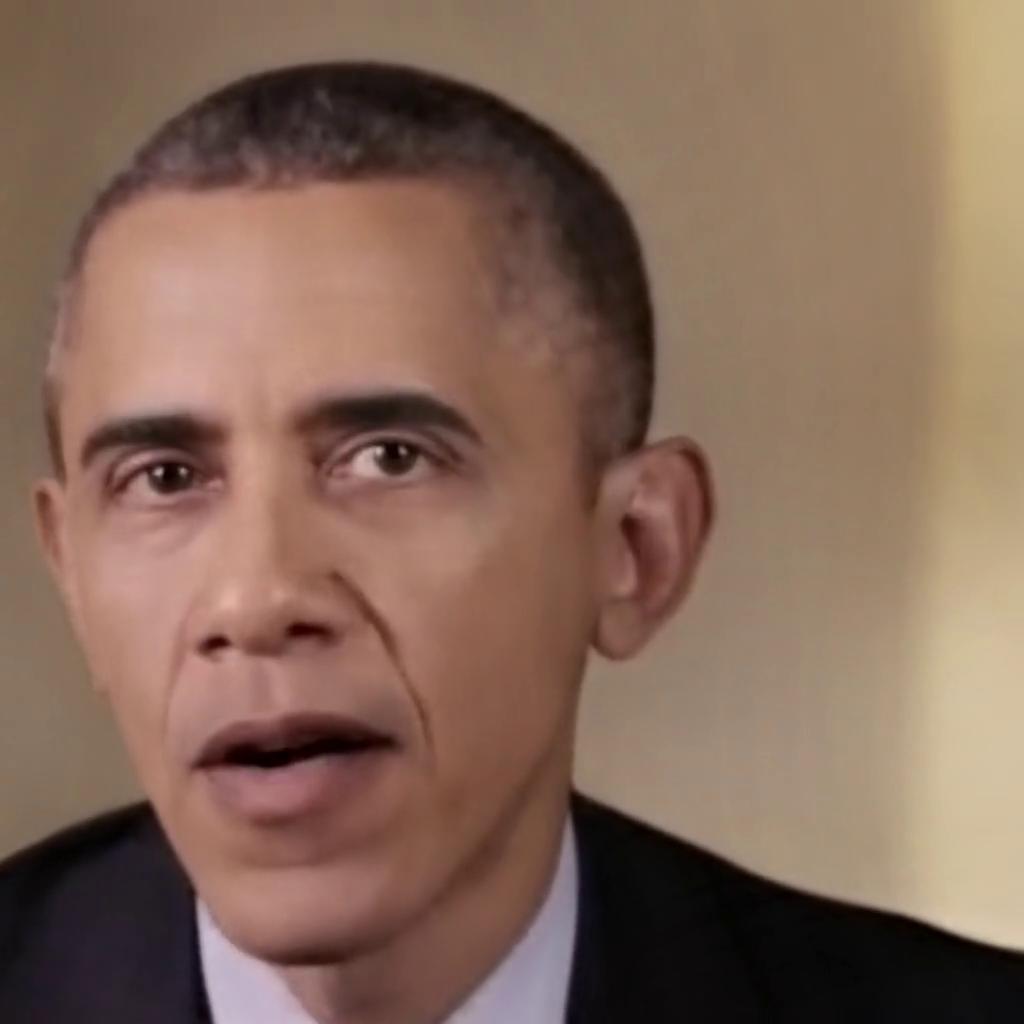} & 
        \includegraphics[width=0.15\columnwidth]{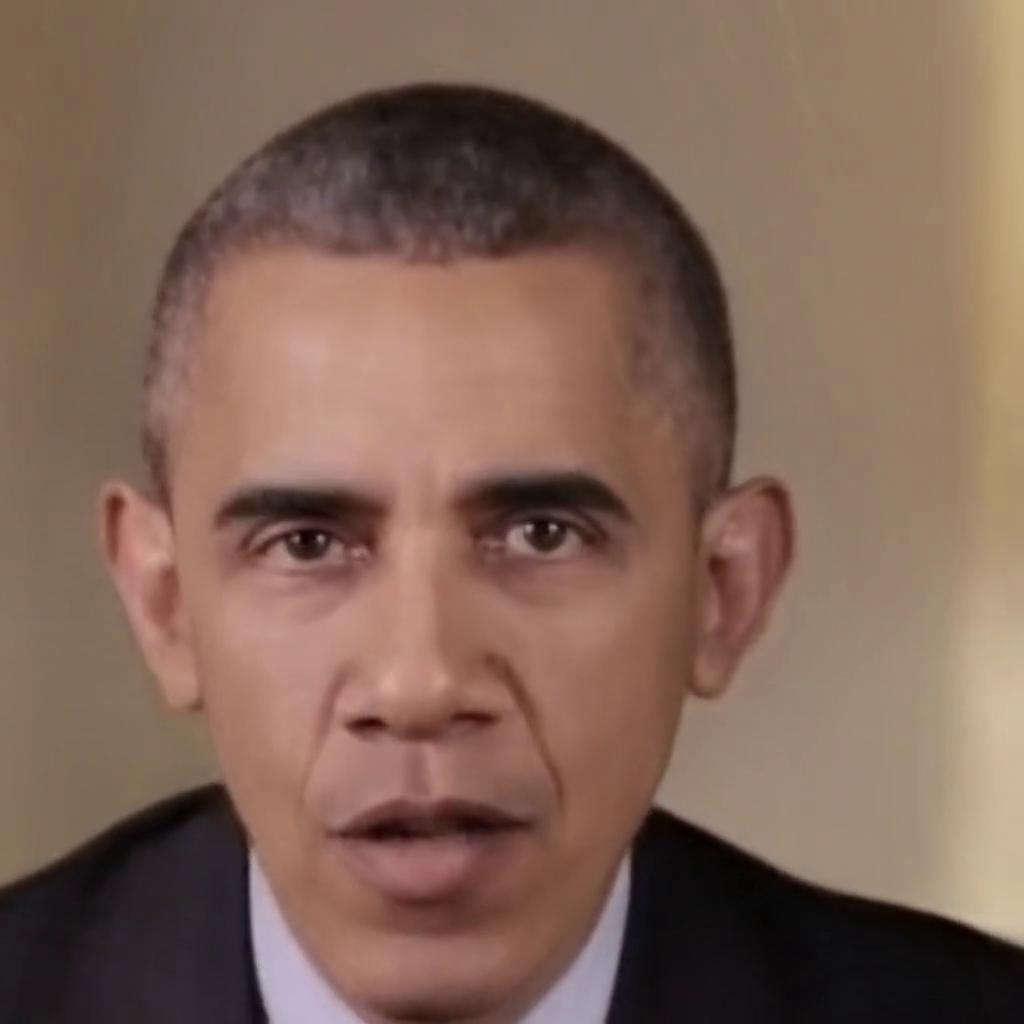} & 
        \includegraphics[width=0.15\columnwidth]{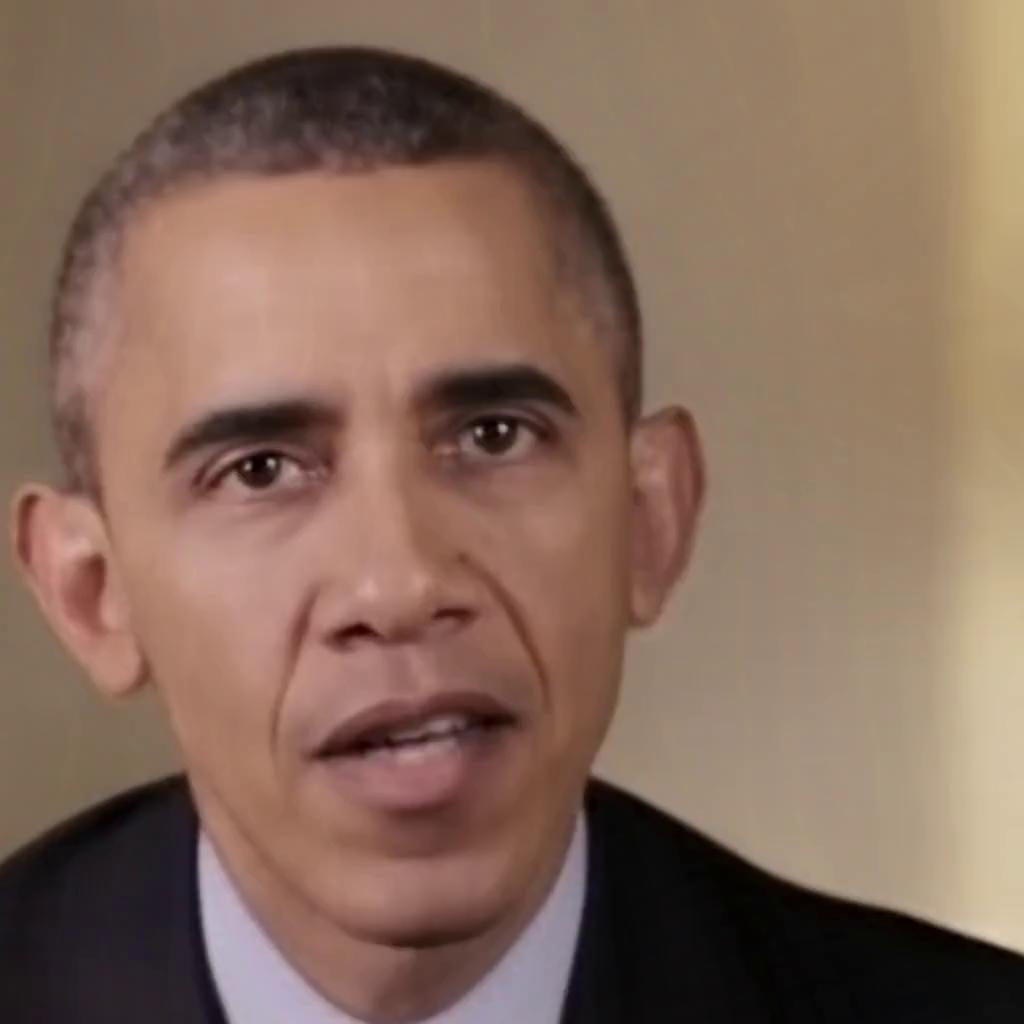} \\

		\raisebox{0.075in}{\rotatebox{90}{$+$ Age}} &
        \includegraphics[width=0.15\columnwidth]{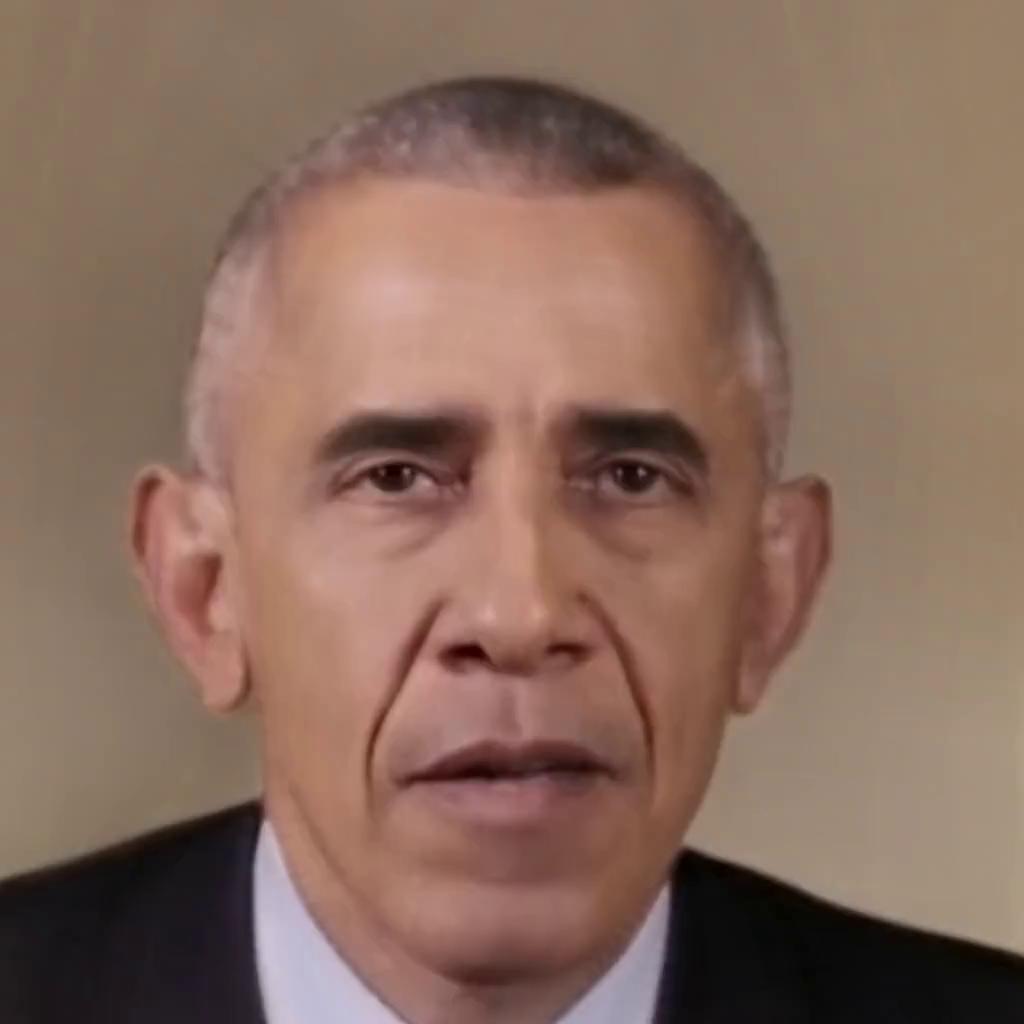} & 
        \includegraphics[width=0.15\columnwidth]{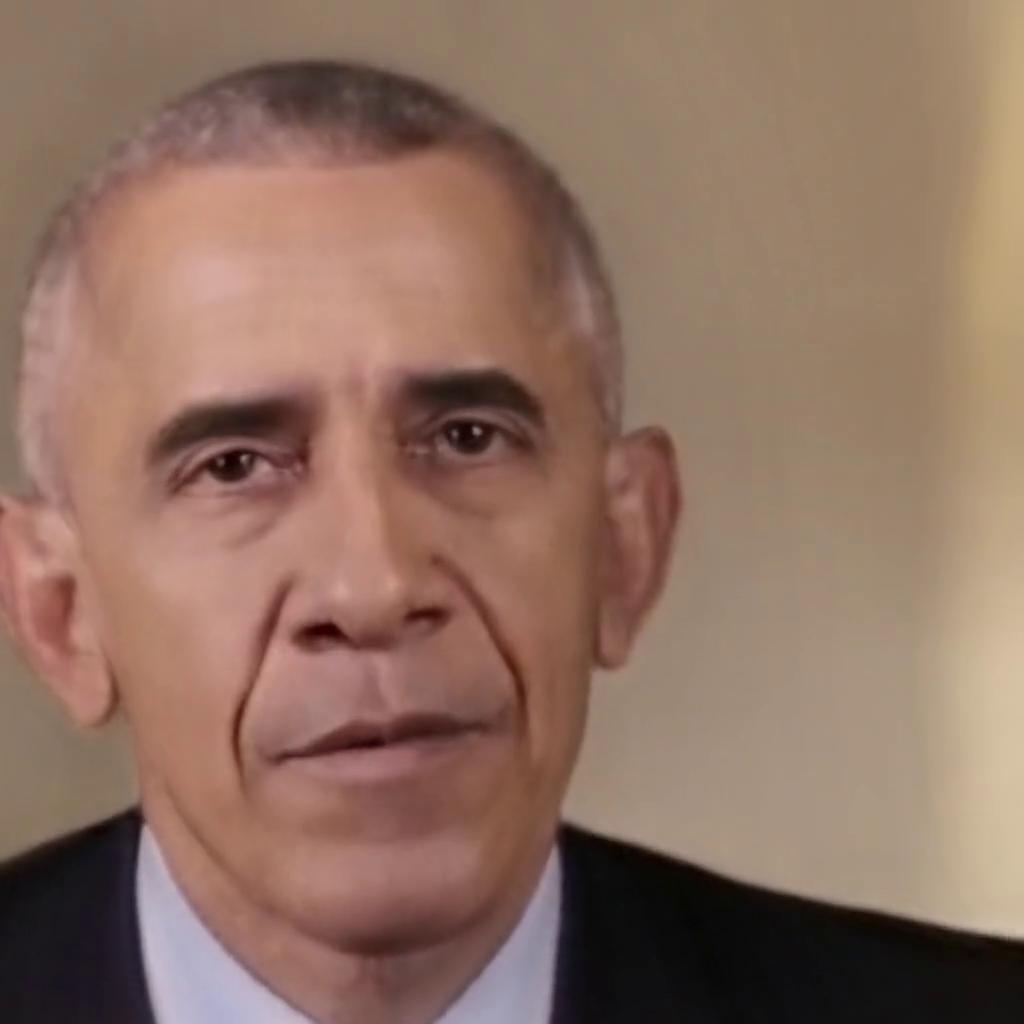} & 
        \includegraphics[width=0.15\columnwidth]{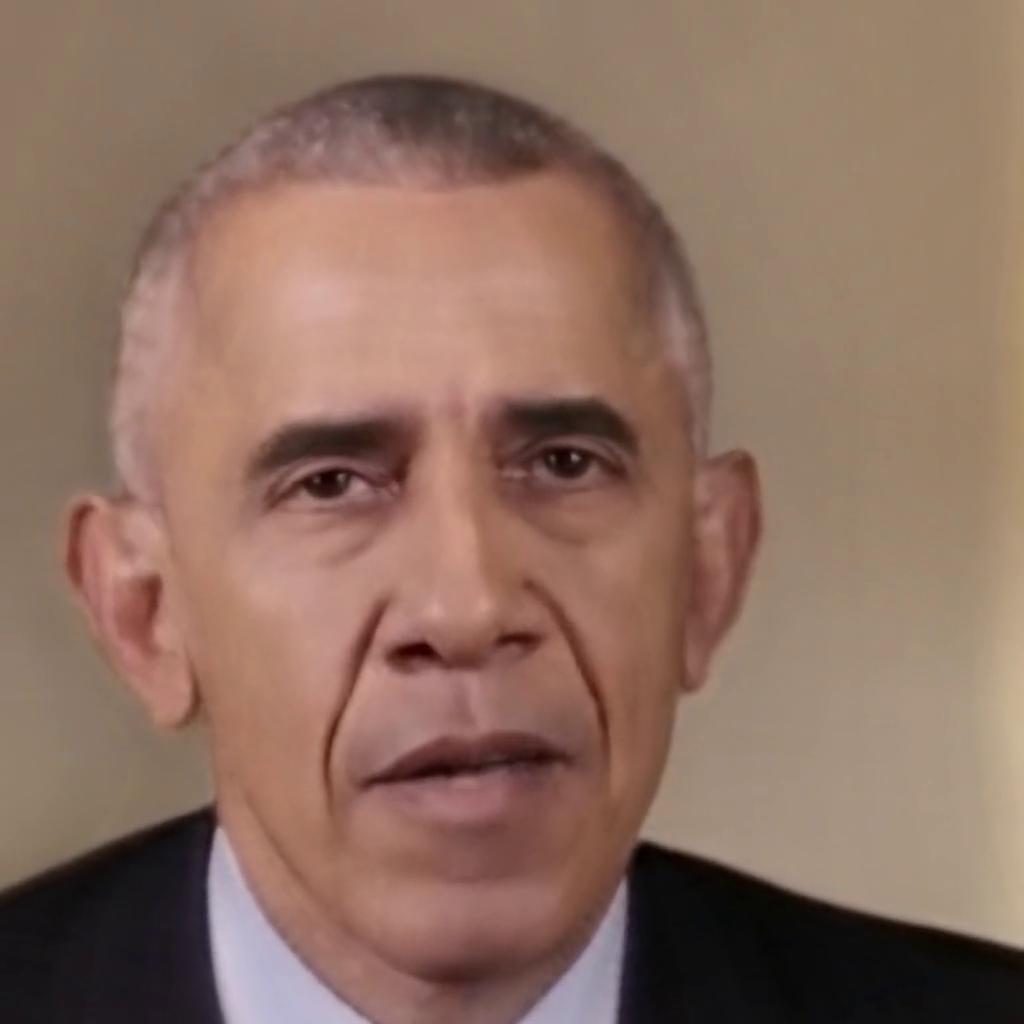} & 
        \includegraphics[width=0.15\columnwidth]{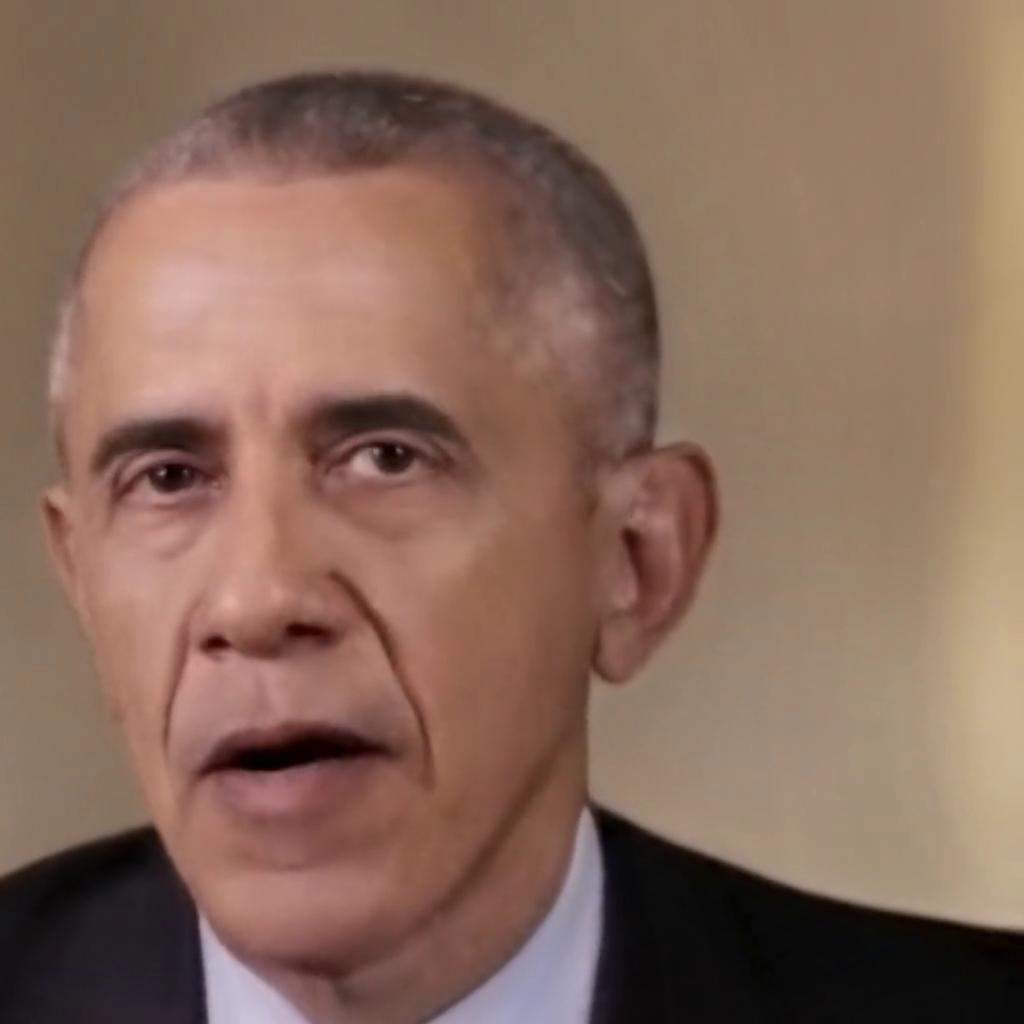} & 
        \includegraphics[width=0.15\columnwidth]{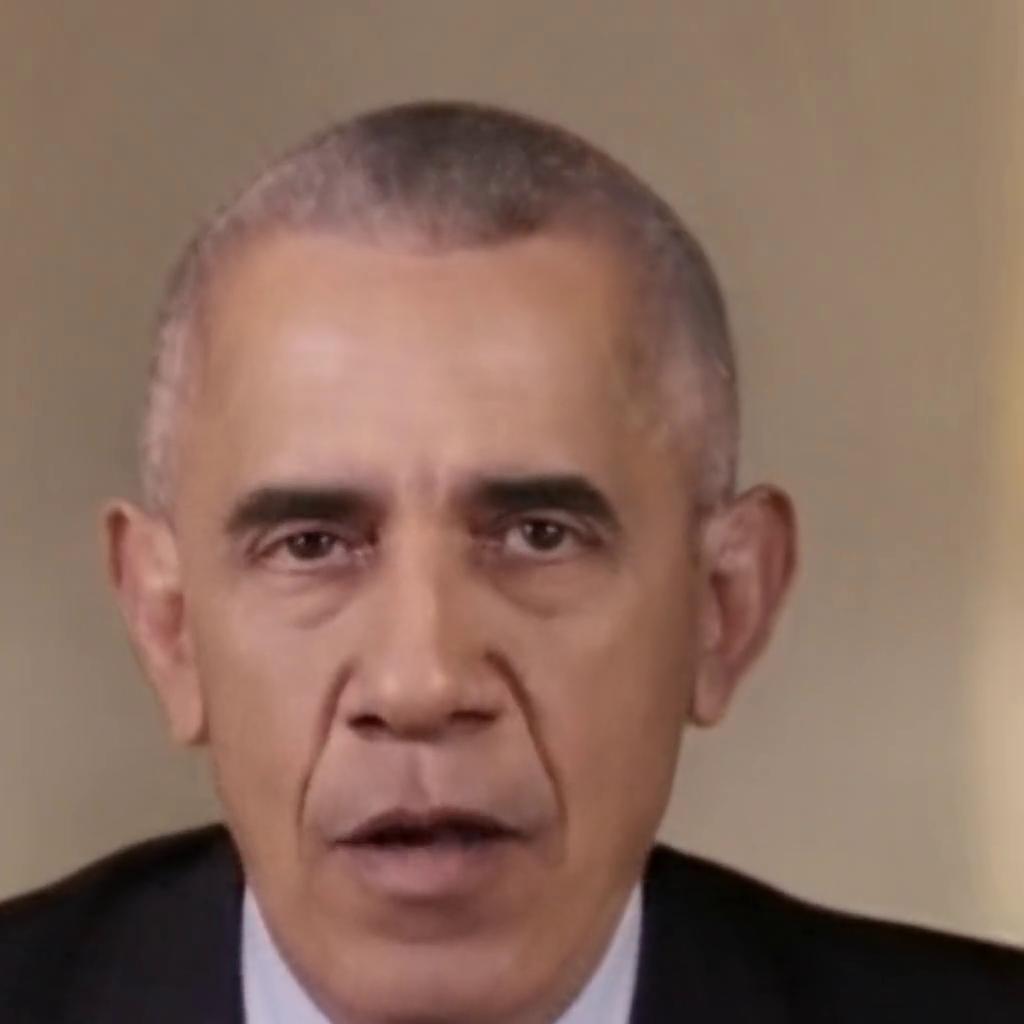} & 
        \includegraphics[width=0.15\columnwidth]{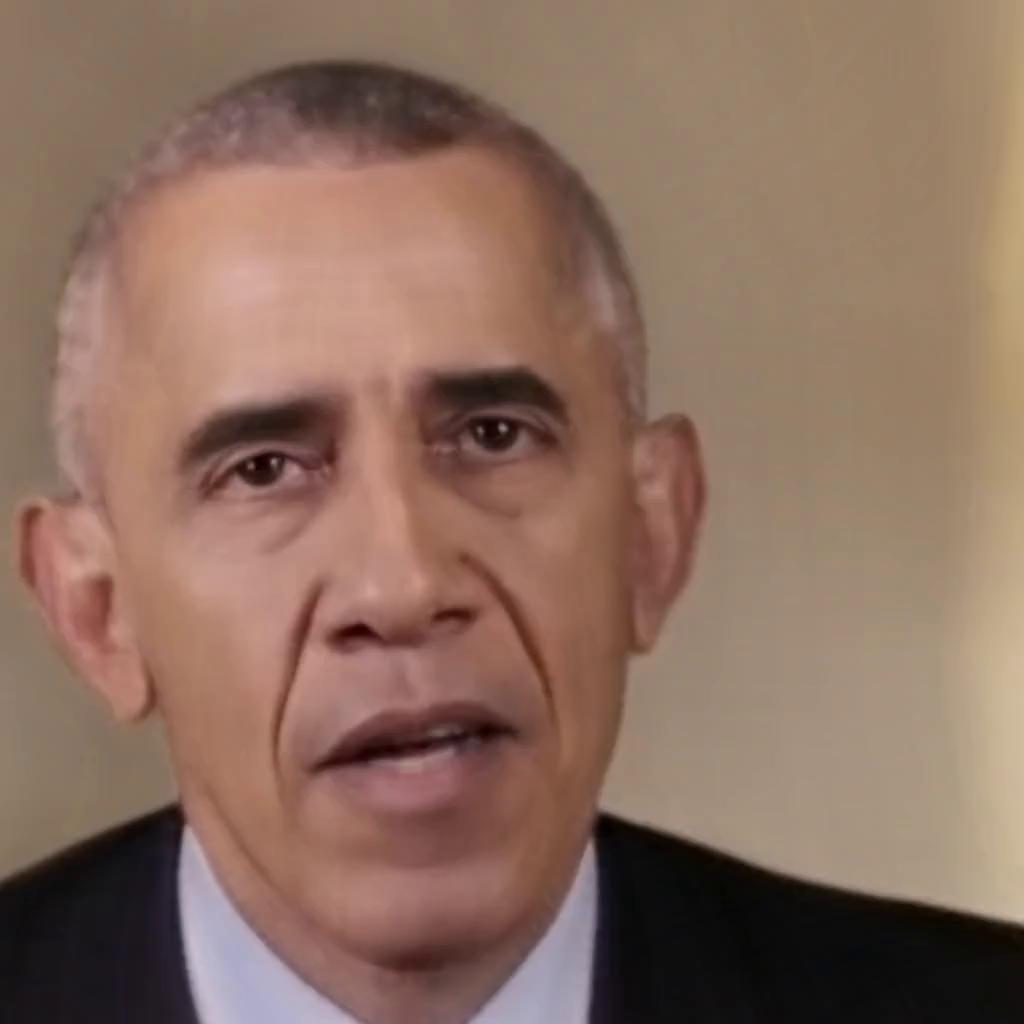} \\

		\raisebox{0.05in}{\rotatebox{90}{$+$ Pixar}} &
        \includegraphics[width=0.15\columnwidth]{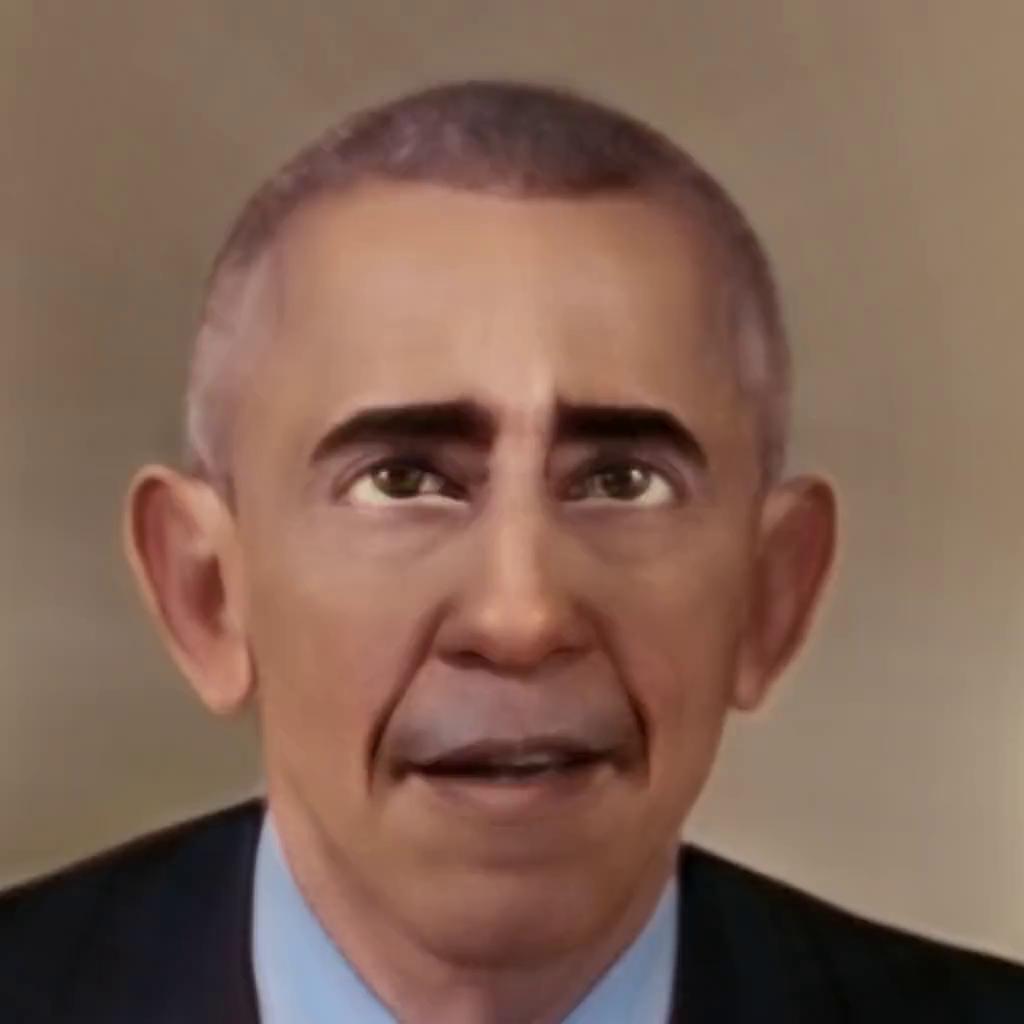} & 
        \includegraphics[width=0.15\columnwidth]{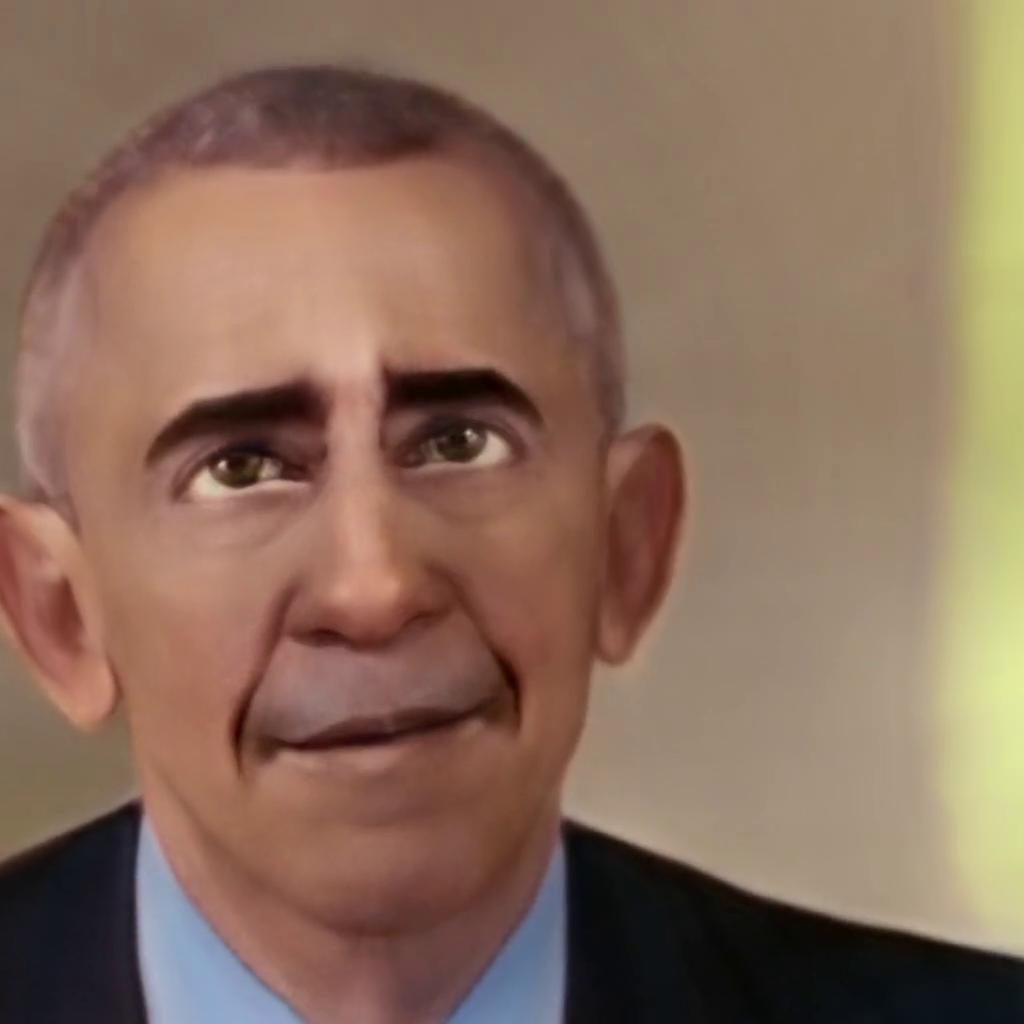} & 
        \includegraphics[width=0.15\columnwidth]{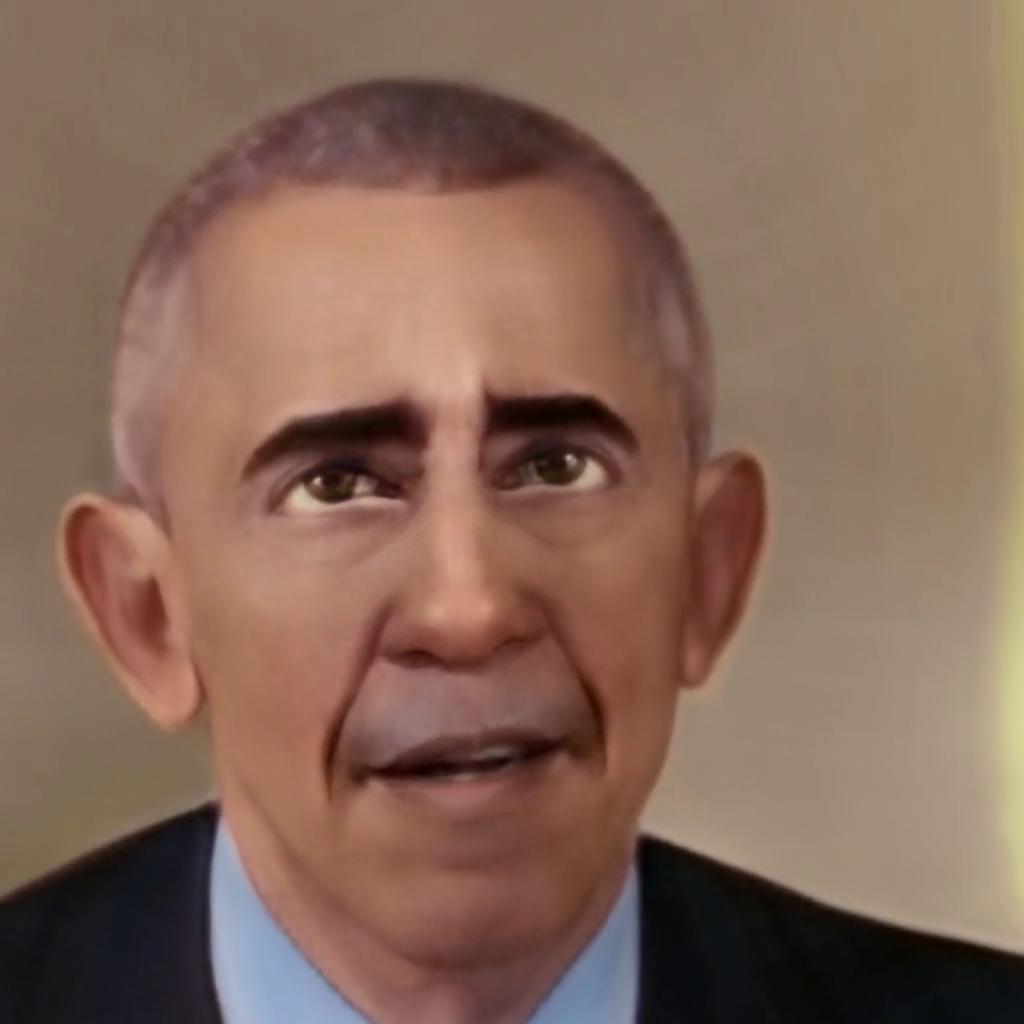} & 
        \includegraphics[width=0.15\columnwidth]{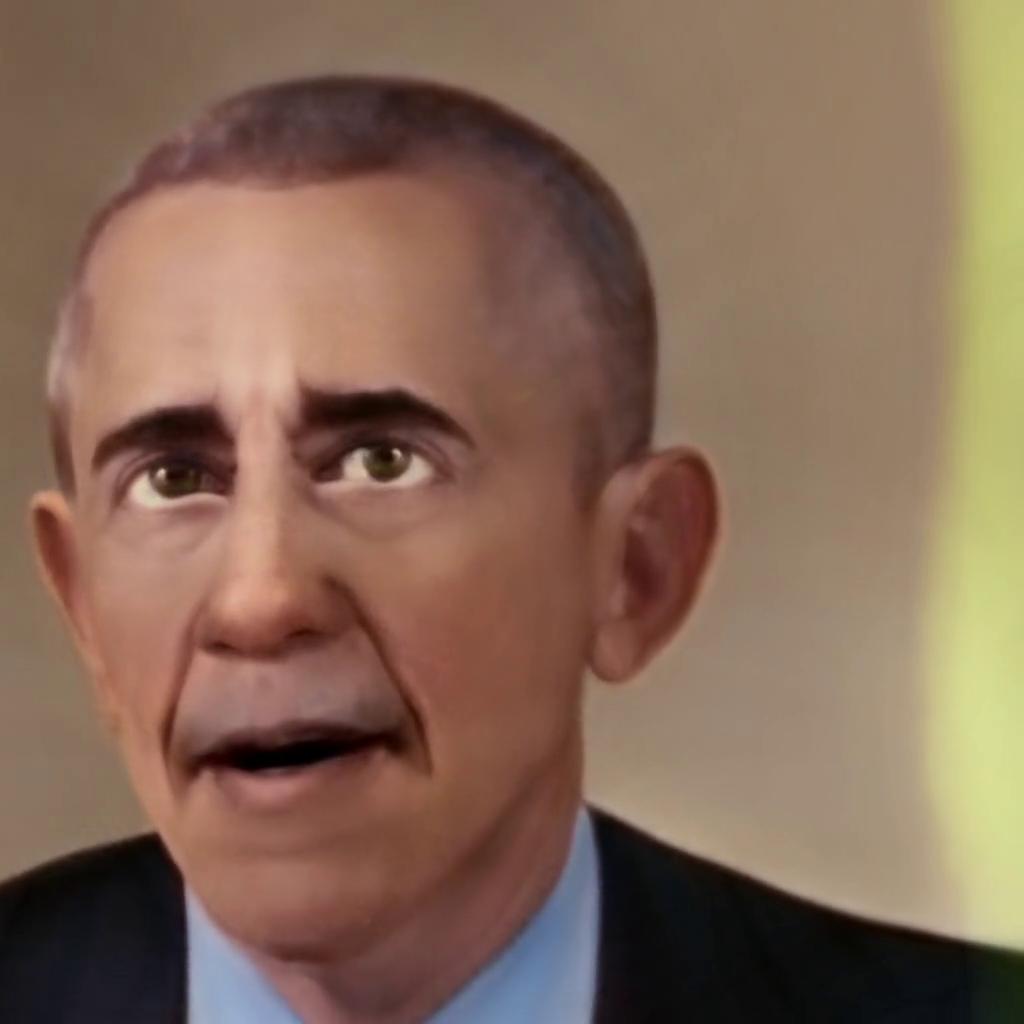} & 
        \includegraphics[width=0.15\columnwidth]{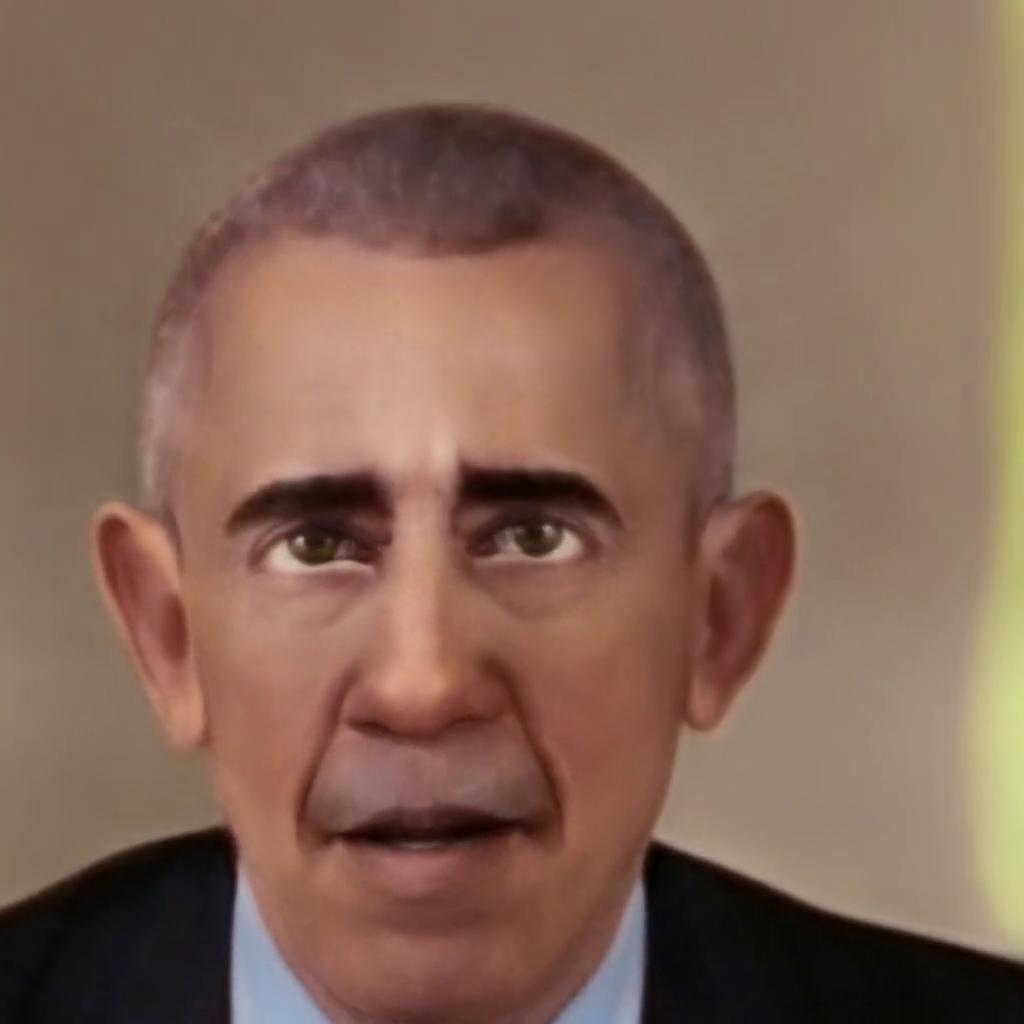} & 
        \includegraphics[width=0.15\columnwidth]{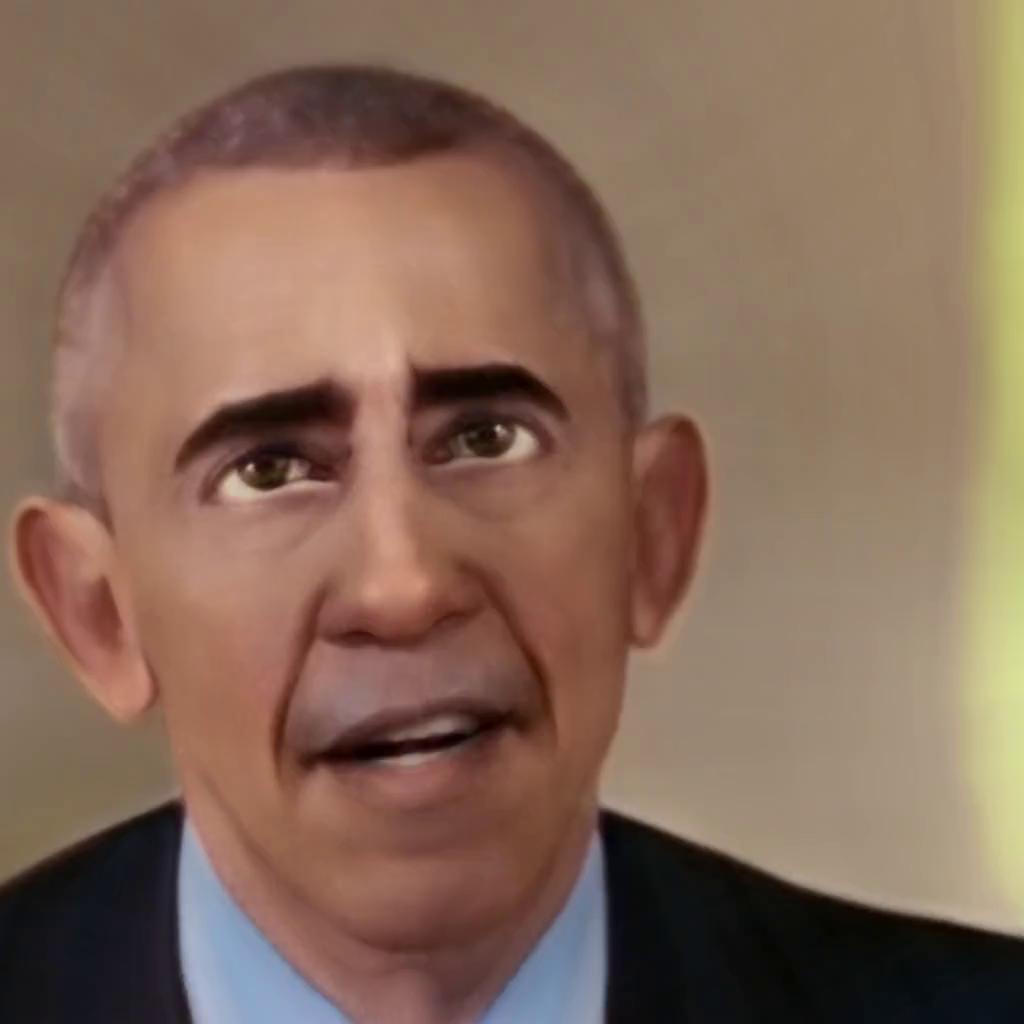} \\

	\end{tabular}
	}
	\vspace{-0.3cm}
	\caption{
    Sample results of our full video encoding and editing pipeline with StyleGAN3. Additional full video results and editing examples are provided in the supplementary materials.
    }
    \vspace{-0.15cm}
	\label{fig:video_results}
\end{figure}

%% file: figures/wide_dwight.tex
\begin{figure}[tb]
	\centering
	\setlength{\tabcolsep}{1pt}	

	{\footnotesize
	\begin{tabular}{c c c c c c c c}
        
		\raisebox{0.075in}{\rotatebox{90}{Original}} &
        \includegraphics[width=0.18\columnwidth]{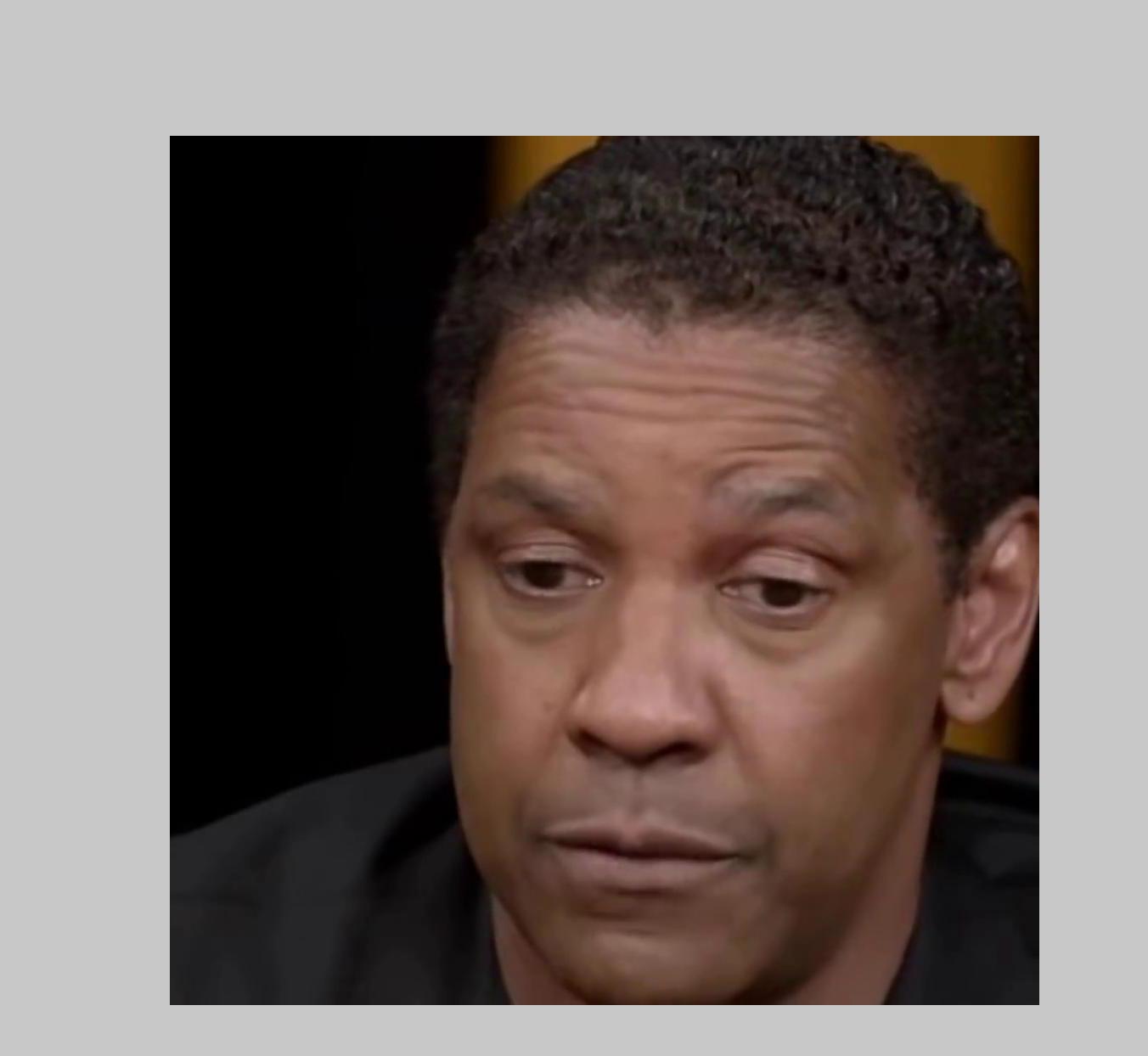} & 
        \includegraphics[width=0.18\columnwidth]{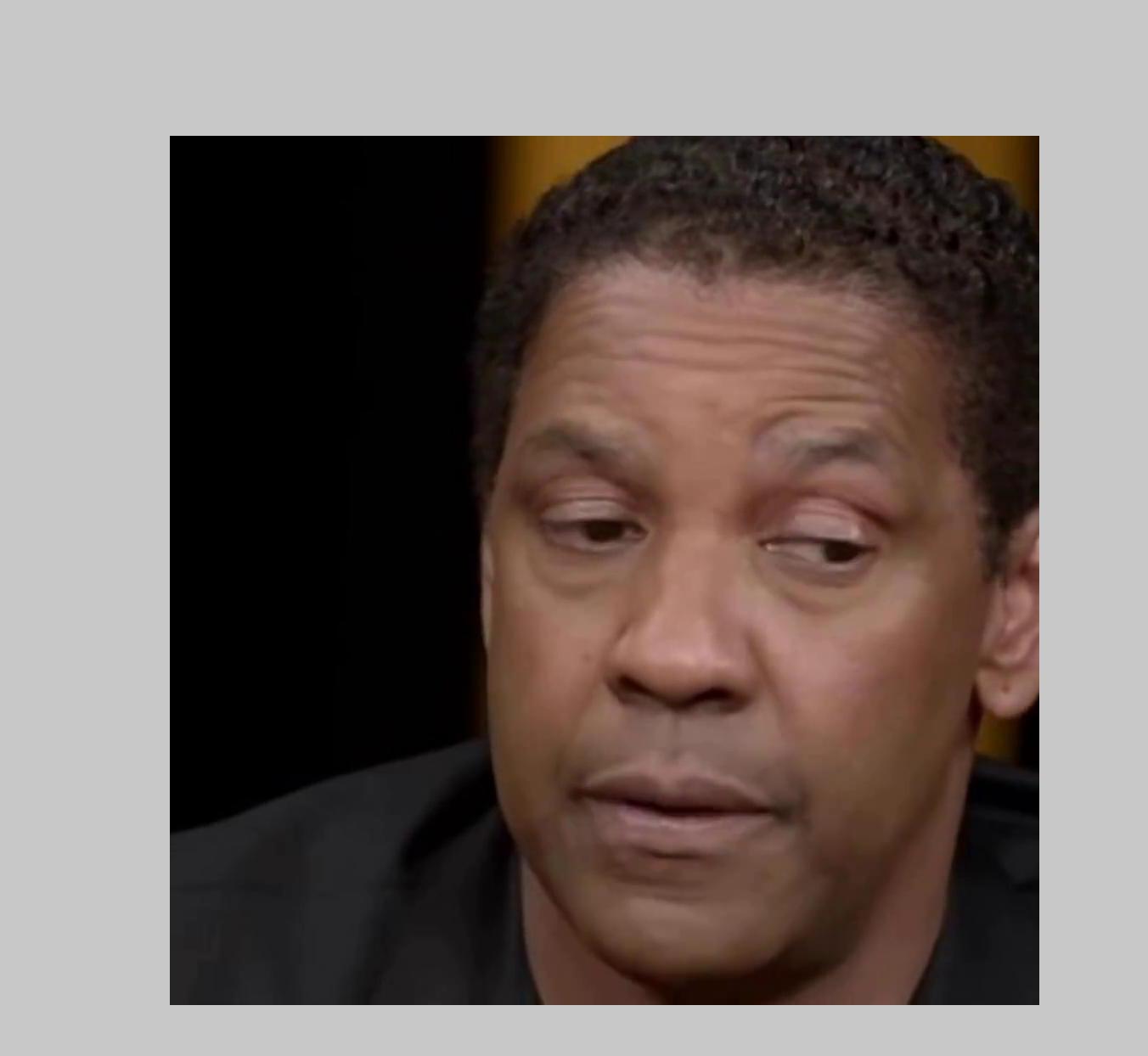} & 
        \includegraphics[width=0.18\columnwidth]{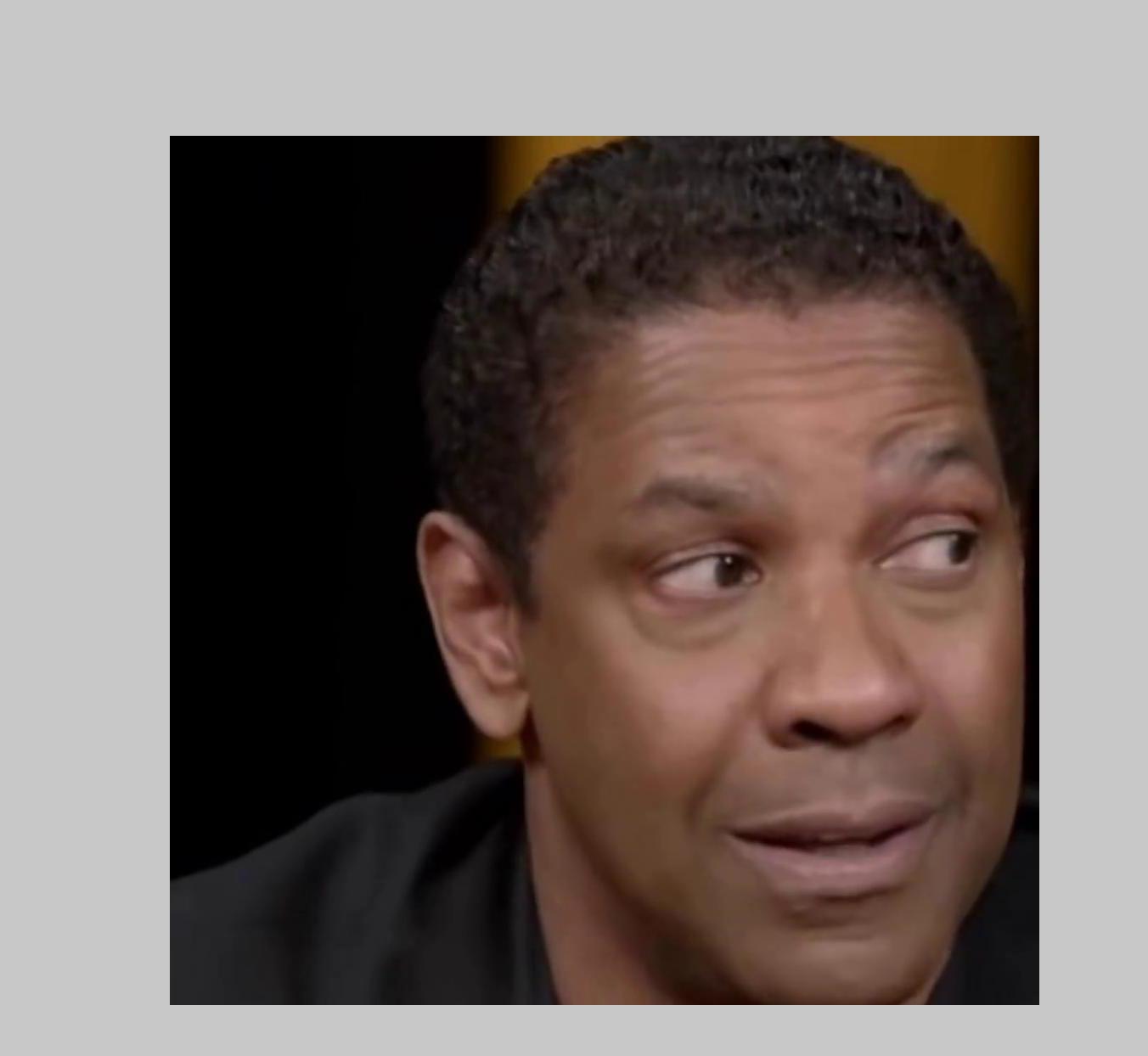} & 
        \includegraphics[width=0.18\columnwidth]{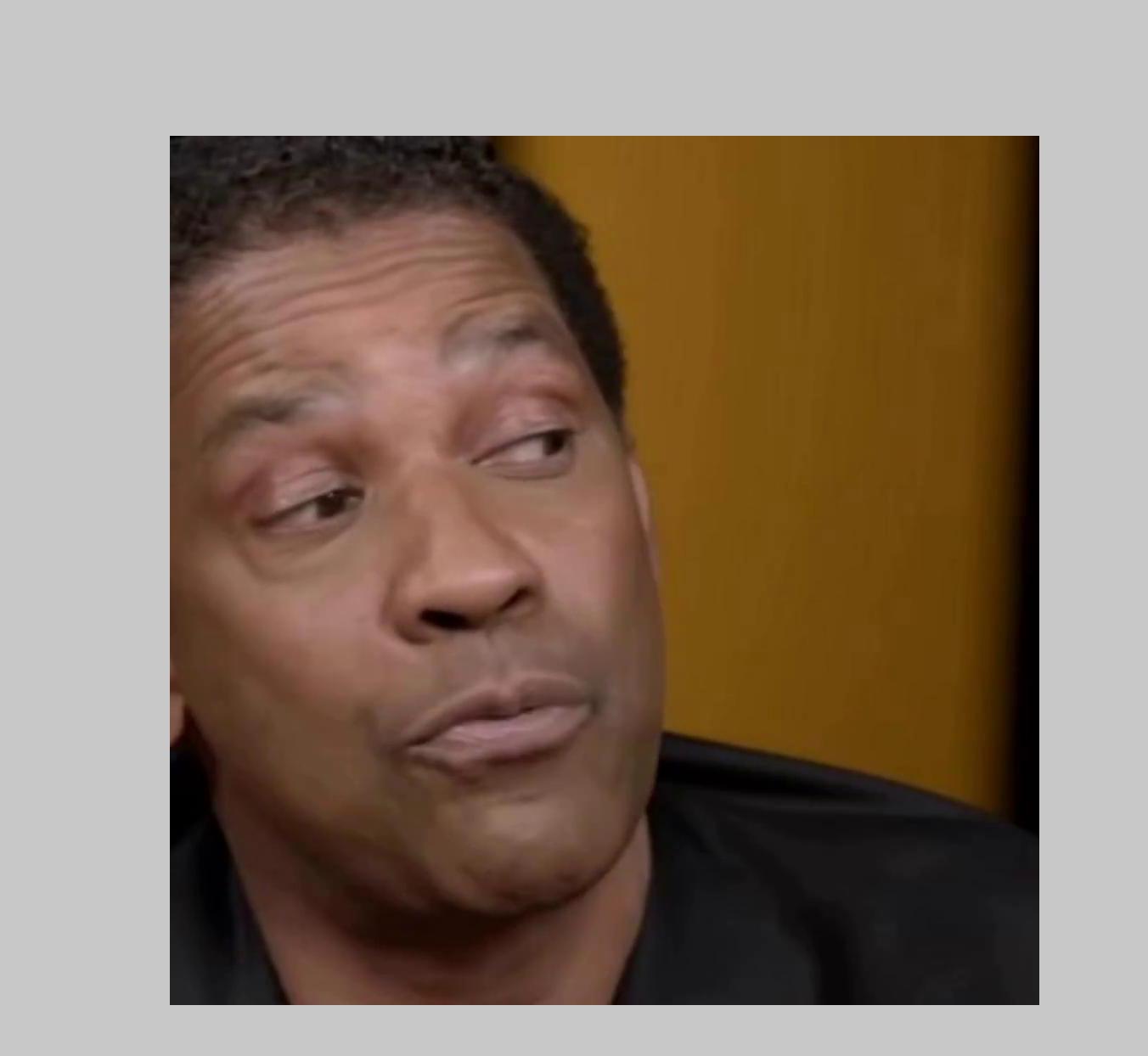} &
        \includegraphics[width=0.18\columnwidth]{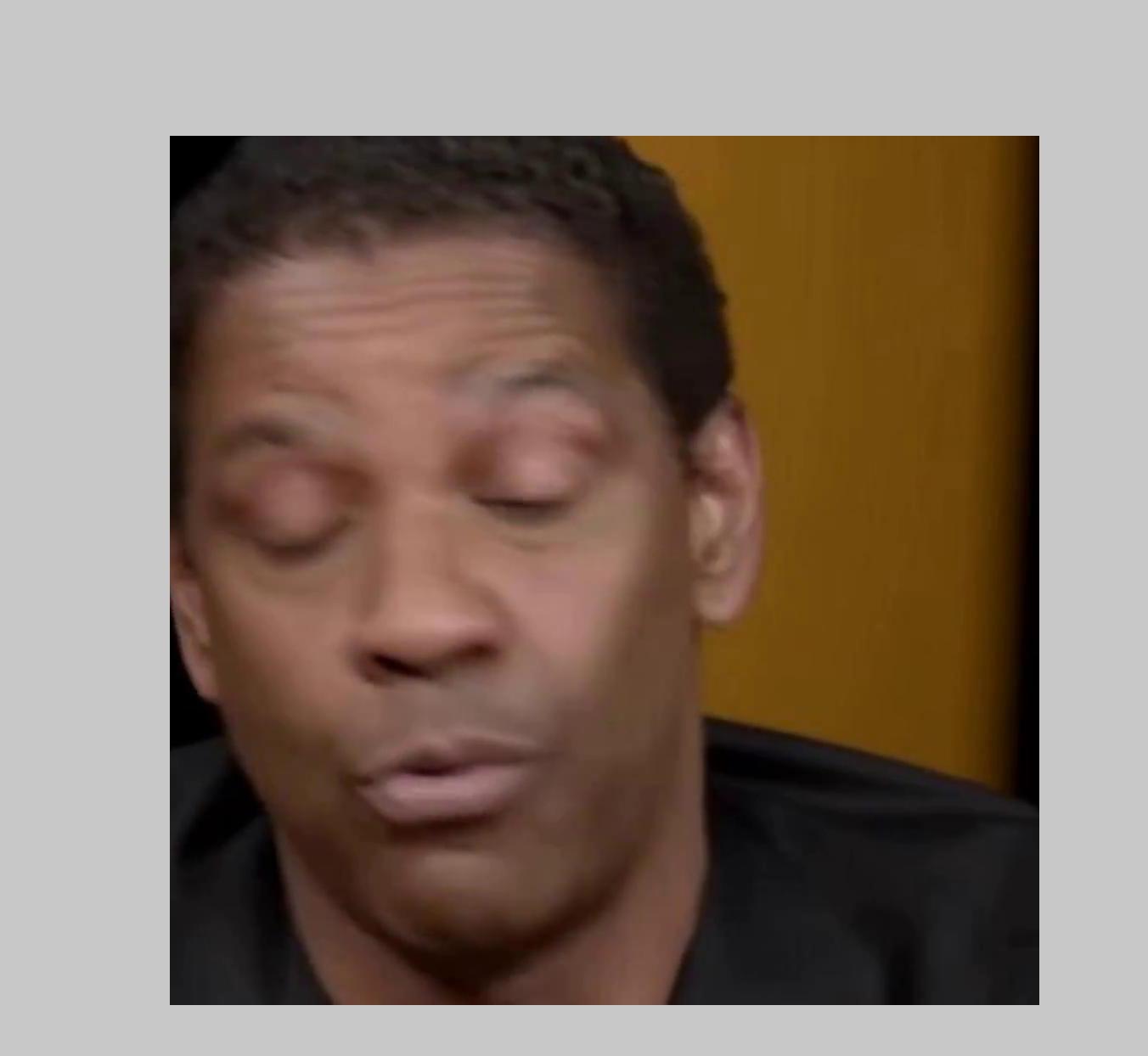} \\

		\raisebox{0.1in}{\rotatebox{90}{Recon.}} &
        \includegraphics[width=0.18\columnwidth]{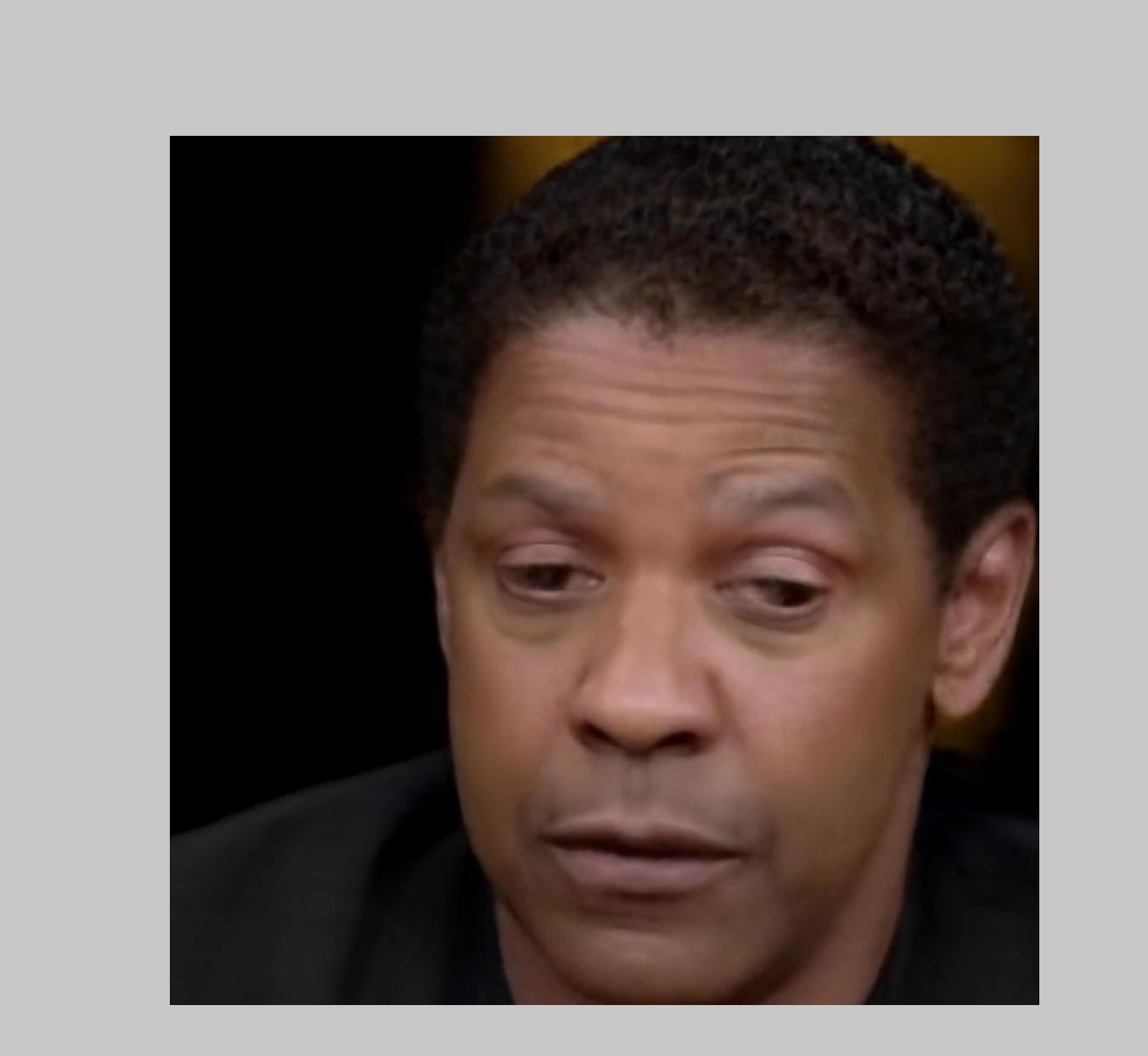} & 
        \includegraphics[width=0.18\columnwidth]{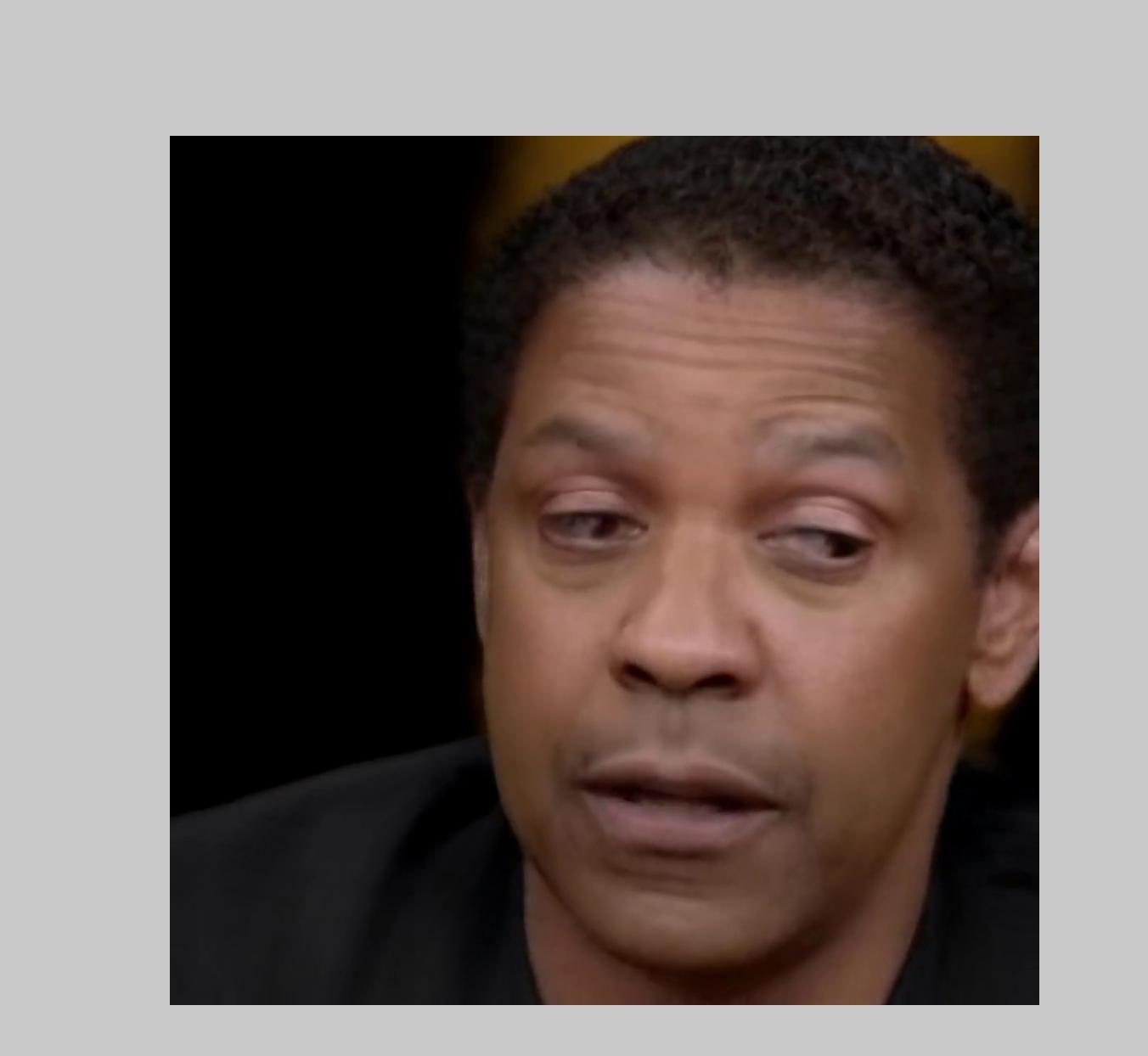} & 
        \includegraphics[width=0.18\columnwidth]{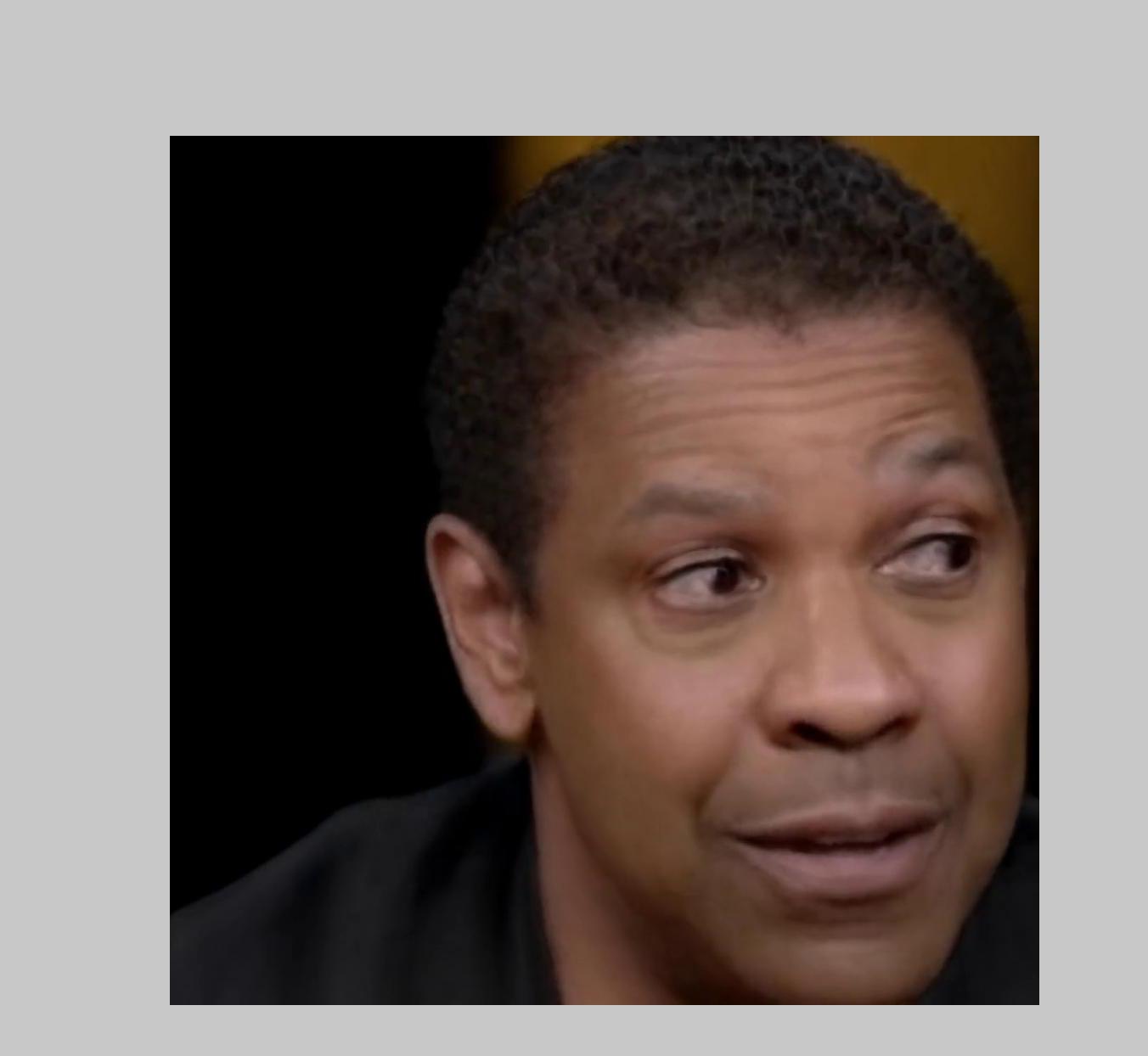} & 
        \includegraphics[width=0.18\columnwidth]{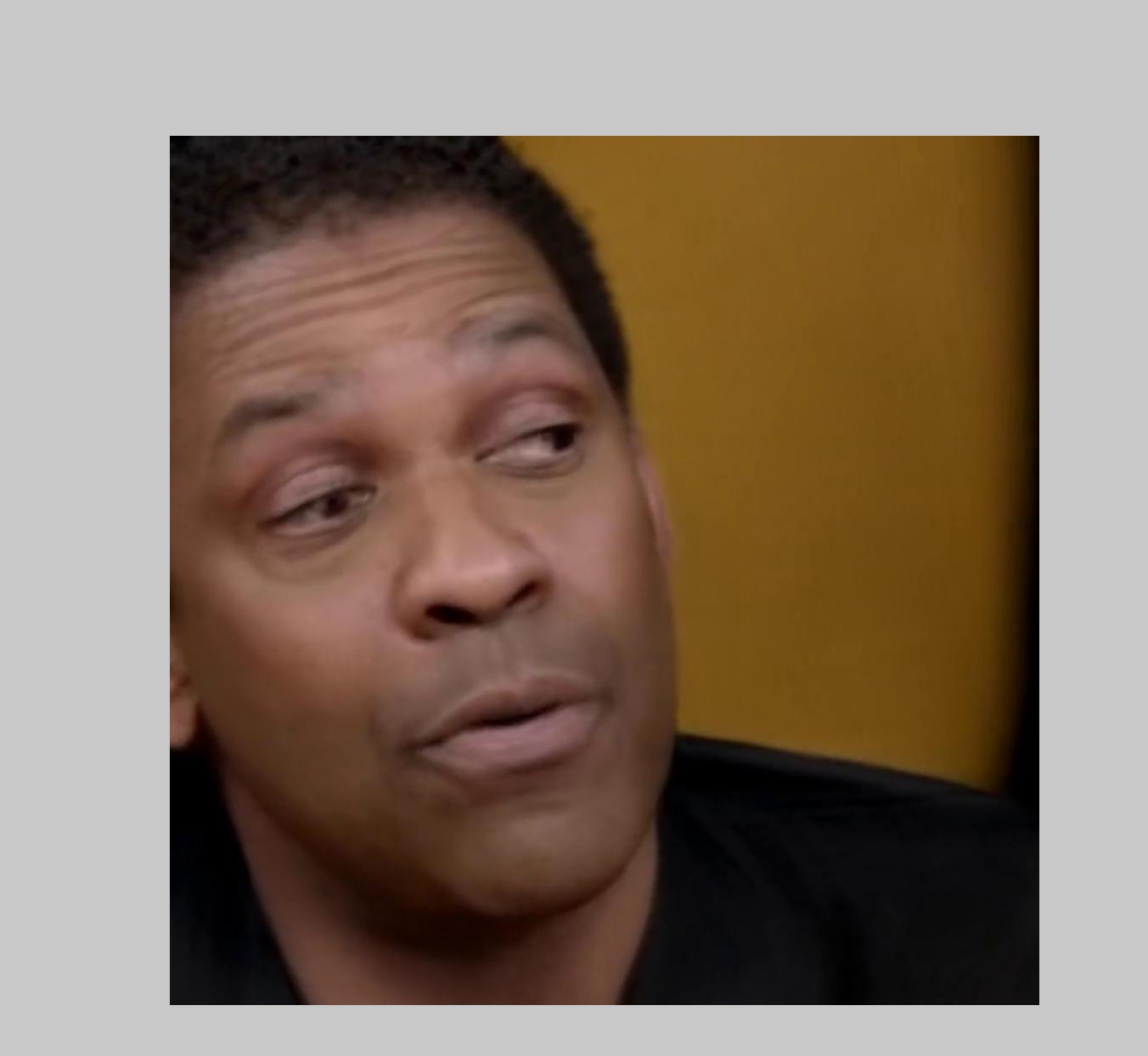} & 
        \includegraphics[width=0.18\columnwidth]{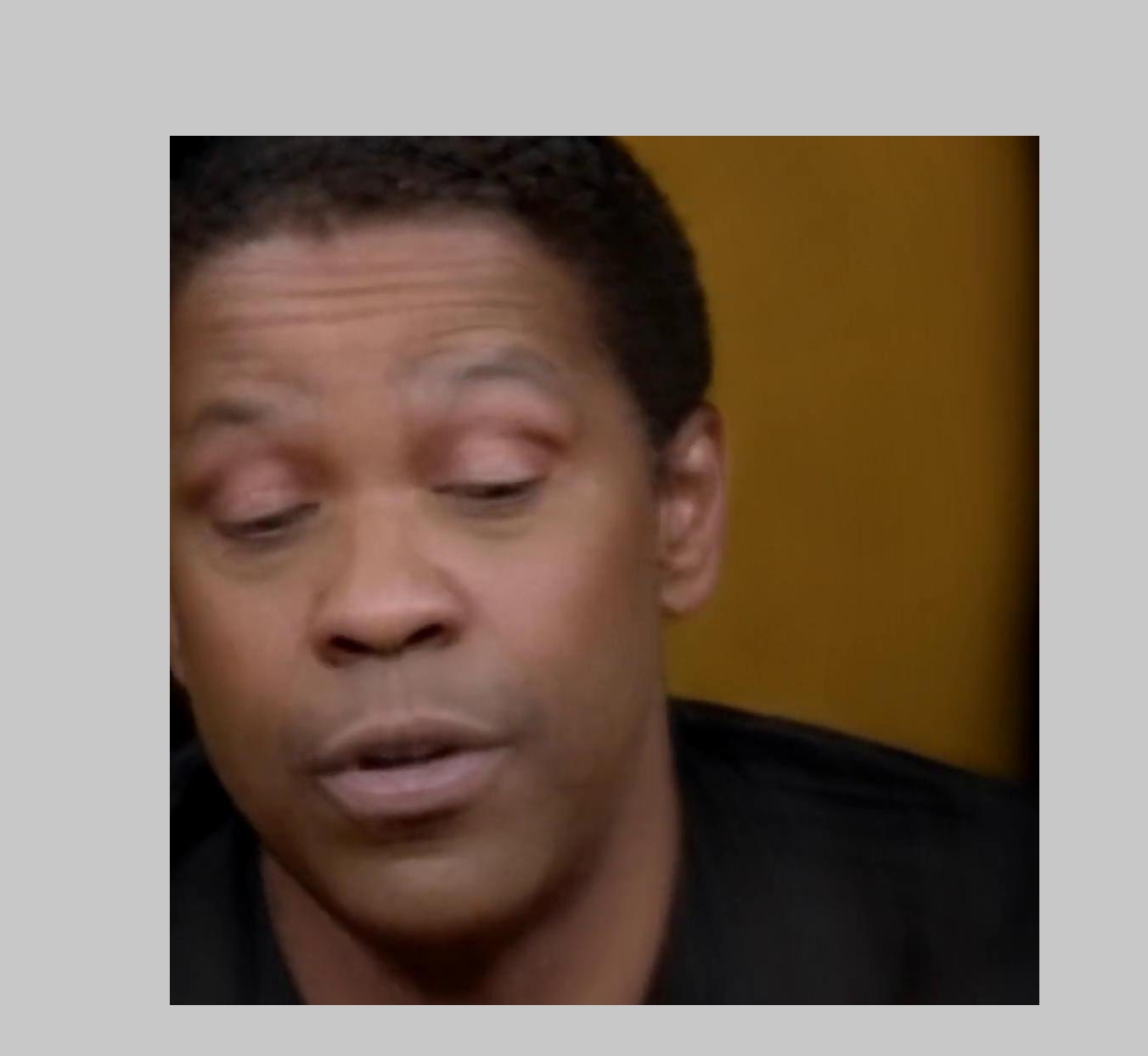} \\

		\raisebox{0.025in}{\rotatebox{90}{Wide View}} &
        \includegraphics[width=0.18\columnwidth]{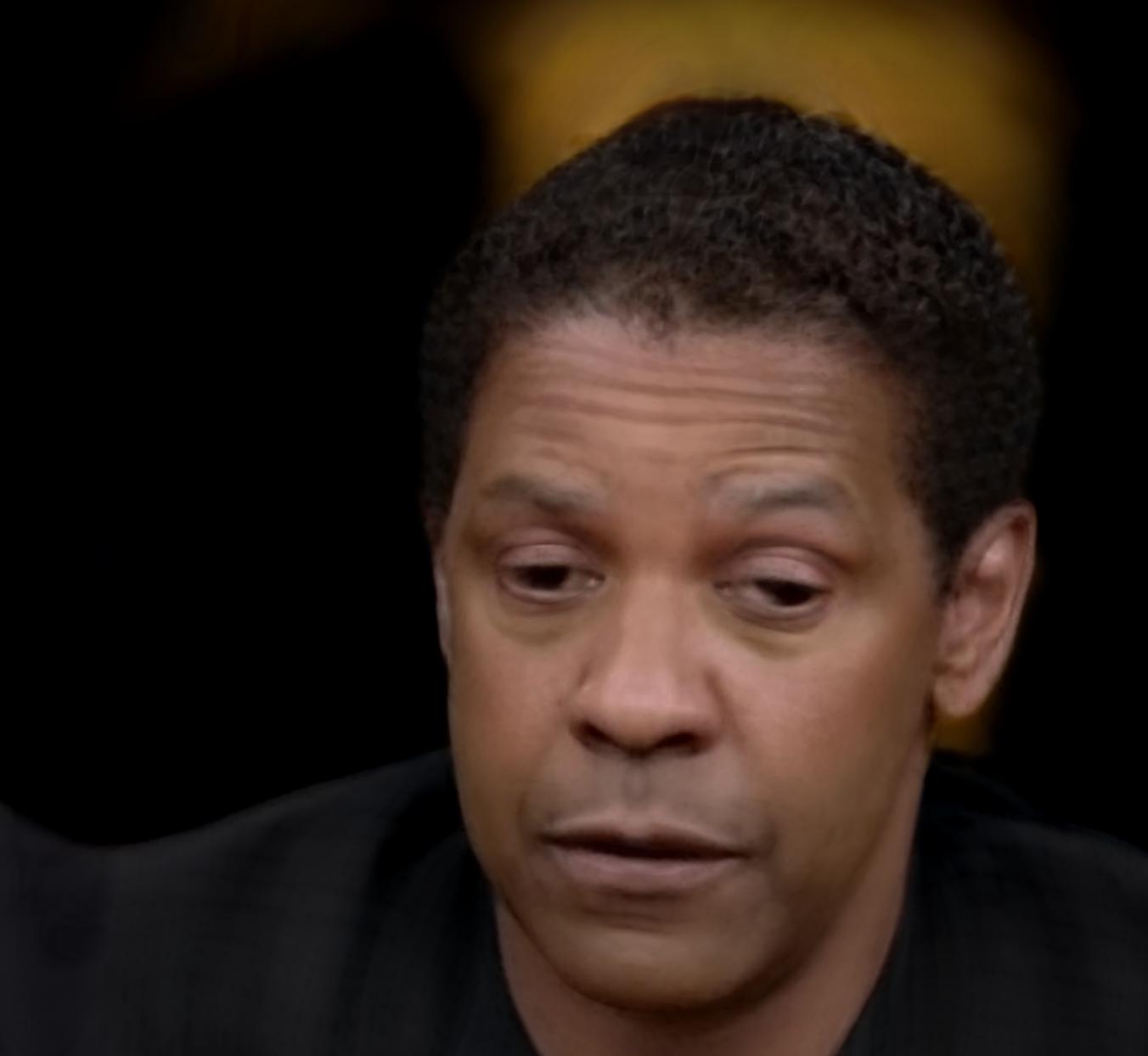} & 
        \includegraphics[width=0.18\columnwidth]{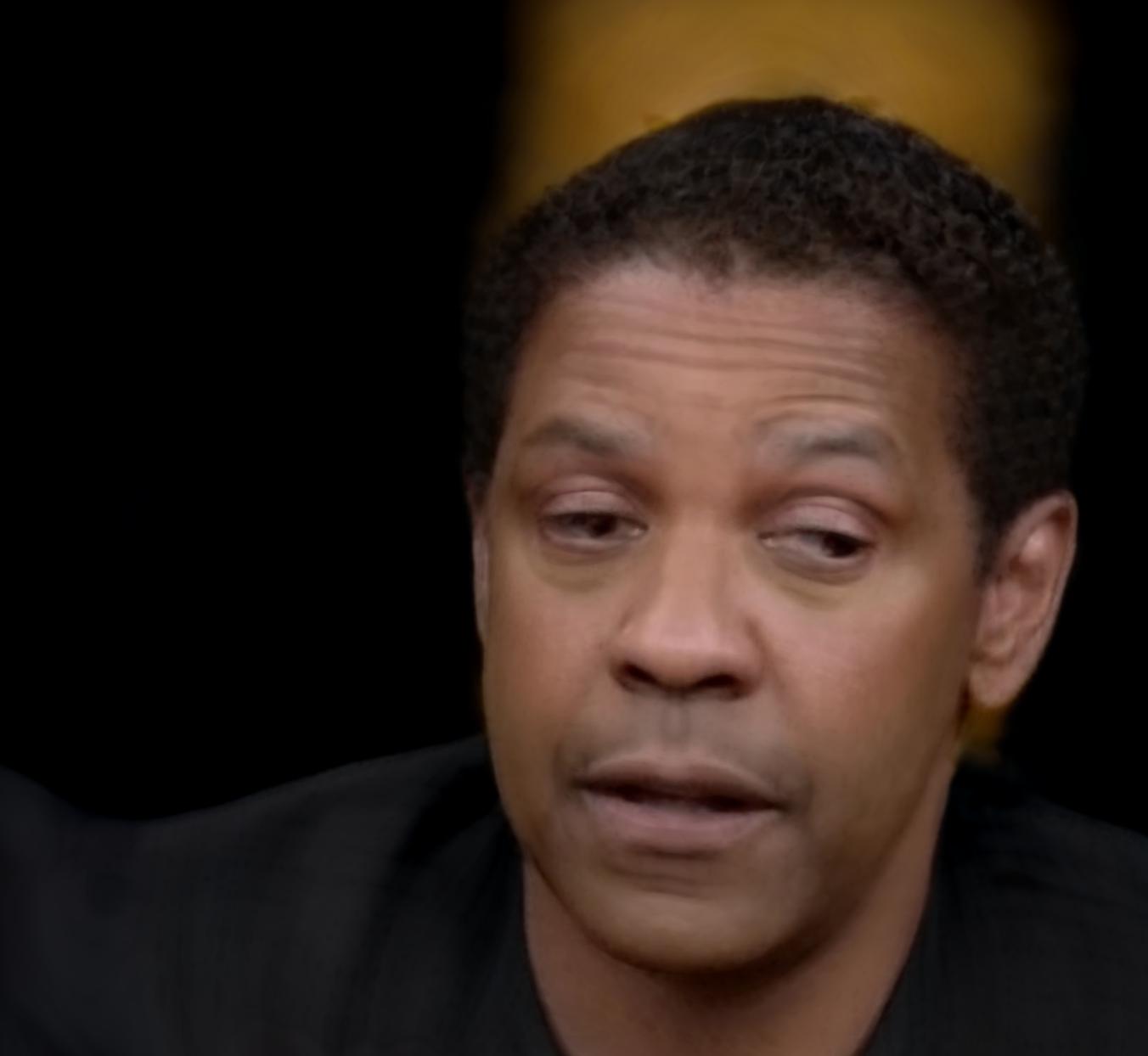} & 
        \includegraphics[width=0.18\columnwidth]{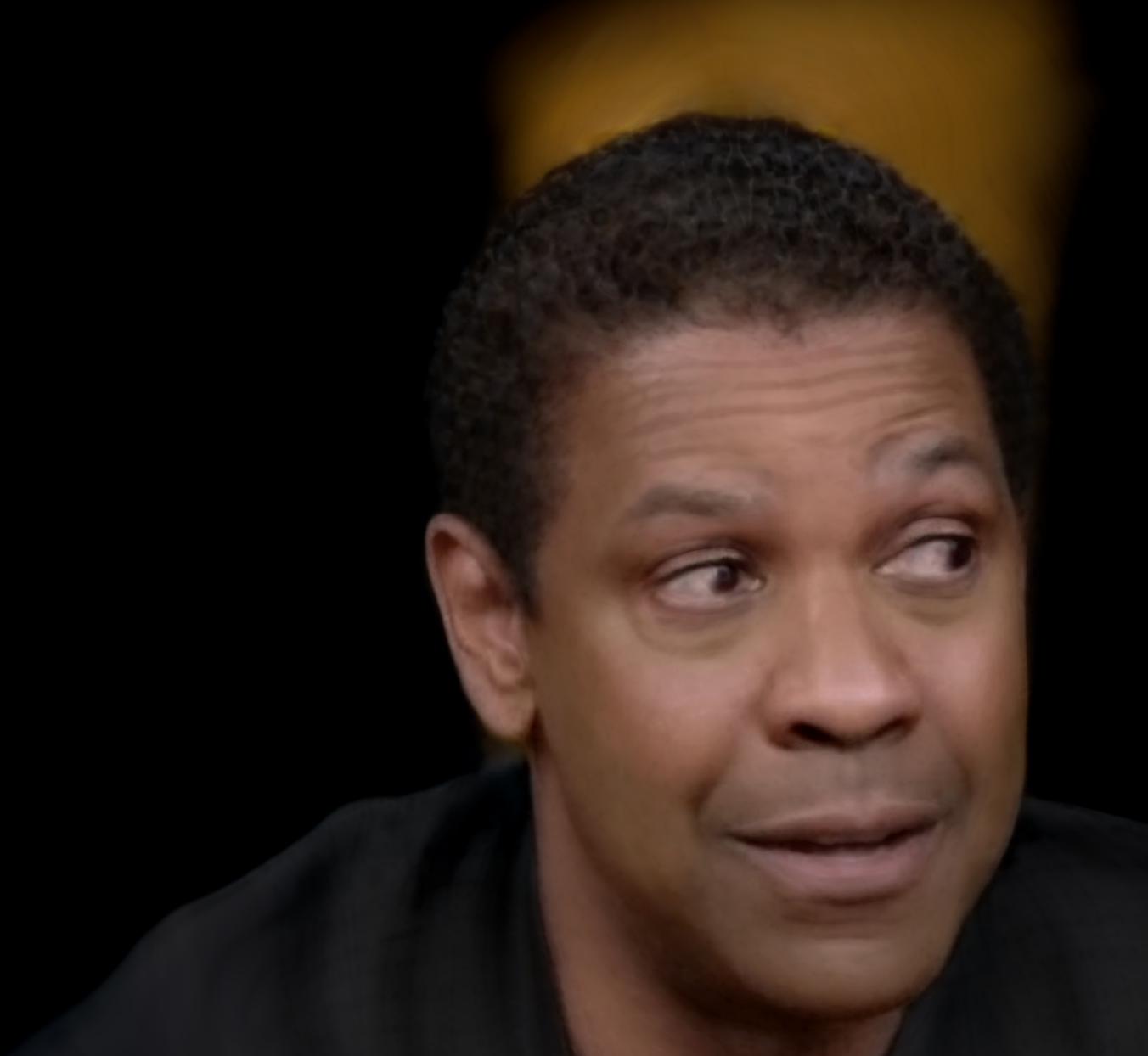} & 
        \includegraphics[width=0.18\columnwidth]{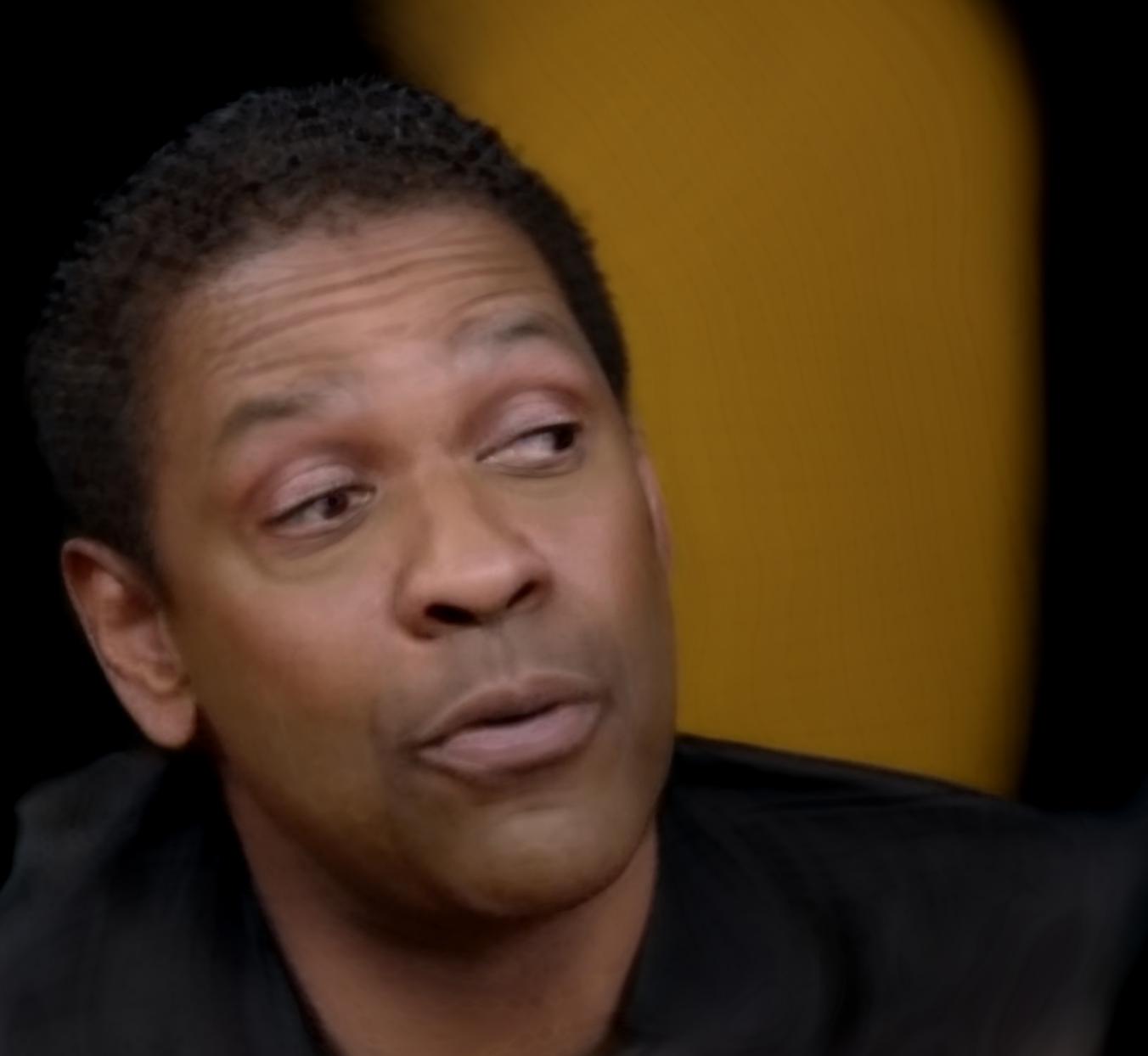} & 
        \includegraphics[width=0.18\columnwidth]{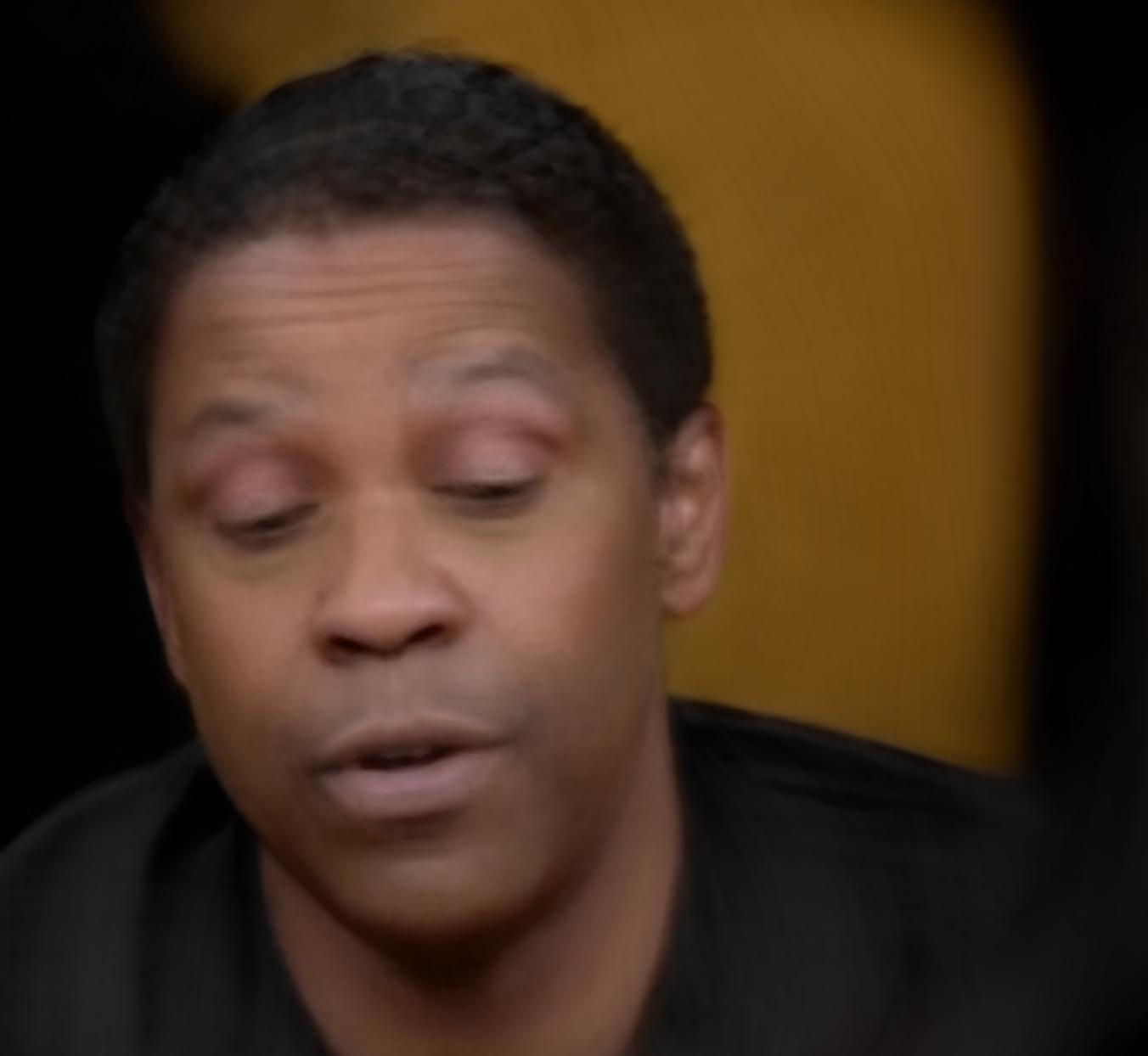} \\

	\end{tabular}
	}
	
	\vspace{-0.35cm}
	\caption{
	Video reconstruction results obtained using our field of view expansion technique. In row $2$ we provide the original unaligned reconstructions while in row $3$ we provide the expanded video reconstruction. 
	}
    \vspace{-0.425cm}
	\label{fig:wide_results}
\end{figure}

%% file: conclusion.tex
\section{Conclusions}
In this work, we have explored the competence of StyleGAN3, wondering whether indeed ``the third time is the charm''. We feel the answer is still unclear and more research may be required for a definite answer. On the one hand, the ability of StyleGAN3 to control the translation and rotation of generated images opens new intriguing opportunities. 
The prominent example explored is the generative field of view expansion, which allows one to apply StyleGAN editing on cropped video frames in a more consistent manner, alleviating the need for cumbersome and challenging seamless stitching.

On the other hand, the benefits of StyleGAN3 do come with limitations. Generally speaking, its latent space is somewhat more entangled than that of its predecessors. This makes the inversion task more challenging, affecting the robustness of frame inversions along a video. We have shown that this may be alleviated by applying the inversion on aligned images and exploiting the transformation control to compensate for the alignments. Moreover, as we have shown, training the encoder solely on aligned images does not introduce additional overheads, and even gains higher-quality synthesis.

We have naturally focused on facial images and videos. More research is required to investigate the power of StyleGAN3 for other domains. In particular, avoiding texture-sticking may be significant in videos of outdoor scenes containing high-frequency textures, like foliage or water streams. Another intriguing direction is to consider an encoder architecture that mirrors the StyleGAN3 generator, and might also employ $1\!\times\!1$ convolutions and Fourier features.

%% file: appendix.tex
\section{Background and Related Works}~\label{sec:related_appendix}

\subsection{StyleGAN Latent Spaces}

In StyleGAN2 numerous latent spaces have extensively been explored~\cite{abdal2019image2stylegan,abdal2020image2stylegan++,wu2021stylespace, Zhu2020ImprovedSE,zhu2021barbershop}, and were shown to be semantically-rich and disentangled. 
Each of the commonly-used spaces --- $\mathcal{W}, \mathcal{W+}$, and $\mathcal{S}$ --- is disentangled by a different degree and may therefore be better suited for different tasks. 
The $\mathcal{W}$ latent space, obtained from StyleGAN's mapping function, has been shown \cite{karras2020analyzing} to be more disentangled than the original $\mathcal{Z}$ space, and is therefore better suited for image editing~\cite{tov2021designing,roich2021pivotal,Zhu2020ImprovedSE}.
To edit real images, an extension of $\mathcal{W}$ is needed. Commonly, $\mathcal{W+}$~\cite{abdal2019image2stylegan} is used for StyleGAN inversion, in which a different latent code is inserted into each layer of the synthesis network.
While $\mathcal{W}$ offers disentangled editing capabilities, $\mathcal{S}$ \cite{wu2021stylespace} has been shown to be even more disentangled. 

\subsection{Editing Images with StyleGAN2}
Owing to these rich latent spaces, StyleGAN2~\cite{karras2020analyzing} has been heavily studied for achieving diverse edits over images. Early works focused on fully supervised techniques using semantic labels and facial priors~\cite{shen2020interpreting,abdal2020styleflow,goetschalckx2019ganalyze,tewari2020pie,tewari2020stylerig,hou2020guidedstyle,wu2021stylespace} for discovering latent directions. To reduce supervision, others have explored both unsupervised approaches~\cite{harkonen2020ganspace,shen2020closedform,wang2021a,bau2020units,alharbi2020disentangled} and self-supervised approaches~\cite{jahanian2019steerability,plumerault2020controlling,voynov2020unsupervised}. To achieve more fine-grained control, many works have explored the mixing of latent codes~\cite{kafri2021stylefusion,collins2020editing,Chong_2021_ICCV,shoshan2021gan} and local-based editing via semantic maps~\cite{ling2021editgan,abdel2021barbershop} or reference images~\cite{lewis2020vogue,kim2021stylemapgan}. Finally, to achieve text-based editing some have leveraged powerful contrast language-image (CLIP) models~\cite{patashnik2021styleclip,gal2021stylegannada,chefer2021imagebased,abdal2021clip2stylegan,bau2021paint}. 

\subsection{Inverting Images with StyleGAN2}
To apply the aforementioned editing techniques on real images, one must first achieve an accurate inversion of the GAN~\cite{abdal2019image2stylegan,zhu2016generative}. Inversion methods typically directly optimize the latent vector to minimize the reconstruction error of a given image~\cite{lipton2017precise,creswell2018inverting,abdal2019image2stylegan,abdal2020image2stylegan++,semantic2019bau,zhu2020improved,zhu2016generative,yeh2017semantic,gu2020image}, train an encoder to map an image to its latent representation~\cite{perarnau2016invertible, luo2017learning, guan2020collaborative,pidhorskyi2020adversarial,richardson2020encoding,tov2021designing,alaluf2021restyle,kang2021gan,kim2021exploiting,wang2021HFGI}, or design a hybrid approach combining both. While most techniques keep the generator fixed, recent works have proposed tuning the generator to achieve more accurate image inversions, either via a per-image optimization~\cite{semantic2019bau,roich2021pivotal} or a learned hypernetwork~\cite{alaluf2021hyperstyle,dinh2021hyperinverter}. 

\section{Additional Analysis}

\input{figures/nada_editing}

\subsection{Preservation of Properties Under Fine-Tuning}
Multiple works leverage the fine-tuning of StyleGAN2 generators for various applications, such as image-to-image translation~\cite{gal2021stylegannada,wu2021stylealign,zhu2021mind,pinkney2020resolution,song2021agilegan} and inversion~\cite{roich2021pivotal,semantic2019bau,alaluf2021hyperstyle}. These works show that under certain conditions, the fine-tuned generator (\emph{child}) faithfully preserves key properties of the original generator (\emph{parent}), due to the alignment between their latent spaces. In other words, the semantics of latent codes and directions in the latent spaces are often unchanged.
Here, we examine whether this preservation holds true in StyleGAN3. As shown in \cref{fig:StyleganNADA-edits}, one can use the same editing directions on various generators fine-tuned with StyleGAN-NADA~\cite{gal2021stylegannada}, indicating that this latent space alignment is indeed retained. Additional generations and editing results obtained with StyleGAN-NADA are illustrated in \cref{fig:stylegan_nada_adaptation_supplementary,fig:stylegan_nada_editing_adaptation_supp}, respectively.

\subsection{Disentanglement Analysis of \texorpdfstring{$\mathcal{W}+$}{W+}}
In \cref{sec:analysis-disentanglement}, we performed a quantitative analysis of the disentanglement of the different latent spaces of StyleGAN3 and computed the DCI metrics for each space. 
Recall, \textit{disentanglement} measures the extent to which each latent channel controls a single attribute, while \textit{completeness} measures the degree to which each attribute is controlled by a single latent channel. Finally, \textit{informativeness} assesses the accuracy of the attribute classifiers for a given latent representation.
It is of interest to quantify the disentanglement of the $\mathcal{W}+$ latent space, often used for inversion. However, we find that randomly generated images in $\mathcal{W}+$ are unnatural, making the computation of the DCI metric unreliable. In \cref{fig:wplus_analysis}, we compare a collection of uncurated samples from $\mathcal{W}+$ for both StyleGAN2 and StyleGAN3 generators trained on aligned facial images. As shown, while both sets of images are unrealistic, those of StyleGAN3 contain significantly more artifacts. As such, we choose to omit the computation of the DCI metrics of $\mathcal{W}+$. 

\input{figures/supplementary/wplus_analysis}

\section{Landmarks-Based Transformations}~\label{sec:computing_transforms}
As described in \cref{sec:inversion}, our StyleGAN3 encoders are trained solely on aligned images. To invert a given unaligned image, we first align the image and invert the resulting image using the encoder. We then generate the unaligned image reconstruction by utilizing the user-defined transformation passed to the synthesis network along with the inverted latent code. 

To compute the transformation triplet $(r, t_x, t_y)$ for a given unaligned image $x_{unaligned}$, we employ an off-the-self landmark detection tool~\cite{dlib09} and build on the landmark parsing procedure used for creating the FFHQ~\cite{karras2019style} dataset. 
Specifically, we first align the image to obtain an aligned version of the input, denoted by $x_{aligned}$.

We then detect the eyes of both $x_{unaligned}$ and $x_{aligned}$, and compute the rotation and the translation between the two sets. Here, the rotation is given by the angle between the lines that connect the eyes in both images. For computing translation, we rotate the aligned image and measure the vertical and horizontal distances between the left eye of the rotated aligned image and unaligned image. These three values explicitly define the user-specified transformations that are passed to the Fourier features of StyleGAN3's synthesis network. This process is illustrated in \cref{fig:predicting_transforms}.

It should be noted that not all unaligned images can be inverted. As the unaligned FFHQ StyleGAN3 generator was trained on unaligned images of a certain facial size, we can only invert faces of this size. Therefore, before computing the landmarks for a given image, we first crop it so that the face is of the size suited for StyleGAN3. This process is done similarly to the process for creating the official FFHQ-U dataset.

\input{figures/supplementary/transforms_prediction}

\section{Editing: Additional Details}
\paragraph{\textbf{InterFaceGAN. }}
For editing images in $\mathcal{W}$, we use the official implementation of InterFaceGAN~\cite{shen2020interpreting} for training the linear boundaries using off-the-shelf classifiers. We apply HopeNet~\cite{Ruiz_2018_CVPR_Workshops} for pose, Rothe~\etal~\cite{Rothe-ICCVW-2015,Rothe-IJCV-2018} for age, and the classifier from Lin~\etal~\cite{lin2021anycost} for the remaining attributes.

\section{StyleGAN3 Encoding Scheme}
\subsection{Additional Training Details}~\label{sec:encoder_details}
Our encoders for inverting StyleGAN3 are based on the pSp~\cite{richardson2020encoding} and e4e~\cite{tov2021designing} encoding schemes. We apply the same encoder architectures as those used for inverting StyleGAN2 generators. Following our insights that a pre-trained aligned StyleGAN3 generator can synthesize both aligned and unaligned images, our encoders are trained solely on aligned images, significantly simplifying the training objective of the encoder.

Due to the larger memory consumption required by StyleGAN3, our encoders are trained using a batch size of $2$. To match the batch size used in the official implementations of the StyleGAN2 pSp and e4e encoders, we apply gradient accumulation to attain an effective batch size of $8$ (i.e., an optimization step is performed every four batches). All encoders are trained using a single NVIDIA P40 GPU.

Training is performed using the same set of losses as used to train the StyleGAN2 encoders. Specifically, we use a weighted combination of the $L_2$ pixel-wise loss, the LPIPS~\cite{zhang2018perceptual} perceptual loss, and an identity-based reconstruction~\cite{deng2019arcface,richardson2020encoding}. The overall loss objective is given by: 
\begin{equation}~\label{eq:rec_loss}
    \mathcal{L}(x) =  \lambda_{l2}\mathcal{L}_{2}(x) + \lambda_{lpips}\mathcal{L}_{LPIPS}(x) + \lambda_{id}\mathcal{L}_{\text{id}}(x),
\end{equation}
where we set $\lambda_{l2}=1$, $\lambda_{lpips}=0.8$, and $\lambda_{id}=0.1$. 

For training the $\text{ReStyle}_{e4e}$, encoder we additionally remove the progressive training scheme used in the official implementation. Instead, all $16$ latent codes are predicted simultaneously by the encoder from the start of training. We find this leads to faster convergence. 

\vspace{0.2cm}
\subsection{Baselines and Comparisons} 
In our work we compare our $\text{ReStyle}_{pSp}$ and $\text{ReStyle}_{e4e}$ StyleGAN3 encoders with their StyleGAN2 counterparts from Alaluf~\etal~\cite{alaluf2021restyle}. All encoders are trained on the aligned FFHQ dataset~\cite{karras2019style} consisting of $70,000$ images. For a quantitative comparison of the encoders, we compute the reconstruction metrics on the aligned CelebA-HQ~\cite{karras2017progressive,liu2015deep} test set. Finally, for our qualitative images, we display the aligned outputs for StyleGAN2 encoders and the unaligned outputs for the StyleGAN3 encoders.

\subsection{Ablation Study}~\label{sec:ablation_study}
\vspace{-0.8cm}
\paragraph{\textbf{Designing an Encoder for Unaligned Generators. }}
If one were to design an encoder for encoding unaligned images into StyleGAN3's latent space, a natural first attempt at doing so would be to train an encoder on unaligned images using existing encoding schemes for inverting an unaligned generator. Specifically, assume we have pairs of images $ \{ ( x^{i}_{aligned},~x^{i}_{unaligned} ) \}_{i=1}^{N}$, we can train the encoder to solve the following objective: 
\begin{align}~\label{eq:encoding_objective}
\sum_{i=1}^N \; \mathcal{L} ( x^i_{unaligned}, G ( E ( x^i_{unaligned} ) ) ),
\end{align}
where $G$ is a pre-trained $\textit{unaligned}$ generator.

Yet, an immediate challenge arises: employing the identity loss, which incorporates a pre-trained facial recognition network~\cite{deng2019arcface}, is non-trivial. This facial recognition network is trained on images that are aligned and cropped to the inner facial region. Hence, applying this network on unaligned images may lead to unpredictable results. One may mitigate this by using off-the-shelf facial detectors~\cite{zhang2016joint,dlib09} to detect and align the images, but applying these networks during training is impractical. In Richardson~\etal~\cite{richardson2020encoding}, the authors overcome this challenge by using a heuristic that roughly crops the face before passing the image through the facial recognition network. Yet, they assume the images are pre-aligned. When training on unaligned images, as in our case, this heuristic is no longer applicable. 

To overcome this challenge we can perform the pseudo-alignment trick described in \cref{sec:preliminaries}. Consider the latent code $w = (w_0,w_1,...,w_{15})$  outputted by our encoder for some unaligned input $x_{unaligned}$. We can replace the first latent code $w_0$ with the generator's average latent code to obtain the pseudo-aligned latent representation $w_{aligned} = (\overline{w},w_1,...,w_{15})$, corresponding to the pseudo-aligned image $y_{aligned} = G(w_{aligned}; (0,0, 0))$.  

Given the pseudo-aligned image, we are now more accurately able to compute the identity loss between $x_{aligned}$ and $y_{aligned}$. Additionally, we may compute the $L_2$ and LPIPS reconstruction losses between the original the original unaligned image $x_{unaligned}$ and the unaligned reconstruction $y_{unaligned} = G(w; (0,0), 0)$. 

\input{figures/supplementary/ablation_unaligned_encoder}

\paragraph{\textbf{Qualitative Comparisons. }}
We now compare the unaligned encoding scheme above to our aligned scheme presented in \cref{sec:inversion}. As illustrated in \cref{fig:ablation_unaligned_encoder}, our scheme achieves superior reconstructions compared to the unaligned version. We attribute this improvement to the simpler training task of our aligned encoder: rather than needing to capture \textit{both} the input identity and position, our encoder can focus on reconstructing only the former with the desired pose provided via the user-defined transformations predicted using the procedure described in \cref{sec:computing_transforms}.

\section{Video Inversion Scheme: Additional Details}~\label{sec:video_details}
\vspace{-0.3cm}
\subsection{Video Preprocessing}
As mentioned in \cref{sec:video}, before feeding a given frame to our encoder, we align and crop the frame. The landmark detector used for aligning the image and computing the Fourier features transformations uses either the distance between the two eyes or the distances between the eyes and the mouth. Since the distance between the eyes and mouth may change along the video, we choose to use only the distance between the eyes for all frames for this. We find that doing so gives a slightly move stable video reconstruction. Note, that the cropping procedure assumes the distance between the camera and the input face does not change along the video. 

\subsection{Pivotal Tuning of Videos}
For inverting and editing a given video we perform a per-video fine-tuning of the StyleGAN3 generator network using the pivotal tuning technique from Roich~\etal~\cite{roich2021pivotal}. For each video, training is performed for a total of $8,000$ optimization steps with a batch size of $2$ using the $L_2$ pixel-wise loss and the LPIPS~\cite{roich2021pivotal} loss, both with equal weight coefficients. For example, given a video consisting of $200$ frames, each frame is observed an average of $40$ times during training. During training, we do not alter the weights of the input Fourier features layer of the generator.

\subsection{Latent Vector Smoothing}
We refer the reader to the accompanying video results for a comparison of video inversions and reconstructions with and without the latent vector smoothing operation. As can be seen, when no latent smoothing is performed, the resulting video is unstable due to the pre-alignment step and the per-frame inversions of the encoder.

\subsection{Field-of-View Expansion}~\label{sec:fov_expansion}
In the main paper, we describe how to expand the field of view when reconstructing and editing a given input frame. In the overview example, we performed an expansion toward a single direction (e.g., extending the top of the video). Yet, it is also possible to extend the field-of-view toward \textit{multiple} directions. For each direction we wish to expand the image in, we generate another image shifted toward that direction. For example, for expanding an image to the right by $\Delta$ uses the transformations parameters $(0,-\Delta,0)$ while expanding an image at the bottom uses parameters $(0,0,-\Delta)$. Given the generated image for each direction, we then copy the non-overlapping parts from the sifted images and join them with the original reconstructed image.
Note, for cases where we wish to expand an image both horizontally and vertically, we add an additional shift in both directions for filling in the corner regions. We demonstrate results of this field-of-view expansion in \cref{fig:video_results_wide_supplementary}.

\setlength{\parindent}{0pt}
\section{Additional Results}~\label{sec:additional_results}
We provide additional results and comparisons, as follows:
\begin{enumerate}
    \item \cref{fig:styleclip-edits-supp} demonstrates non-linear editing performed in the $\mathcal{W}+$ latent space of an aligned StyleGAN3 generator using StyleCLIP's mapping technique~\cite{patashnik2021styleclip}. 
    \item \cref{fig:styleclip-global-directions-supplementary,fig:edit_afhq_supp,fig:edit_landscapes_supp} illustrate edits obtained across various domains using StyleCLIP's~\cite{patashnik2021styleclip} global editing technique applied in StyleGAN3's $\mathcal{S}$ space. 
    \item In \cref{fig:stylegan_nada_adaptation_supplementary}, we demonstrate domain adaptation results obtained by fine-tuning pre-trained StyleGAN3 generators using StyleGAN-NADA~\cite{gal2021stylegannada} across various domains.
    \item \cref{fig:stylegan_nada_editing_adaptation_supp} shows editing results obtained over various fine-tuned child generators showing that the alignment of latent spaces is preserved under the fine-tuning of StyleGAN3.
    \item \cref{fig:inversions_supplementary,fig:inversions_supplementary_2} provide additional reconstruction comparisons between our StyleGAN3 encoders and StyleGAN2 encoders from Alaluf~\etal~\cite{alaluf2021restyle}. 
    \item \cref{fig:editing_reals_supplementary} shows additional editing results obtained over real images using the editing techniques from Shen~\etal~\cite{shen2020interpreting} and Patashnik~\etal~\cite{patashnik2021styleclip}.
    \item \cref{fig:video_results_supplementary_1,fig:video_results_supplementary_2} show video reconstruction and editing results obtained with our full StyleGAN3 encoding pipeline. In addition, we demonstrate domain adaptation results applied over the edited videos, allowing us to generate edited videos in various styles. 
    \item \cref{fig:video_results_wide_supplementary} demonstrates our field-of-view expansion technique on multiple video sequences.
    \item Finally, we invite the reader to visit our project page where we provide full video reconstructions and edits on various inputs. 

\end{enumerate}

\clearpage

\input{figures/supplementary/styleclip_mapper}
\input{figures/supplementary/styleclip_edit_ffhq}
\input{figures/supplementary/edits_afhq}
\input{figures/supplementary/edits_lands}

\input{figures/supplementary/nada_images}
\input{figures/supplementary/nada_editing}

\input{figures/supplementary/inversion_comparison}
\input{figures/supplementary/inversion_comparison_2}
\input{figures/supplementary/edits_reals_comparison}

\input{figures/supplementary/video_results}
\input{figures/supplementary/video_results_2}
\input{figures/supplementary/video_results_wide}

%% file: figures/nada_editing.tex
\begin{figure}[tb]
	\centering
	\setlength{\tabcolsep}{1.5pt}	

	{\small
	\begin{tabular}{c c@{} c c@{} c c@{} c}
	
		\raisebox{0.125in}{\rotatebox{90}{Pixar}} &
		\includegraphics[width=0.155\columnwidth]{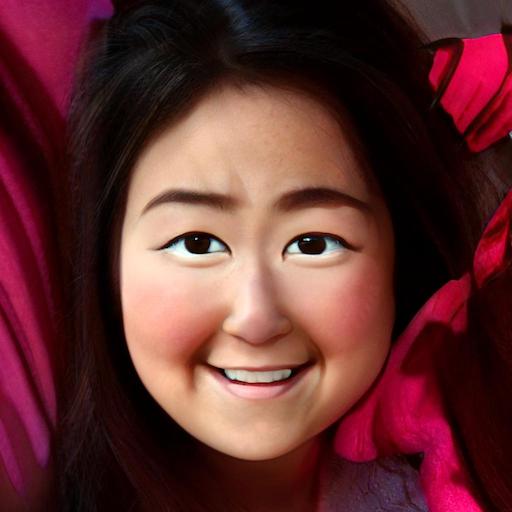} & 
		\includegraphics[width=0.155\columnwidth]{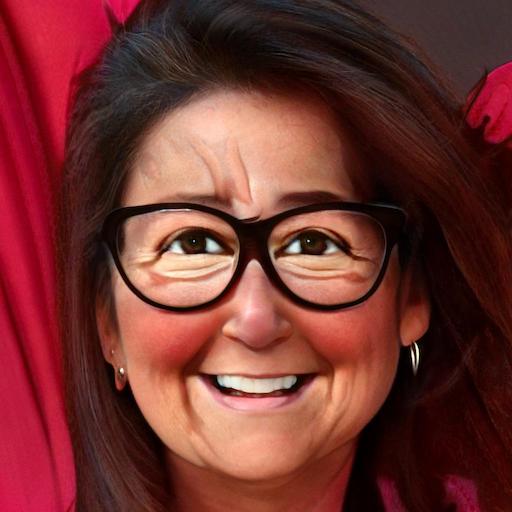} & 
		\includegraphics[width=0.155\columnwidth]{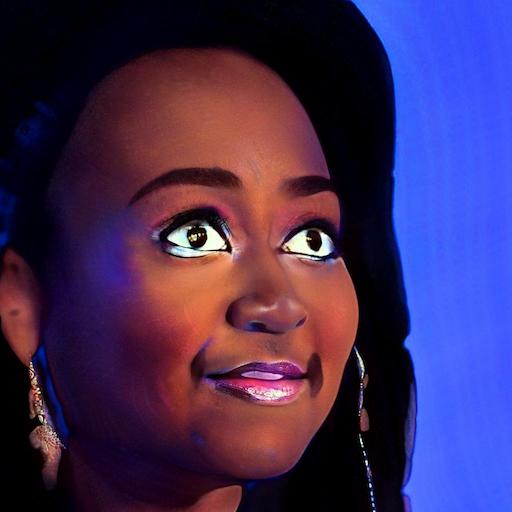} & 
		\includegraphics[width=0.155\columnwidth]{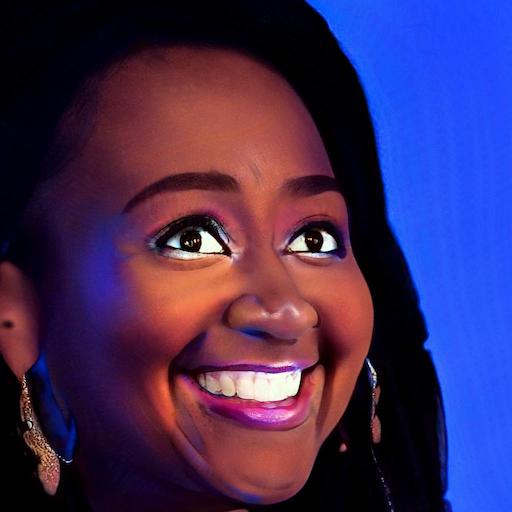} & 
		\includegraphics[width=0.155\columnwidth]{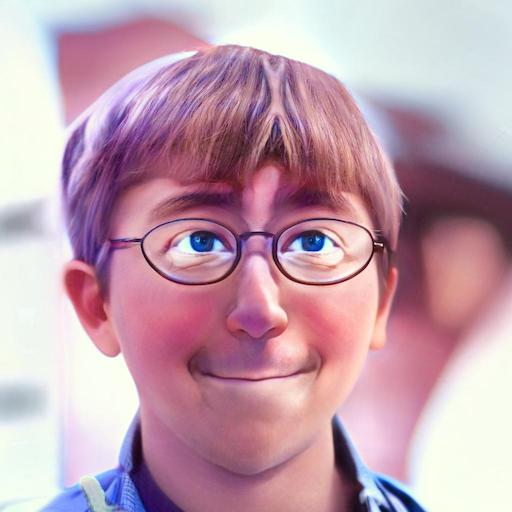} & 
		\includegraphics[width=0.155\columnwidth]{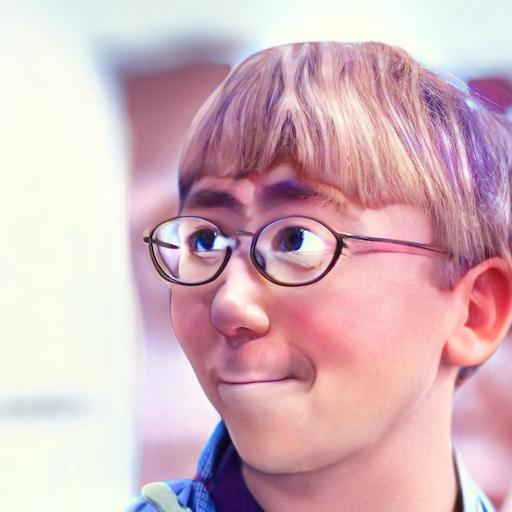} \\ 

		\raisebox{0.1in}{\rotatebox{90}{Sketch}} &
		\includegraphics[width=0.155\columnwidth]{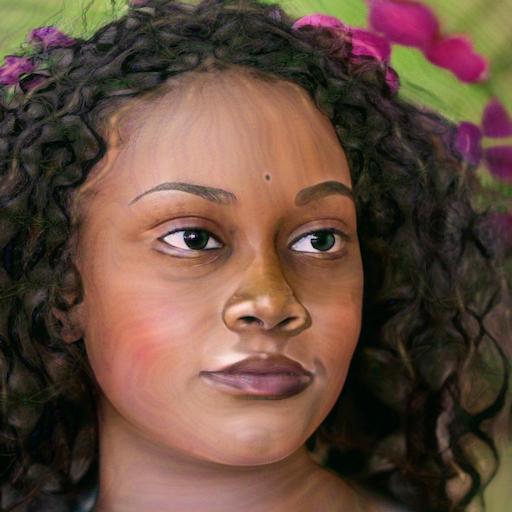} & 
		\includegraphics[width=0.155\columnwidth]{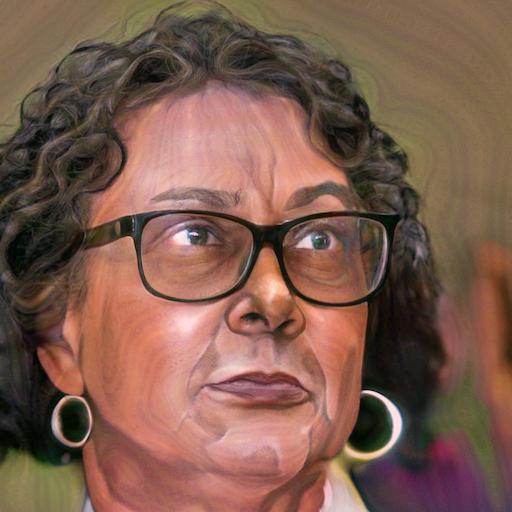} & 
		\includegraphics[width=0.155\columnwidth]{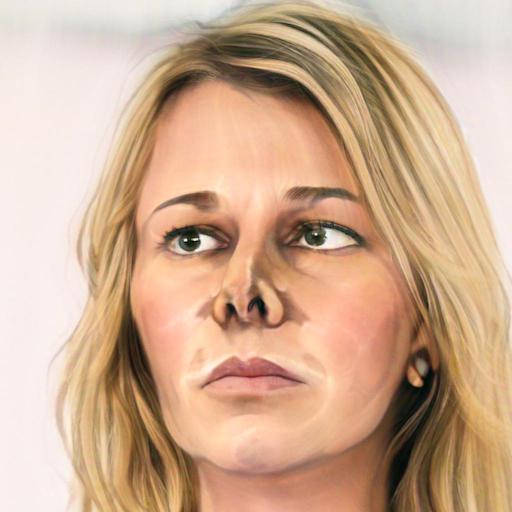} & 
		\includegraphics[width=0.155\columnwidth]{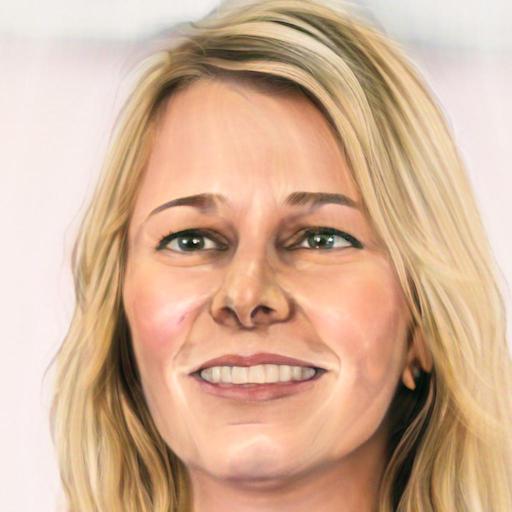} & 
		\includegraphics[width=0.155\columnwidth]{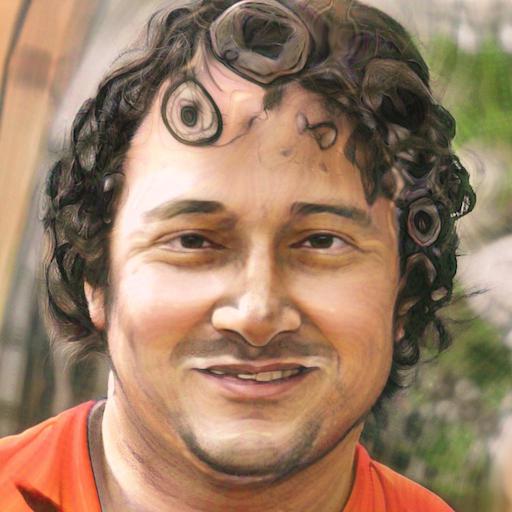} & 
		\includegraphics[width=0.155\columnwidth]{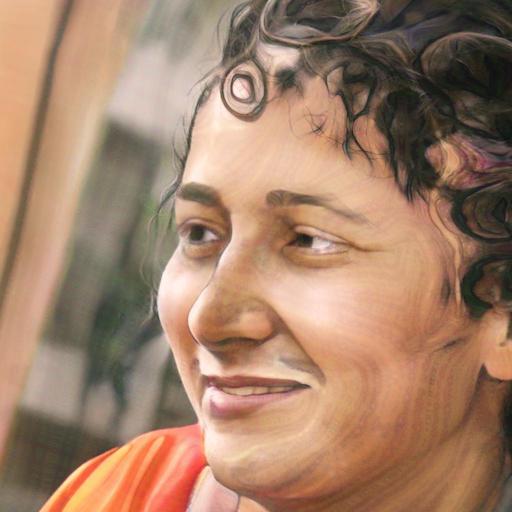} 
		\\

        & \mc{Age} & \mc{Smile} & \mc{Pose} \\

	\end{tabular}
	}
	
	\vspace{-0.35cm}
	\caption{Editing using fine-tuned generators. The editing directions in the latent space of the parent (FFHQ) StyleGAN3 generator have a similar effect in the fine-tuned child generators trained with StyleGAN-NADA~\cite{gal2021stylegannada}, indicating the latent spaces remain semantically aligned.}
	\vspace{-0.2cm}
	\label{fig:StyleganNADA-edits}
\end{figure}

%% file: figures/supplementary/wplus_analysis.tex
\begin{figure}[tb]
	\centering
	\setlength{\tabcolsep}{1pt}	
	{\footnotesize
	\begin{tabular}{c c c c c c c}
		\raisebox{0.025in}{\rotatebox{90}{StyleGAN2}} &
		\includegraphics[width=0.18\columnwidth]{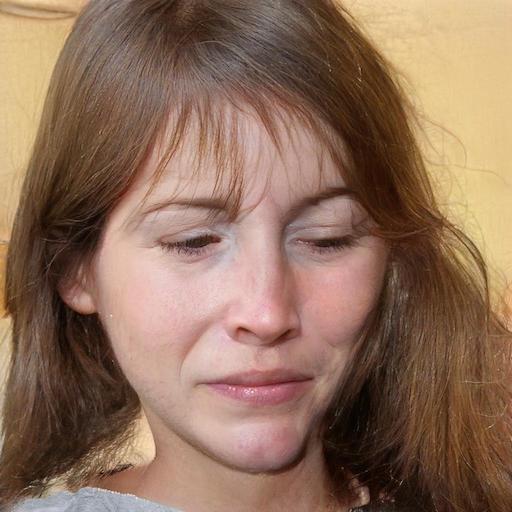} & 
		\includegraphics[width=0.18\columnwidth]{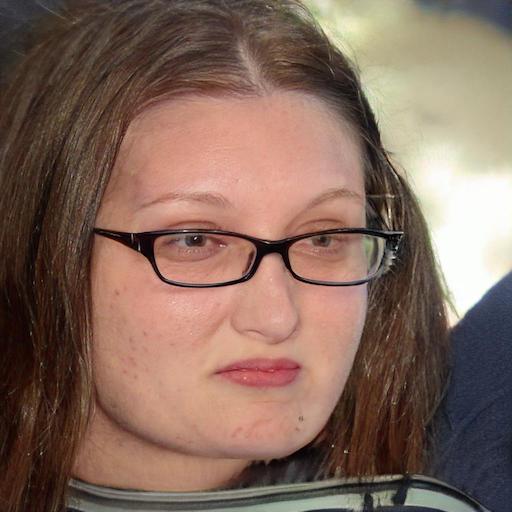} &
		\includegraphics[width=0.18\columnwidth]{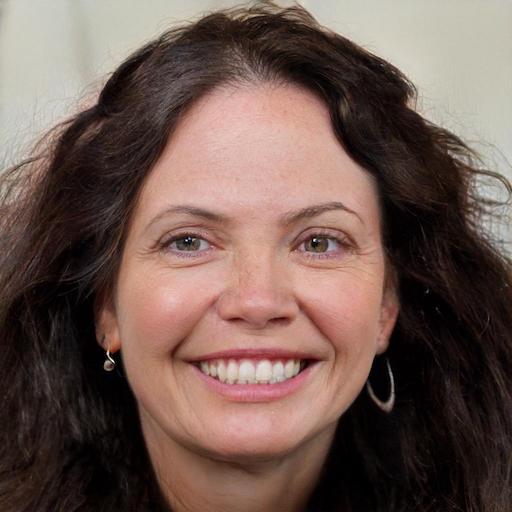} &
		\includegraphics[width=0.18\columnwidth]{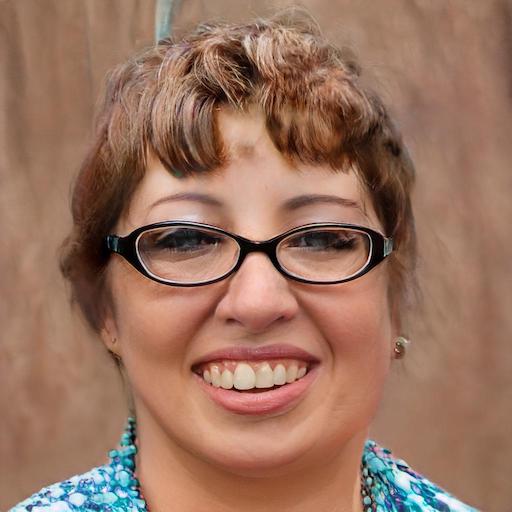} &
		\includegraphics[width=0.18\columnwidth]{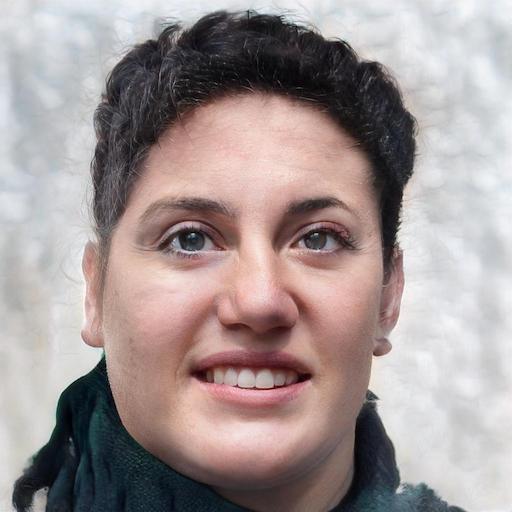} \\
		\raisebox{0.025in}{\rotatebox{90}{StyleGAN3}} &
		\includegraphics[width=0.18\columnwidth]{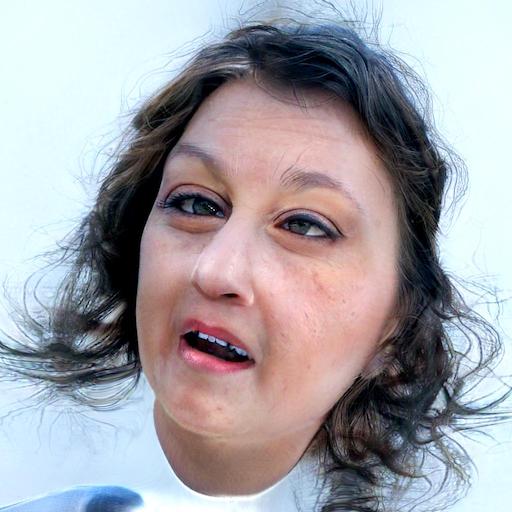} & 
		\includegraphics[width=0.18\columnwidth]{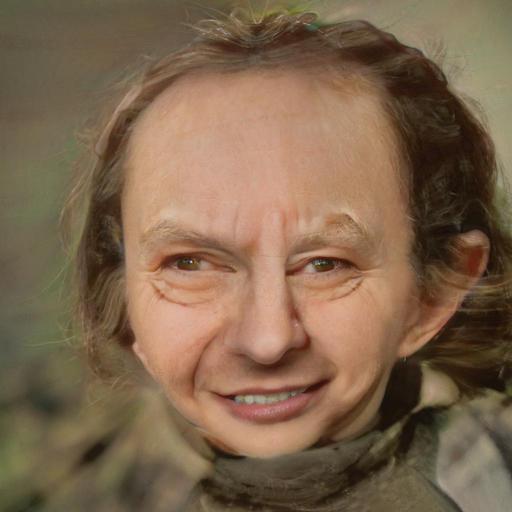} &
		\includegraphics[width=0.18\columnwidth]{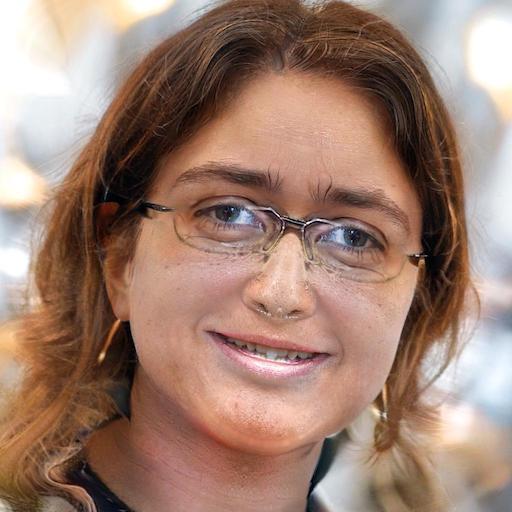} &
		\includegraphics[width=0.18\columnwidth]{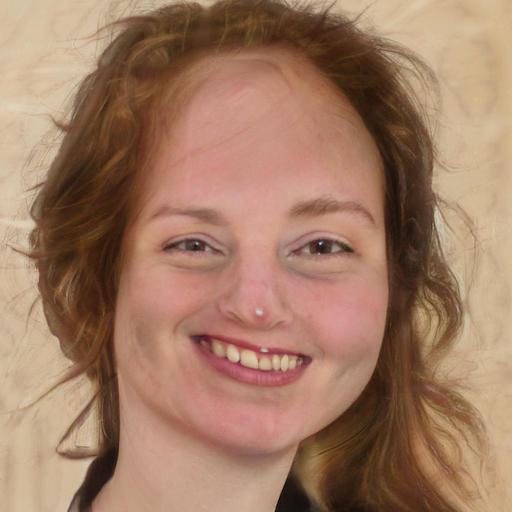} &
		\includegraphics[width=0.18\columnwidth]{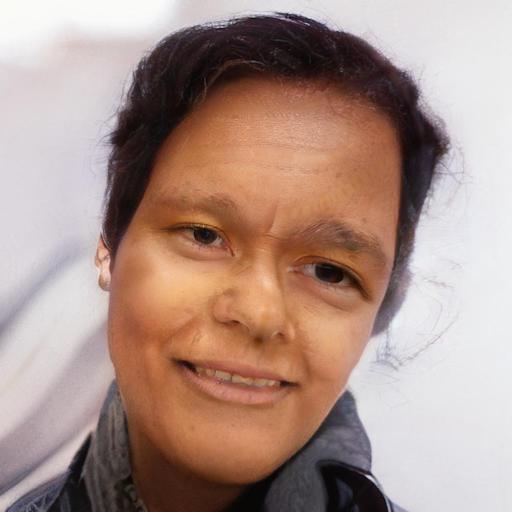} \\

	\end{tabular}
	}
	\vspace{-0.2cm}
	\caption{
		An uncurated set of images from aligned StyleGAN2 and StyleGAN3 generators generated by randomly sampling latent codes from $\mathcal{W}+$. 
    }
	\label{fig:wplus_analysis}
	\vspace{-0.4cm}
\end{figure}

%% file: figures/supplementary/transforms_prediction.tex
\begin{figure}
    \centering
    \setlength{\belowcaptionskip}{-5pt}
    \includegraphics[width=\linewidth]{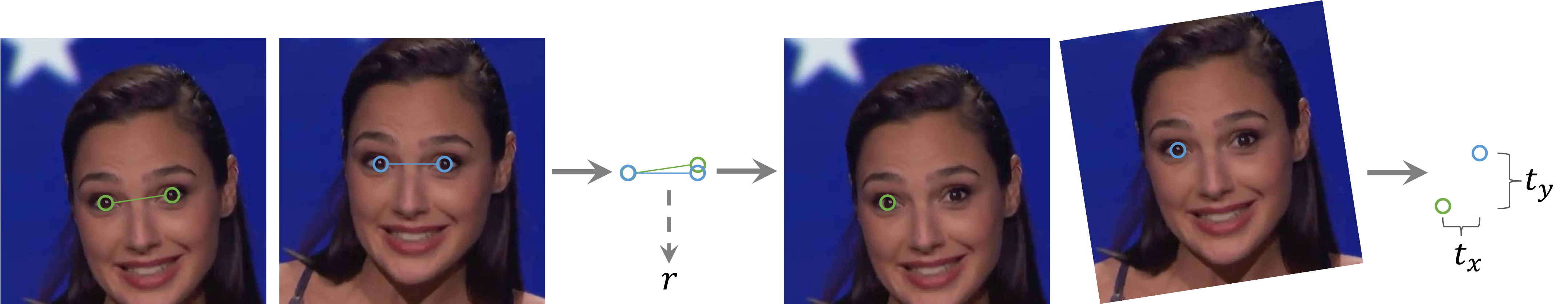}
    \vspace{-0.35cm}
    \caption{
        Predicting the user-specified transformation parameters passed to StyleGAN3's input Fourier features. Using an off-the-shelf landmark detector~\cite{dlib09} we compute the rotation $r$ and translation parameters $(t_x, t_y)$ used for controlling the generated position and in-plane rotation.
    }
    \label{fig:predicting_transforms}
\end{figure}

%% file: figures/supplementary/ablation_unaligned_encoder.tex
\begin{figure}[tb]
	\centering
	\setlength{\tabcolsep}{1pt}	
	{\small
	\begin{tabular}{c c c c c c c}
	
		\raisebox{0.125in}{\rotatebox{90}{Source}} &
		\includegraphics[width=0.18\columnwidth]{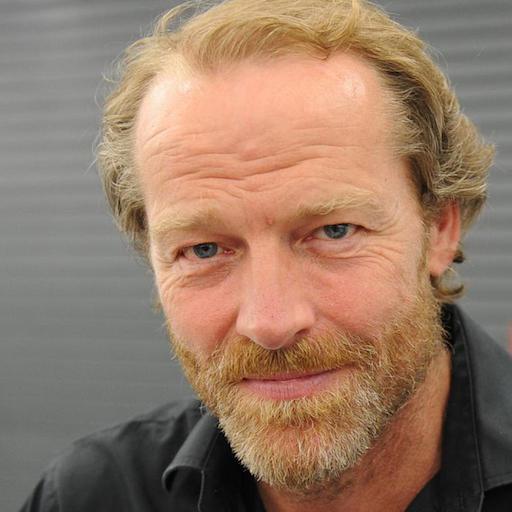} & 
		\includegraphics[width=0.18\columnwidth]{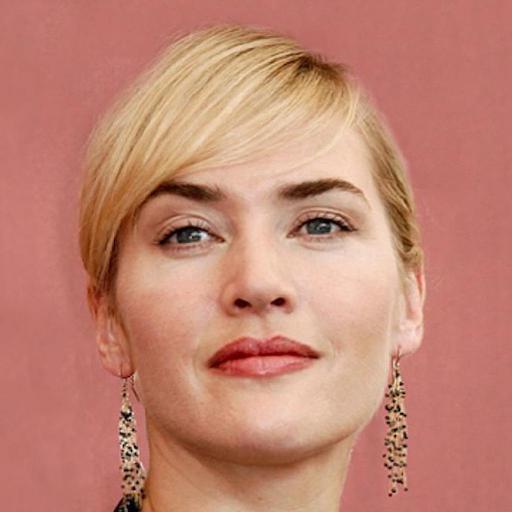} & 
		\includegraphics[width=0.18\columnwidth]{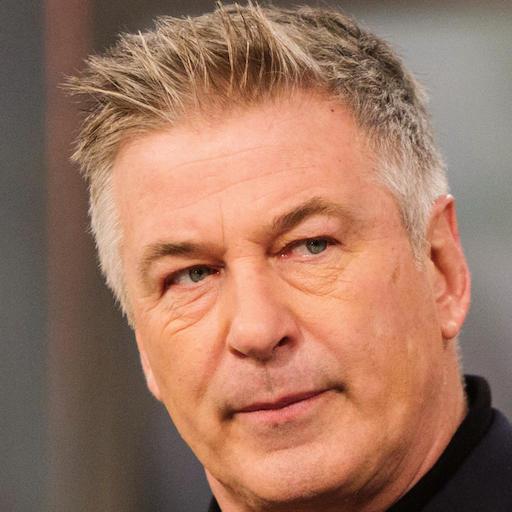} & 
		\includegraphics[width=0.18\columnwidth]{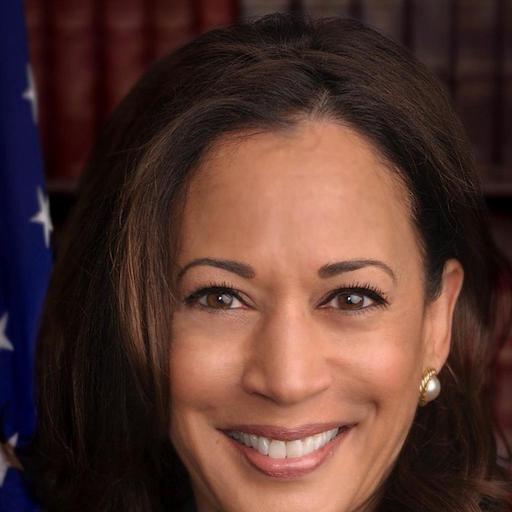} & 
		\includegraphics[width=0.18\columnwidth]{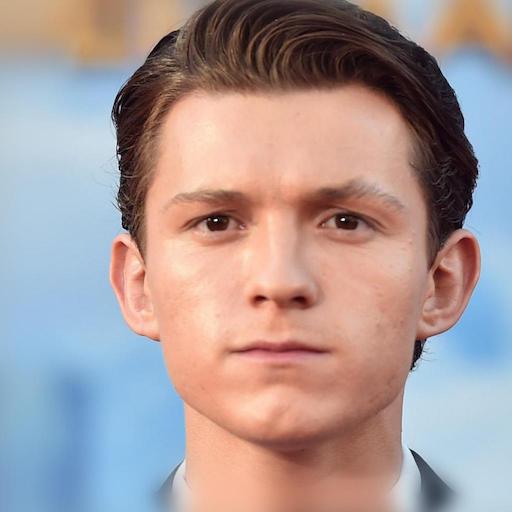} \\
		
		\raisebox{0.05in}{\rotatebox{90}{Unaligned}} &
		\includegraphics[width=0.18\columnwidth]{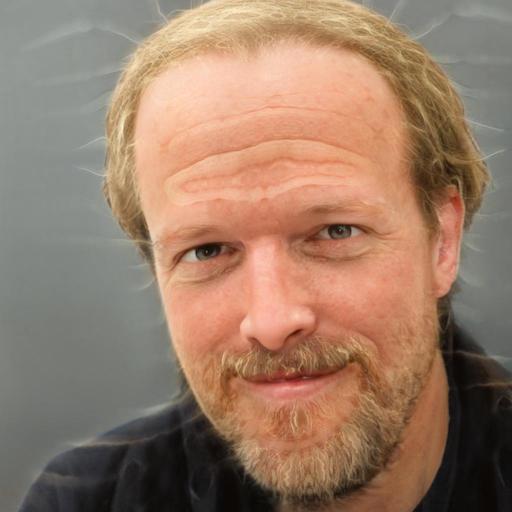} & 
		\includegraphics[width=0.18\columnwidth]{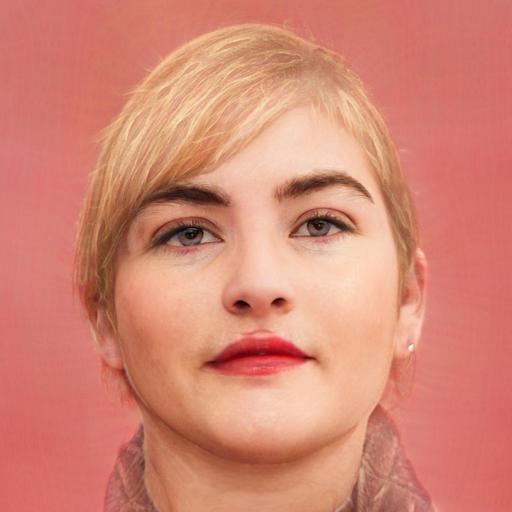} & 
		\includegraphics[width=0.18\columnwidth]{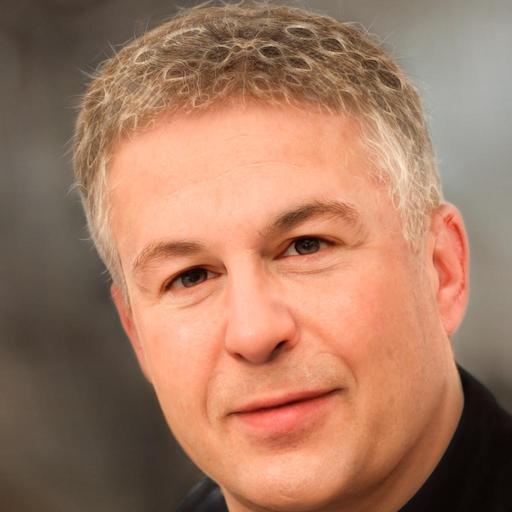} & 
		\includegraphics[width=0.18\columnwidth]{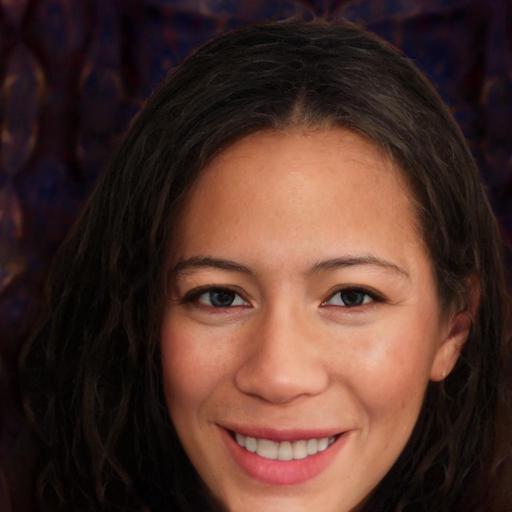} & 
		\includegraphics[width=0.18\columnwidth]{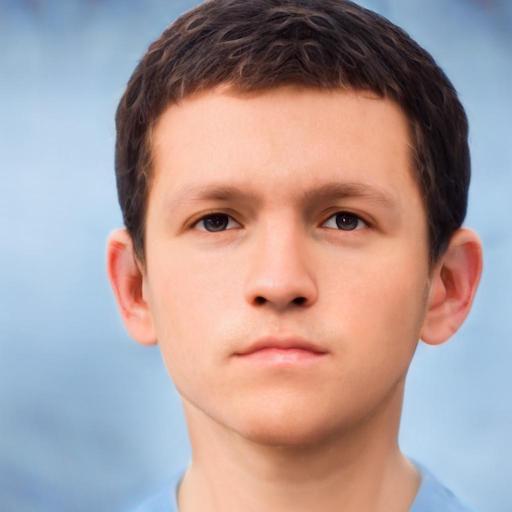} \\
		
		\raisebox{0.1in}{\rotatebox{90}{Aligned}} &
		\includegraphics[width=0.18\columnwidth]{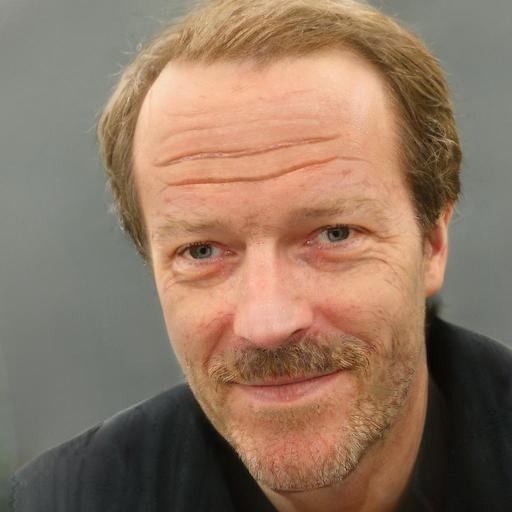} & 
		\includegraphics[width=0.18\columnwidth]{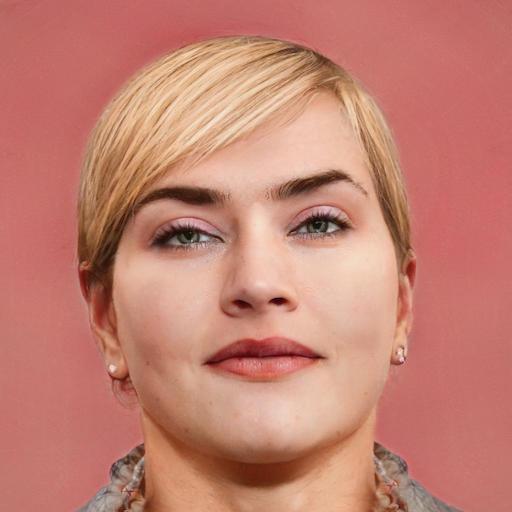} & 
		\includegraphics[width=0.18\columnwidth]{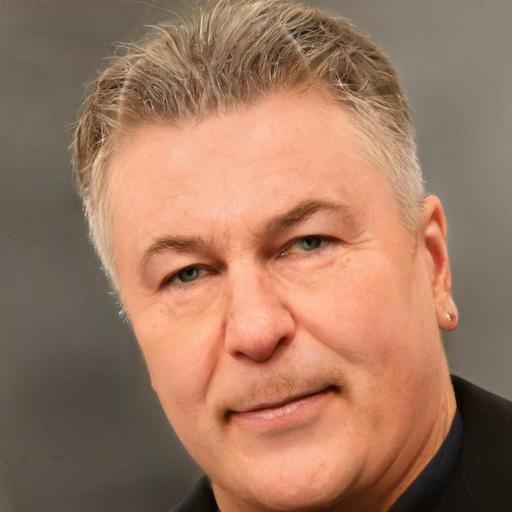} & 
		\includegraphics[width=0.18\columnwidth]{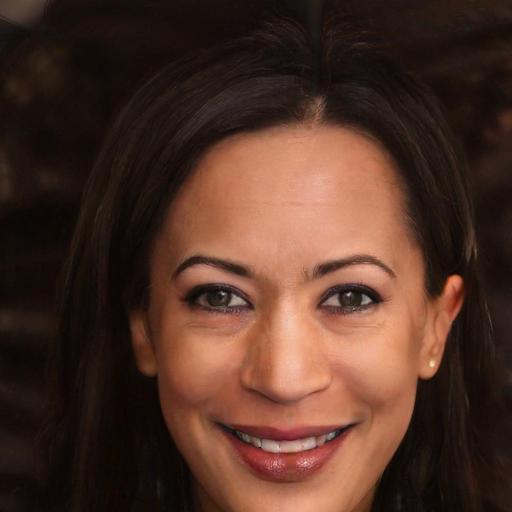} & 
		\includegraphics[width=0.18\columnwidth]{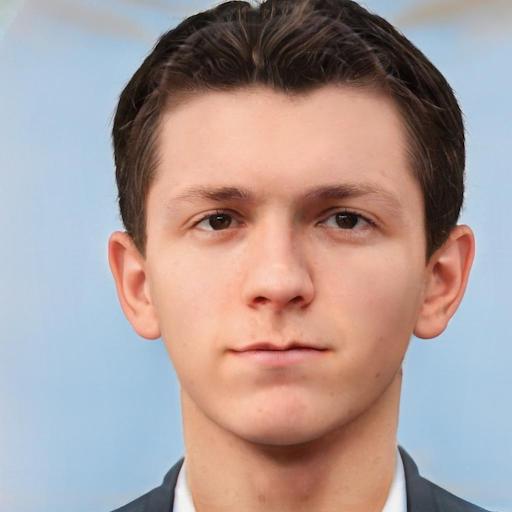} \\

	\end{tabular}
	}
	\vspace{-0.4cm}
	\caption{We compare the reconstruction quality of an encoder trained on unaligned images (row $2$) to our encoder trained on aligned images (row $3$). Our encoder achieves superior reconstructions while faithfully capturing the unaligned poses by leveraging the user-defined transformations supported by StyleGAN3. 
    }
	\label{fig:ablation_unaligned_encoder}
	\vspace{-0.2cm}
\end{figure}

%% file: figures/supplementary/styleclip_mapper.tex
\begin{figure*}[tb]
	\centering
	\setlength{\tabcolsep}{1pt}	
	{\small
	\begin{tabular}{c c c c c c}
	
	\\ \\ \\ \\
	
		\includegraphics[width=0.22\columnwidth]{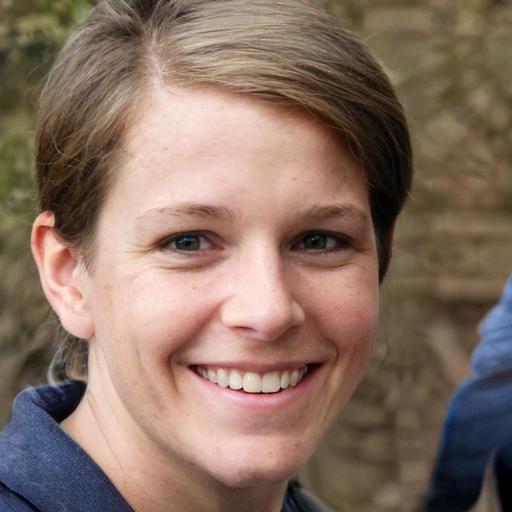} & 
		\includegraphics[width=0.22\columnwidth]{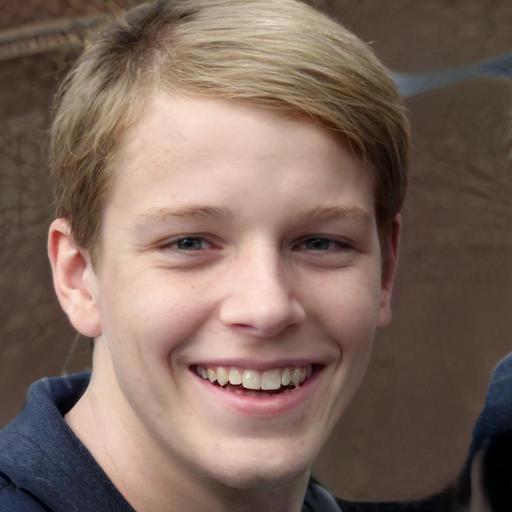} & 
		\includegraphics[width=0.22\columnwidth]{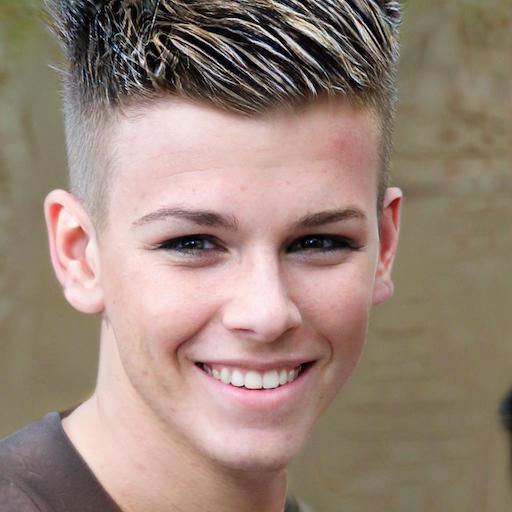} & 
		\includegraphics[width=0.22\columnwidth]{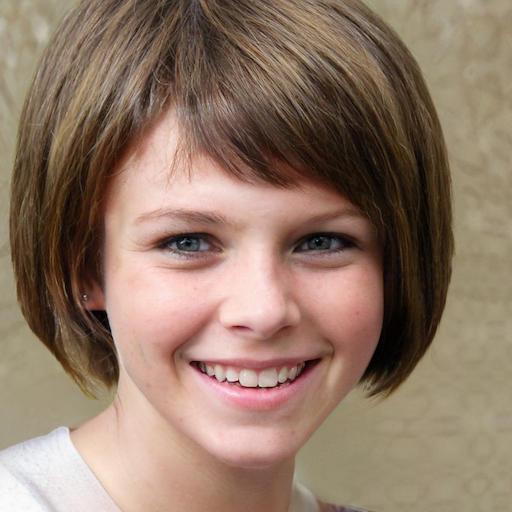} & 
		\includegraphics[width=0.22\columnwidth]{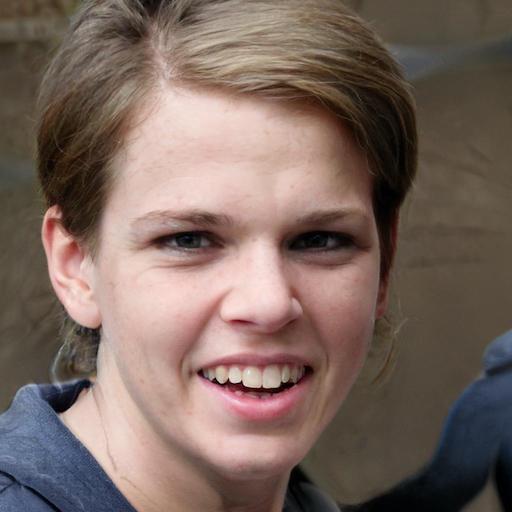} & 
		\includegraphics[width=0.22\columnwidth]{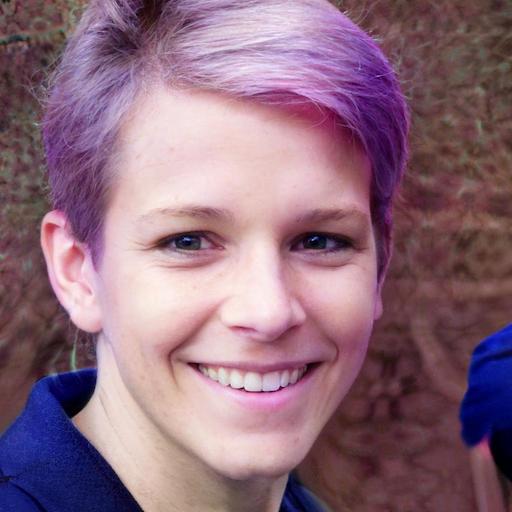}
		\\
		
		\includegraphics[width=0.22\columnwidth]{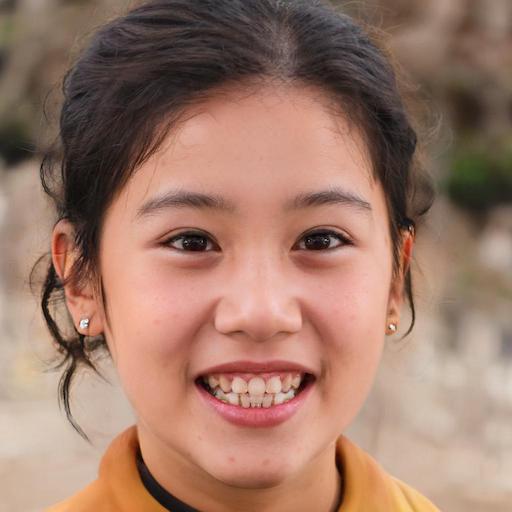} & 
		\includegraphics[width=0.22\columnwidth]{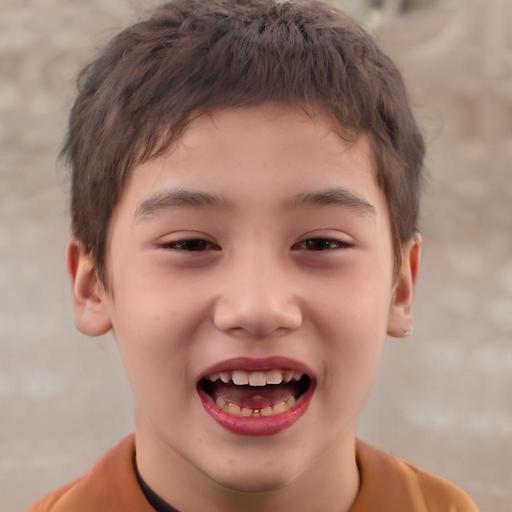} & 
		\includegraphics[width=0.22\columnwidth]{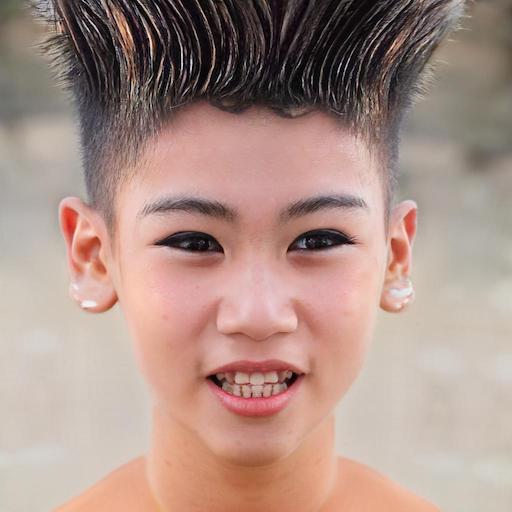} & 
		\includegraphics[width=0.22\columnwidth]{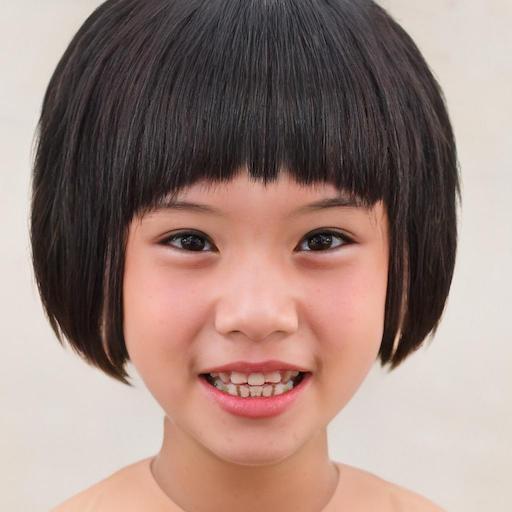} & 
		\includegraphics[width=0.22\columnwidth]{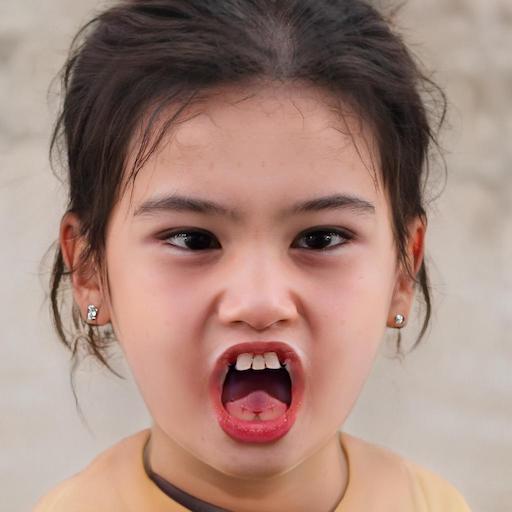} & 
		\includegraphics[width=0.22\columnwidth]{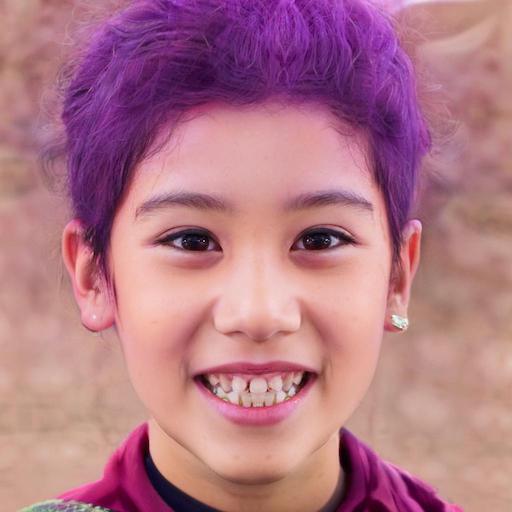}
		\\
		\includegraphics[width=0.22\columnwidth]{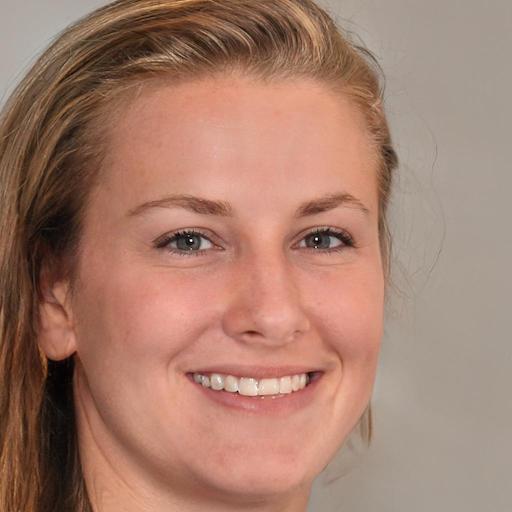} & 
		\includegraphics[width=0.22\columnwidth]{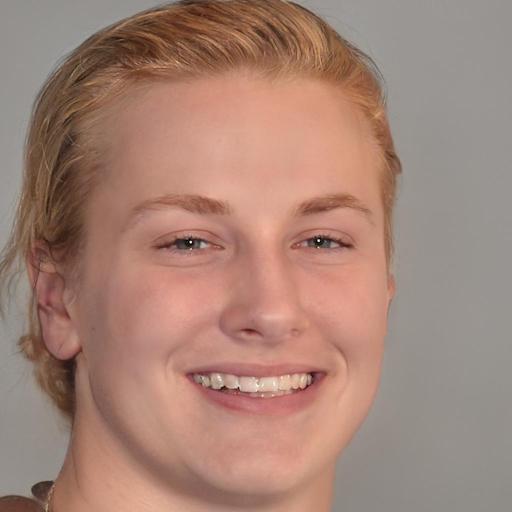} & 
		\includegraphics[width=0.22\columnwidth]{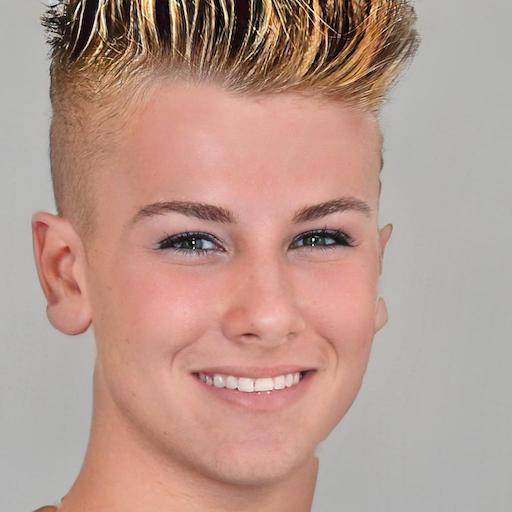} & 
		\includegraphics[width=0.22\columnwidth]{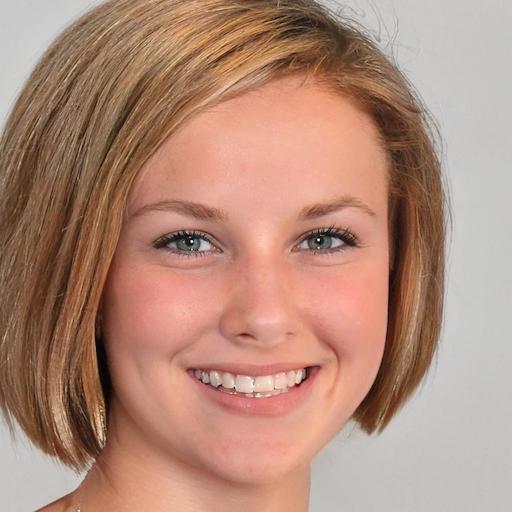} & 
		\includegraphics[width=0.22\columnwidth]{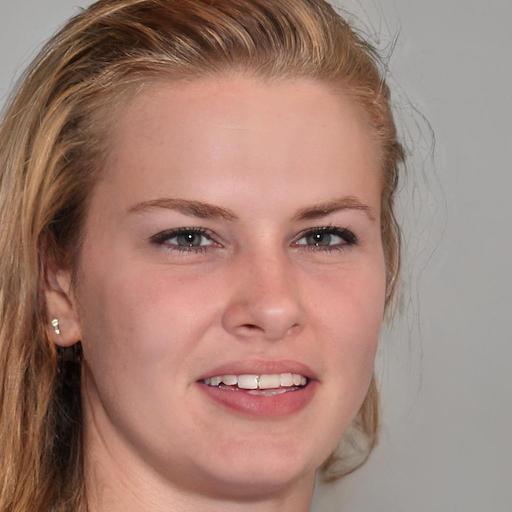} & 
		\includegraphics[width=0.22\columnwidth]{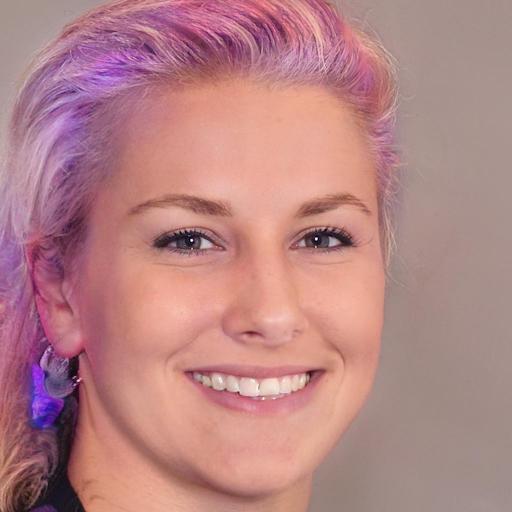}
		\\
		
		\includegraphics[width=0.22\columnwidth]{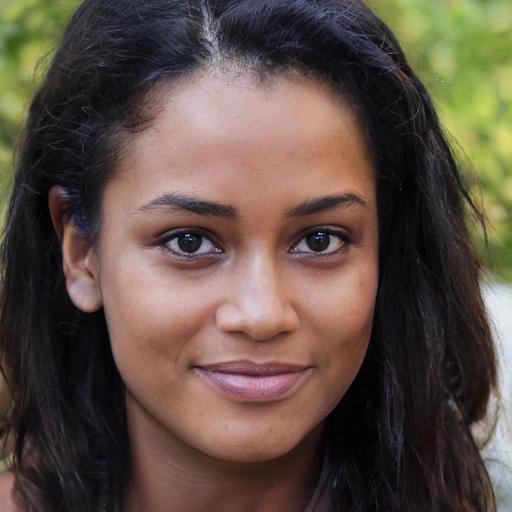} & 
		\includegraphics[width=0.22\columnwidth]{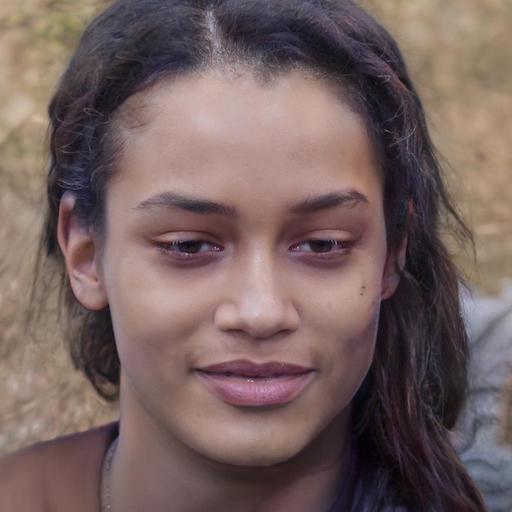} & 
		\includegraphics[width=0.22\columnwidth]{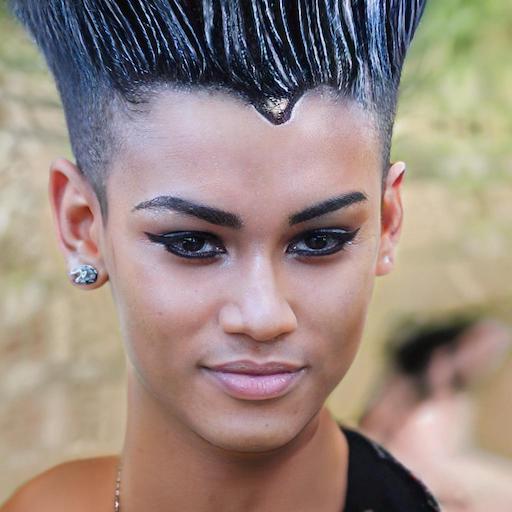} & 
		\includegraphics[width=0.22\columnwidth]{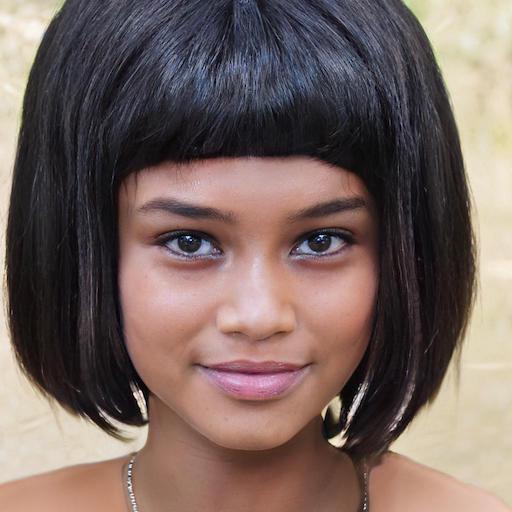} & 
		\includegraphics[width=0.22\columnwidth]{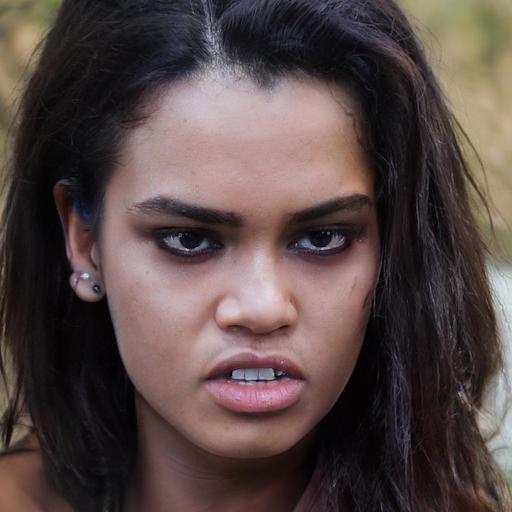} & 
		\includegraphics[width=0.22\columnwidth]{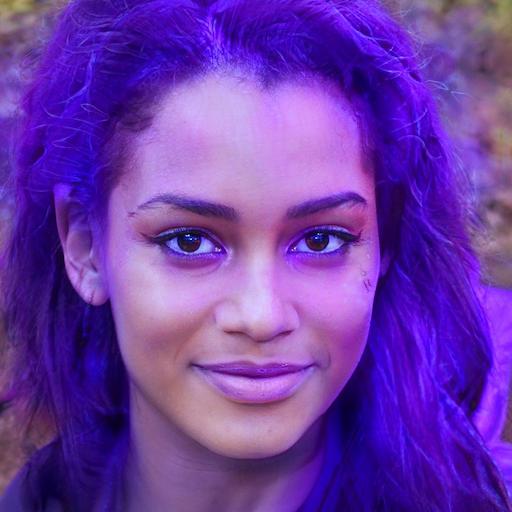}
		\\

		\includegraphics[width=0.22\columnwidth]{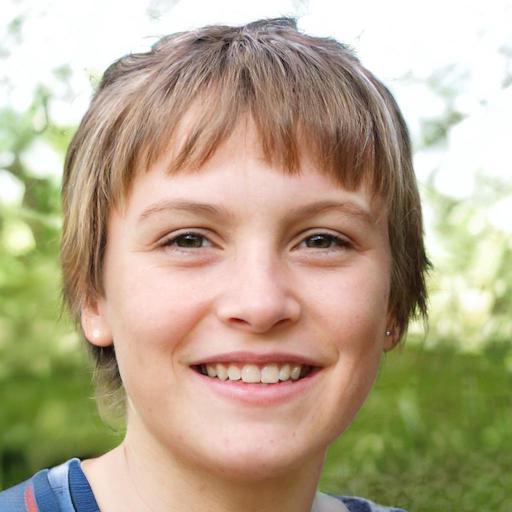} & 
		\includegraphics[width=0.22\columnwidth]{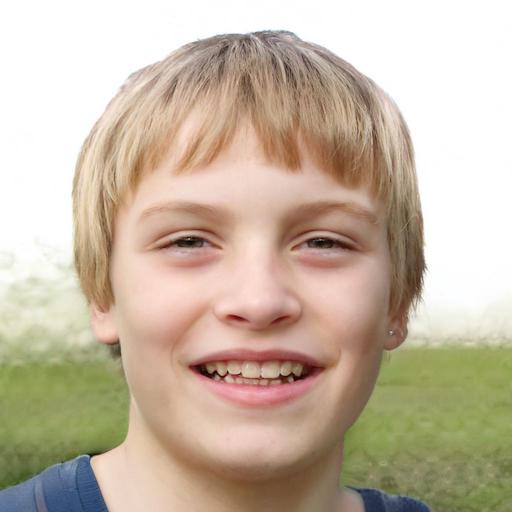} & 
		\includegraphics[width=0.22\columnwidth]{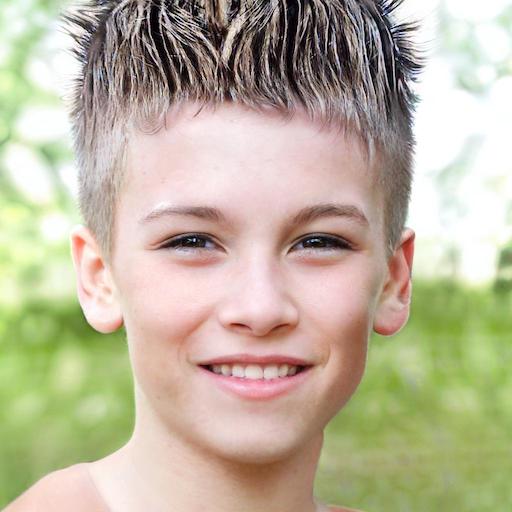} & 
		\includegraphics[width=0.22\columnwidth]{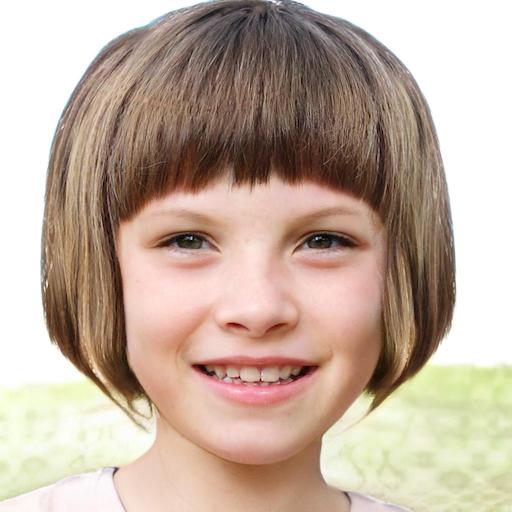} & 
		\includegraphics[width=0.22\columnwidth]{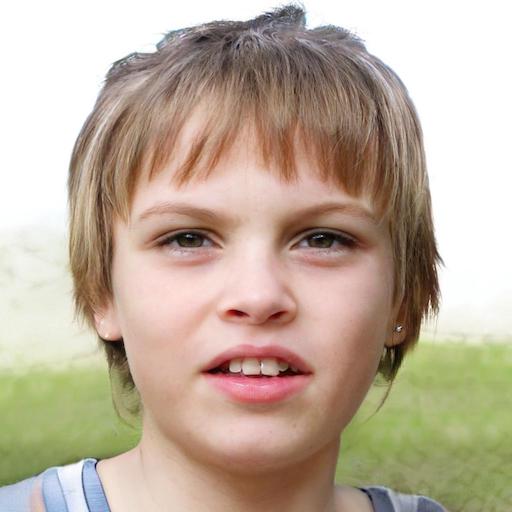} & 
		\includegraphics[width=0.22\columnwidth]{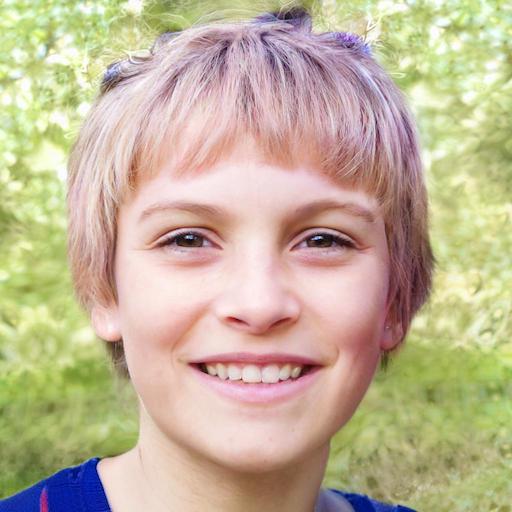}
		\\
		
		\includegraphics[width=0.22\columnwidth]{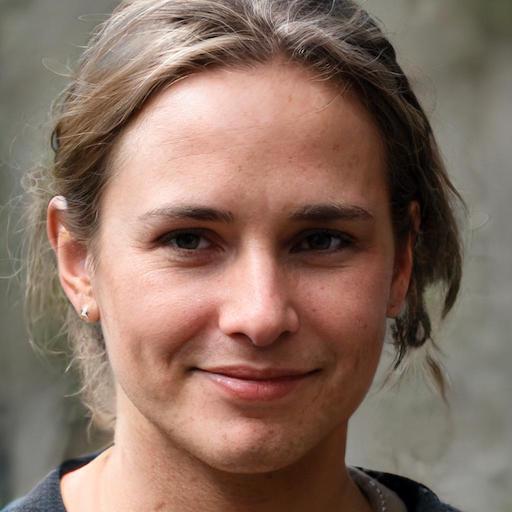} & 
		\includegraphics[width=0.22\columnwidth]{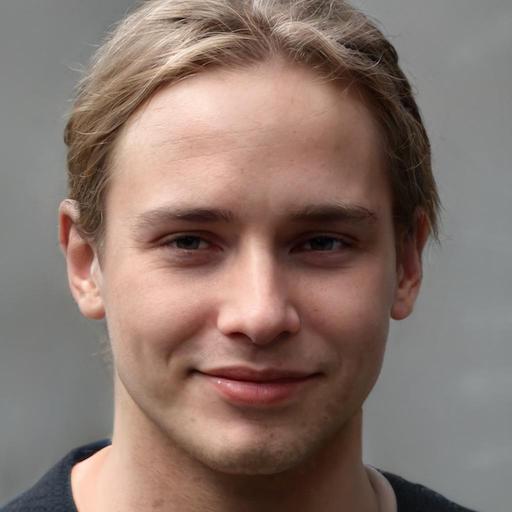} & 
		\includegraphics[width=0.22\columnwidth]{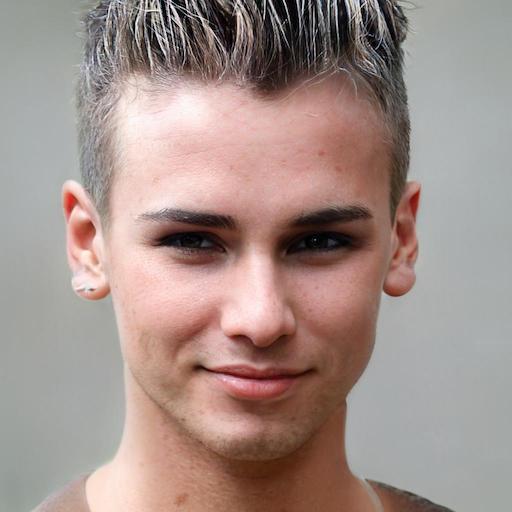} & 
		\includegraphics[width=0.22\columnwidth]{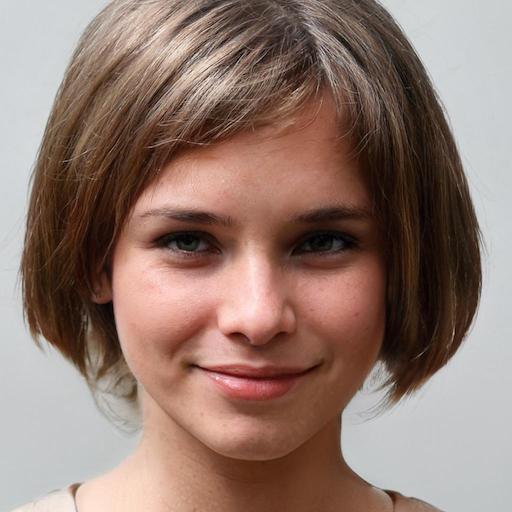} & 
		\includegraphics[width=0.22\columnwidth]{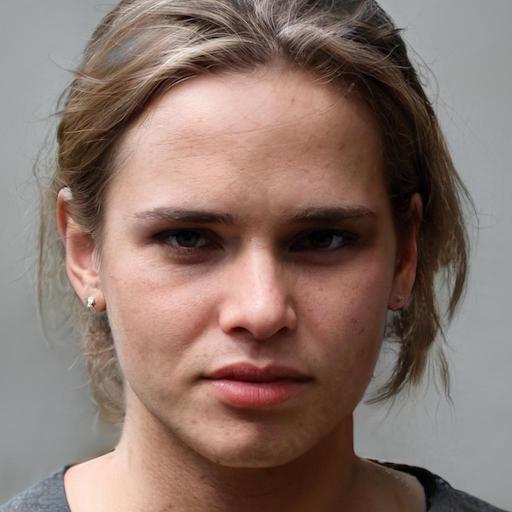} & 
		\includegraphics[width=0.22\columnwidth]{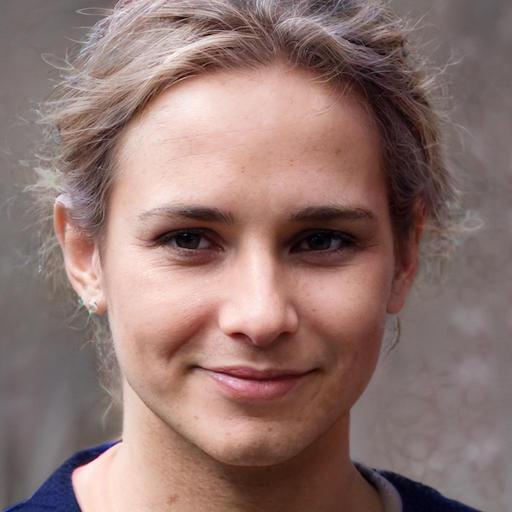}
		\\
		
		\includegraphics[width=0.22\columnwidth]{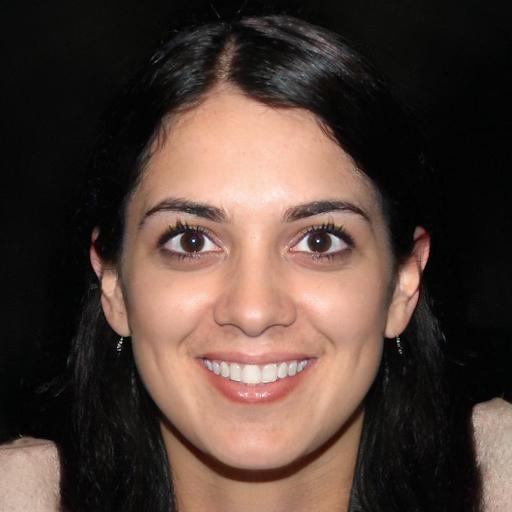} & 
		\includegraphics[width=0.22\columnwidth]{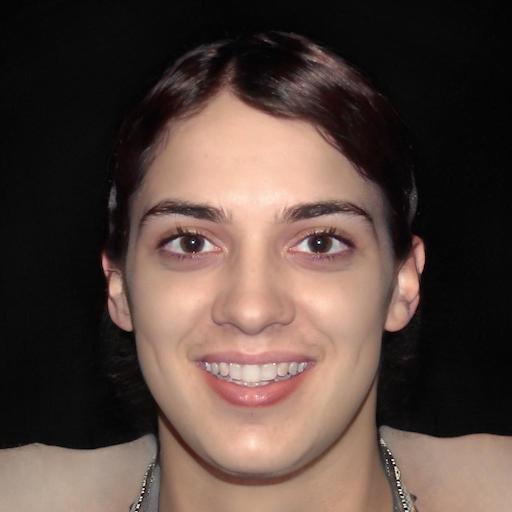} & 
		\includegraphics[width=0.22\columnwidth]{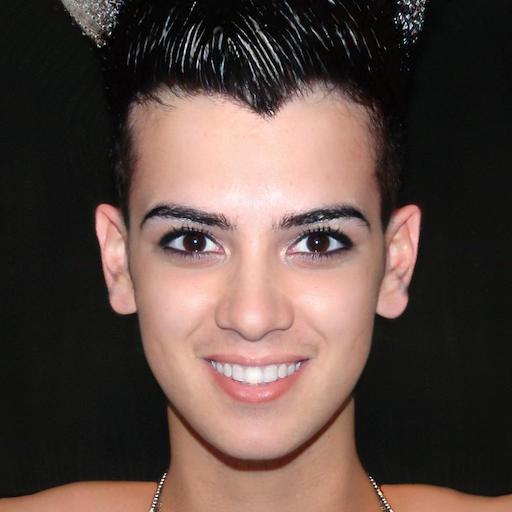} & 
		\includegraphics[width=0.22\columnwidth]{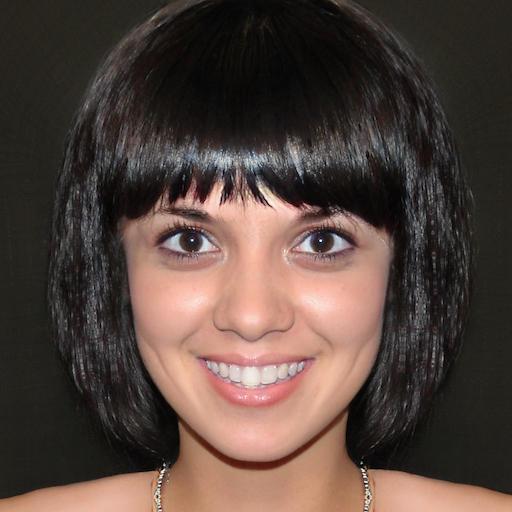} &
		\includegraphics[width=0.22\columnwidth]{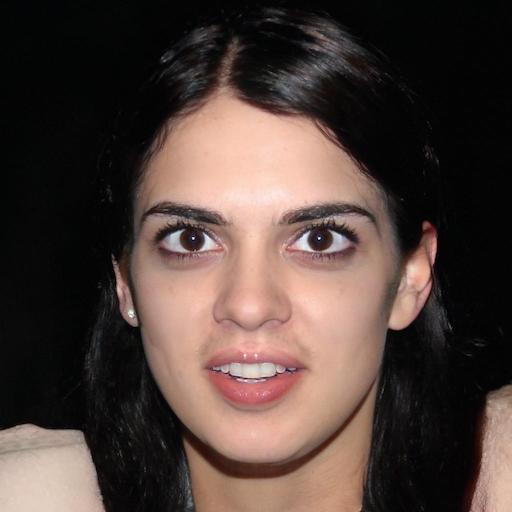} & 
		\includegraphics[width=0.22\columnwidth]{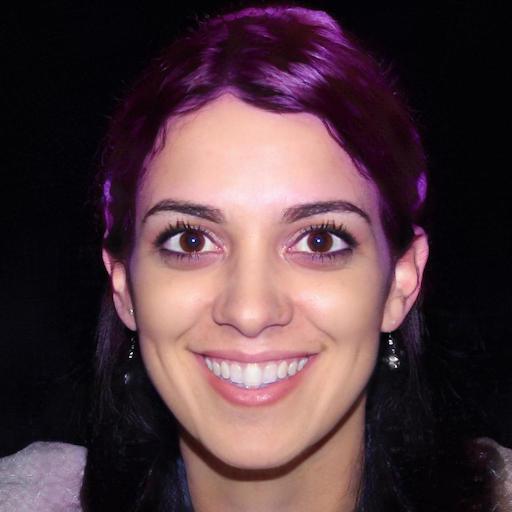}
		\\
		
		\includegraphics[width=0.22\columnwidth]{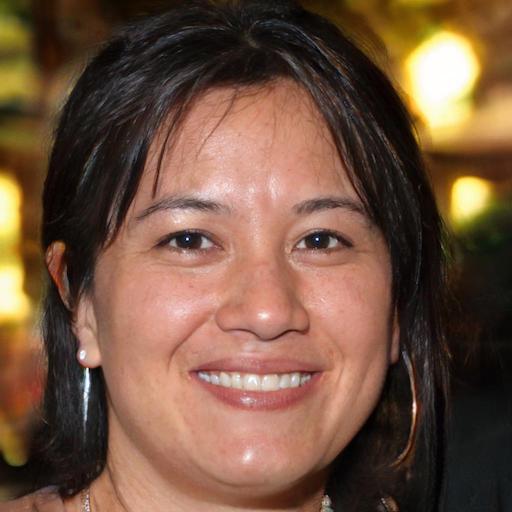} & 
		\includegraphics[width=0.22\columnwidth]{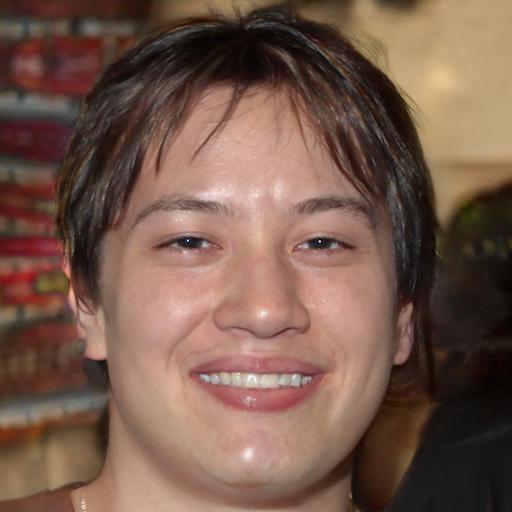} & 
		\includegraphics[width=0.22\columnwidth]{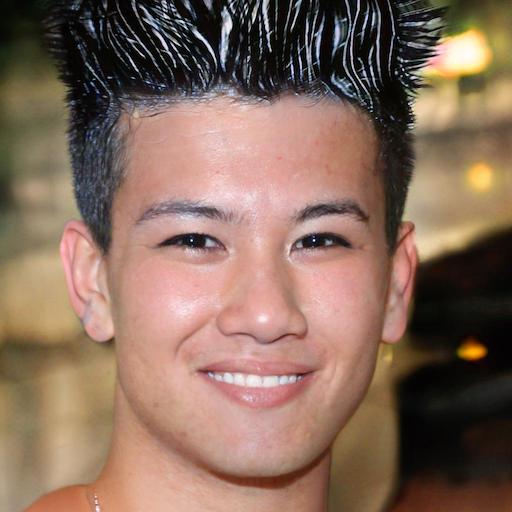} & 
		\includegraphics[width=0.22\columnwidth]{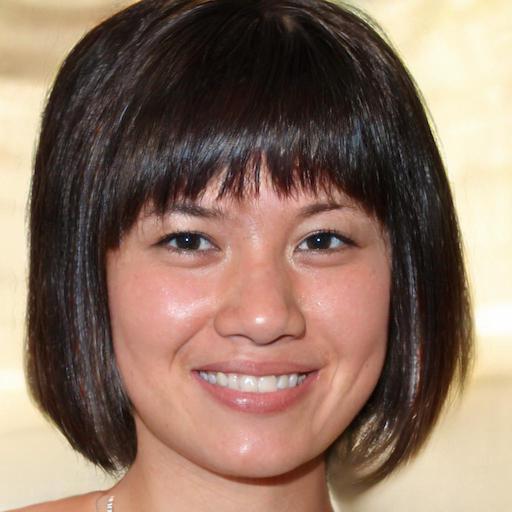} & 
		\includegraphics[width=0.22\columnwidth]{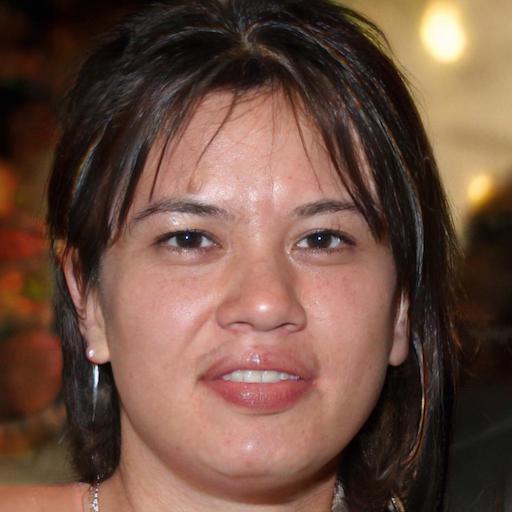} & 
		\includegraphics[width=0.22\columnwidth]{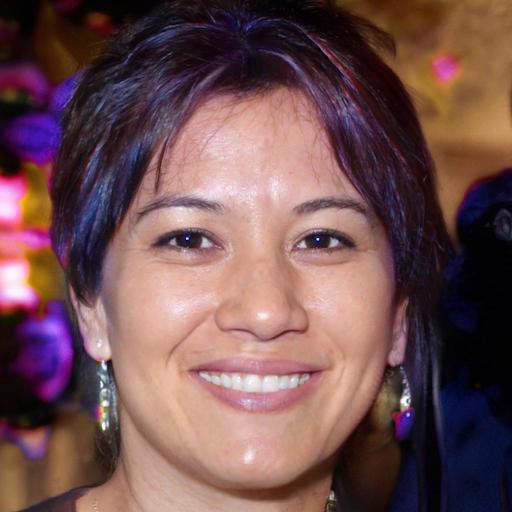}
		\\
		
		\includegraphics[width=0.22\columnwidth]{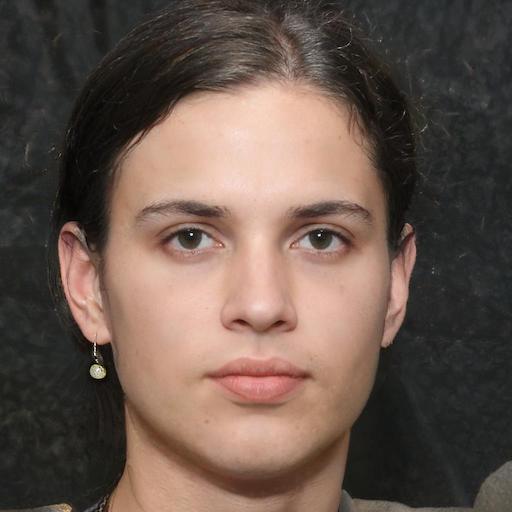} & 
		\includegraphics[width=0.22\columnwidth]{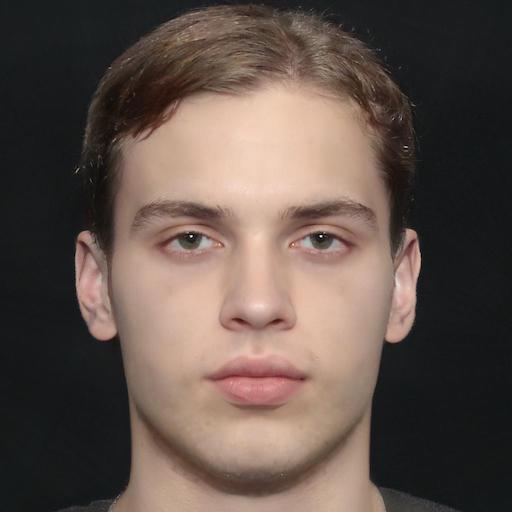} & 
		\includegraphics[width=0.22\columnwidth]{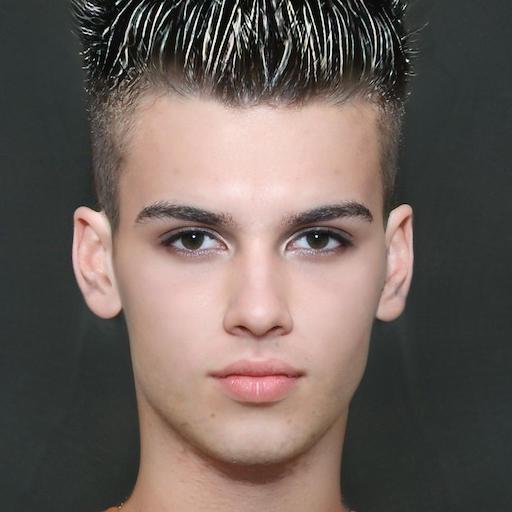} & 
		\includegraphics[width=0.22\columnwidth]{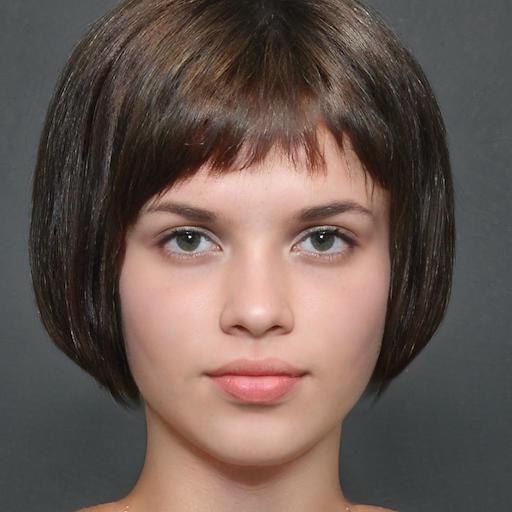} & 
		\includegraphics[width=0.22\columnwidth]{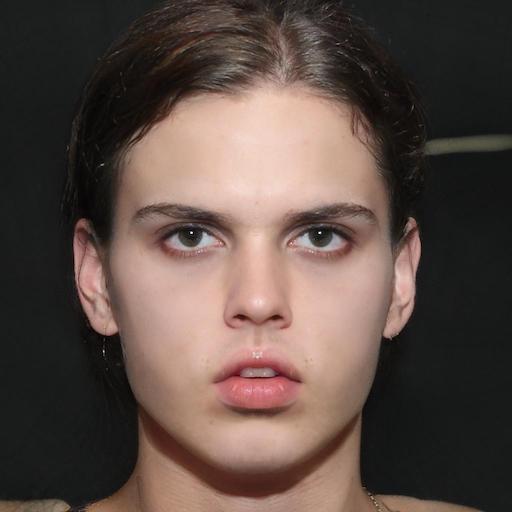} & 
		\includegraphics[width=0.22\columnwidth]{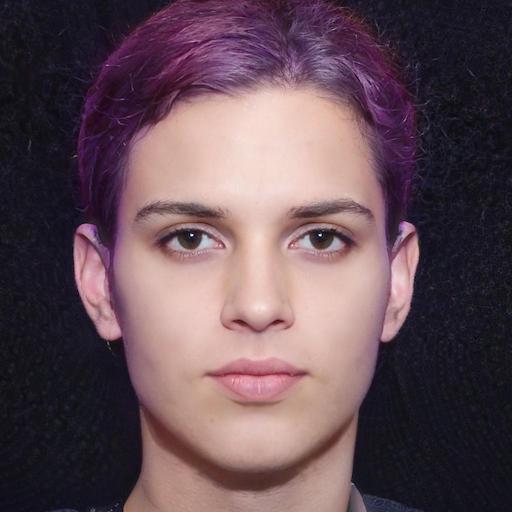}
		\\

		Source & Man & Mohawk & Bobcut & Angry & Purple Hair
	\end{tabular}
	}
	\vspace{-0.1cm}
	\caption{Non-linear editing in $\mathcal{W}+$. We edit images using the StyleCLIP mapping technique with StyleGAN3 trained on aligned faces. Even with non-linear editing paths, the edits are still entangled: local edits (e.g., expression/hairstyle) alter other attributes (e.g., background/identity).}
	\vspace{-0.4cm}
	\label{fig:styleclip-edits-supp}
\end{figure*}

%% file: figures/supplementary/styleclip_edit_ffhq.tex
\begin{figure*}[tb]
	\centering
	\setlength{\tabcolsep}{1pt}
	
	{\small
	\begin{tabular}{c c c c c c}
        \\
		\includegraphics[width=0.22\columnwidth]{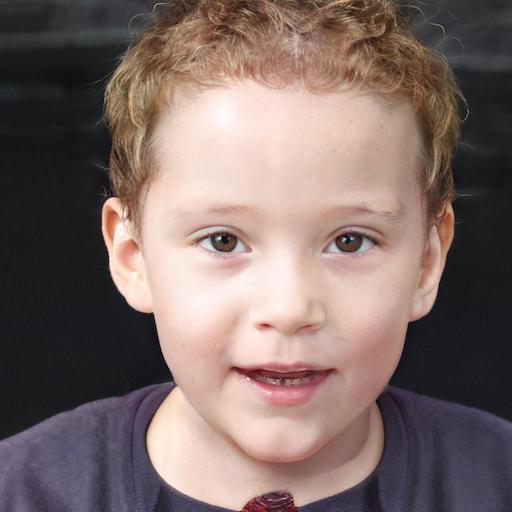} & 
        \includegraphics[width=0.22\columnwidth]{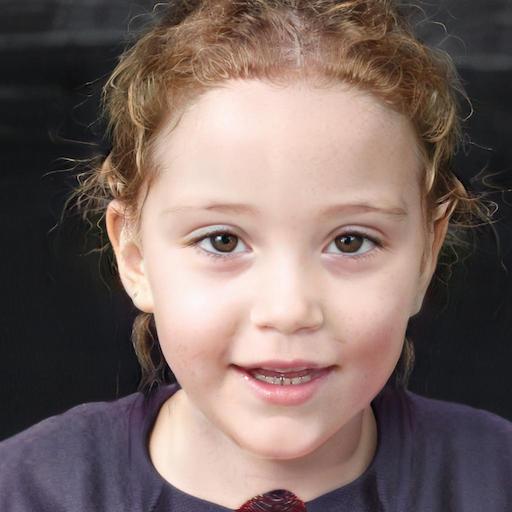} & 
        \includegraphics[width=0.22\columnwidth]{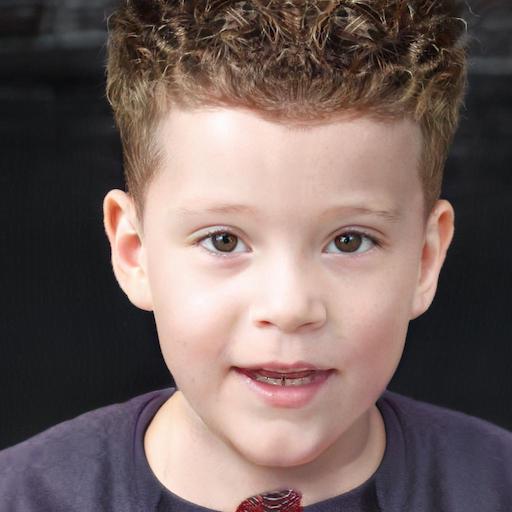} & 
        \includegraphics[width=0.22\columnwidth]{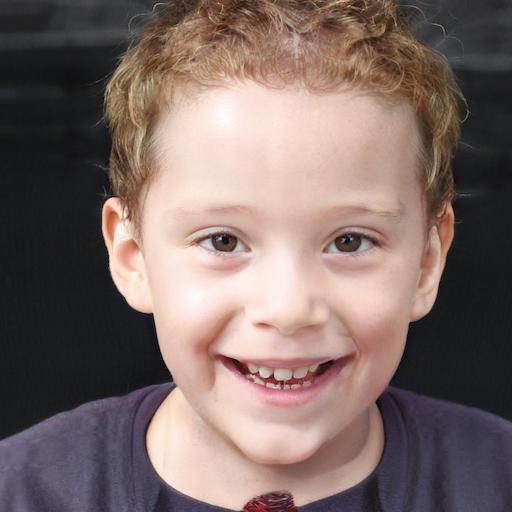} & 
        \includegraphics[width=0.22\columnwidth]{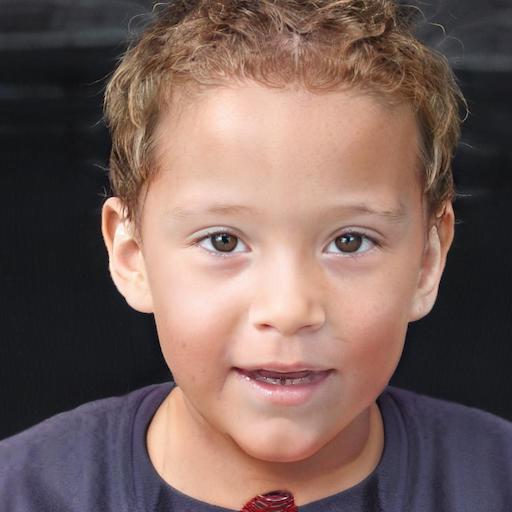} 
        \\
		\includegraphics[width=0.22\columnwidth]{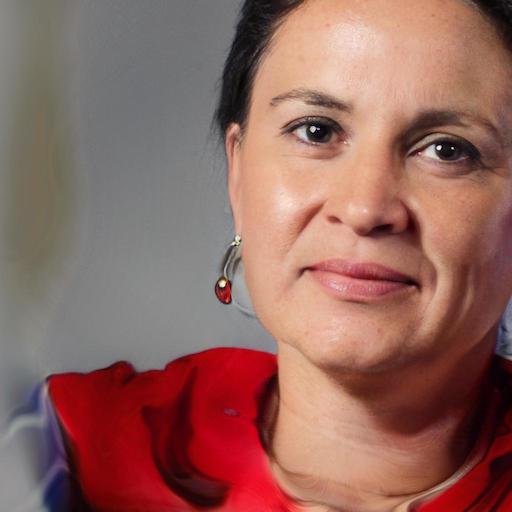} & 
        \includegraphics[width=0.22\columnwidth]{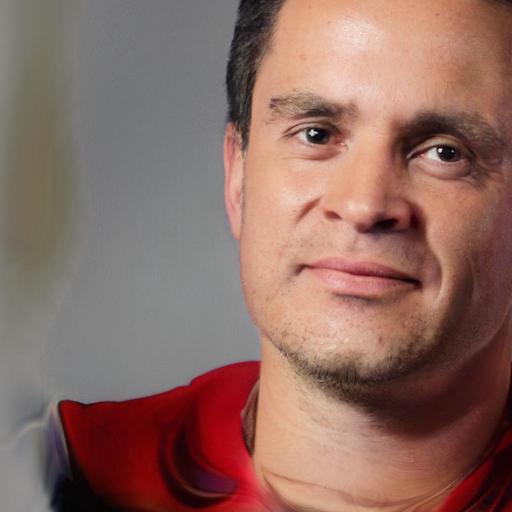} & 
        \includegraphics[width=0.22\columnwidth]{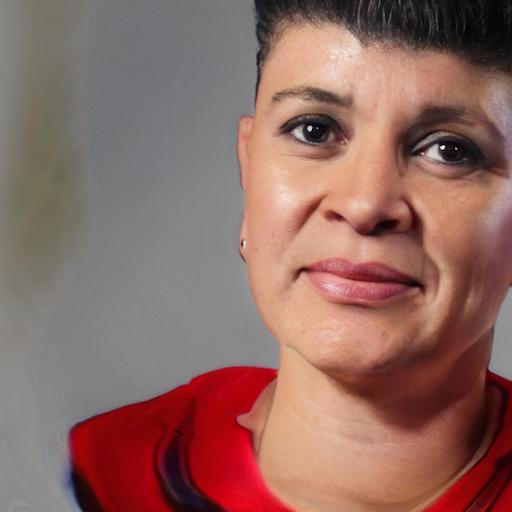} & 
        \includegraphics[width=0.22\columnwidth]{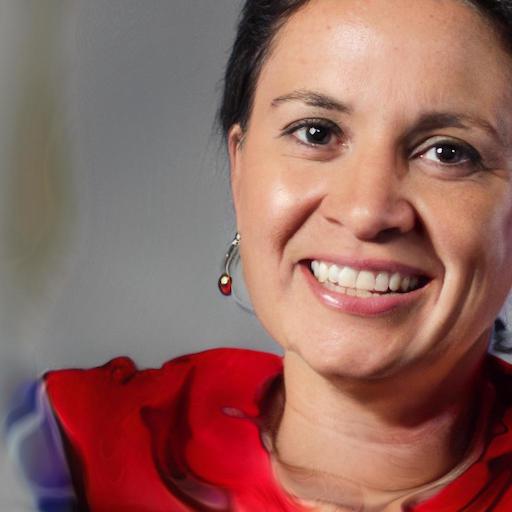} & 
        \includegraphics[width=0.22\columnwidth]{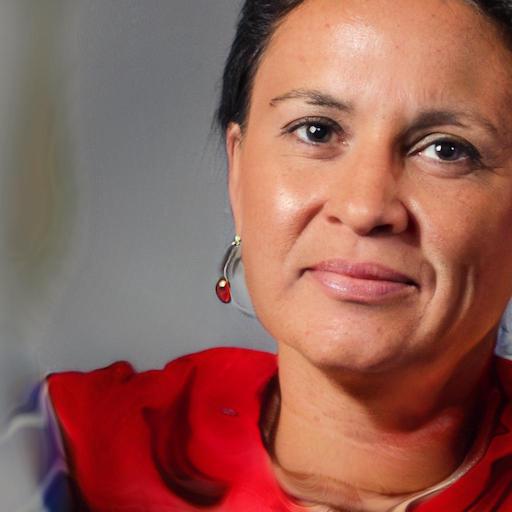} 
        \\
		\includegraphics[width=0.22\columnwidth]{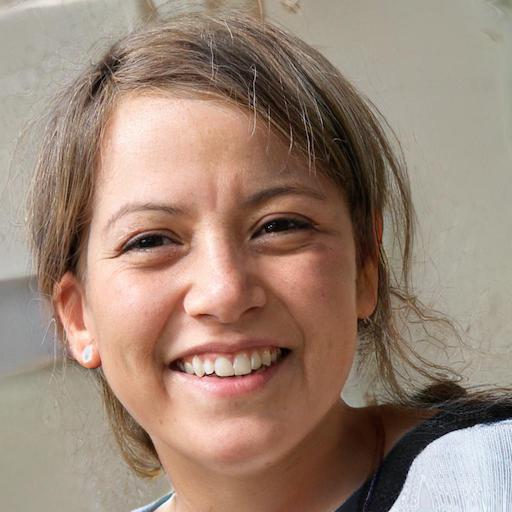} & 
        \includegraphics[width=0.22\columnwidth]{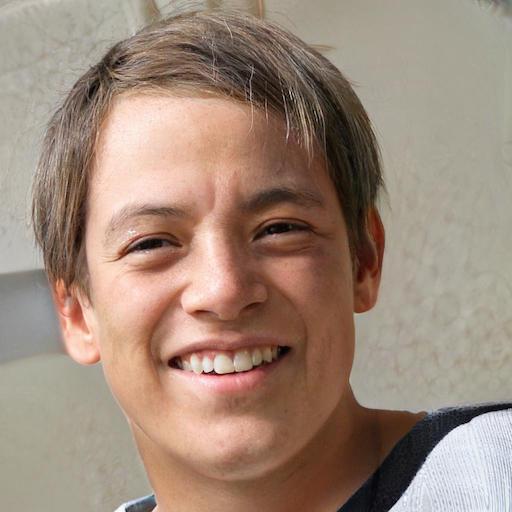} & 
        \includegraphics[width=0.22\columnwidth]{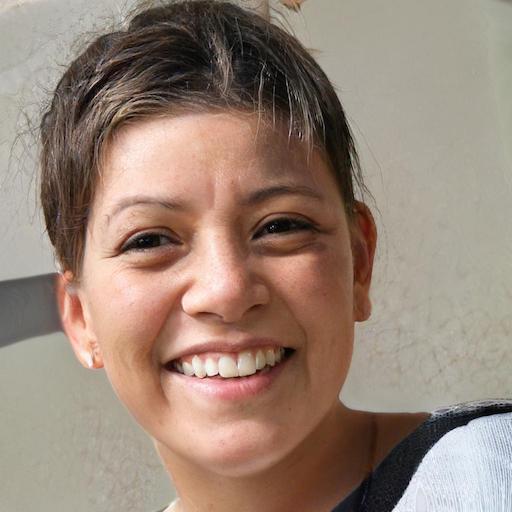} & 
        \includegraphics[width=0.22\columnwidth]{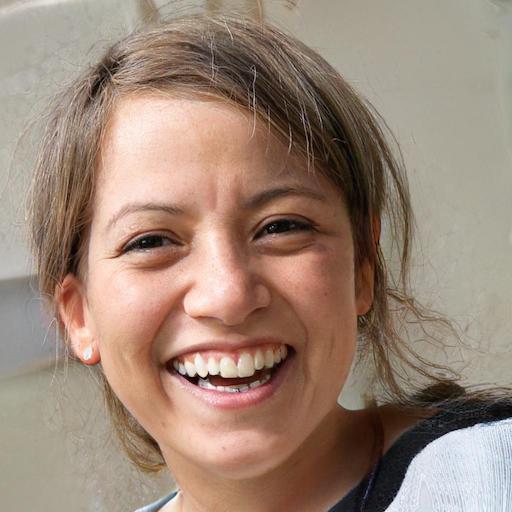} & 
        \includegraphics[width=0.22\columnwidth]{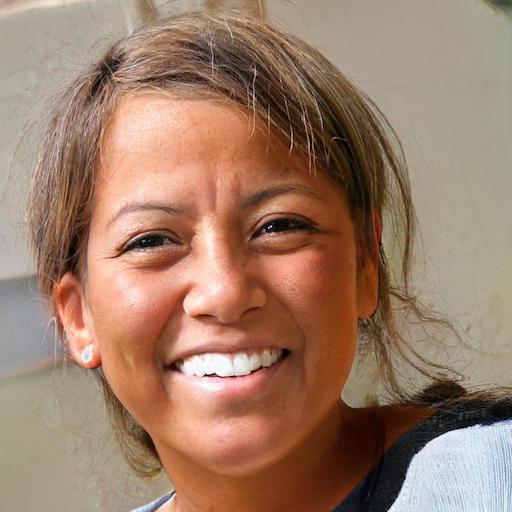} 
        \\
		\includegraphics[width=0.22\columnwidth]{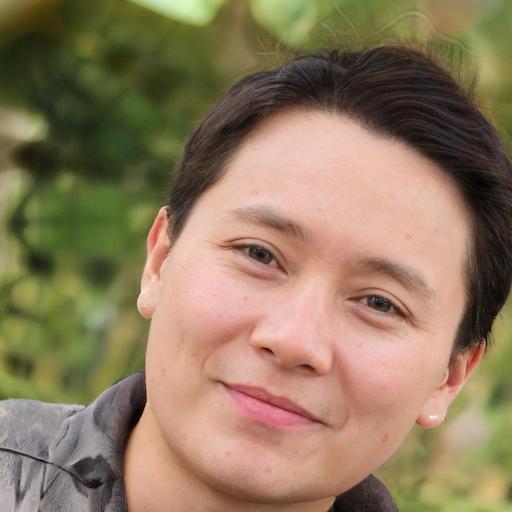} & 
        \includegraphics[width=0.22\columnwidth]{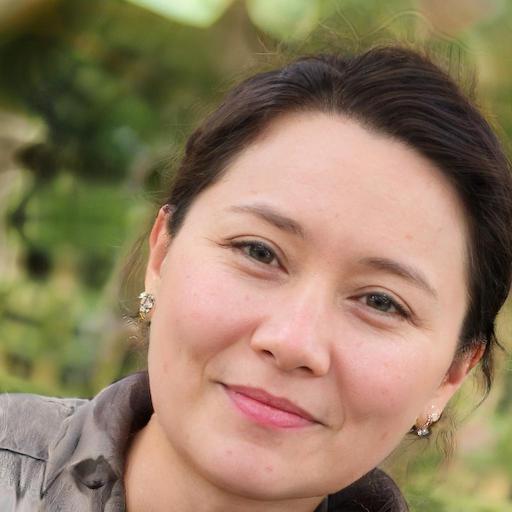} & 
        \includegraphics[width=0.22\columnwidth]{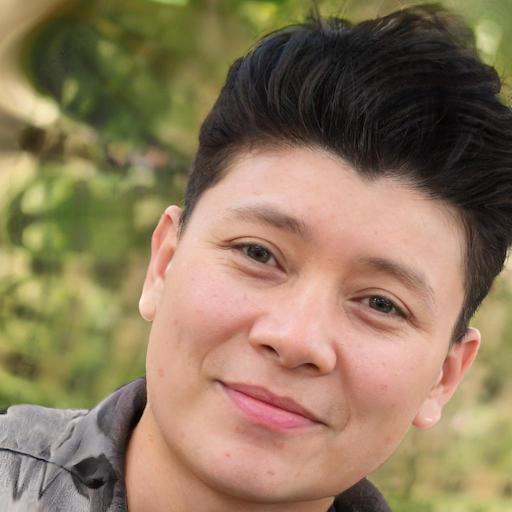} & 
        \includegraphics[width=0.22\columnwidth]{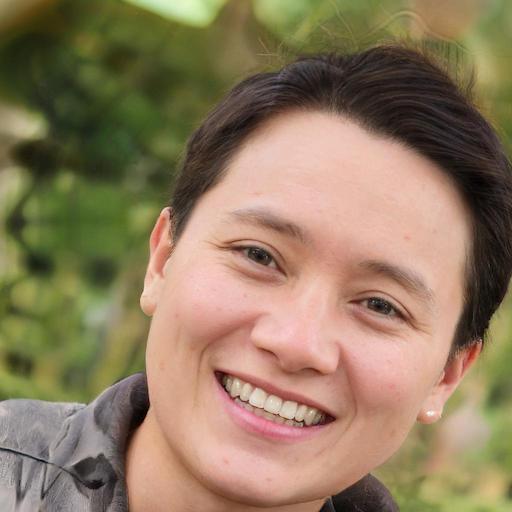} & 
        \includegraphics[width=0.22\columnwidth]{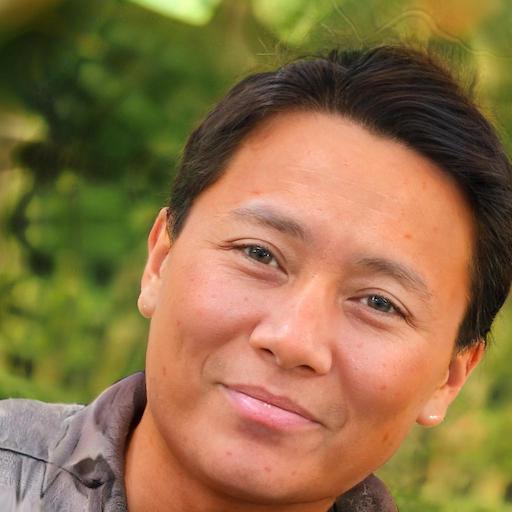} 
        \\
		\includegraphics[width=0.22\columnwidth]{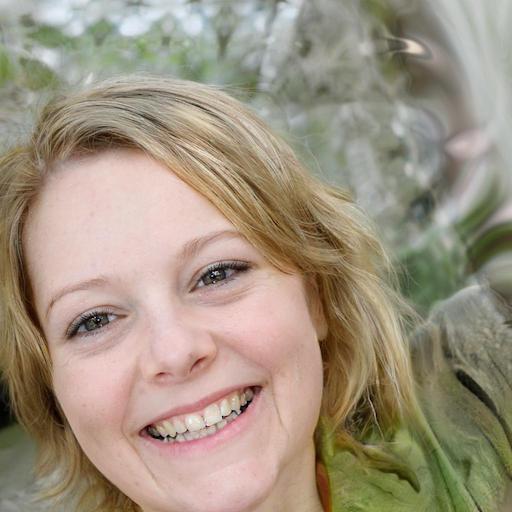} & 
        \includegraphics[width=0.22\columnwidth]{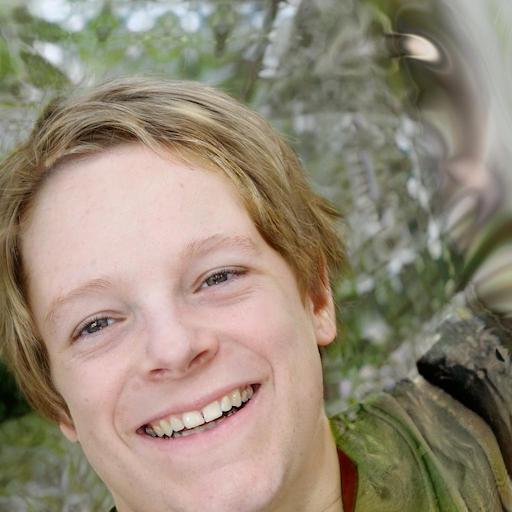} & 
        \includegraphics[width=0.22\columnwidth]{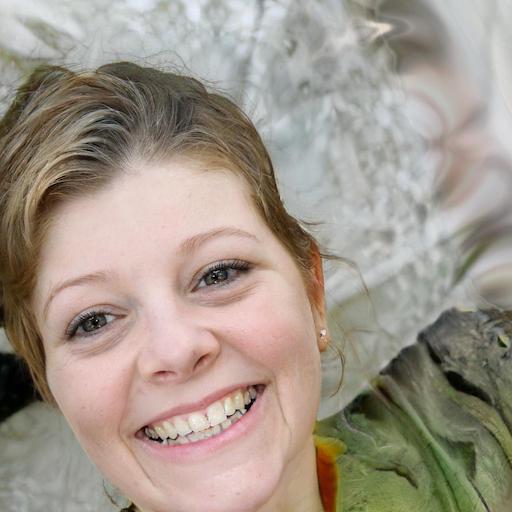} & 
        \includegraphics[width=0.22\columnwidth]{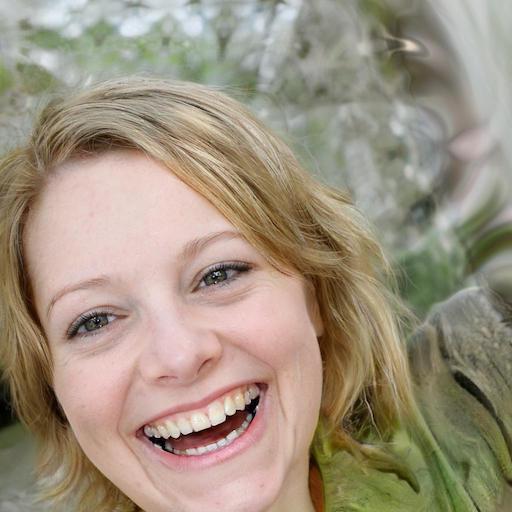} & 
        \includegraphics[width=0.22\columnwidth]{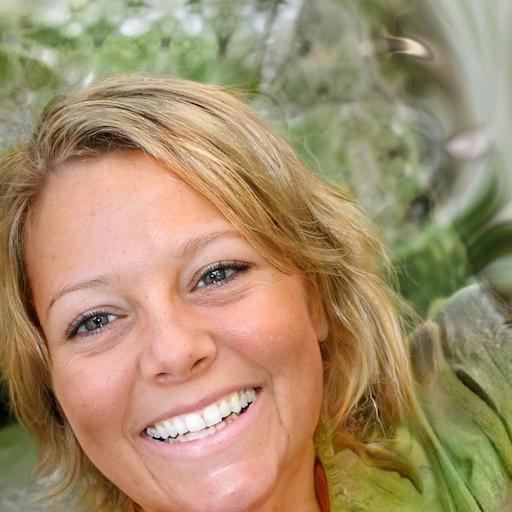} 
        \\
		\includegraphics[width=0.22\columnwidth]{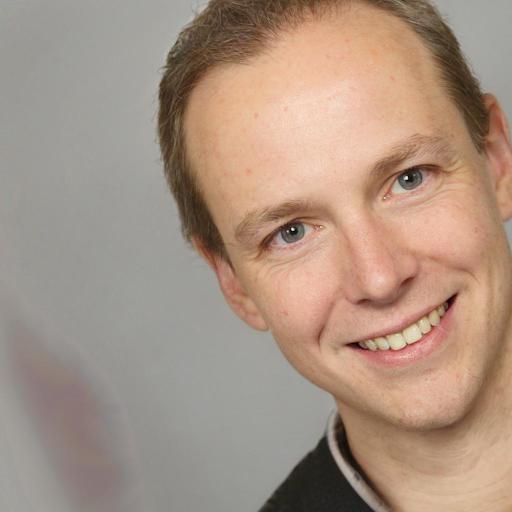} & 
        \includegraphics[width=0.22\columnwidth]{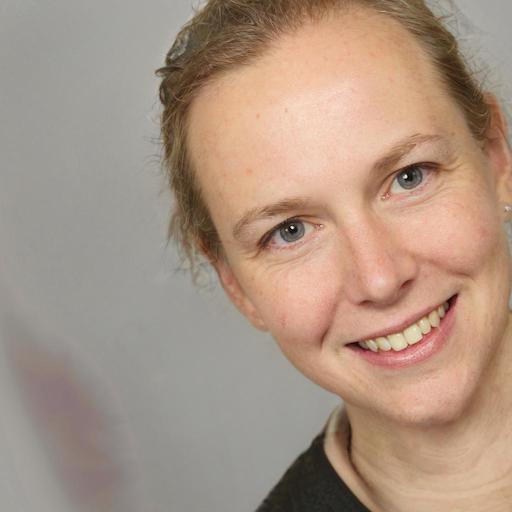} & 
        \includegraphics[width=0.22\columnwidth]{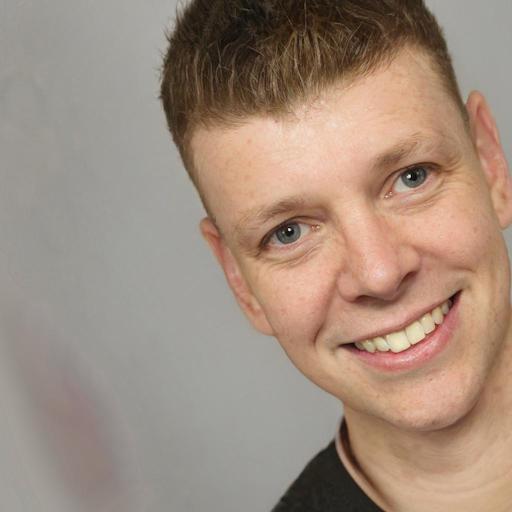} & 
        \includegraphics[width=0.22\columnwidth]{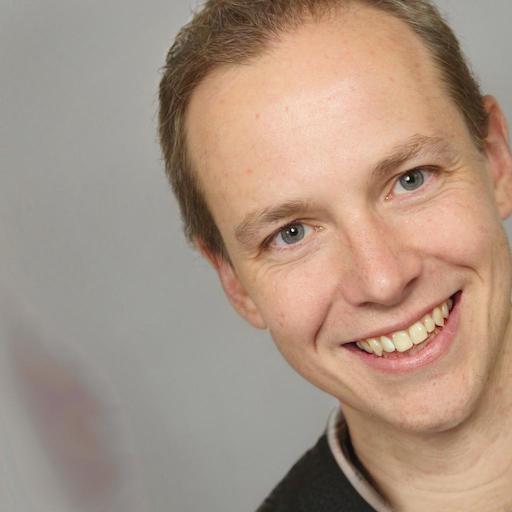} & 
        \includegraphics[width=0.22\columnwidth]{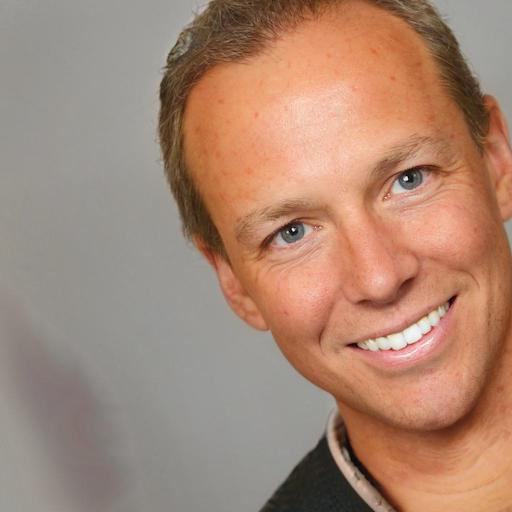} 
        \\
		\includegraphics[width=0.22\columnwidth]{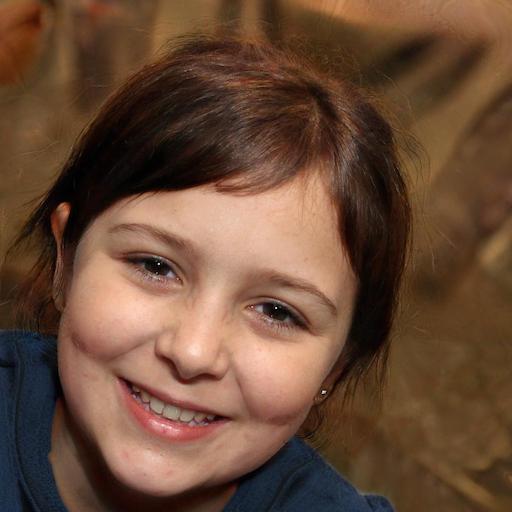} & 
        \includegraphics[width=0.22\columnwidth]{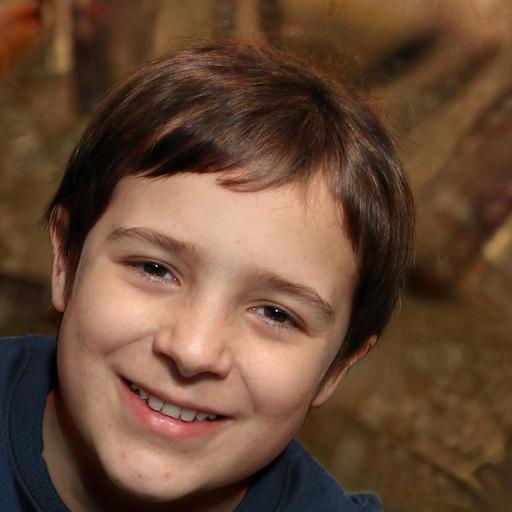} & 
        \includegraphics[width=0.22\columnwidth]{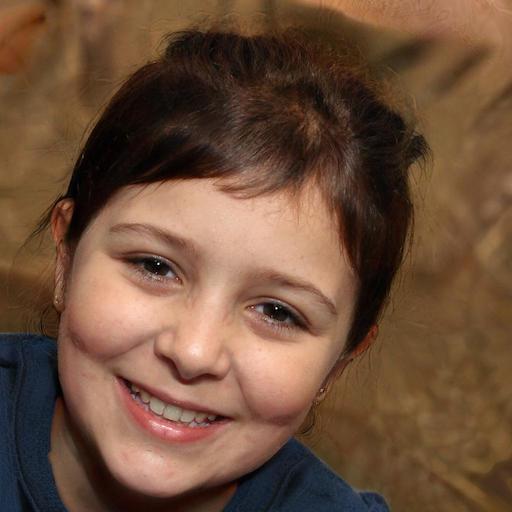} & 
        \includegraphics[width=0.22\columnwidth]{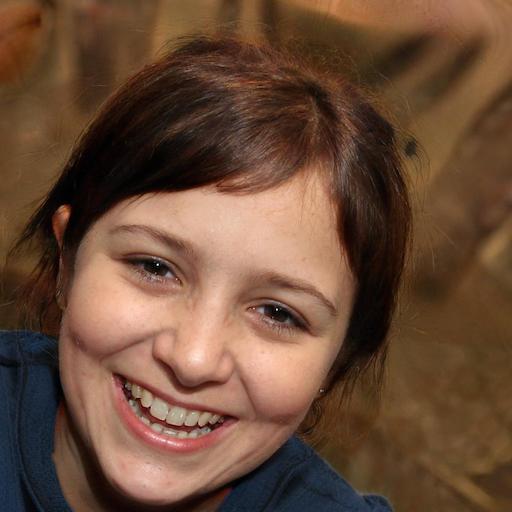} & 
        \includegraphics[width=0.22\columnwidth]{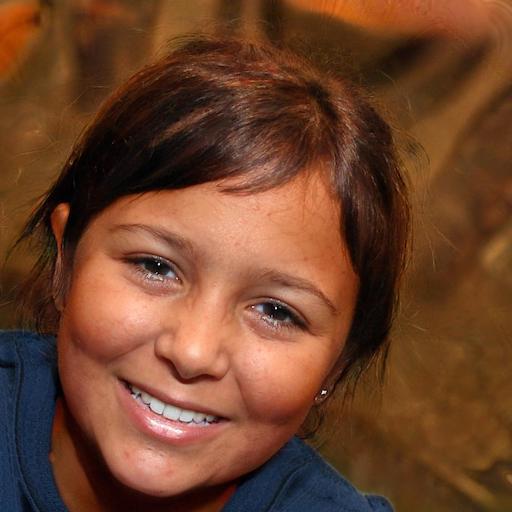} 
        \\
		\includegraphics[width=0.22\columnwidth]{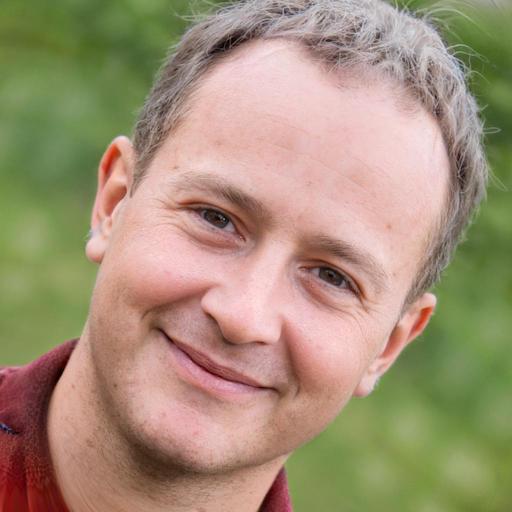} & 
        \includegraphics[width=0.22\columnwidth]{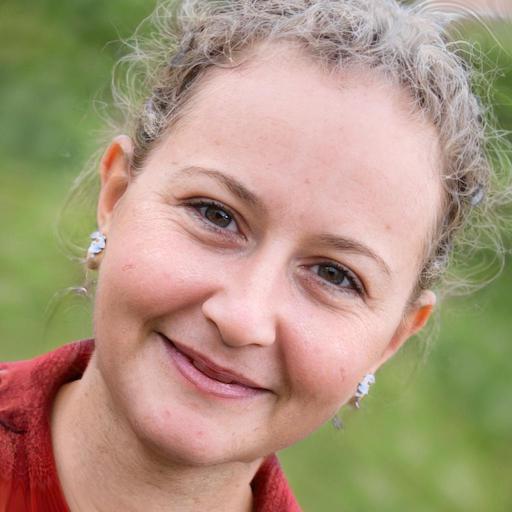} & 
        \includegraphics[width=0.22\columnwidth]{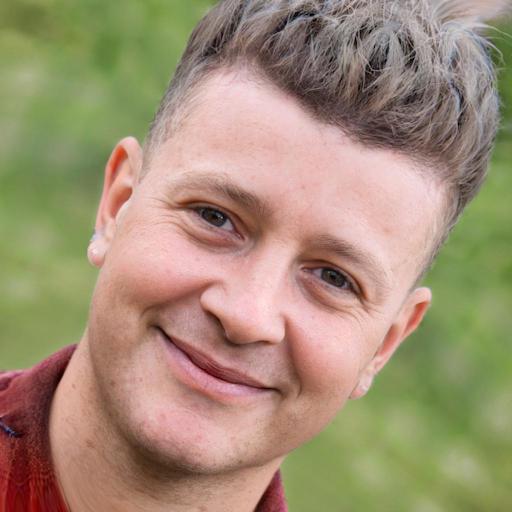} & 
        \includegraphics[width=0.22\columnwidth]{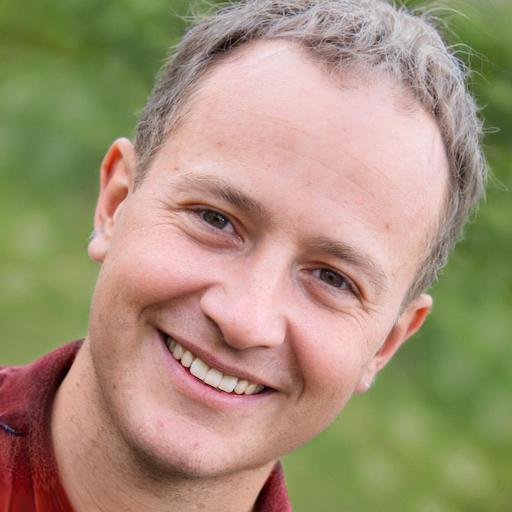} & 
        \includegraphics[width=0.22\columnwidth]{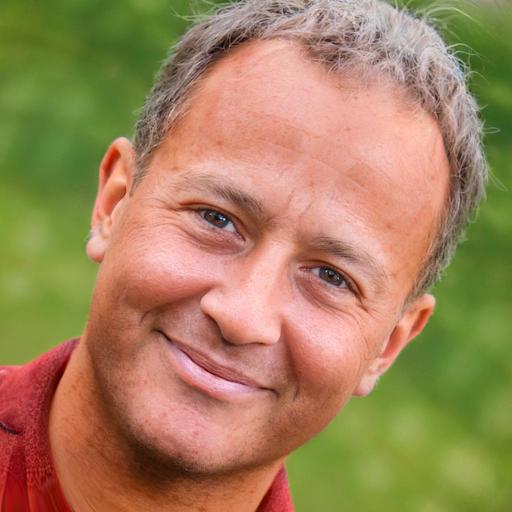} 
        \\
		\includegraphics[width=0.22\columnwidth]{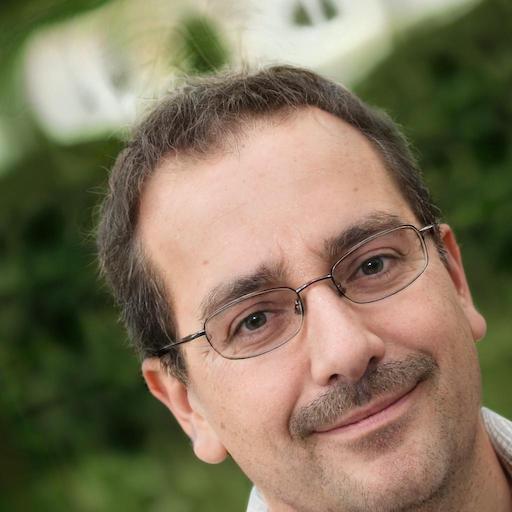} & 
        \includegraphics[width=0.22\columnwidth]{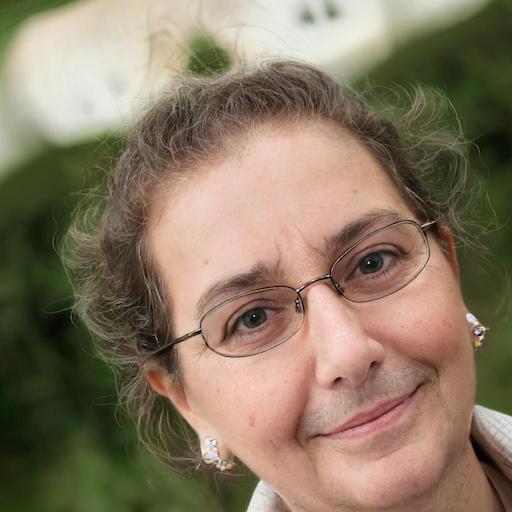} & 
        \includegraphics[width=0.22\columnwidth]{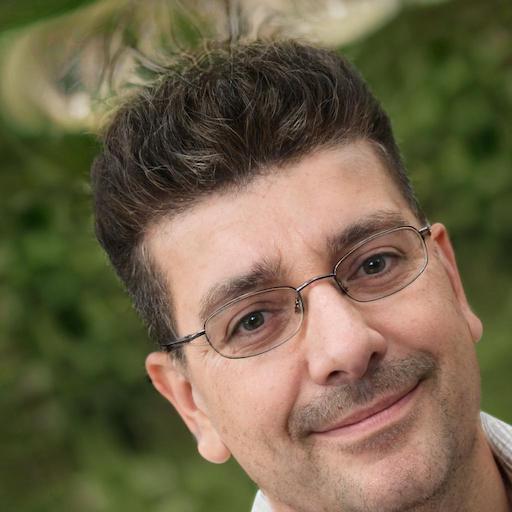} & 
        \includegraphics[width=0.22\columnwidth]{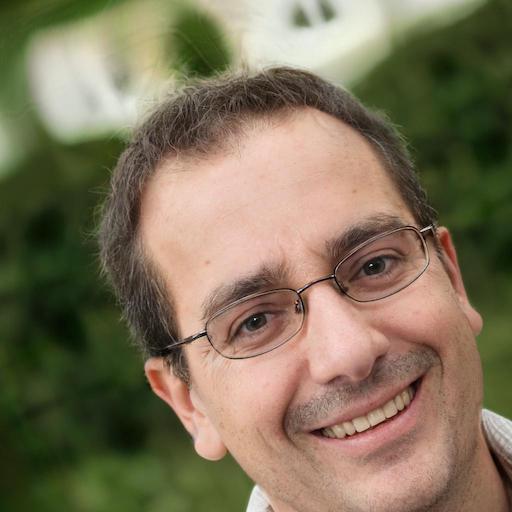} & 
        \includegraphics[width=0.22\columnwidth]{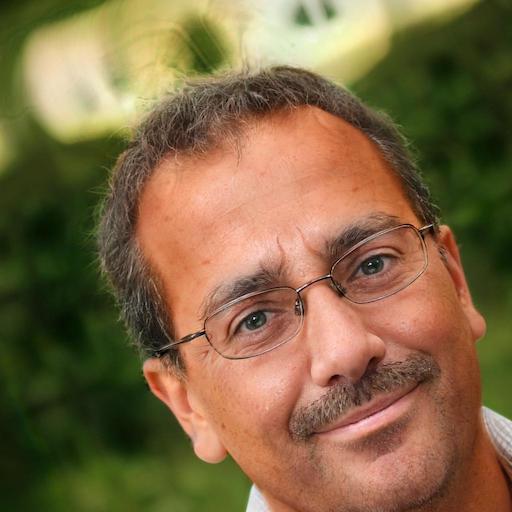} 
        \\
		\includegraphics[width=0.22\columnwidth]{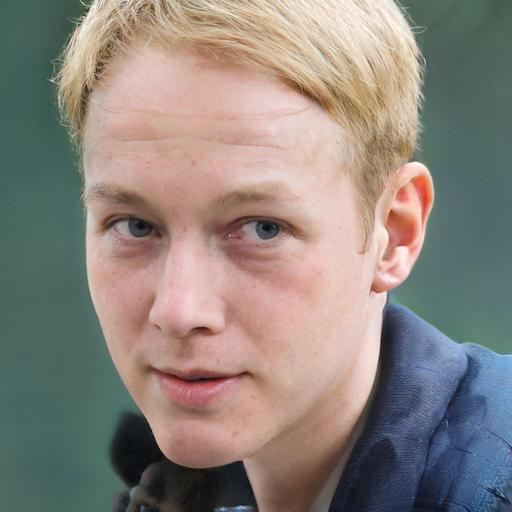} & 
        \includegraphics[width=0.22\columnwidth]{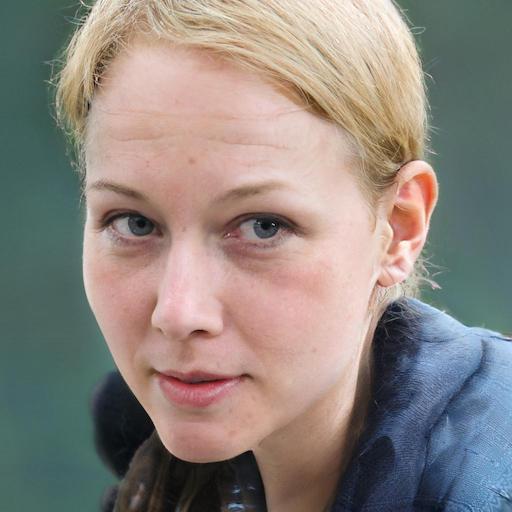} & 
        \includegraphics[width=0.22\columnwidth]{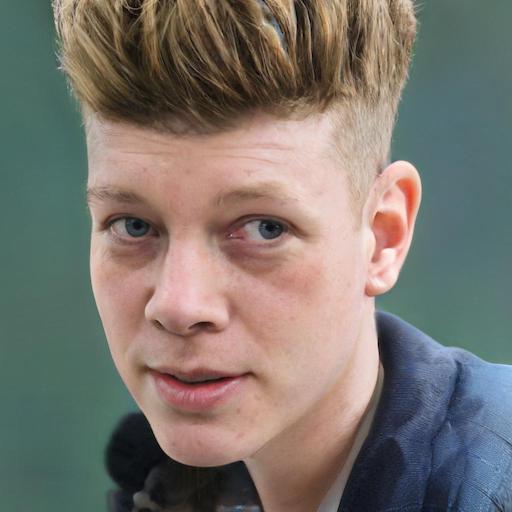} & 
        \includegraphics[width=0.22\columnwidth]{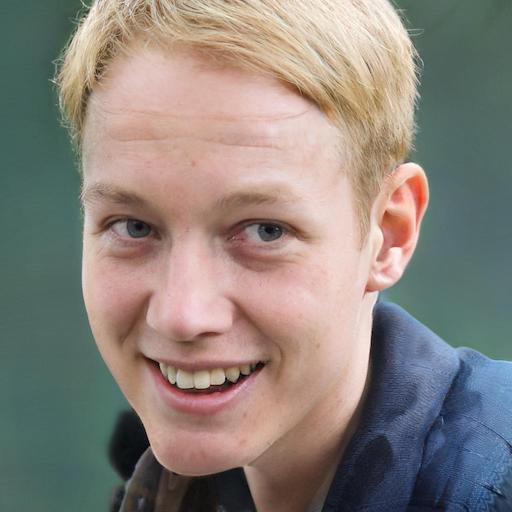} & 
        \includegraphics[width=0.22\columnwidth]{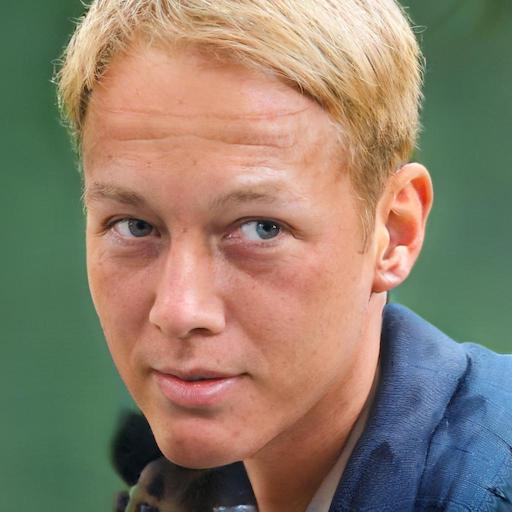} 
        \\

		 Source & Gender & Hi-top Fade & Expression & Tanned
	\end{tabular}
	}

	\vspace{-0.3cm}
	\caption{Editing in $\mathcal{S}$. Editing results on synthetic images obtained using StyleCLIP's global direction technique~\cite{patashnik2021styleclip} applied over an aligned StyleGAN3 generator trained on the aligned FFHQ~\cite{karras2019style} dataset. The unaligned results are obtained by randomly sampling the rotation and translation parameters $(r,t_x,t_y)$ applied over the input Fourier features of the generator.
	}
	\label{fig:styleclip-global-directions-supplementary}
\end{figure*}

%% file: figures/supplementary/edits_afhq.tex
\begin{figure*}[tb]
	\centering
	\setlength{\tabcolsep}{1pt}	
	{\small
	\begin{tabular}{c c c c c c}

        \includegraphics[width=0.25\columnwidth]{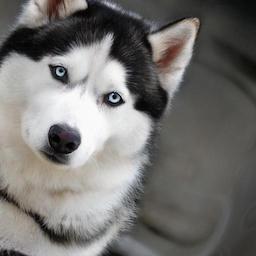} &
        \includegraphics[width=0.25\columnwidth]{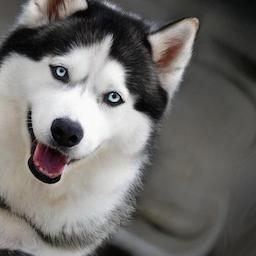} &
        \includegraphics[width=0.25\columnwidth]{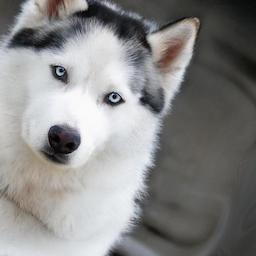} &
        \includegraphics[width=0.25\columnwidth]{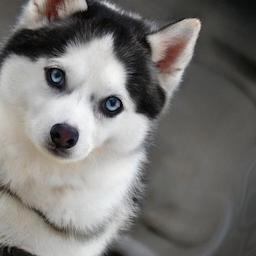} &
        \includegraphics[width=0.25\columnwidth]{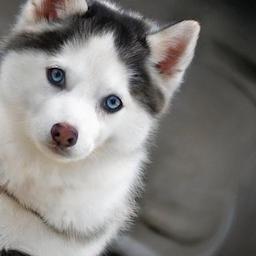} 
        \\
        \includegraphics[width=0.25\columnwidth]{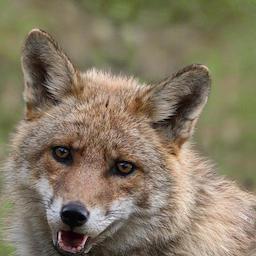} &
        \includegraphics[width=0.25\columnwidth]{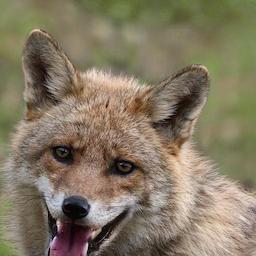} &
        \includegraphics[width=0.25\columnwidth]{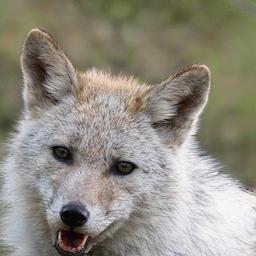} &
        \includegraphics[width=0.25\columnwidth]{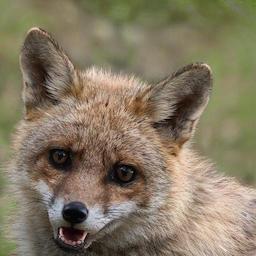} &
        \includegraphics[width=0.25\columnwidth]{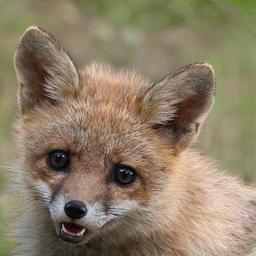} 
        \\
        \includegraphics[width=0.25\columnwidth]{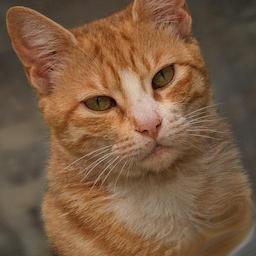} &
        \includegraphics[width=0.25\columnwidth]{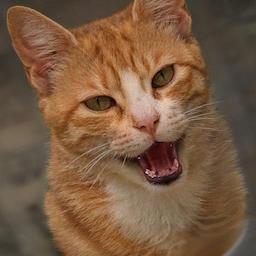} &
        \includegraphics[width=0.25\columnwidth]{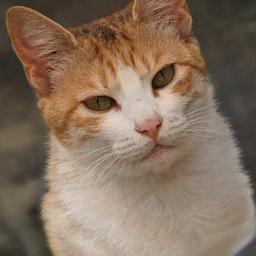} &
        \includegraphics[width=0.25\columnwidth]{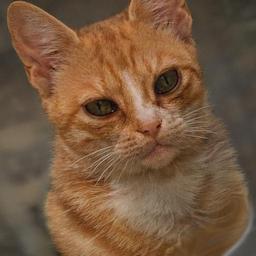} &
        \includegraphics[width=0.25\columnwidth]{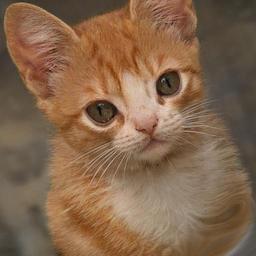} \\
        
        \includegraphics[width=0.25\columnwidth]{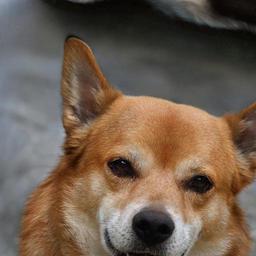} &
        \includegraphics[width=0.25\columnwidth]{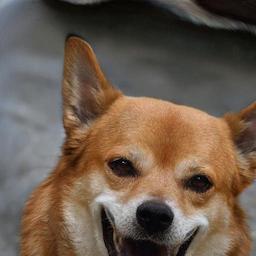} &
        \includegraphics[width=0.25\columnwidth]{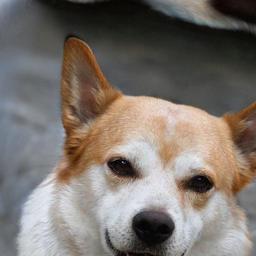} &
        \includegraphics[width=0.25\columnwidth]{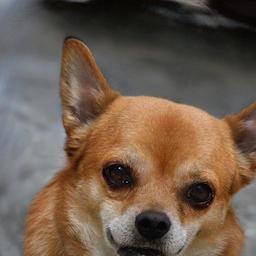} &
        \includegraphics[width=0.25\columnwidth]{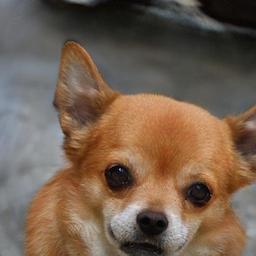} 
        \\
        \includegraphics[width=0.25\columnwidth]{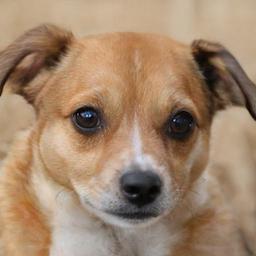} &
        \includegraphics[width=0.25\columnwidth]{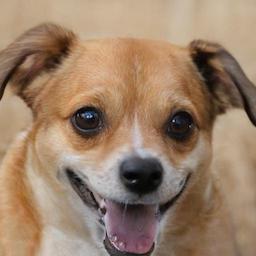} &
        \includegraphics[width=0.25\columnwidth]{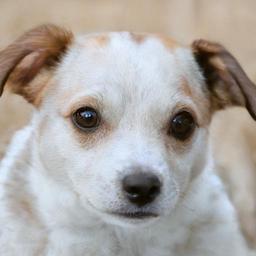} &
        \includegraphics[width=0.25\columnwidth]{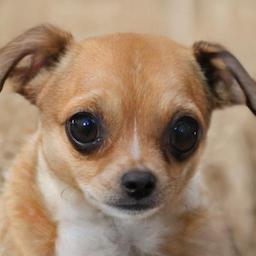} &
        \includegraphics[width=0.25\columnwidth]{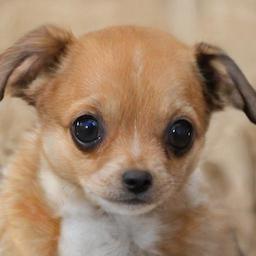} \\
        \includegraphics[width=0.25\columnwidth]{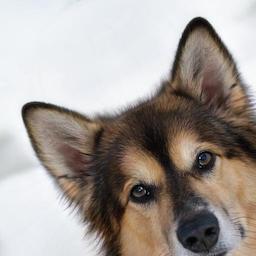} &
        \includegraphics[width=0.25\columnwidth]{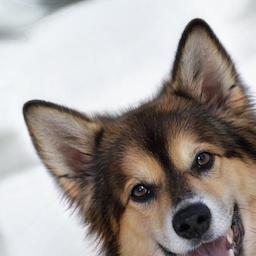} &
        \includegraphics[width=0.25\columnwidth]{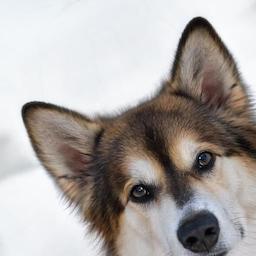} &
        \includegraphics[width=0.25\columnwidth]{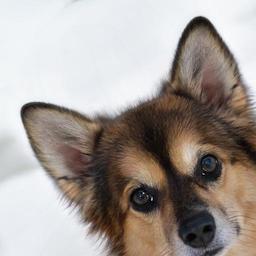} &
        \includegraphics[width=0.25\columnwidth]{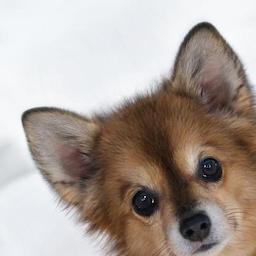} 
        \\
        \includegraphics[width=0.25\columnwidth]{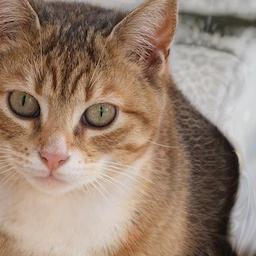} &
        \includegraphics[width=0.25\columnwidth]{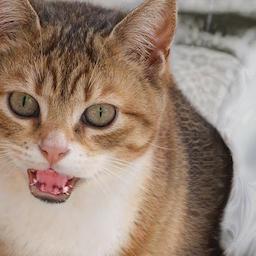} &
        \includegraphics[width=0.25\columnwidth]{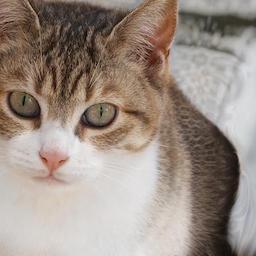} &
        \includegraphics[width=0.25\columnwidth]{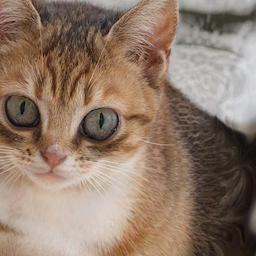} &
        \includegraphics[width=0.25\columnwidth]{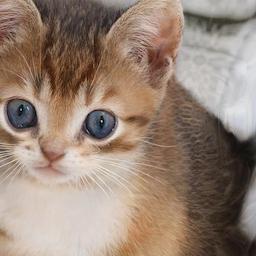} 
        \\

        \includegraphics[width=0.25\columnwidth]{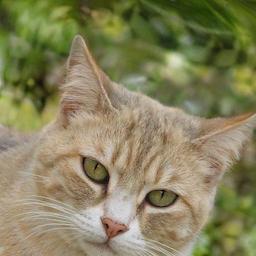} &
        \includegraphics[width=0.25\columnwidth]{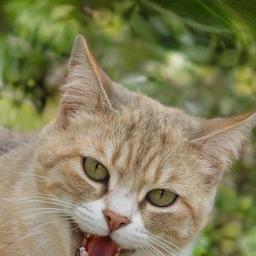} &
        \includegraphics[width=0.25\columnwidth]{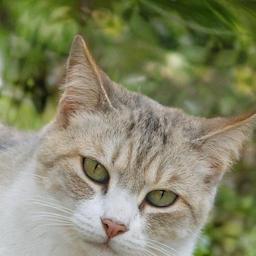} &
        \includegraphics[width=0.25\columnwidth]{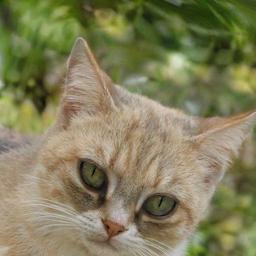} &
        \includegraphics[width=0.25\columnwidth]{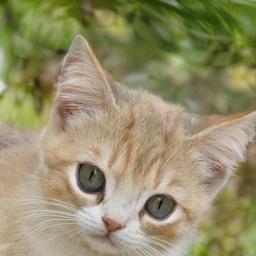} 
        \\
        \includegraphics[width=0.25\columnwidth]{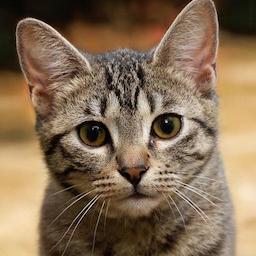} &
        \includegraphics[width=0.25\columnwidth]{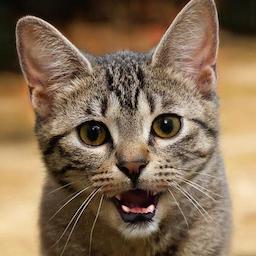} &
        \includegraphics[width=0.25\columnwidth]{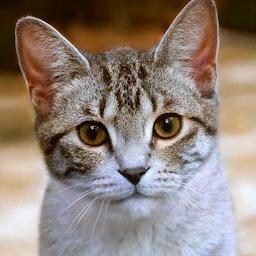} &
        \includegraphics[width=0.25\columnwidth]{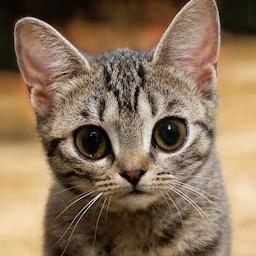} &
        \includegraphics[width=0.25\columnwidth]{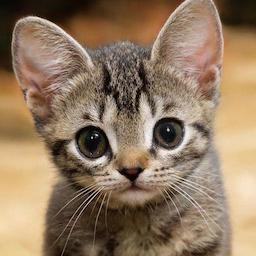} 
        \\
        Source & Happy & White & Big Eyes & Baby
	\end{tabular}
	}
	\vspace{-0.1cm}
	\caption{Editing in $\mathcal{S}$. Editing results on synthetic images obtained using StyleCLIP's global direction technique~\cite{patashnik2021styleclip} applied over an aligned StyleGAN3 generator trained on the AFHQv2~\cite{choi2020stargan,aliasfreeKarras2021} dataset. The unaligned results are obtained by randomly sampling the rotation and translation parameters $(r,t_x,t_y)$ applied over the input Fourier features of the generator.
	}
	\label{fig:edit_afhq_supp}
\end{figure*}

%% file: figures/supplementary/edits_lands.tex
\begin{figure*}[tb]
	\centering
	\setlength{\tabcolsep}{1pt}	
	{\small
	\begin{tabular}{c c c c c c}

        \includegraphics[width=0.24\columnwidth]{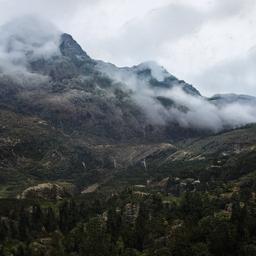} &
        \includegraphics[width=0.24\columnwidth]{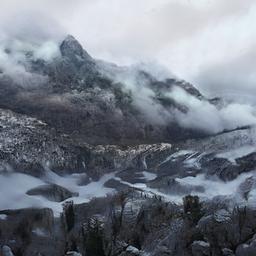} &
        \includegraphics[width=0.24\columnwidth]{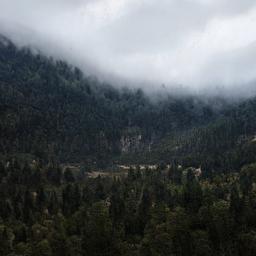} &
        \includegraphics[width=0.24\columnwidth]{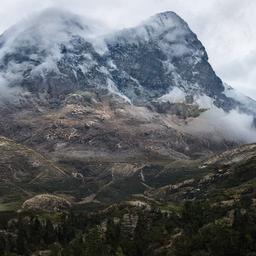} &
        \includegraphics[width=0.24\columnwidth]{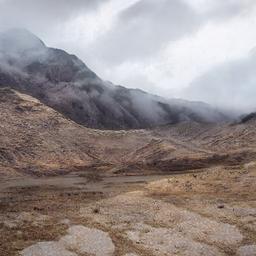} &
        \\
        \includegraphics[width=0.24\columnwidth]{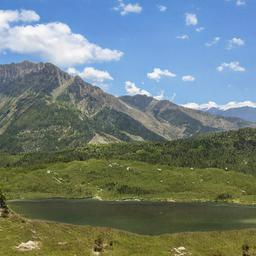} &
        \includegraphics[width=0.24\columnwidth]{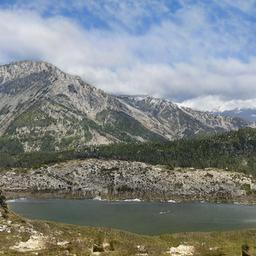} &
        \includegraphics[width=0.24\columnwidth]{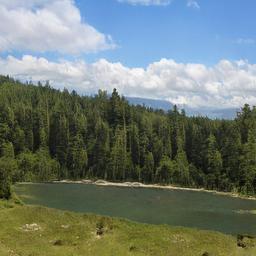} &
        \includegraphics[width=0.24\columnwidth]{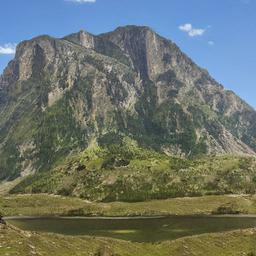} &
        \includegraphics[width=0.24\columnwidth]{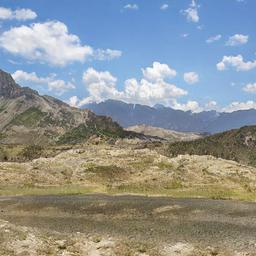} &
        \\
        \includegraphics[width=0.24\columnwidth]{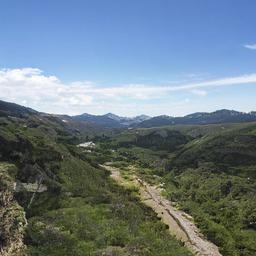} &
        \includegraphics[width=0.24\columnwidth]{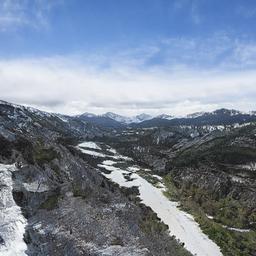} &
        \includegraphics[width=0.24\columnwidth]{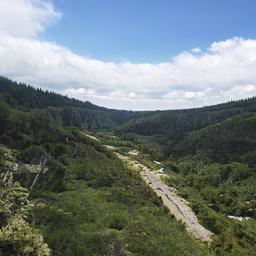} &
        \includegraphics[width=0.24\columnwidth]{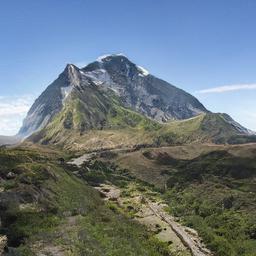} &
        \includegraphics[width=0.24\columnwidth]{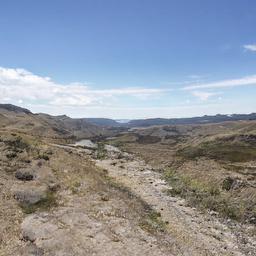} &
        \\
        \includegraphics[width=0.24\columnwidth]{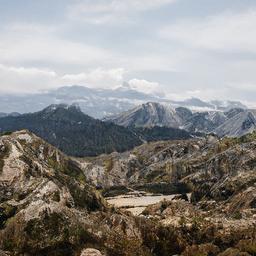} &
        \includegraphics[width=0.24\columnwidth]{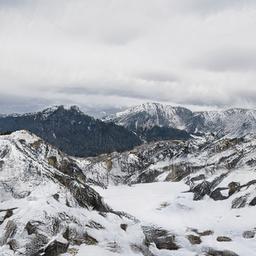} &
        \includegraphics[width=0.24\columnwidth]{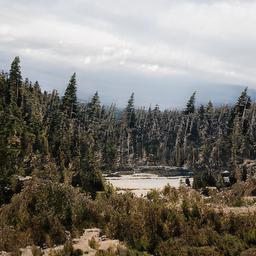} &
        \includegraphics[width=0.24\columnwidth]{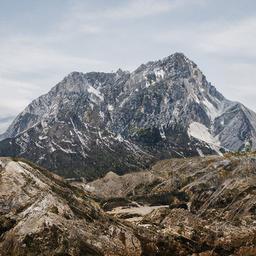} &
        \includegraphics[width=0.24\columnwidth]{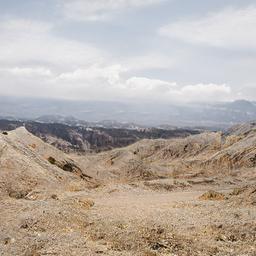} &
        \\
        
        \includegraphics[width=0.24\columnwidth]{images/editing/global_direction/lands/source/005_004.jpg} &
        \includegraphics[width=0.24\columnwidth]{images/editing/global_direction/lands/winter/005_008.jpg} &
        \includegraphics[width=0.24\columnwidth]{images/editing/global_direction/lands/forest/005_007.jpg} &
        \includegraphics[width=0.24\columnwidth]{images/editing/global_direction/lands/mountain/005_002.jpg} &
        \includegraphics[width=0.24\columnwidth]{images/editing/global_direction/lands/desert/005_006.jpg} &
        \\
        \includegraphics[width=0.24\columnwidth]{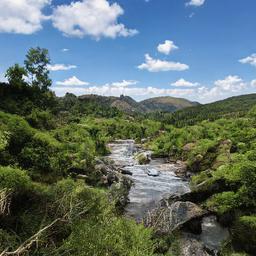} &
        \includegraphics[width=0.24\columnwidth]{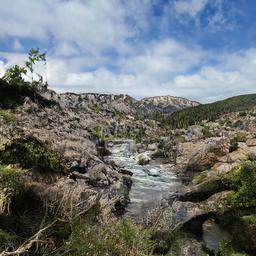} &
        \includegraphics[width=0.24\columnwidth]{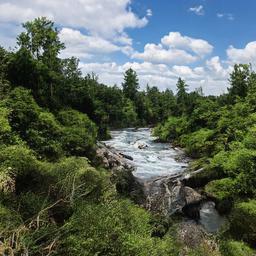} &
        \includegraphics[width=0.24\columnwidth]{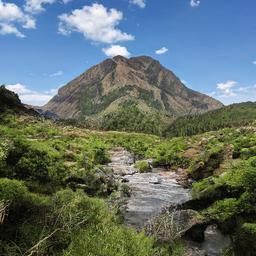} &
        \includegraphics[width=0.24\columnwidth]{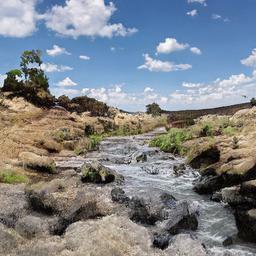} &
        \\
        \includegraphics[width=0.24\columnwidth]{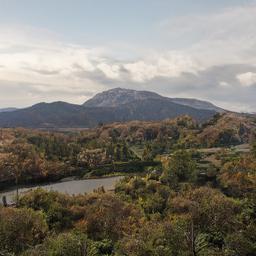} &
        \includegraphics[width=0.24\columnwidth]{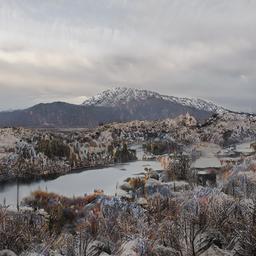} &
        \includegraphics[width=0.24\columnwidth]{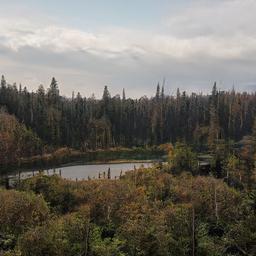} &
        \includegraphics[width=0.24\columnwidth]{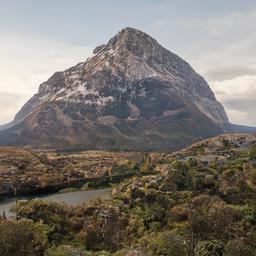} &
        \includegraphics[width=0.24\columnwidth]{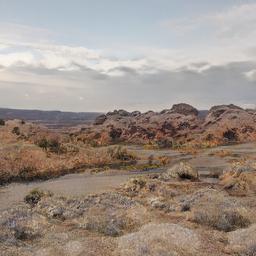} &
        \\
        \includegraphics[width=0.24\columnwidth]{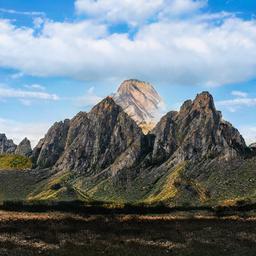} &
        \includegraphics[width=0.24\columnwidth]{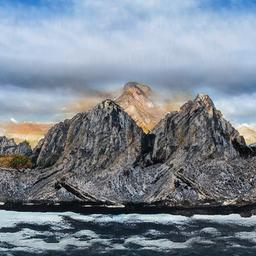} &
        \includegraphics[width=0.24\columnwidth]{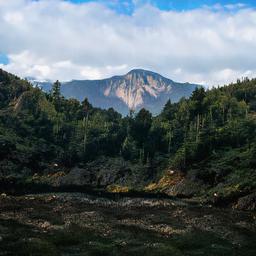} &
        \includegraphics[width=0.24\columnwidth]{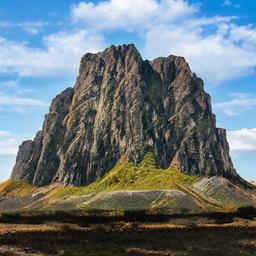} &
        \includegraphics[width=0.24\columnwidth]{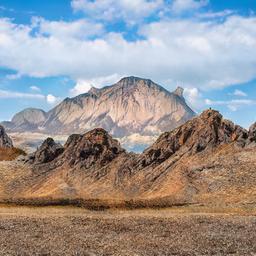} &
        \\
        
        \includegraphics[width=0.24\columnwidth]{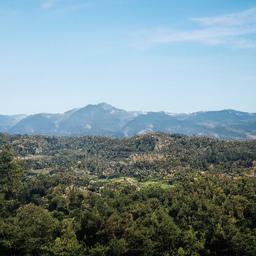} &
        \includegraphics[width=0.24\columnwidth]{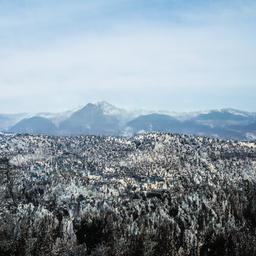} &
        \includegraphics[width=0.24\columnwidth]{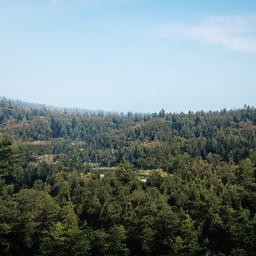} &
        \includegraphics[width=0.24\columnwidth]{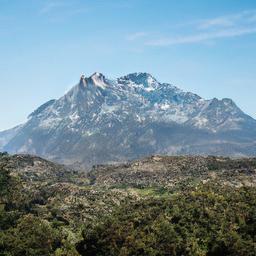} &
        \includegraphics[width=0.24\columnwidth]{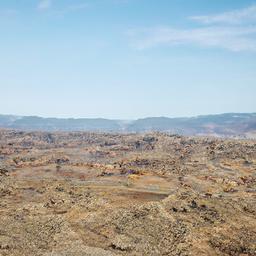} &
        \\
        
        Source & Winter & Forest & Mountain & Desert
	\end{tabular}
	}
	\vspace{-0.3cm}
	\caption{Editing in $\mathcal{S}$. Editing results on synthetic images obtained using StyleCLIP's global direction technique~\cite{patashnik2021styleclip} applied over an aligned StyleGAN3 generator trained on the Landscapes HQ~\cite{ALIS} dataset.}
	\label{fig:edit_landscapes_supp}
\end{figure*}

%% file: figures/supplementary/nada_images.tex
\begin{figure*}[tb]
	\centering
	\setlength{\tabcolsep}{1pt}	

	\begin{tabular}{c c c c c c}
        \raisebox{0.05in}{\rotatebox{90}{}} &
        \includegraphics[width=0.26\columnwidth]{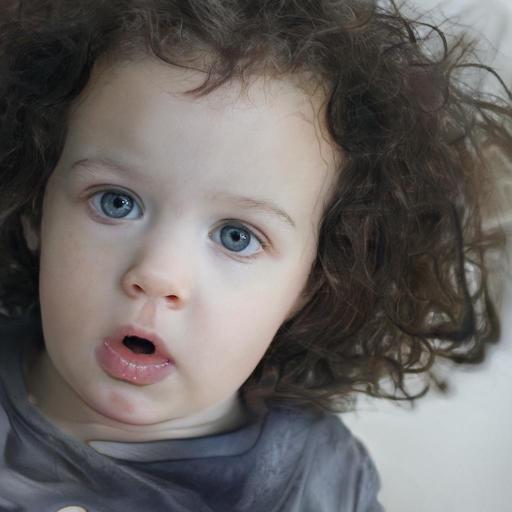} &
        \includegraphics[width=0.26\columnwidth]{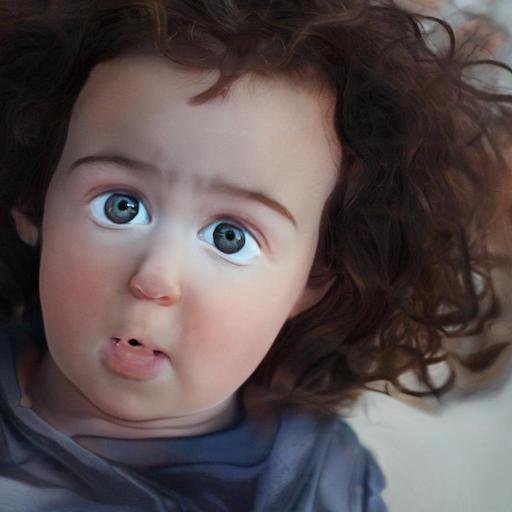} &
        \includegraphics[width=0.26\columnwidth]{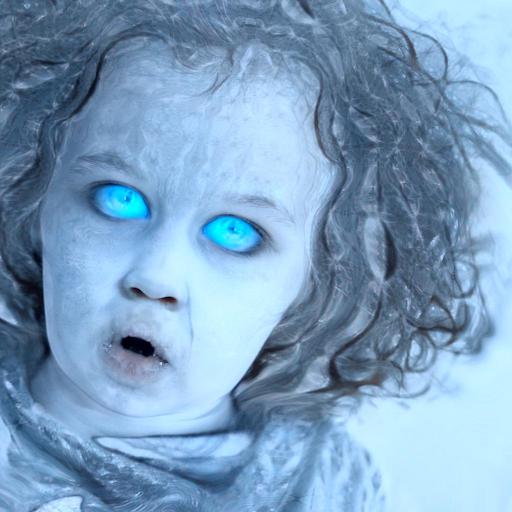} &
        \includegraphics[width=0.26\columnwidth]{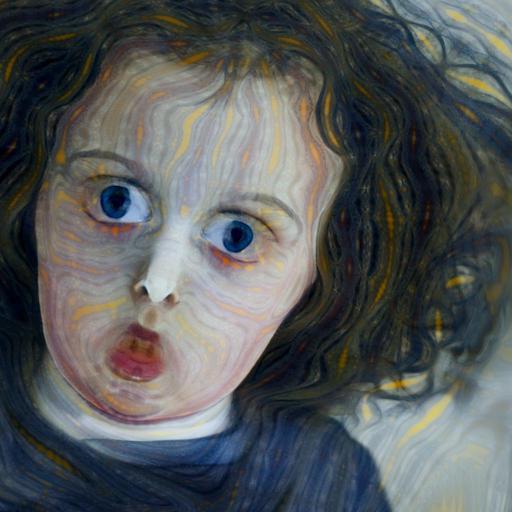} &        \includegraphics[width=0.26\columnwidth]{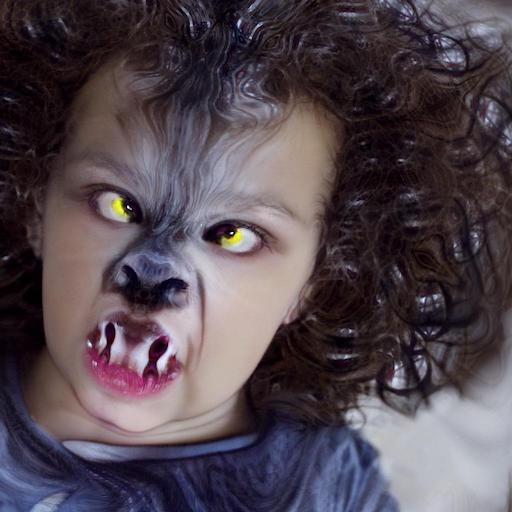}
        \\
        \raisebox{0.05in}{\rotatebox{90}{}} &
        \includegraphics[width=0.26\columnwidth]{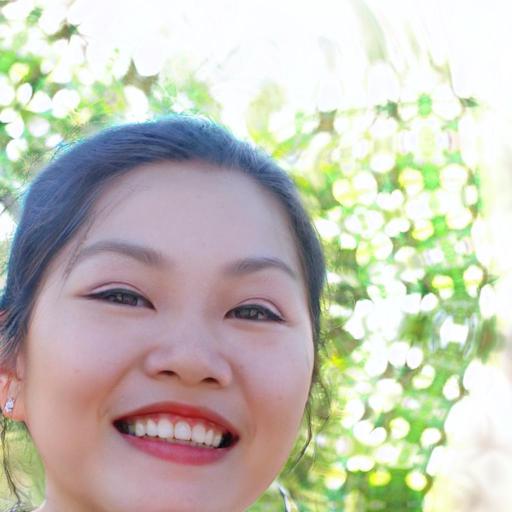} &
        \includegraphics[width=0.26\columnwidth]{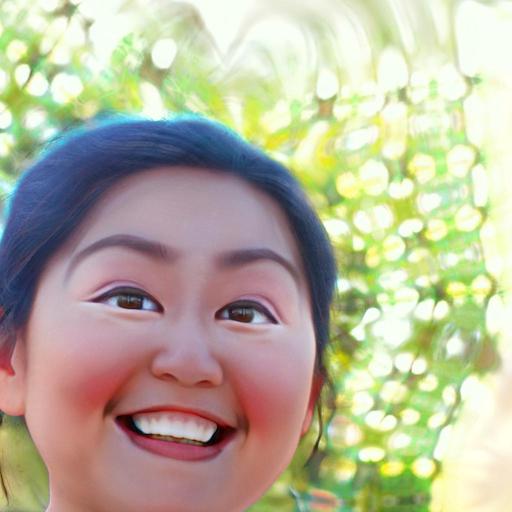} &
        \includegraphics[width=0.26\columnwidth]{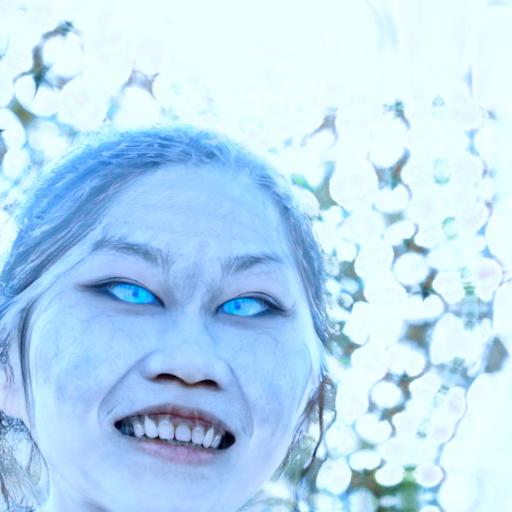} &
        \includegraphics[width=0.26\columnwidth]{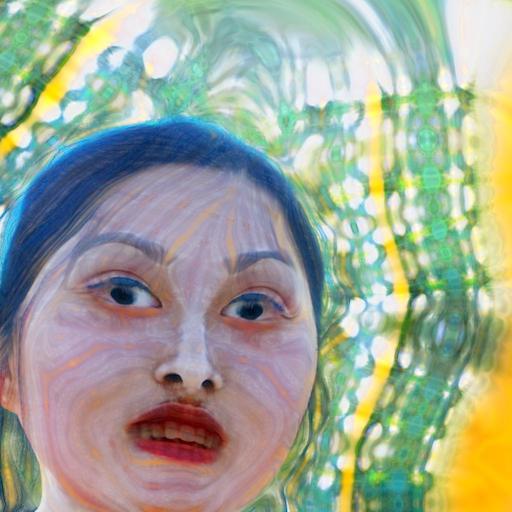} &        \includegraphics[width=0.26\columnwidth]{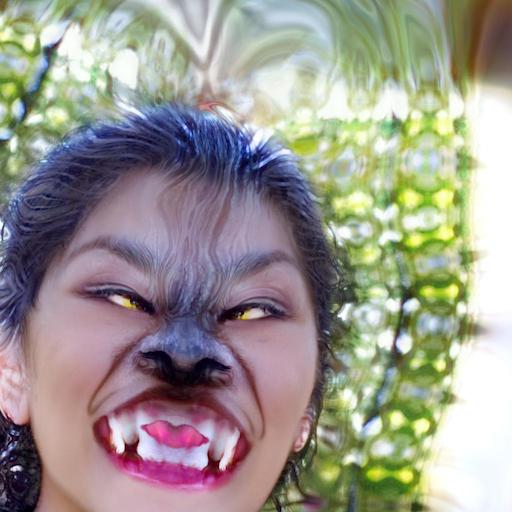}
        
        \\
        \raisebox{0.05in}{\rotatebox{90}{}} &
        \includegraphics[width=0.26\columnwidth]{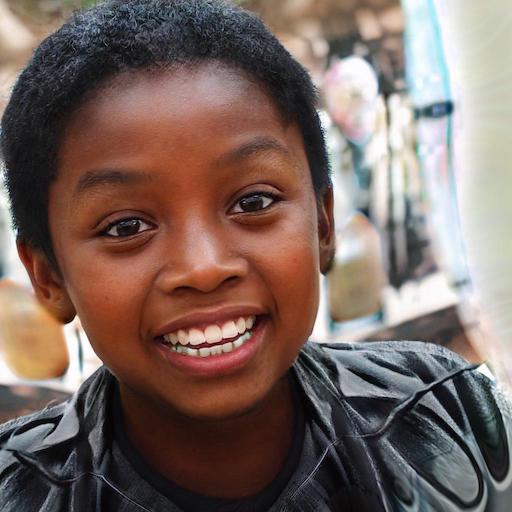} &
        \includegraphics[width=0.26\columnwidth]{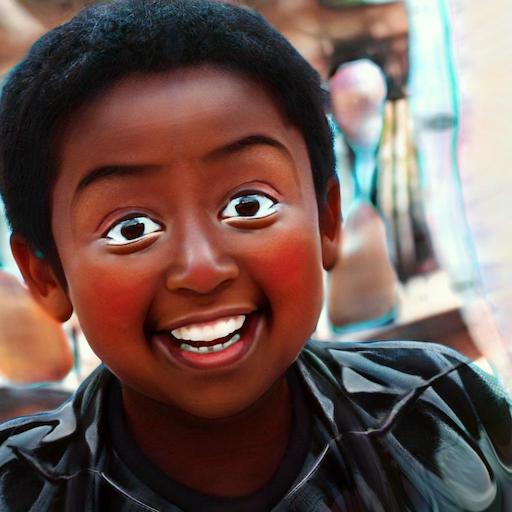} &
        \includegraphics[width=0.26\columnwidth]{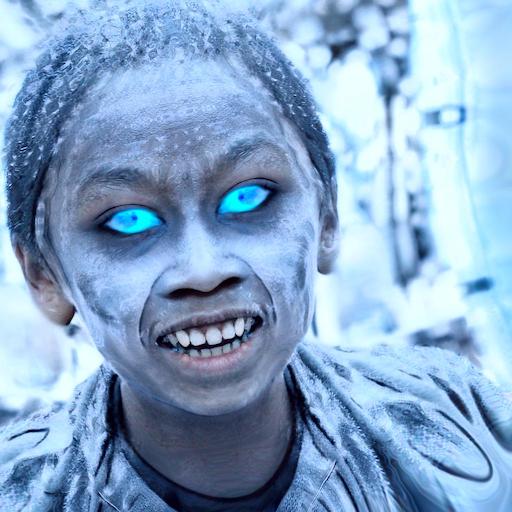} &
        \includegraphics[width=0.26\columnwidth]{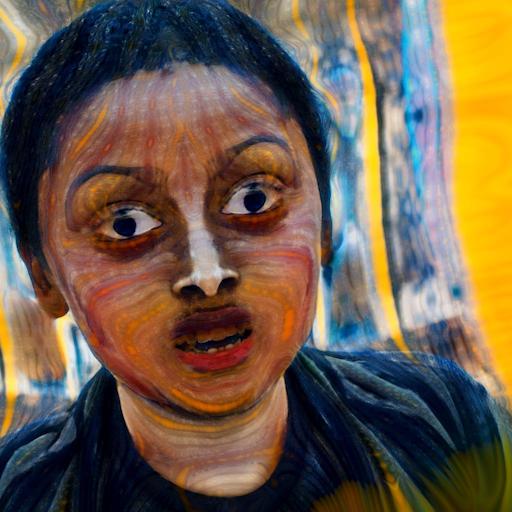} &        \includegraphics[width=0.26\columnwidth]{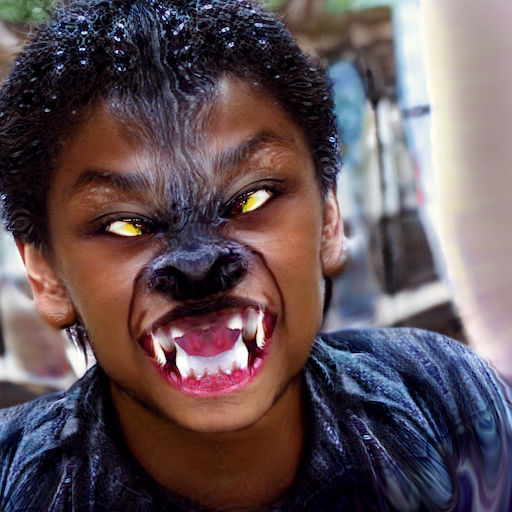}
        \\
        \raisebox{0.05in}{\rotatebox{90}{}} &
        \includegraphics[width=0.26\columnwidth]{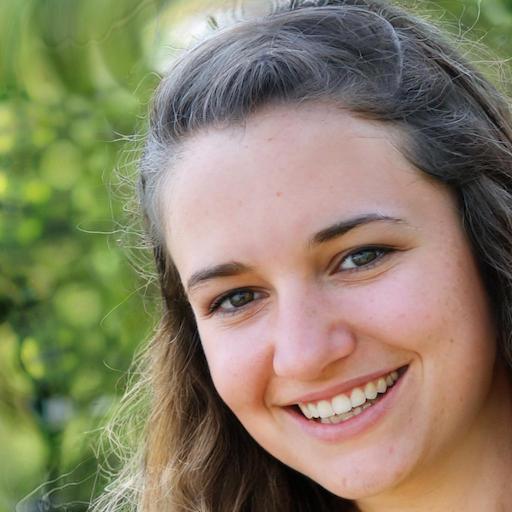} &
        \includegraphics[width=0.26\columnwidth]{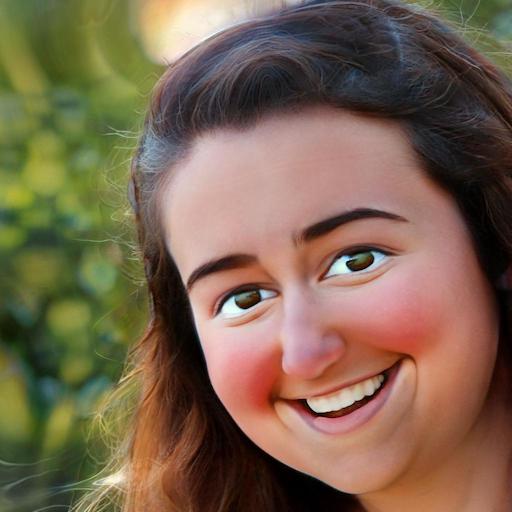} &
        \includegraphics[width=0.26\columnwidth]{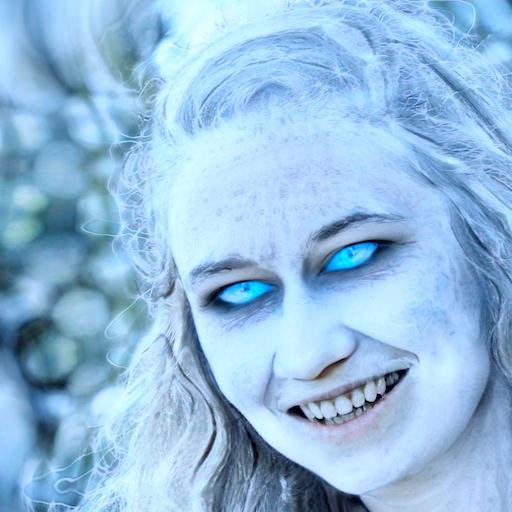} &
        \includegraphics[width=0.26\columnwidth]{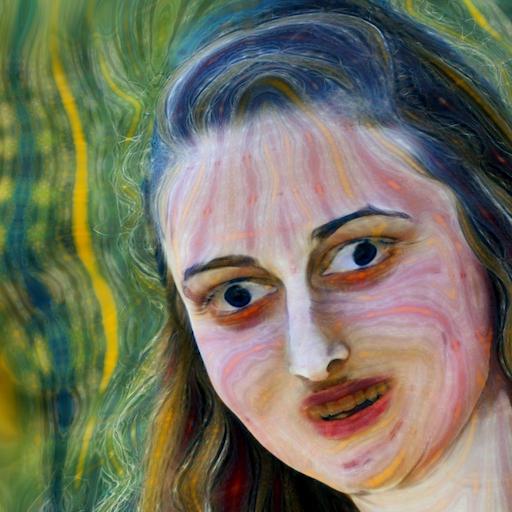} &        \includegraphics[width=0.26\columnwidth]{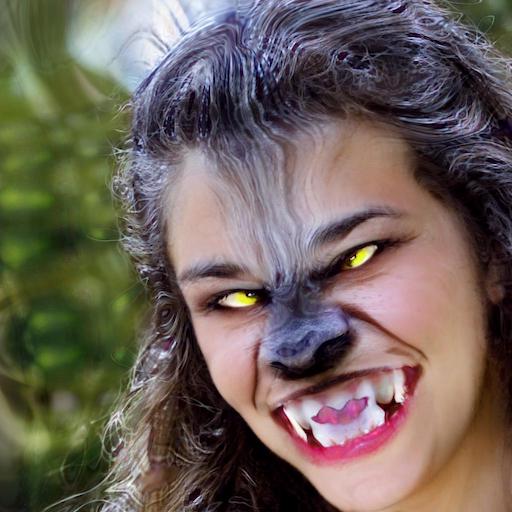}
        
        \\
		 & Source & Pixar & White walker & Munch painting & Werewolf
        
    \\

    \end{tabular}
	
	\begin{tabular}{c c c c c c}
        \raisebox{0.05in}{\rotatebox{90}{}} &
        \includegraphics[width=0.26\columnwidth]{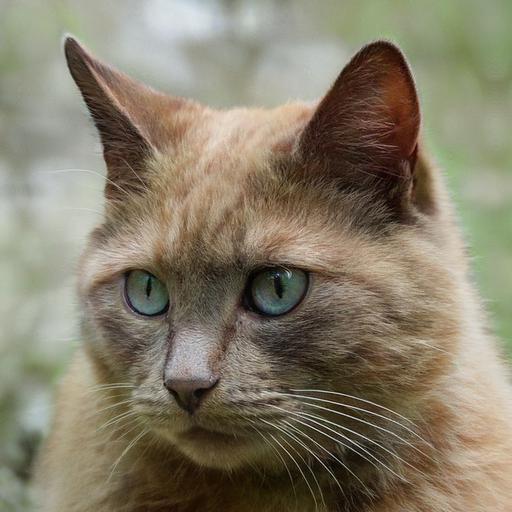} &
        \includegraphics[width=0.26\columnwidth]{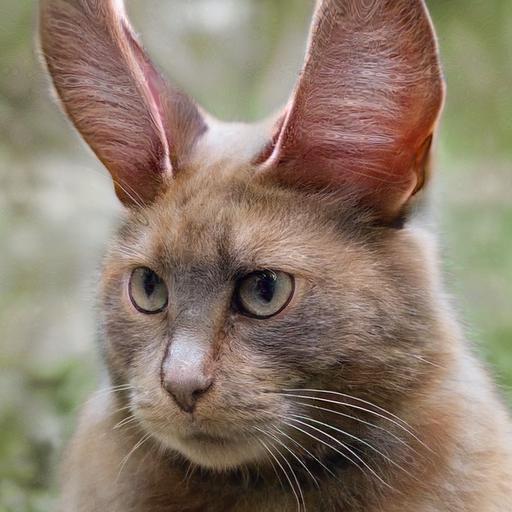} &
        \includegraphics[width=0.26\columnwidth]{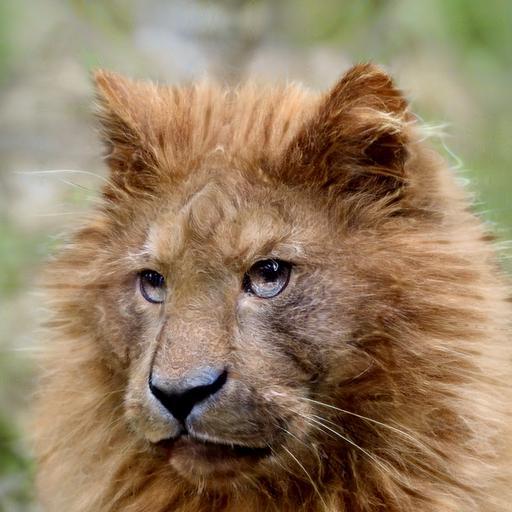} &
        \includegraphics[width=0.26\columnwidth]{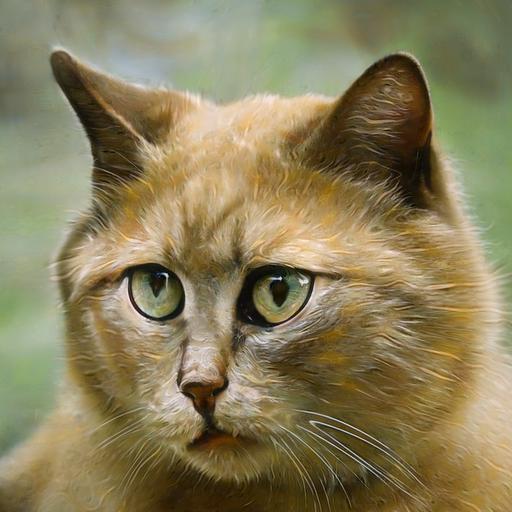} &        \includegraphics[width=0.26\columnwidth]{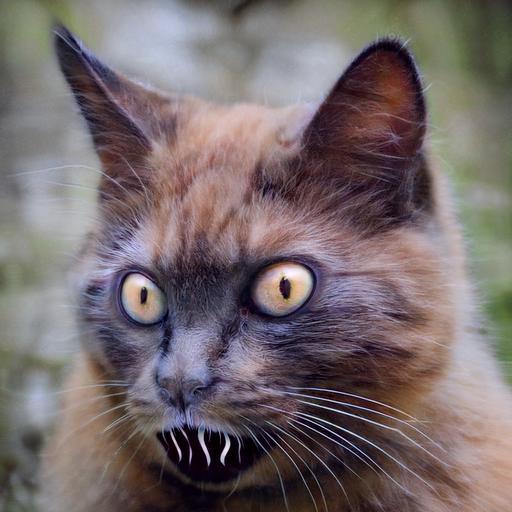}
        \\
        \raisebox{0.05in}{\rotatebox{90}{}} &
        \includegraphics[width=0.26\columnwidth]{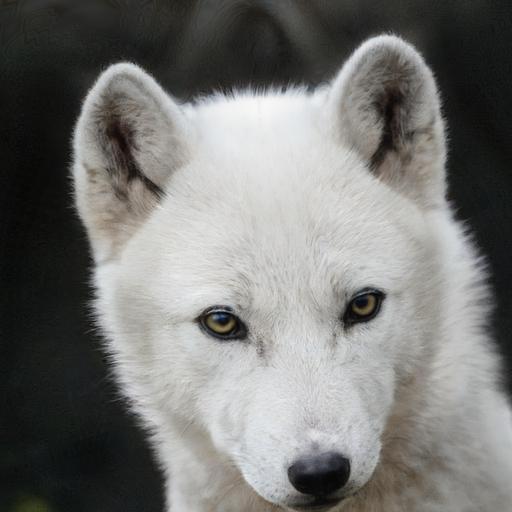} &
        \includegraphics[width=0.26\columnwidth]{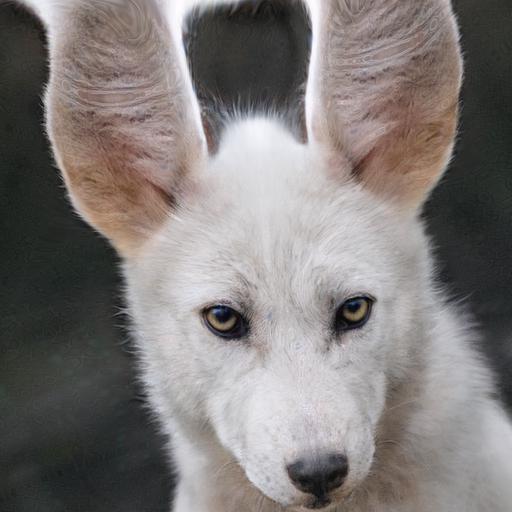} &
        \includegraphics[width=0.26\columnwidth]{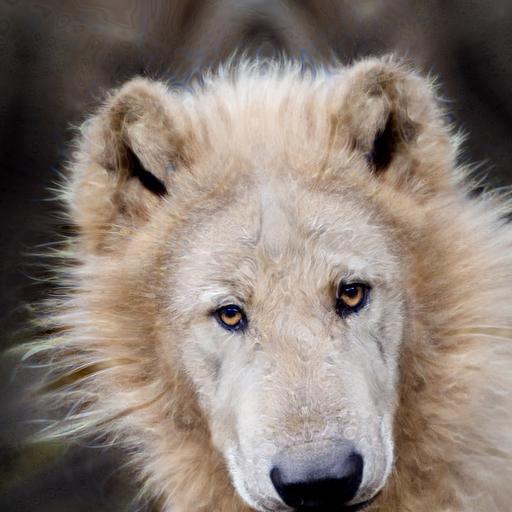} &
        \includegraphics[width=0.26\columnwidth]{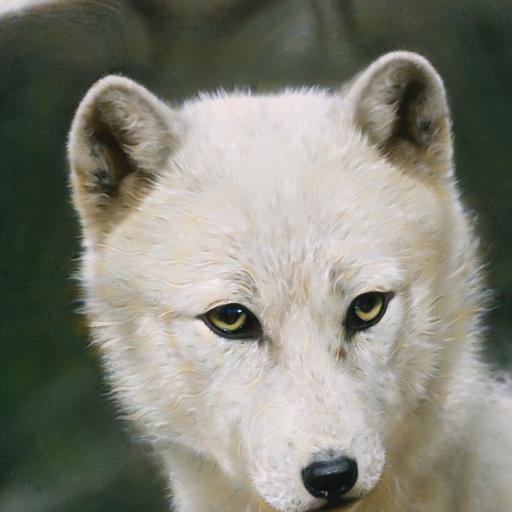} &        \includegraphics[width=0.26\columnwidth]{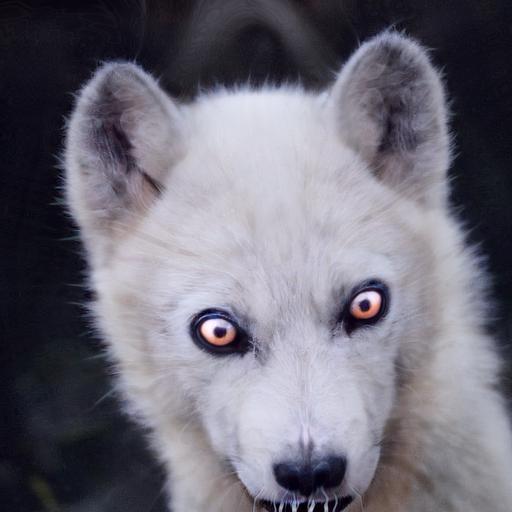}
        \\
        \raisebox{0.05in}{\rotatebox{90}{}} &
        \includegraphics[width=0.26\columnwidth]{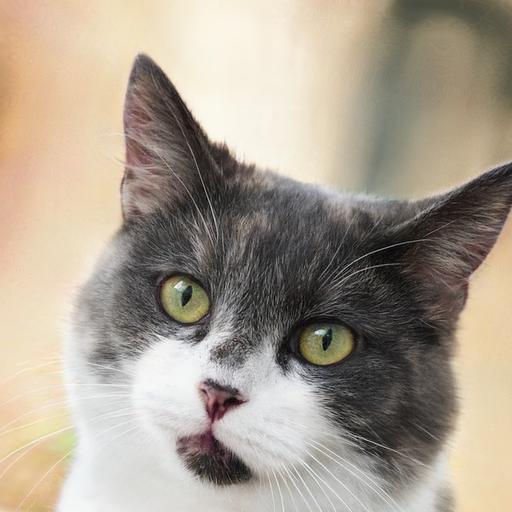} &
        \includegraphics[width=0.26\columnwidth]{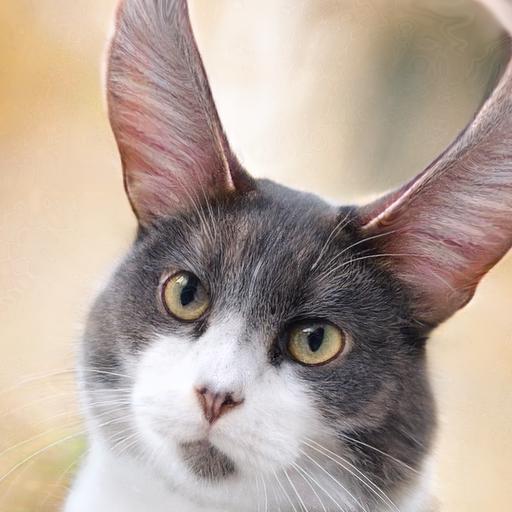} &
        \includegraphics[width=0.26\columnwidth]{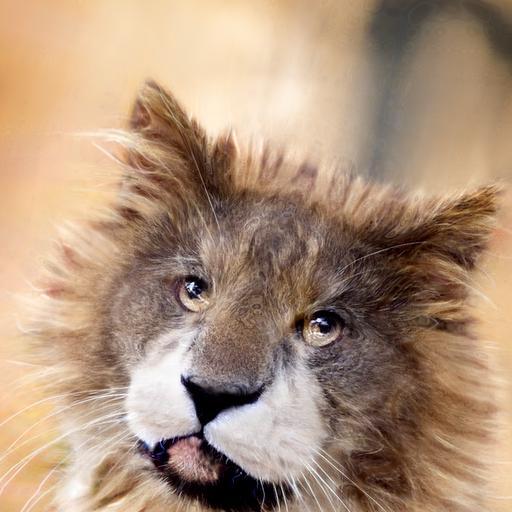} &
        \includegraphics[width=0.26\columnwidth]{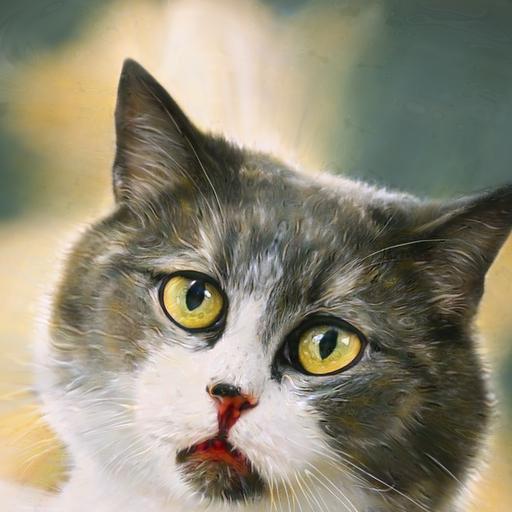} &        \includegraphics[width=0.26\columnwidth]{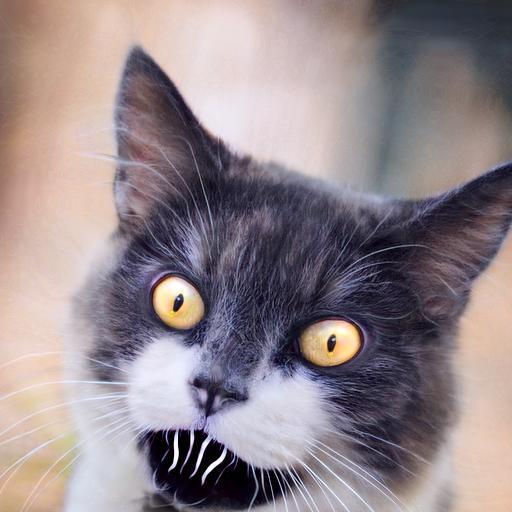}
        \\
        \raisebox{0.05in}{\rotatebox{90}{}} &
        \includegraphics[width=0.26\columnwidth]{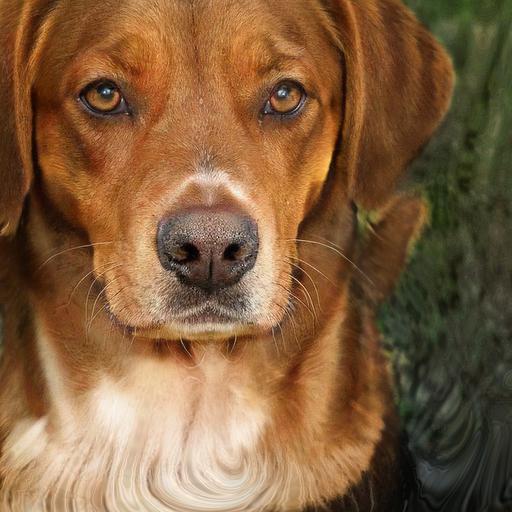} &
        \includegraphics[width=0.26\columnwidth]{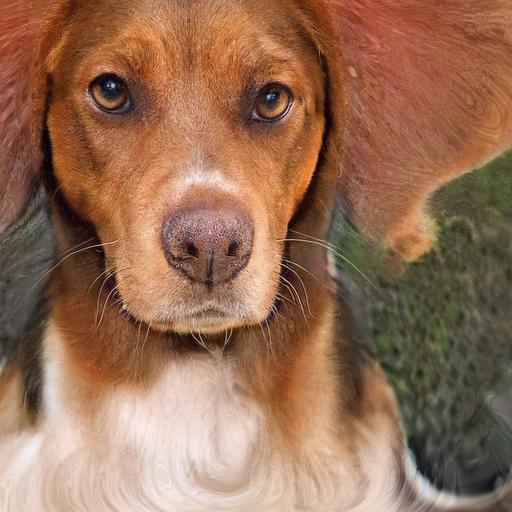} &
        \includegraphics[width=0.26\columnwidth]{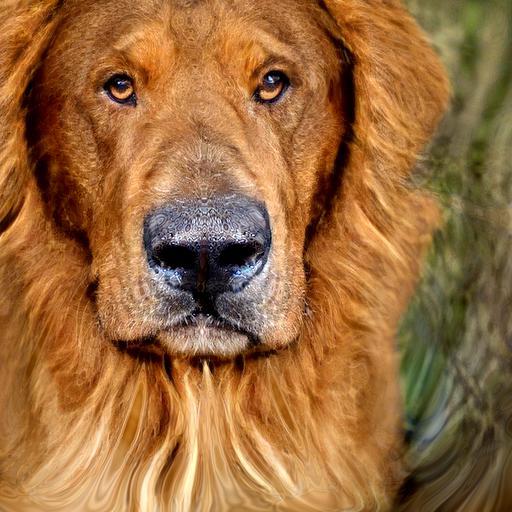} &
        \includegraphics[width=0.26\columnwidth]{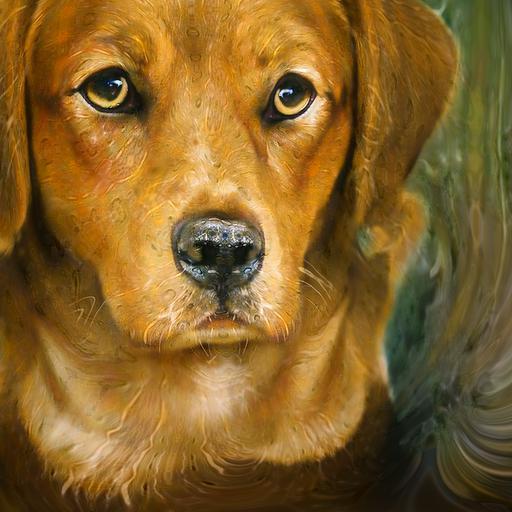} &        \includegraphics[width=0.26\columnwidth]{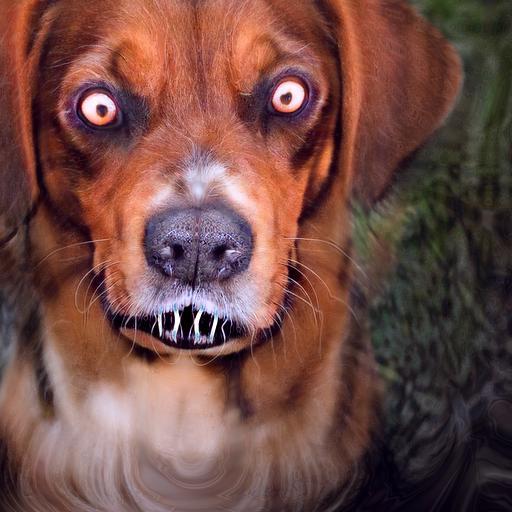}
        \\
		 & Source & Big ears & Lion & Painting & Scary 

	\end{tabular}
	
    \vspace{-0.1cm}
	\caption{Fine-tuning StyleGAN3~\cite{aliasfreeKarras2021} with StyleGAN-NADA~\cite{gal2021stylegannada}. As shown, the latent spaces of the original parent generator and child generators remain aligned under fine-tuning.}
	\label{fig:stylegan_nada_adaptation_supplementary}
\end{figure*}

%% file: figures/supplementary/nada_editing.tex
\begin{figure*}[tb]
	\centering
	\setlength{\tabcolsep}{1pt}	
	{\footnotesize
	\begin{tabular}{c c c c c c}
        \raisebox{0.05in}{\rotatebox{90}{}} &
        \includegraphics[width=0.26\columnwidth]{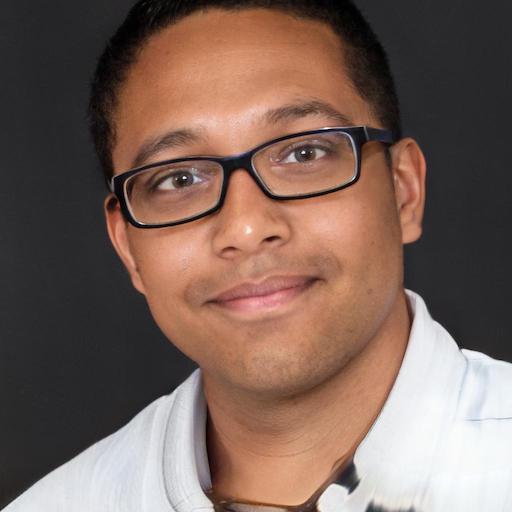} &
        \includegraphics[width=0.26\columnwidth]{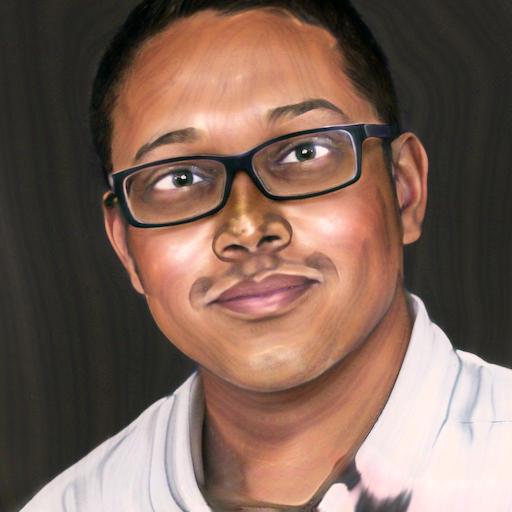} &
        \includegraphics[width=0.26\columnwidth]{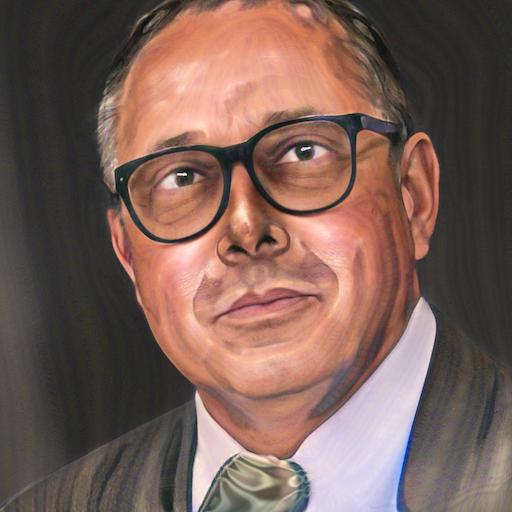} &
        \includegraphics[width=0.26\columnwidth]{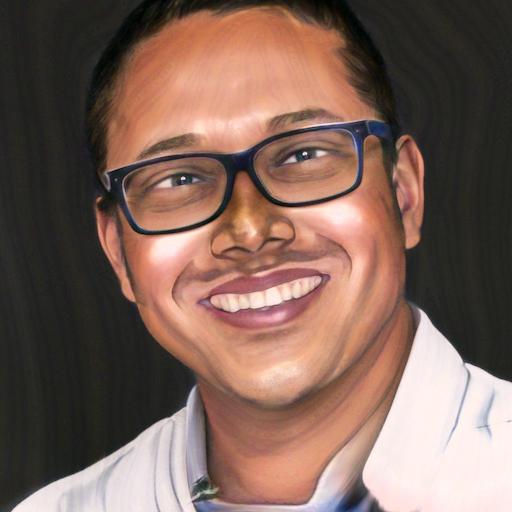} &
        \includegraphics[width=0.26\columnwidth]{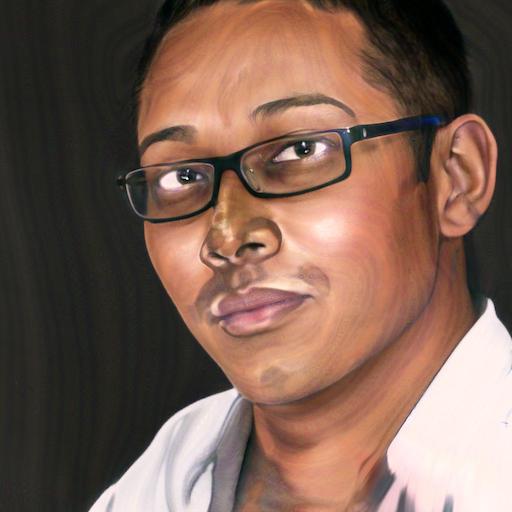}
        \\
        \raisebox{0.05in}{\rotatebox{90}{}} &
        \includegraphics[width=0.26\columnwidth]{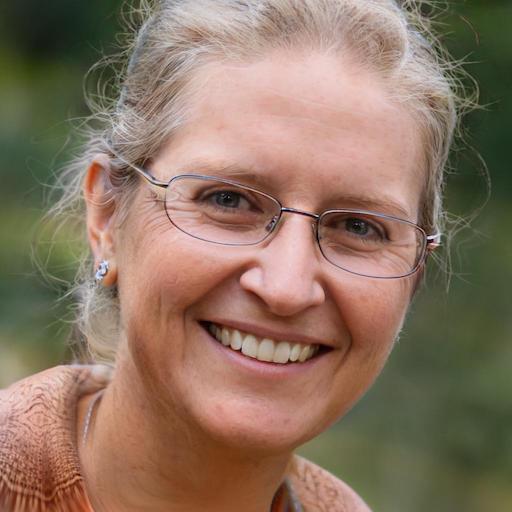} &
        \includegraphics[width=0.26\columnwidth]{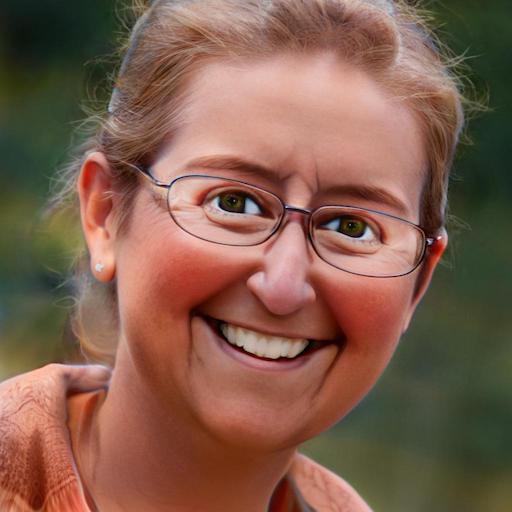} &
        \includegraphics[width=0.26\columnwidth]{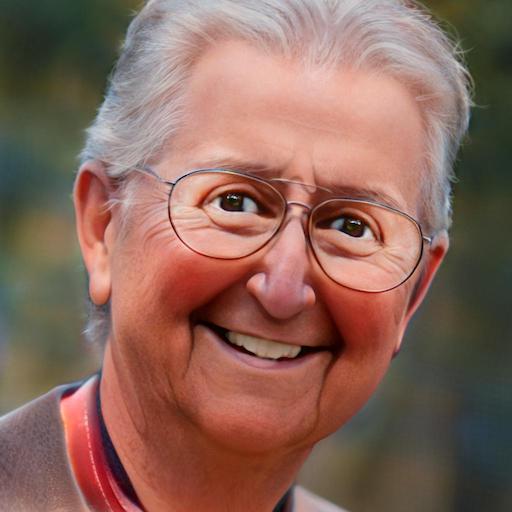} &
        \includegraphics[width=0.26\columnwidth]{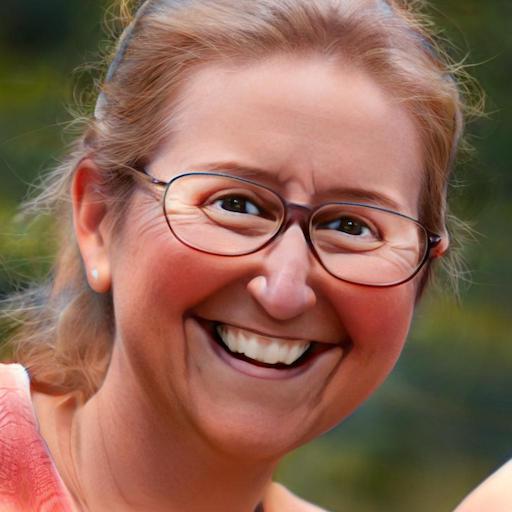} & 
        \includegraphics[width=0.26\columnwidth]{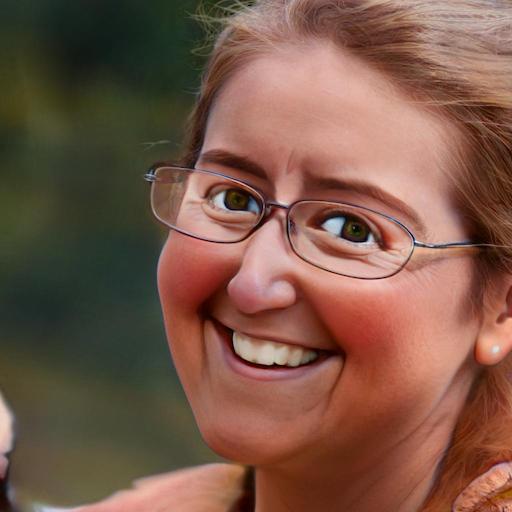}
        
        \\
        \raisebox{0.05in}{\rotatebox{90}{}} &
        \includegraphics[width=0.26\columnwidth]{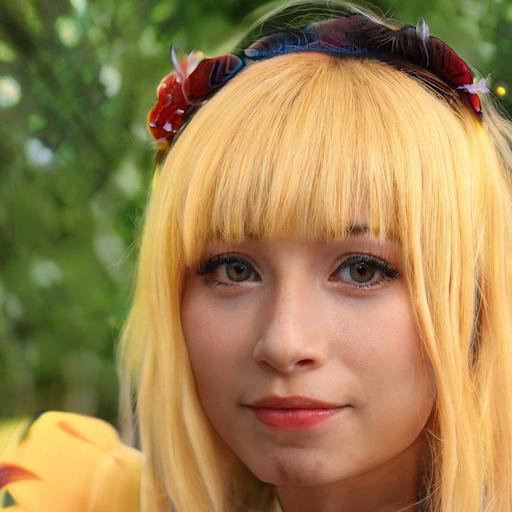} &
        \includegraphics[width=0.26\columnwidth]{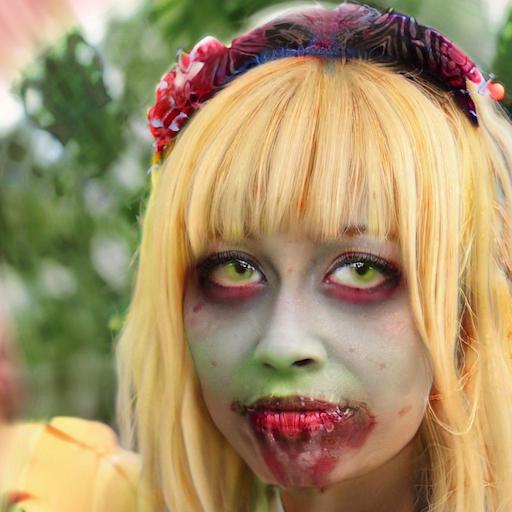} &
        \includegraphics[width=0.26\columnwidth]{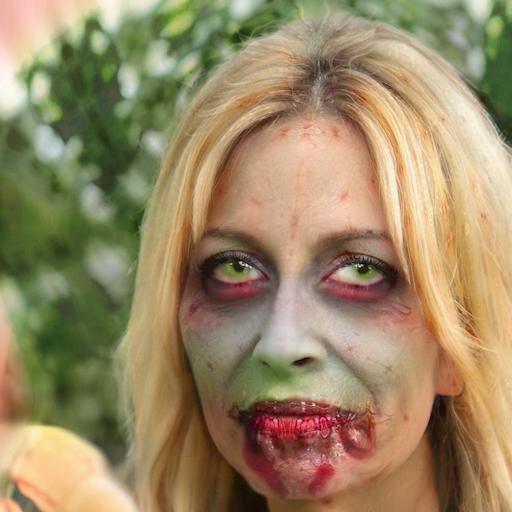} &
        \includegraphics[width=0.26\columnwidth]{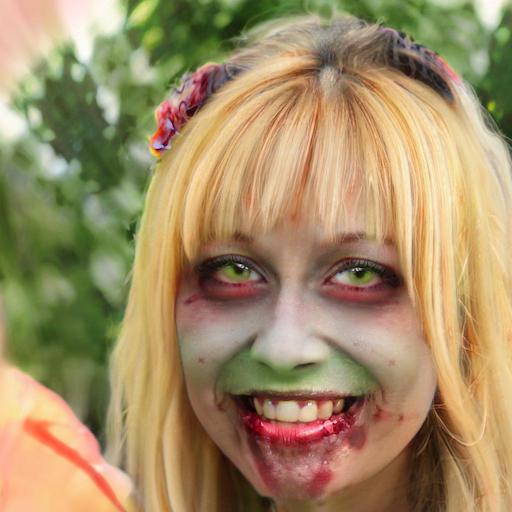} &
        \includegraphics[width=0.26\columnwidth]{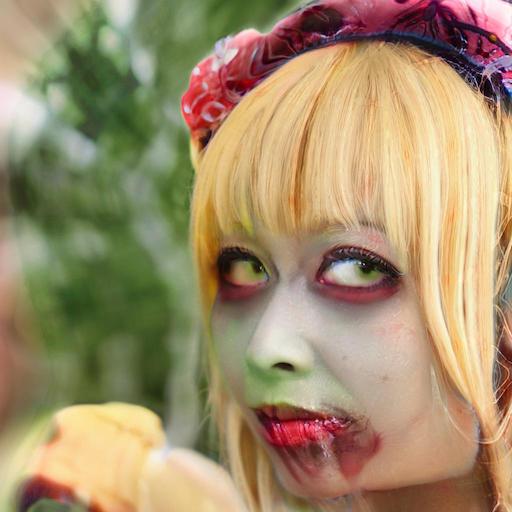}
        \\
        \raisebox{0.05in}{\rotatebox{90}{}} &
        \includegraphics[width=0.26\columnwidth]{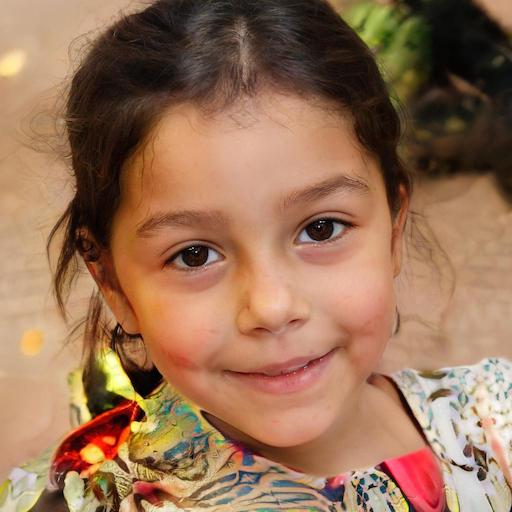} &
        \includegraphics[width=0.26\columnwidth]{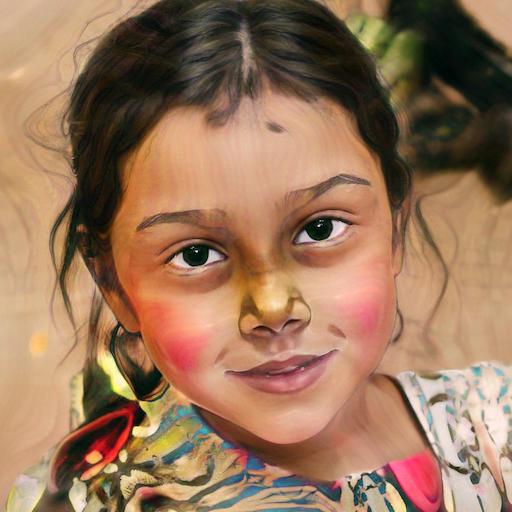} &
        \includegraphics[width=0.26\columnwidth]{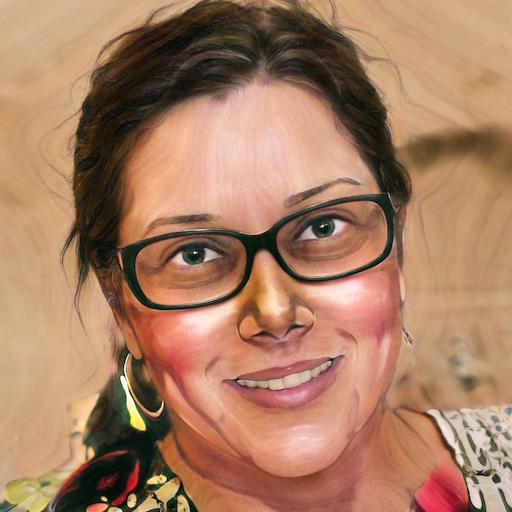} &
        \includegraphics[width=0.26\columnwidth]{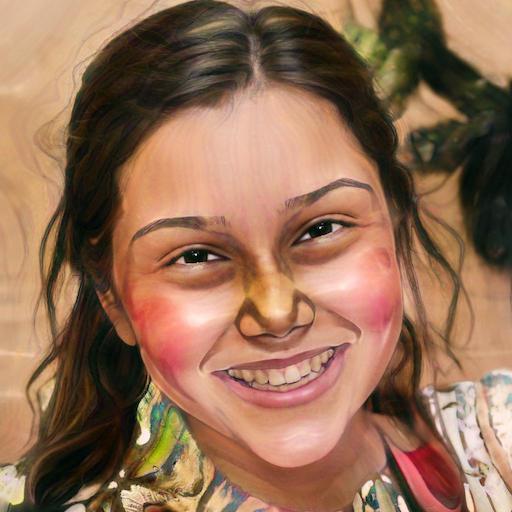} &       
        \includegraphics[width=0.26\columnwidth]{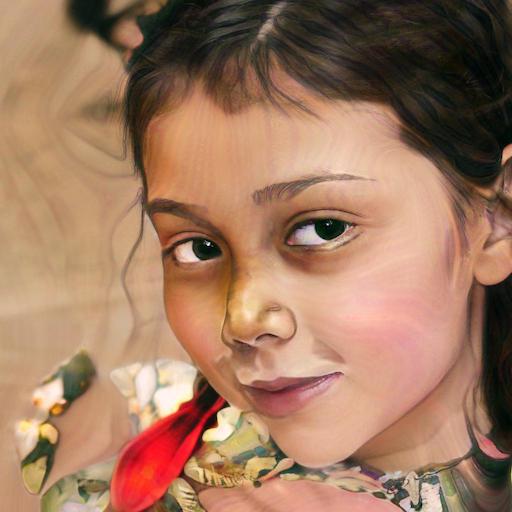}
        
        \\
        \raisebox{0.05in}{\rotatebox{90}{}} &
        \includegraphics[width=0.26\columnwidth]{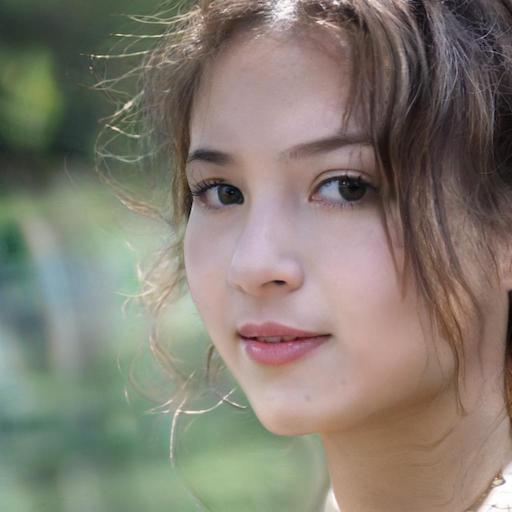} &
        \includegraphics[width=0.26\columnwidth]{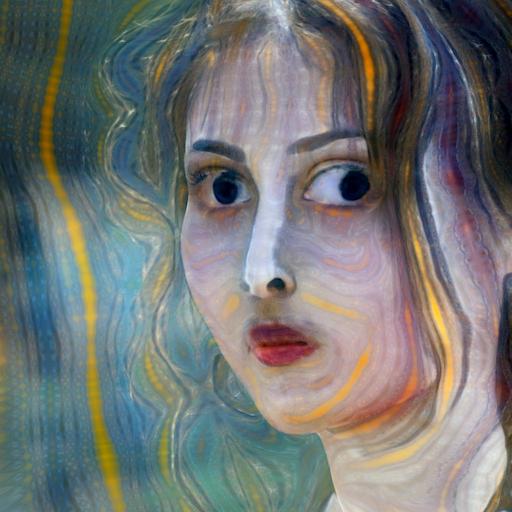} &
        \includegraphics[width=0.26\columnwidth]{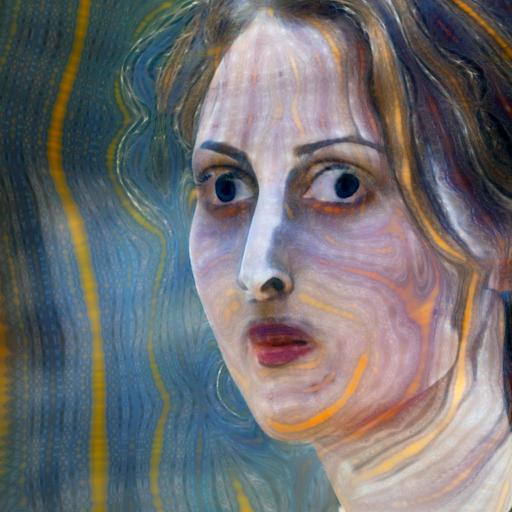} &
        \includegraphics[width=0.26\columnwidth]{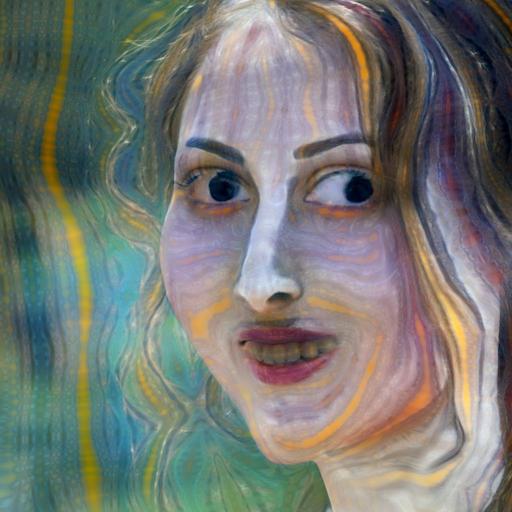} &
        \includegraphics[width=0.26\columnwidth]{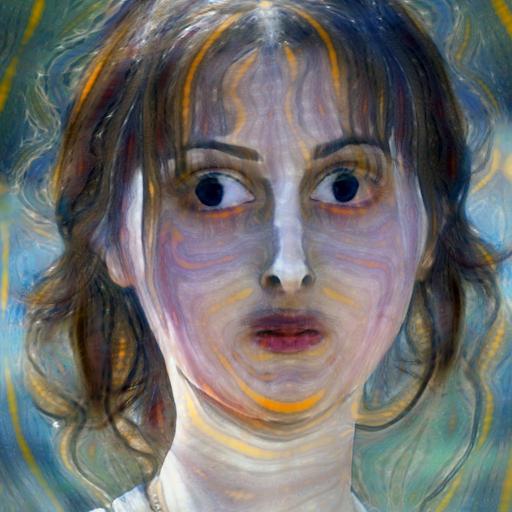}
        
        \\
        \raisebox{0.05in}{\rotatebox{90}{}} &
        \includegraphics[width=0.26\columnwidth]{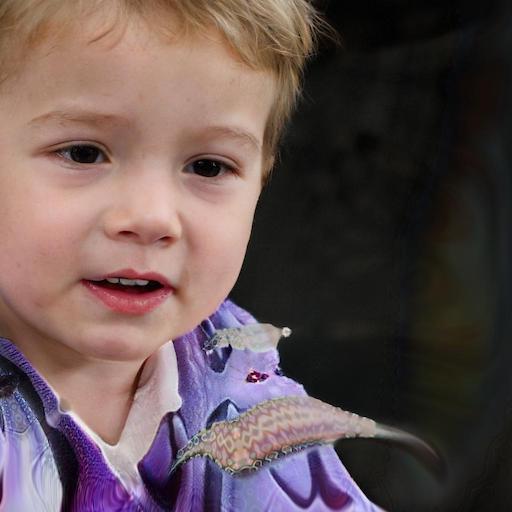} &
        \includegraphics[width=0.26\columnwidth]{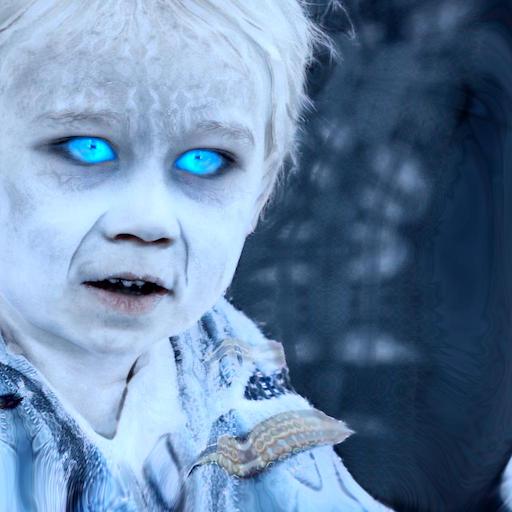} &
        \includegraphics[width=0.26\columnwidth]{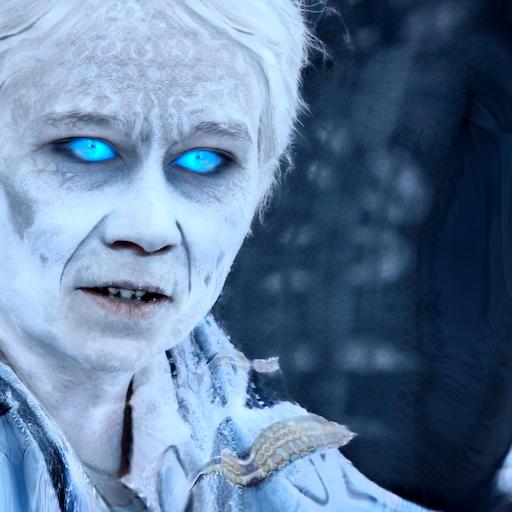} &
        \includegraphics[width=0.26\columnwidth]{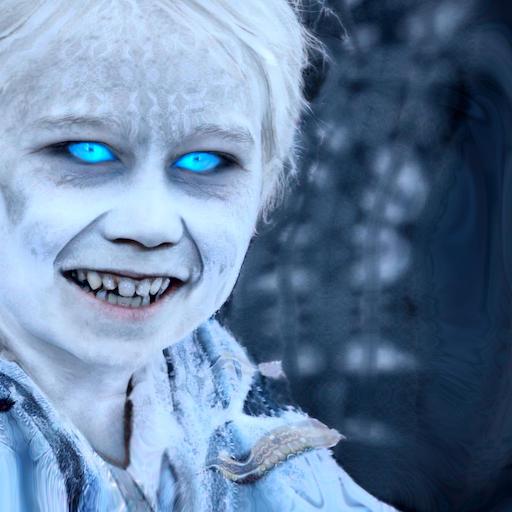} & 
        \includegraphics[width=0.26\columnwidth]{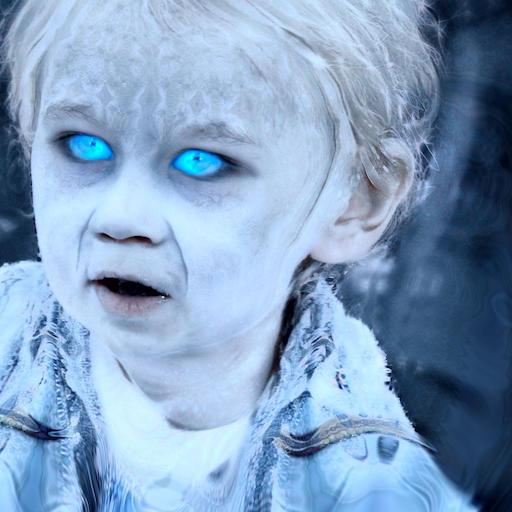}
        
        \\
        \raisebox{0.05in}{\rotatebox{90}{}} &
        \includegraphics[width=0.26\columnwidth]{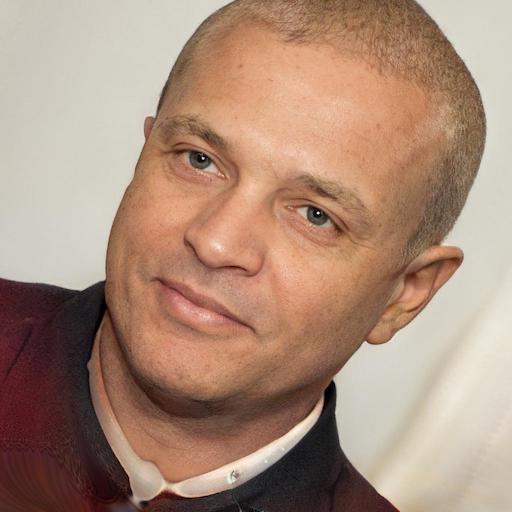} &
        \includegraphics[width=0.26\columnwidth]{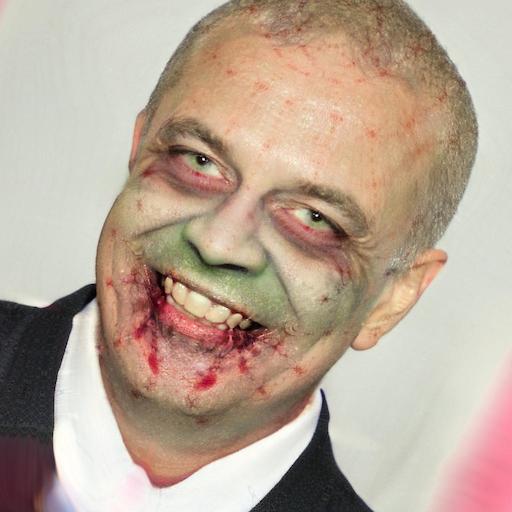} &
        \includegraphics[width=0.26\columnwidth]{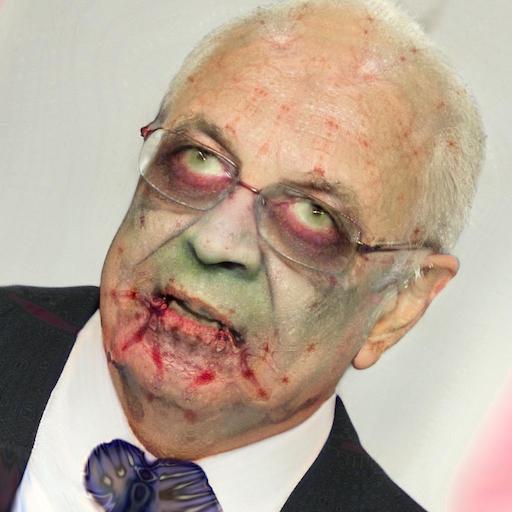} &
        \includegraphics[width=0.26\columnwidth]{images/editing/stylegan_nada/sup/im_seed=248029_zombie_smile_5.jpg} & 
        \includegraphics[width=0.26\columnwidth]{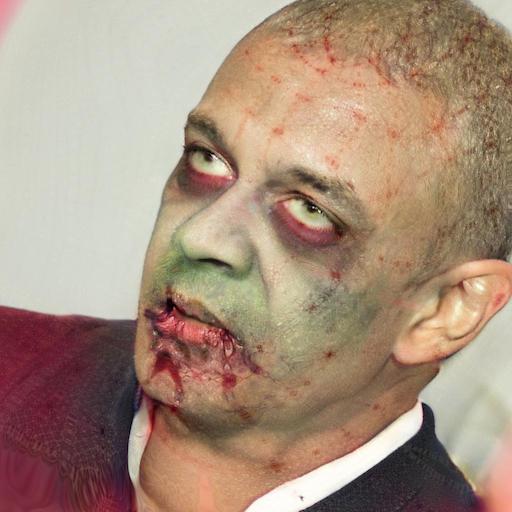}
        
        \\
        \raisebox{0.05in}{\rotatebox{90}{}} &
        \includegraphics[width=0.26\columnwidth]{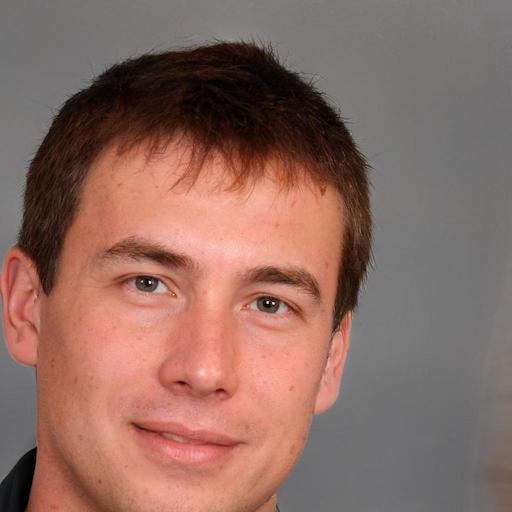} &
        \includegraphics[width=0.26\columnwidth]{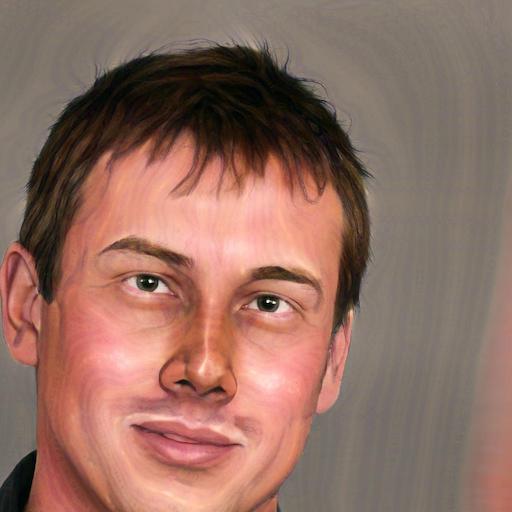} &
        \includegraphics[width=0.26\columnwidth]{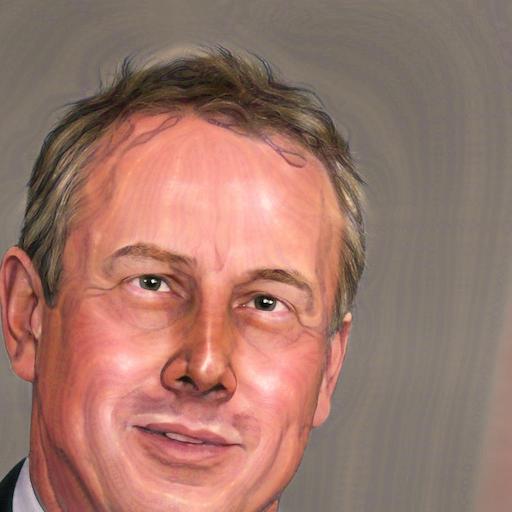} &
        \includegraphics[width=0.26\columnwidth]{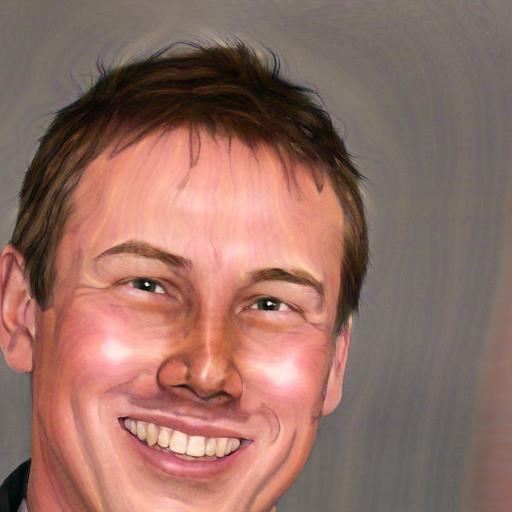} & 
        \includegraphics[width=0.26\columnwidth]{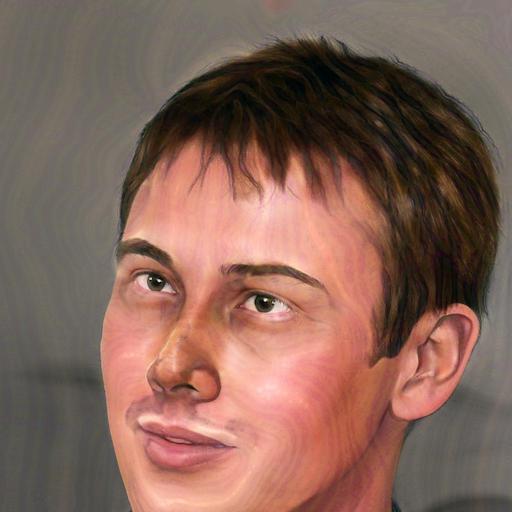}
    
        \\

		 & Source & Style & Age & Smile & Pose
	\end{tabular}
	}
    \vspace{-0.1cm}
	\caption{Editing using fine-tuned generators. The editing directions in the latent space of the parent (FFHQ) StyleGAN3 generator have a similar effect in the fine-tuned child generators trained with StyleGAN-NADA~\cite{gal2021stylegannada}, indicating the latent spaces remain semantically aligned.}
	\label{fig:stylegan_nada_editing_adaptation_supp}
\end{figure*}

%% file: figures/supplementary/inversion_comparison.tex
\begin{figure*}[tb]
	\centering
	\setlength{\tabcolsep}{1pt}	
	{\small
	\begin{tabular}{c c c c c c}

        \includegraphics[width=0.27\columnwidth]{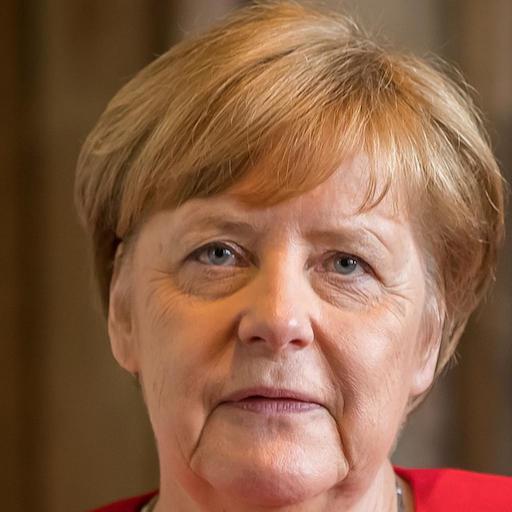} & 
        \includegraphics[width=0.27\columnwidth]{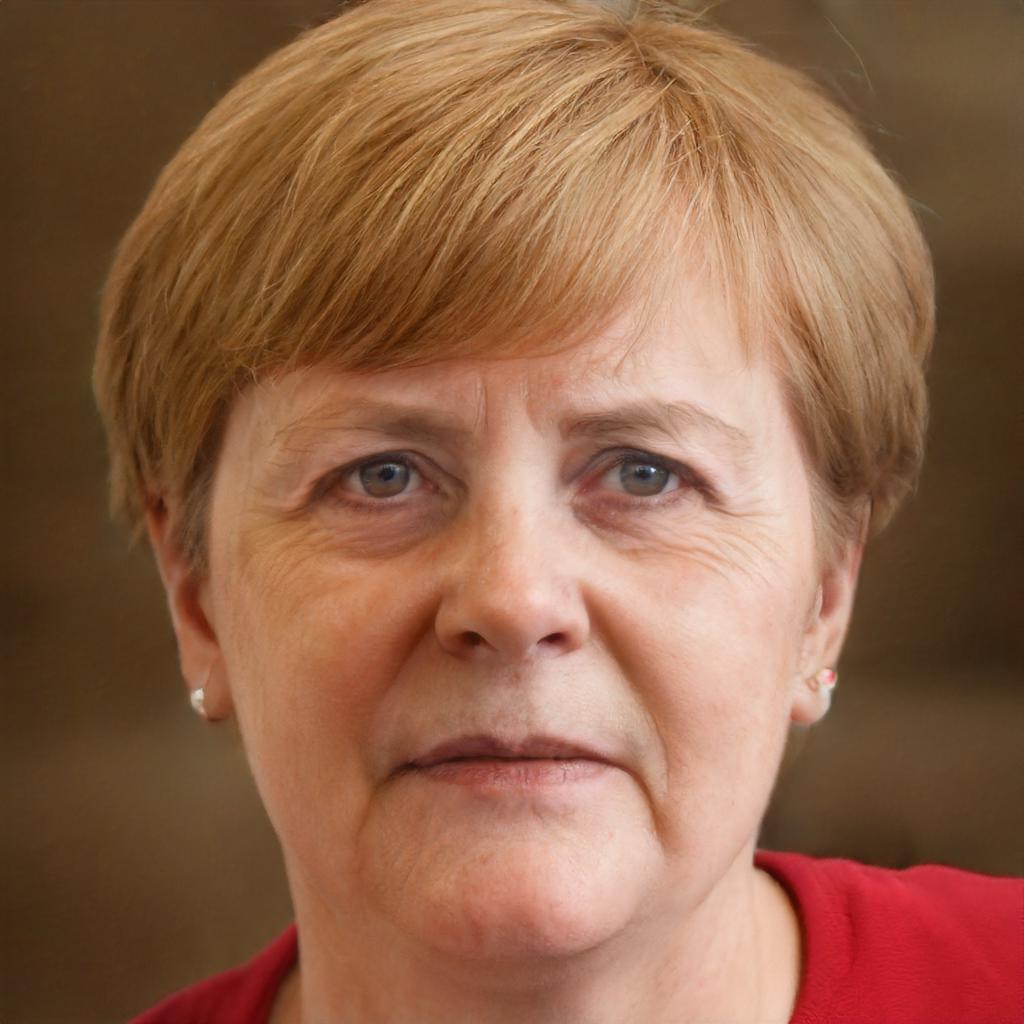} & 
        \includegraphics[width=0.27\columnwidth]{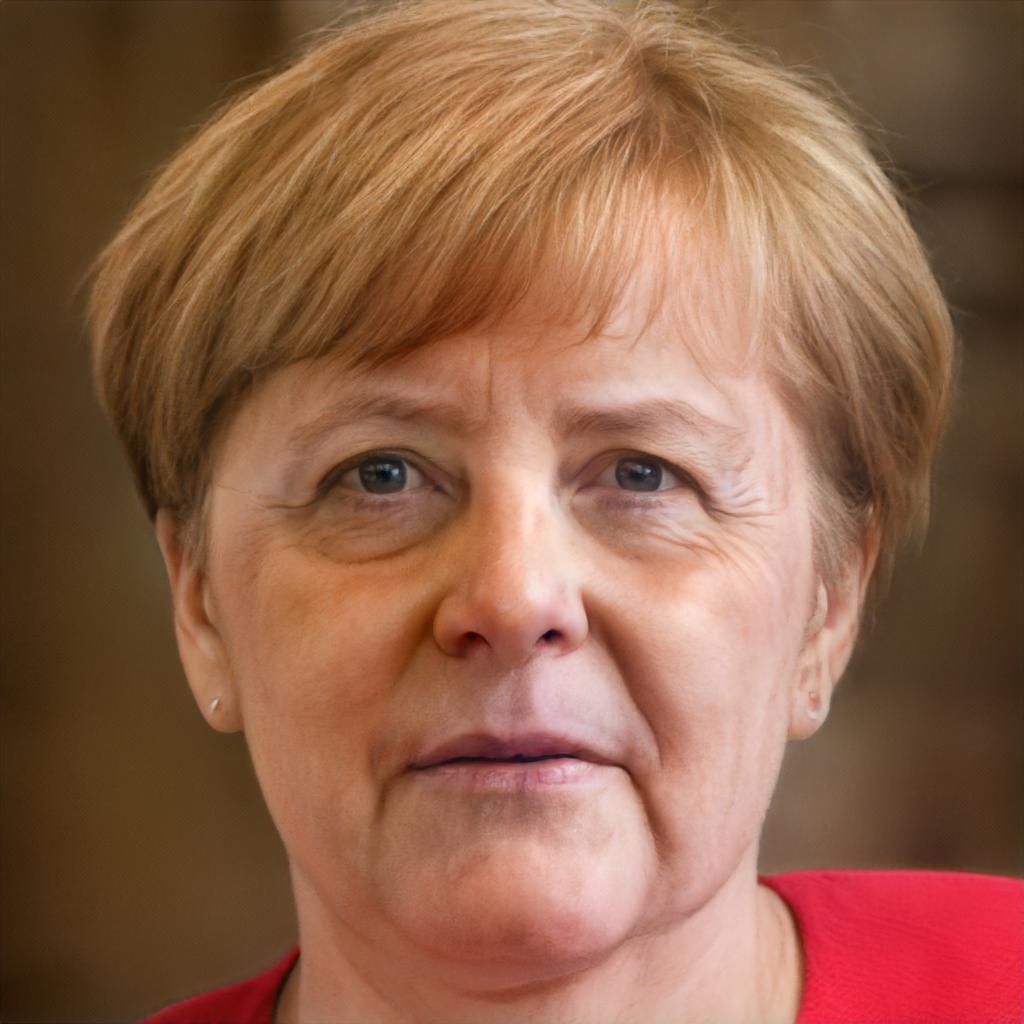} & 
        \includegraphics[width=0.27\columnwidth]{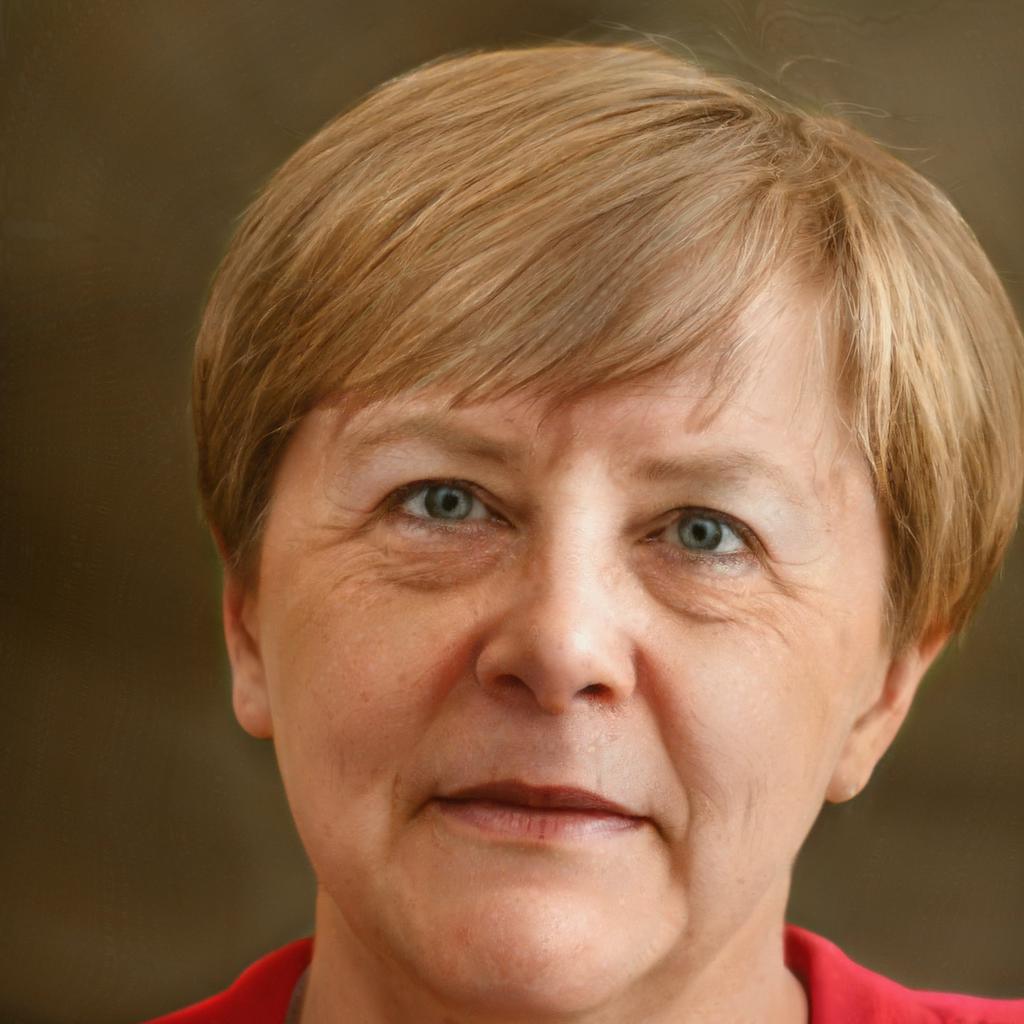} & 
        \includegraphics[width=0.27\columnwidth]{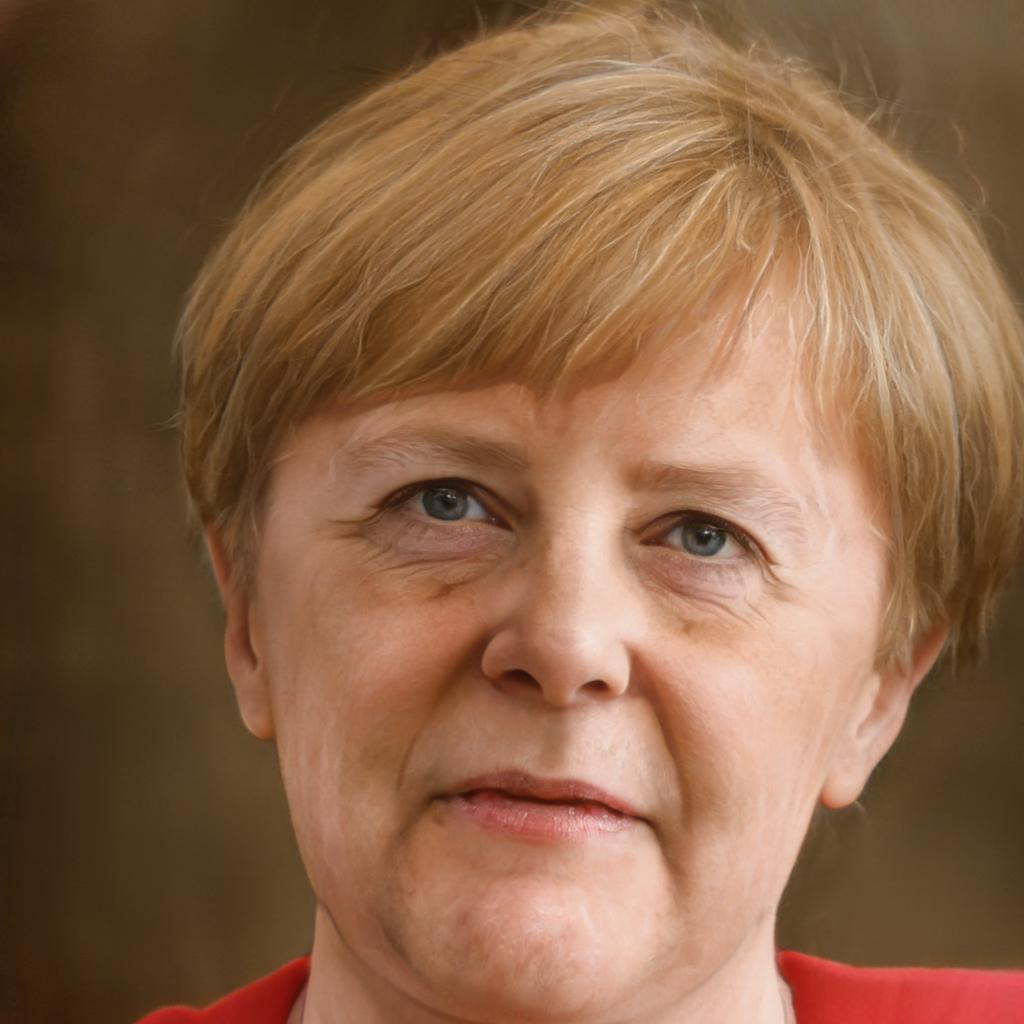} \\

        \includegraphics[width=0.27\columnwidth]{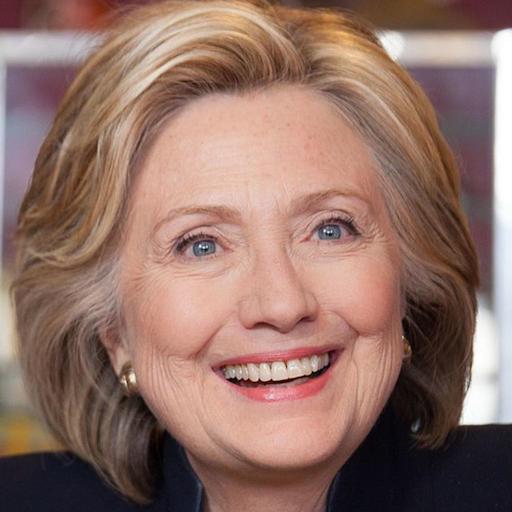} & 
        \includegraphics[width=0.27\columnwidth]{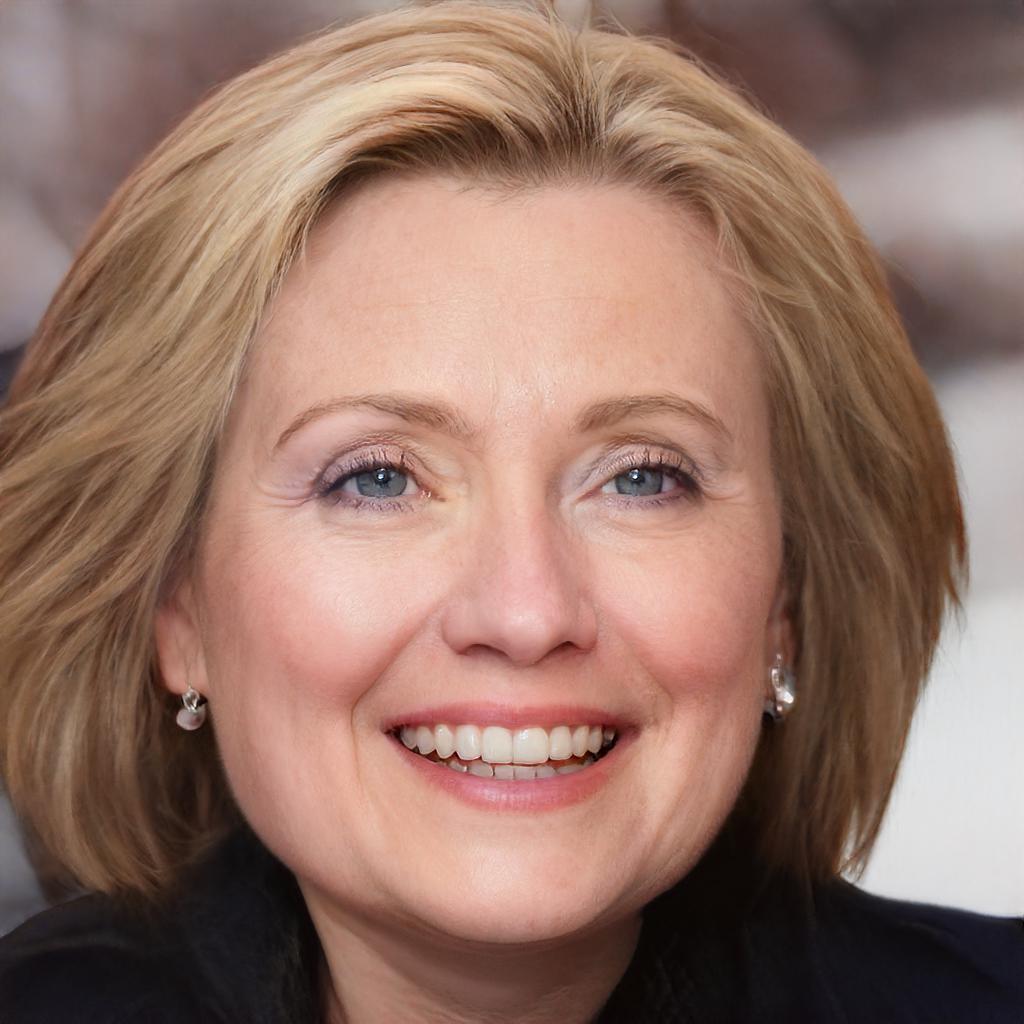} & 
        \includegraphics[width=0.27\columnwidth]{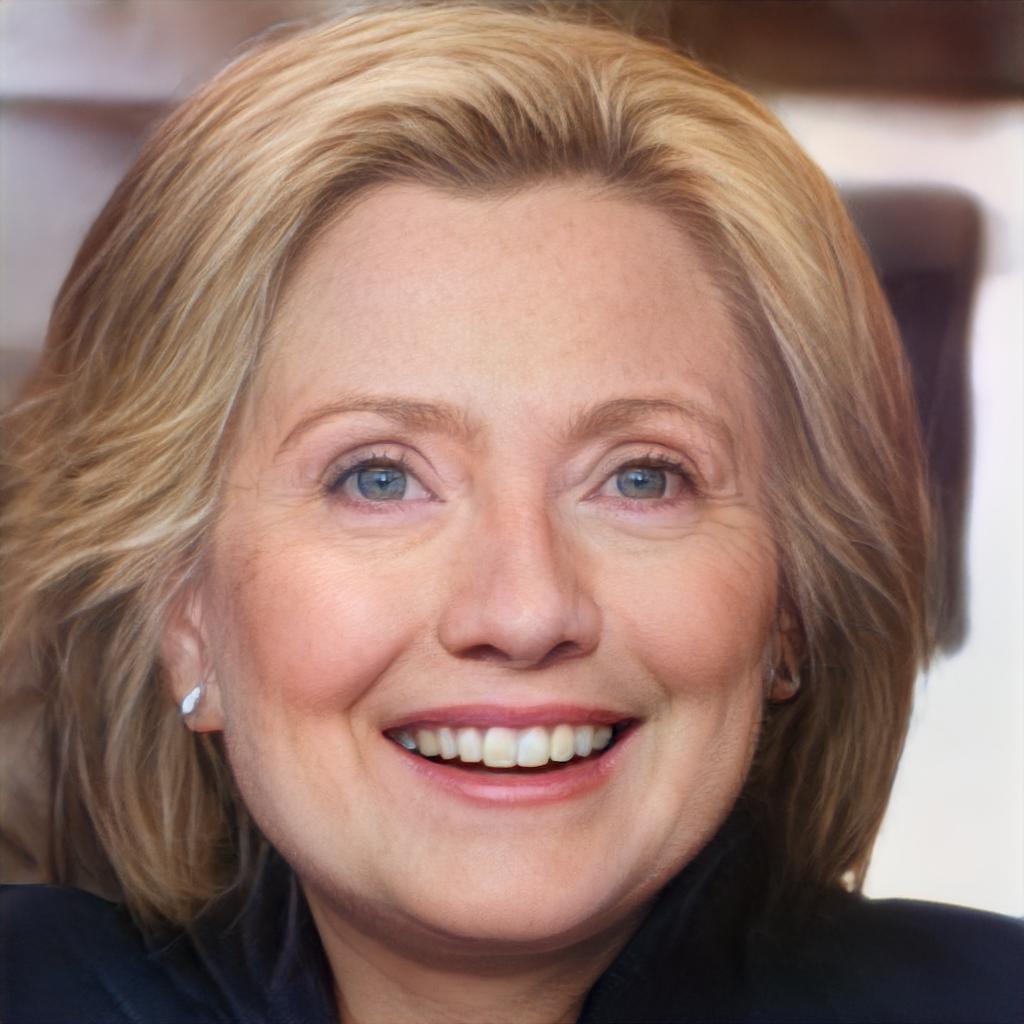} & 
        \includegraphics[width=0.27\columnwidth]{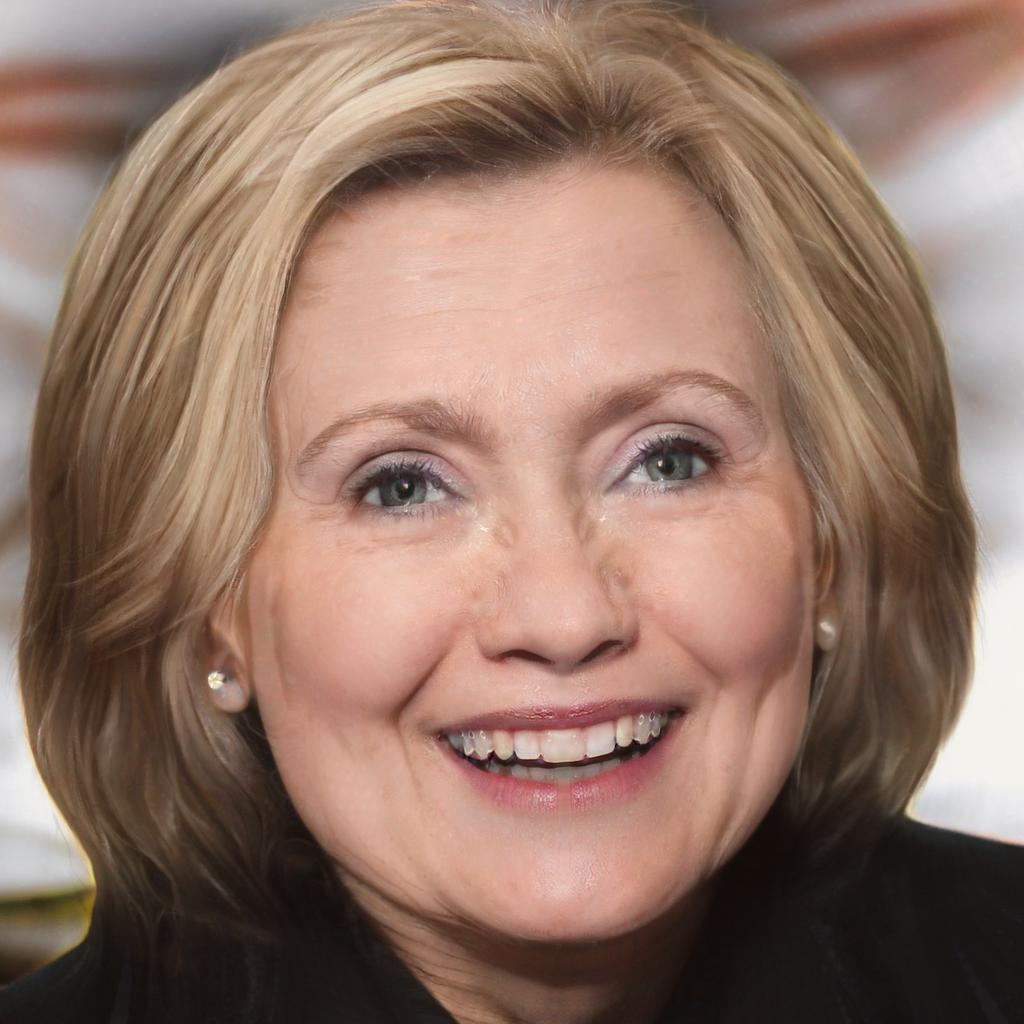} & 
        \includegraphics[width=0.27\columnwidth]{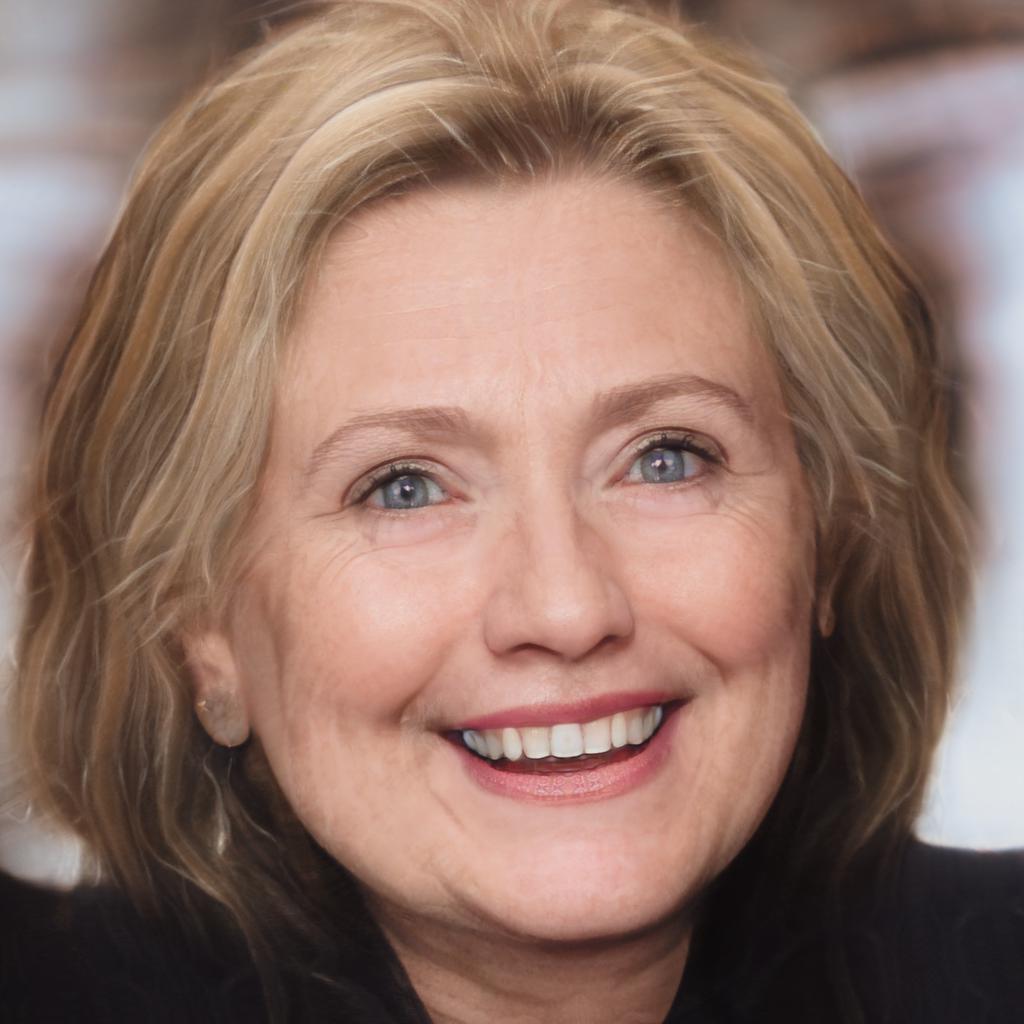} \\
        
        \includegraphics[width=0.27\columnwidth]{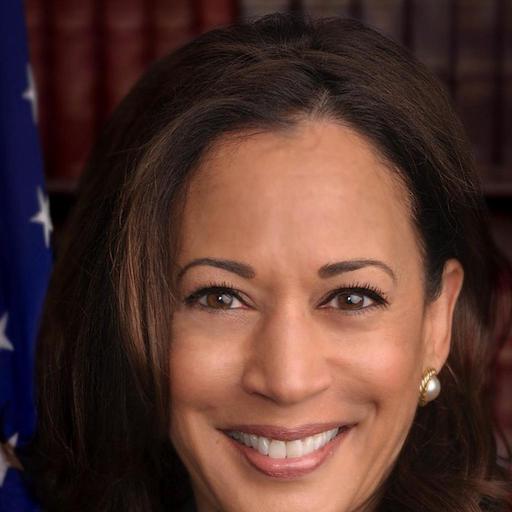} & 
        \includegraphics[width=0.27\columnwidth]{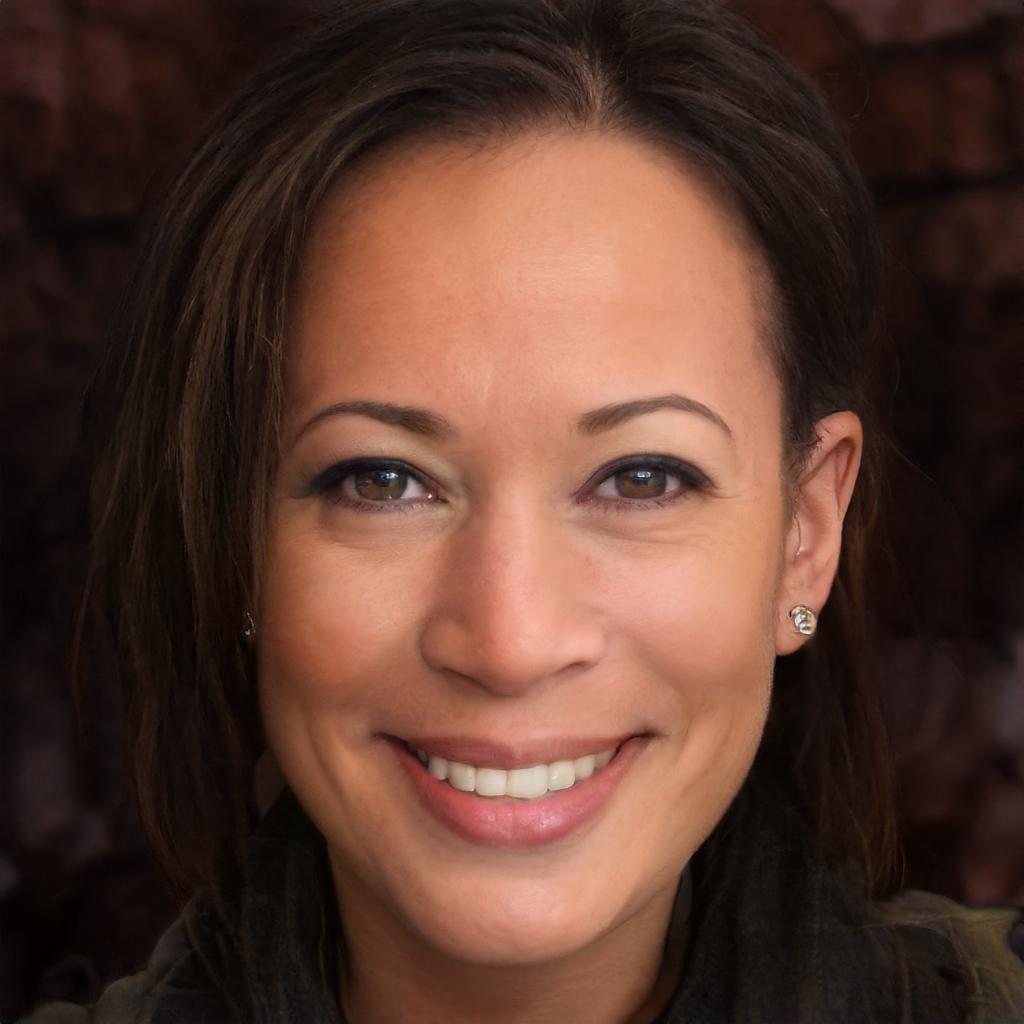} & 
        \includegraphics[width=0.27\columnwidth]{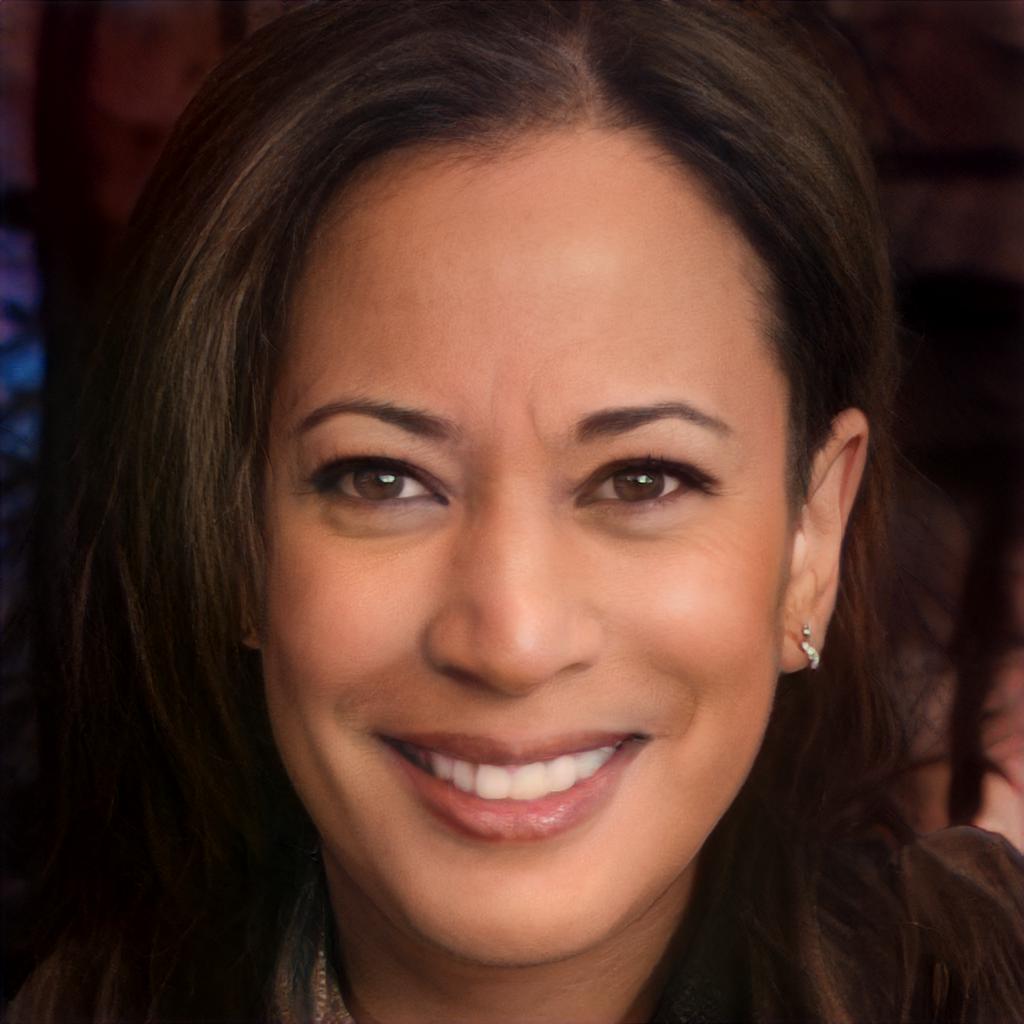} & 
        \includegraphics[width=0.27\columnwidth]{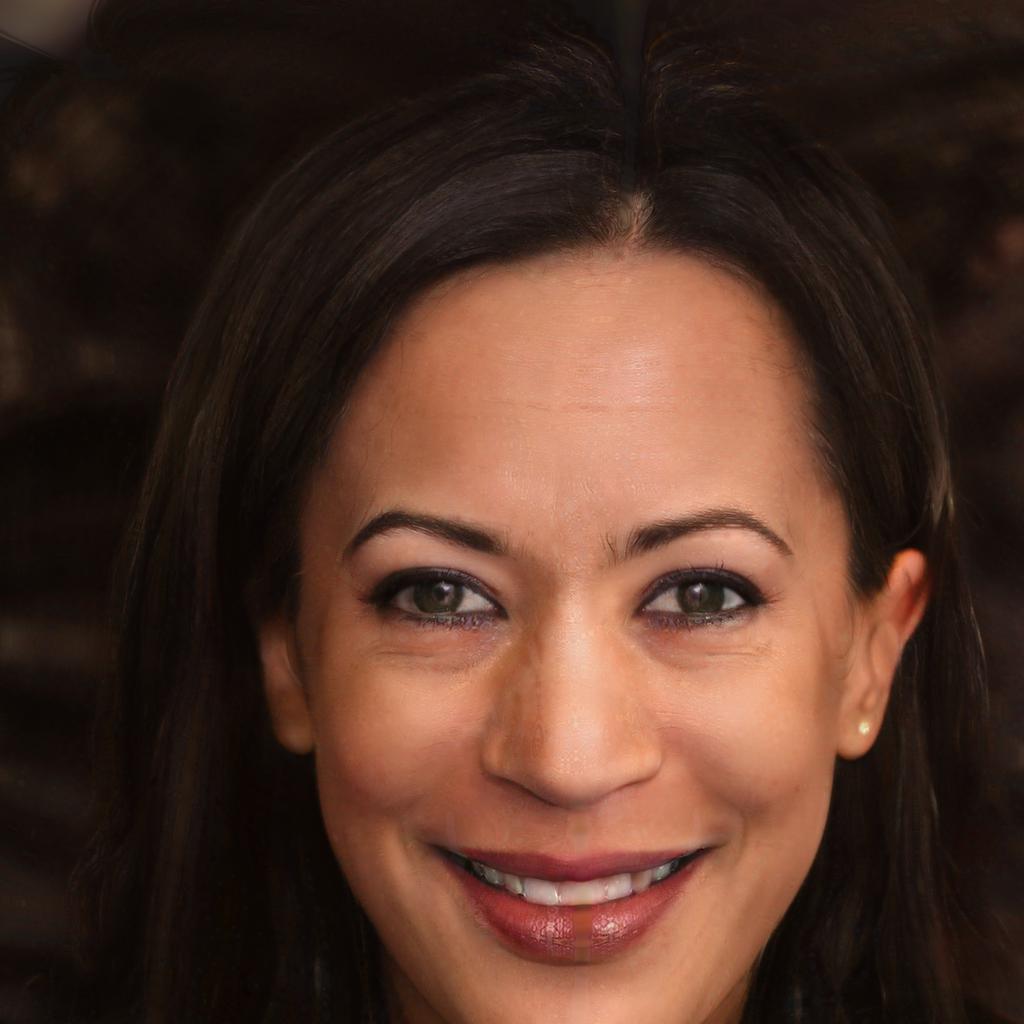} & 
        \includegraphics[width=0.27\columnwidth]{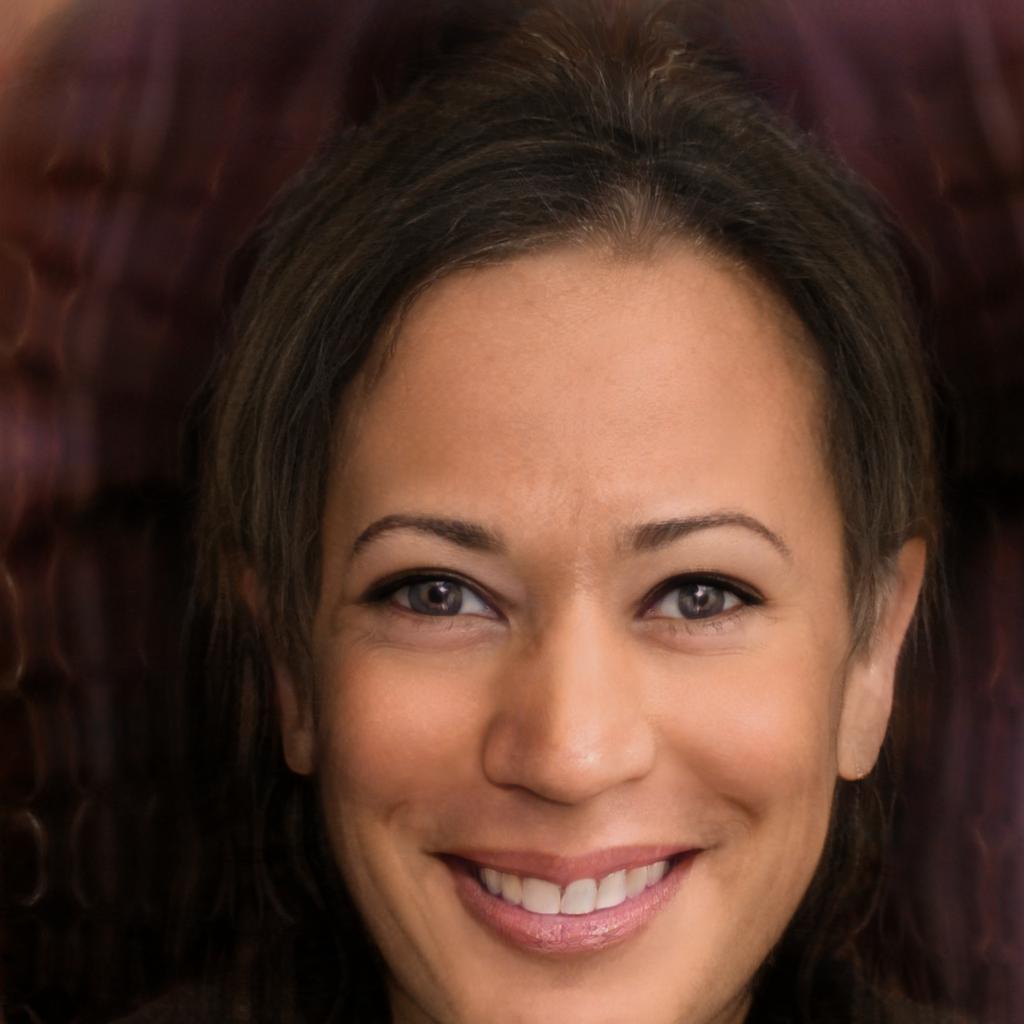} \\
        
        \includegraphics[width=0.27\columnwidth]{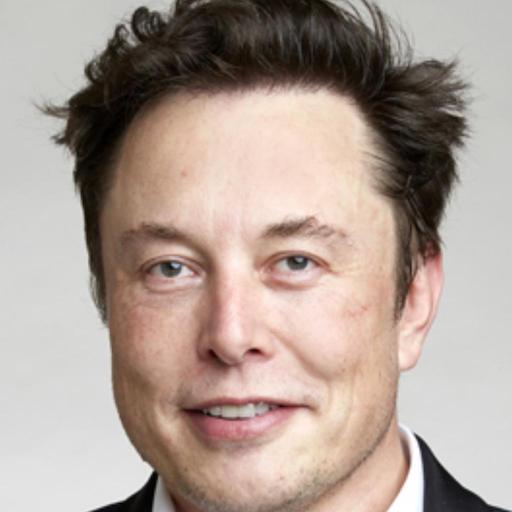} & 
        \includegraphics[width=0.27\columnwidth]{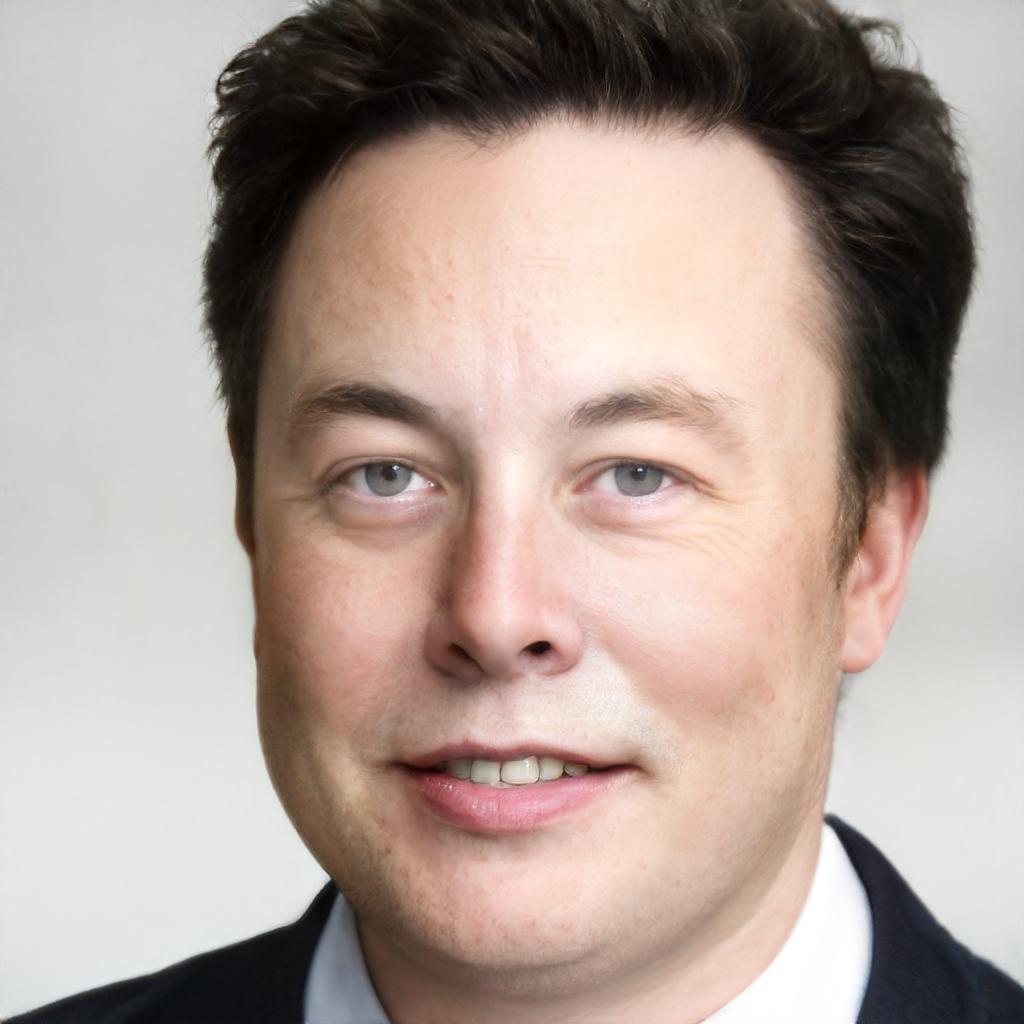} & 
        \includegraphics[width=0.27\columnwidth]{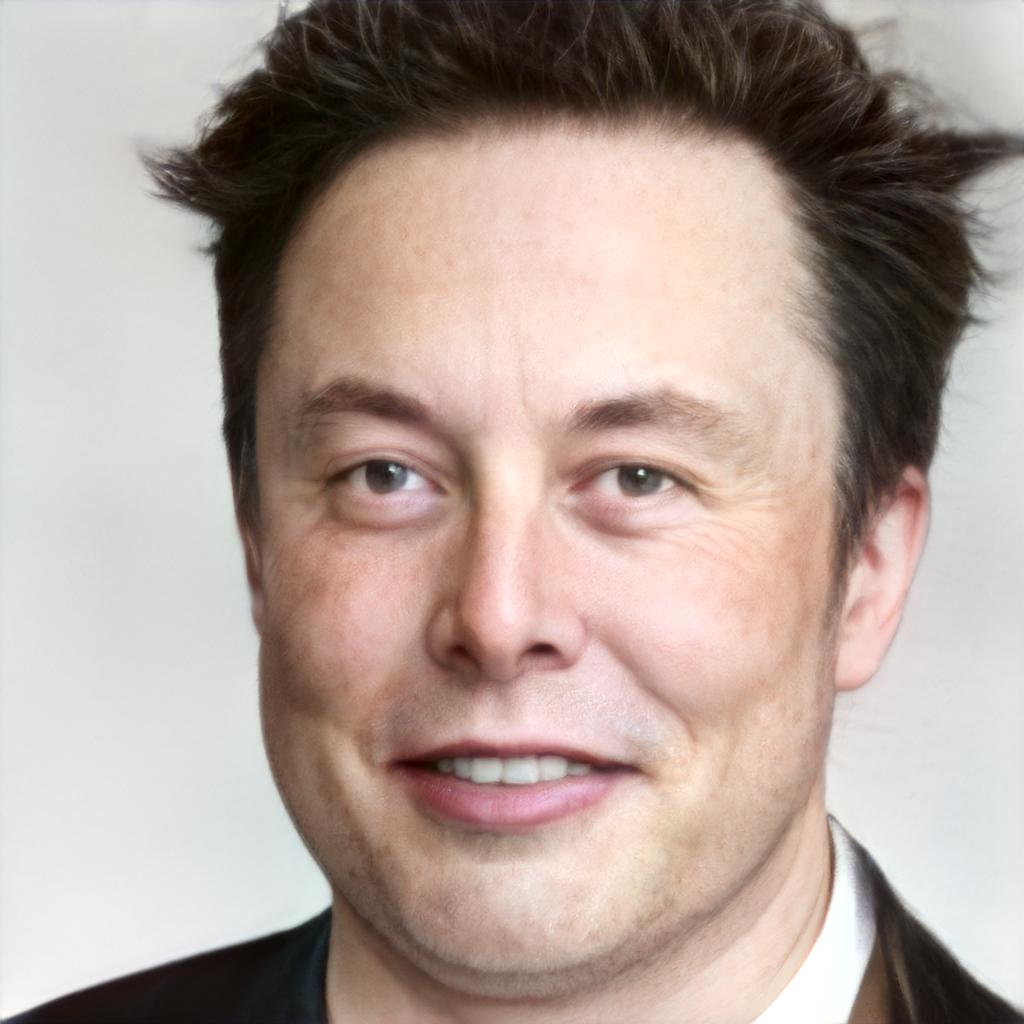} & 
        \includegraphics[width=0.27\columnwidth]{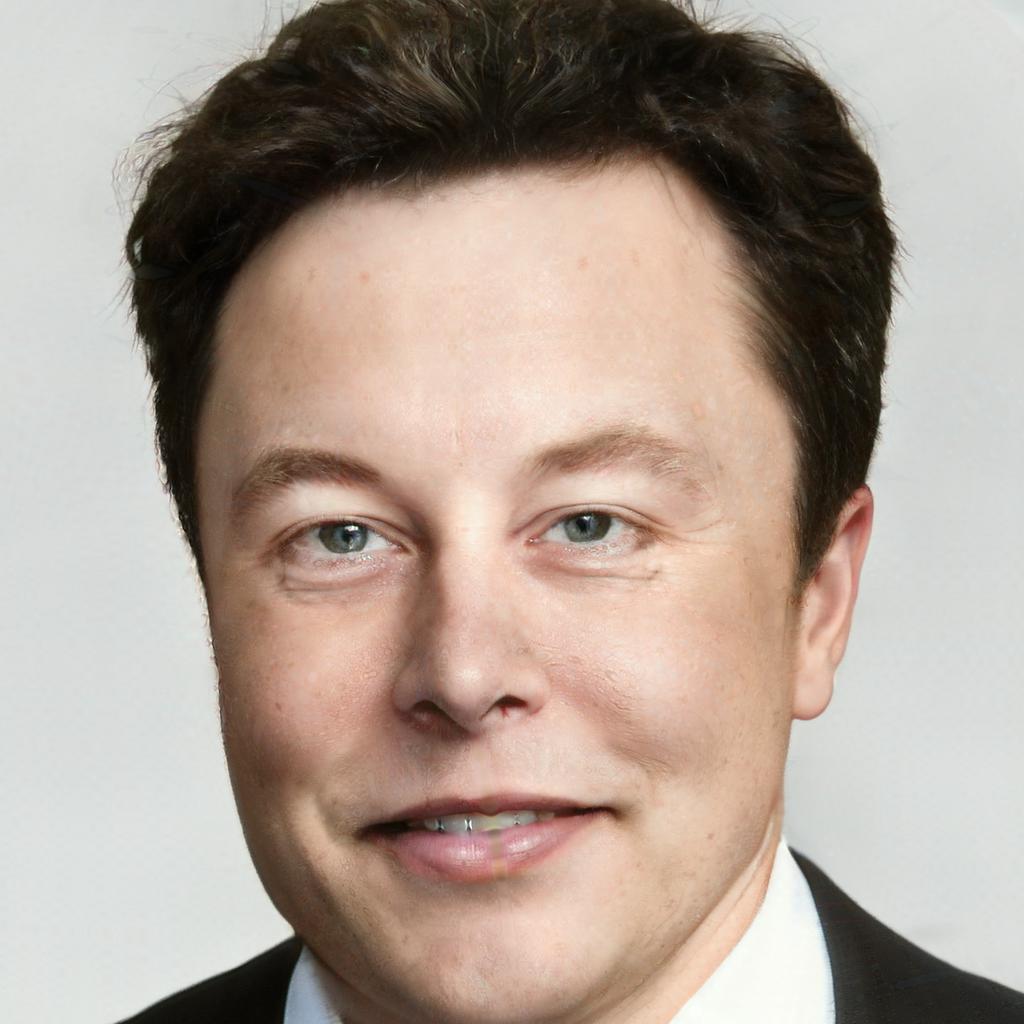} & 
        \includegraphics[width=0.27\columnwidth]{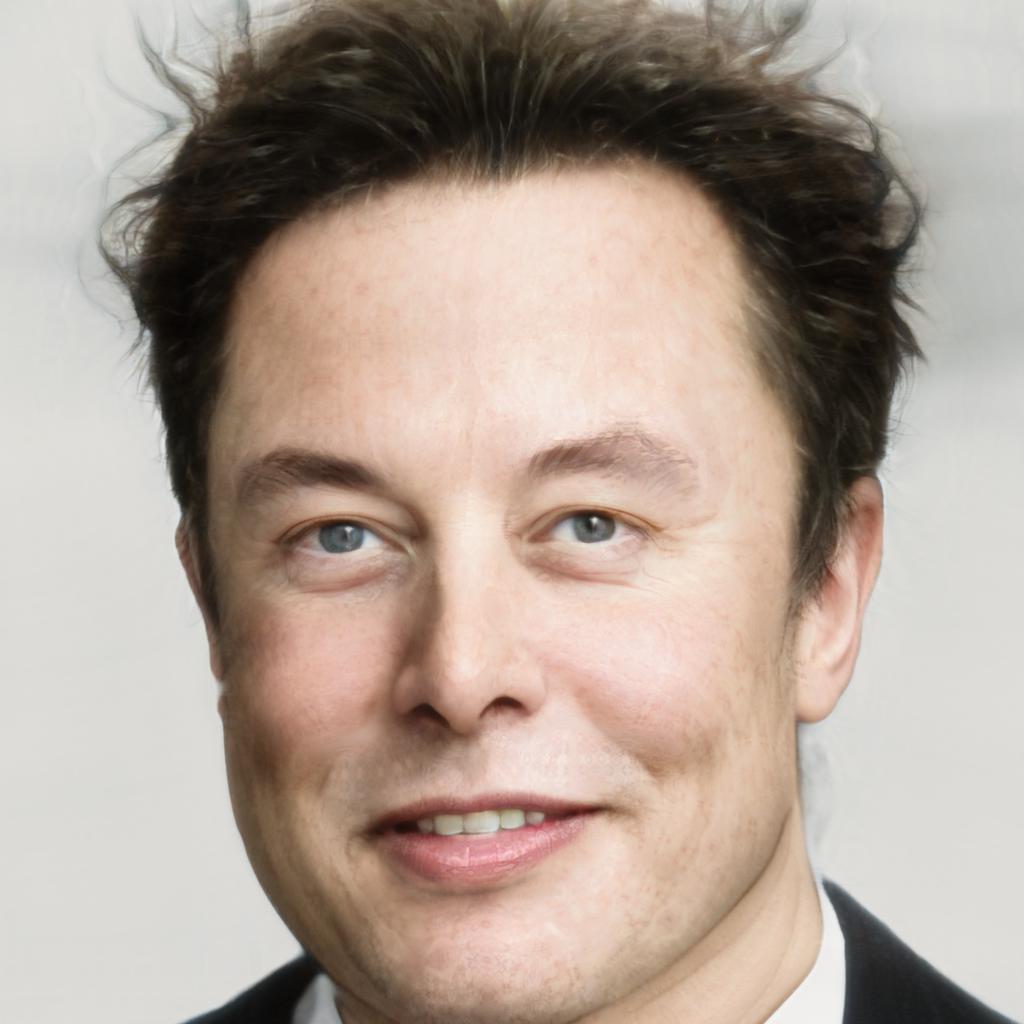} \\

        \includegraphics[width=0.27\columnwidth]{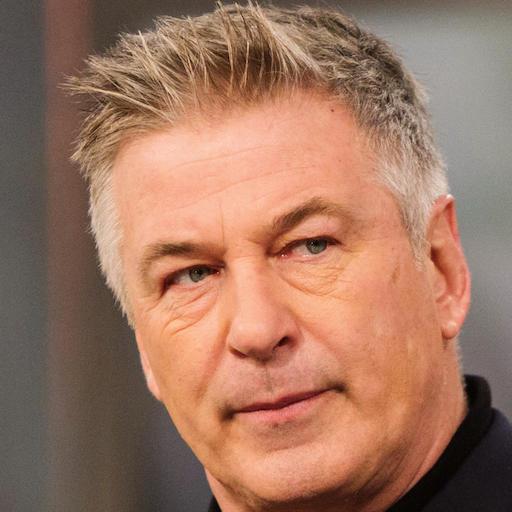} & 
        \includegraphics[width=0.27\columnwidth]{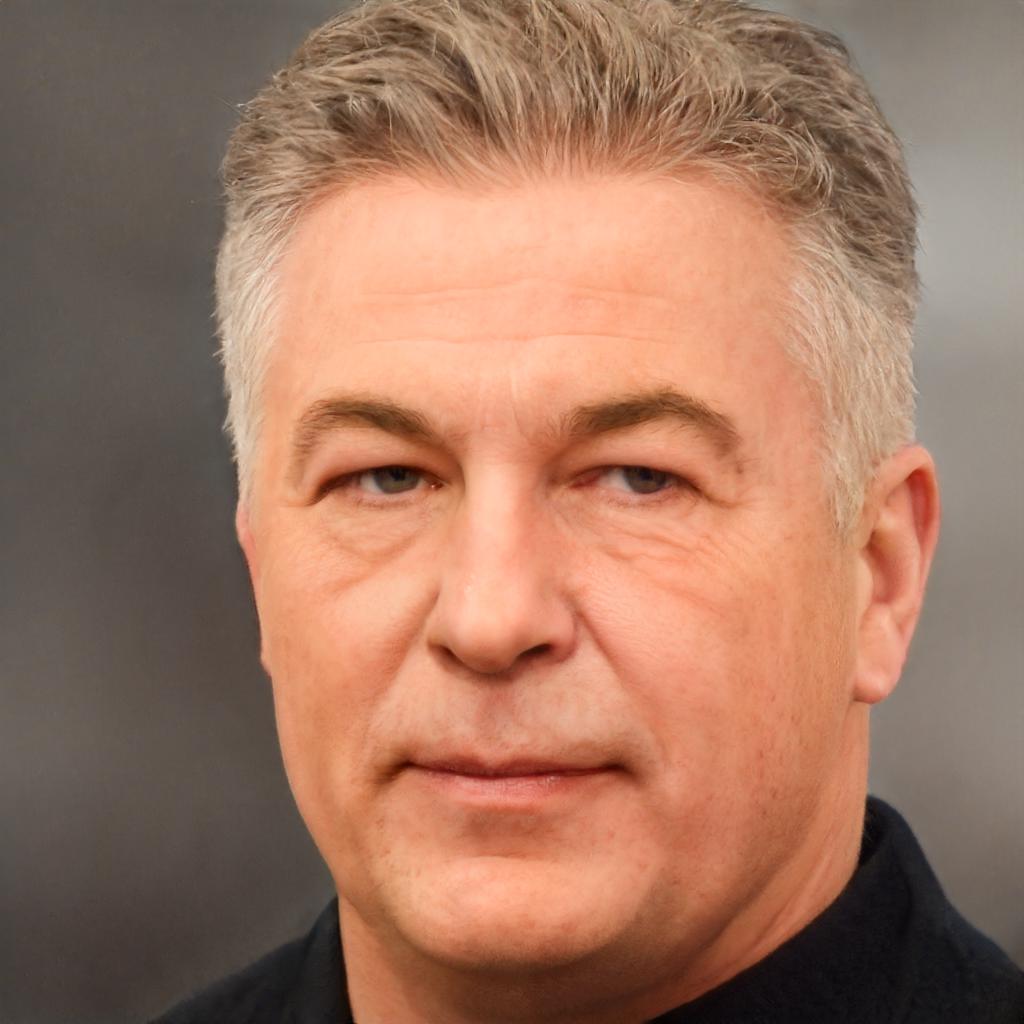} & 
        \includegraphics[width=0.27\columnwidth]{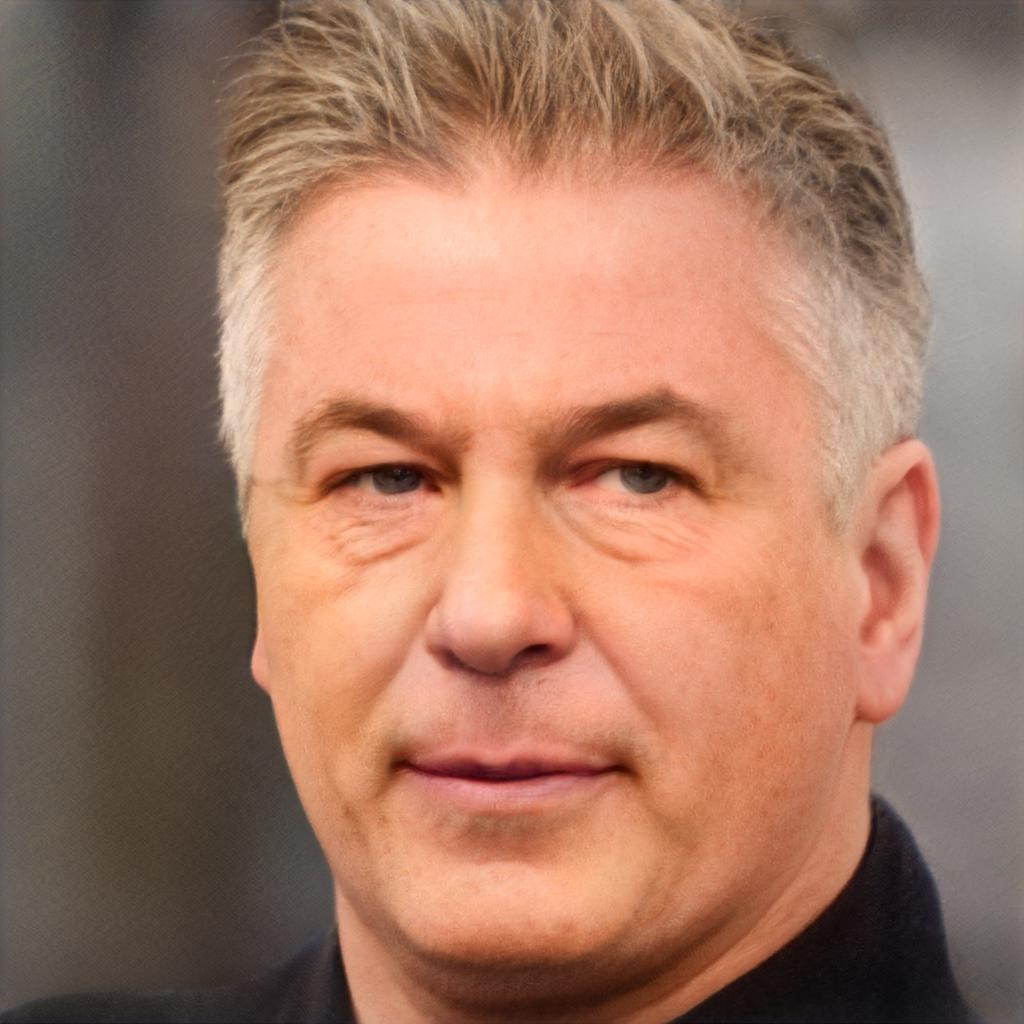} & 
        \includegraphics[width=0.27\columnwidth]{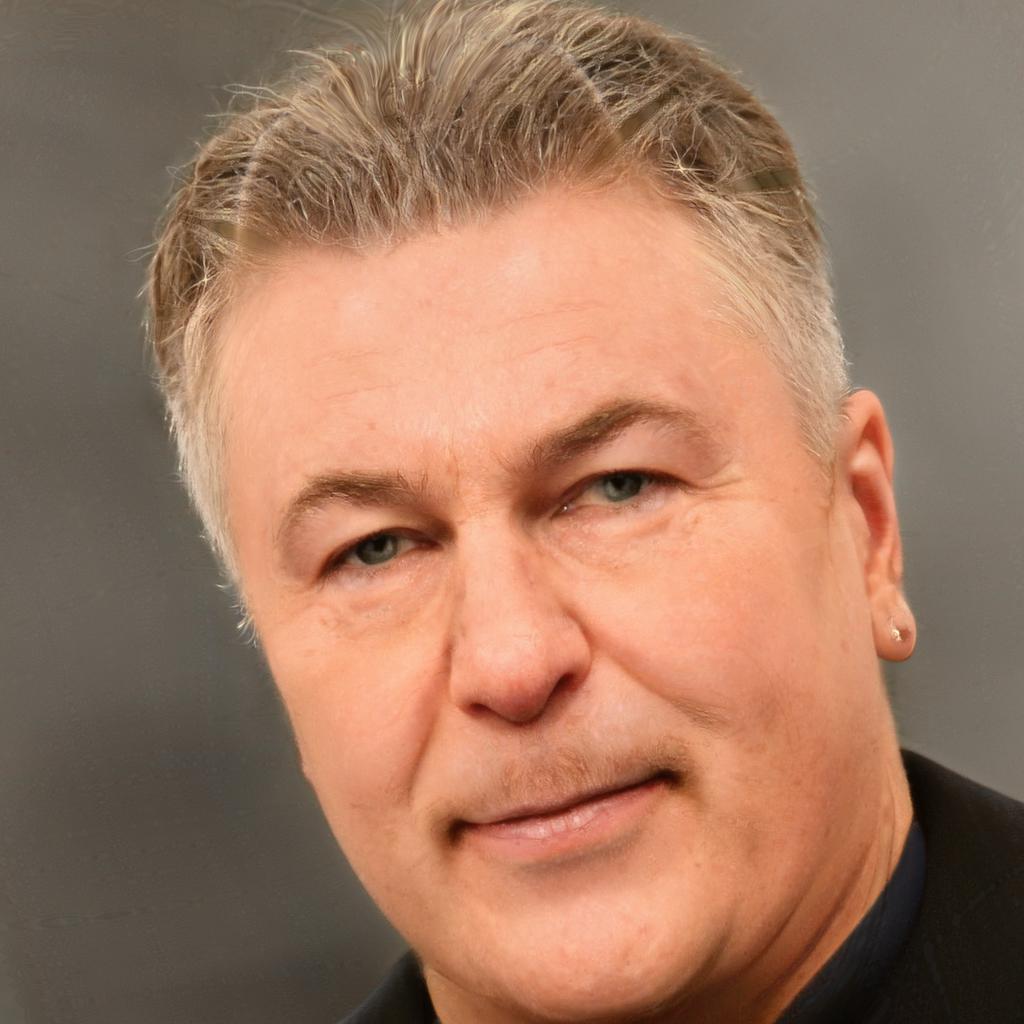} & 
        \includegraphics[width=0.27\columnwidth]{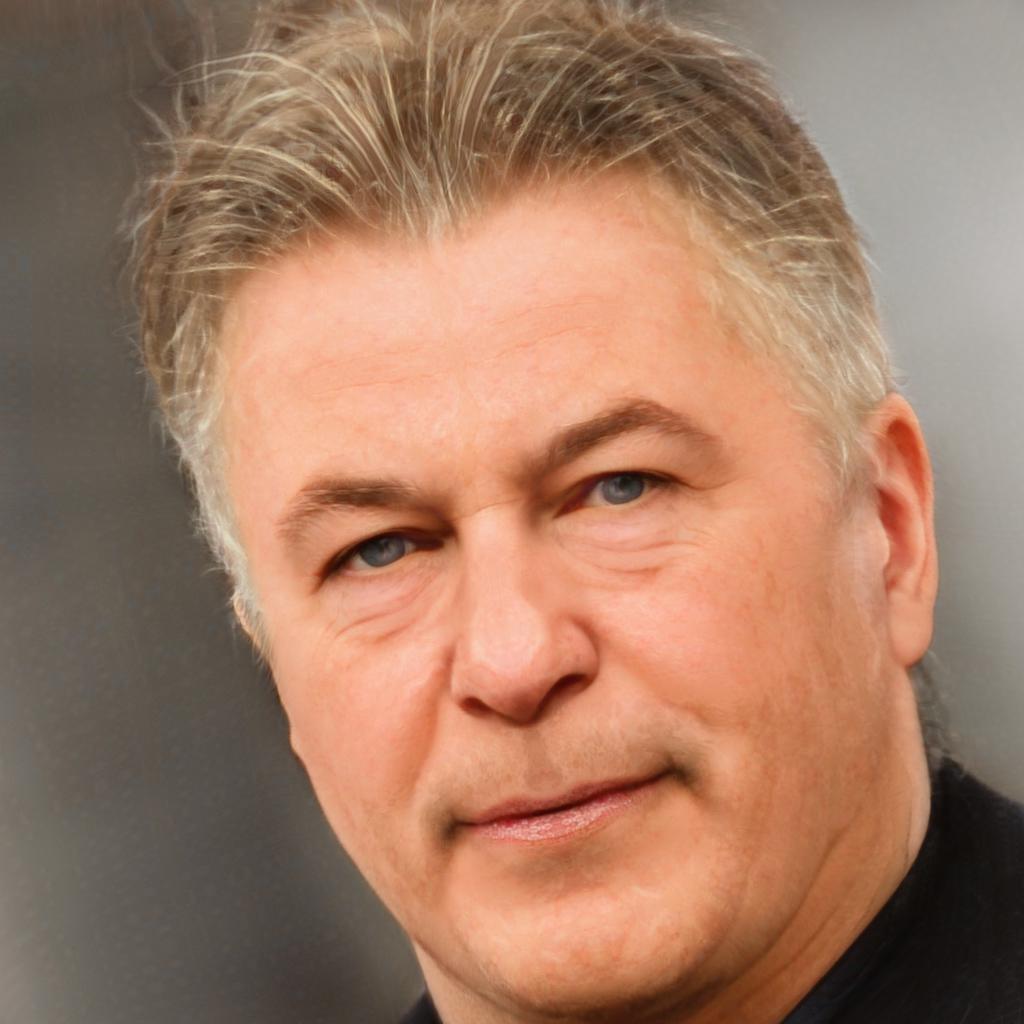} \\

        \includegraphics[width=0.27\columnwidth]{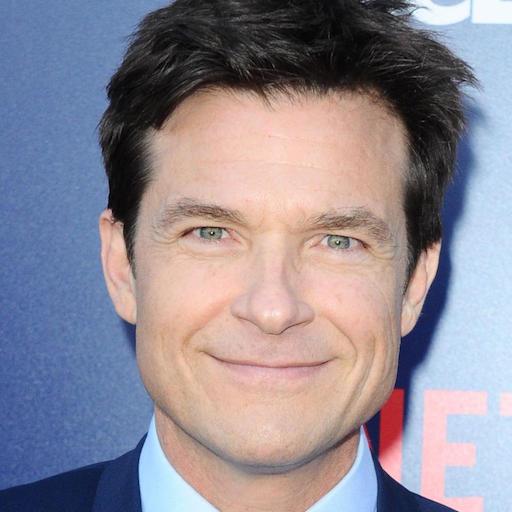} & 
        \includegraphics[width=0.27\columnwidth]{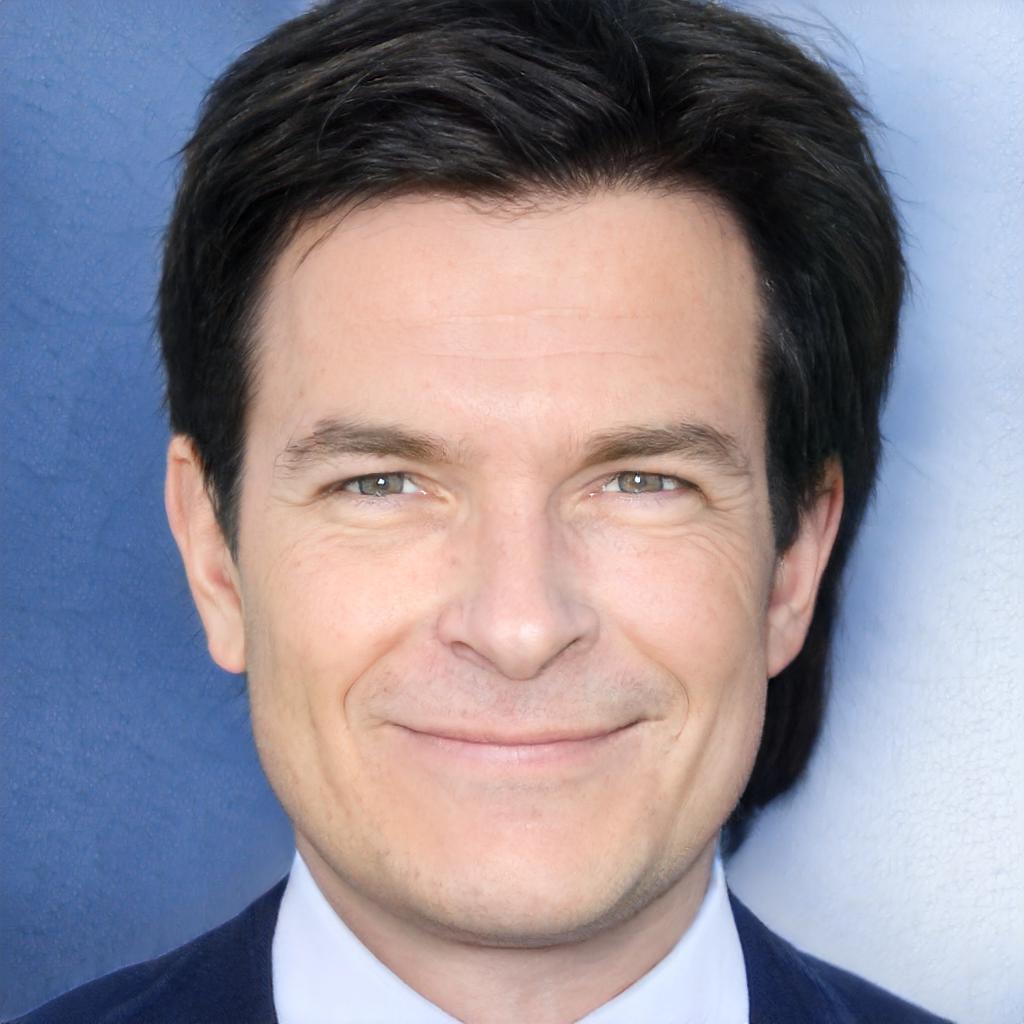} & 
        \includegraphics[width=0.27\columnwidth]{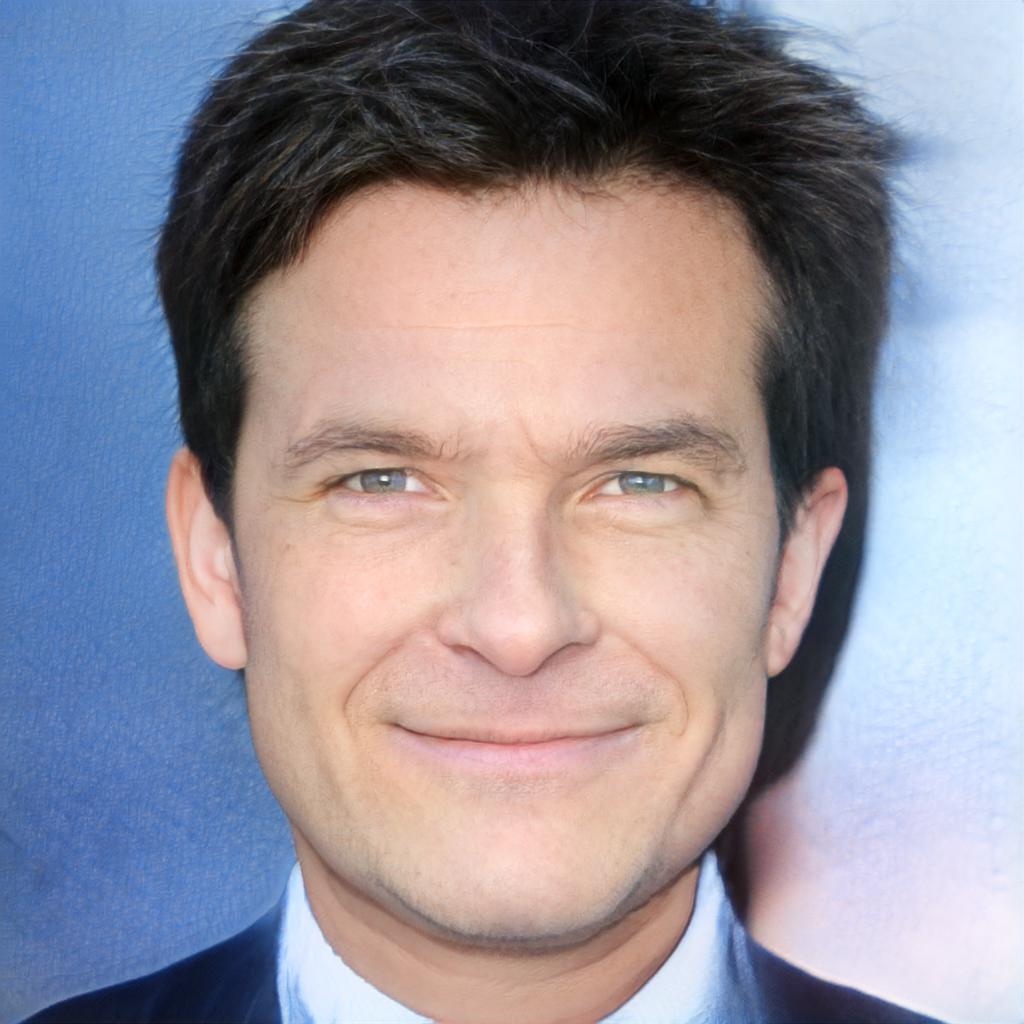} & 
        \includegraphics[width=0.27\columnwidth]{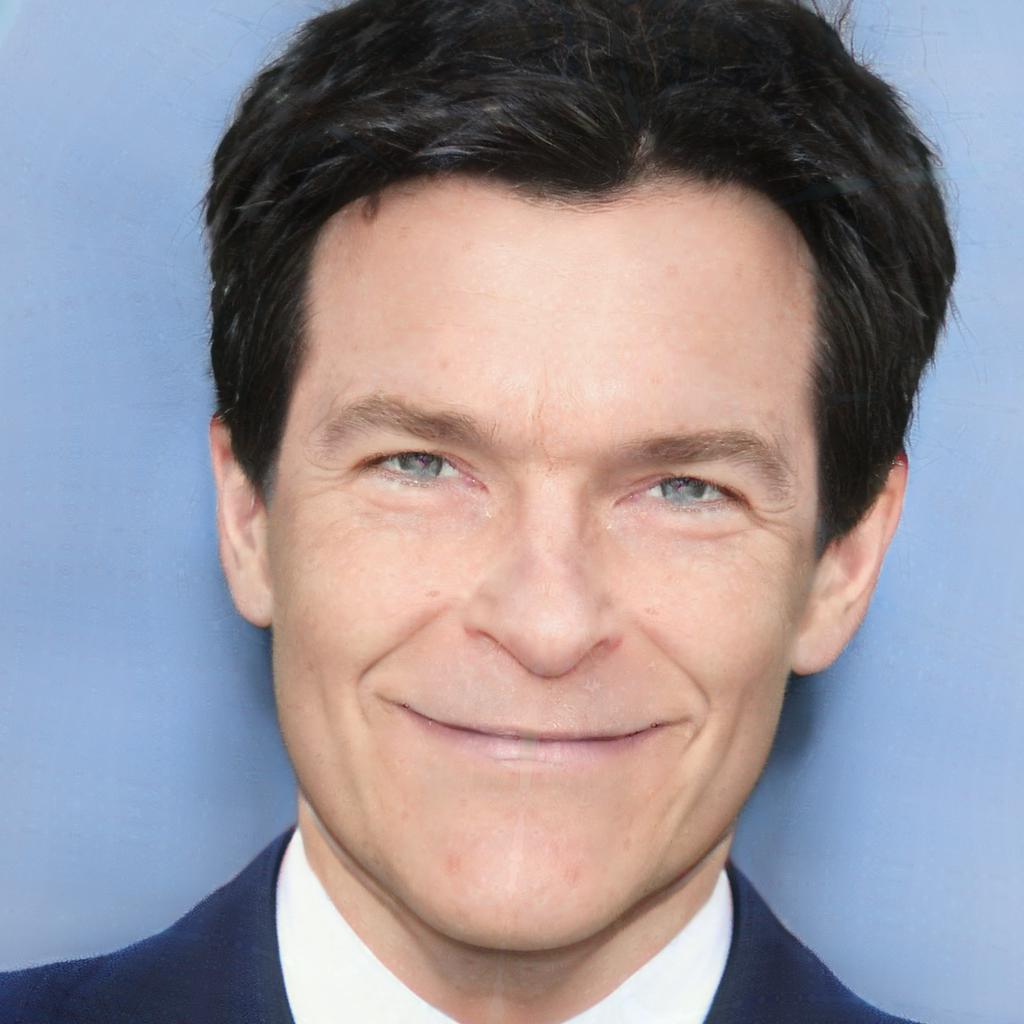} & 
        \includegraphics[width=0.27\columnwidth]{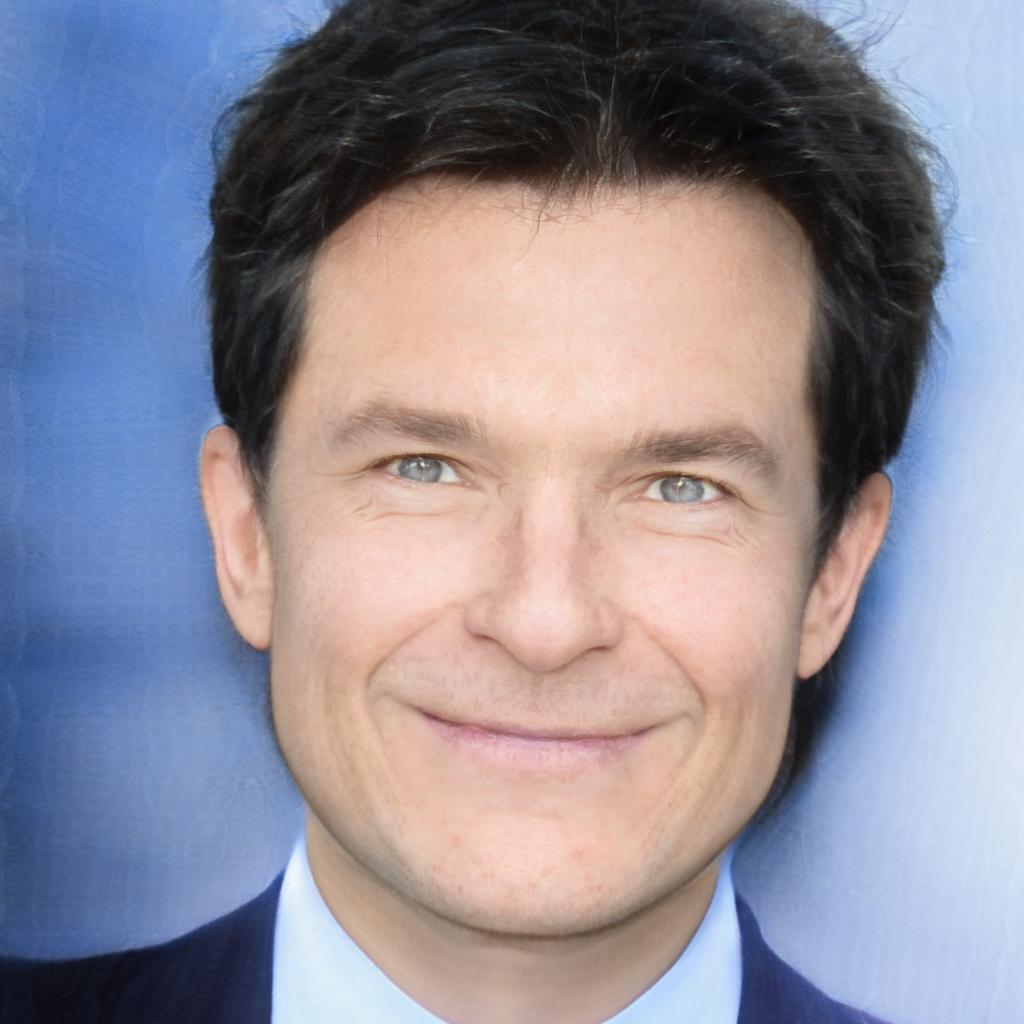} \\
        
        \includegraphics[width=0.27\columnwidth]{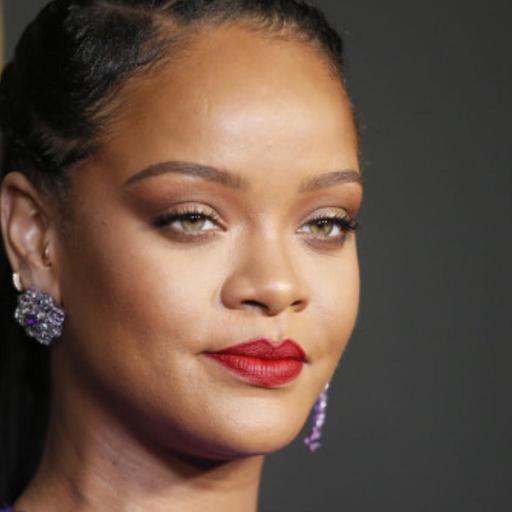} & 
        \includegraphics[width=0.27\columnwidth]{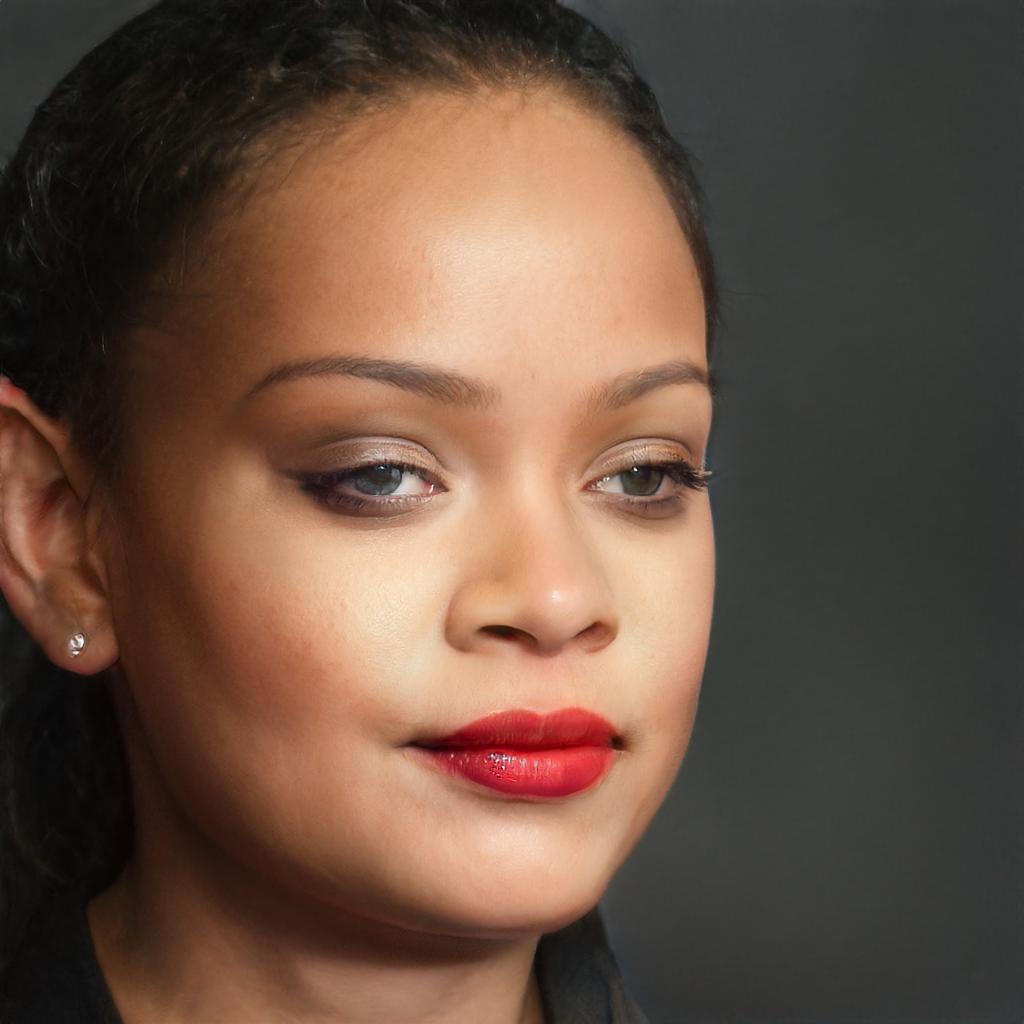} & 
        \includegraphics[width=0.27\columnwidth]{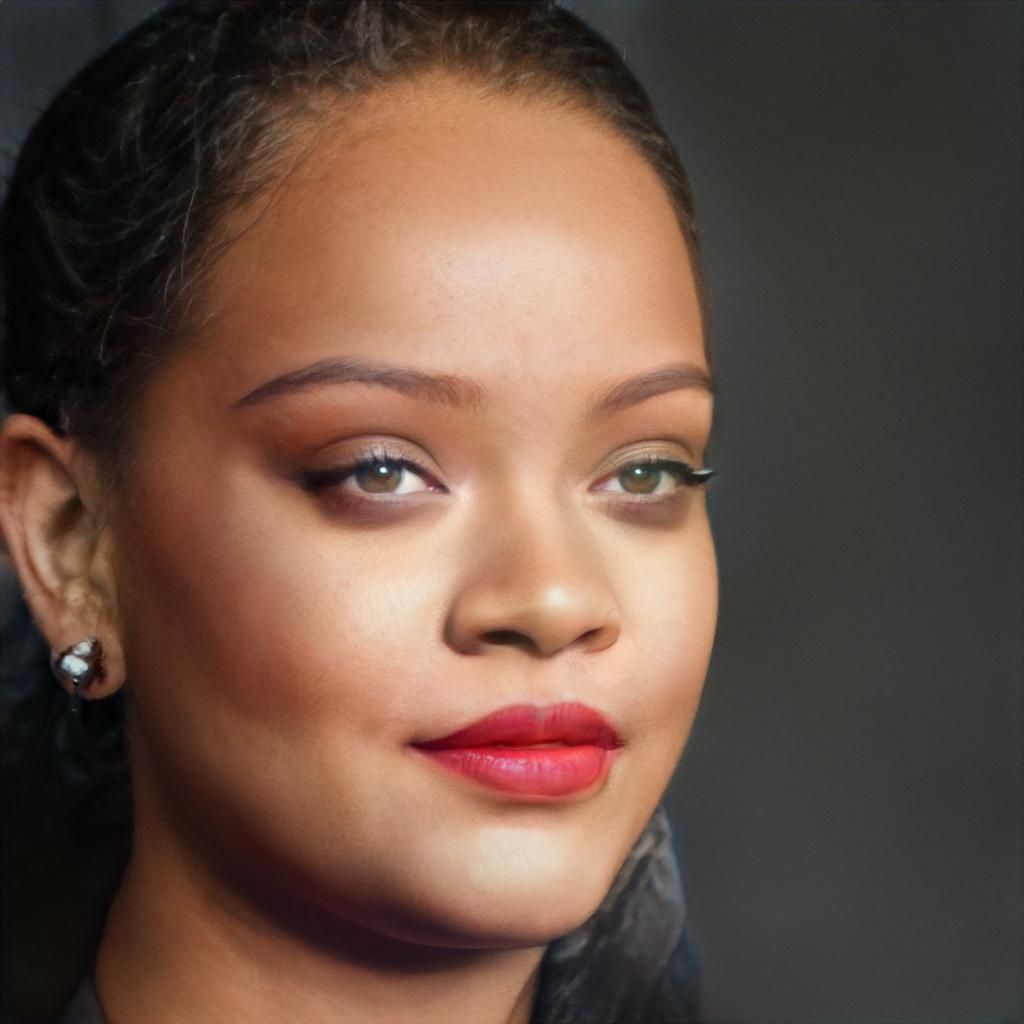} & 
        \includegraphics[width=0.27\columnwidth]{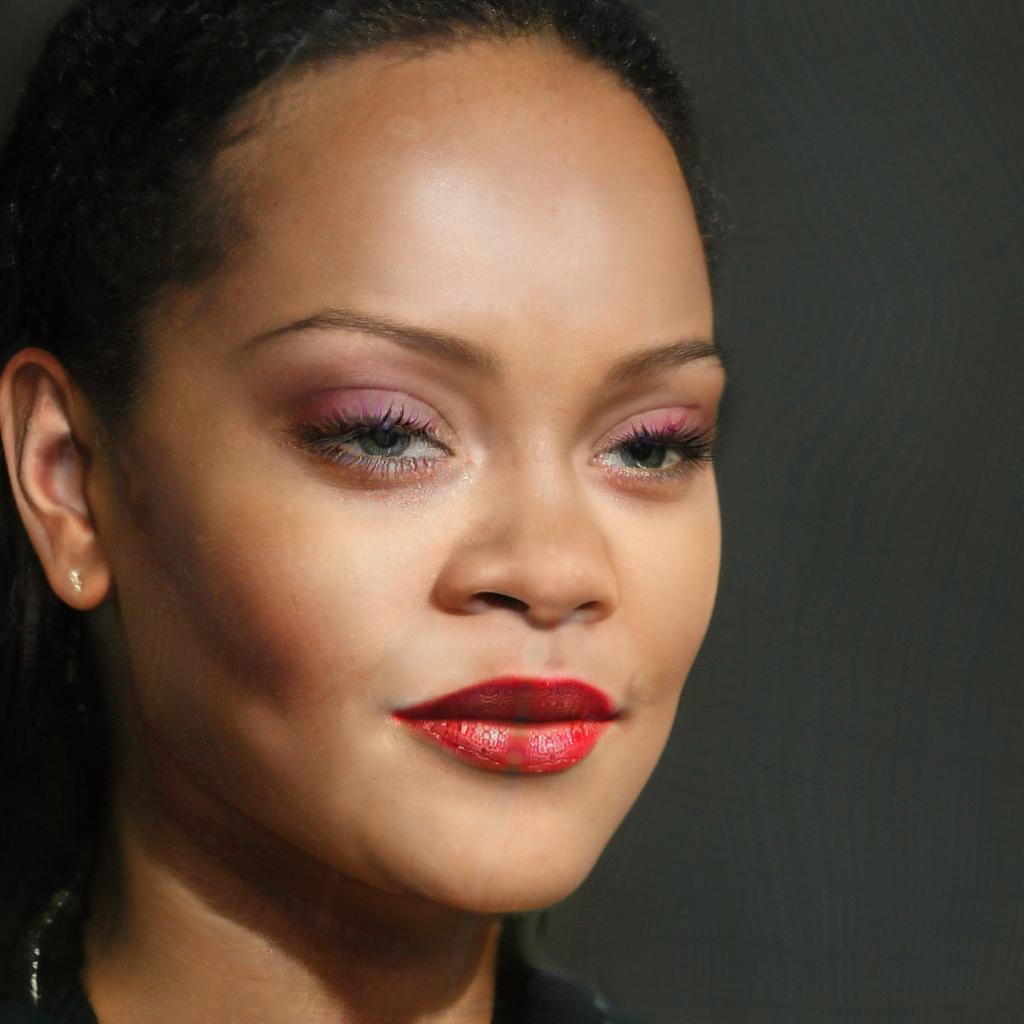} & 
        \includegraphics[width=0.27\columnwidth]{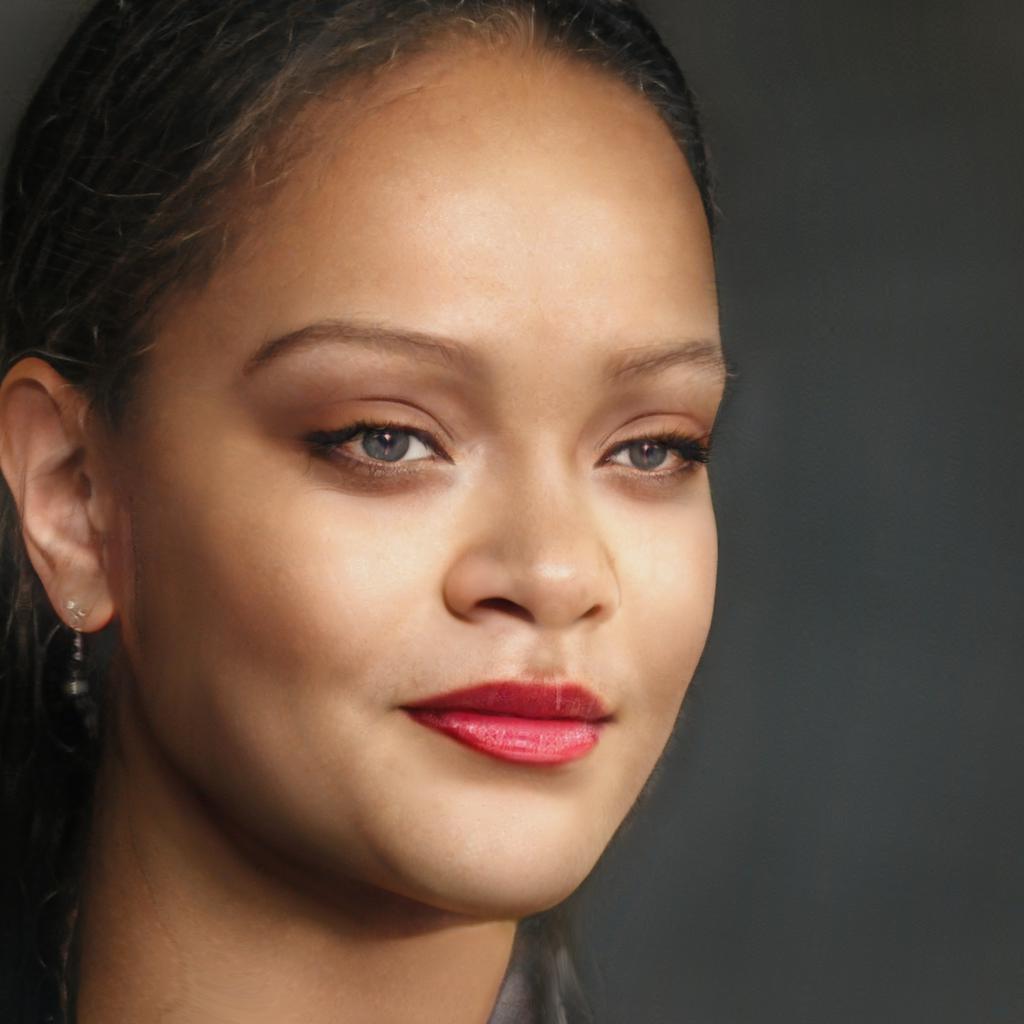} \\
        
        \includegraphics[width=0.27\columnwidth]{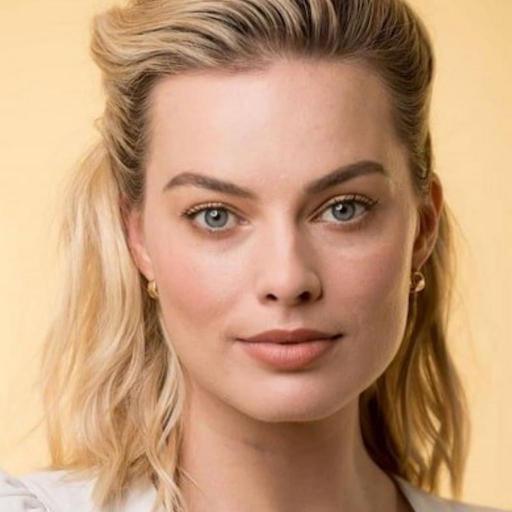} & 
        \includegraphics[width=0.27\columnwidth]{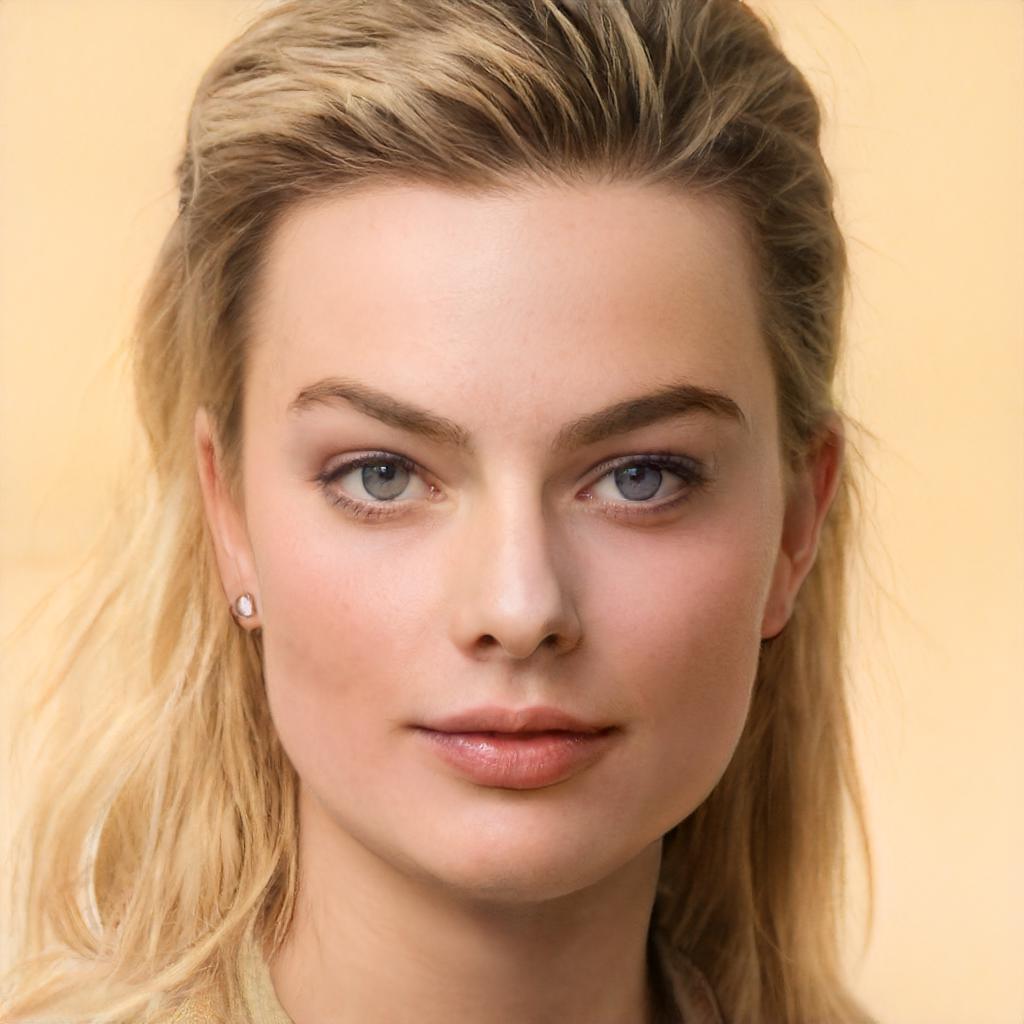} & 
        \includegraphics[width=0.27\columnwidth]{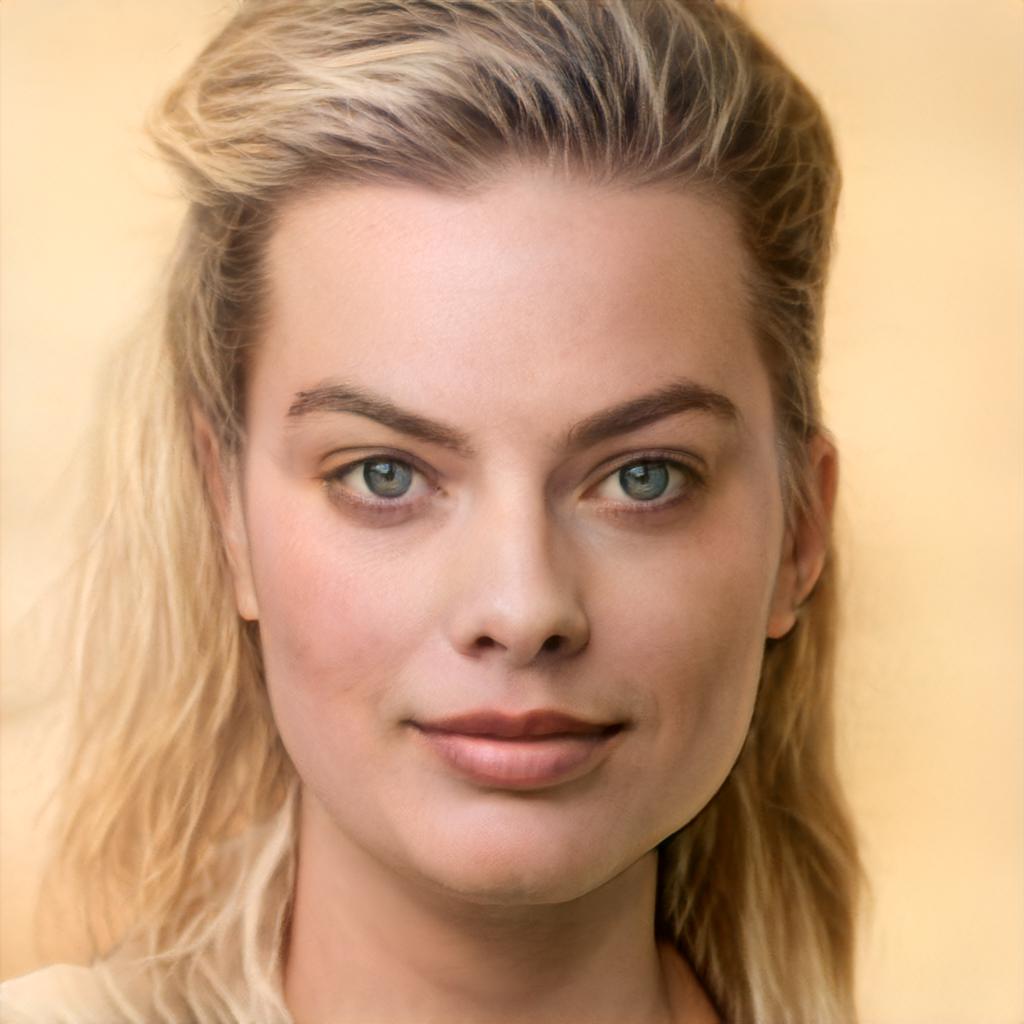} & 
        \includegraphics[width=0.27\columnwidth]{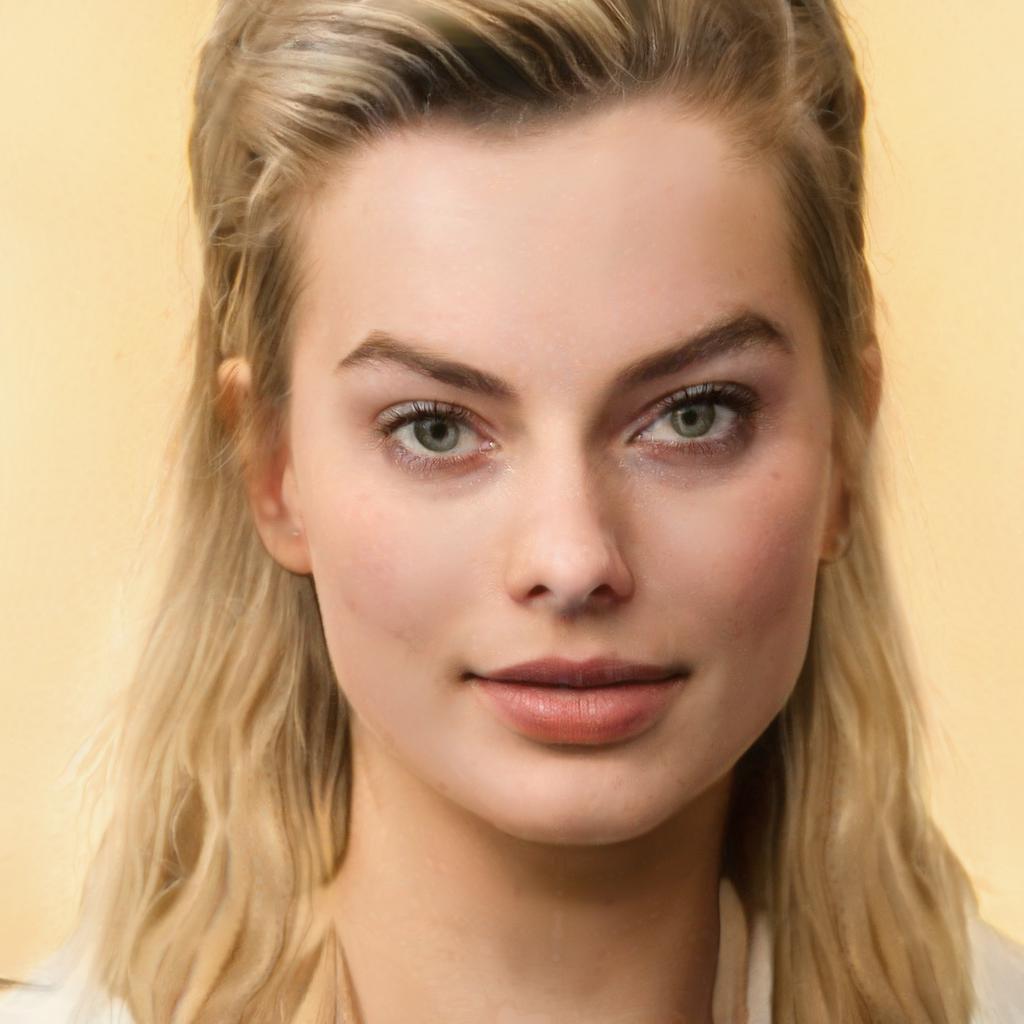} & 
        \includegraphics[width=0.27\columnwidth]{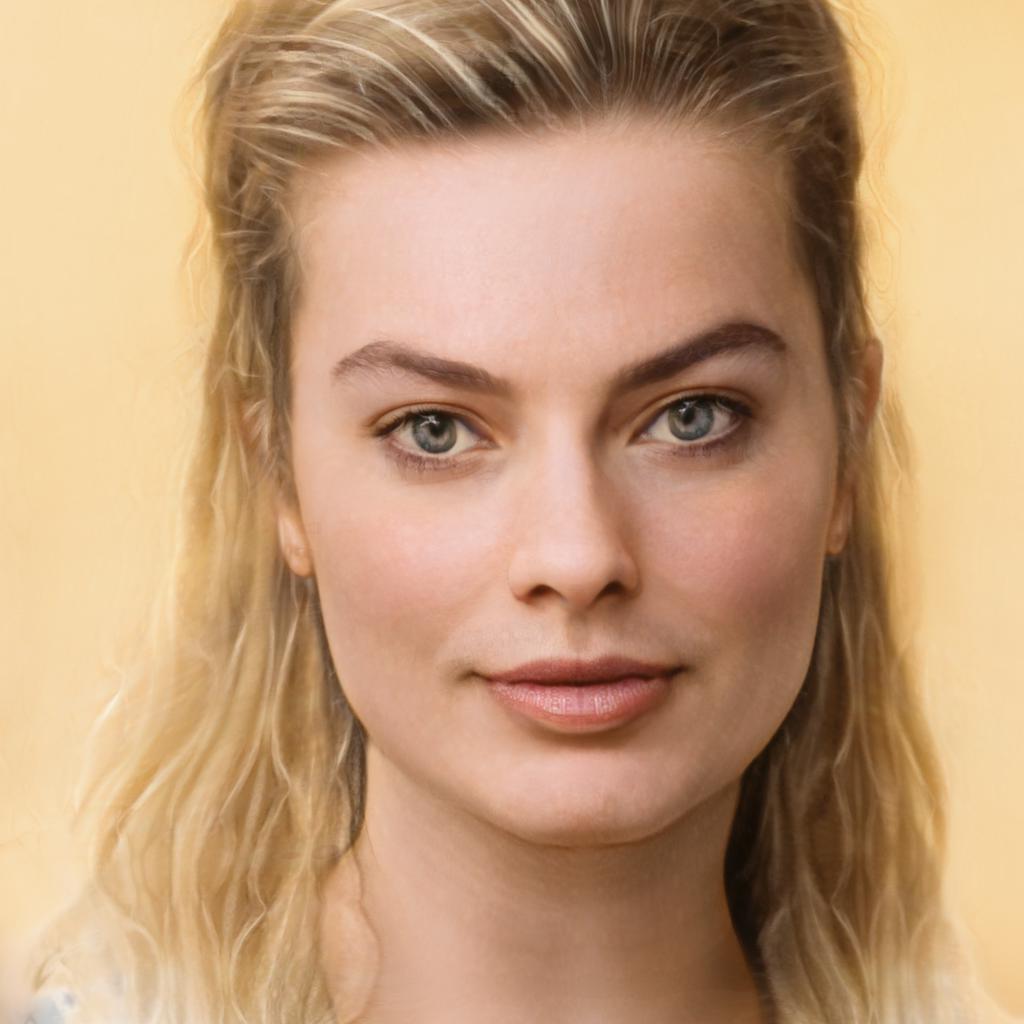} \\

		 Unaligned & SG2 $\text{ReStyle}_{e4e}$ & SG2 $\text{ReStyle}_{pSp}$ & SG3 $\text{ReStyle}_{e4e}$ & SG3 $\text{ReStyle}_{pSp}$
	\end{tabular}
	}
	\vspace{-0.1cm}
	\caption{
	Reconstruction quality comparisons between encoders trained for inverting StyleGAN2 and StyleGAN3 generators. When given unaligned source images, our StyleGAN3 encoders are able to achieve comparable reconstruction quality to their StyleGAN2 counterparts. 
	}
	\label{fig:inversions_supplementary}
\end{figure*}

%% file: figures/supplementary/inversion_comparison_2.tex
\begin{figure*}[tb]
	\centering
	\setlength{\tabcolsep}{1pt}	
	{\small
	\begin{tabular}{c c c c c c}

        \includegraphics[width=0.27\columnwidth]{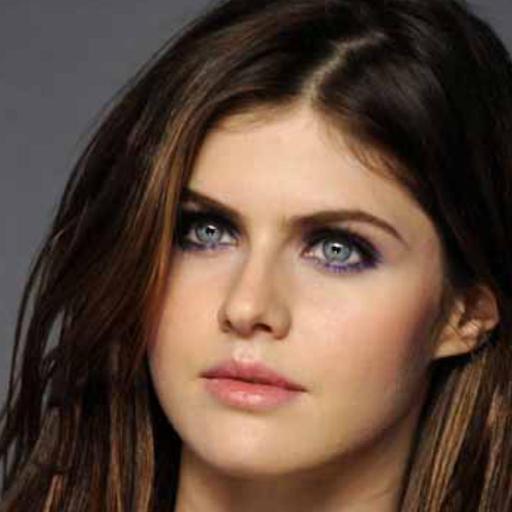} & 
        \includegraphics[width=0.27\columnwidth]{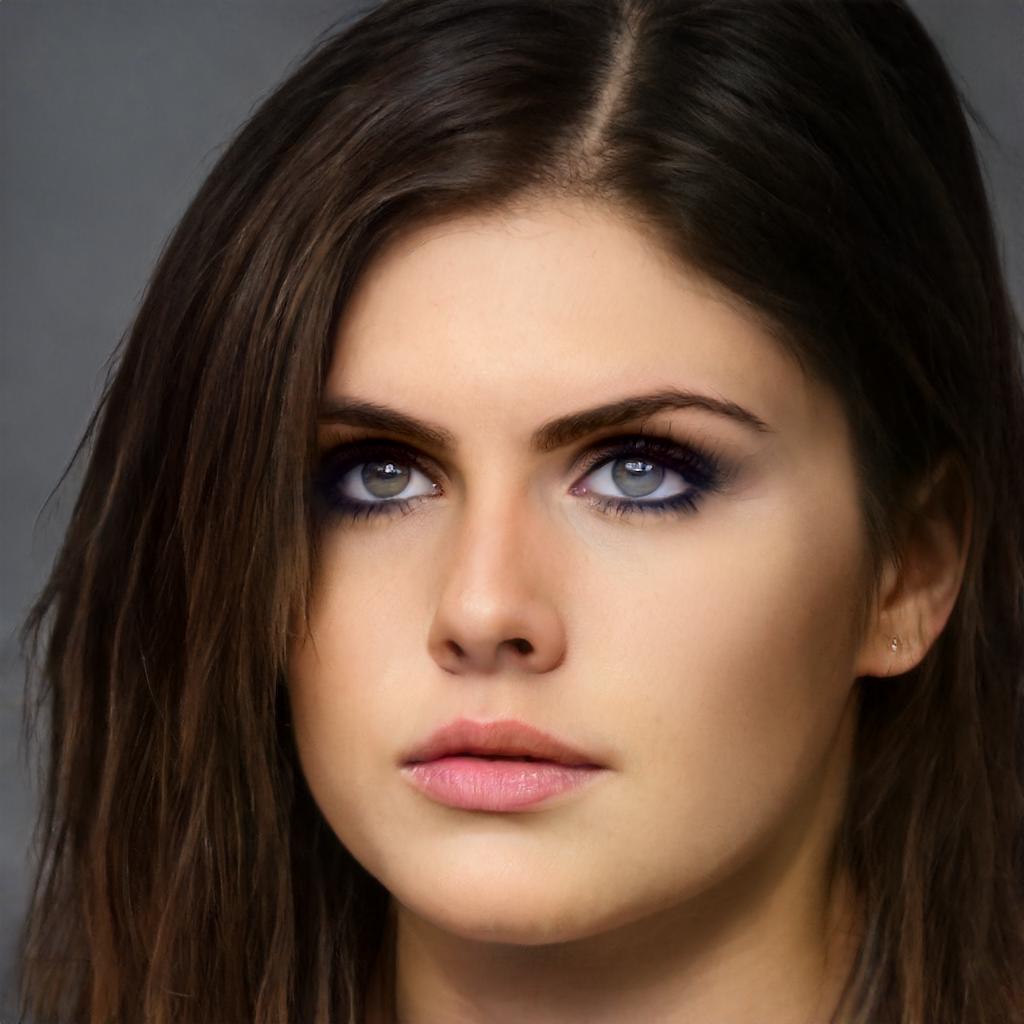} & 
        \includegraphics[width=0.27\columnwidth]{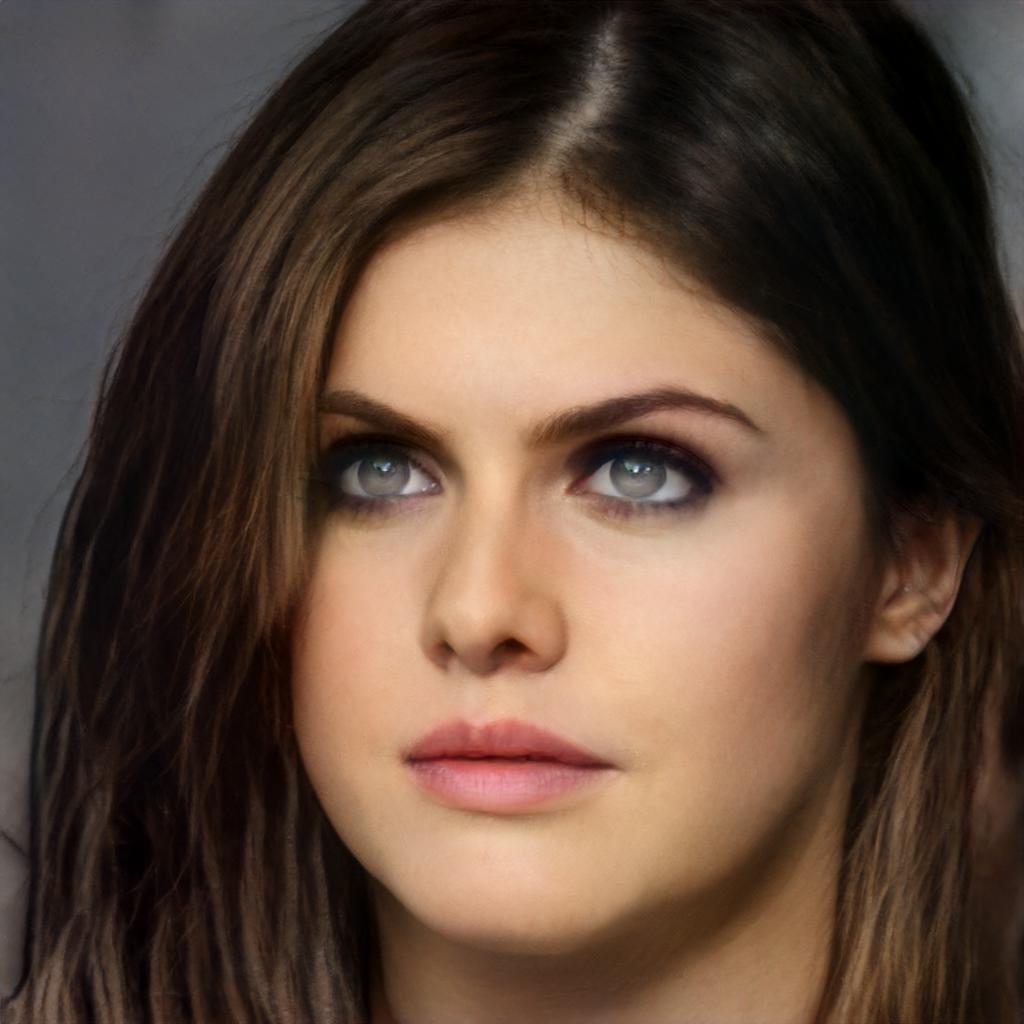} & 
        \includegraphics[width=0.27\columnwidth]{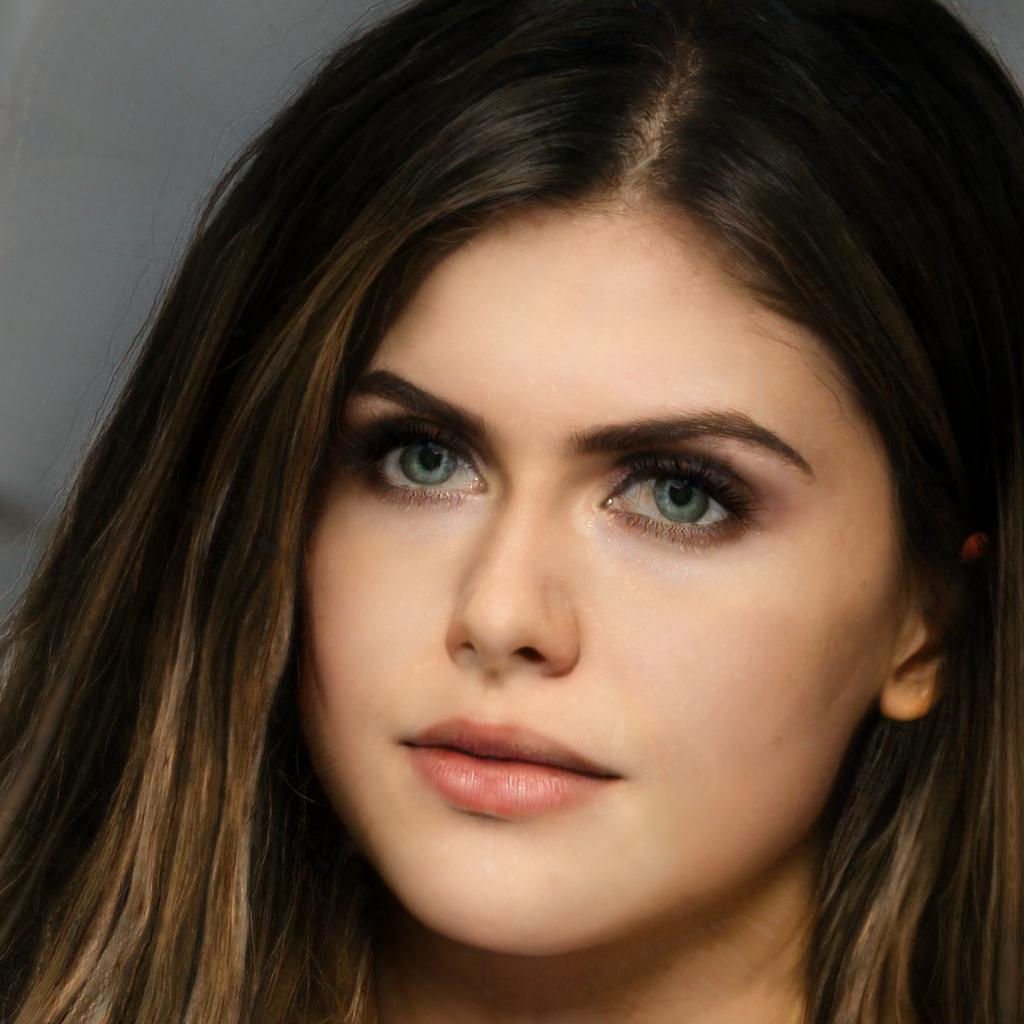} & 
        \includegraphics[width=0.27\columnwidth]{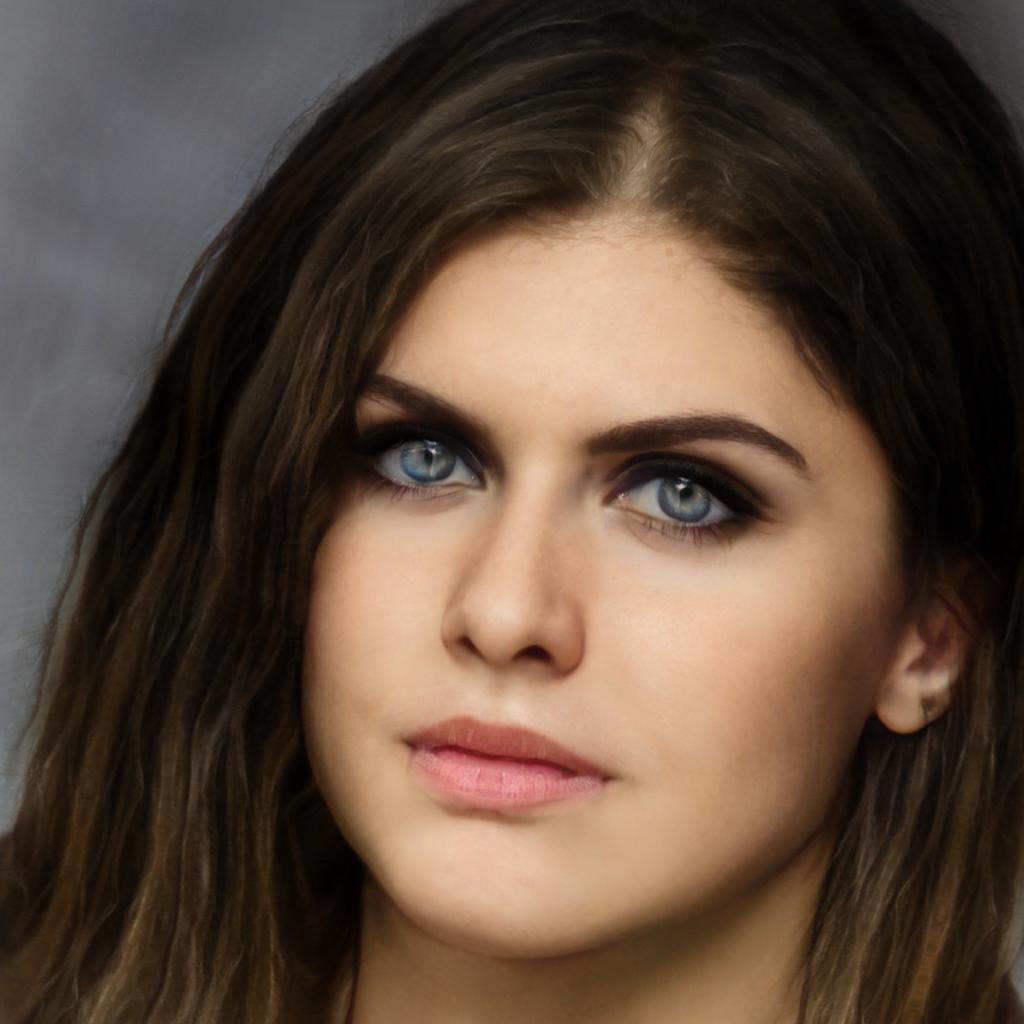} \\

        \includegraphics[width=0.27\columnwidth]{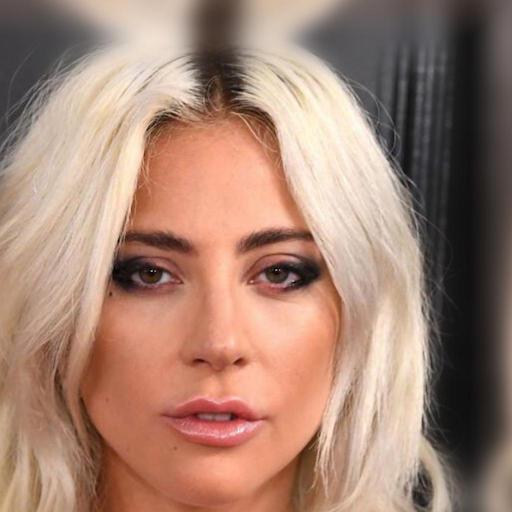} & 
        \includegraphics[width=0.27\columnwidth]{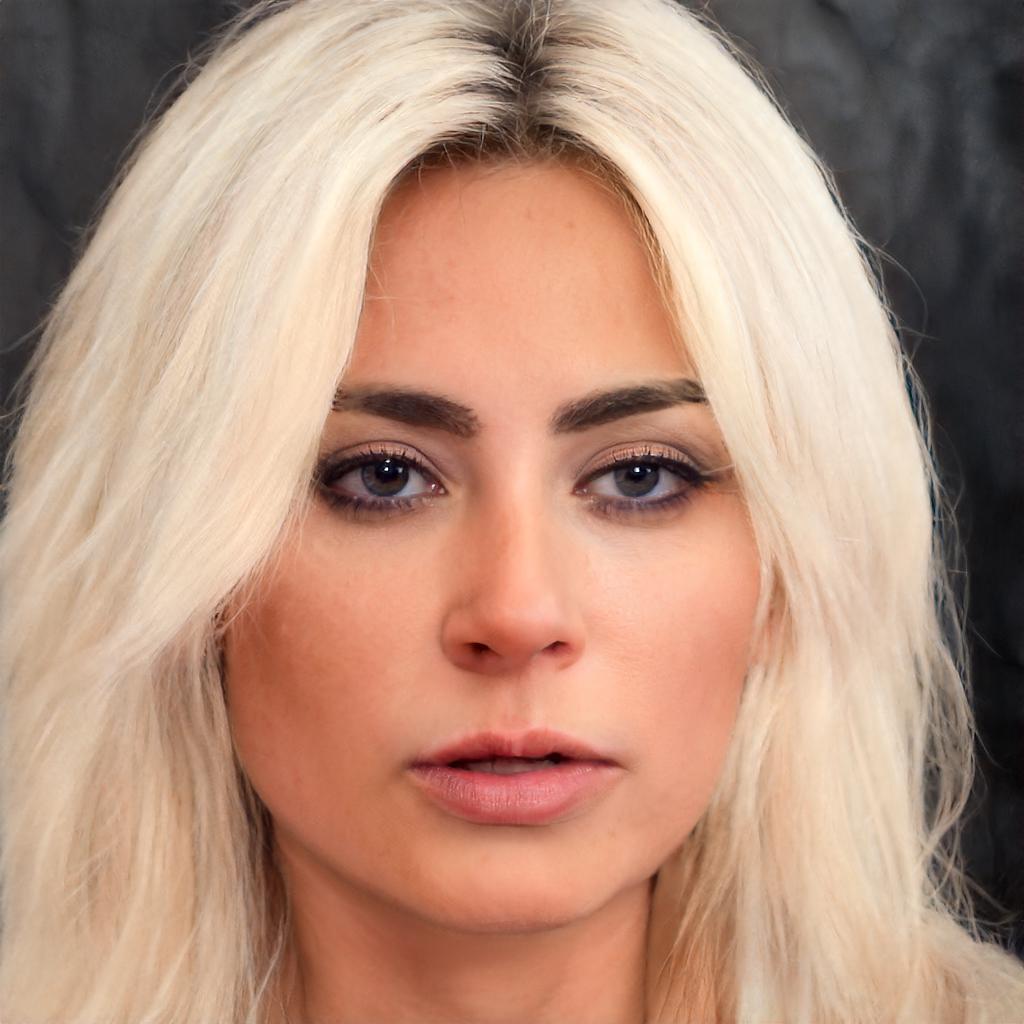} & 
        \includegraphics[width=0.27\columnwidth]{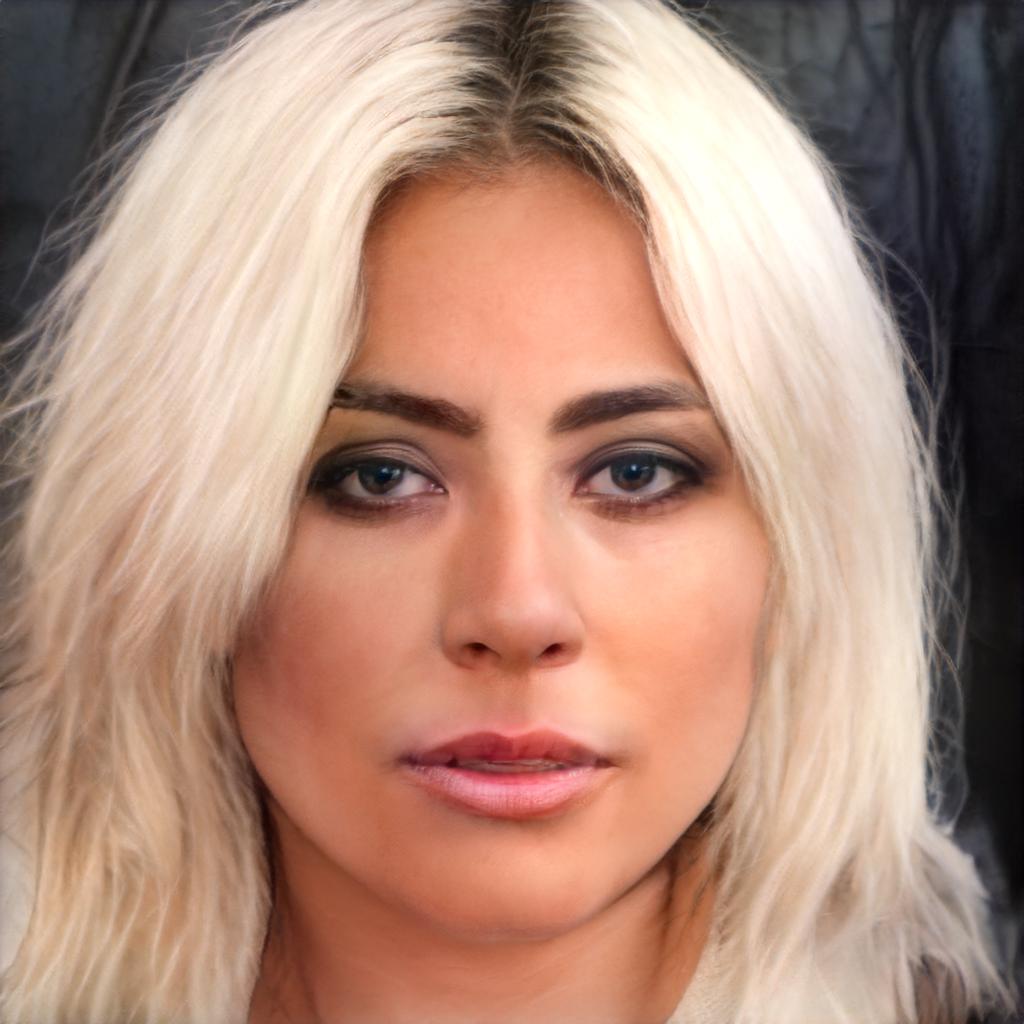} & 
        \includegraphics[width=0.27\columnwidth]{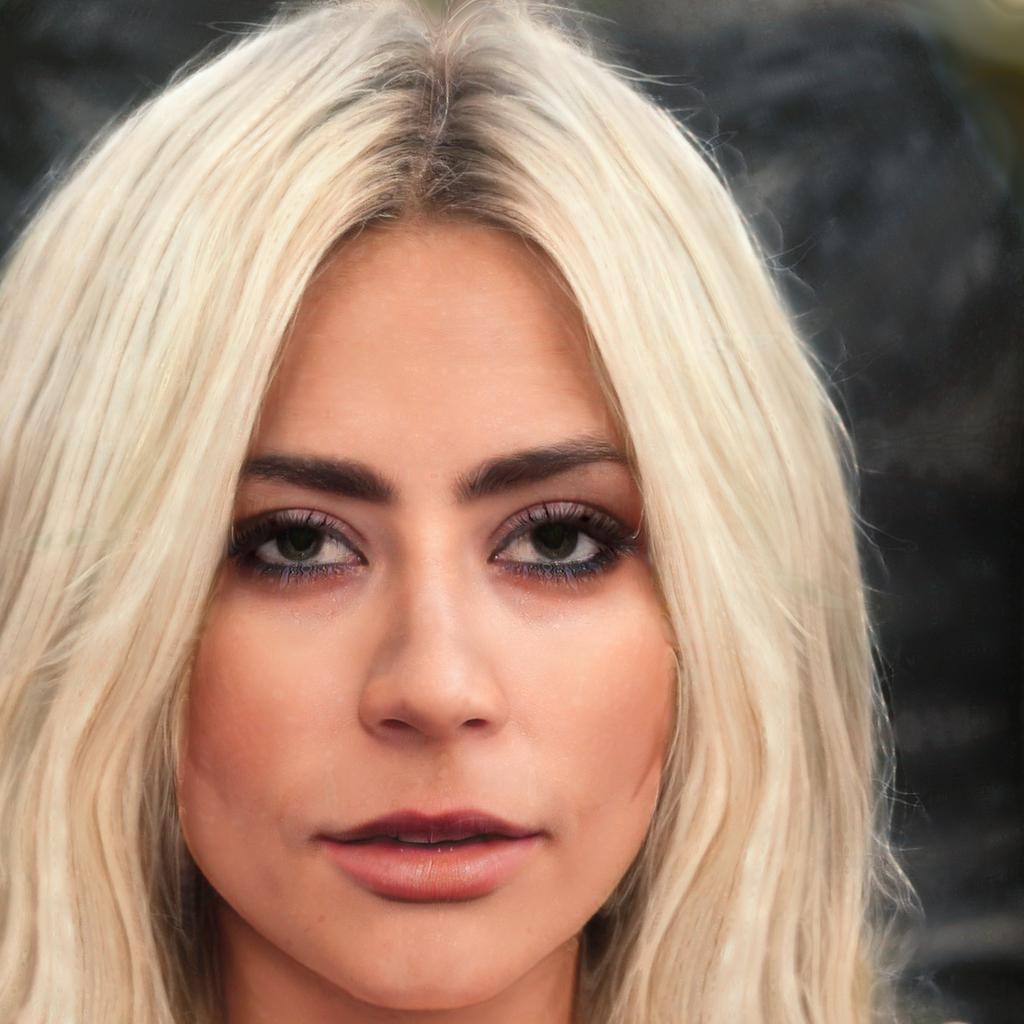} & 
        \includegraphics[width=0.27\columnwidth]{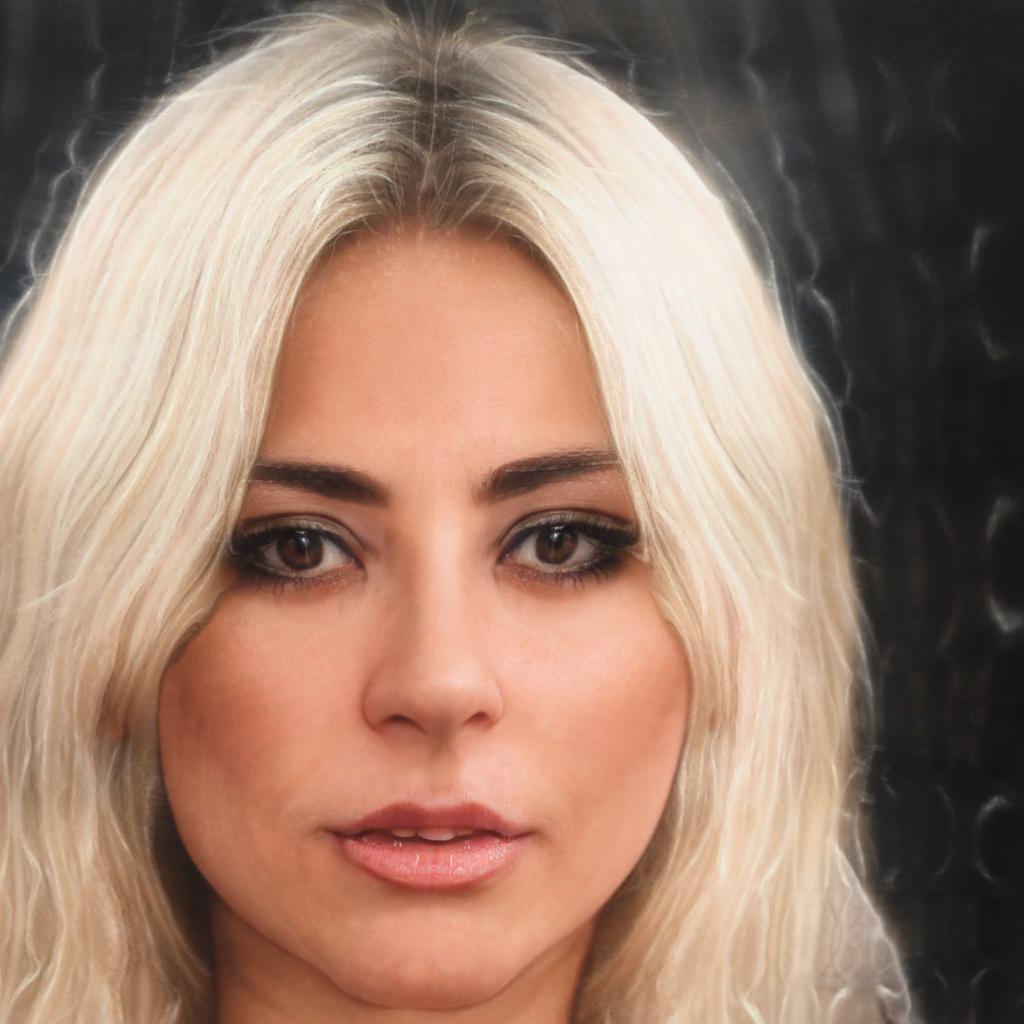} \\

        \includegraphics[width=0.27\columnwidth]{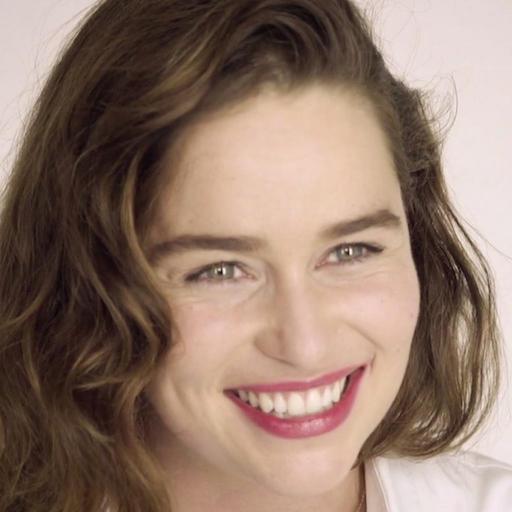} & 
        \includegraphics[width=0.27\columnwidth]{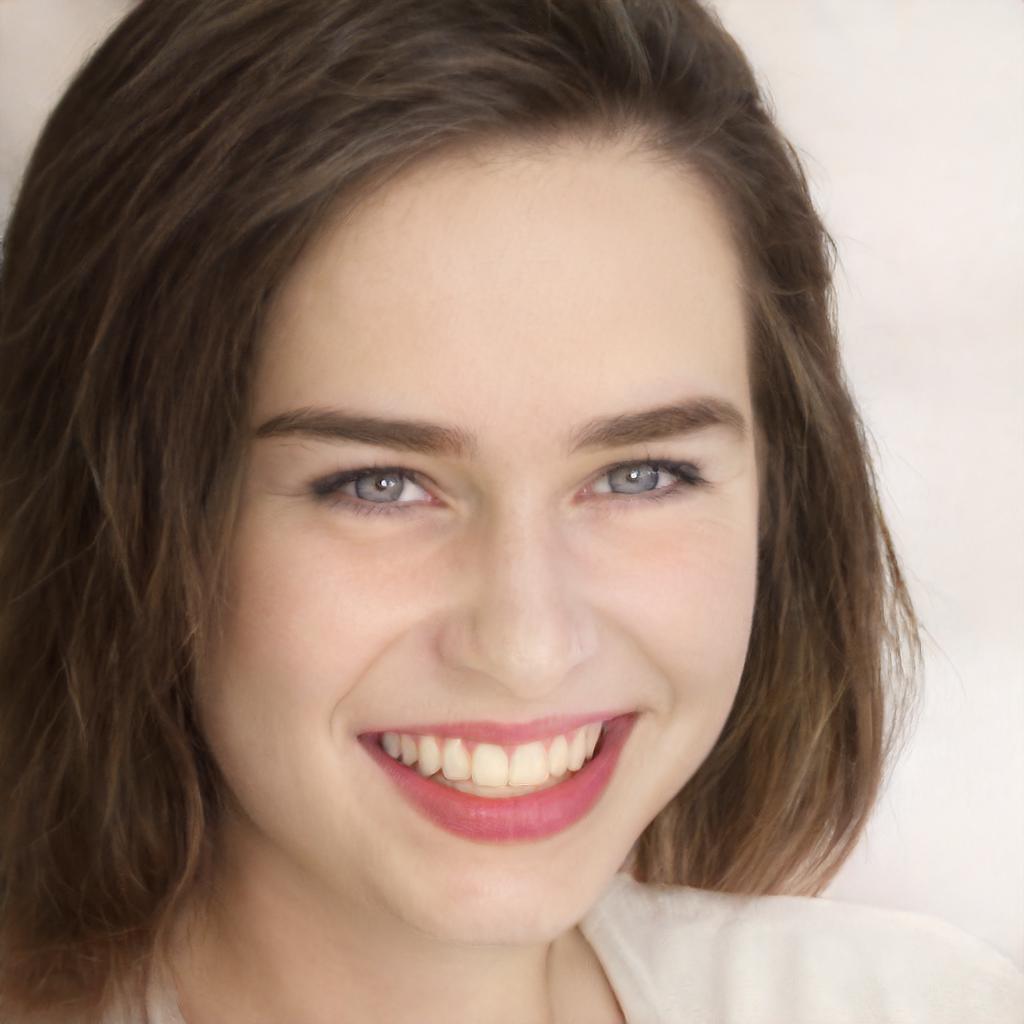} & 
        \includegraphics[width=0.27\columnwidth]{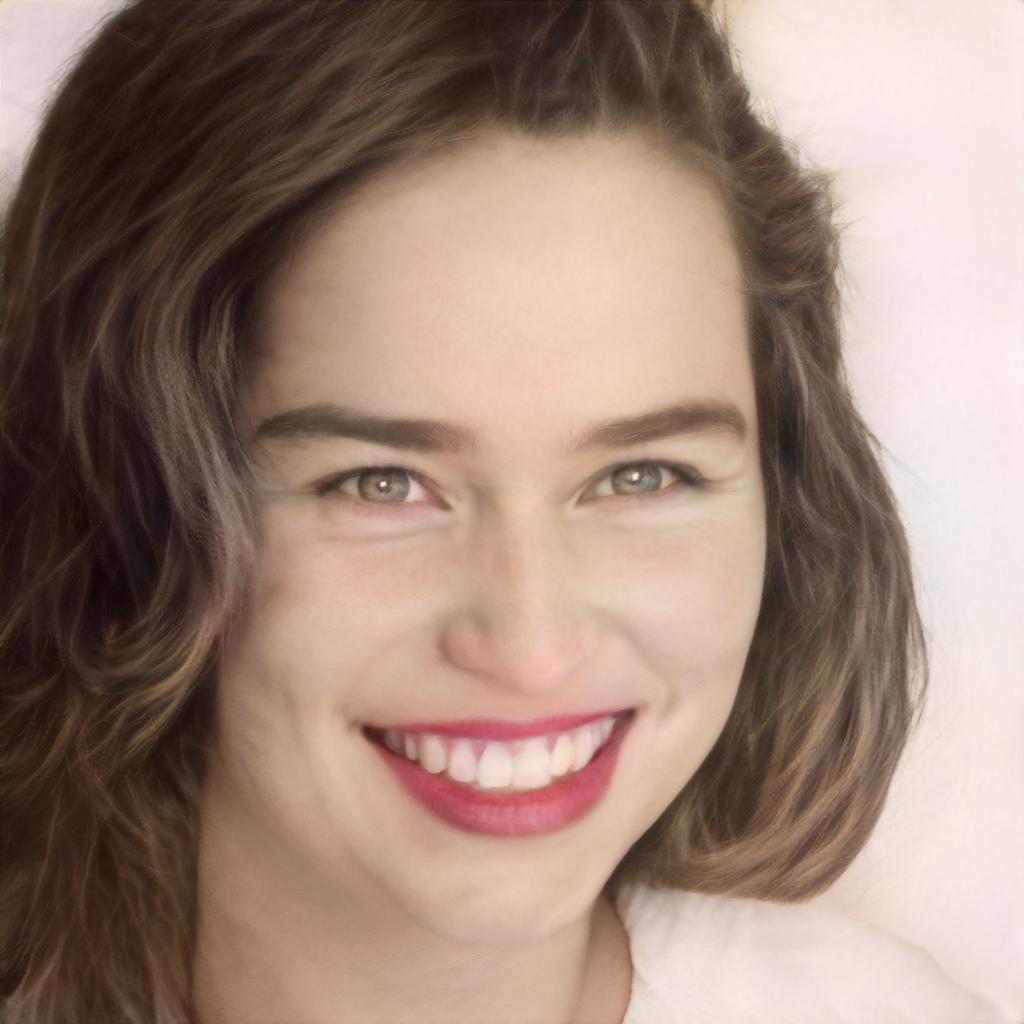} & 
        \includegraphics[width=0.27\columnwidth]{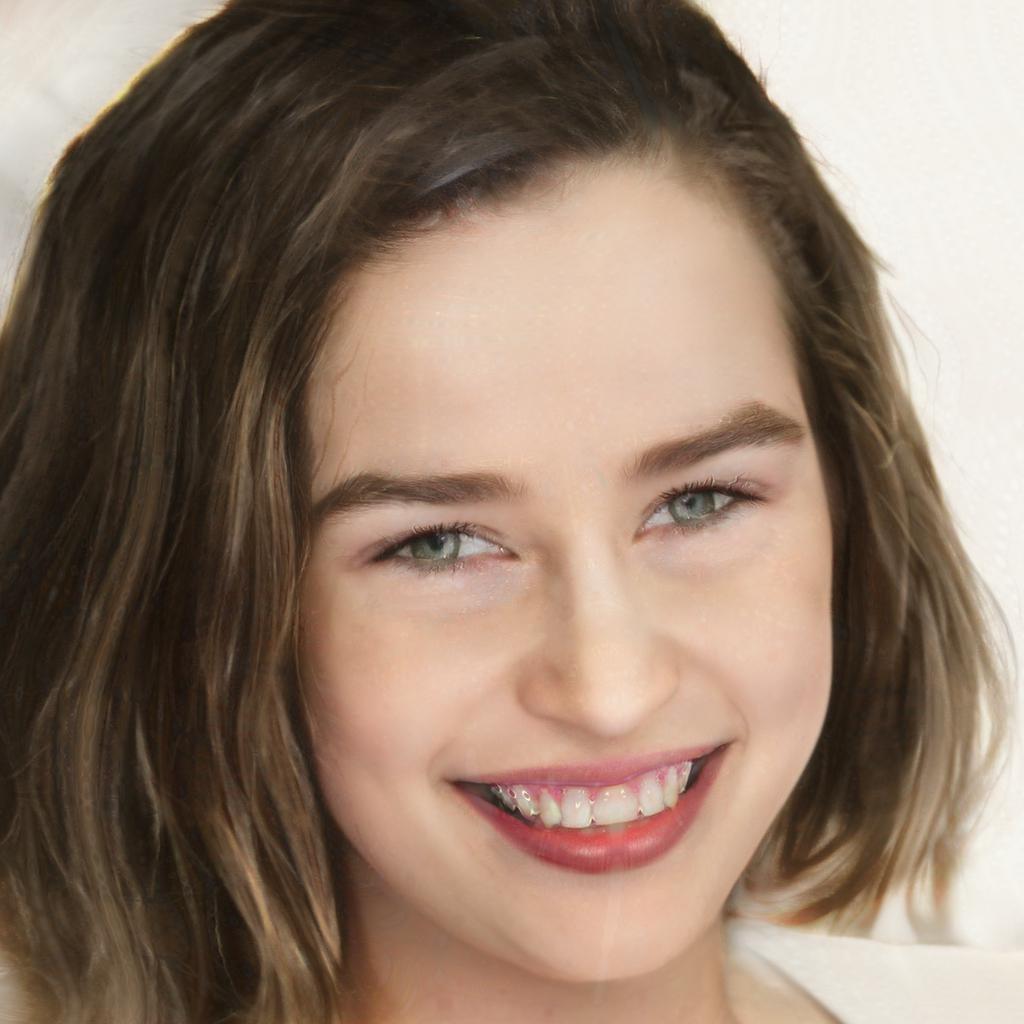} & 
        \includegraphics[width=0.27\columnwidth]{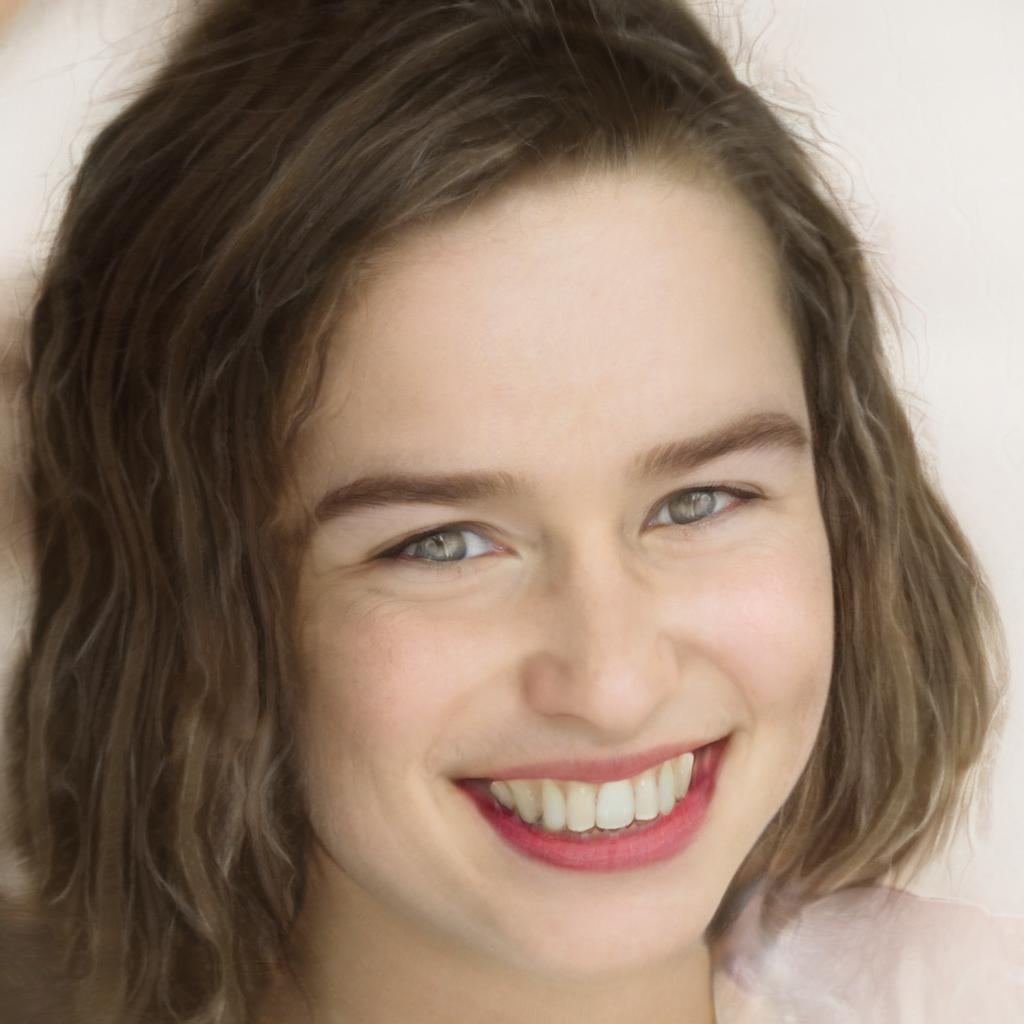} \\
        
        \includegraphics[width=0.27\columnwidth]{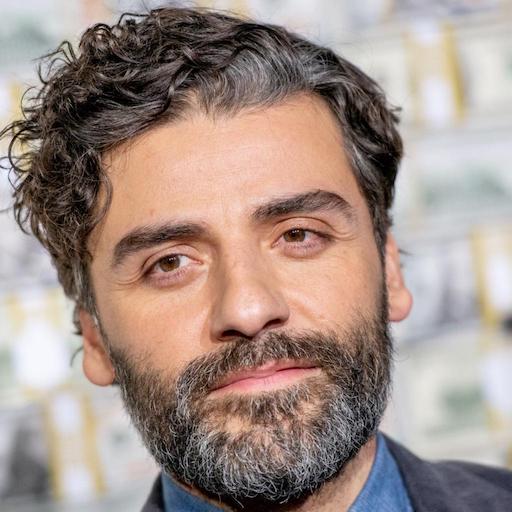} & 
        \includegraphics[width=0.27\columnwidth]{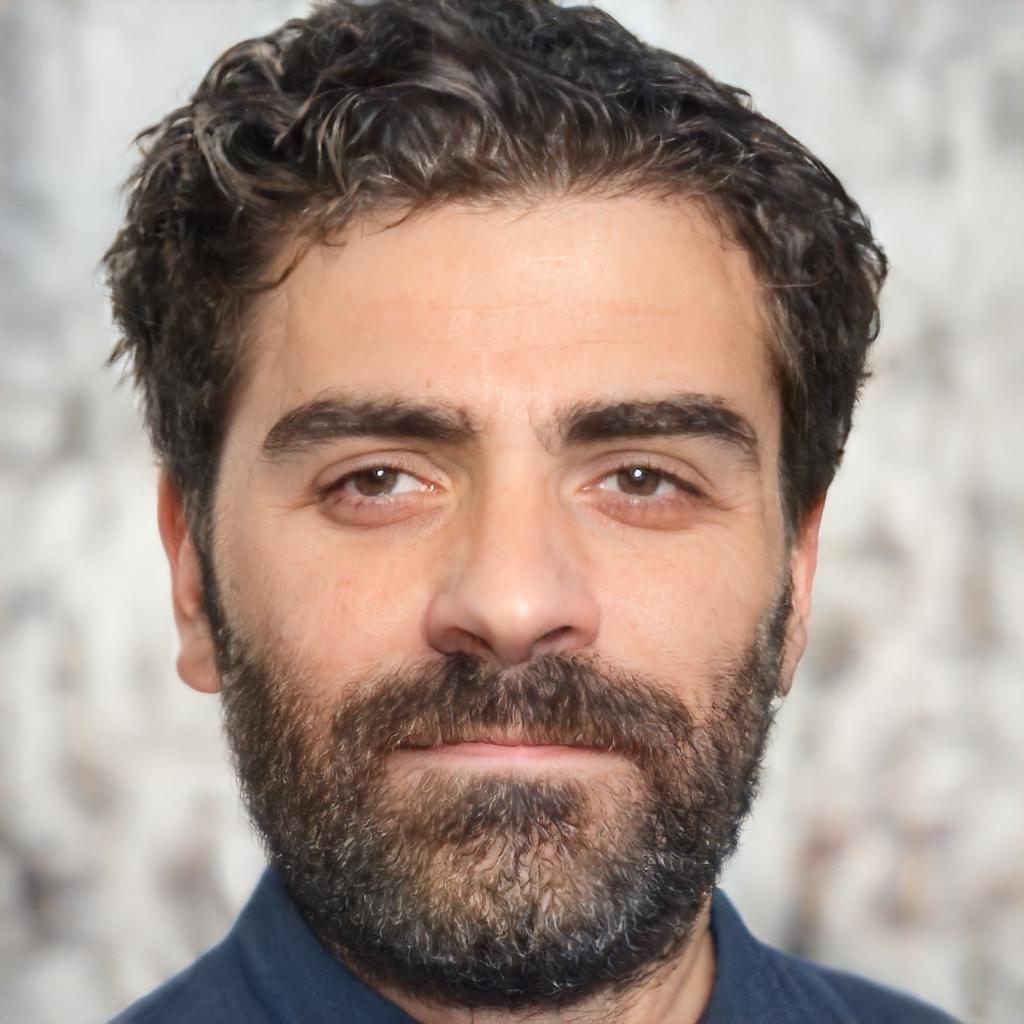} & 
        \includegraphics[width=0.27\columnwidth]{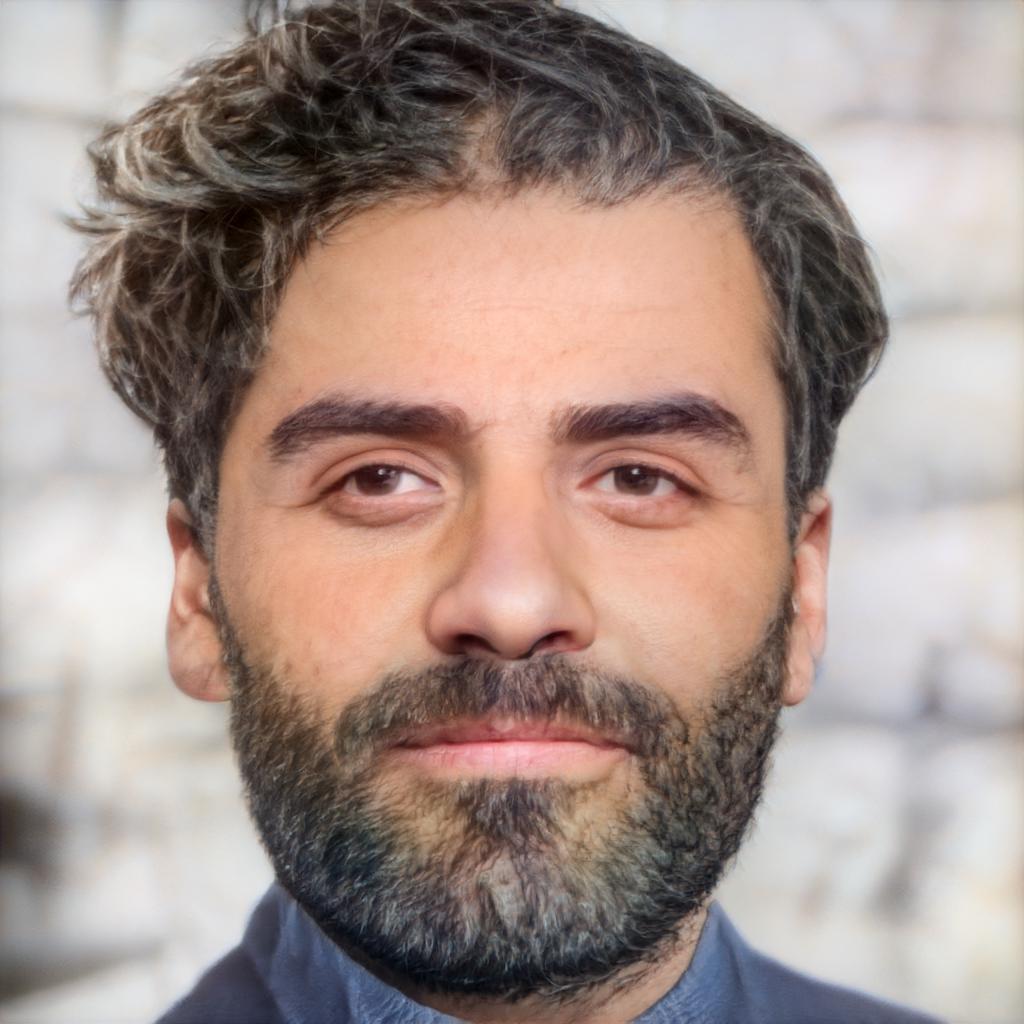} & 
        \includegraphics[width=0.27\columnwidth]{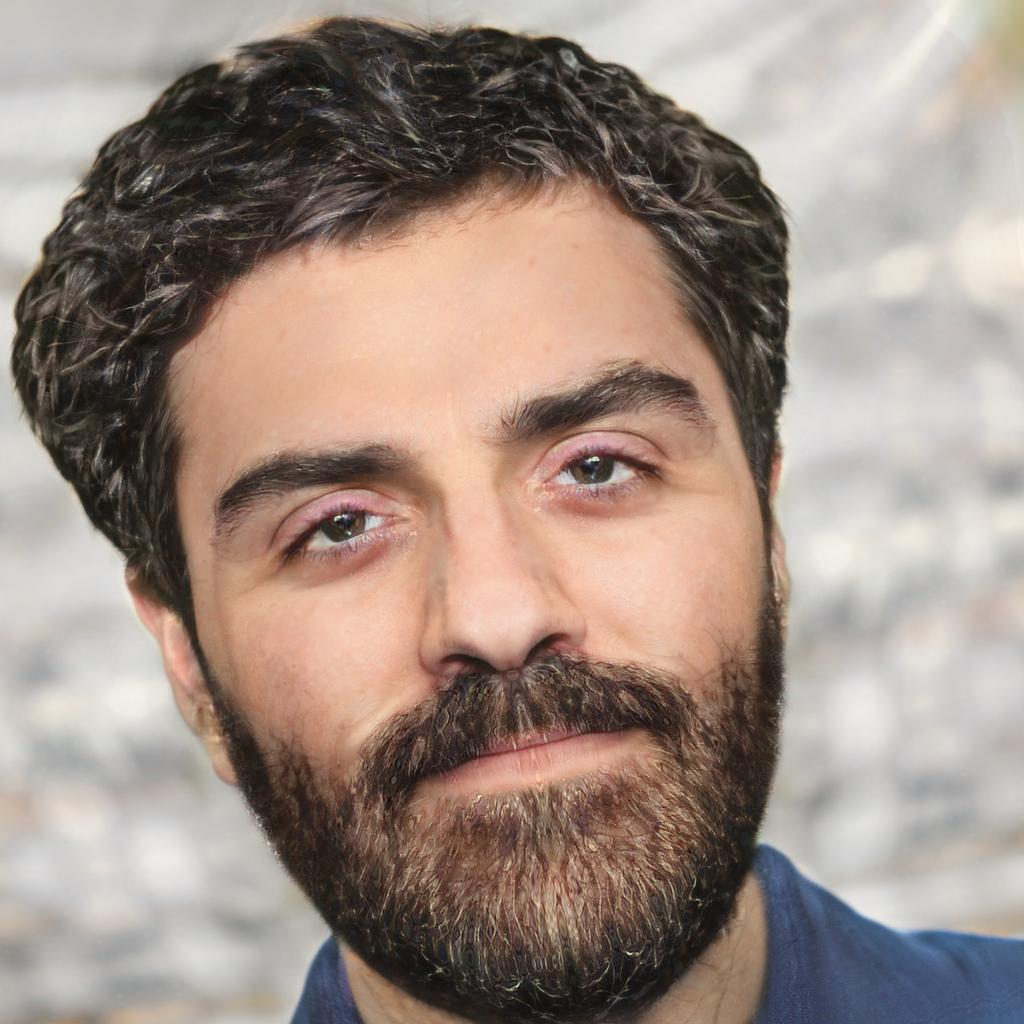} & 
        \includegraphics[width=0.27\columnwidth]{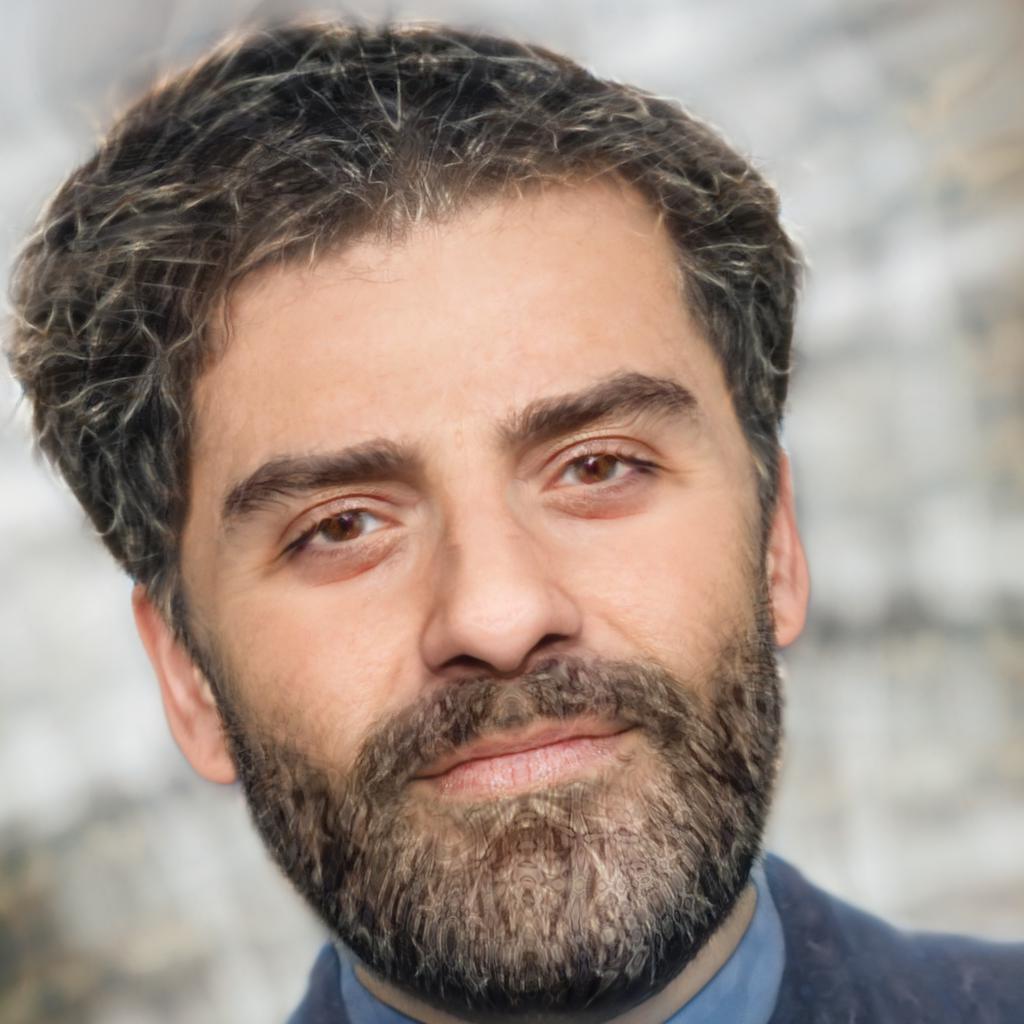} \\

        \includegraphics[width=0.27\columnwidth]{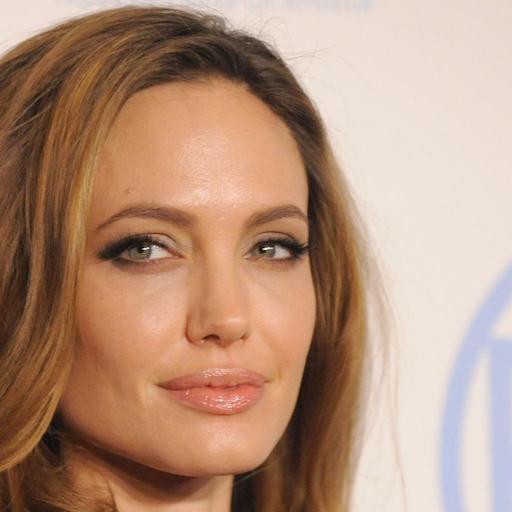} & 
        \includegraphics[width=0.27\columnwidth]{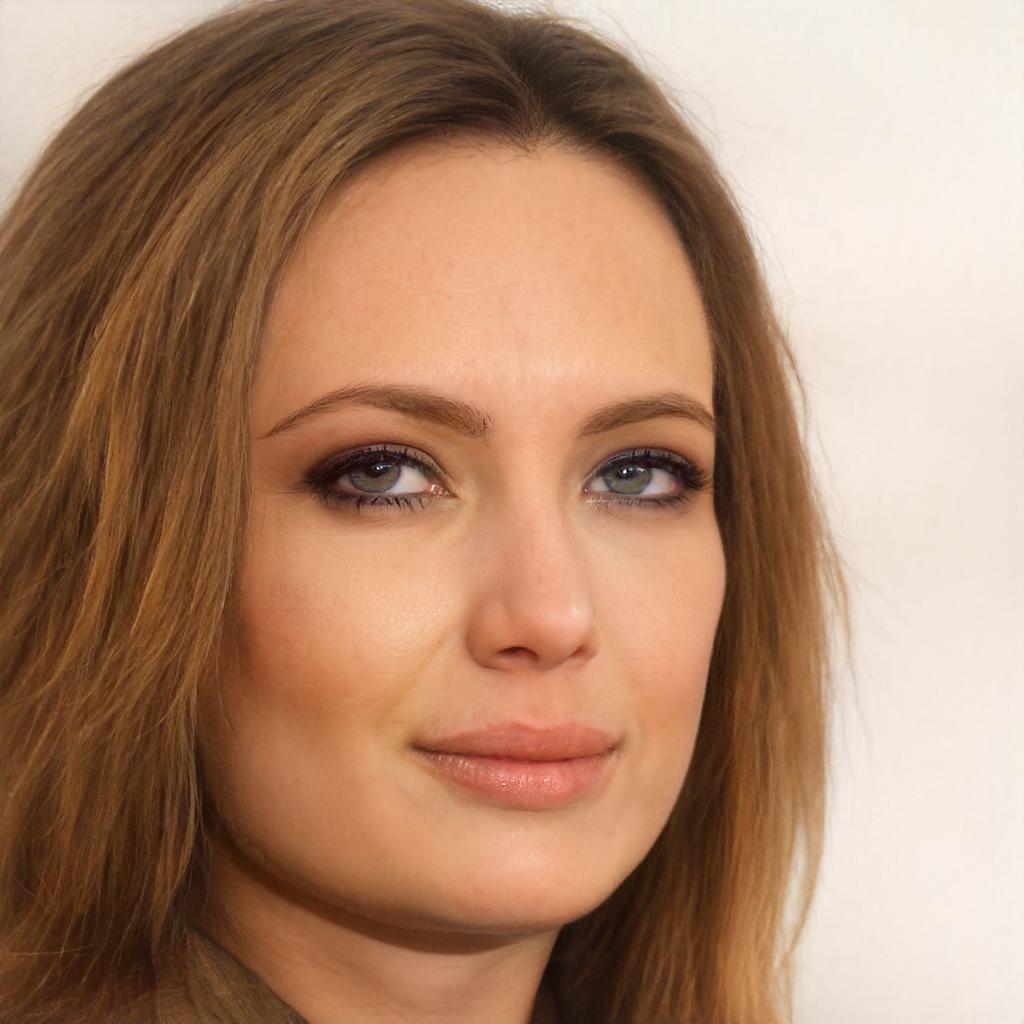} & 
        \includegraphics[width=0.27\columnwidth]{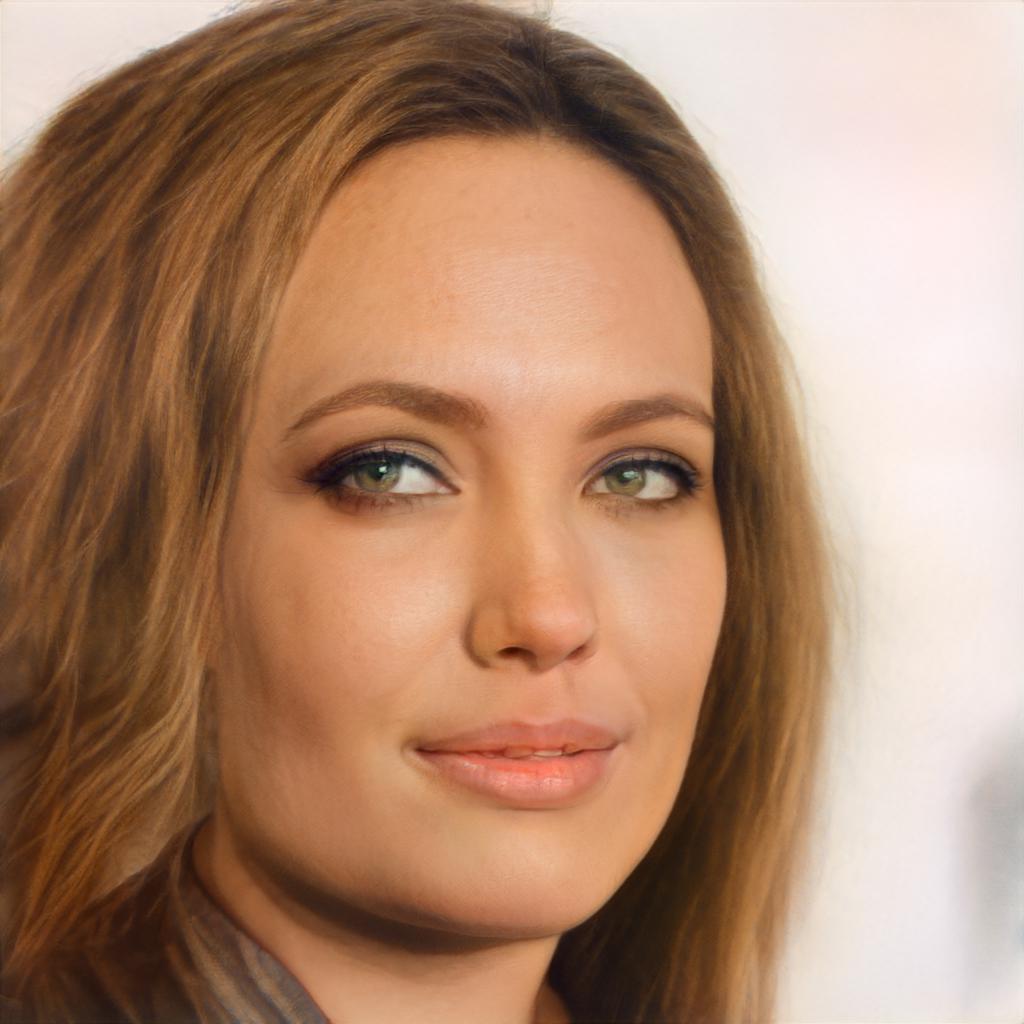} & 
        \includegraphics[width=0.27\columnwidth]{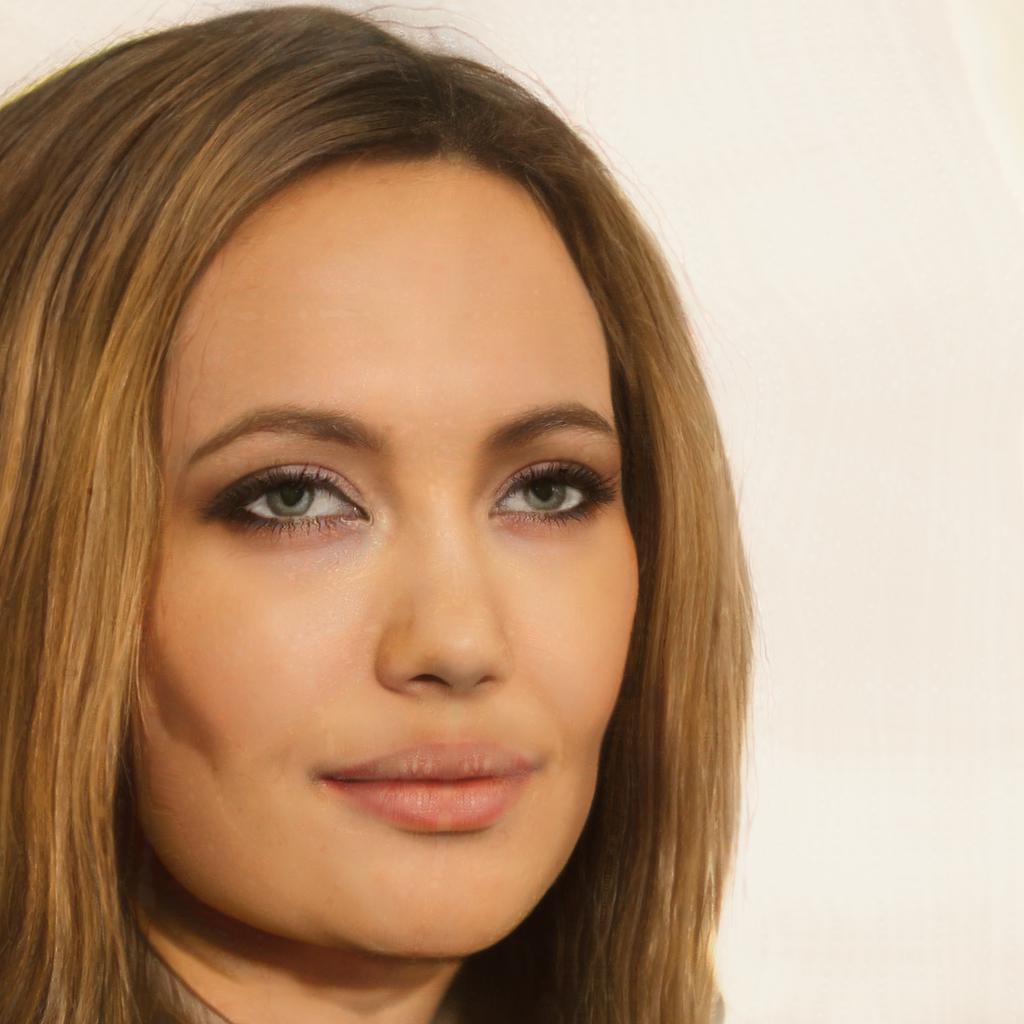} & 
        \includegraphics[width=0.27\columnwidth]{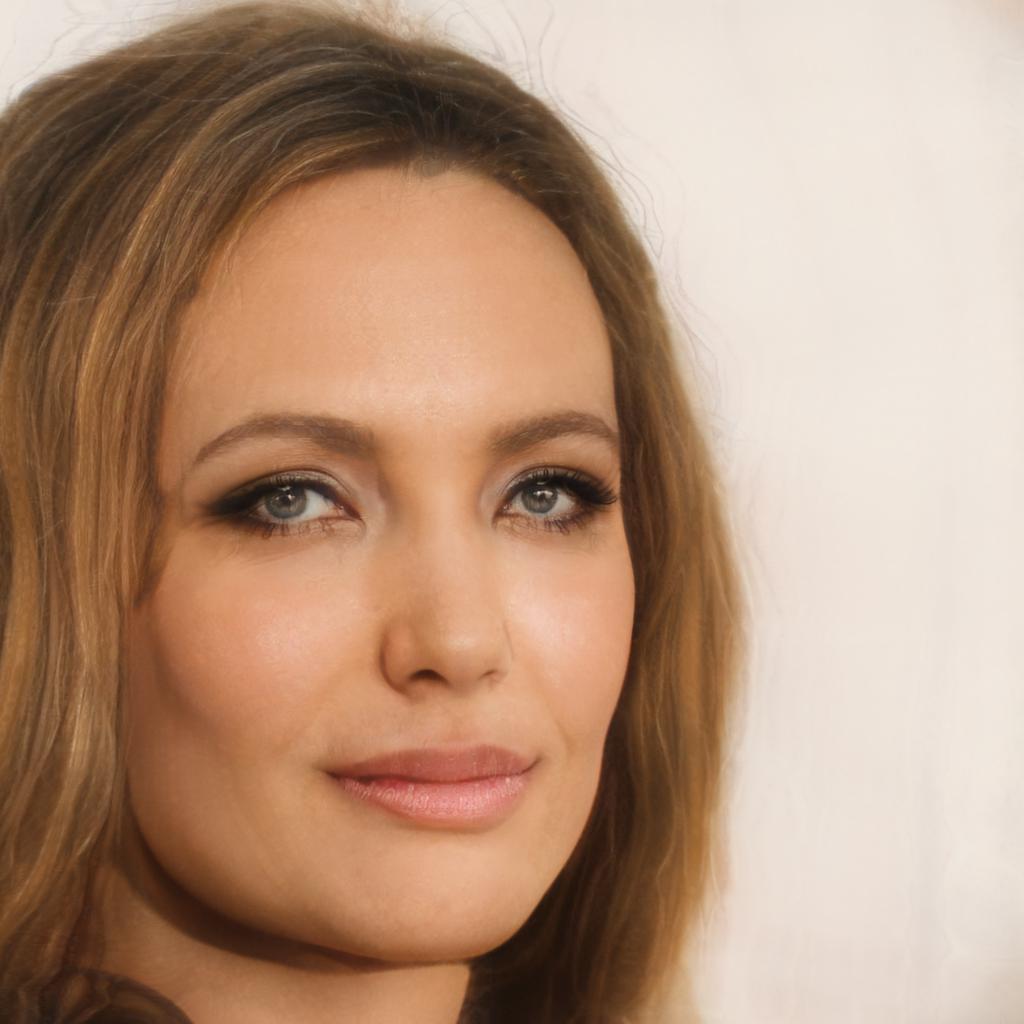} \\

        \includegraphics[width=0.27\columnwidth]{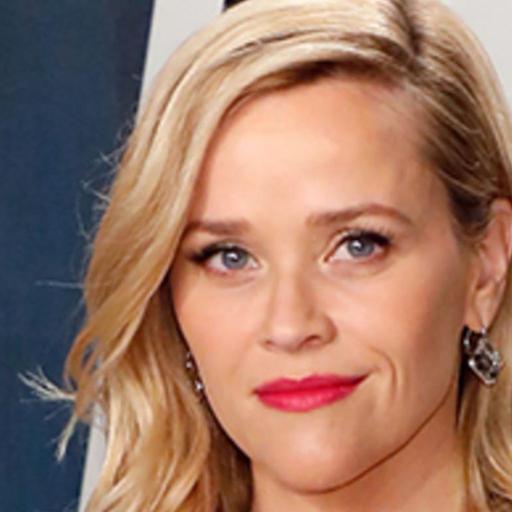} & 
        \includegraphics[width=0.27\columnwidth]{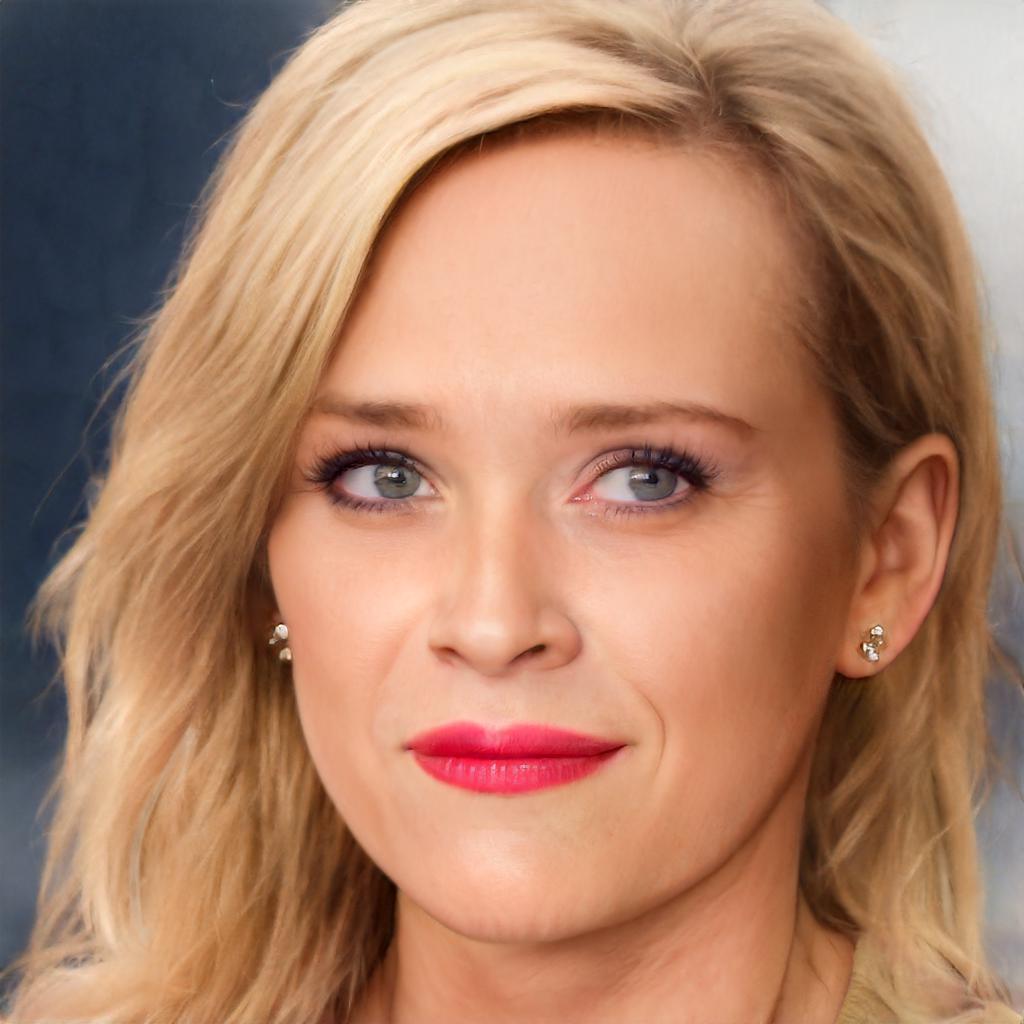} & 
        \includegraphics[width=0.27\columnwidth]{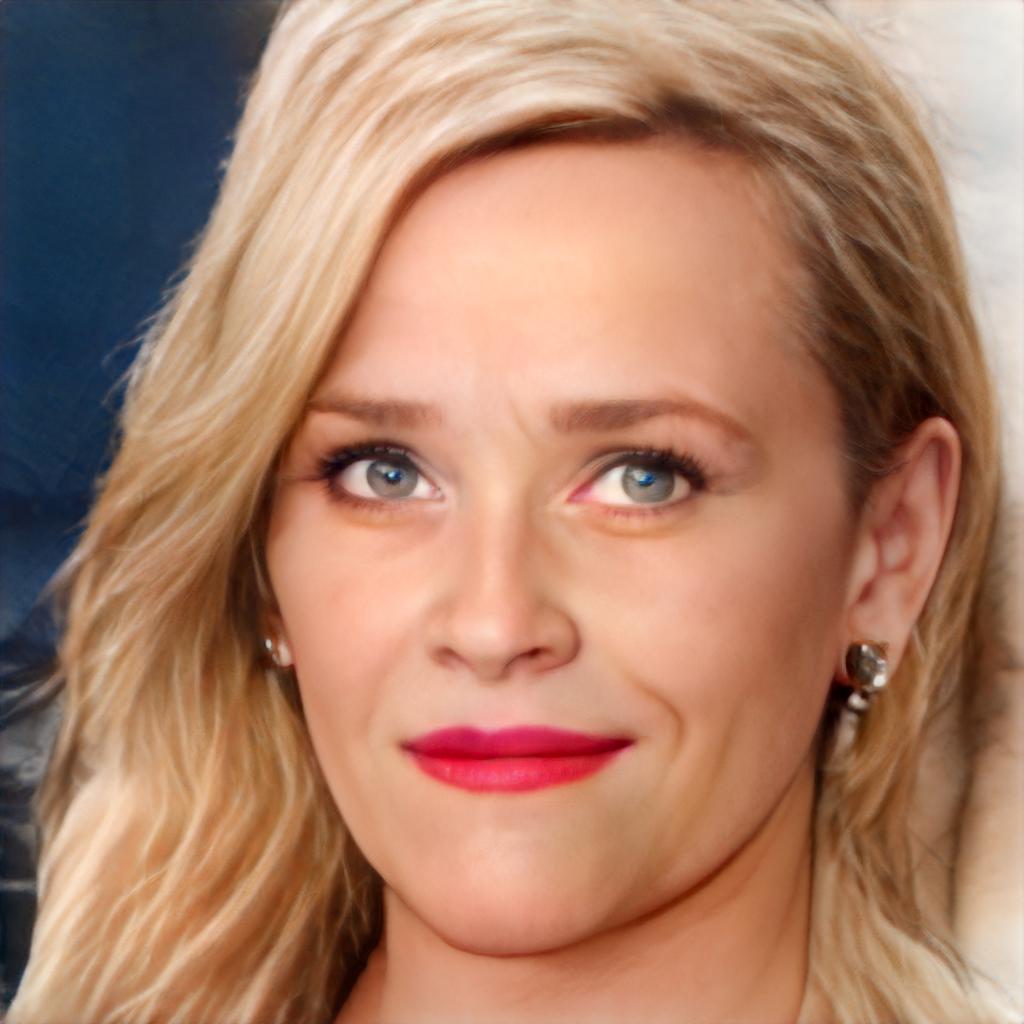} & 
        \includegraphics[width=0.27\columnwidth]{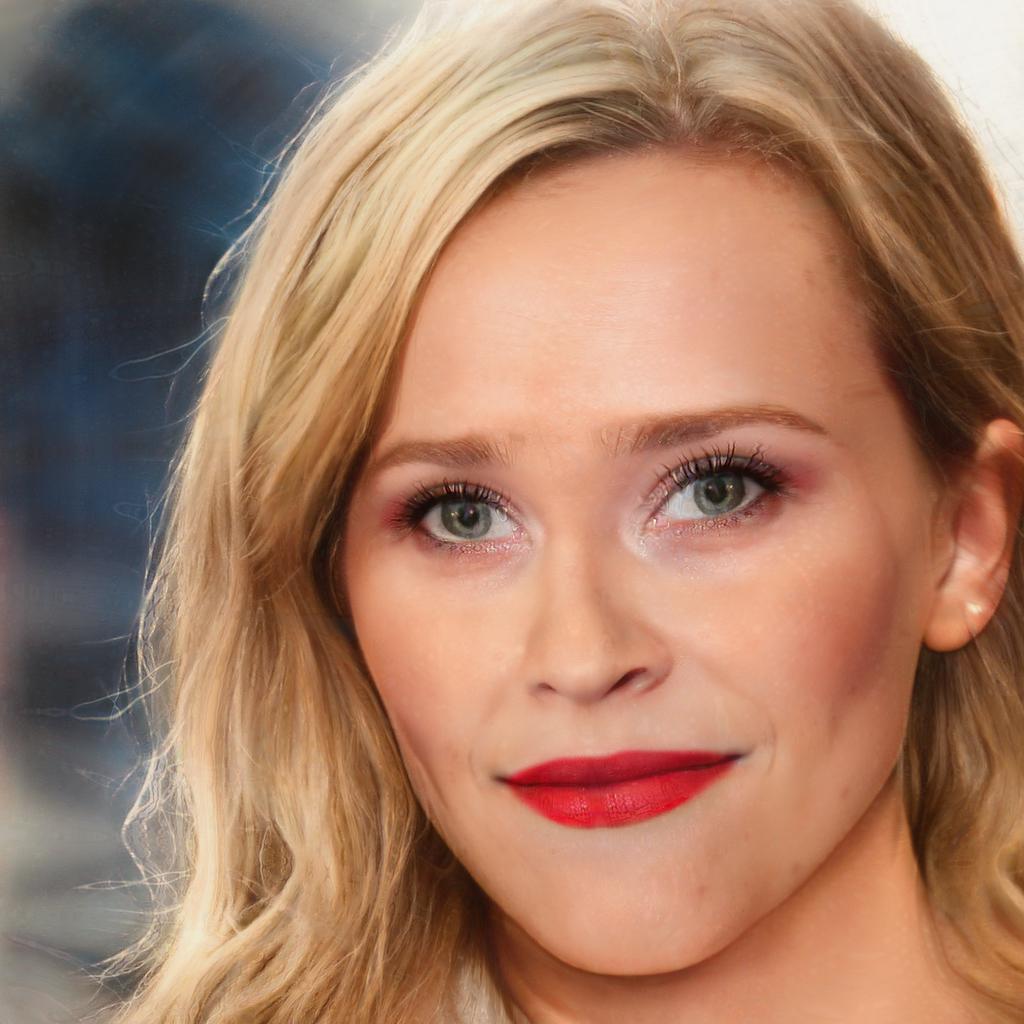} & 
        \includegraphics[width=0.27\columnwidth]{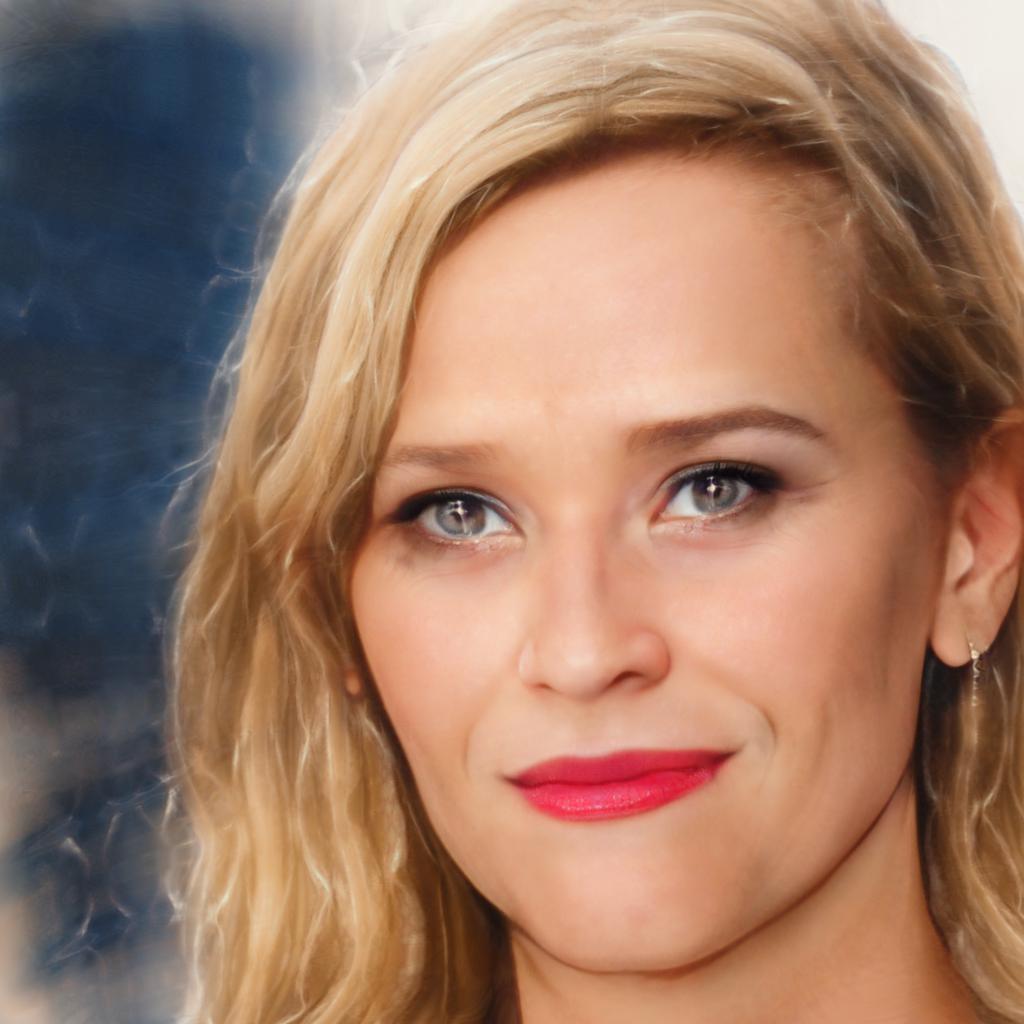} \\
        
        \includegraphics[width=0.27\columnwidth]{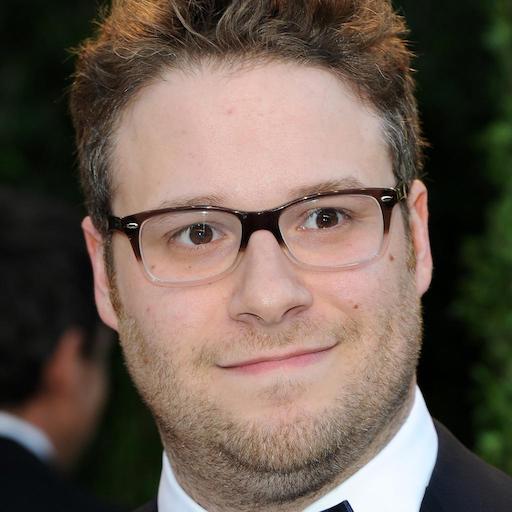} & 
        \includegraphics[width=0.27\columnwidth]{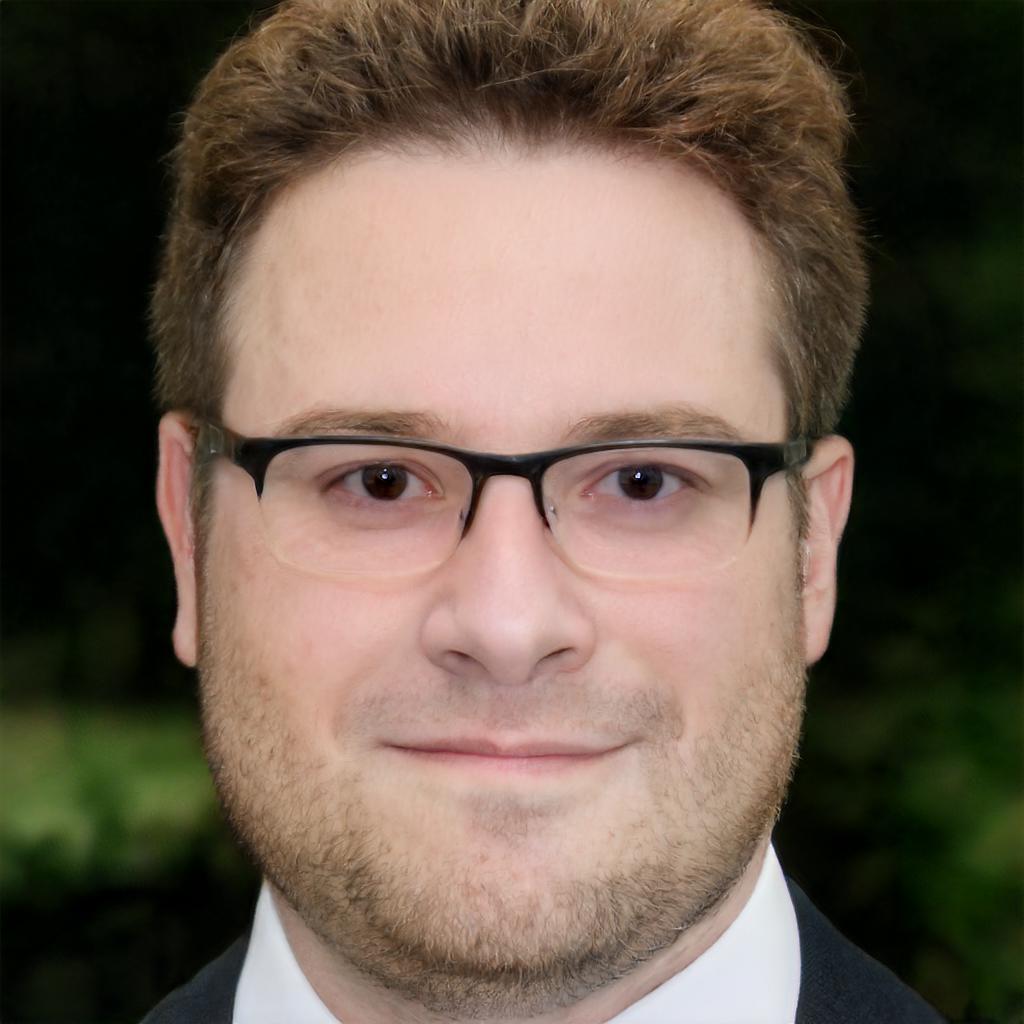} & 
        \includegraphics[width=0.27\columnwidth]{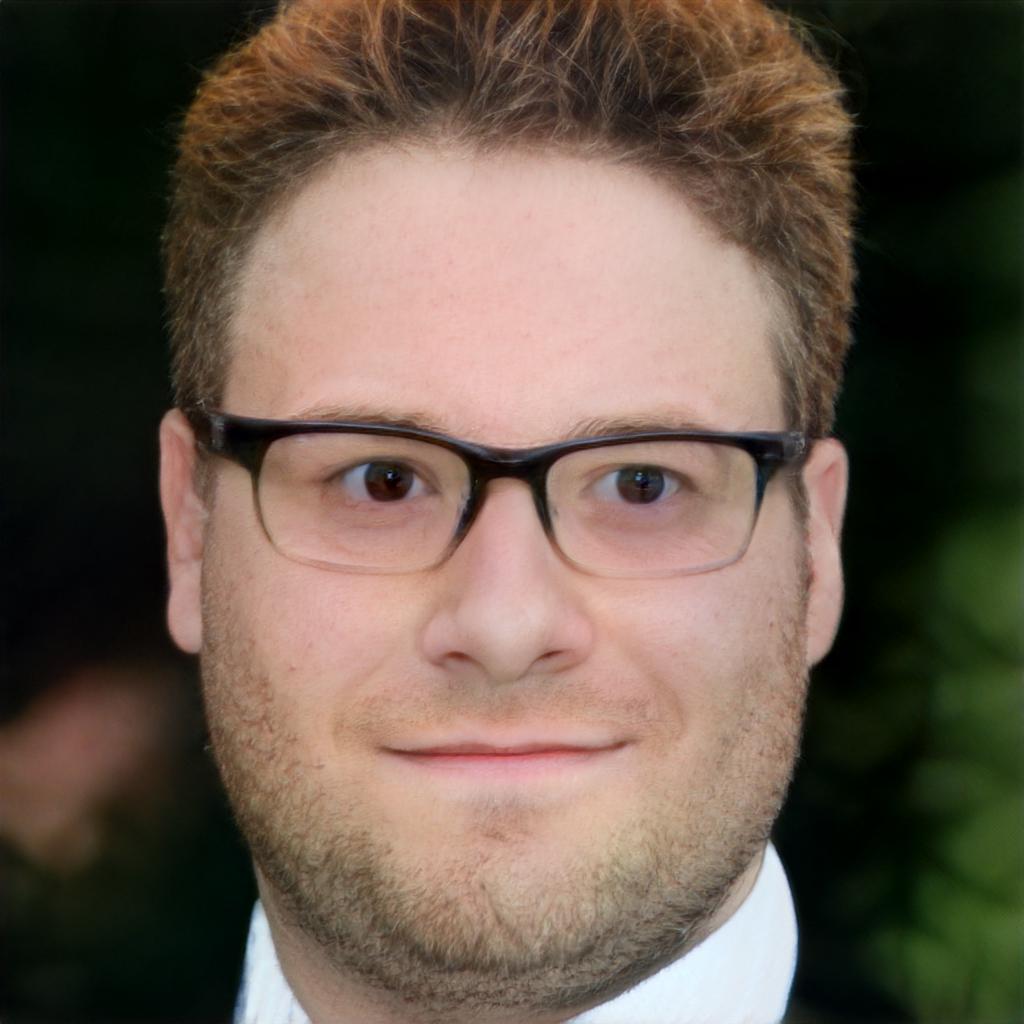} & 
        \includegraphics[width=0.27\columnwidth]{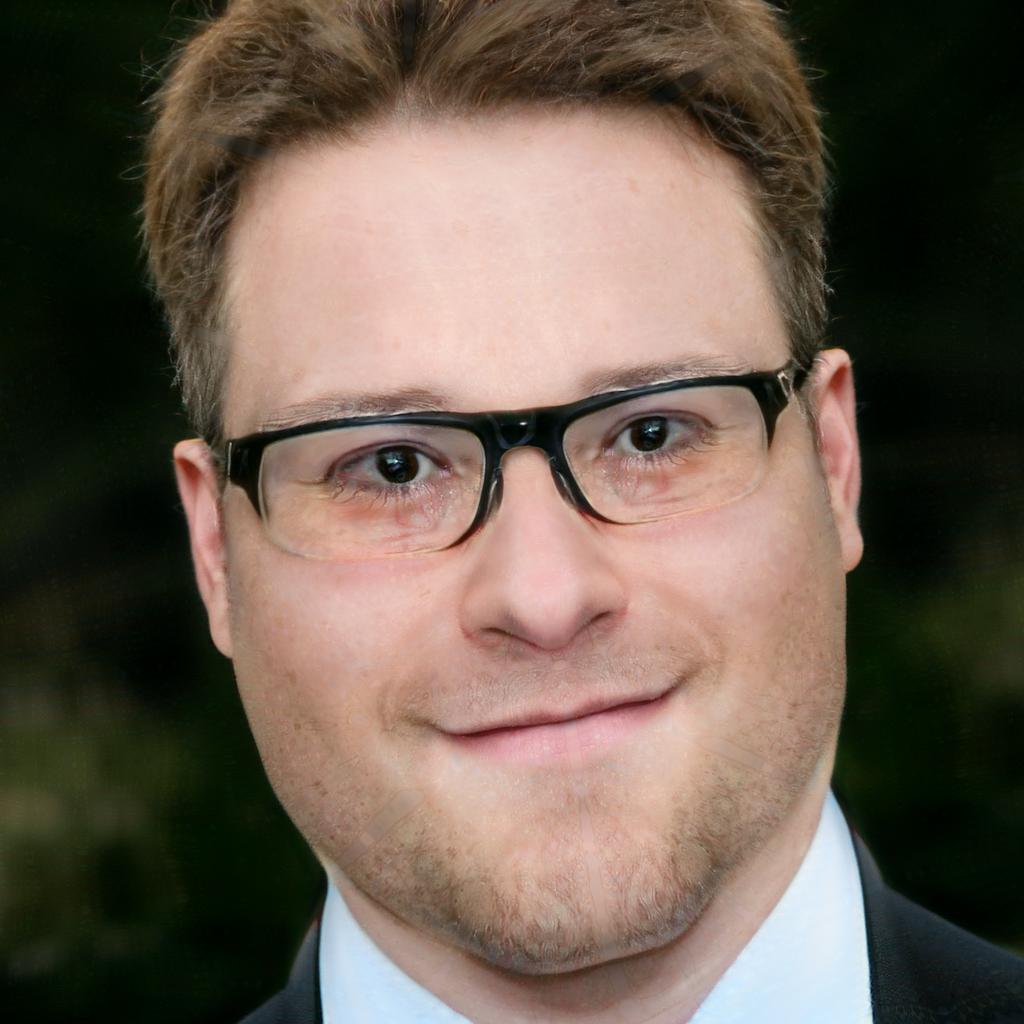} & 
        \includegraphics[width=0.27\columnwidth]{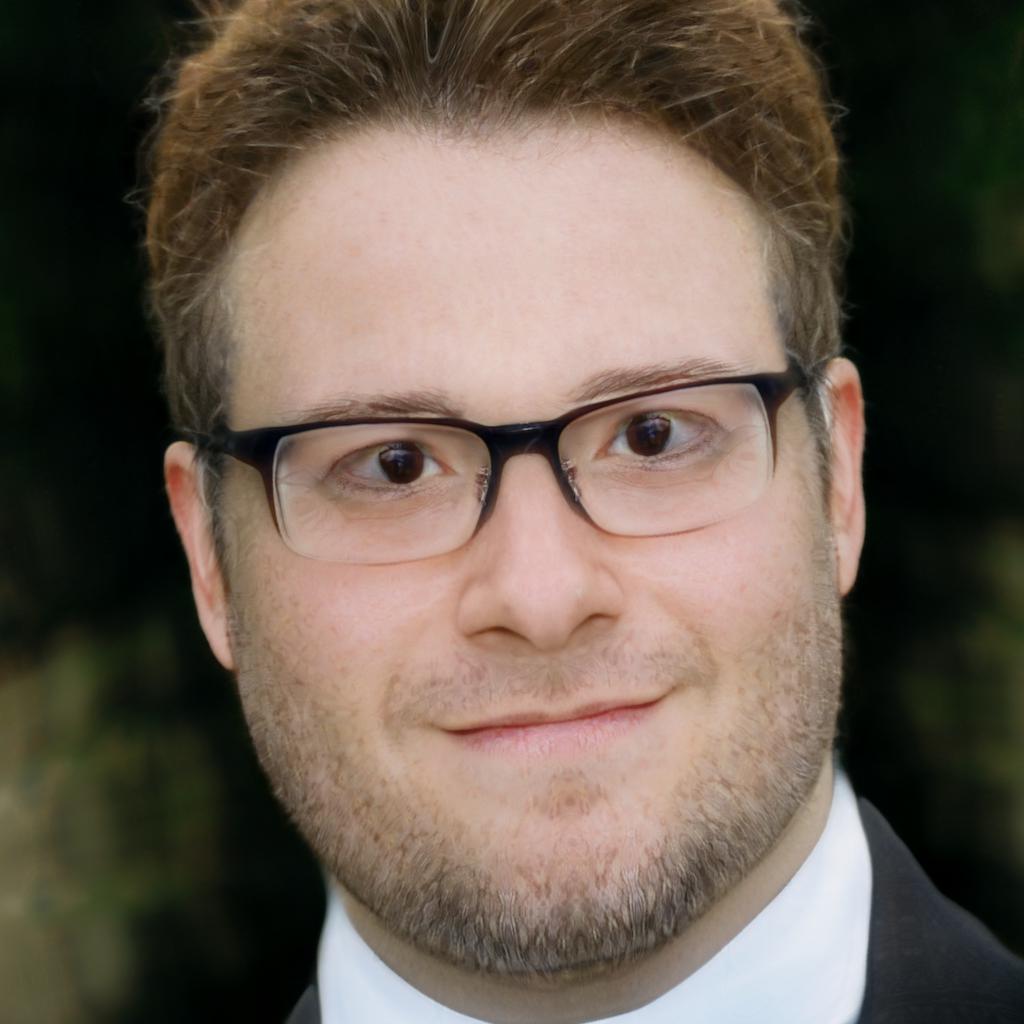} \\
        
        \includegraphics[width=0.27\columnwidth]{images/inversions/watson_unaligned.jpg} & 
        \includegraphics[width=0.27\columnwidth]{images/inversions/watson_restyle_e4e_sg2.jpg} & 
        \includegraphics[width=0.27\columnwidth]{images/inversions/watson_restyle_psp_sg2.jpg} & 
        \includegraphics[width=0.27\columnwidth]{images/inversions/watson_restyle_e4e_sg3.jpg} & 
        \includegraphics[width=0.27\columnwidth]{images/inversions/watson_restyle_psp_sg3.jpg} \\

		 Unaligned & SG2 $\text{ReStyle}_{e4e}$ & SG2 $\text{ReStyle}_{pSp}$ & SG3 $\text{ReStyle}_{e4e}$ & SG3 $\text{ReStyle}_{pSp}$
	\end{tabular}
	}
	\vspace{-0.1cm}
	\caption{
	Additional reconstruction quality comparisons between encoders trained for inverting StyleGAN2 and StyleGAN3 generators. When given unaligned source images, our StyleGAN3 encoders are able to achieve comparable reconstruction quality to their StyleGAN2 counterparts. 
	}
	\label{fig:inversions_supplementary_2}
\end{figure*}

%% file: figures/supplementary/edits_reals_comparison.tex
\begin{figure*}[tb]
	\centering
	\setlength{\tabcolsep}{1pt}	
	{\footnotesize
	\begin{tabular}{c c c c c c c}

		\raisebox{0.2in}{\rotatebox{90}{$+$ Age}} &
        \includegraphics[width=0.24\columnwidth]{images/editing/reals/scarlett_johanson_unaligned.jpg} & 
        \includegraphics[width=0.24\columnwidth]{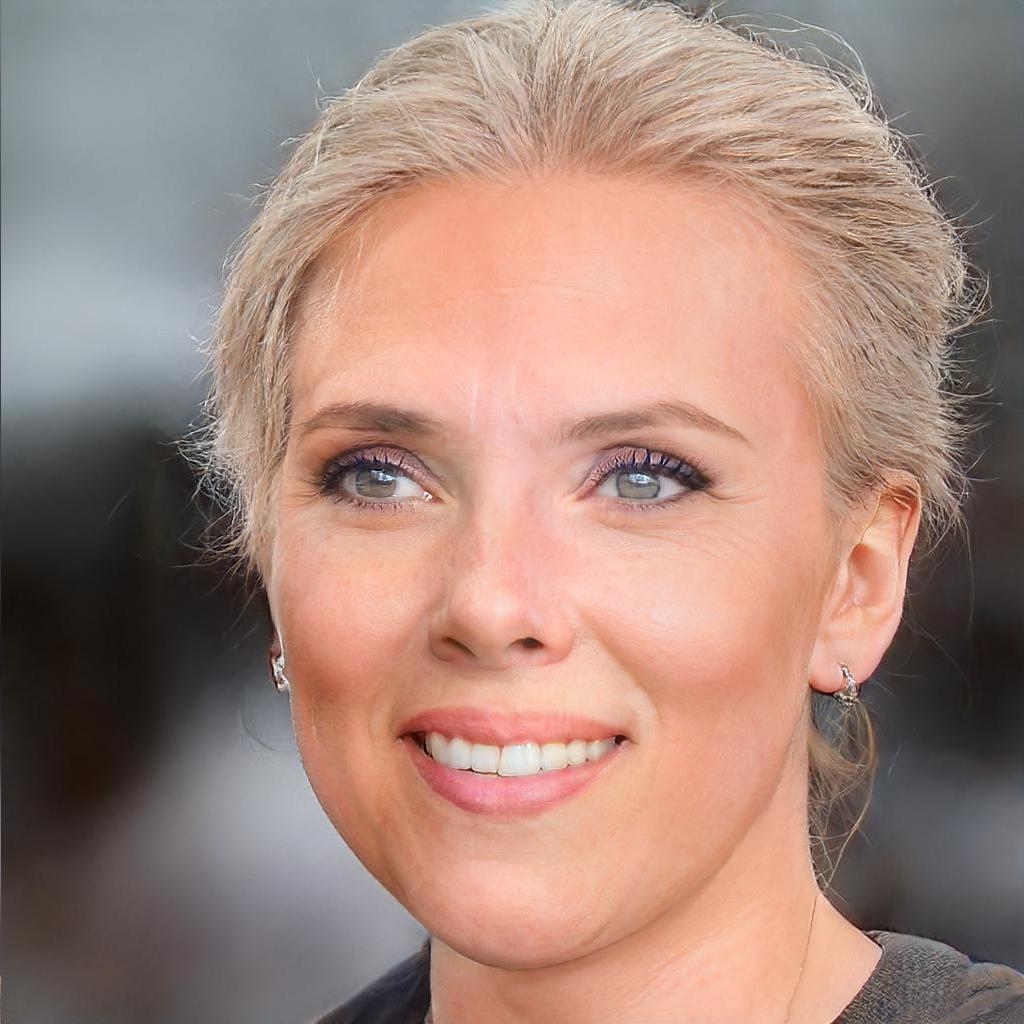} & 
        \includegraphics[width=0.24\columnwidth]{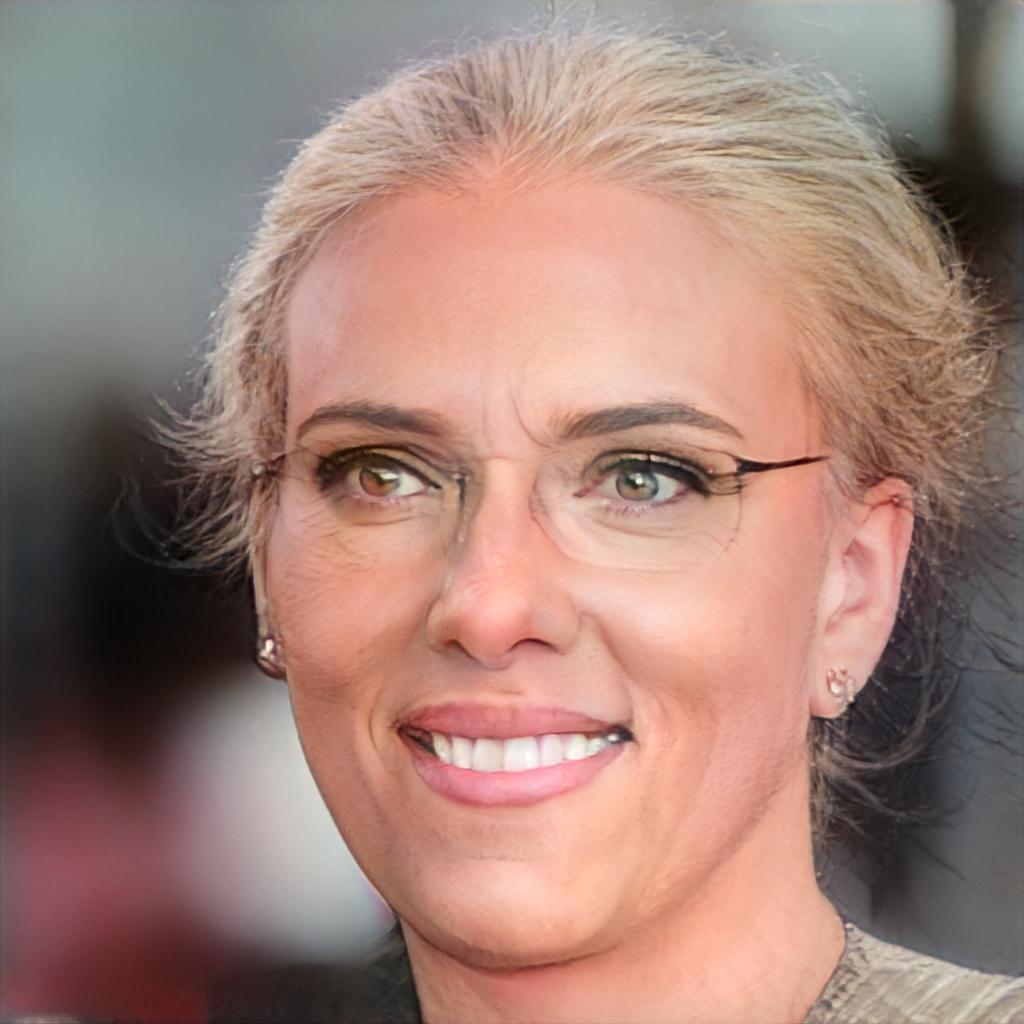} & 
        \includegraphics[width=0.24\columnwidth]{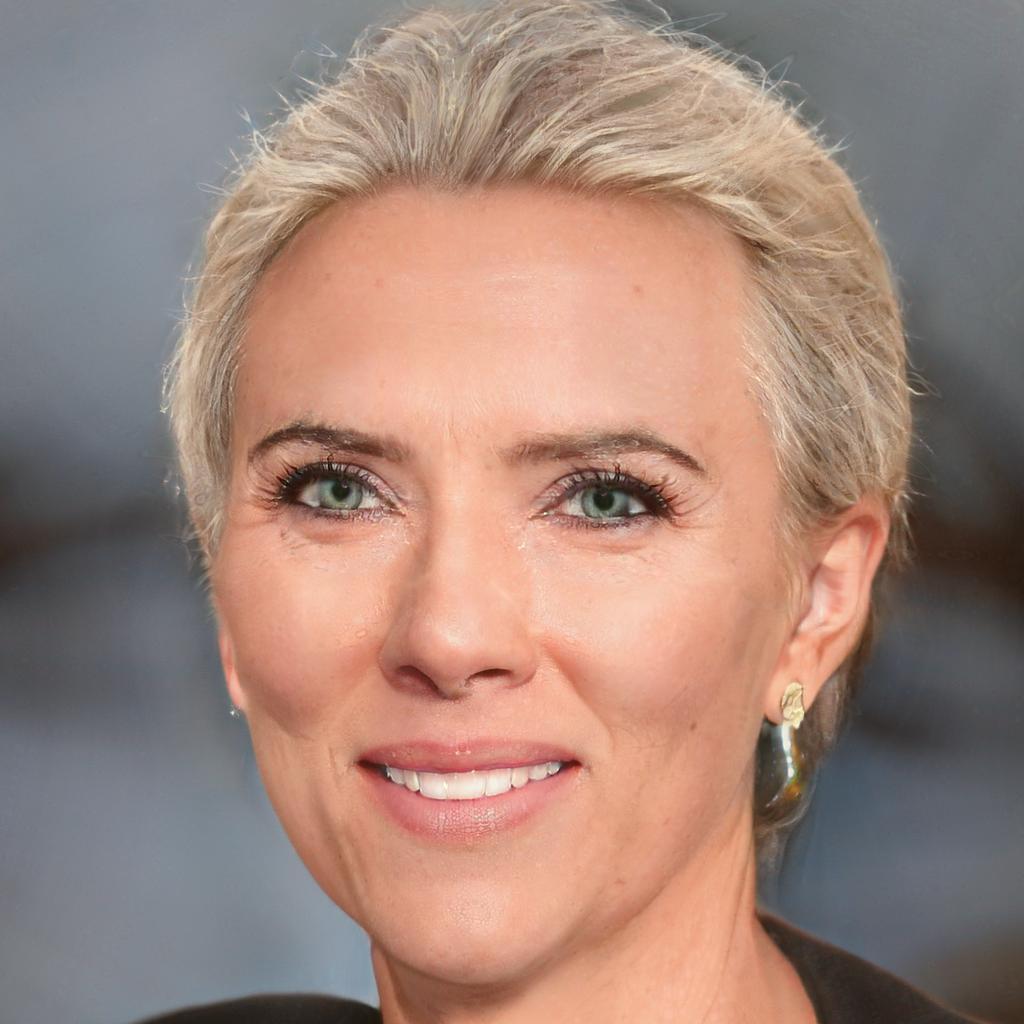} & 
        \includegraphics[width=0.24\columnwidth]{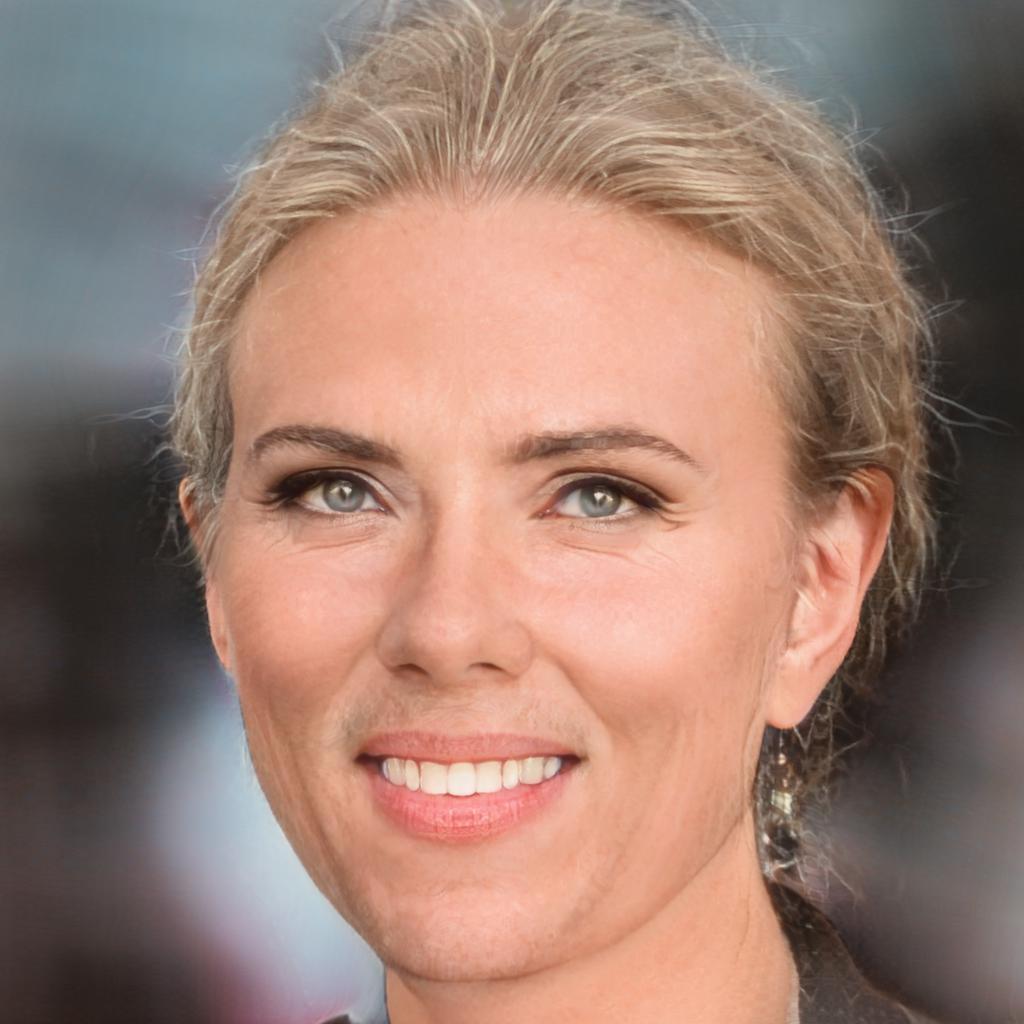} \\

		\raisebox{0.2in}{\rotatebox{90}{$-$ Age}} &
        \includegraphics[width=0.24\columnwidth]{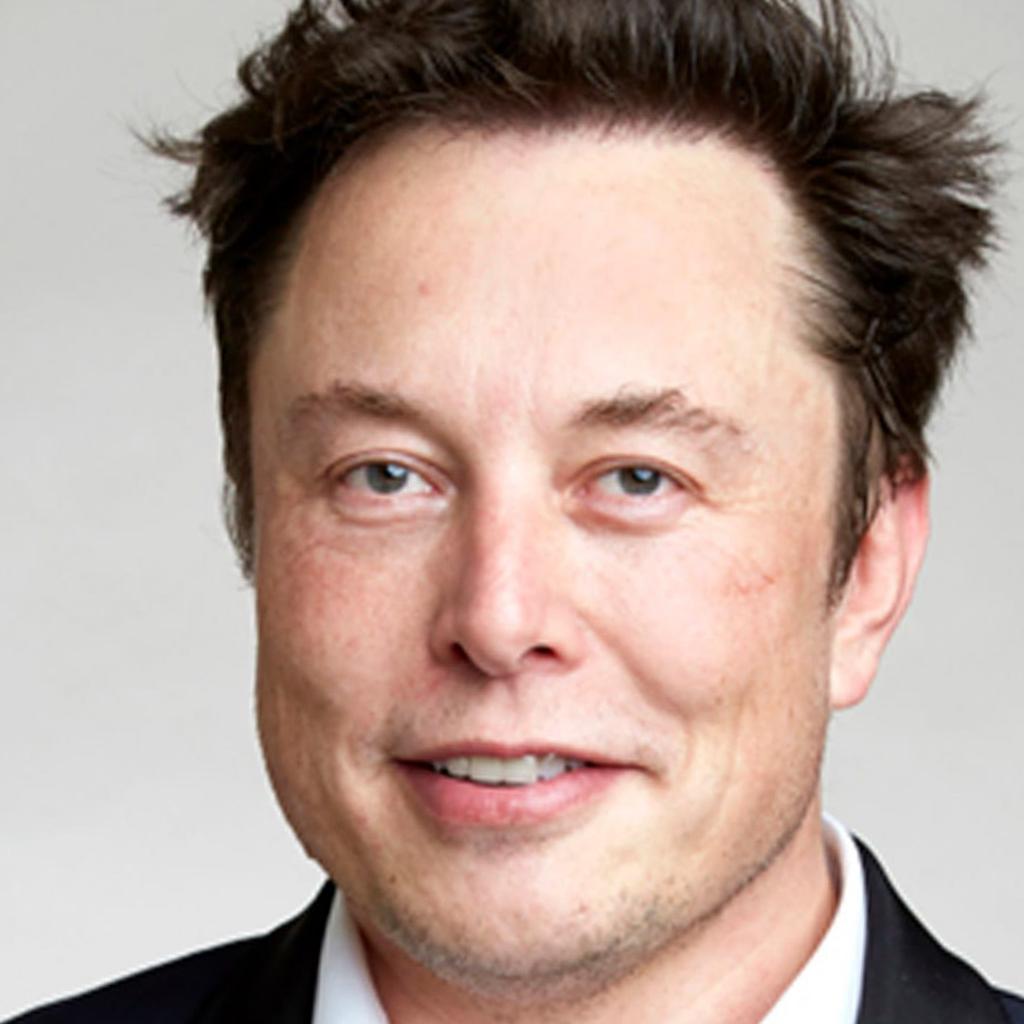} & 
        \includegraphics[width=0.24\columnwidth]{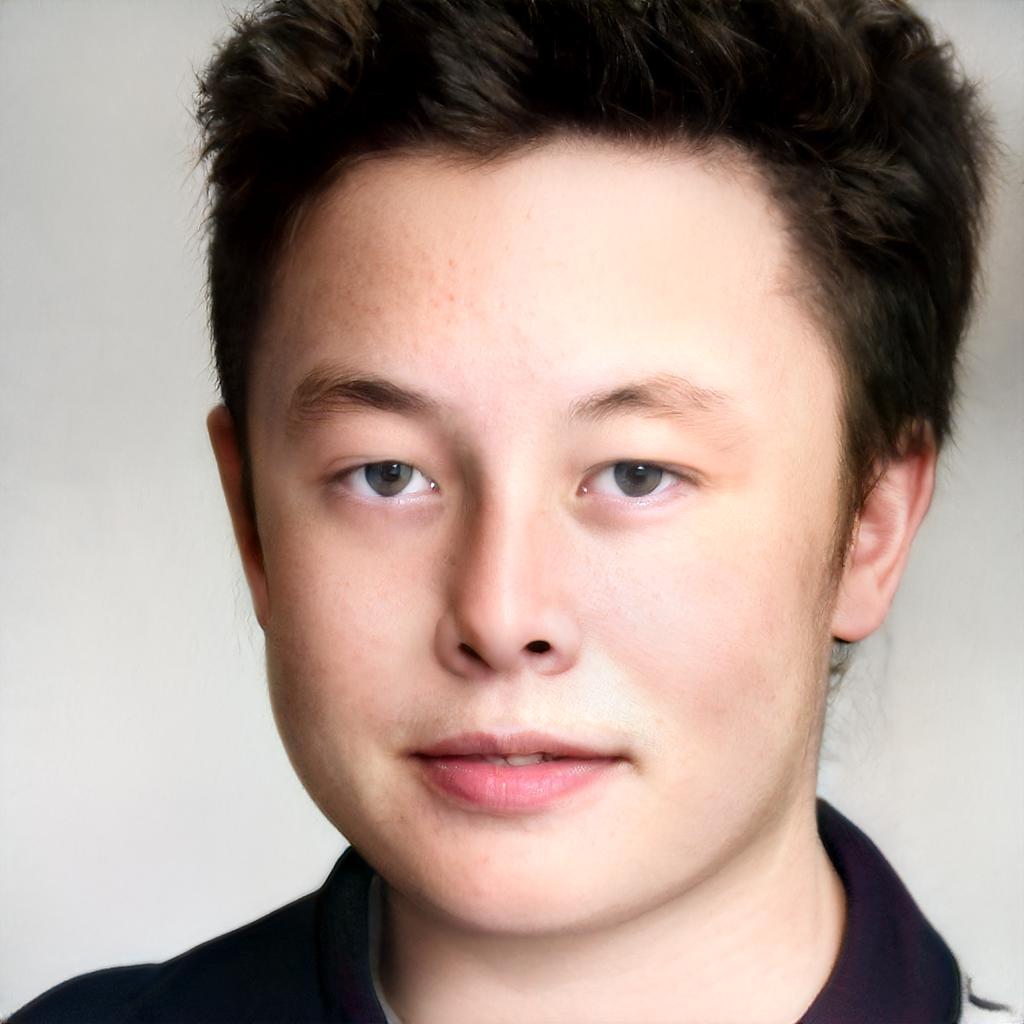} & 
        \includegraphics[width=0.24\columnwidth]{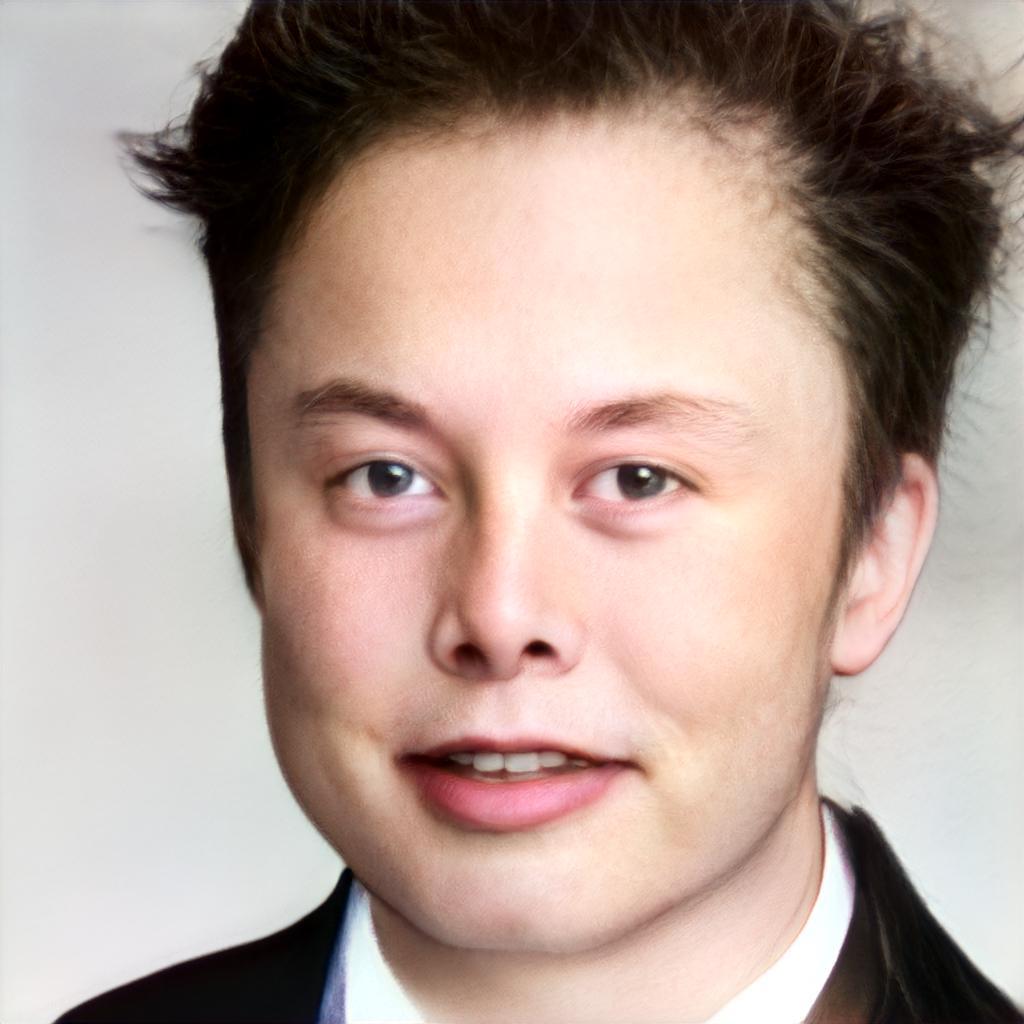} & 
        \includegraphics[width=0.24\columnwidth]{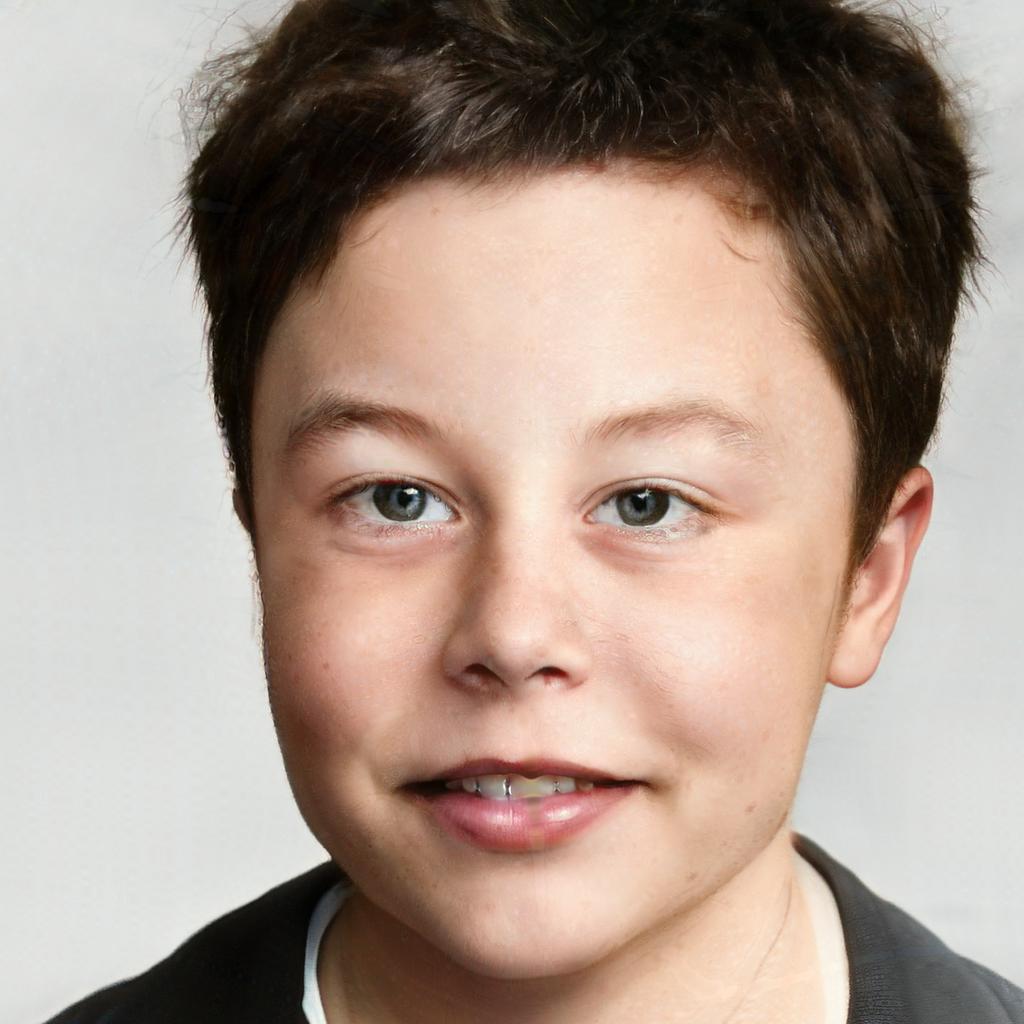} & 
        \includegraphics[width=0.24\columnwidth]{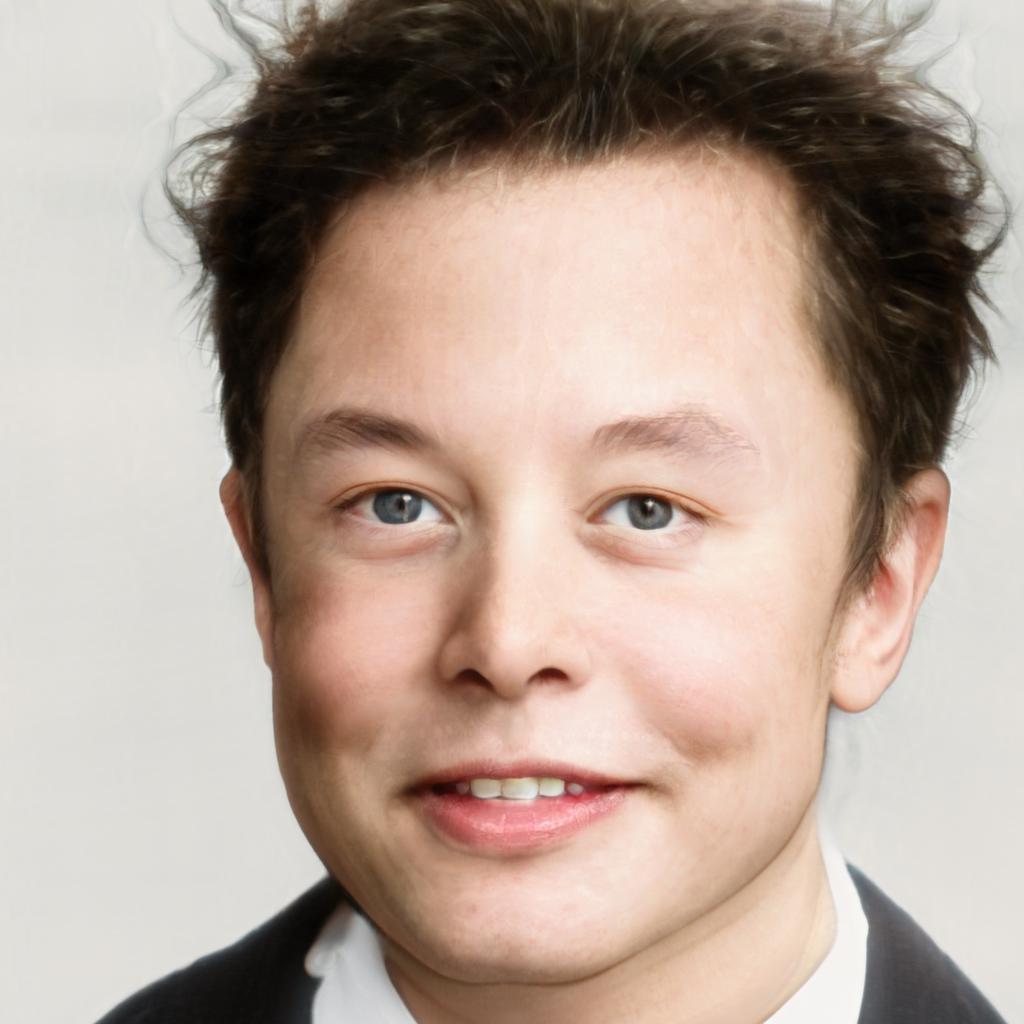} \\

		\raisebox{0.2in}{\rotatebox{90}{$+$ Smile}} &
        \includegraphics[width=0.24\columnwidth]{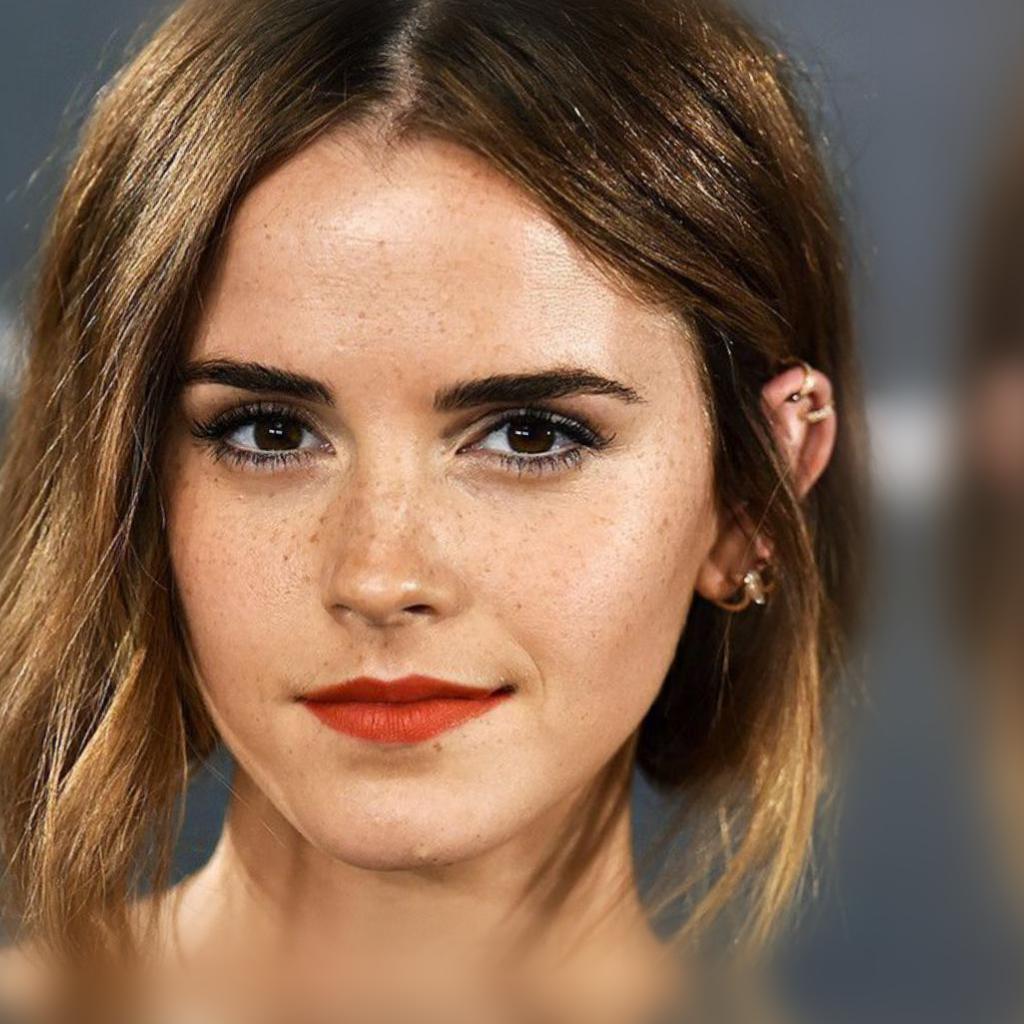} & 
        \includegraphics[width=0.24\columnwidth]{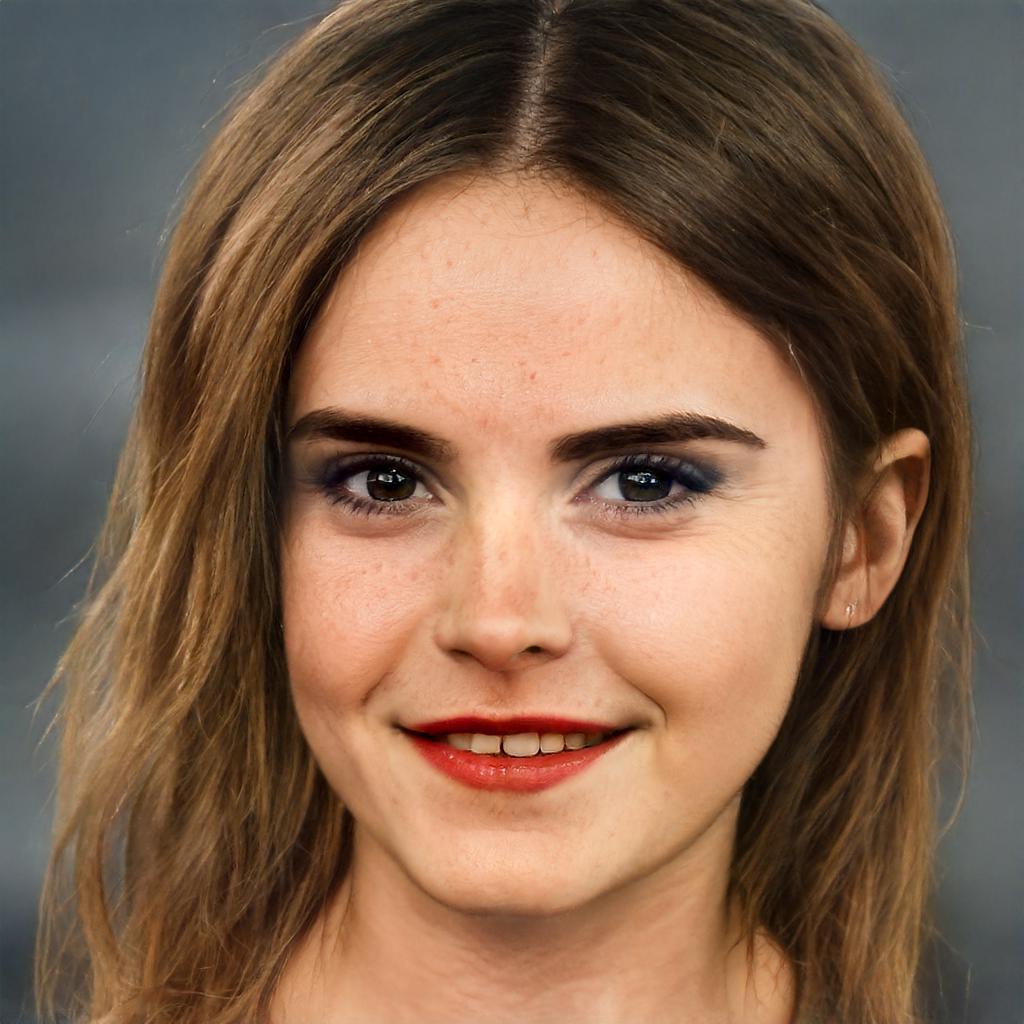} & 
        \includegraphics[width=0.24\columnwidth]{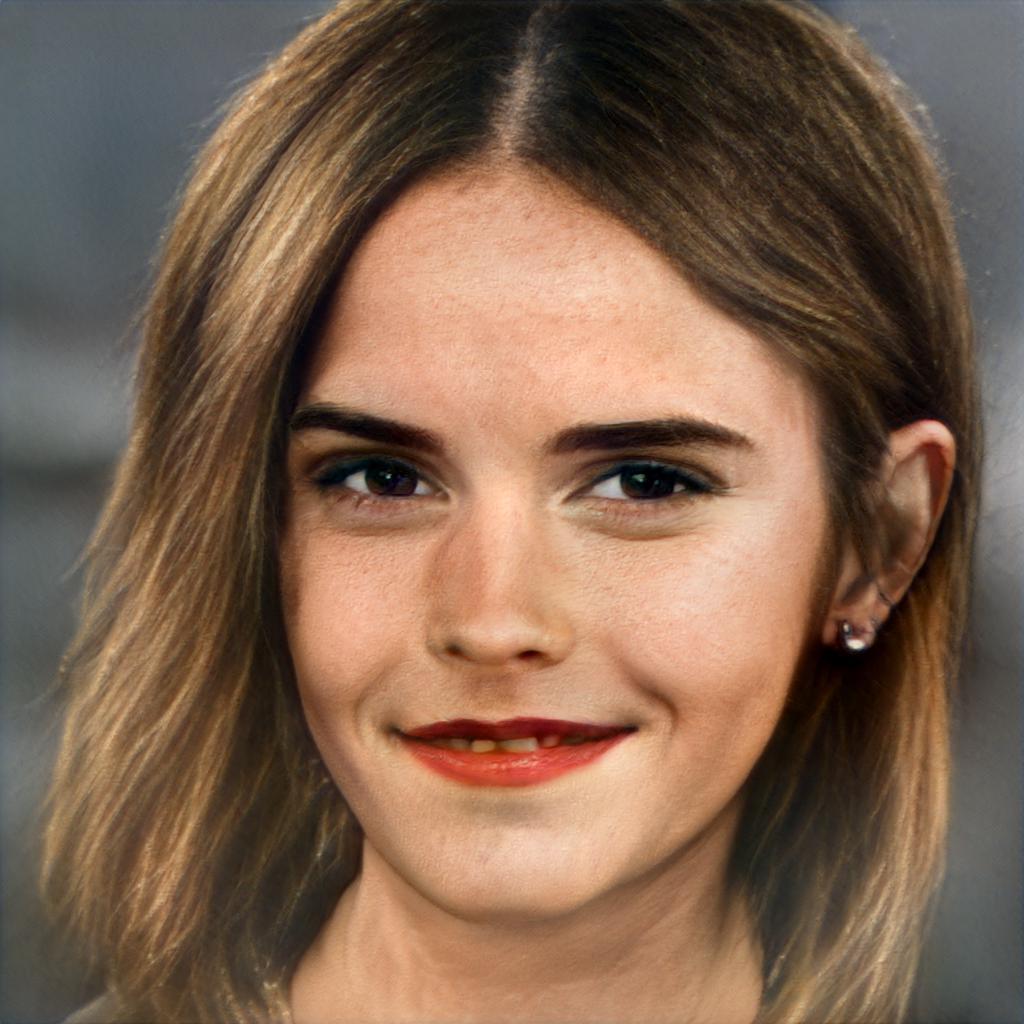} & 
        \includegraphics[width=0.24\columnwidth]{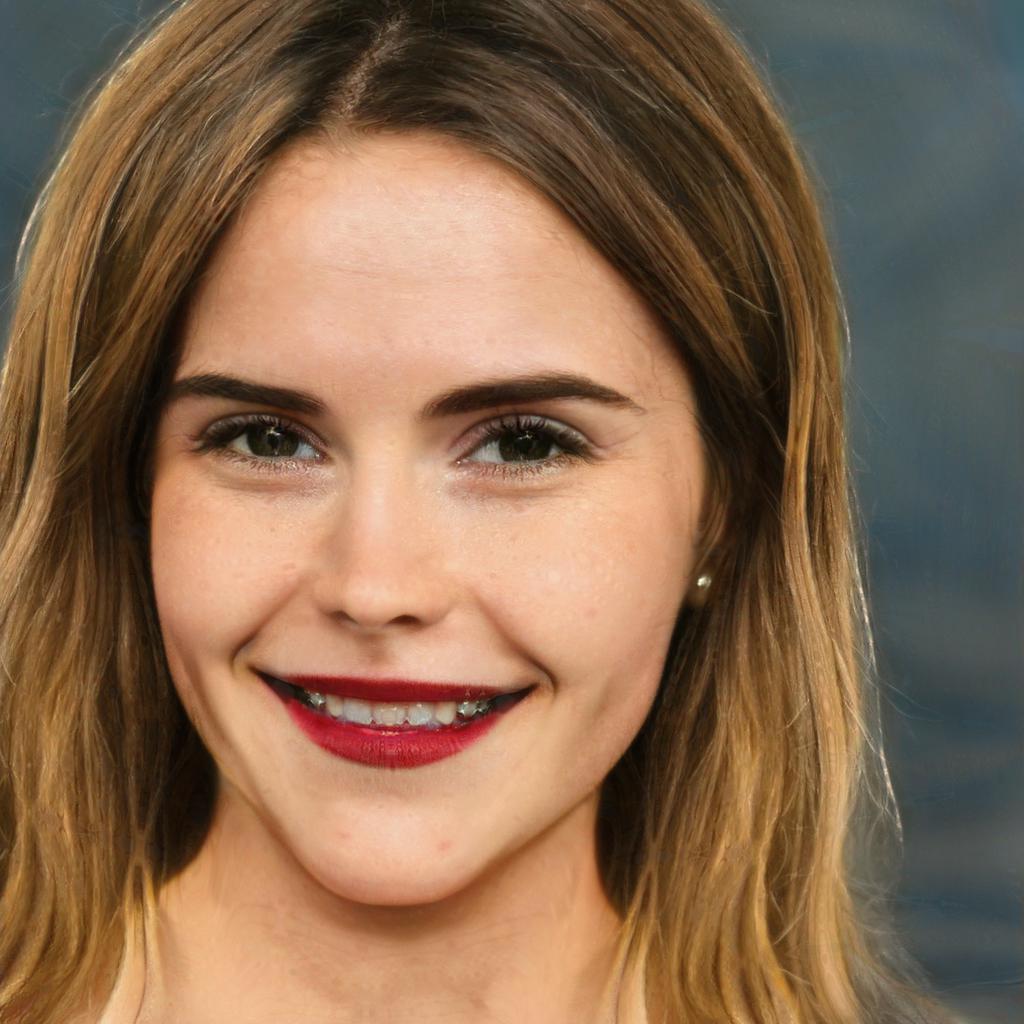} & 
        \includegraphics[width=0.24\columnwidth]{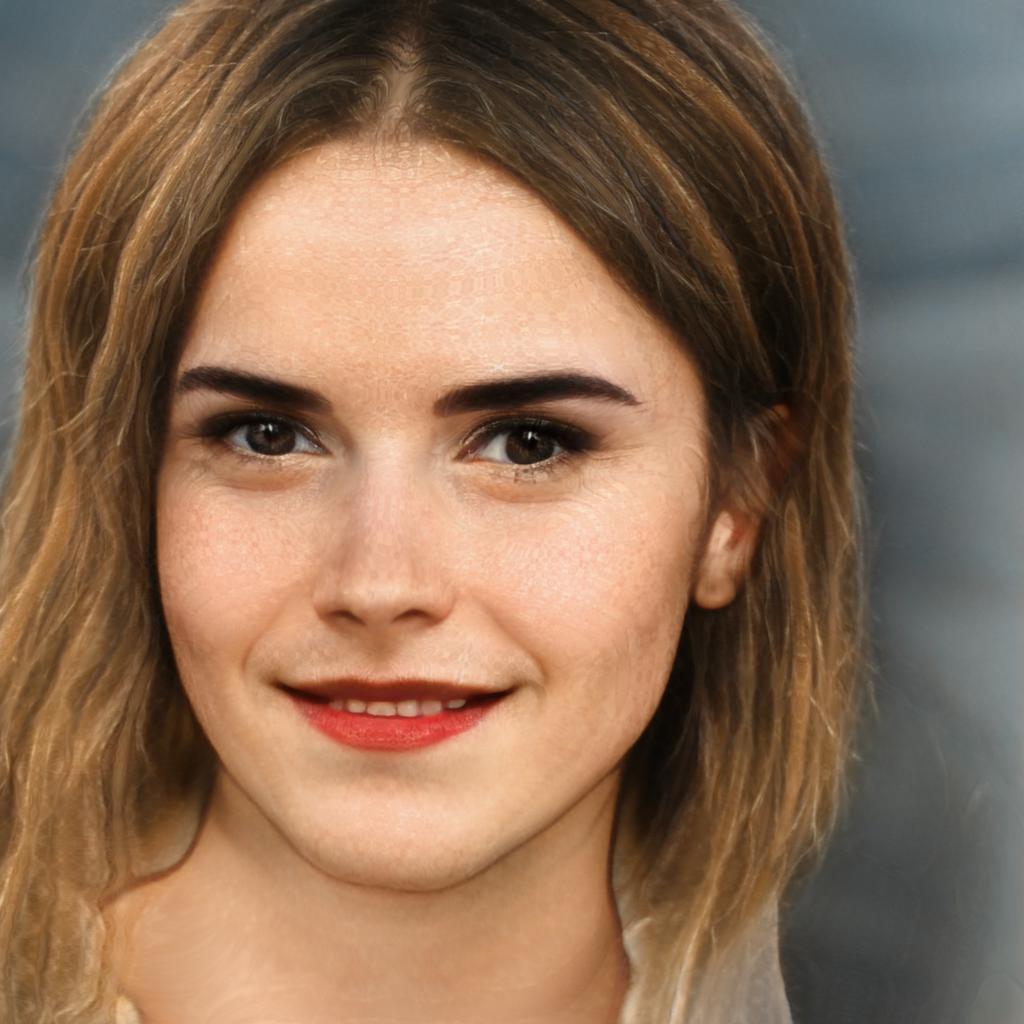} \\
        
		\raisebox{0.2in}{\rotatebox{90}{$+$ Smile}} &
        \includegraphics[width=0.24\columnwidth]{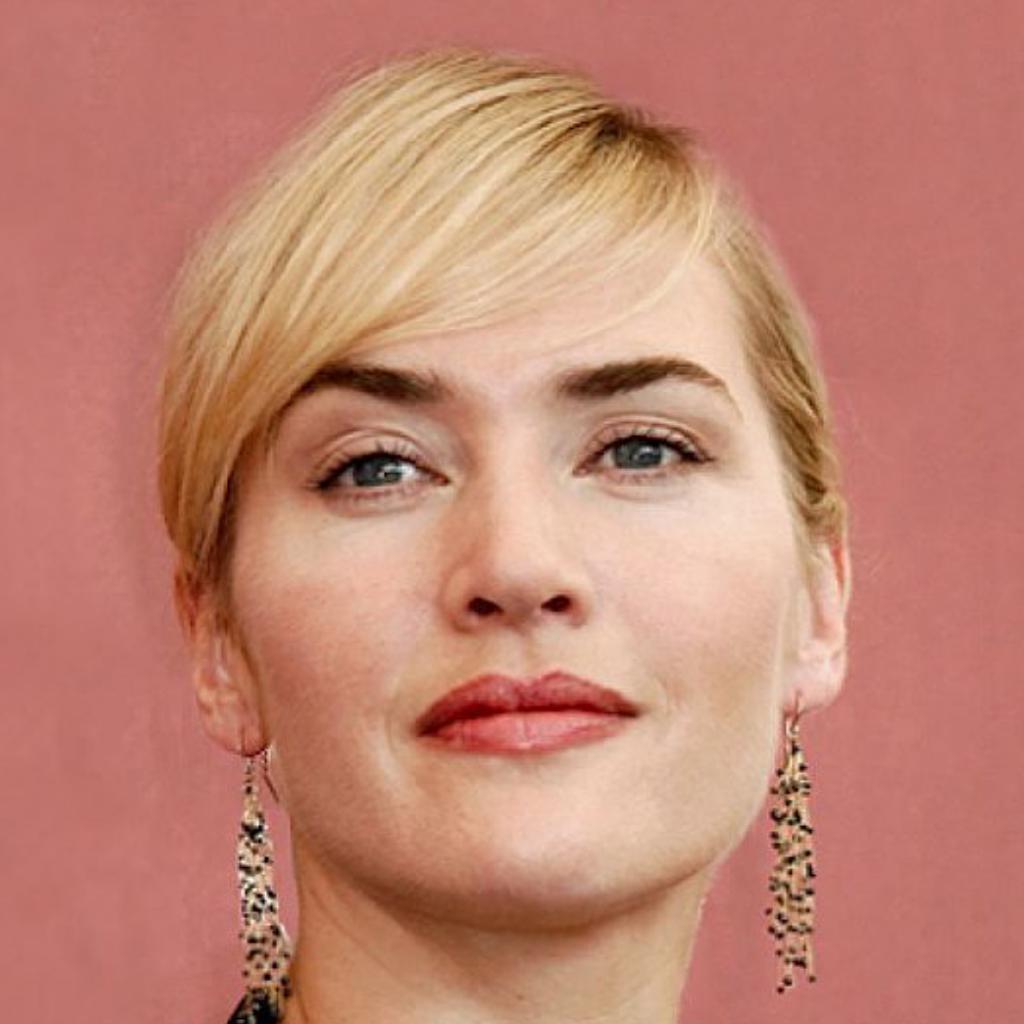} & 
        \includegraphics[width=0.24\columnwidth]{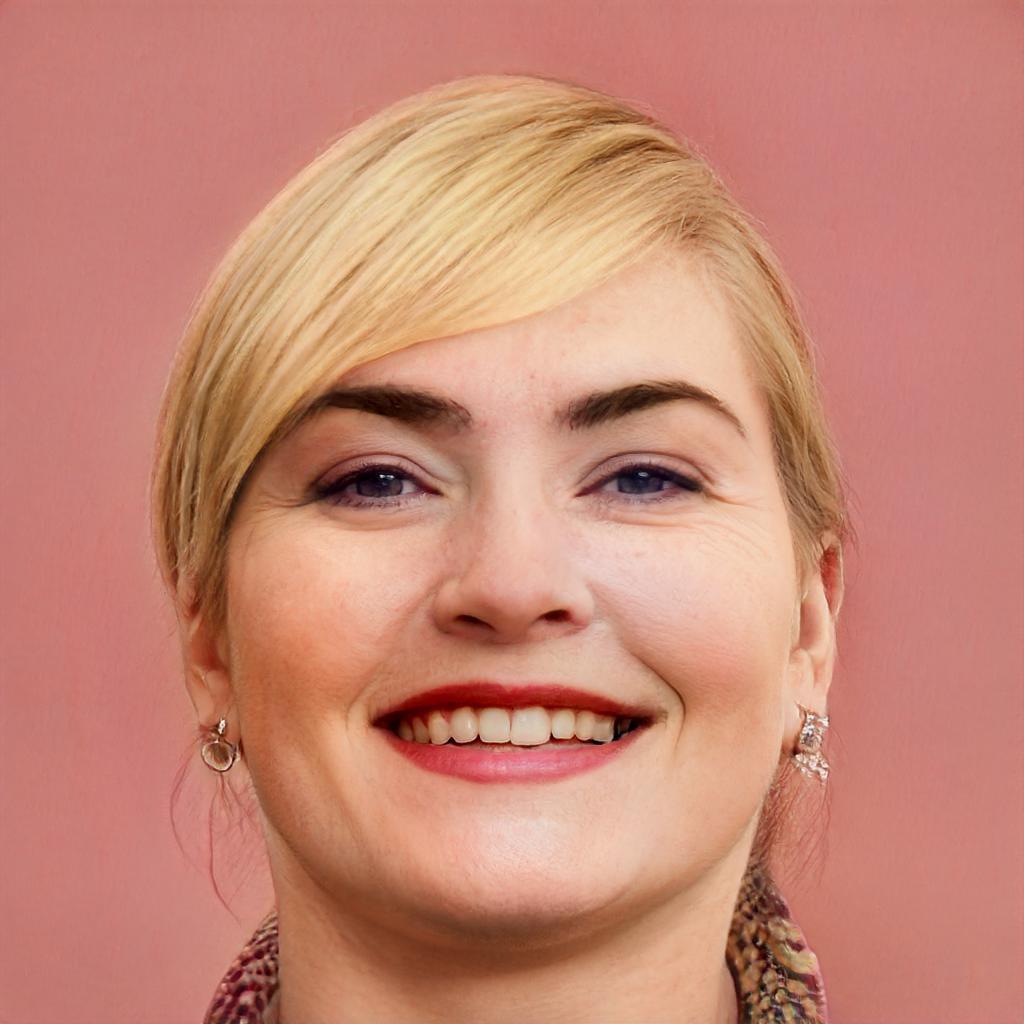} & 
        \includegraphics[width=0.24\columnwidth]{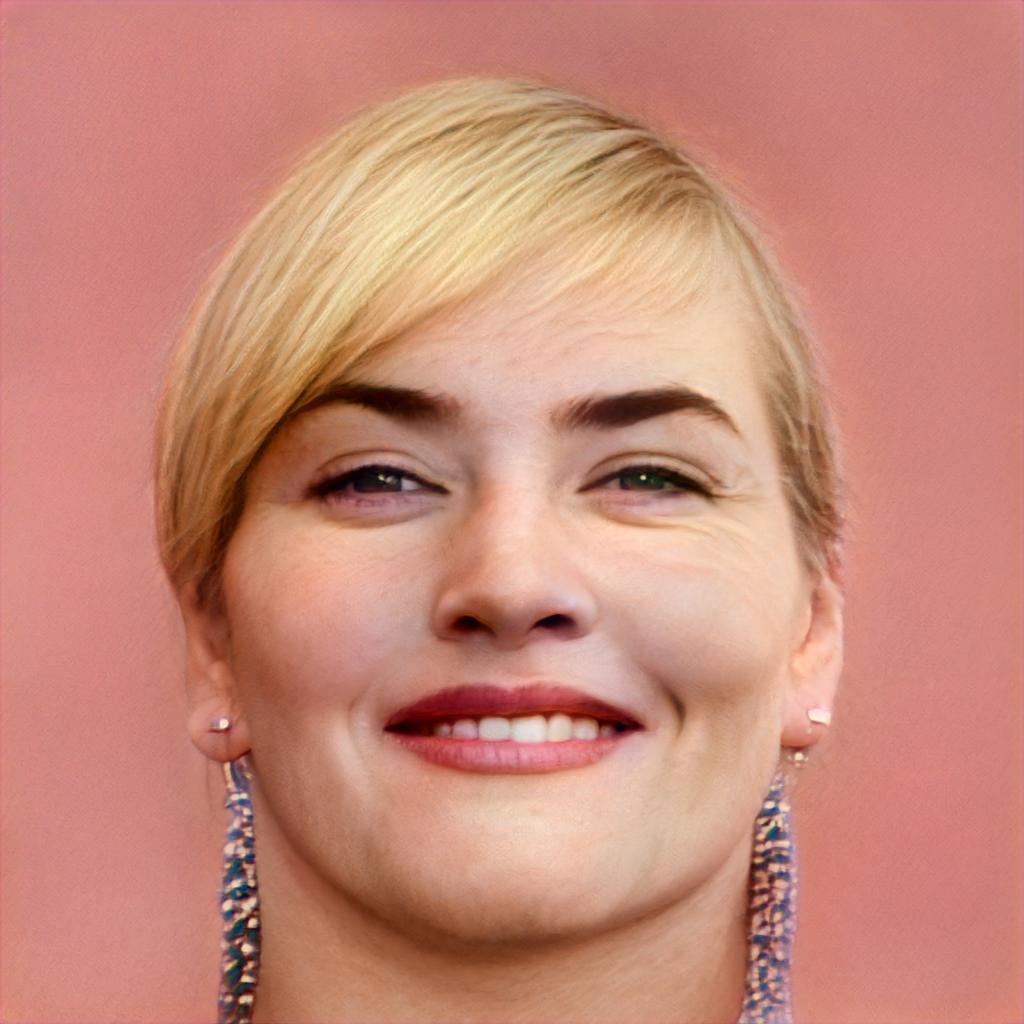} & 
        \includegraphics[width=0.24\columnwidth]{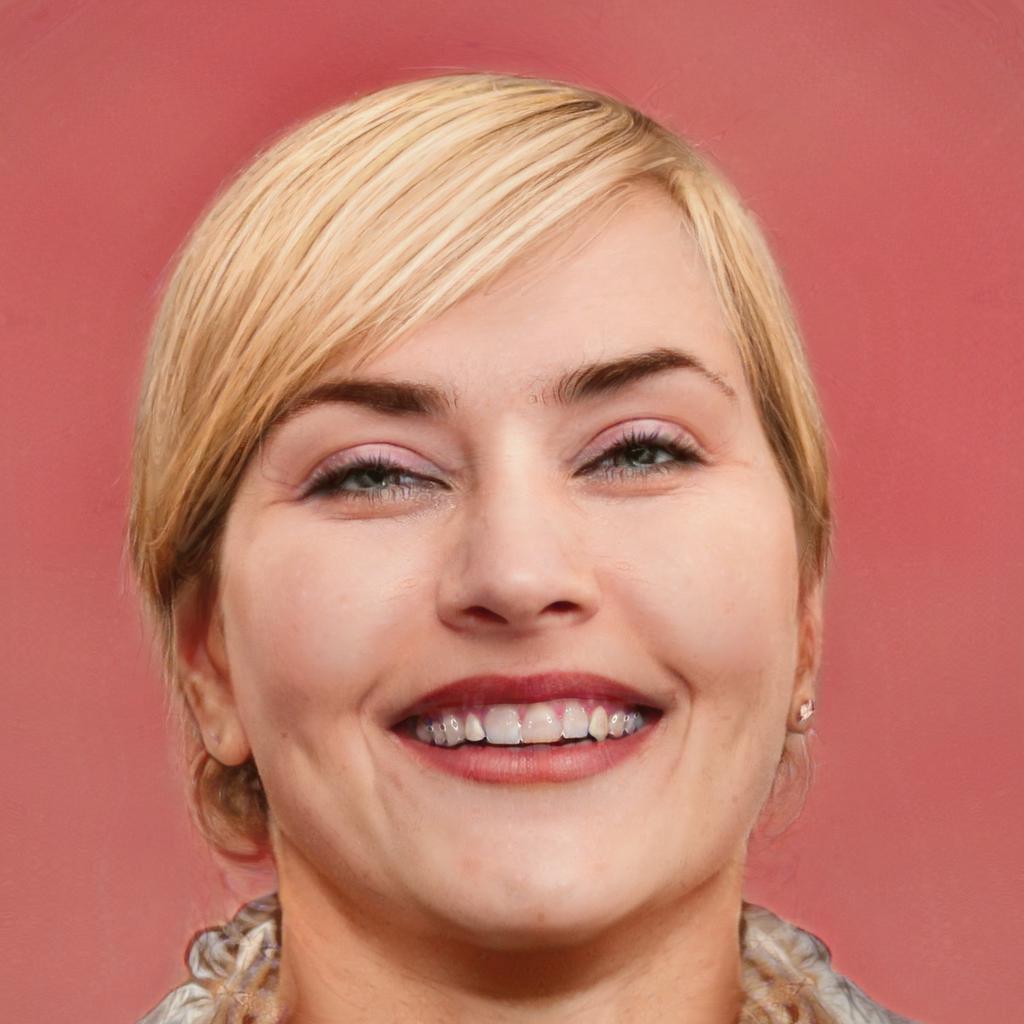} & 
        \includegraphics[width=0.24\columnwidth]{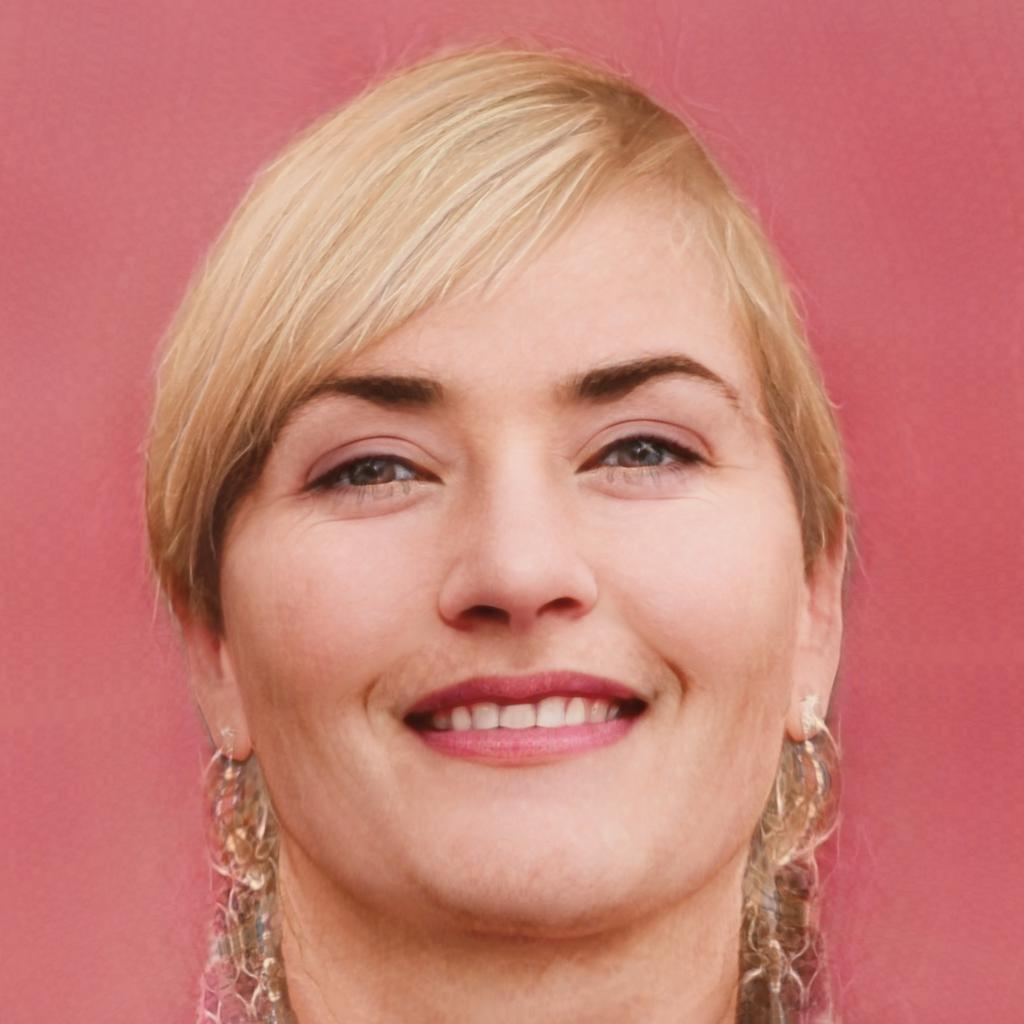} \\
        
		\raisebox{0.2in}{\rotatebox{90}{$+$ Pose}} &
        \includegraphics[width=0.24\columnwidth]{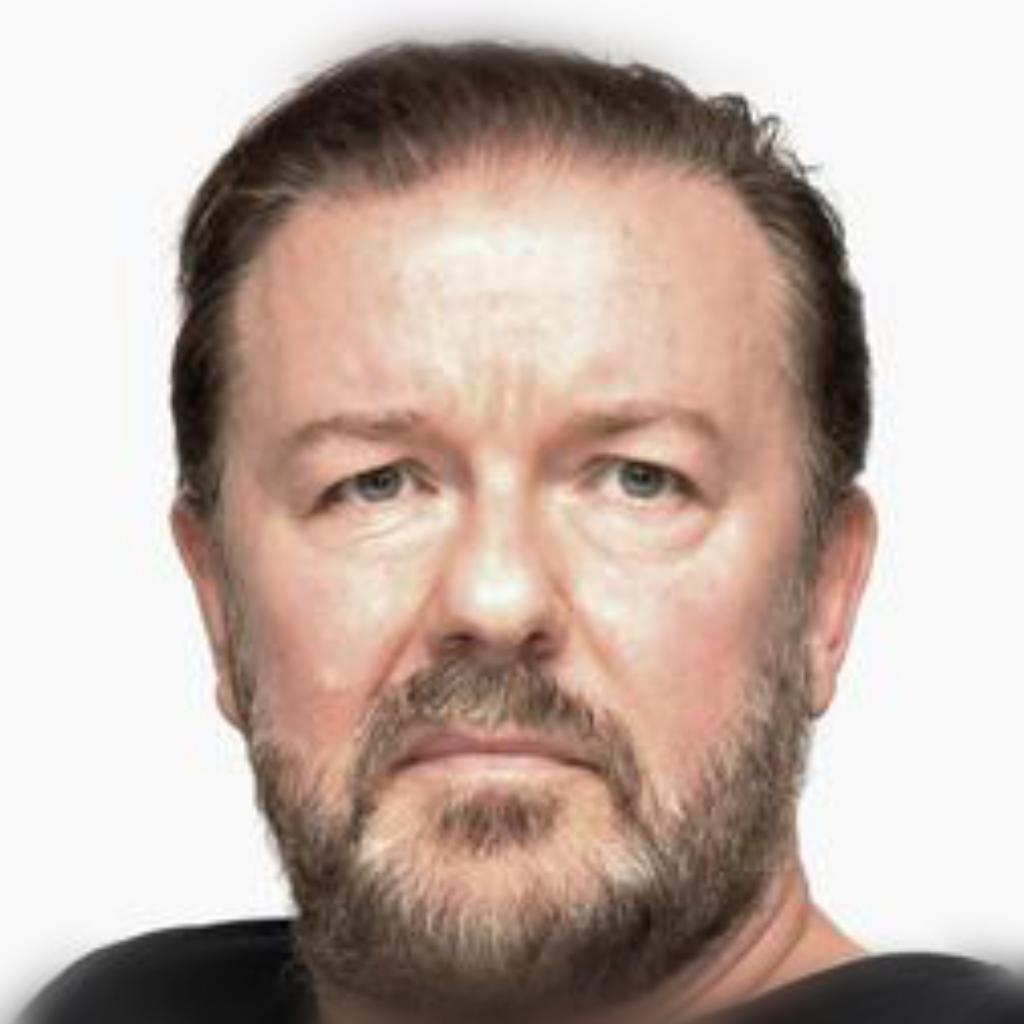} & 
        \includegraphics[width=0.24\columnwidth]{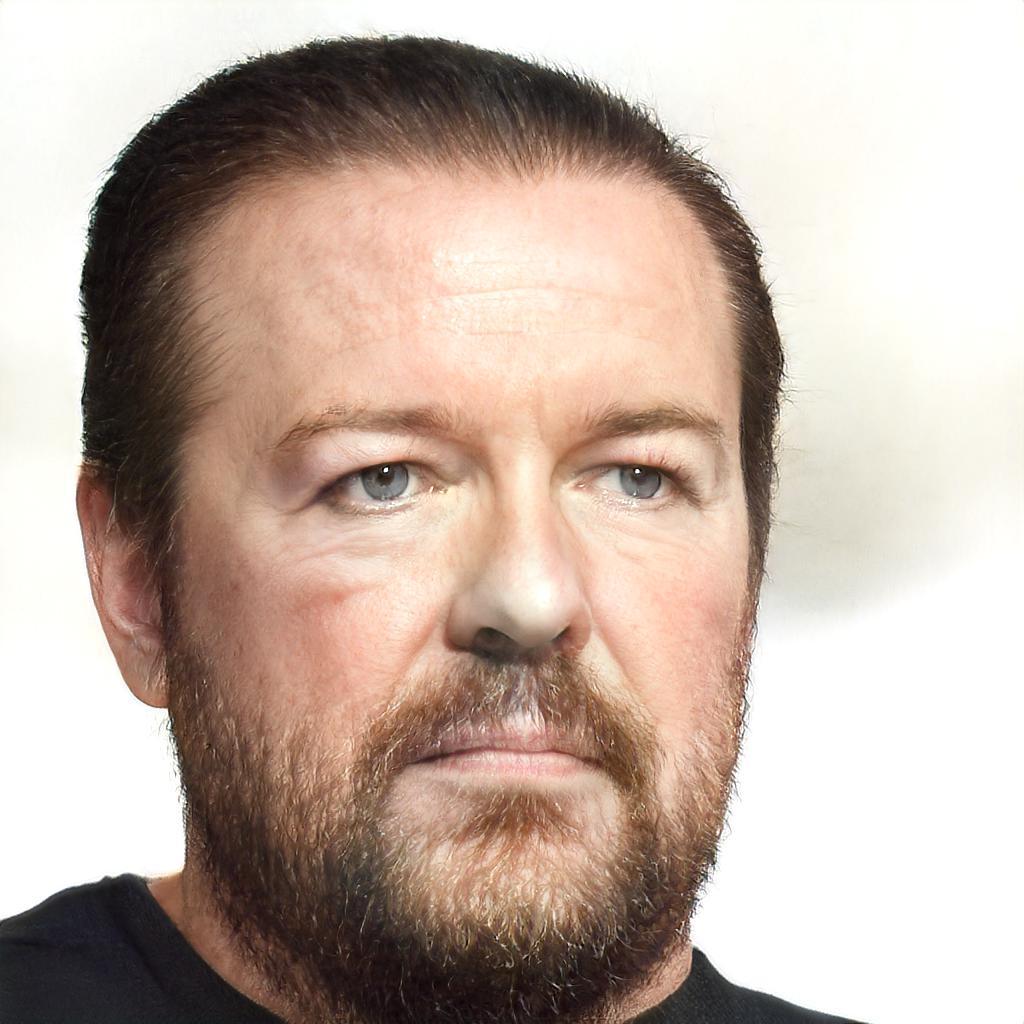} & 
        \includegraphics[width=0.24\columnwidth]{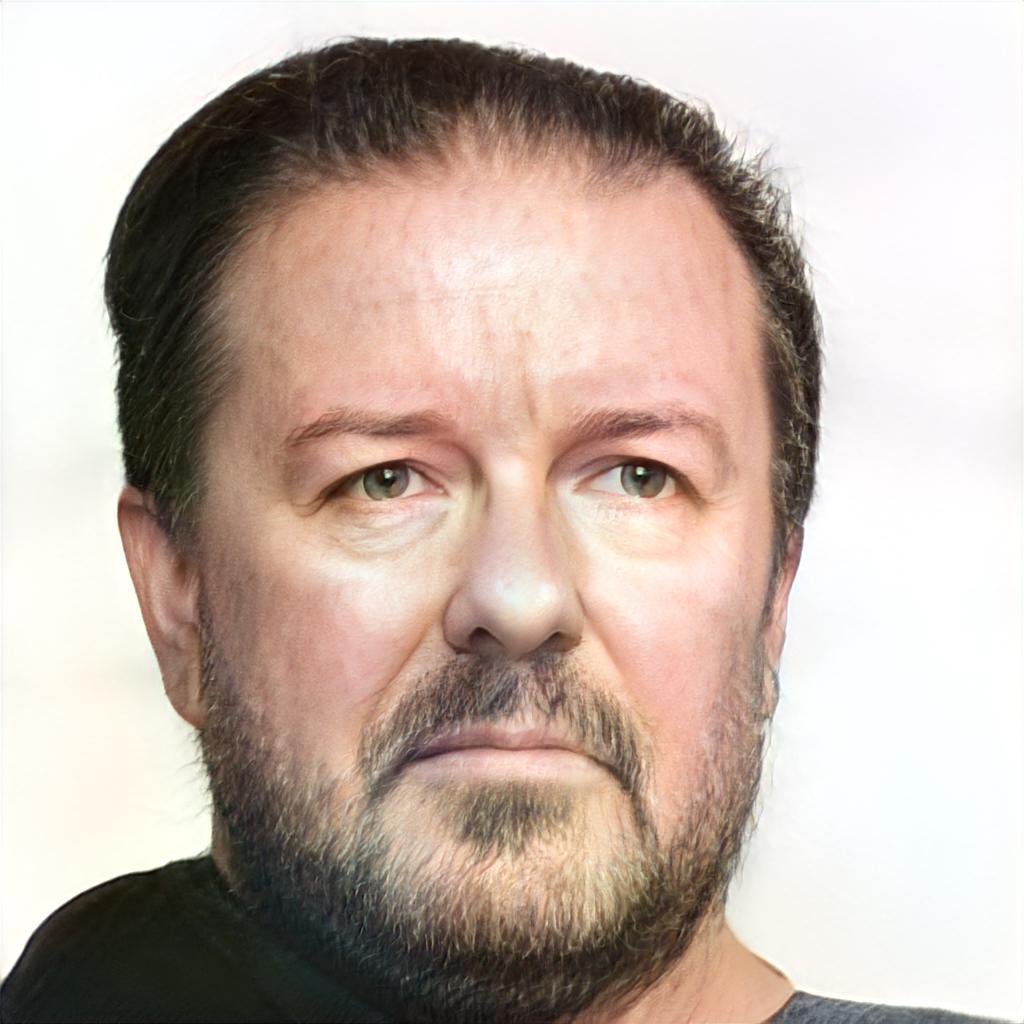} & 
        \includegraphics[width=0.24\columnwidth]{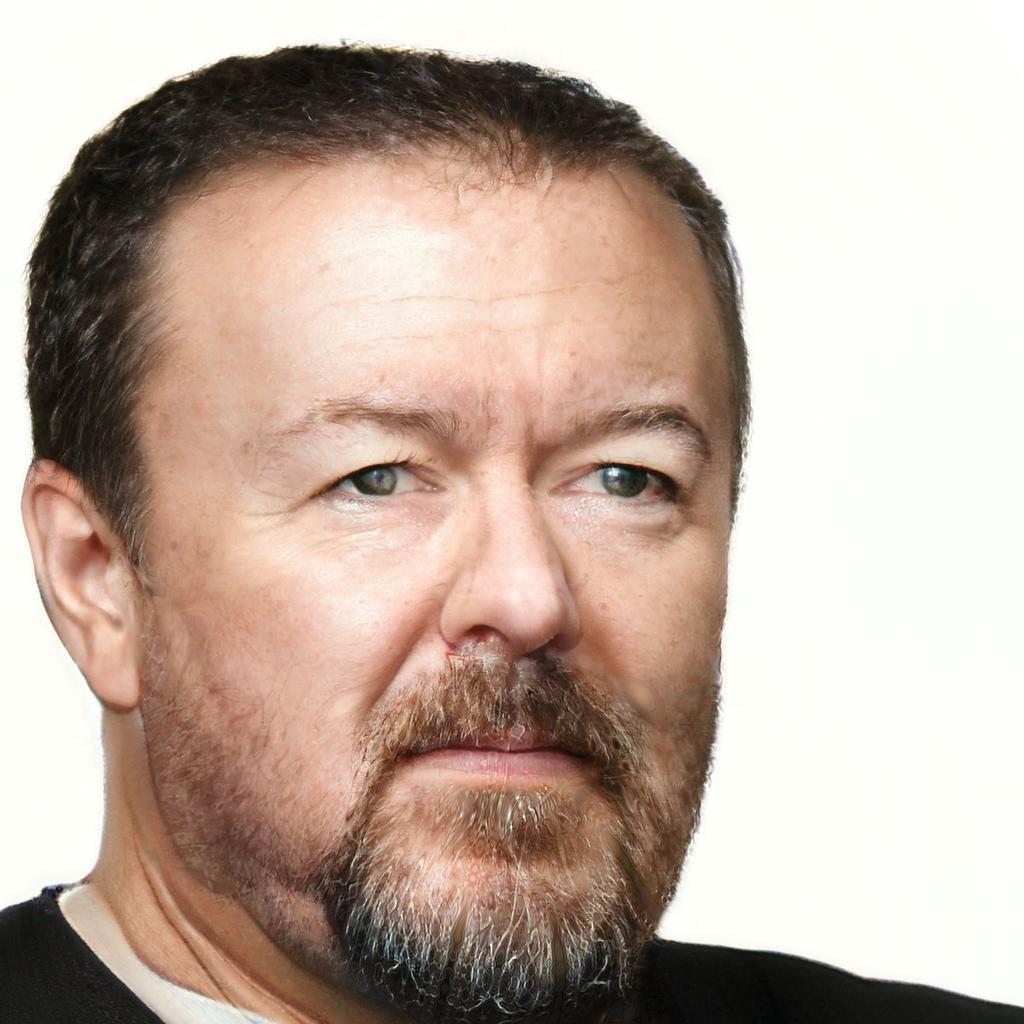} & 
        \includegraphics[width=0.24\columnwidth]{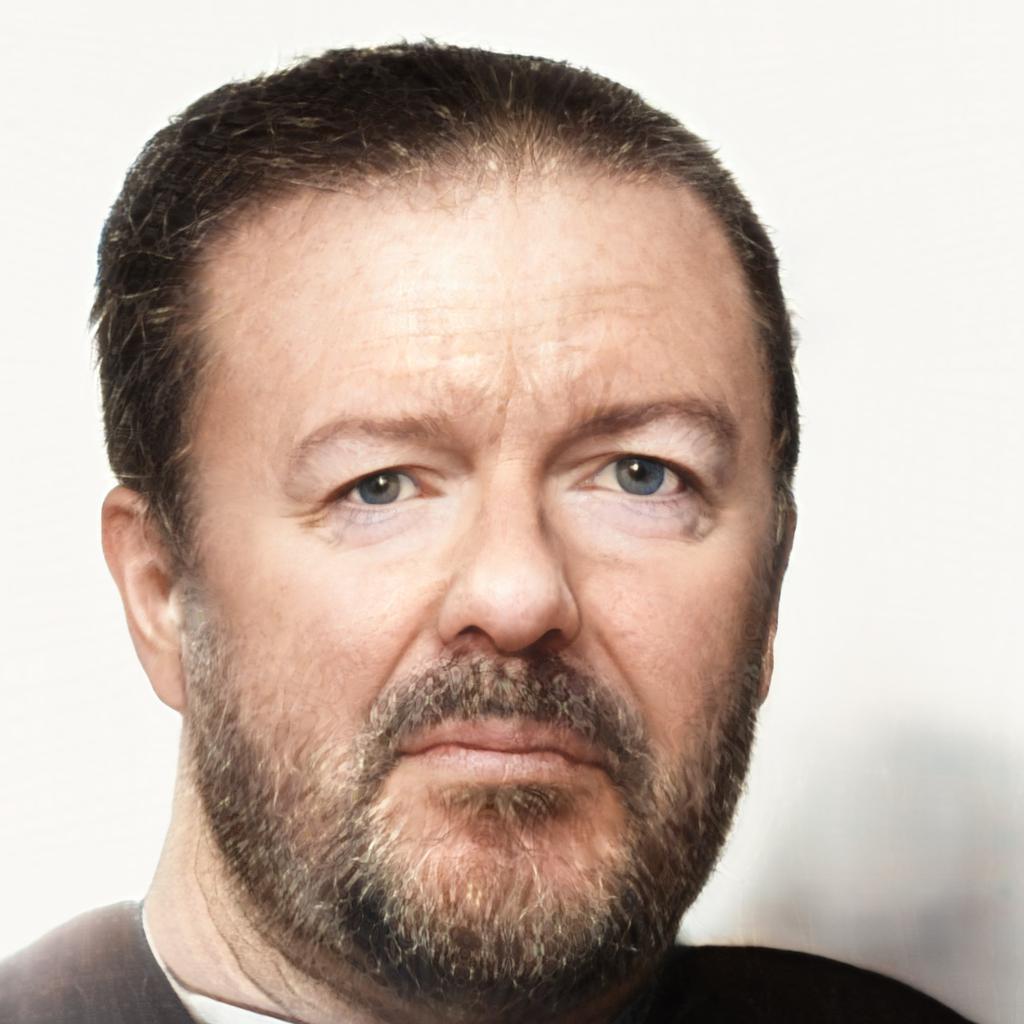} \\

		\raisebox{0.2in}{\rotatebox{90}{$-$ Pose}} &
        \includegraphics[width=0.24\columnwidth]{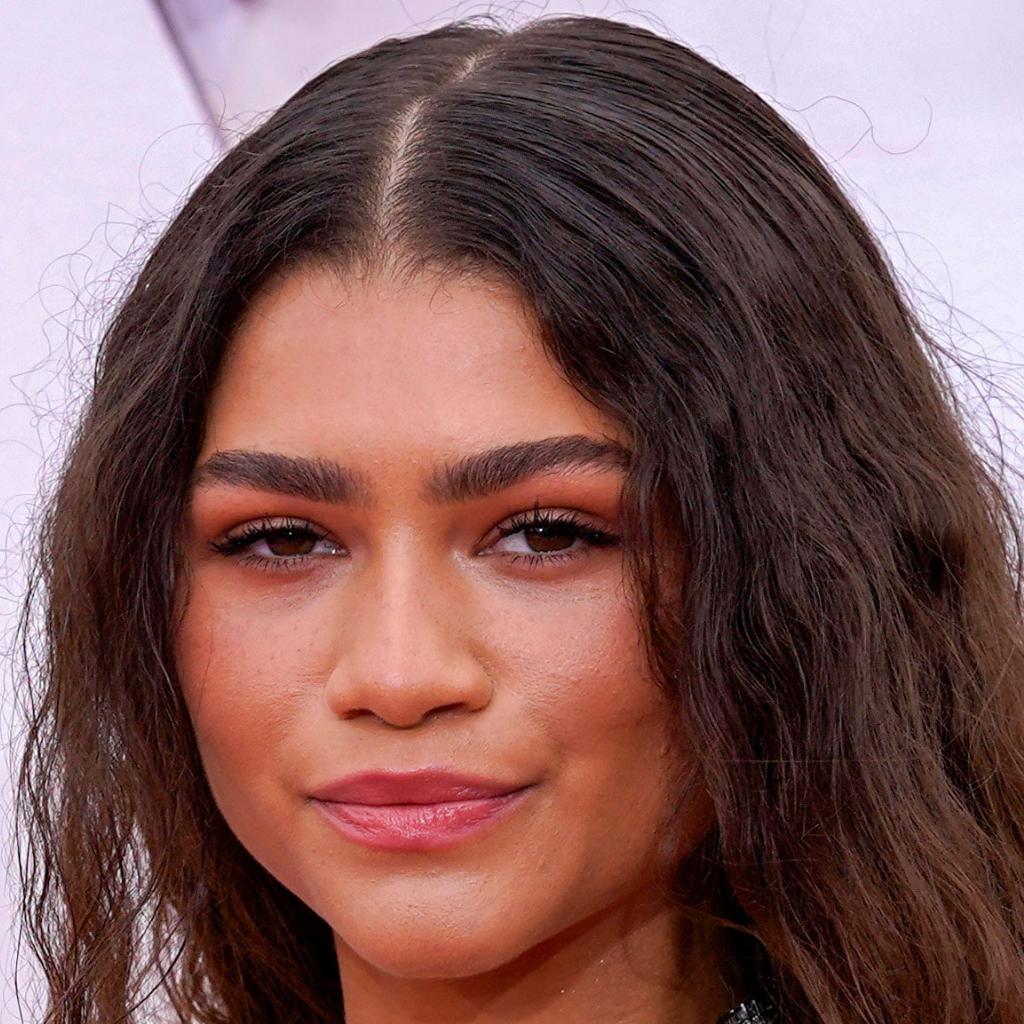} & 
        \includegraphics[width=0.24\columnwidth]{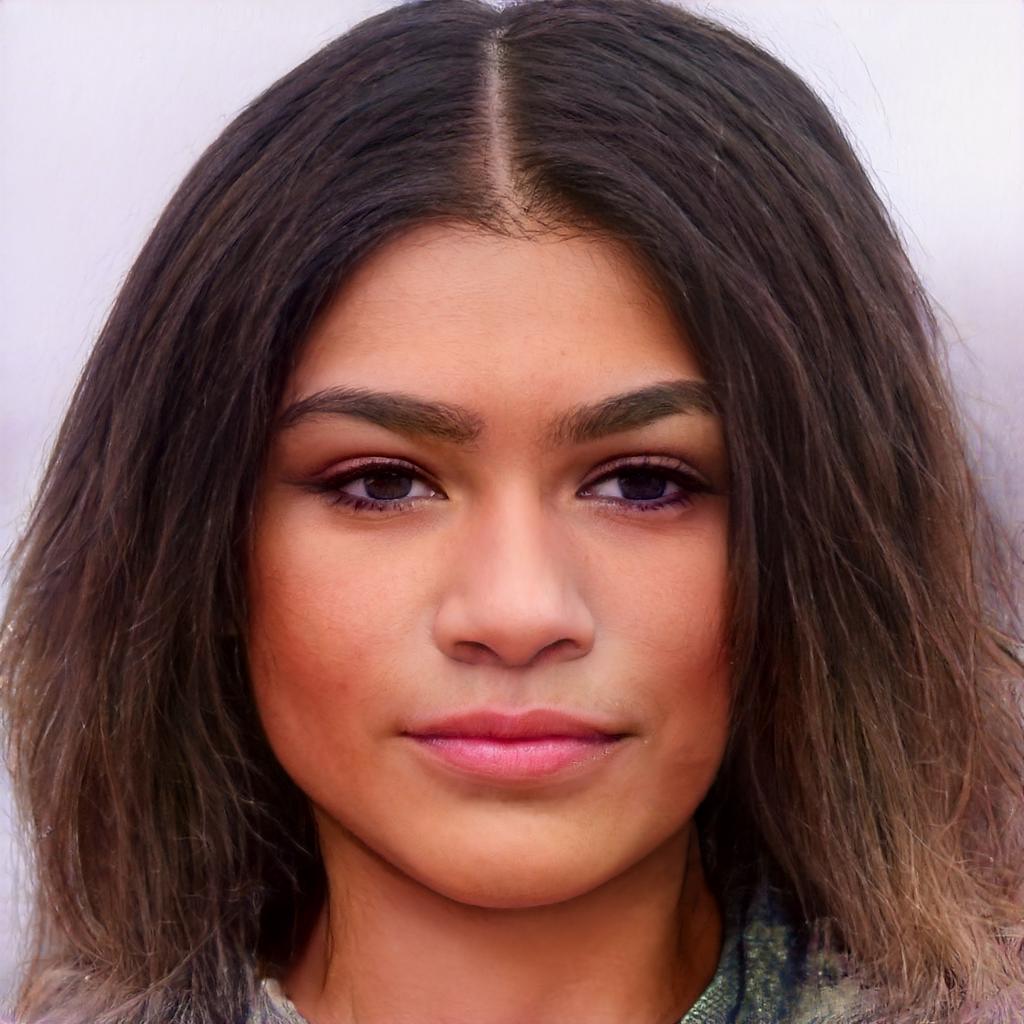} & 
        \includegraphics[width=0.24\columnwidth]{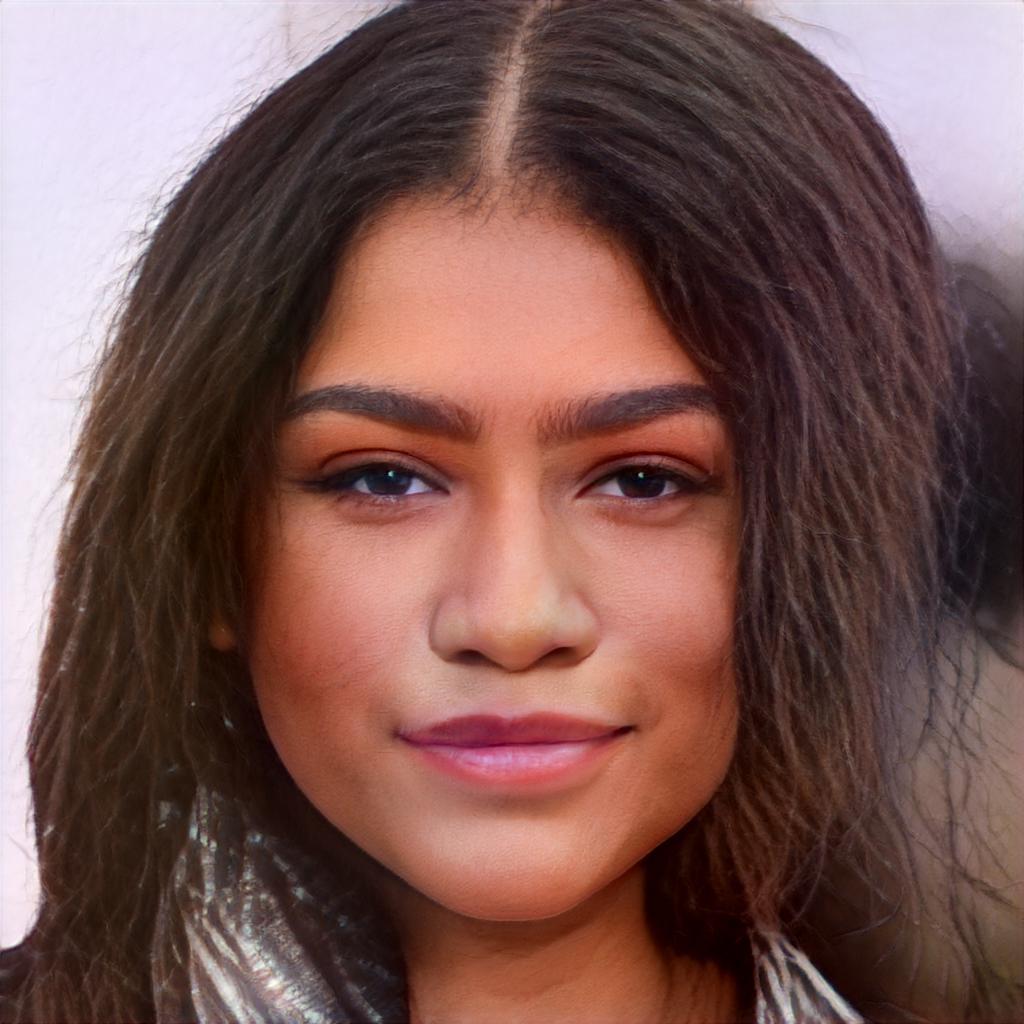} & 
        \includegraphics[width=0.24\columnwidth]{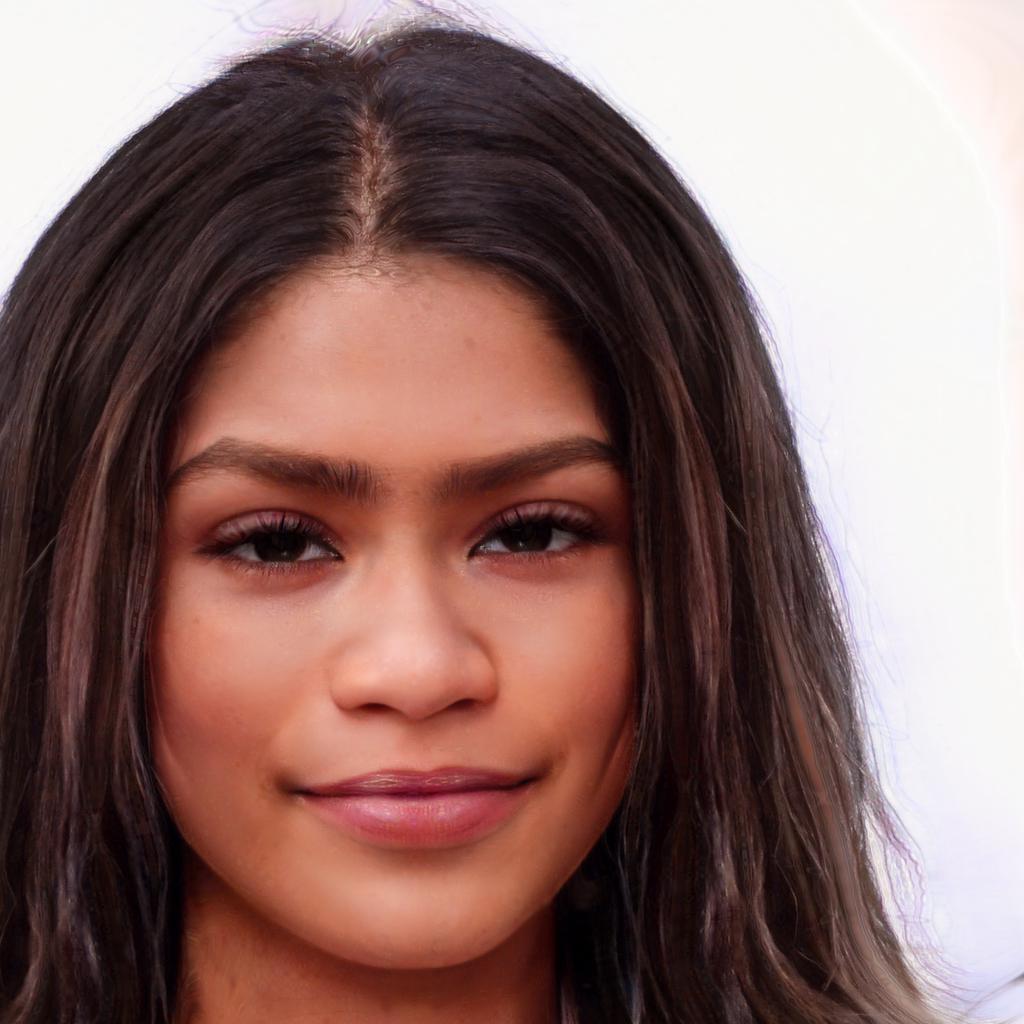} & 
        \includegraphics[width=0.24\columnwidth]{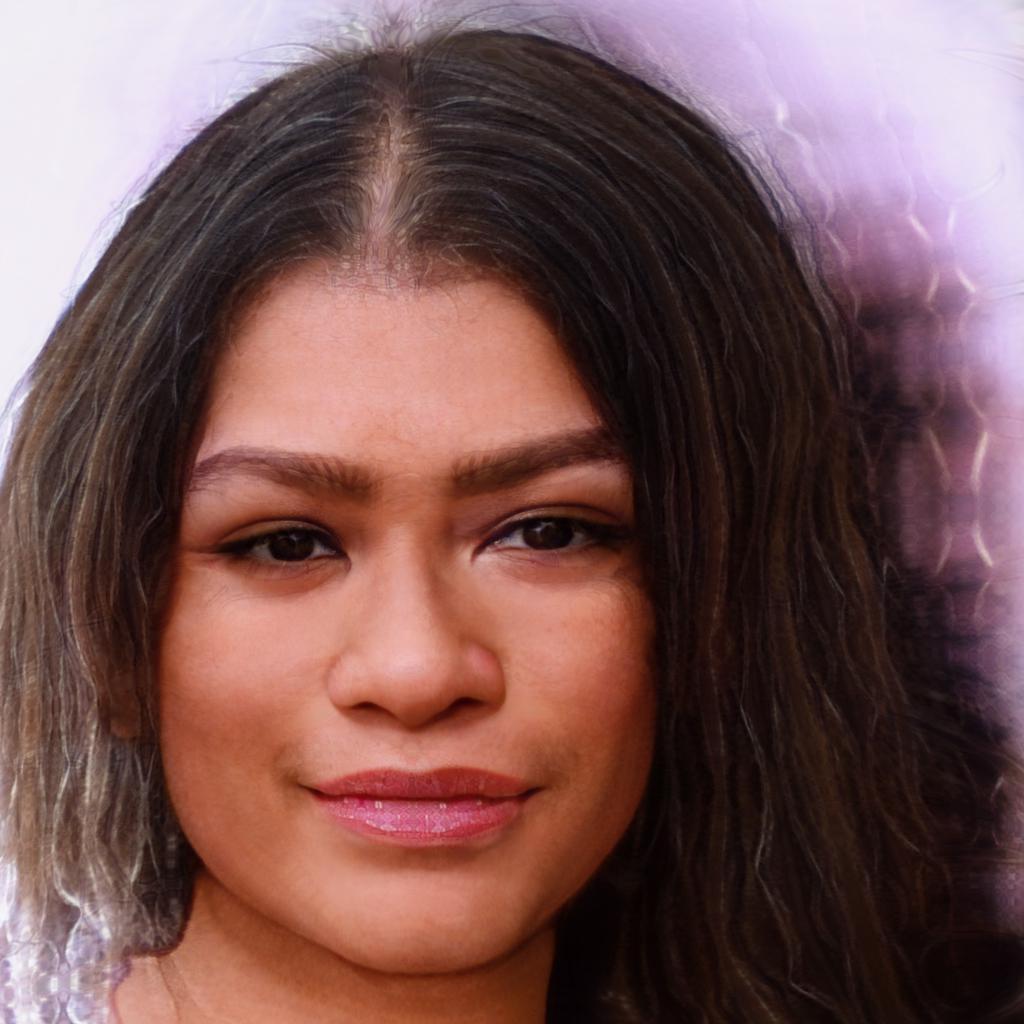} \\

		\raisebox{0.1in}{\rotatebox{90}{$+$ Red Lipstick}} &
        \includegraphics[width=0.24\columnwidth]{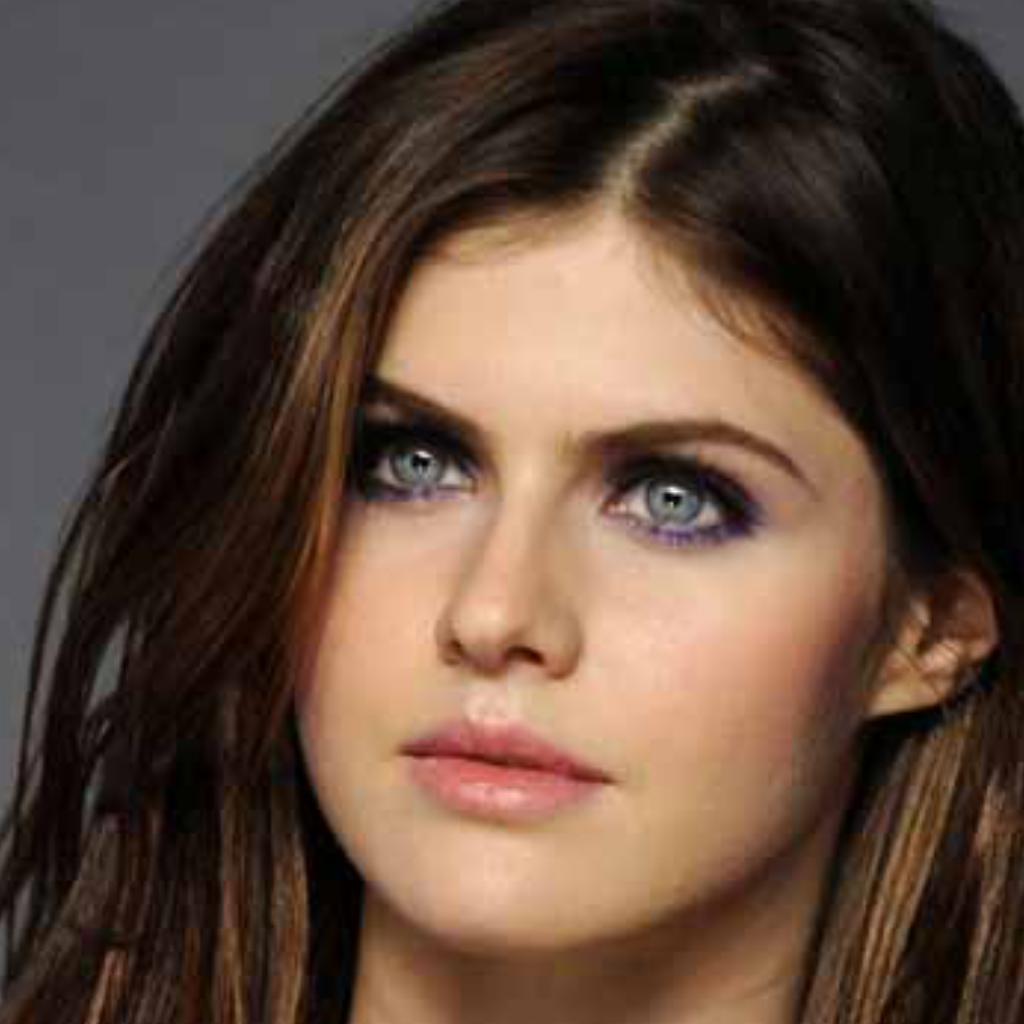} & 
        \includegraphics[width=0.24\columnwidth]{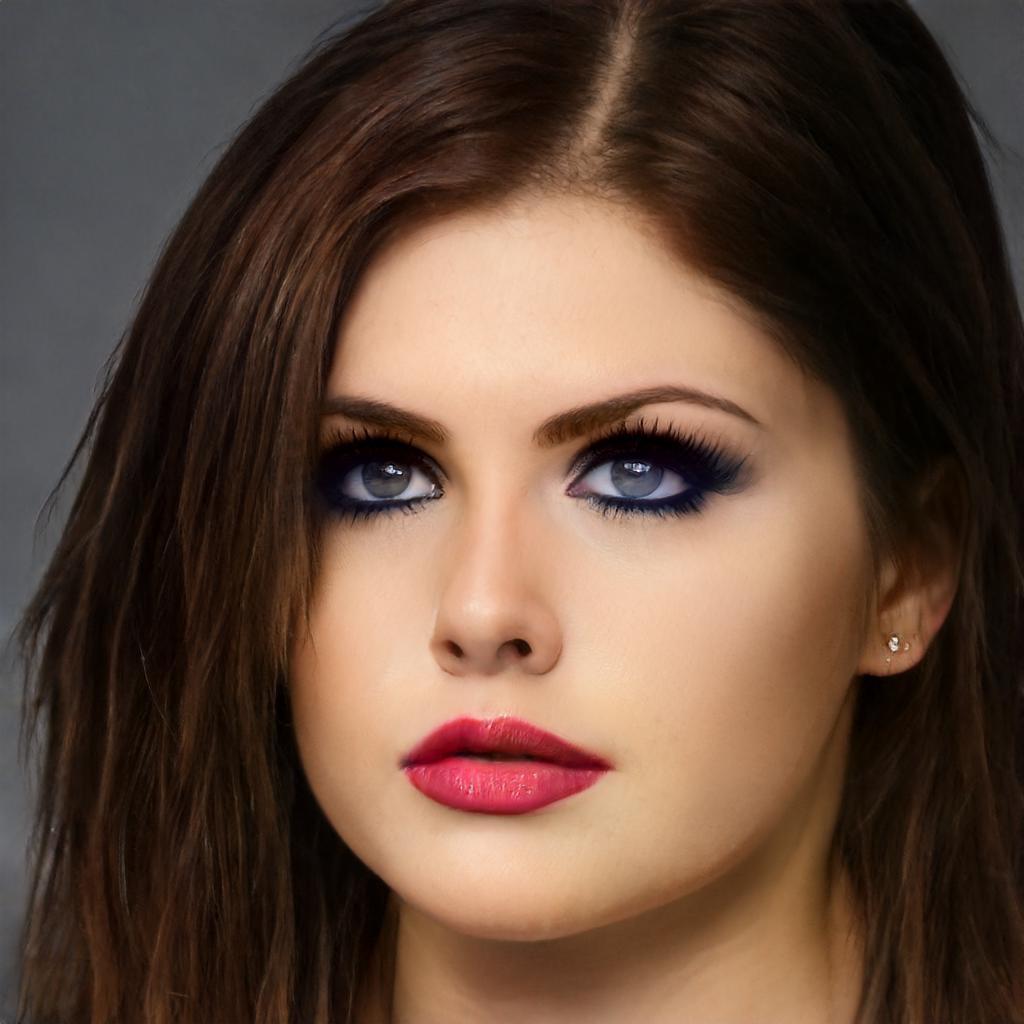} & 
        \includegraphics[width=0.24\columnwidth]{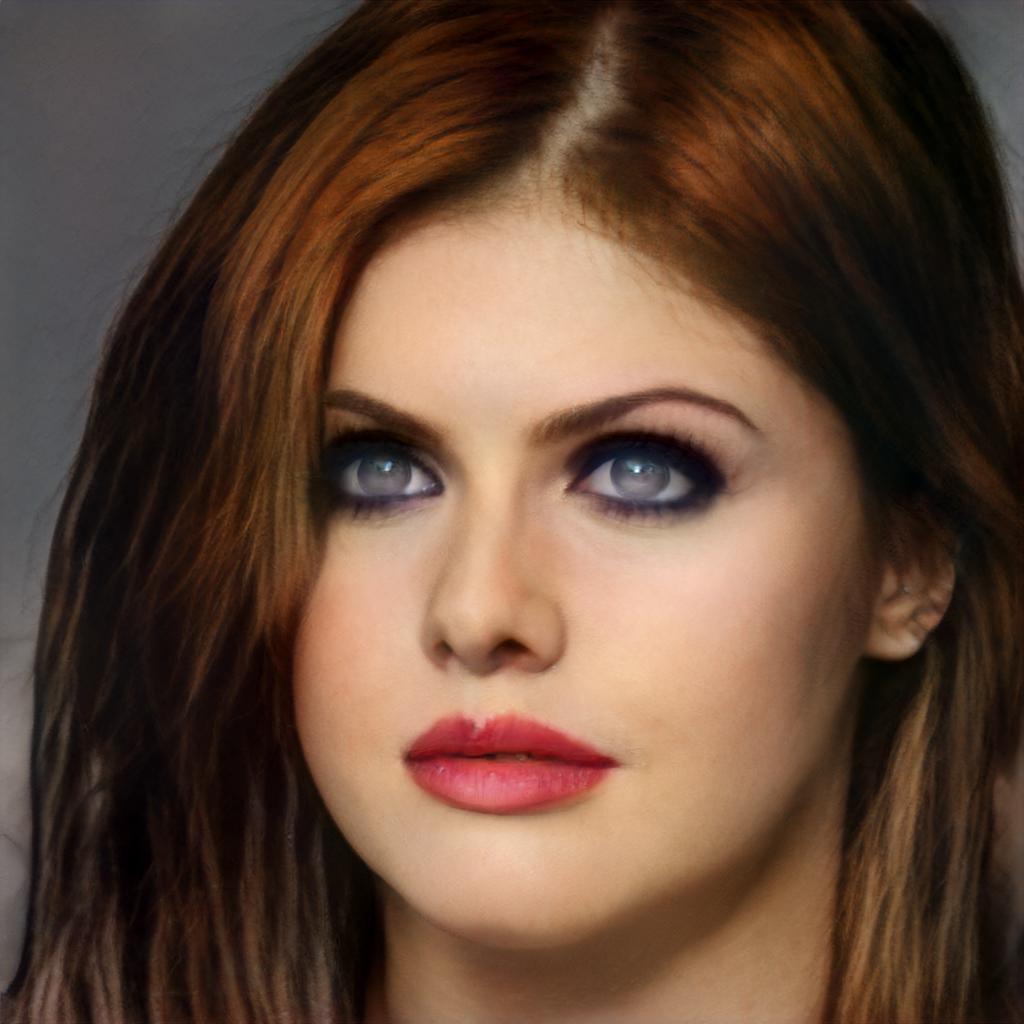} & 
        \includegraphics[width=0.24\columnwidth]{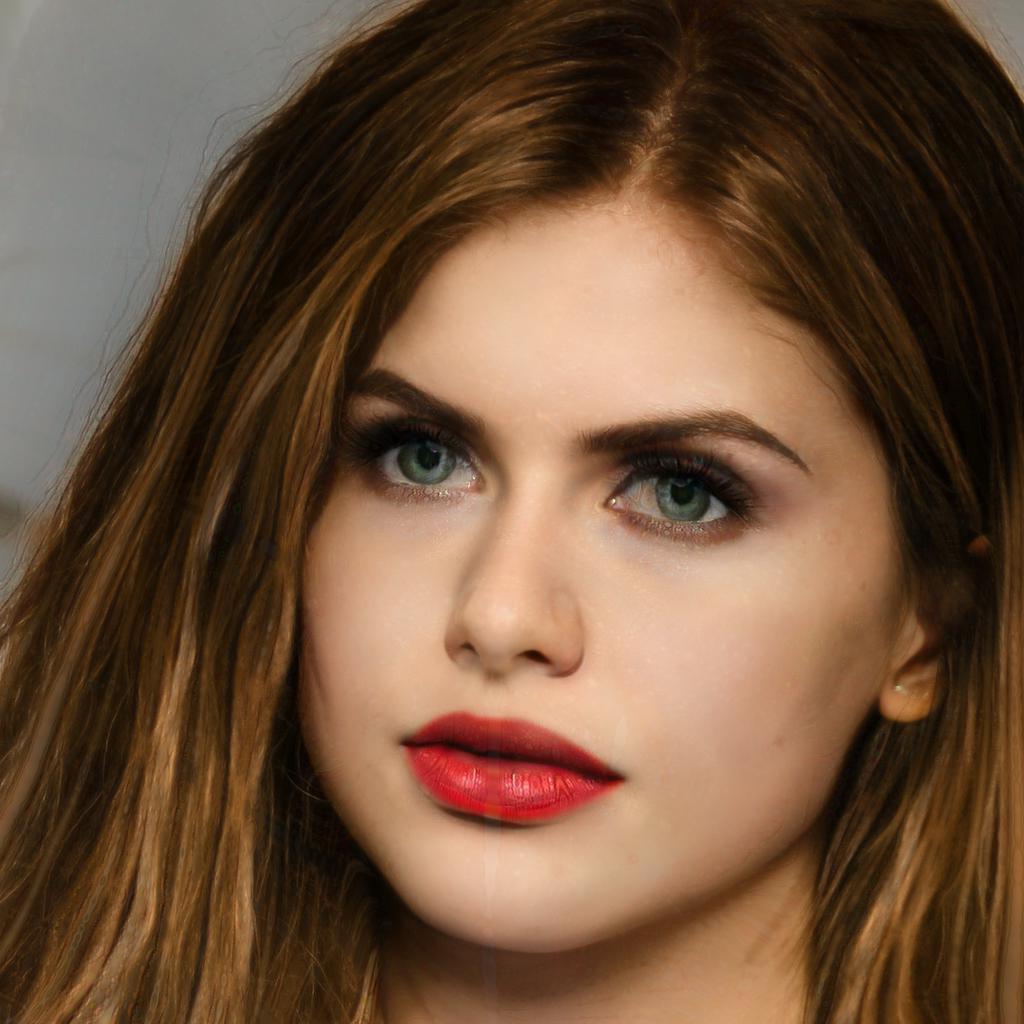} & 
        \includegraphics[width=0.24\columnwidth]{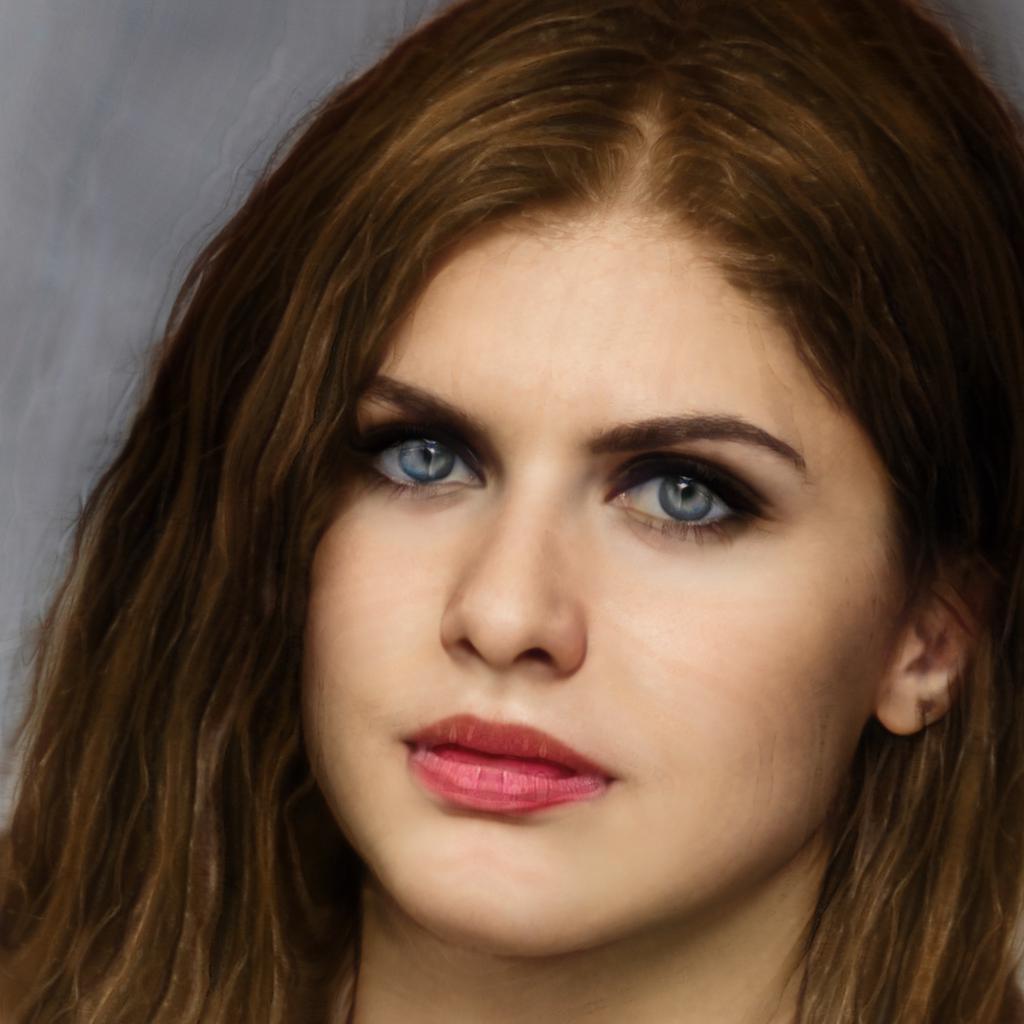} \\

		\raisebox{0.2in}{\rotatebox{90}{$+$ Afro}} &
        \includegraphics[width=0.24\columnwidth]{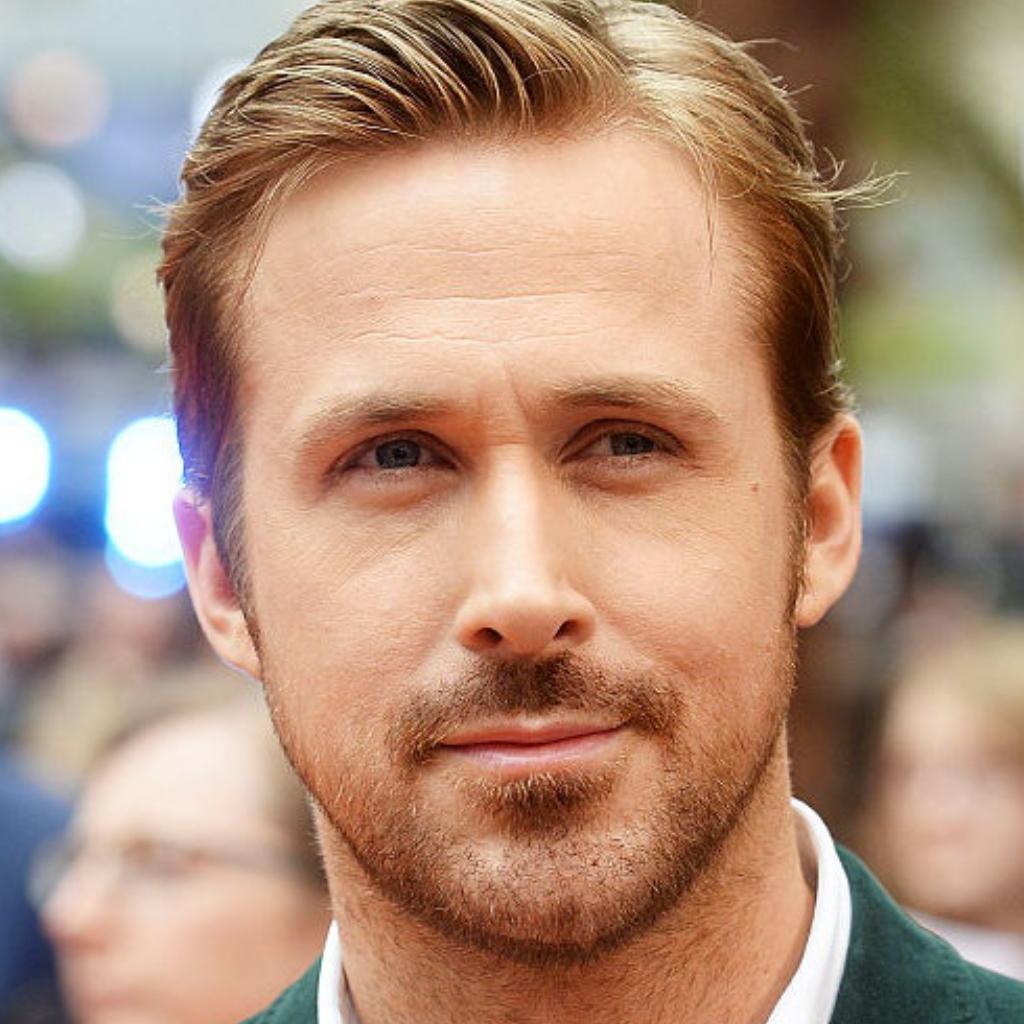} & 
        \includegraphics[width=0.24\columnwidth]{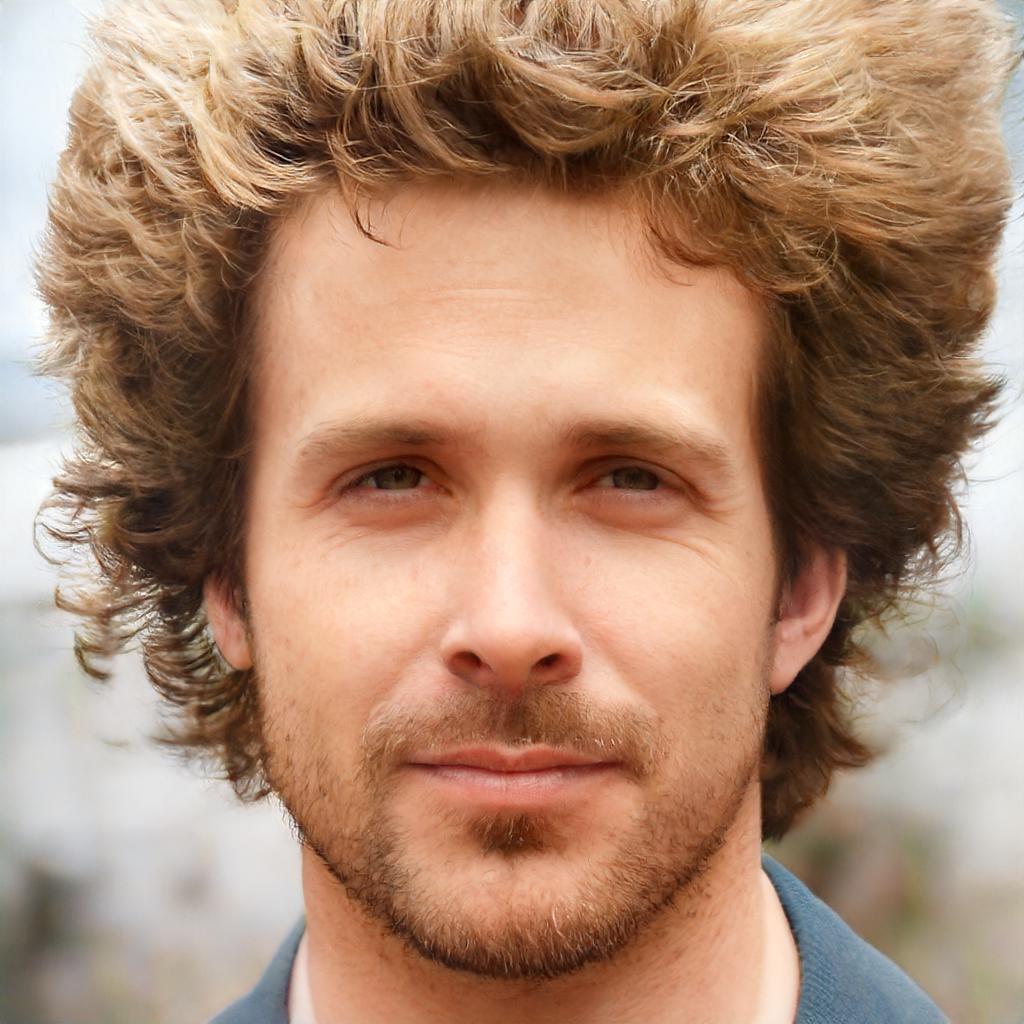} & 
        \includegraphics[width=0.24\columnwidth]{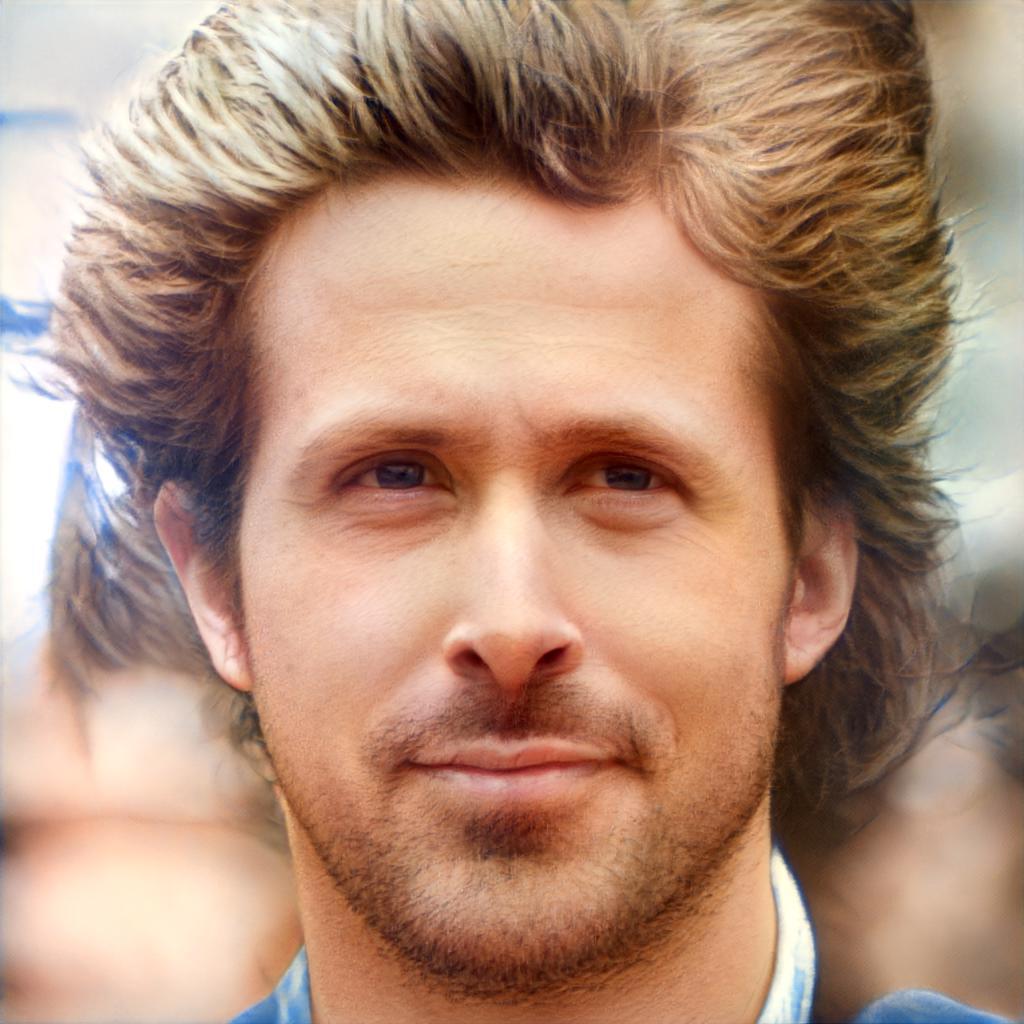} & 
        \includegraphics[width=0.24\columnwidth]{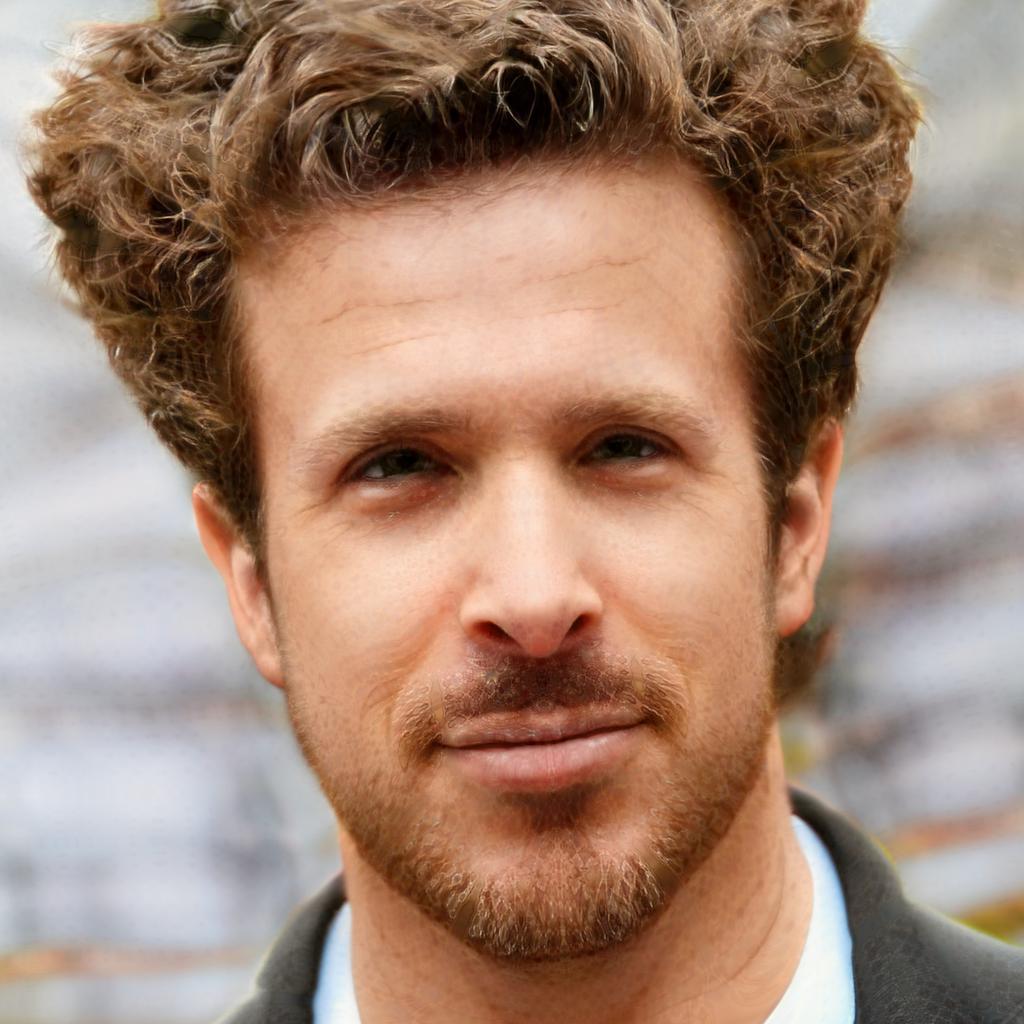} & 
        \includegraphics[width=0.24\columnwidth]{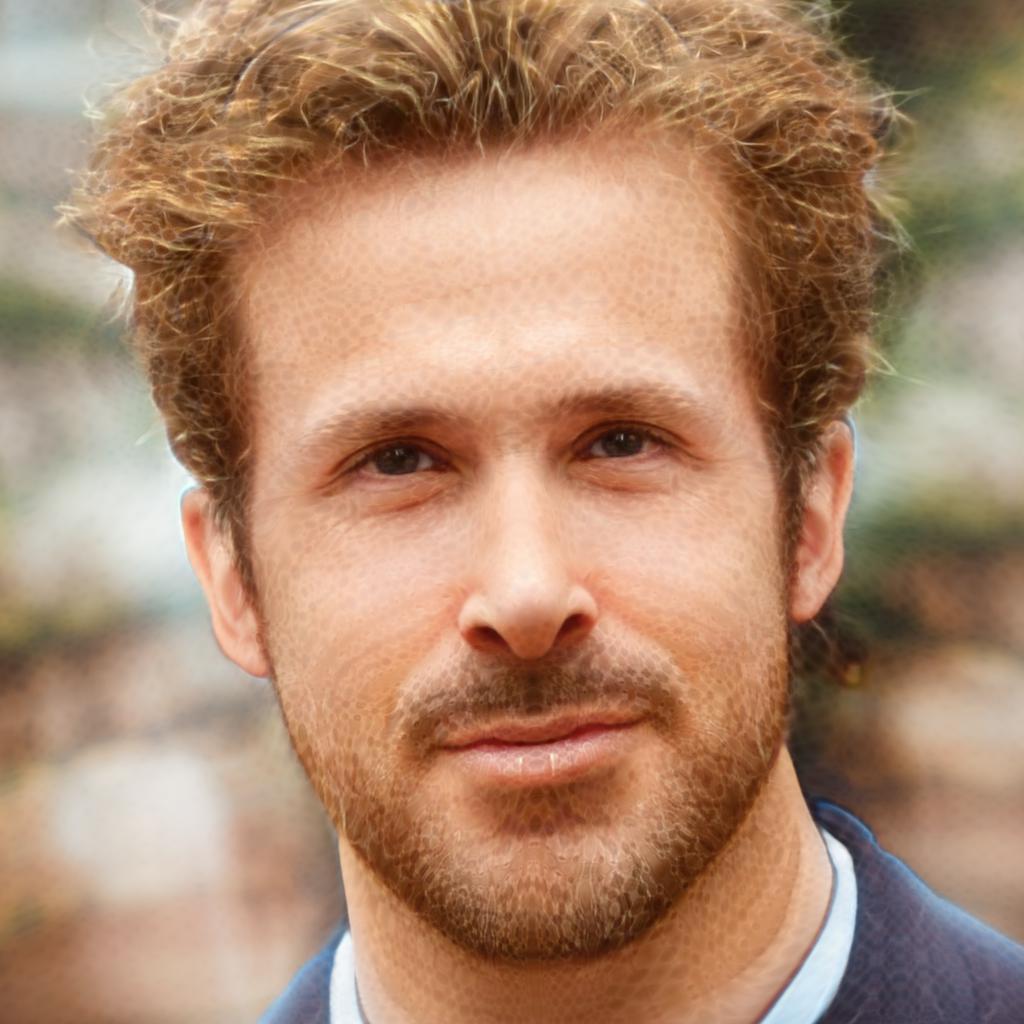} \\
        
		\raisebox{0.2in}{\rotatebox{90}{$+$ Tan}} &
        \includegraphics[width=0.24\columnwidth]{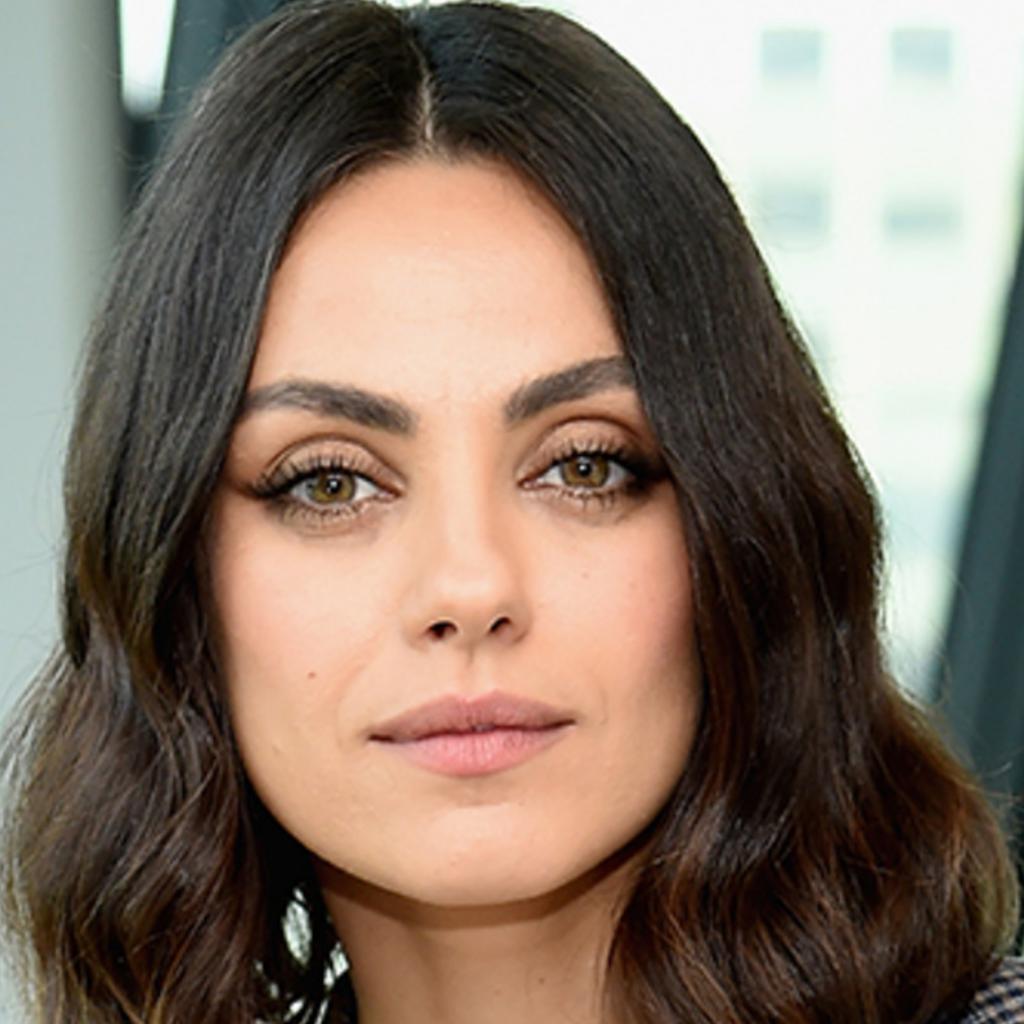} & 
        \includegraphics[width=0.24\columnwidth]{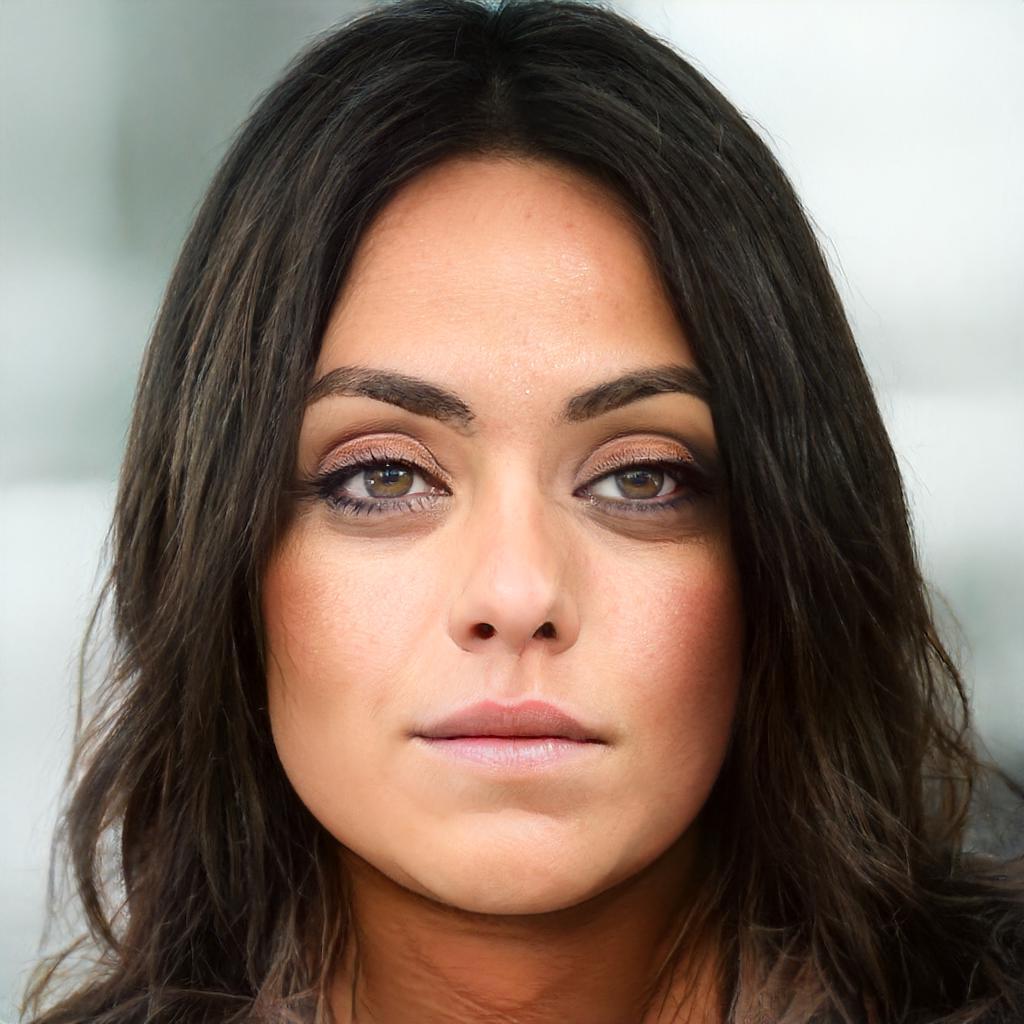} & 
        \includegraphics[width=0.24\columnwidth]{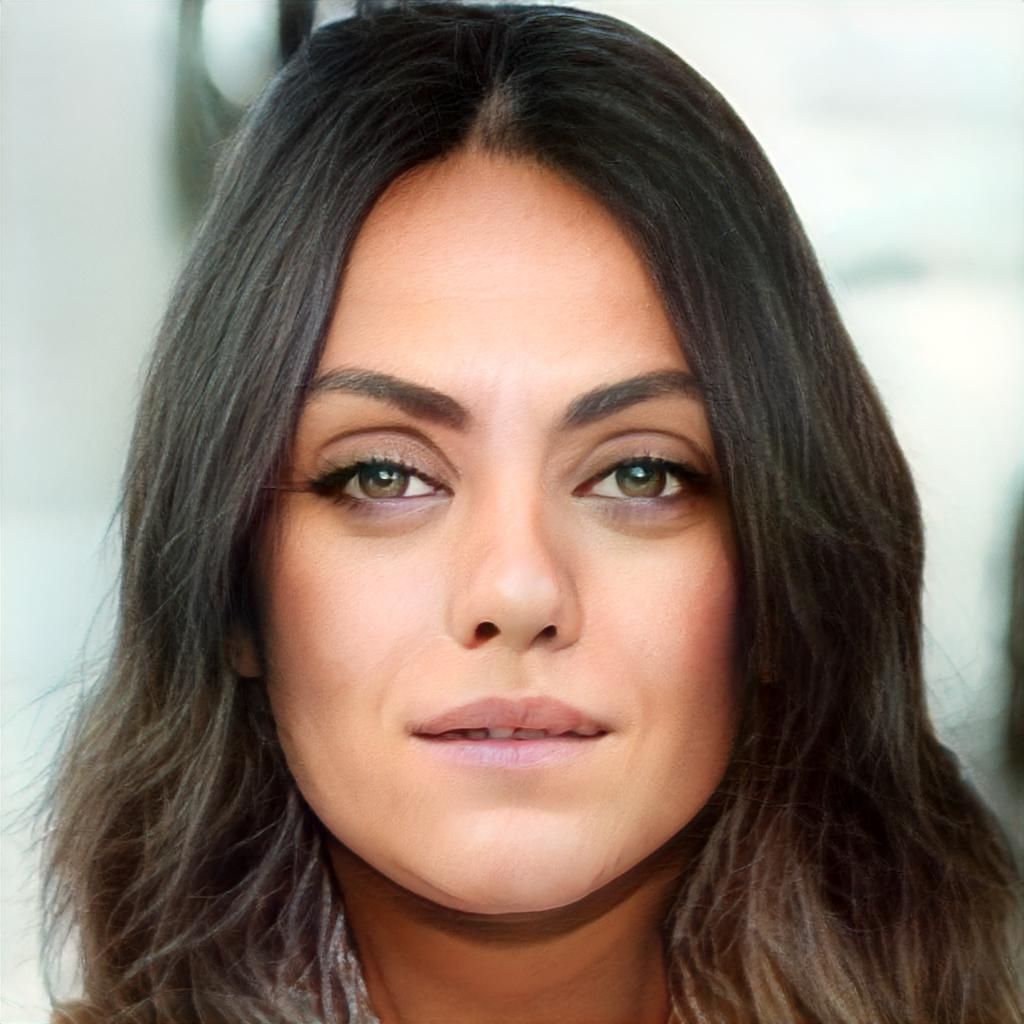} & 
        \includegraphics[width=0.24\columnwidth]{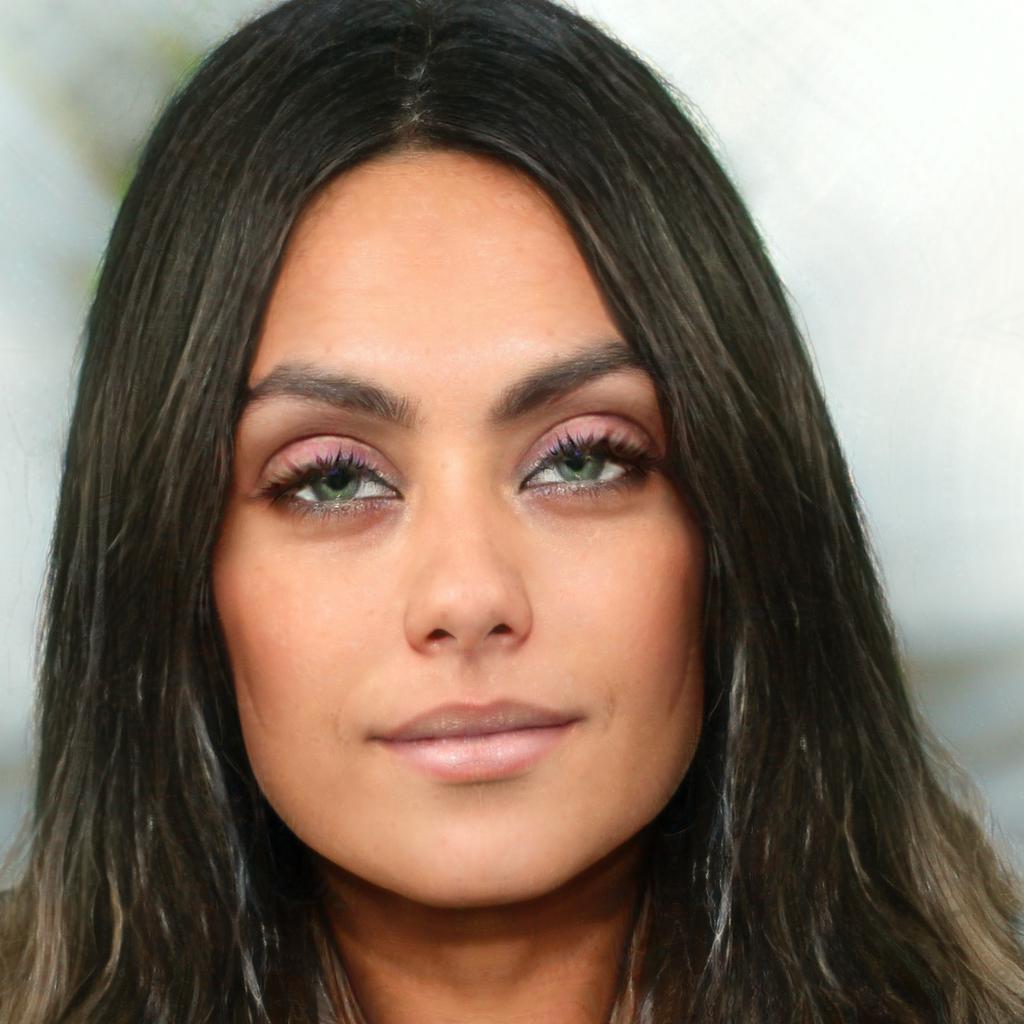} & 
        \includegraphics[width=0.24\columnwidth]{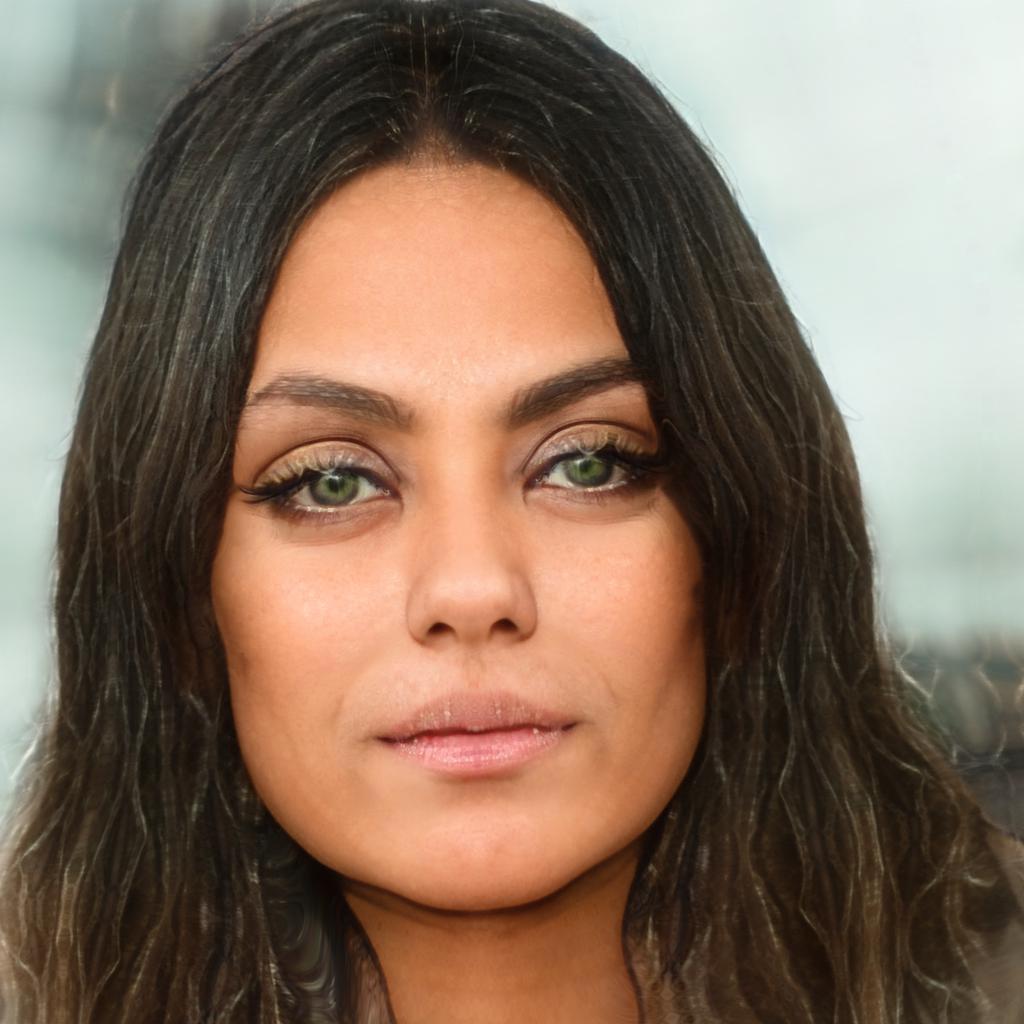} \\

		 & Unaligned & SG2 $\text{ReStyle}_{e4e}$ & SG2 $\text{ReStyle}_{pSp}$ & SG3 $\text{ReStyle}_{e4e}$ & SG3 $\text{ReStyle}_{pSp}$

	\end{tabular}
	}
	\vspace{-0.3cm}
	\caption{
	Editing quality comparison. We perform various edits~\cite{patashnik2021styleclip,shen2020interpreting} over latent codes obtained by each inversion method. When encoding images with our StyleGAN3 encoders, we reconstruct the original unaligned images using the predicted landmark-based user-transformations. 
	}
	\label{fig:editing_reals_supplementary}
\end{figure*}

%% file: figures/supplementary/video_results.tex
\begin{figure*}[tb]
	\centering
	\setlength{\tabcolsep}{1pt}	
	{\footnotesize
	\begin{tabular}{c c c c c c c c c c}

        \\ 
        \\ 
        \\
        \\ 

		\raisebox{0.15in}{\rotatebox{90}{Original}} &
        \includegraphics[width=0.215\columnwidth]{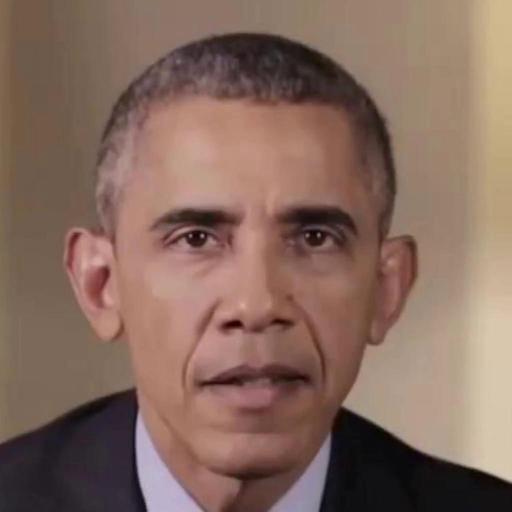} & 
        \includegraphics[width=0.215\columnwidth]{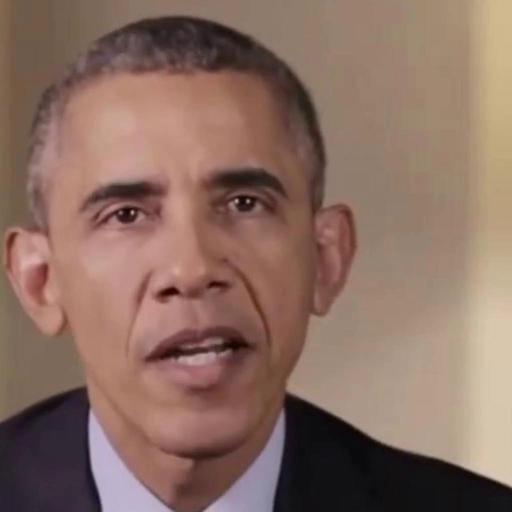} & 
        \includegraphics[width=0.215\columnwidth]{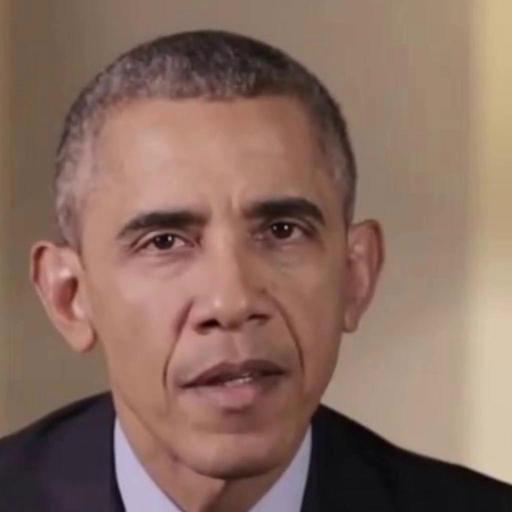} & 
        \includegraphics[width=0.215\columnwidth]{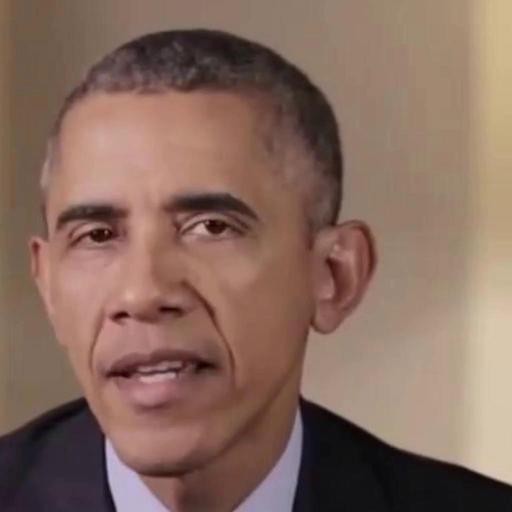} & 
        \includegraphics[width=0.215\columnwidth]{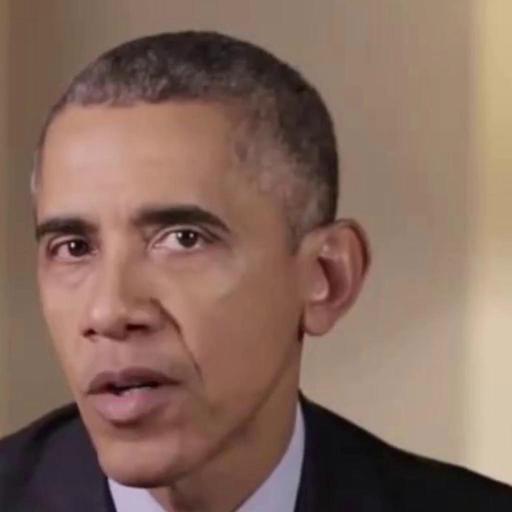} & 
        \includegraphics[width=0.215\columnwidth]{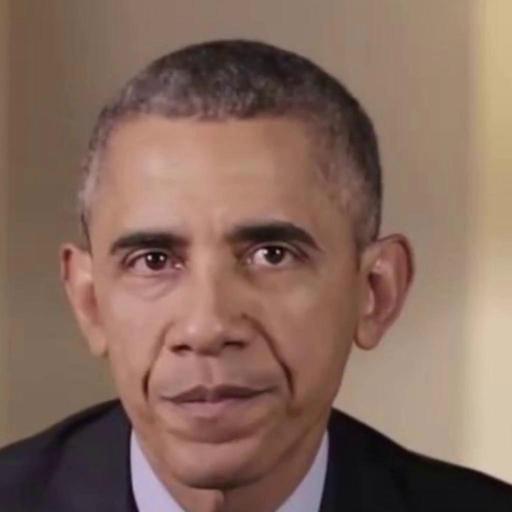} & 
        \includegraphics[width=0.215\columnwidth]{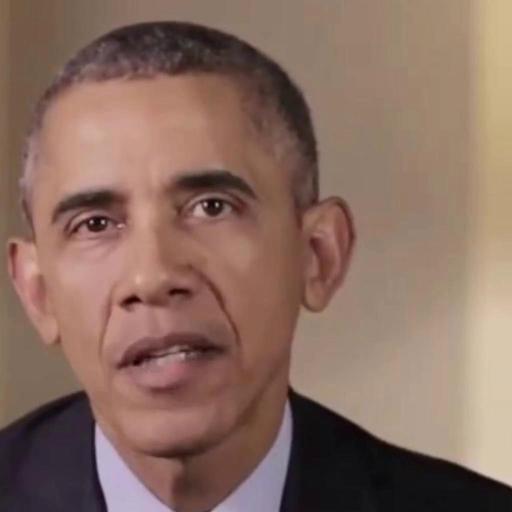} & 
        \includegraphics[width=0.215\columnwidth]{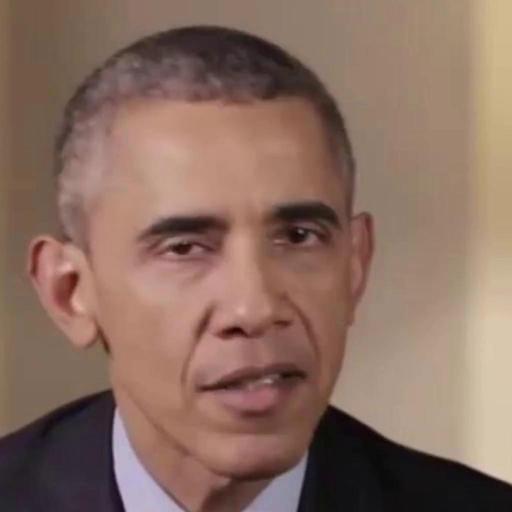} & 
        \includegraphics[width=0.215\columnwidth]{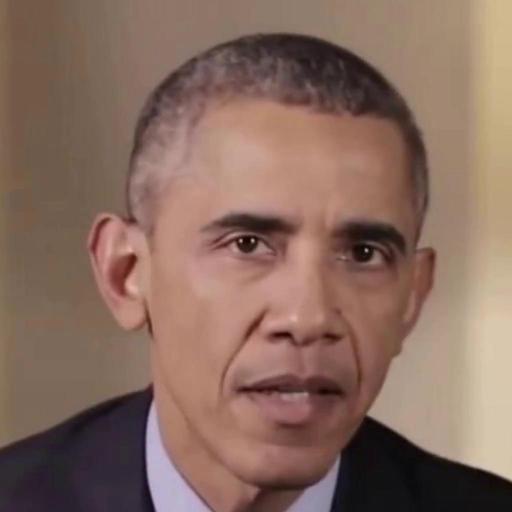} \\

		\raisebox{0.05in}{\rotatebox{90}{Reconstruction}} &
        \includegraphics[width=0.215\columnwidth]{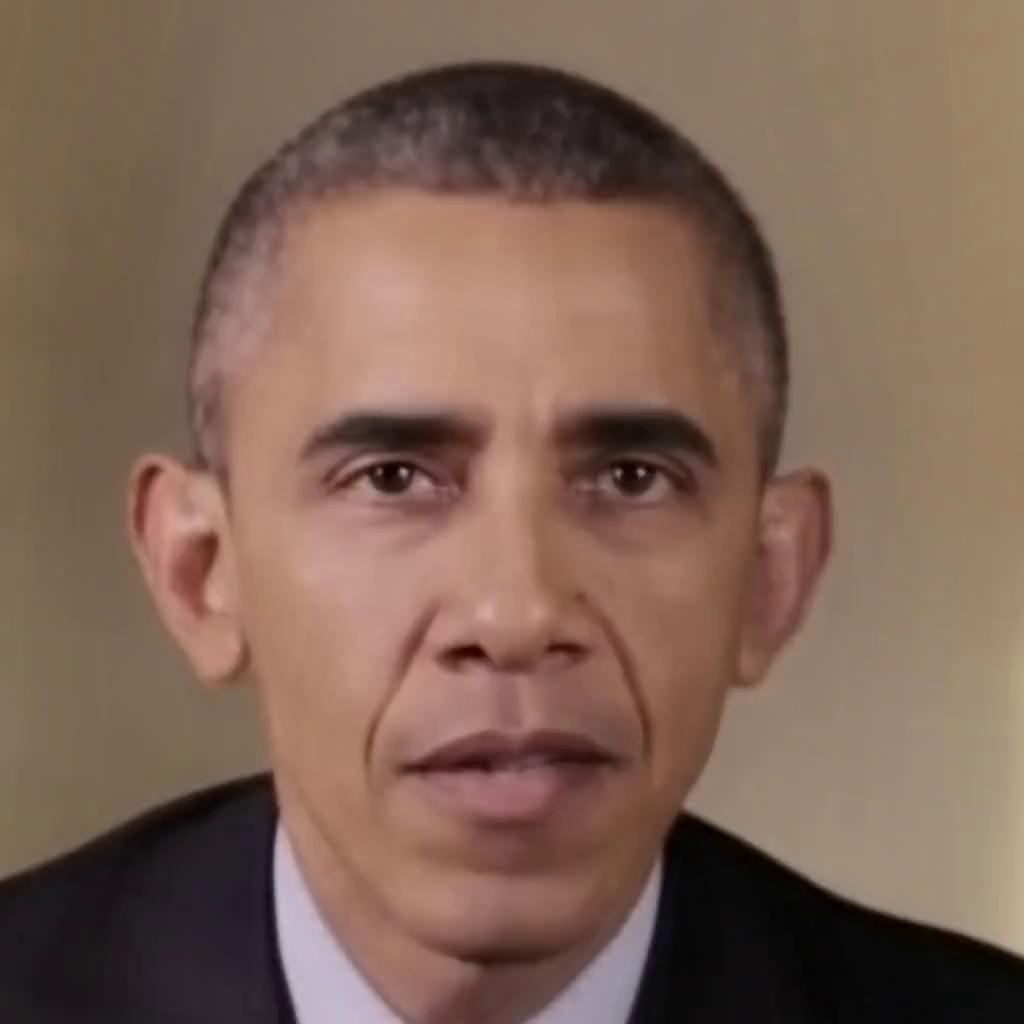} & 
        \includegraphics[width=0.215\columnwidth]{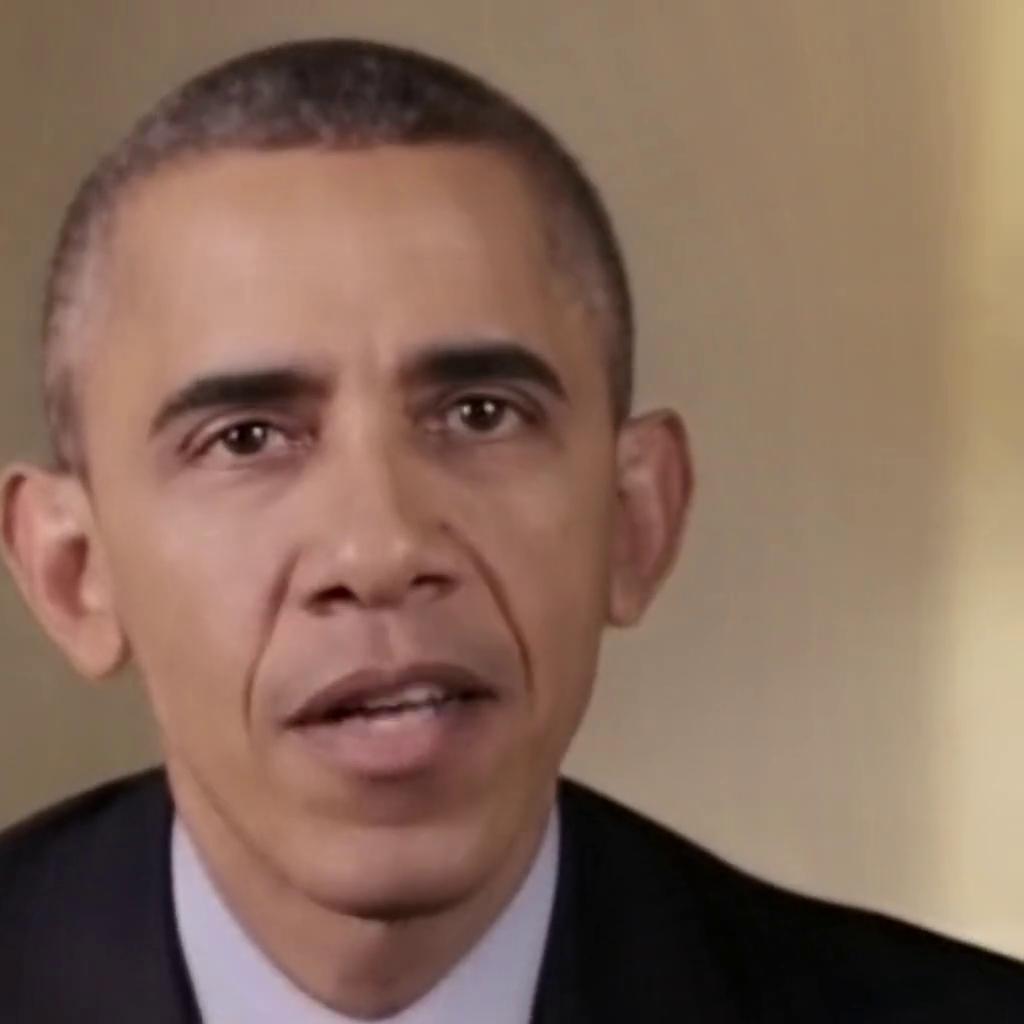} & 
        \includegraphics[width=0.215\columnwidth]{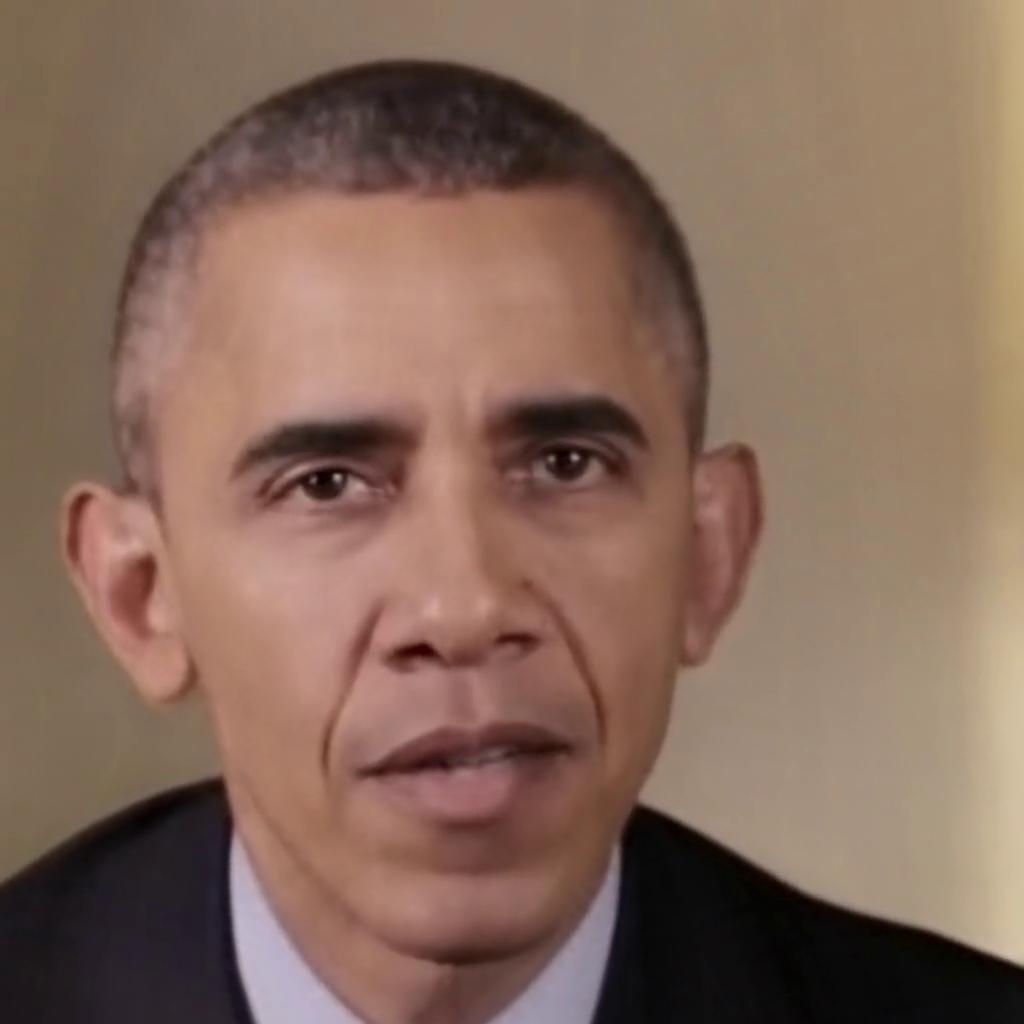} & 
        \includegraphics[width=0.215\columnwidth]{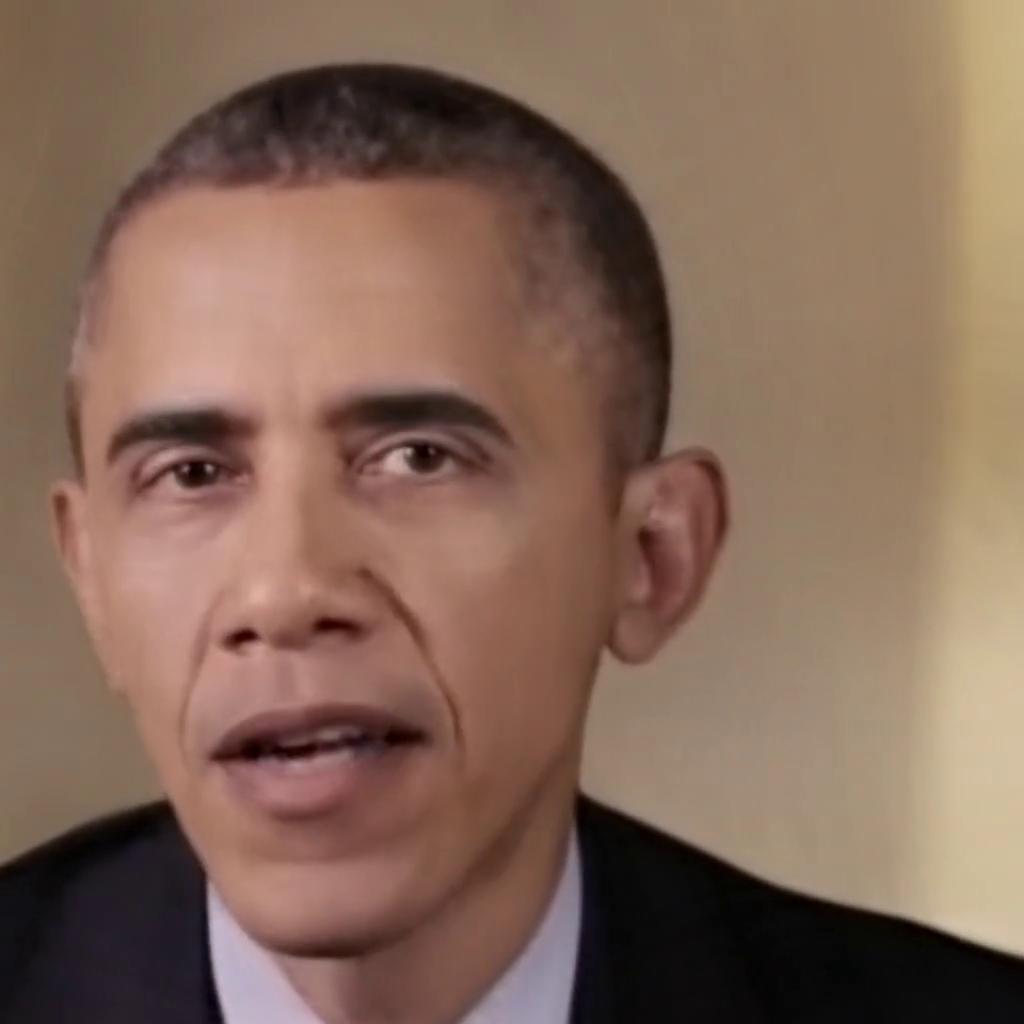} & 
        \includegraphics[width=0.215\columnwidth]{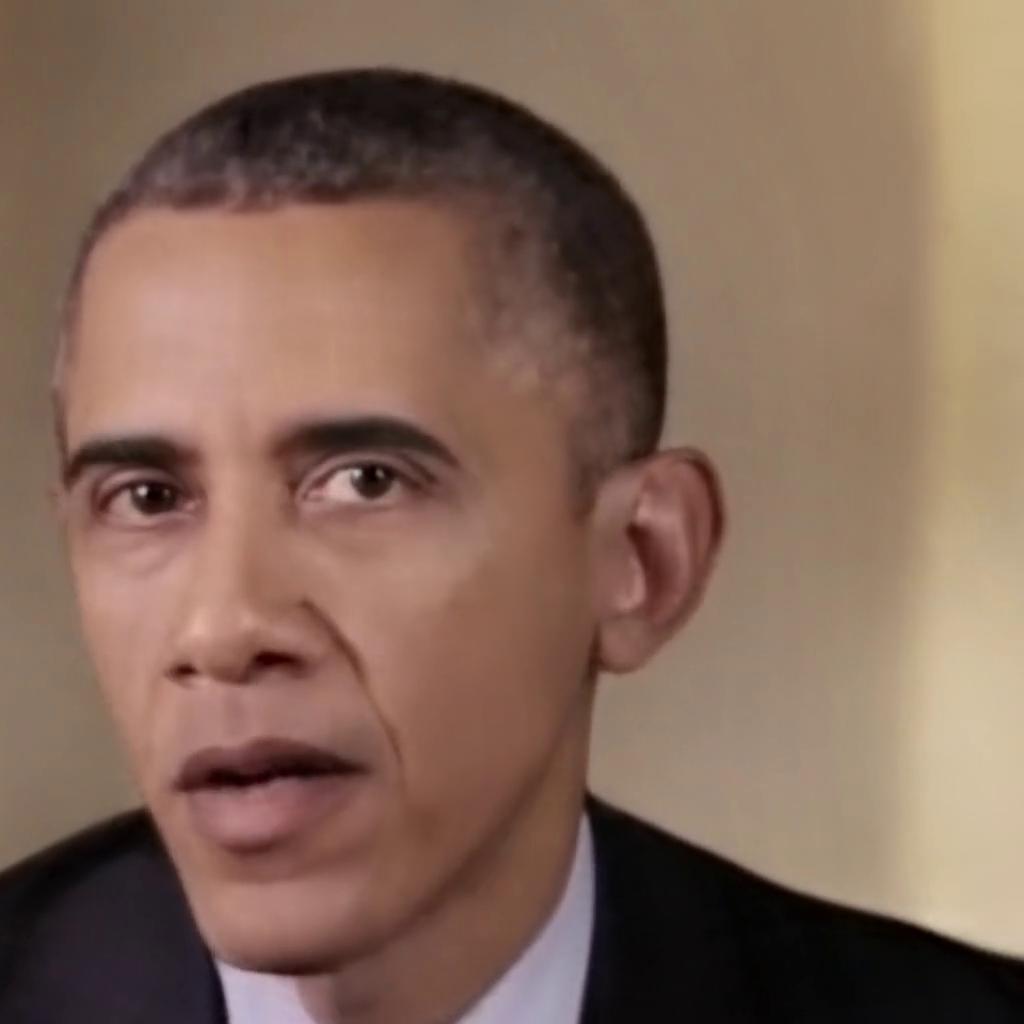} & 
        \includegraphics[width=0.215\columnwidth]{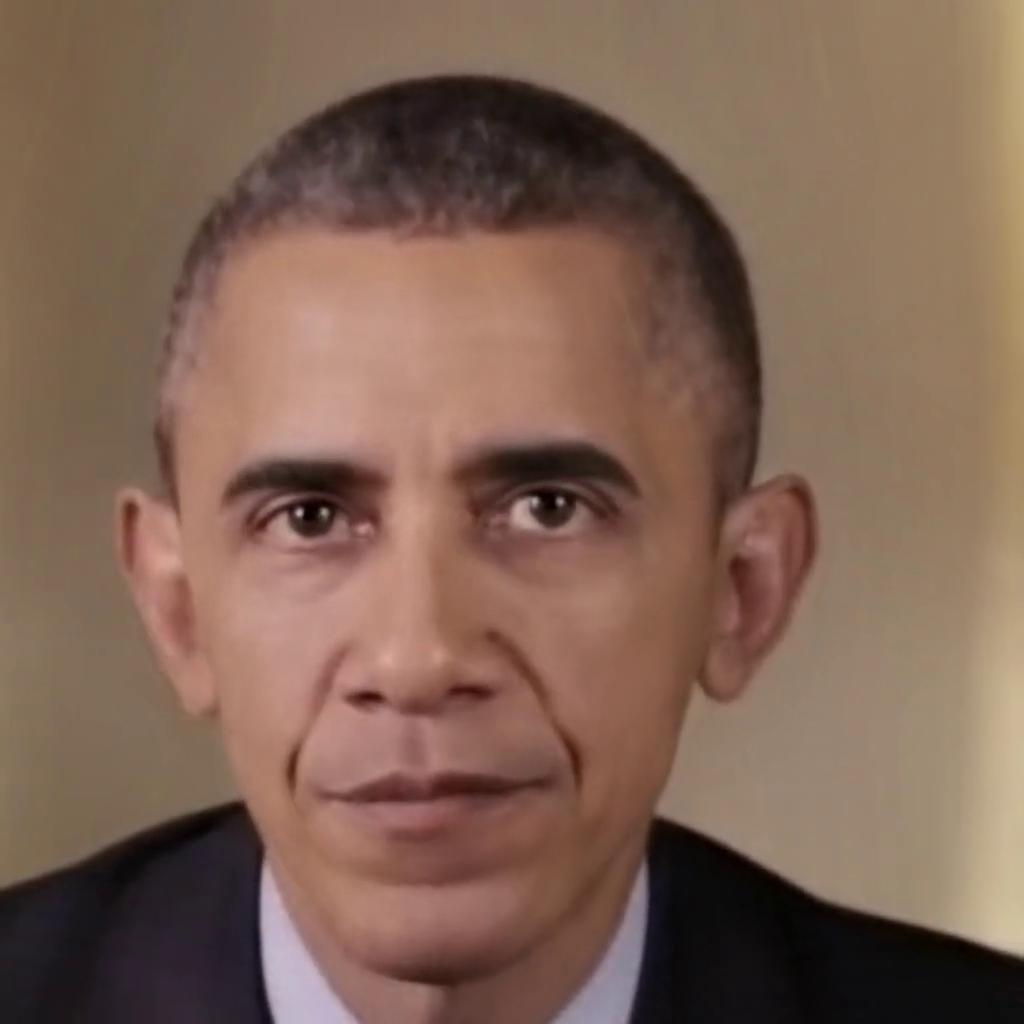} & 
        \includegraphics[width=0.215\columnwidth]{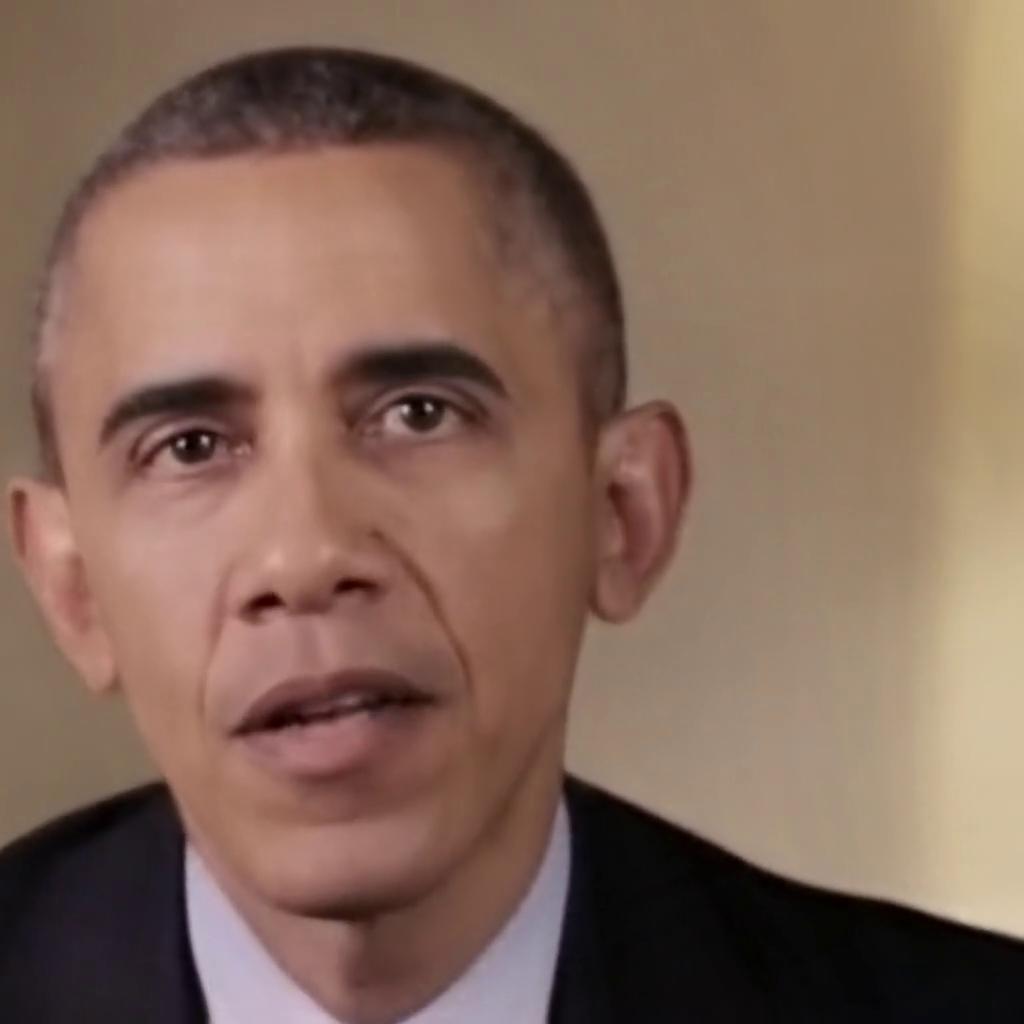} & 
        \includegraphics[width=0.215\columnwidth]{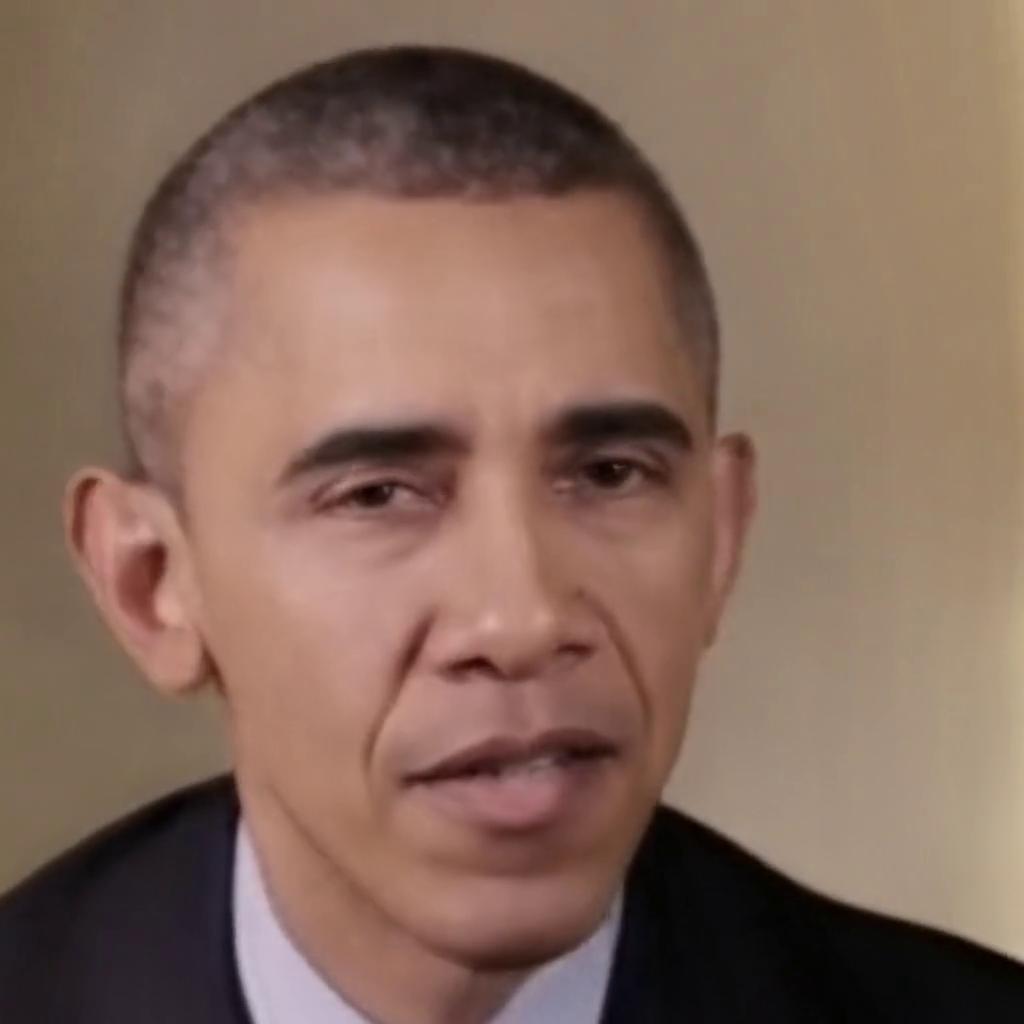} & 
        \includegraphics[width=0.215\columnwidth]{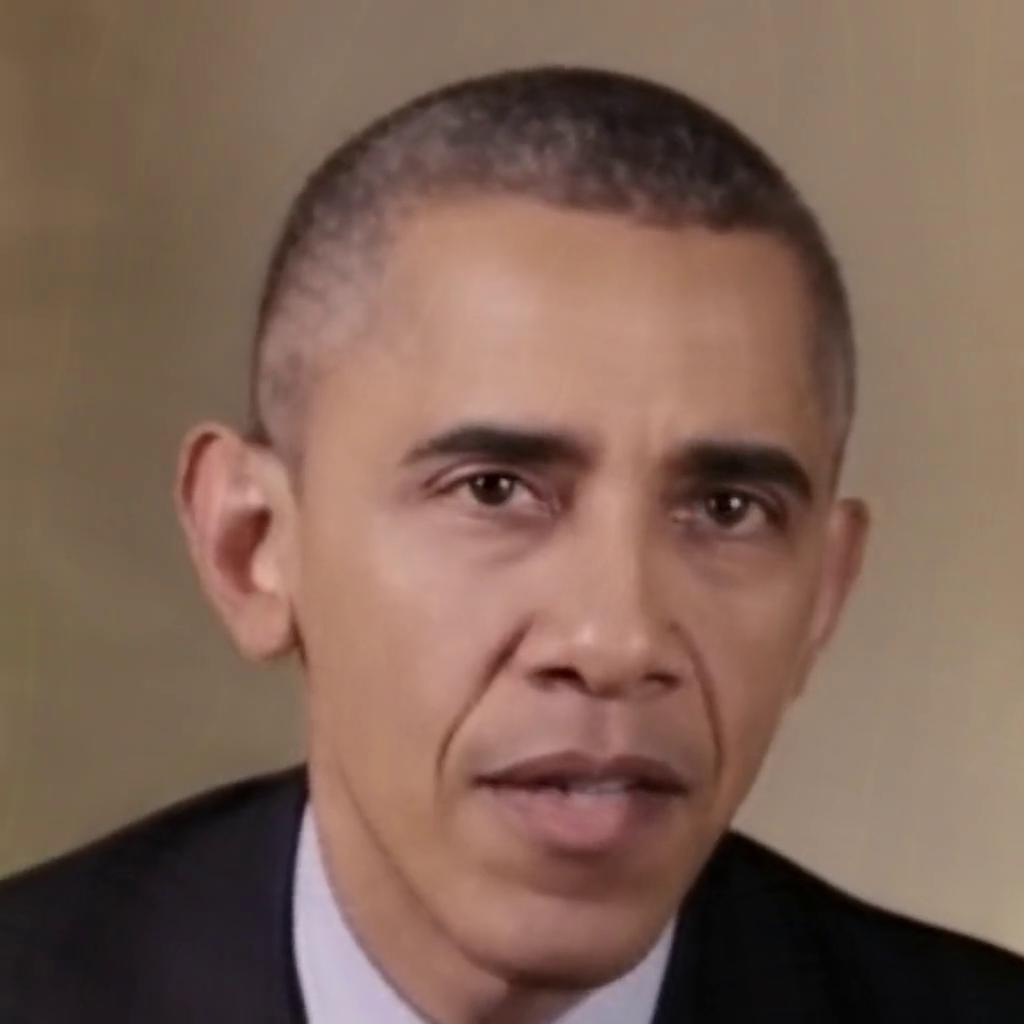} \\

		\raisebox{0.2in}{\rotatebox{90}{$+$ Afro}} &
        \includegraphics[width=0.215\columnwidth]{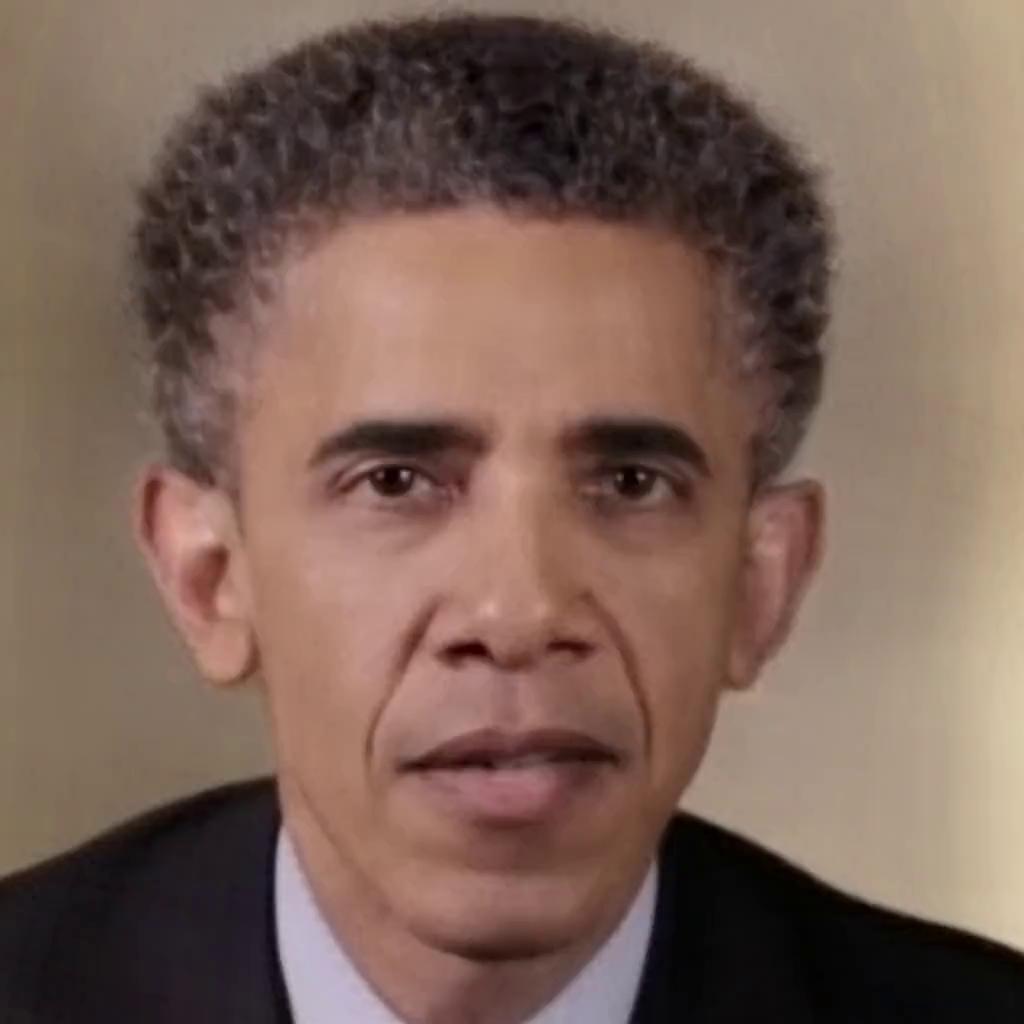} & 
        \includegraphics[width=0.215\columnwidth]{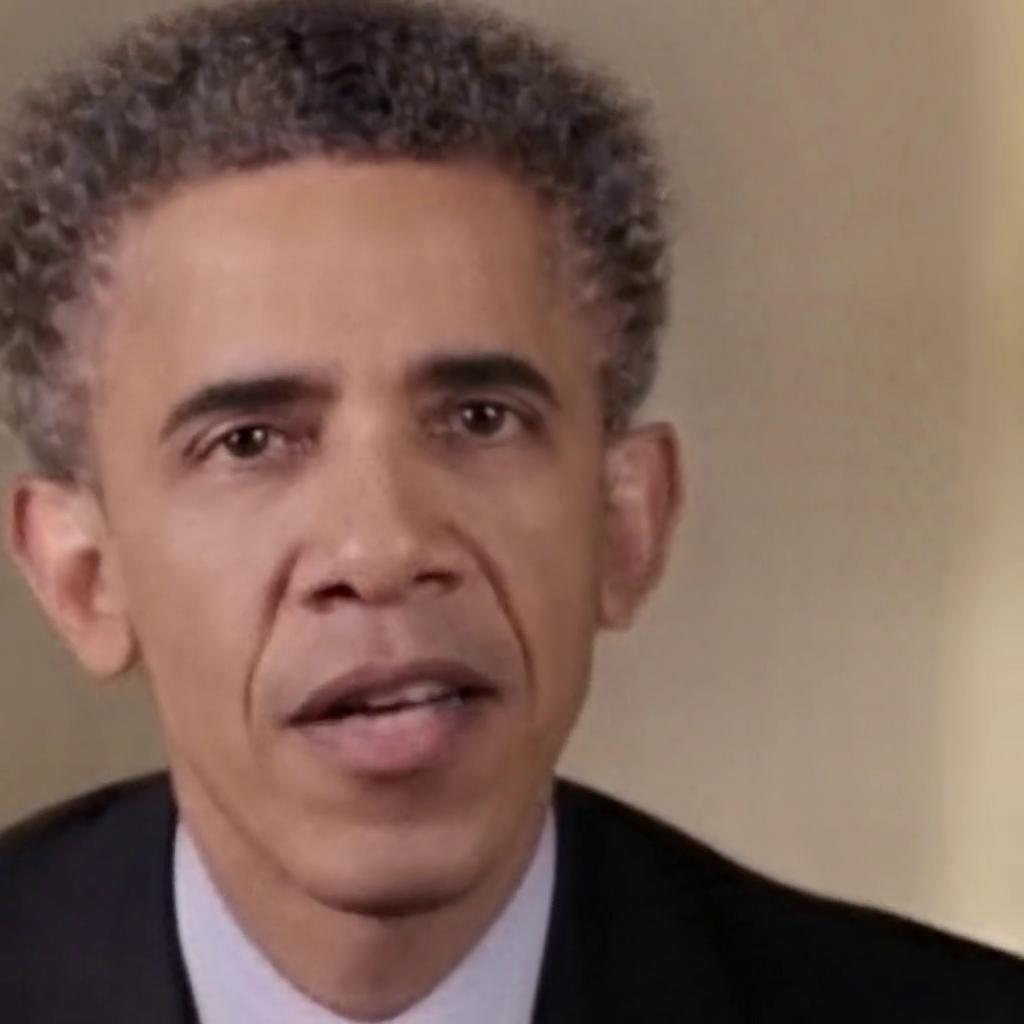} & 
        \includegraphics[width=0.215\columnwidth]{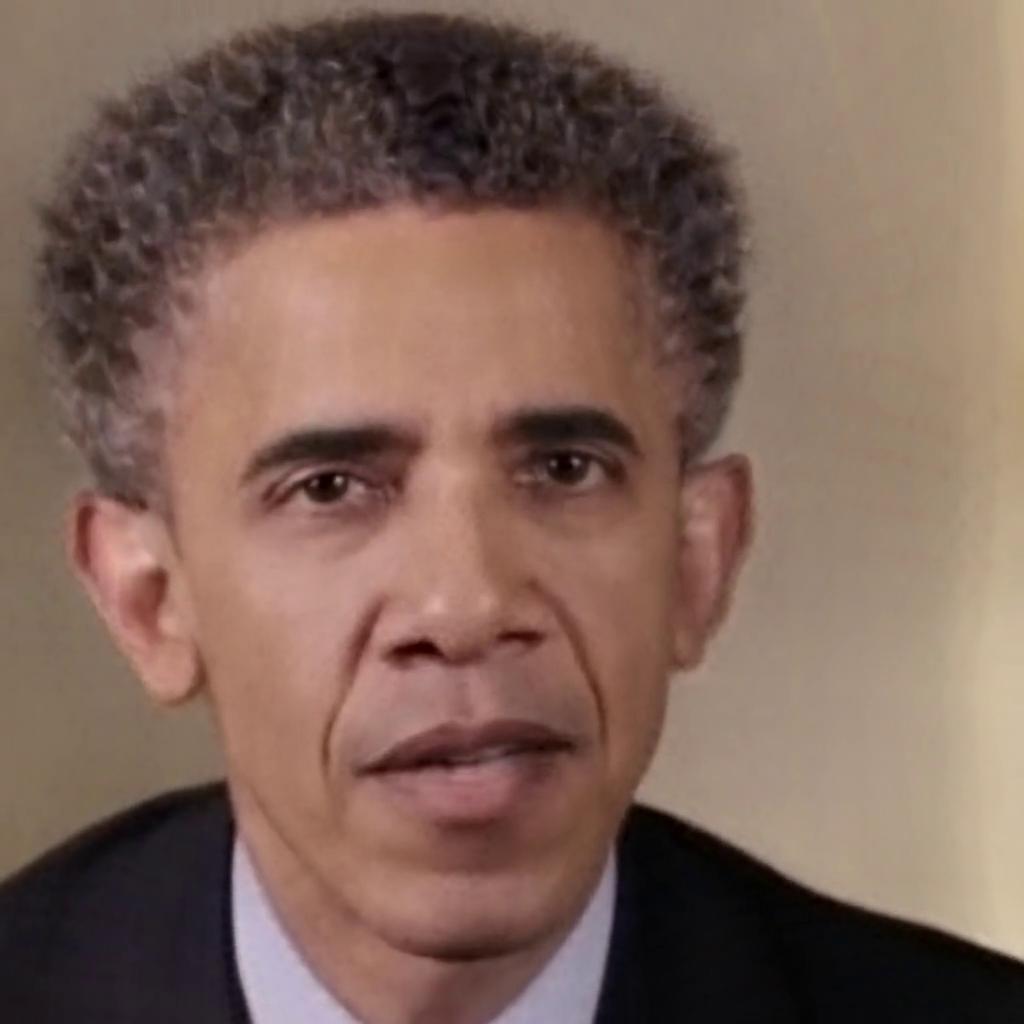} & 
        \includegraphics[width=0.215\columnwidth]{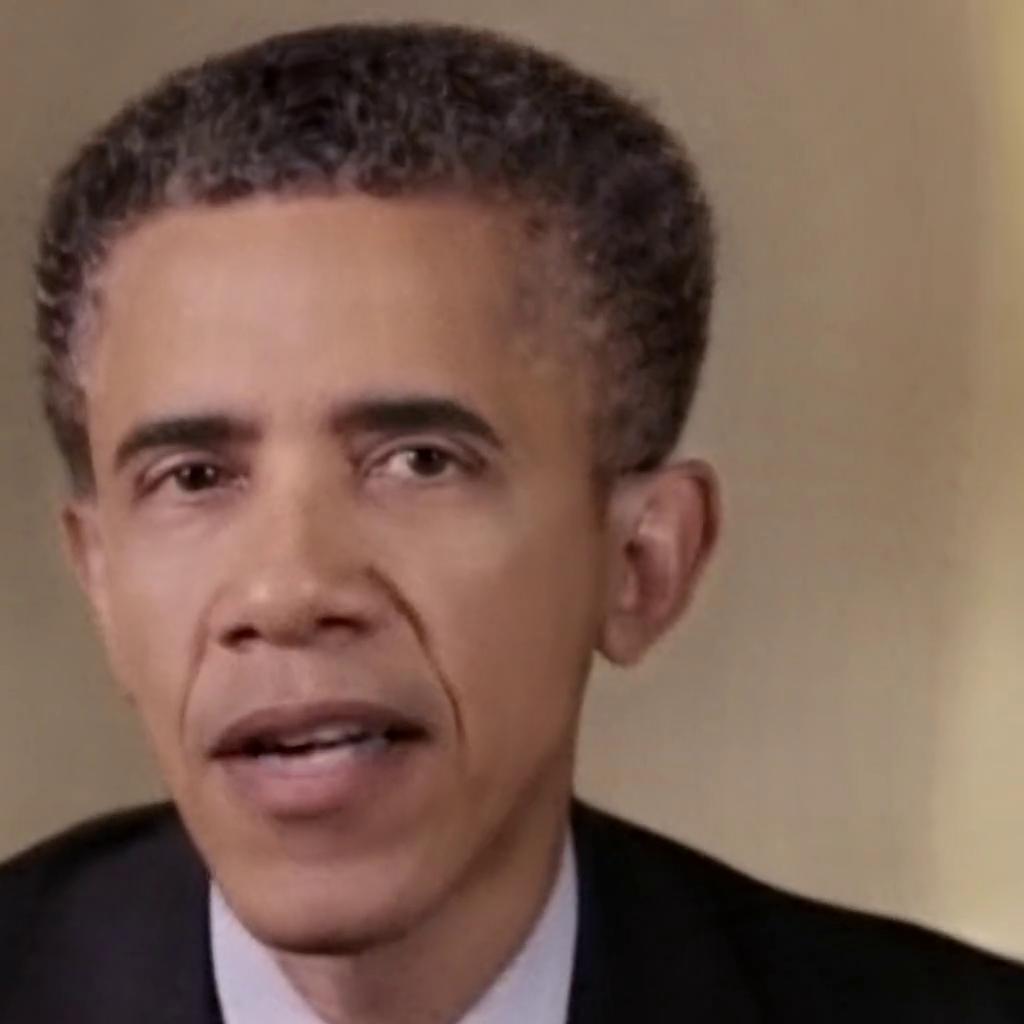} & 
        \includegraphics[width=0.215\columnwidth]{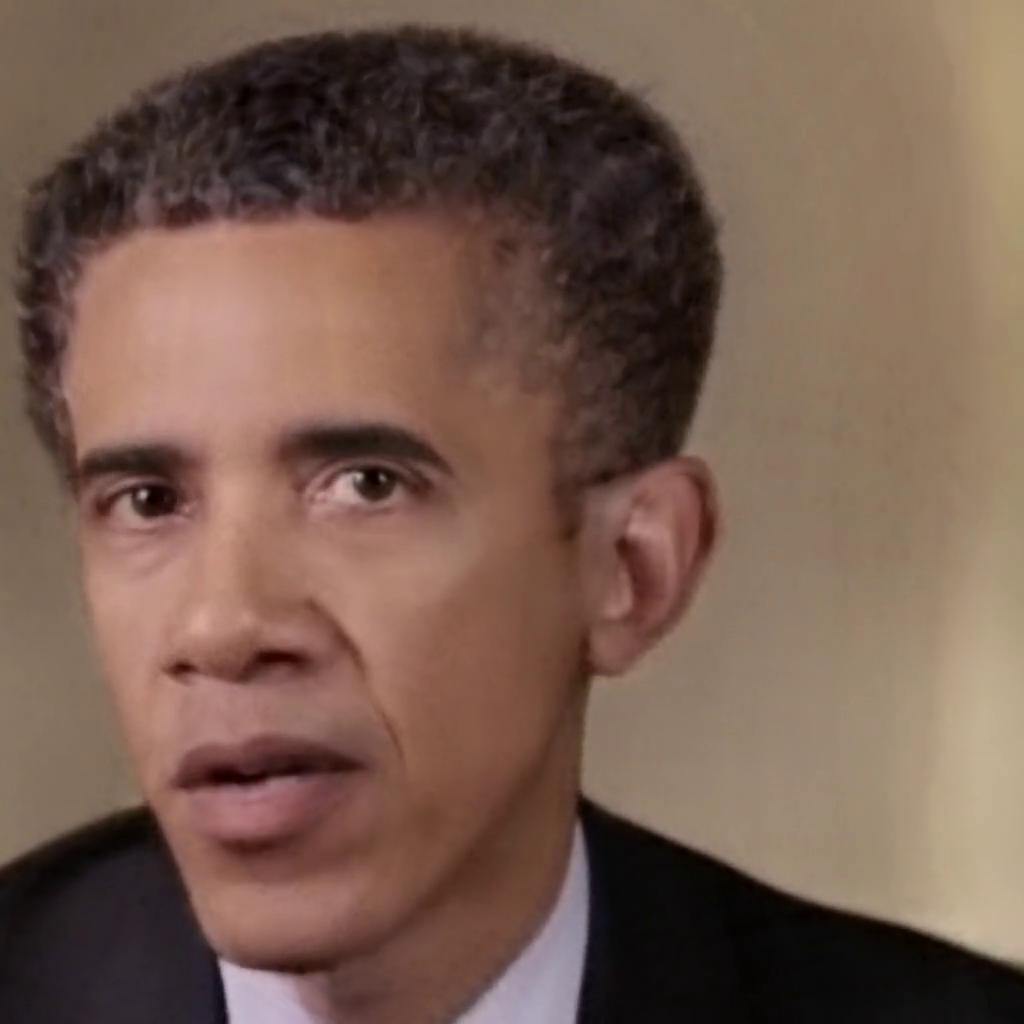} & 
        \includegraphics[width=0.215\columnwidth]{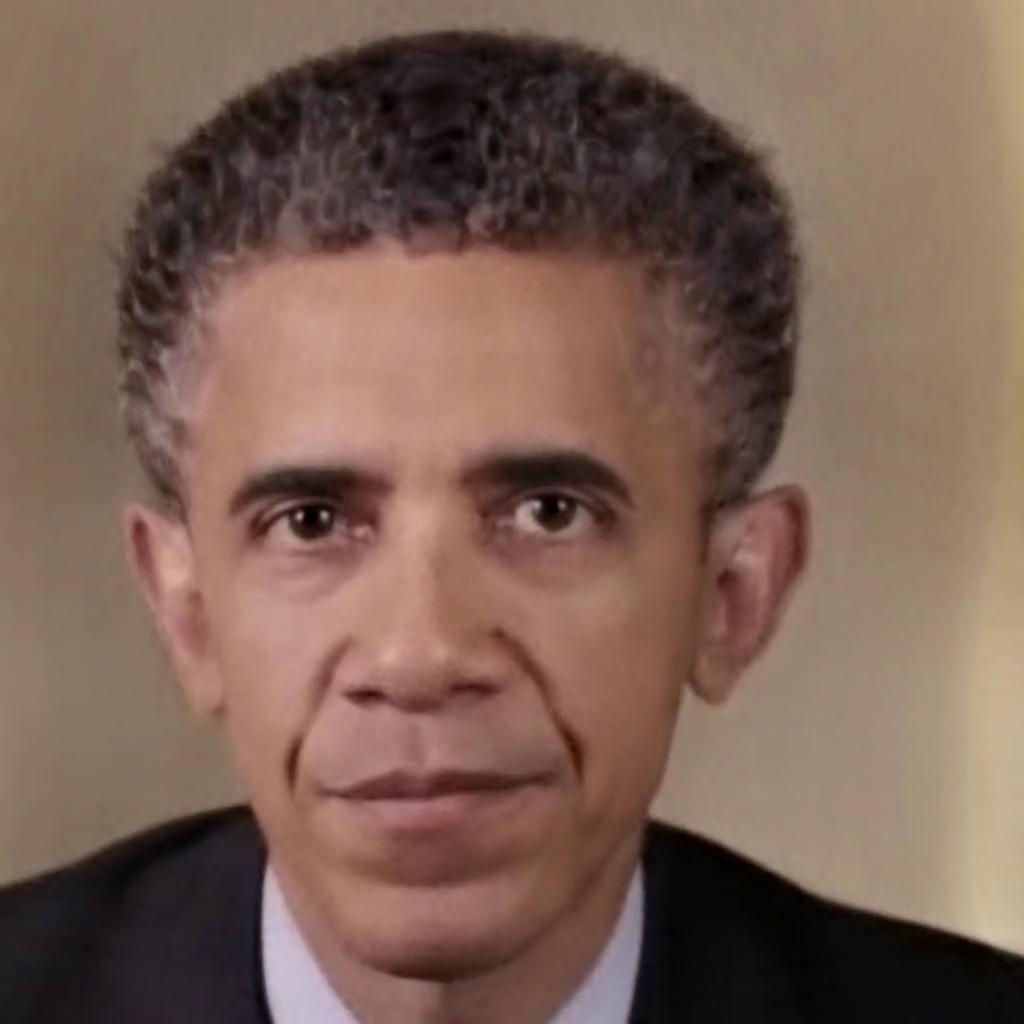} & 
        \includegraphics[width=0.215\columnwidth]{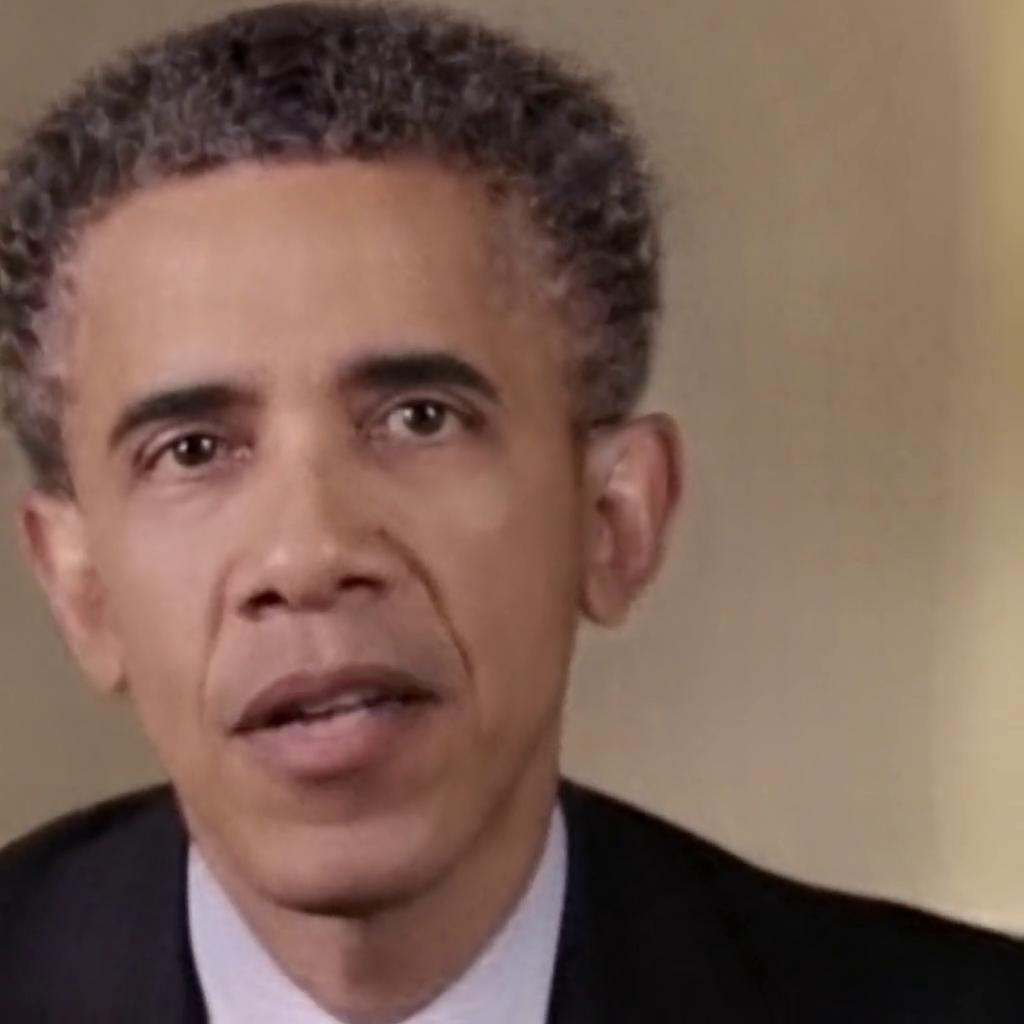} & 
        \includegraphics[width=0.215\columnwidth]{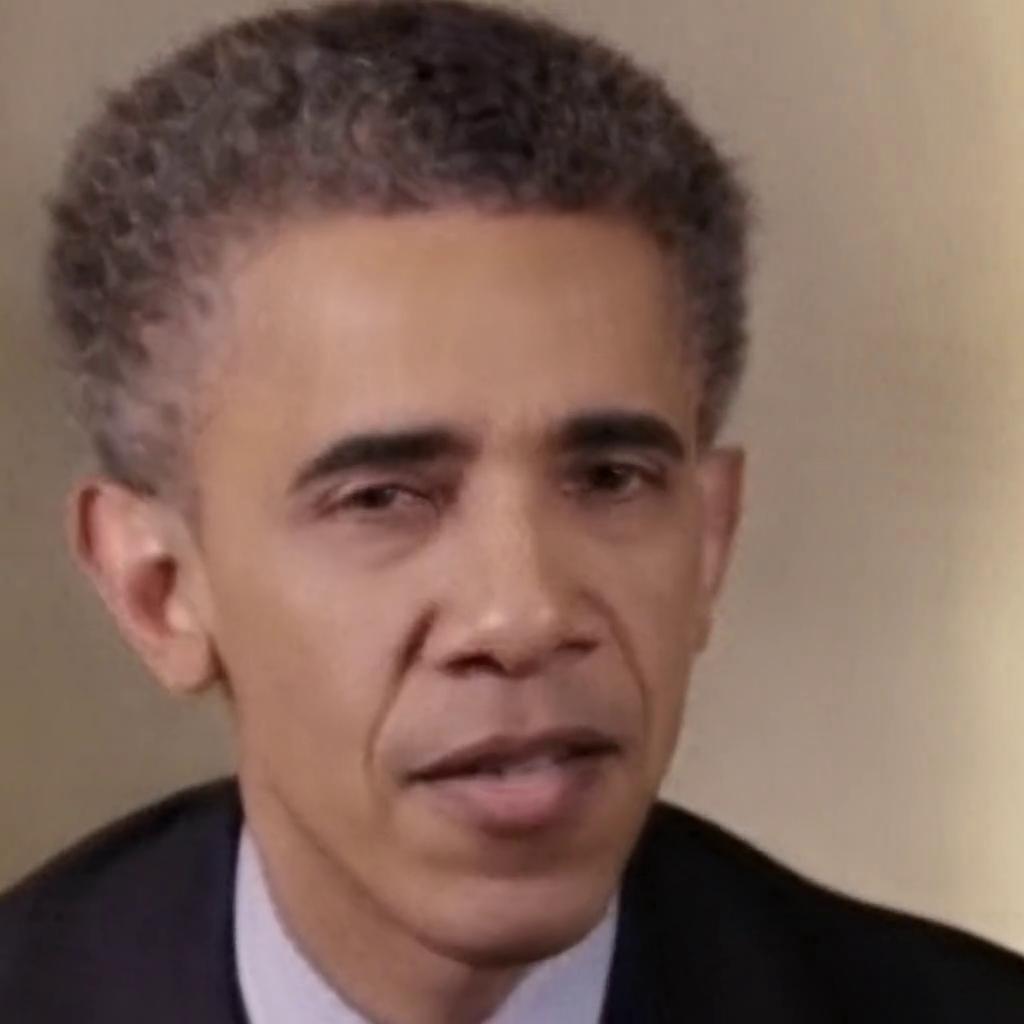} & 
        \includegraphics[width=0.215\columnwidth]{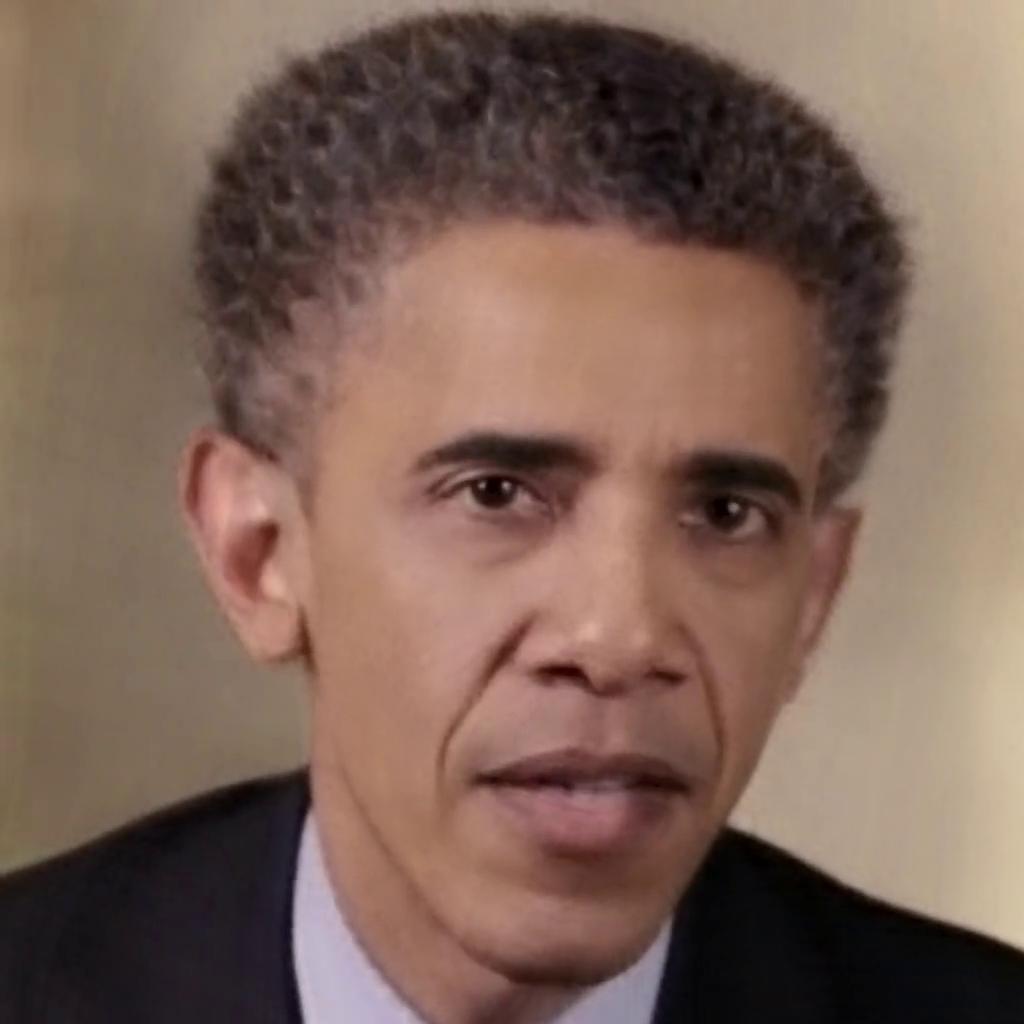} \\
        
		\raisebox{0.175in}{\rotatebox{90}{$+$ Pixar}} &
        \includegraphics[width=0.215\columnwidth]{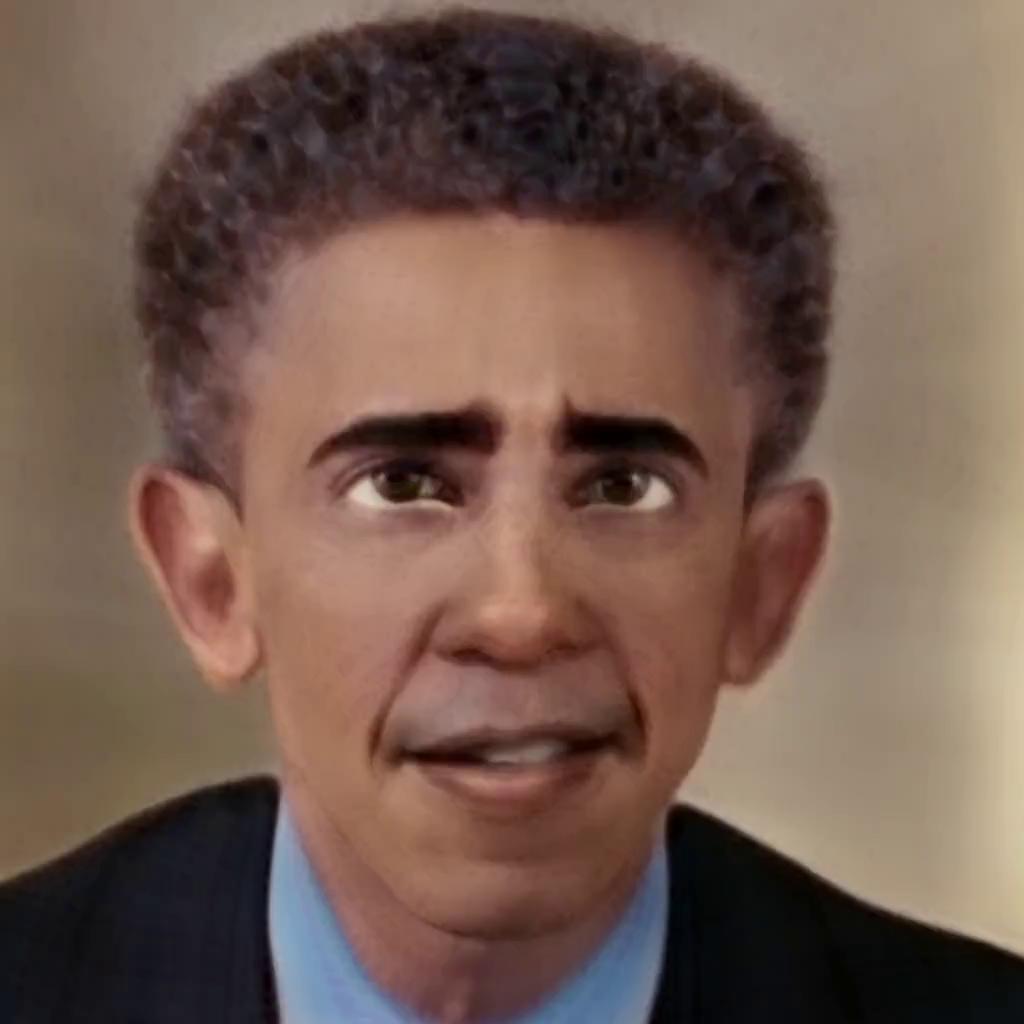} & 
        \includegraphics[width=0.215\columnwidth]{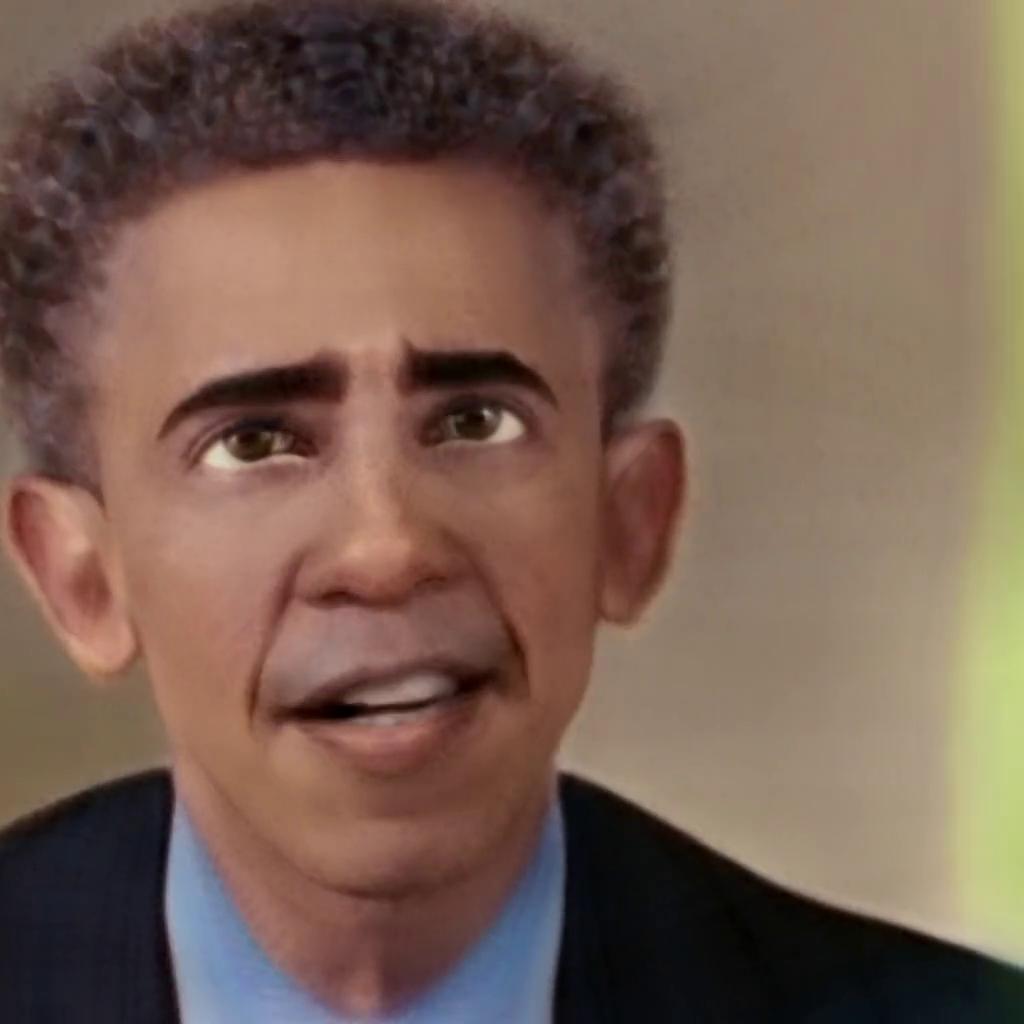} & 
        \includegraphics[width=0.215\columnwidth]{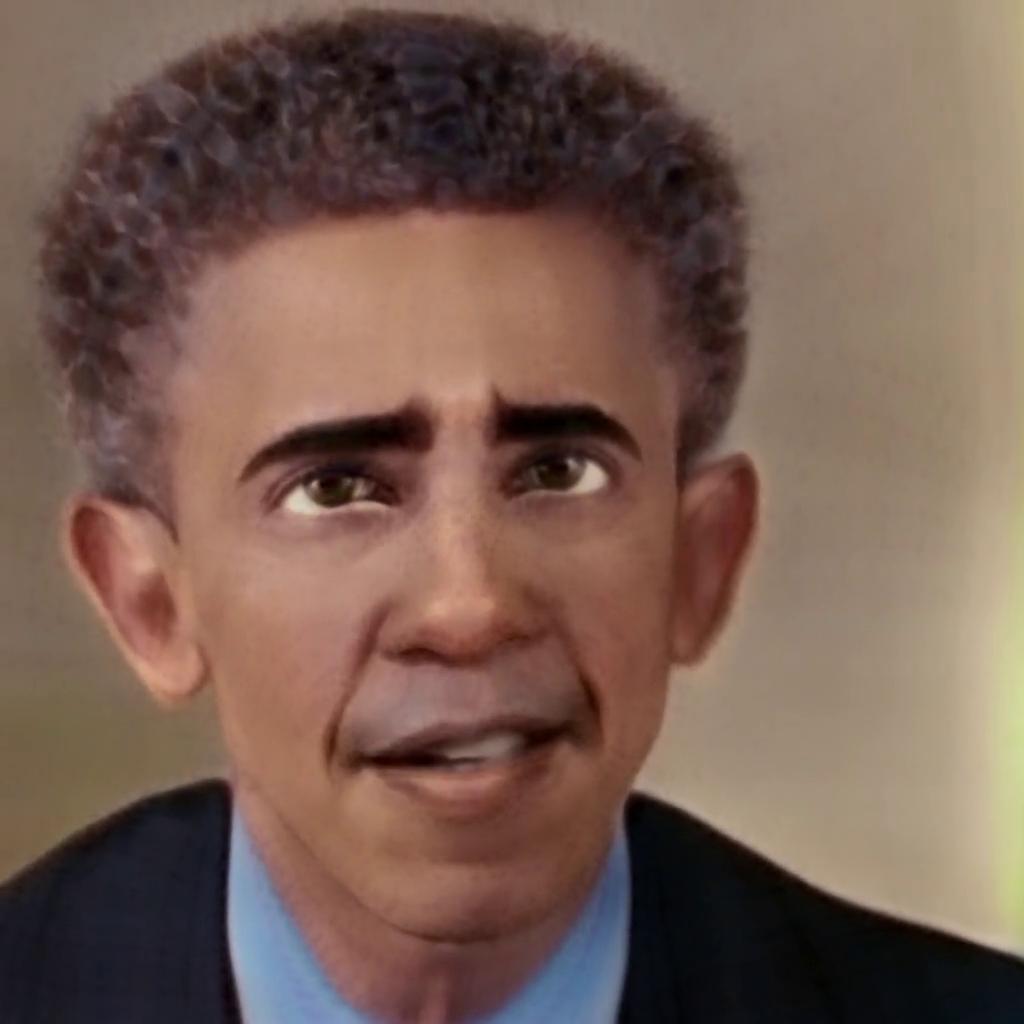} & 
        \includegraphics[width=0.215\columnwidth]{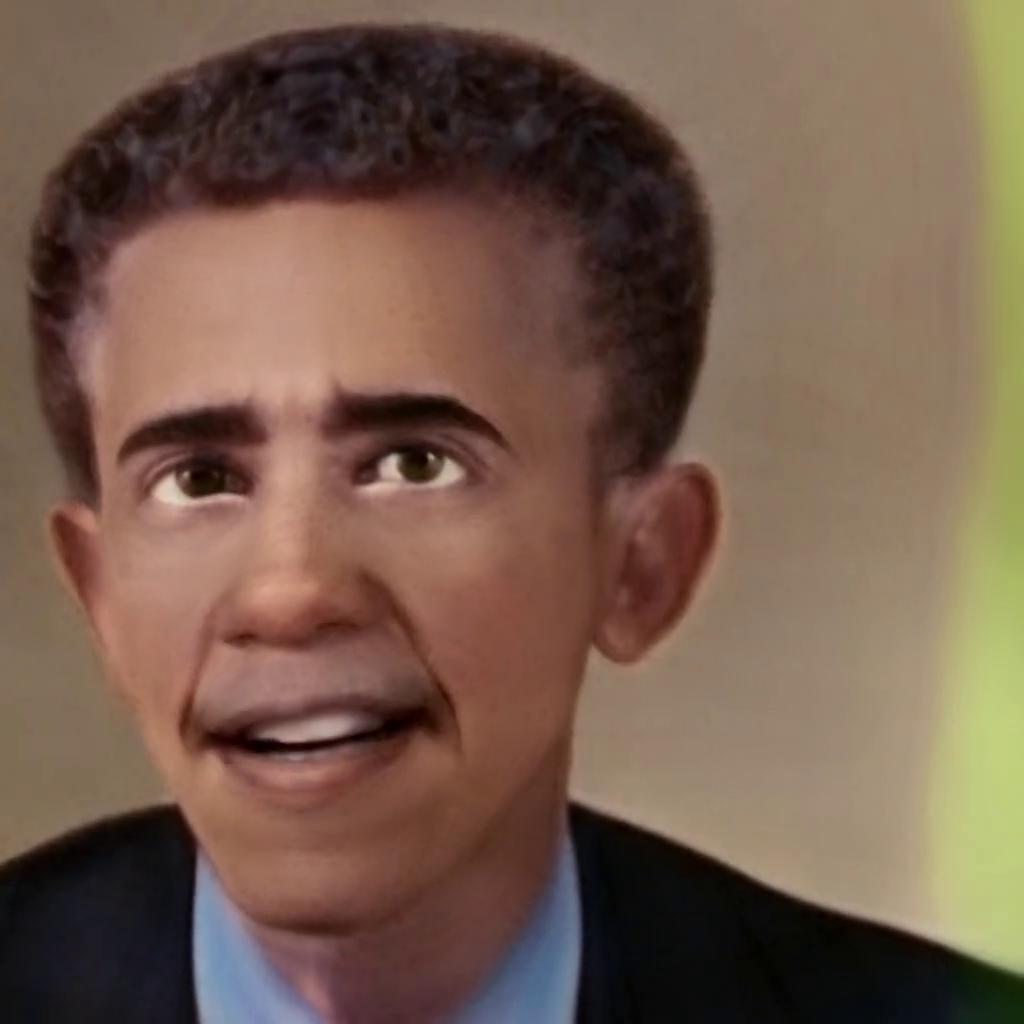} & 
        \includegraphics[width=0.215\columnwidth]{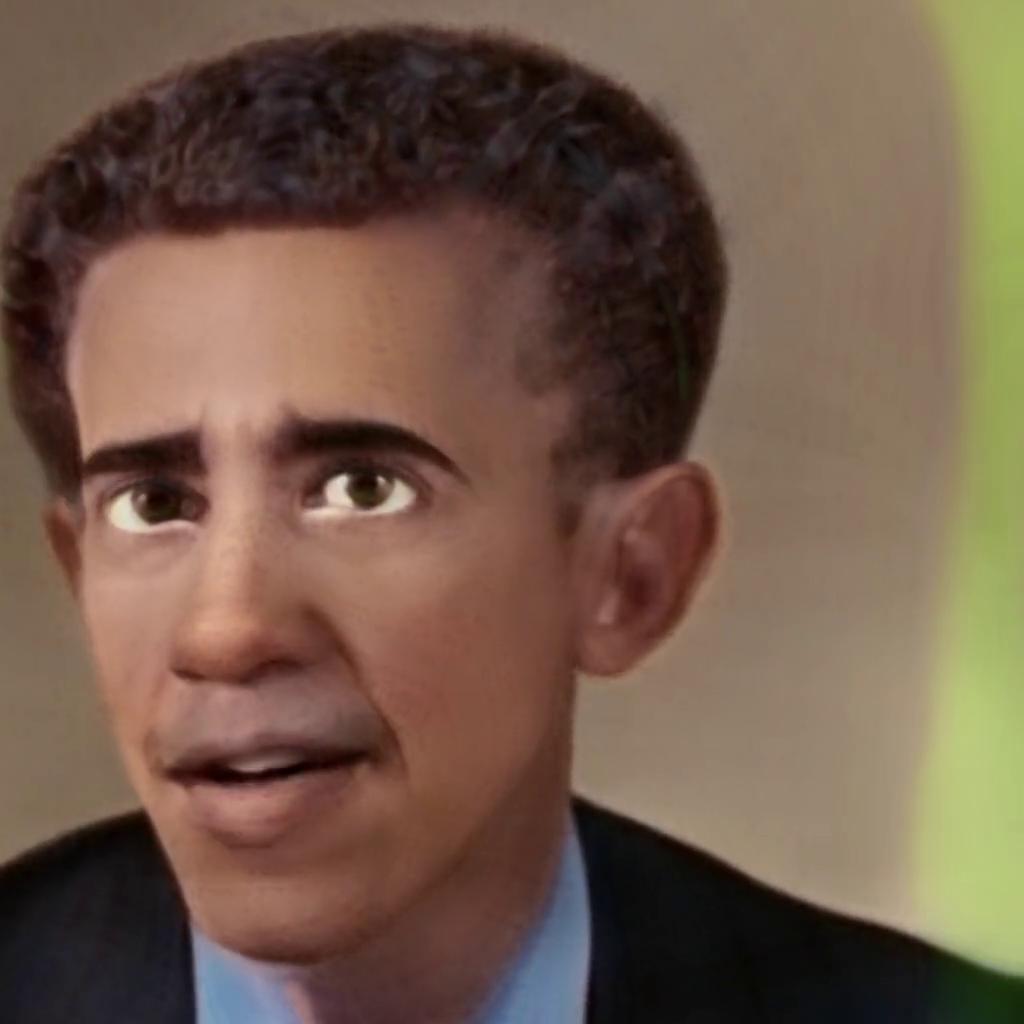} & 
        \includegraphics[width=0.215\columnwidth]{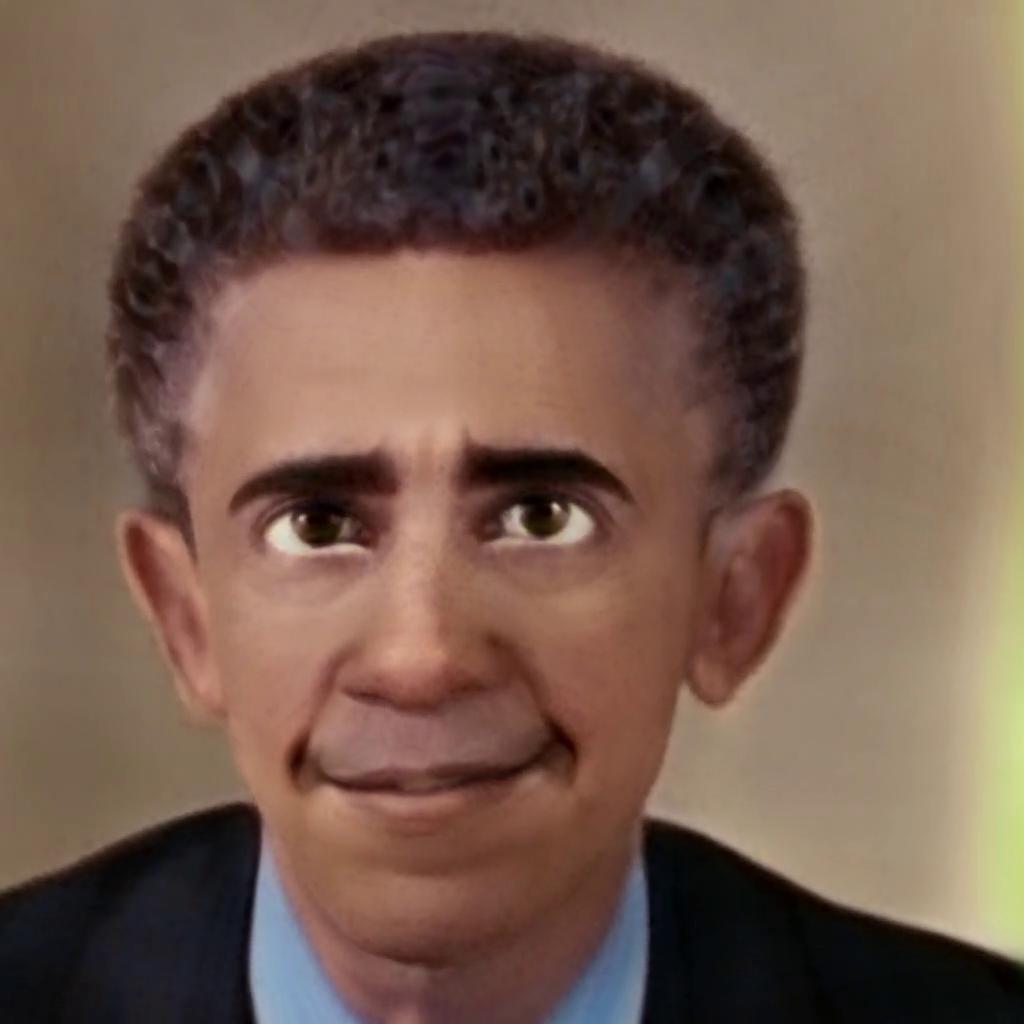} & 
        \includegraphics[width=0.215\columnwidth]{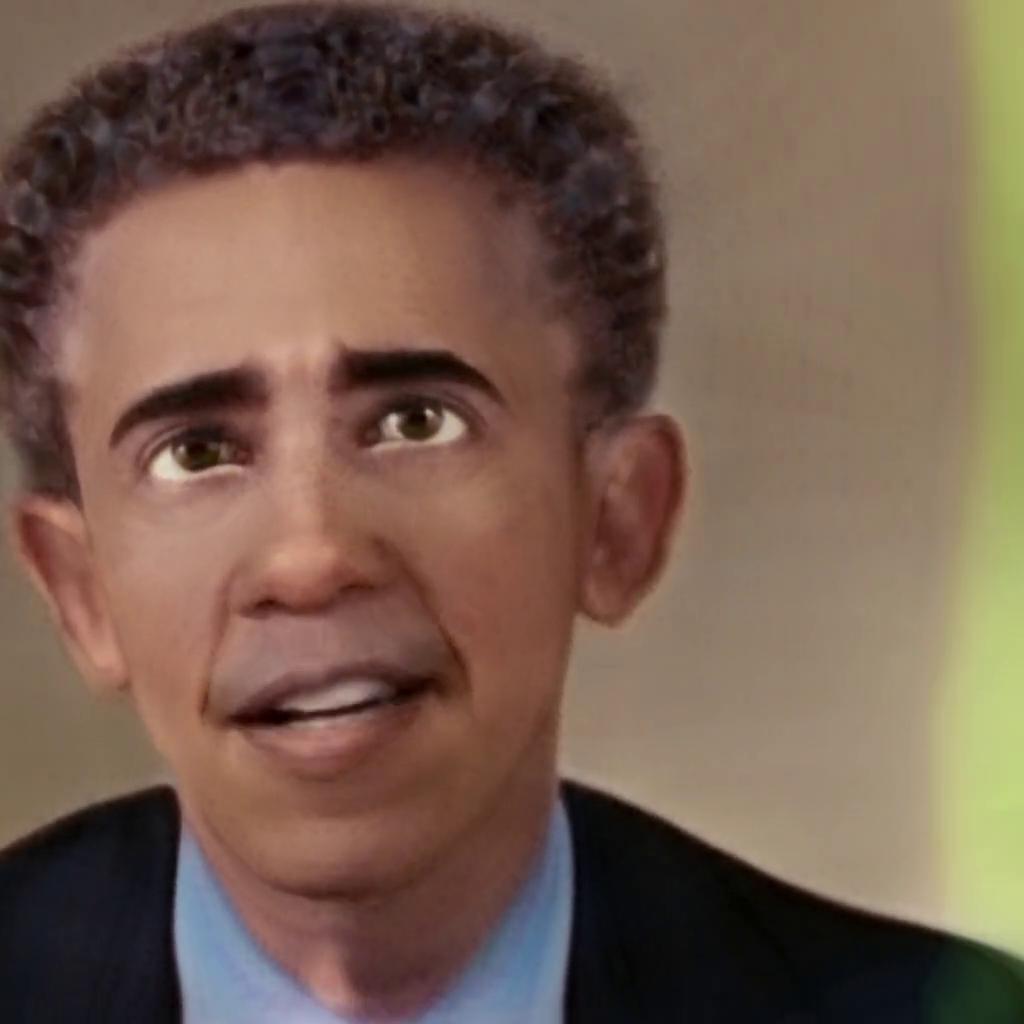} & 
        \includegraphics[width=0.215\columnwidth]{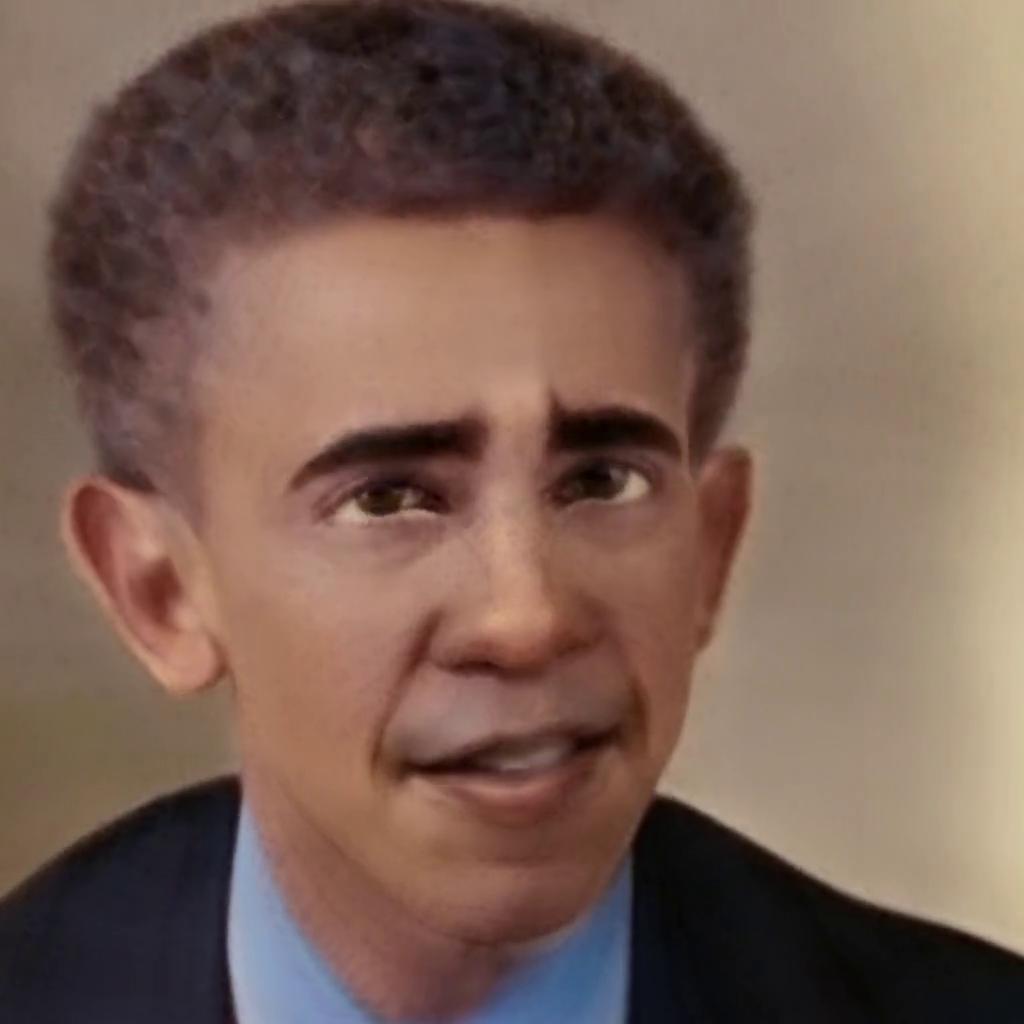} & 
        \includegraphics[width=0.215\columnwidth]{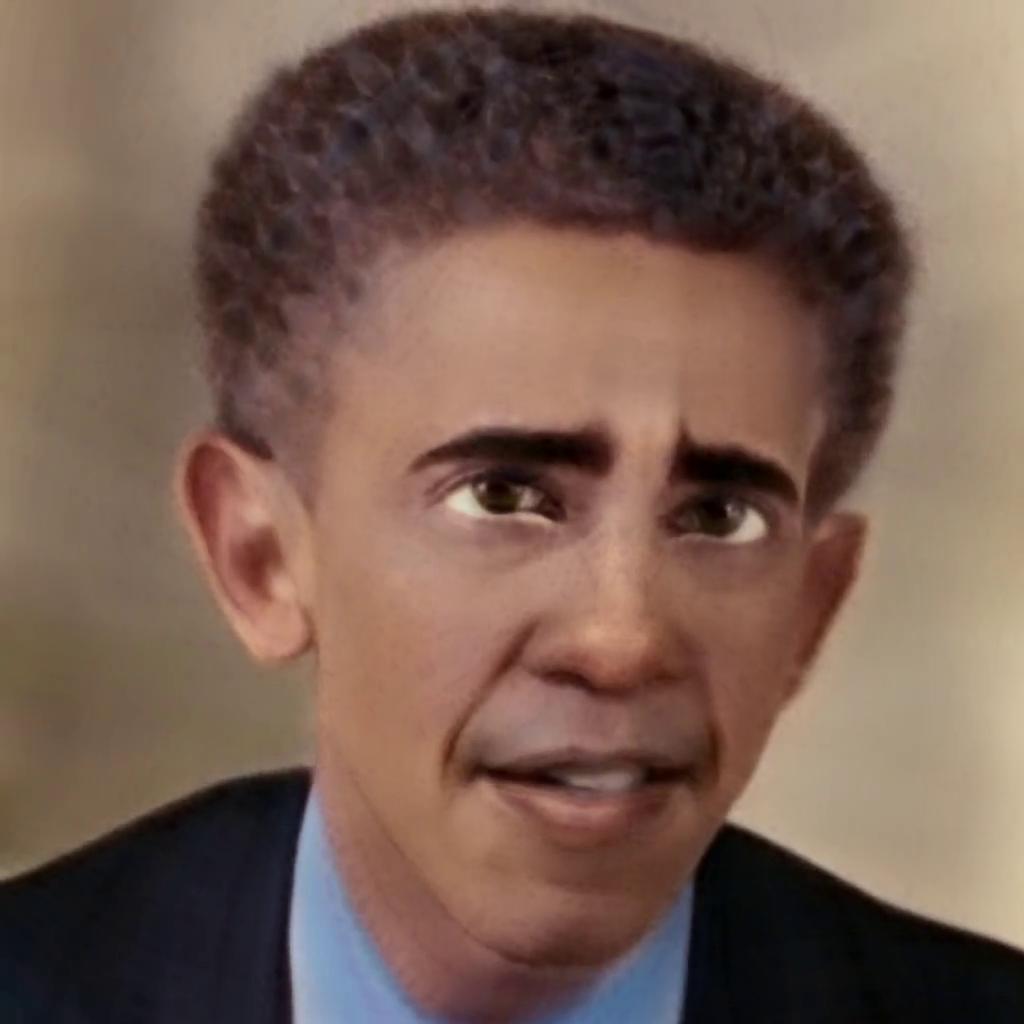} \\

	\end{tabular}
	}
	
	{\footnotesize
	\begin{tabular}{c c c c c c c c c c}
        
        \\ \\

		\raisebox{0.15in}{\rotatebox{90}{Original}} &
        \includegraphics[width=0.215\columnwidth]{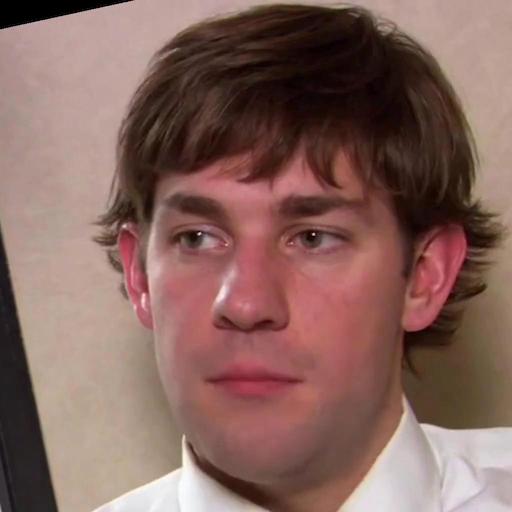} & 
        \includegraphics[width=0.215\columnwidth]{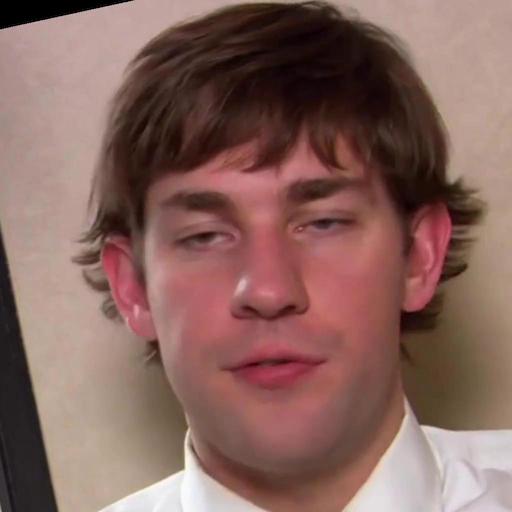} & 
        \includegraphics[width=0.215\columnwidth]{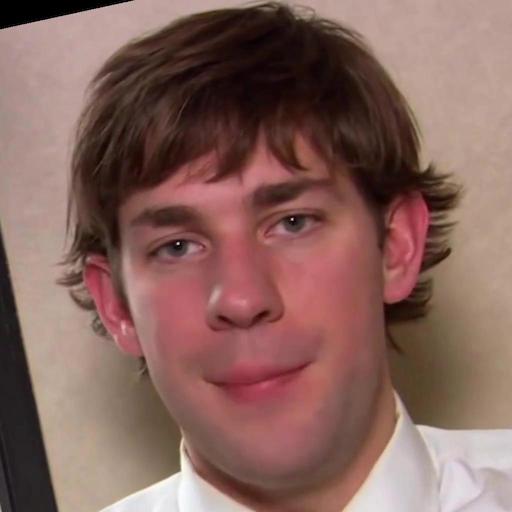} & 
        \includegraphics[width=0.215\columnwidth]{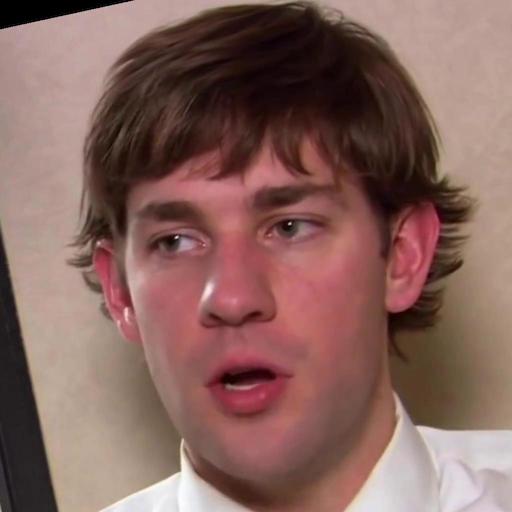} & 
        \includegraphics[width=0.215\columnwidth]{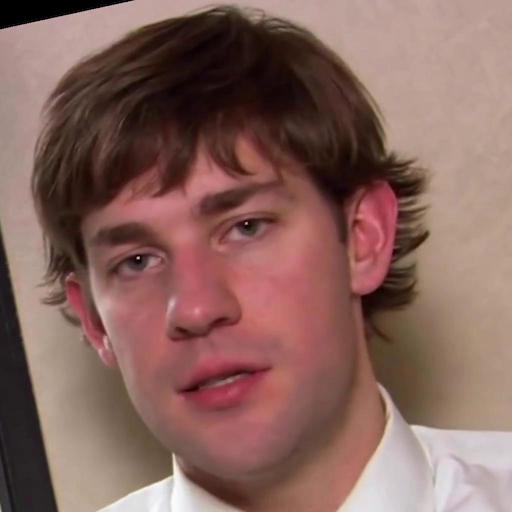} & 
        \includegraphics[width=0.215\columnwidth]{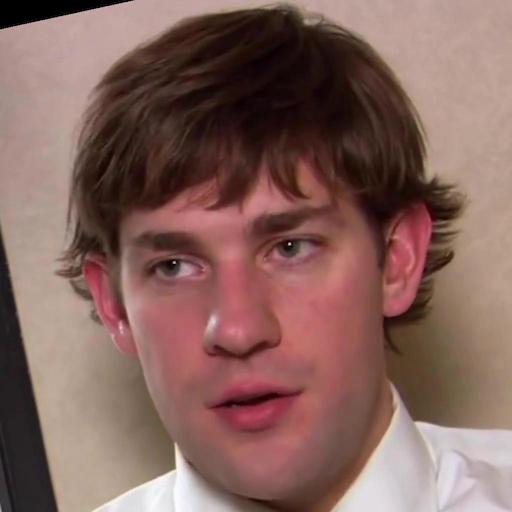} & 
        \includegraphics[width=0.215\columnwidth]{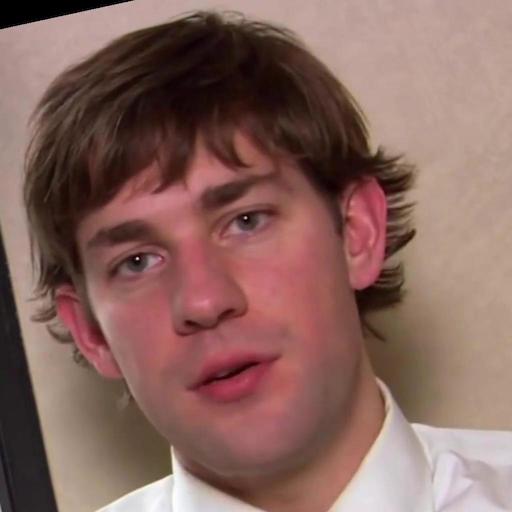} & 
        \includegraphics[width=0.215\columnwidth]{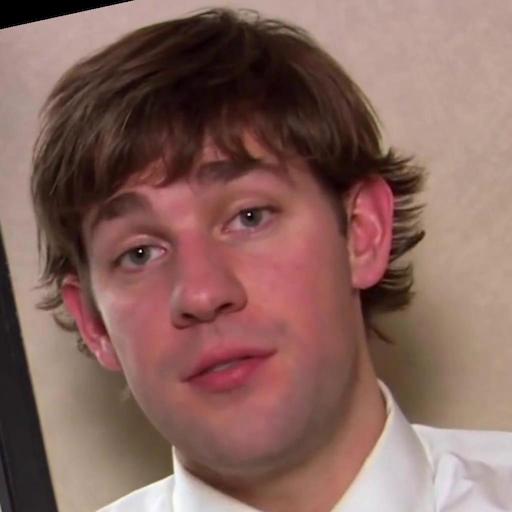} & 
        \includegraphics[width=0.215\columnwidth]{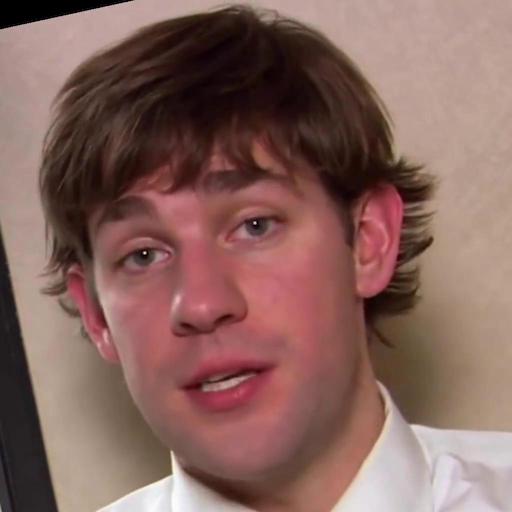} \\

		\raisebox{0.05in}{\rotatebox{90}{Reconstruction}} &
        \includegraphics[width=0.215\columnwidth]{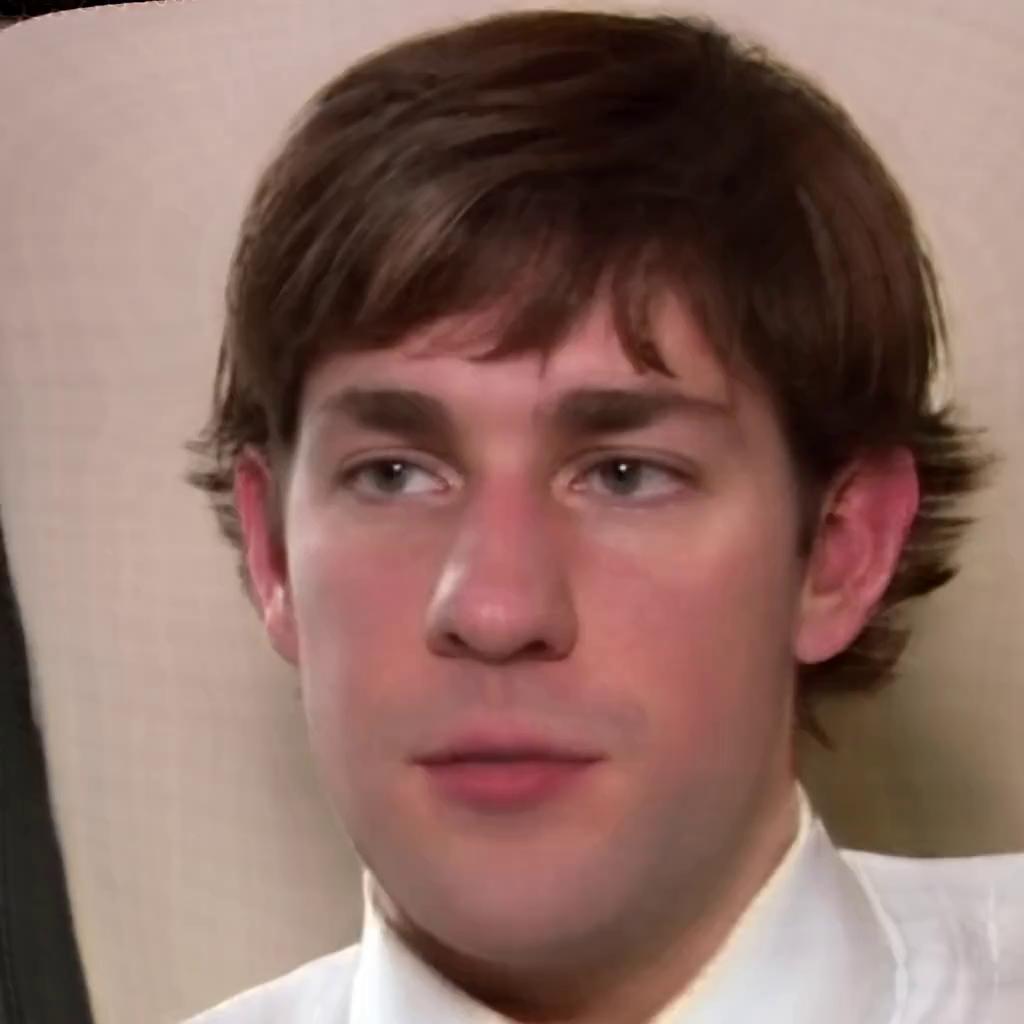} & 
        \includegraphics[width=0.215\columnwidth]{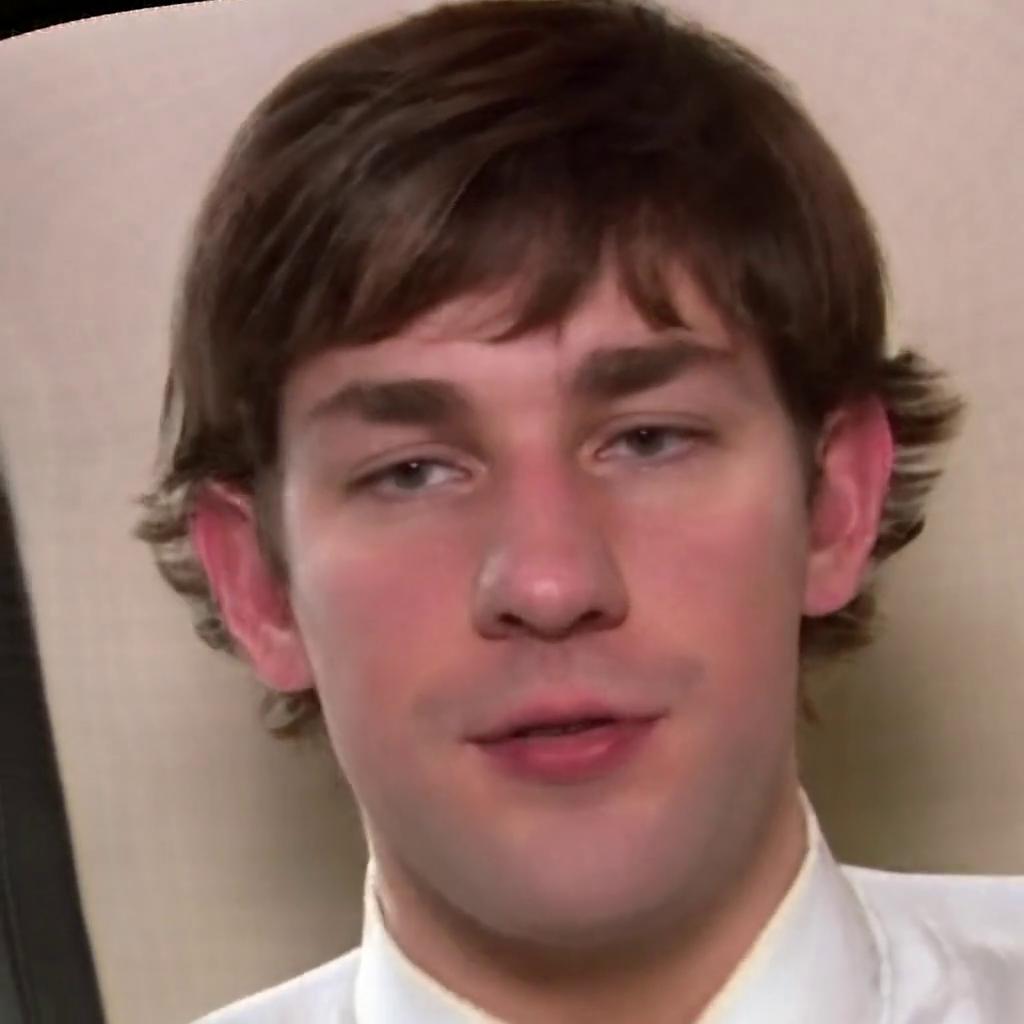} & 
        \includegraphics[width=0.215\columnwidth]{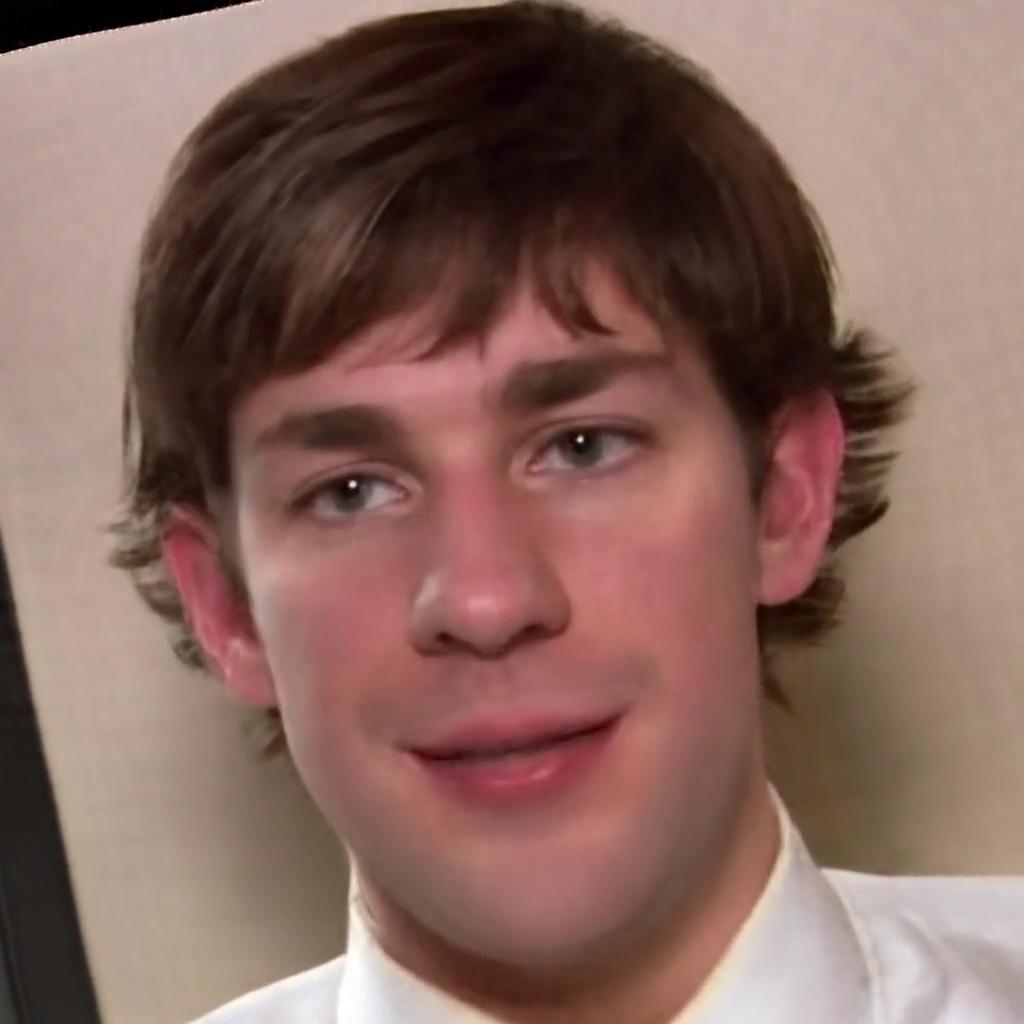} & 
        \includegraphics[width=0.215\columnwidth]{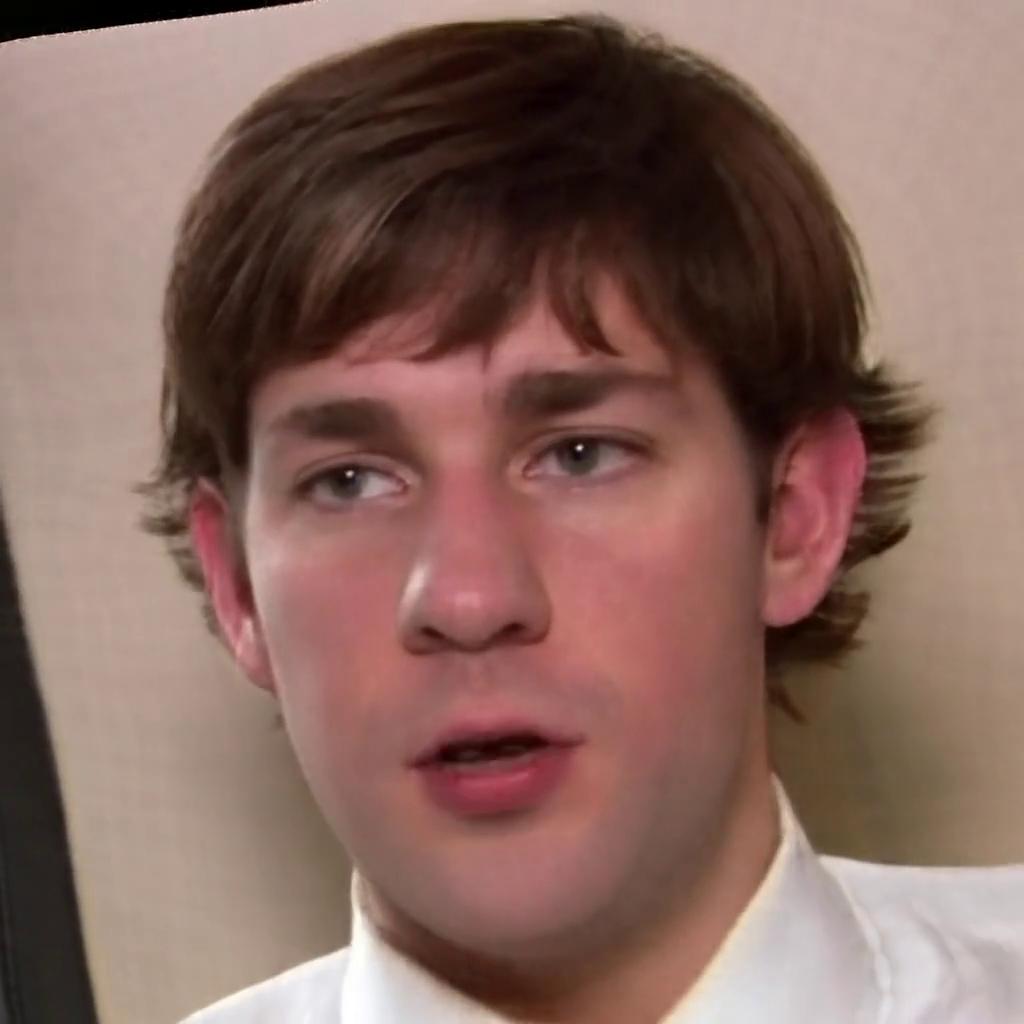} & 
        \includegraphics[width=0.215\columnwidth]{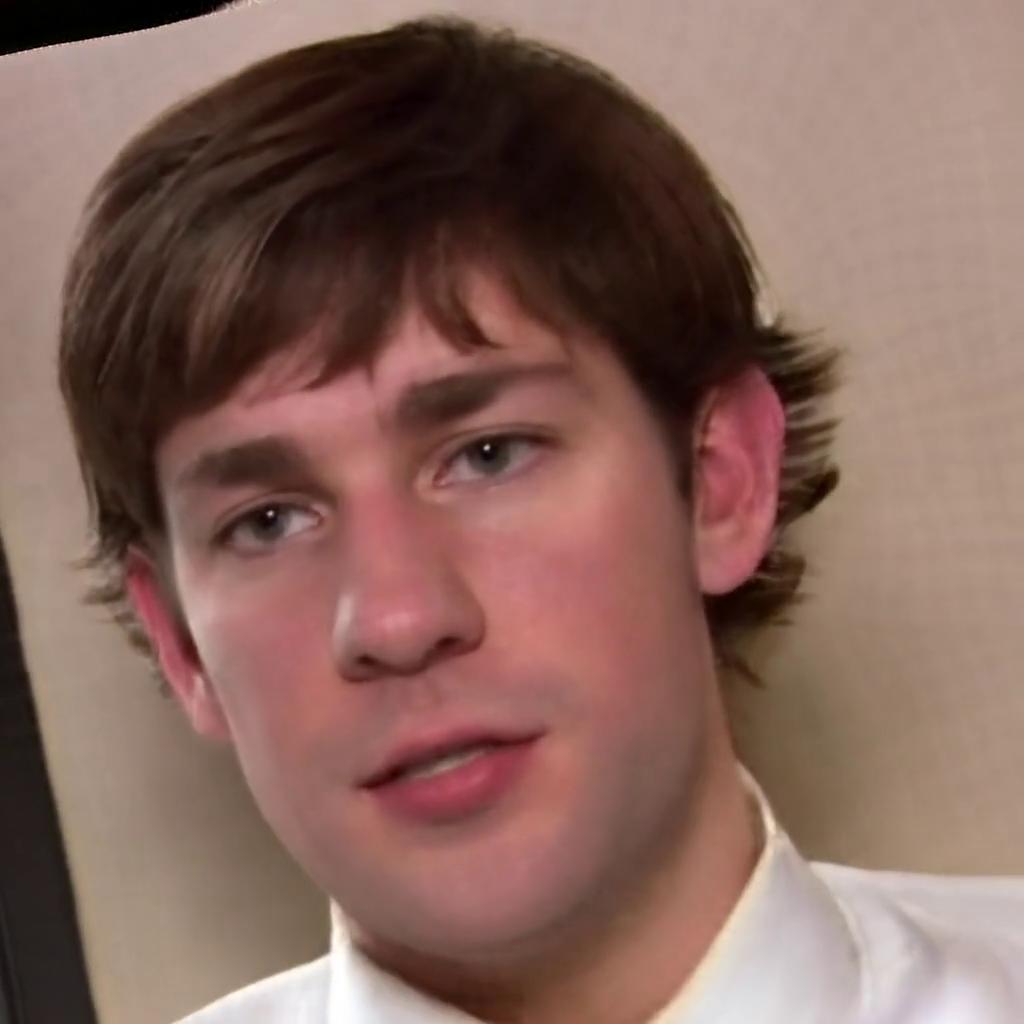} & 
        \includegraphics[width=0.215\columnwidth]{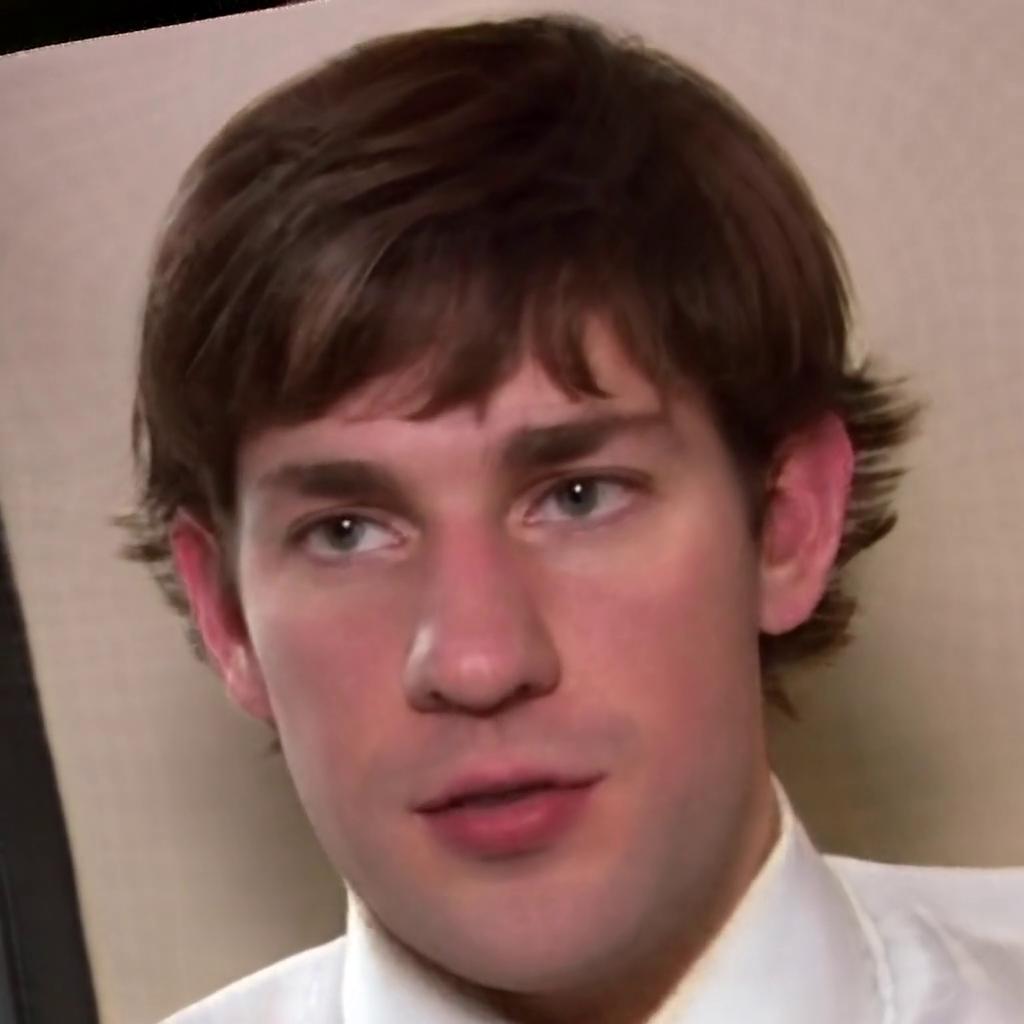} & 
        \includegraphics[width=0.215\columnwidth]{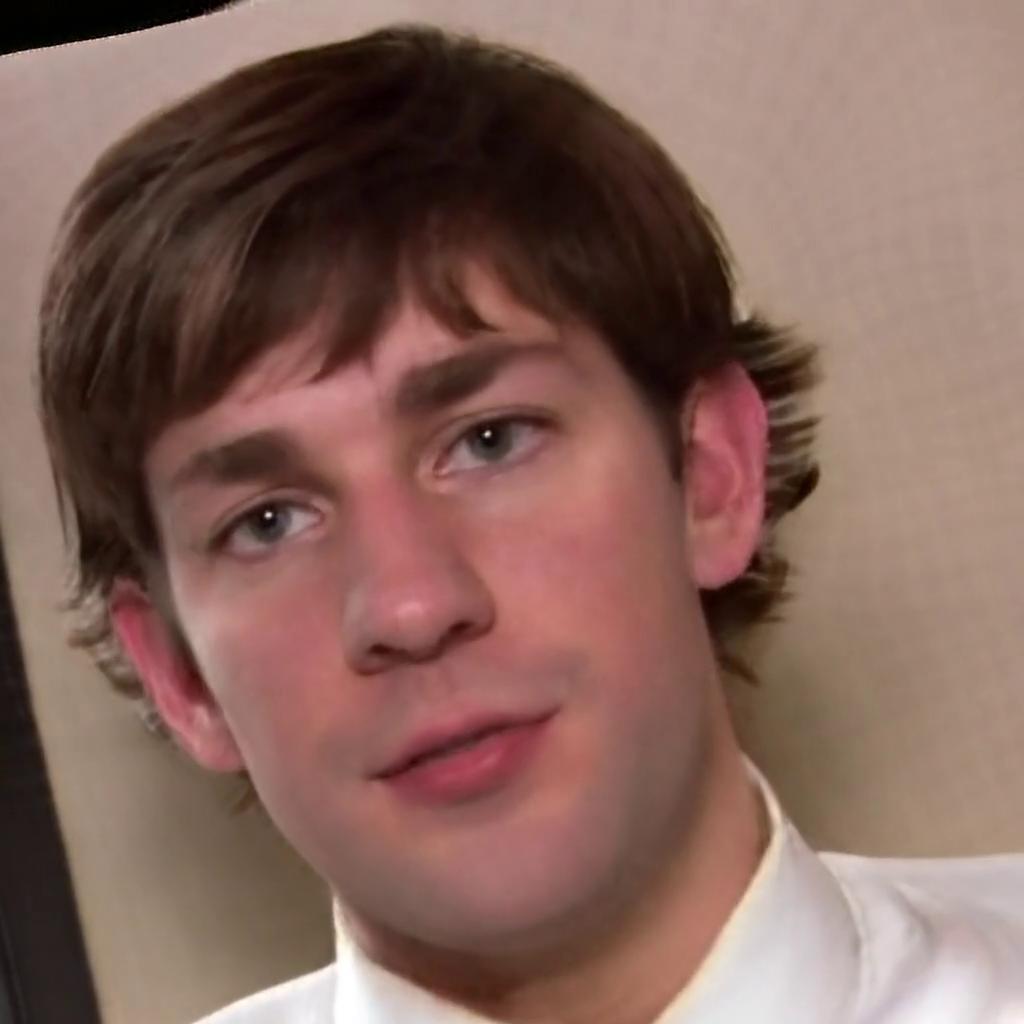} & 
        \includegraphics[width=0.215\columnwidth]{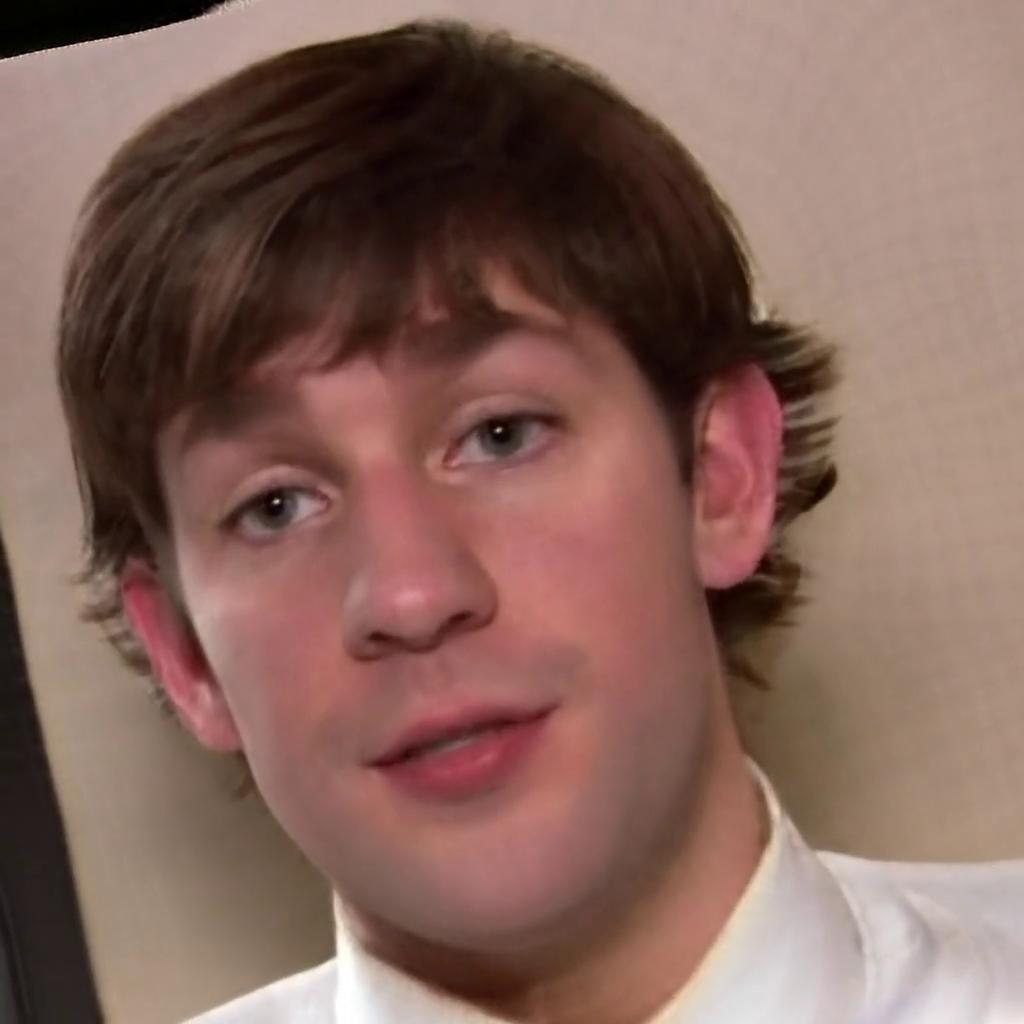} & 
        \includegraphics[width=0.215\columnwidth]{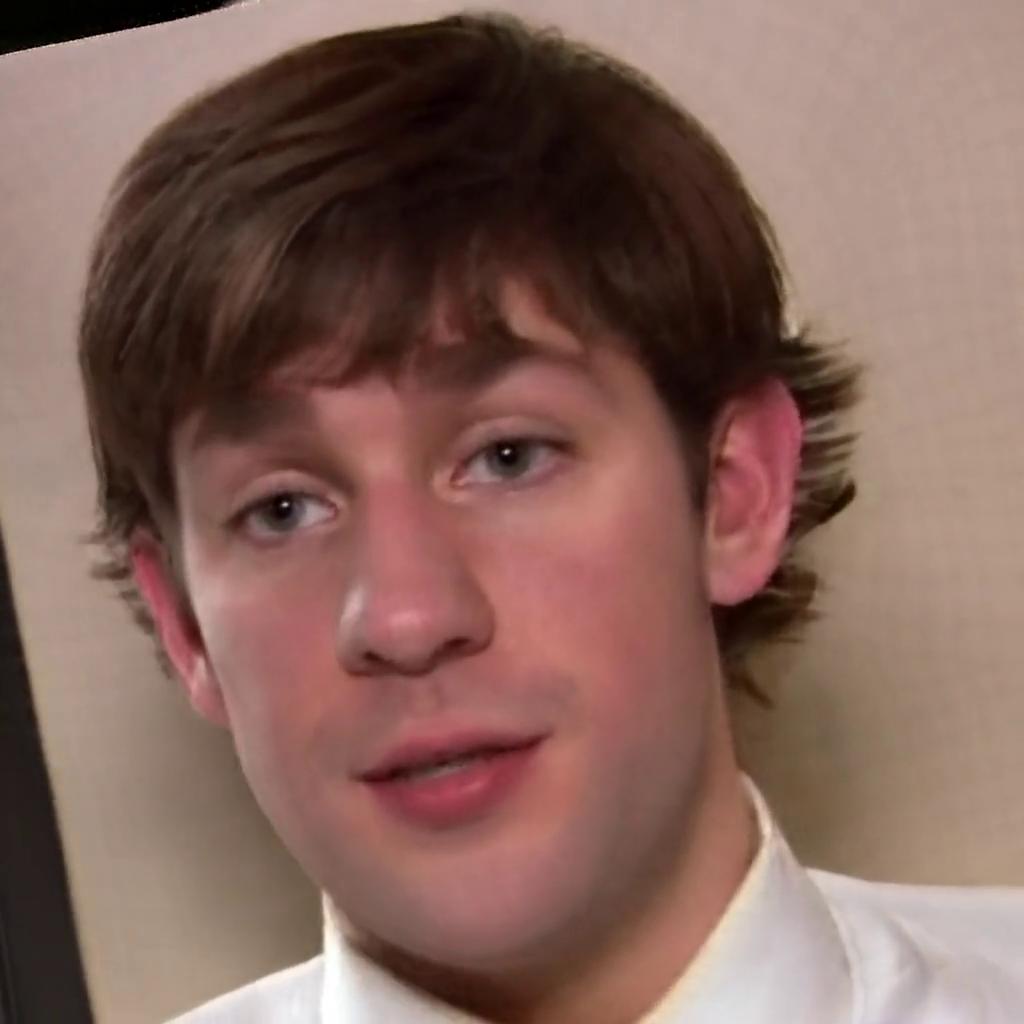} \\

		\raisebox{0.175in}{\rotatebox{90}{$-$ Age}} &
        \includegraphics[width=0.215\columnwidth]{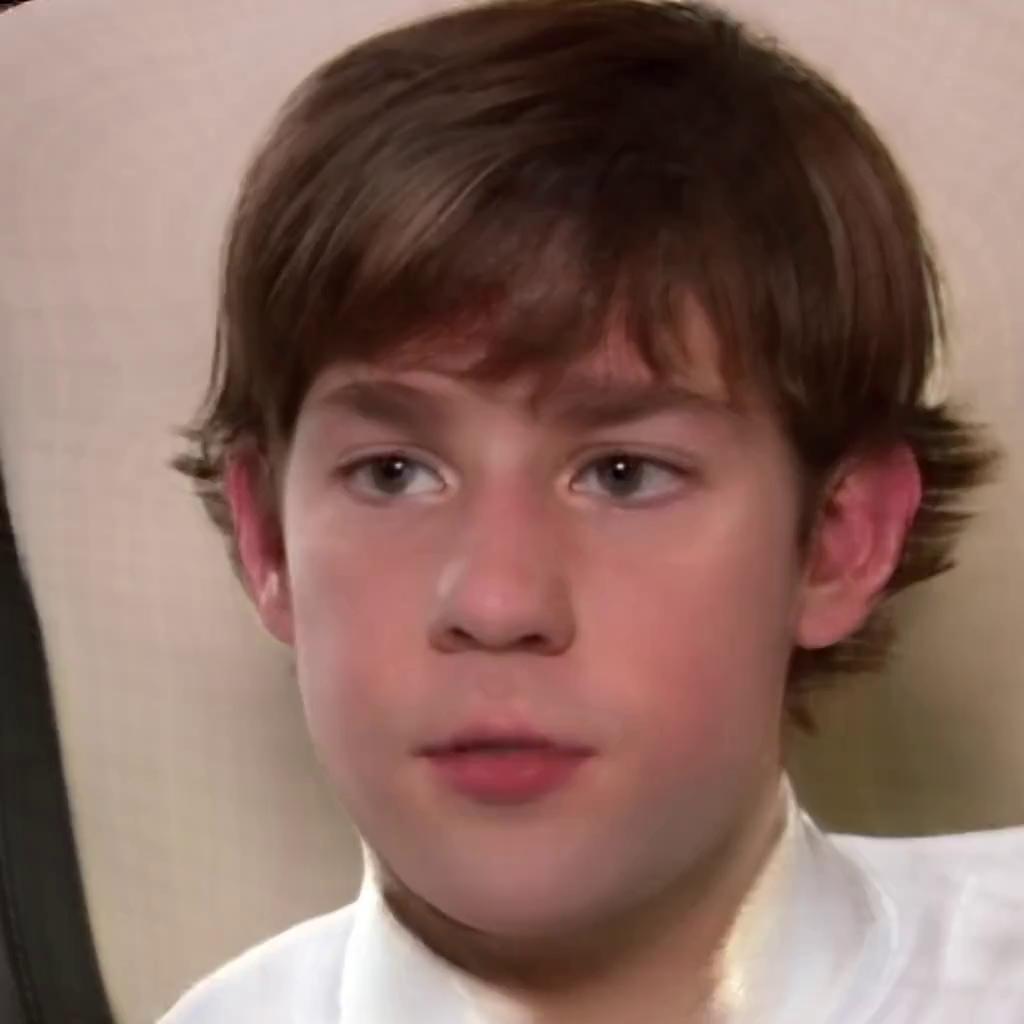} & 
        \includegraphics[width=0.215\columnwidth]{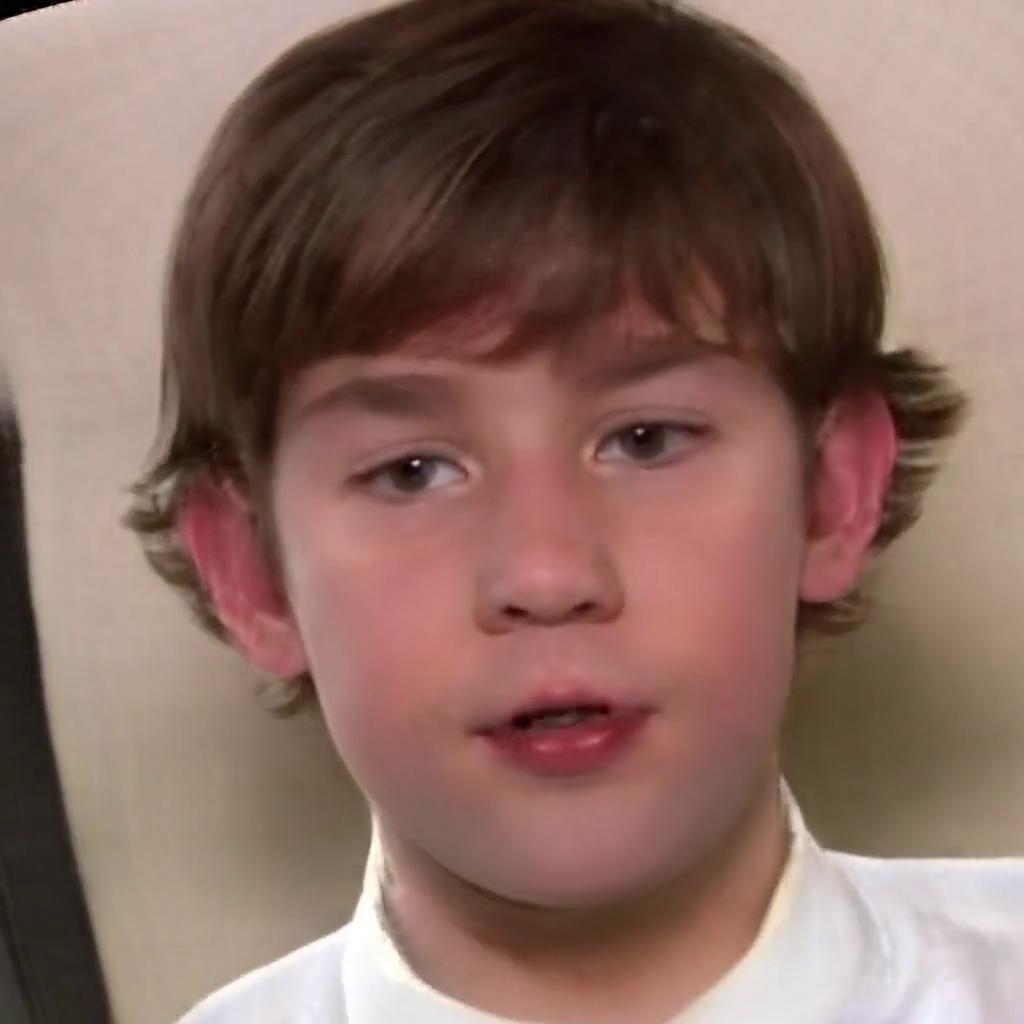} & 
        \includegraphics[width=0.215\columnwidth]{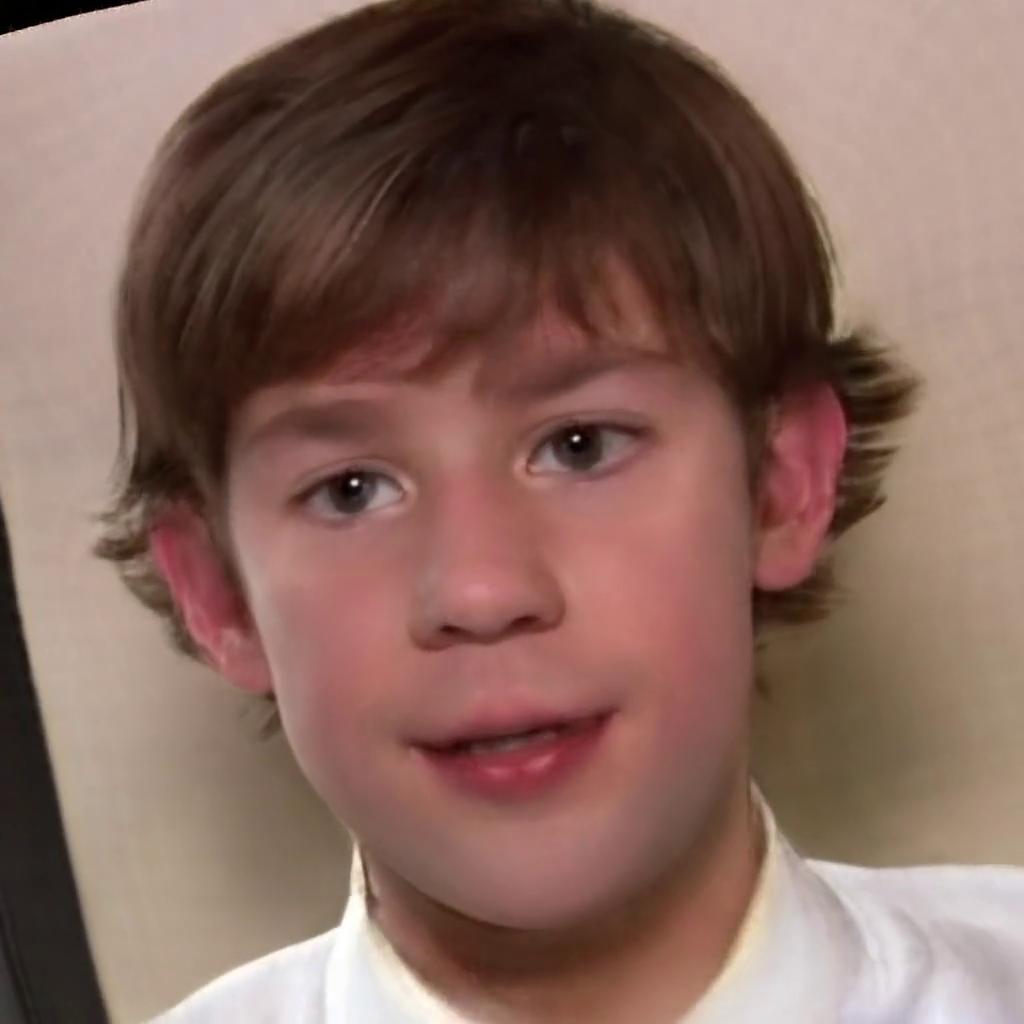} & 
        \includegraphics[width=0.215\columnwidth]{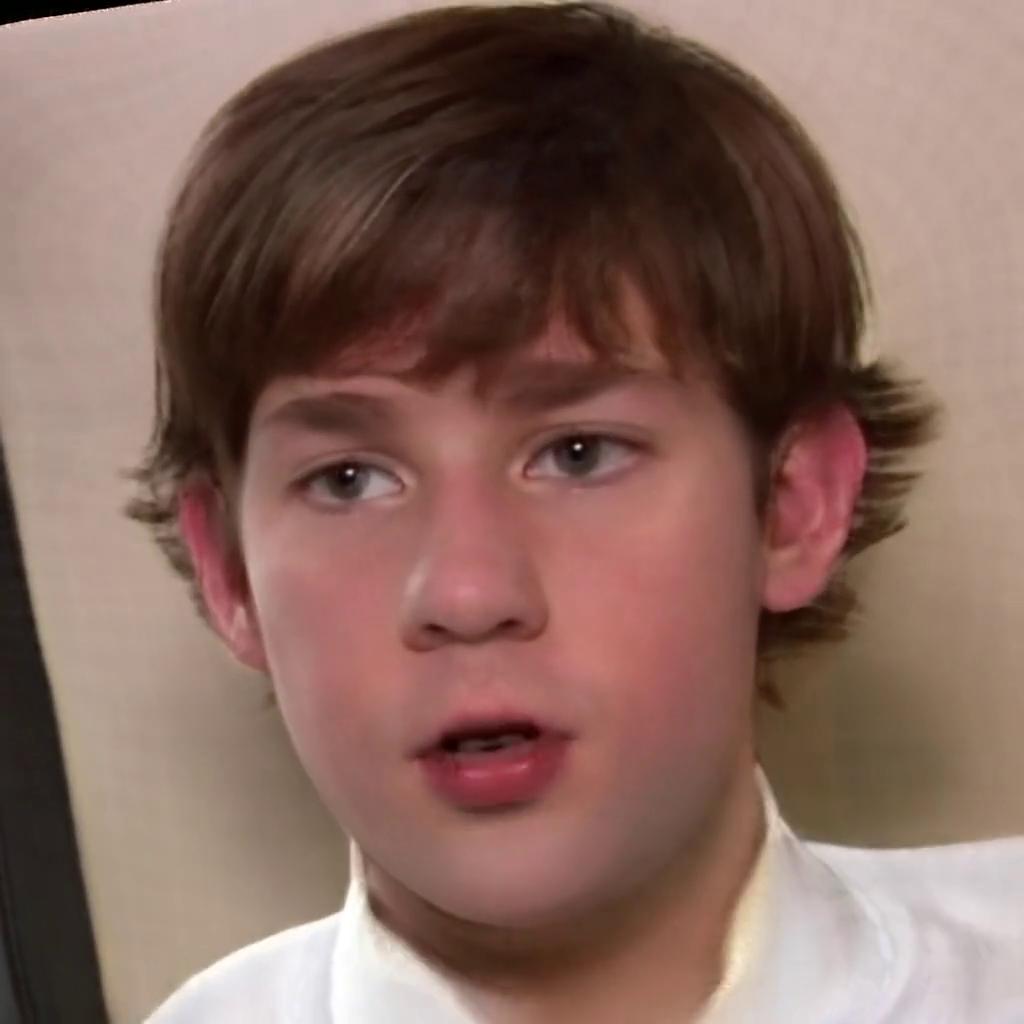} & 
        \includegraphics[width=0.215\columnwidth]{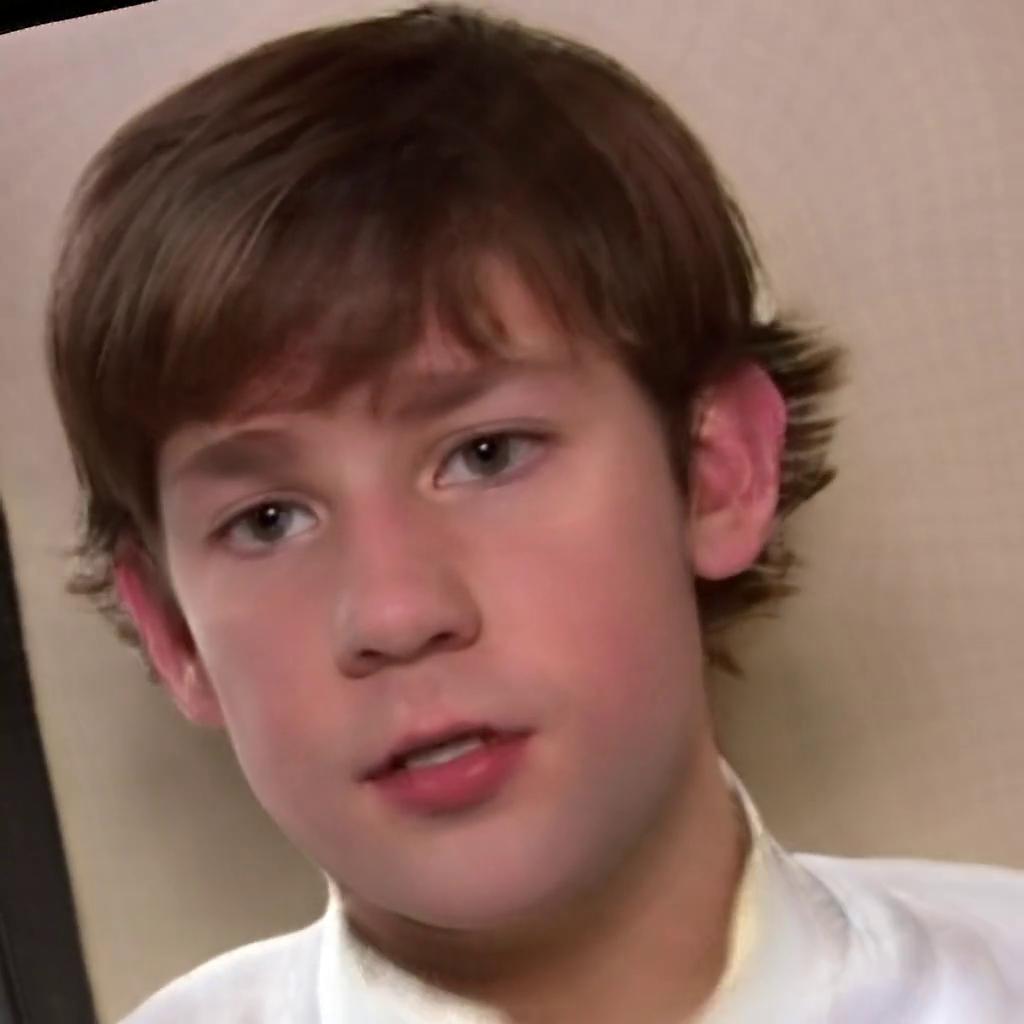} & 
        \includegraphics[width=0.215\columnwidth]{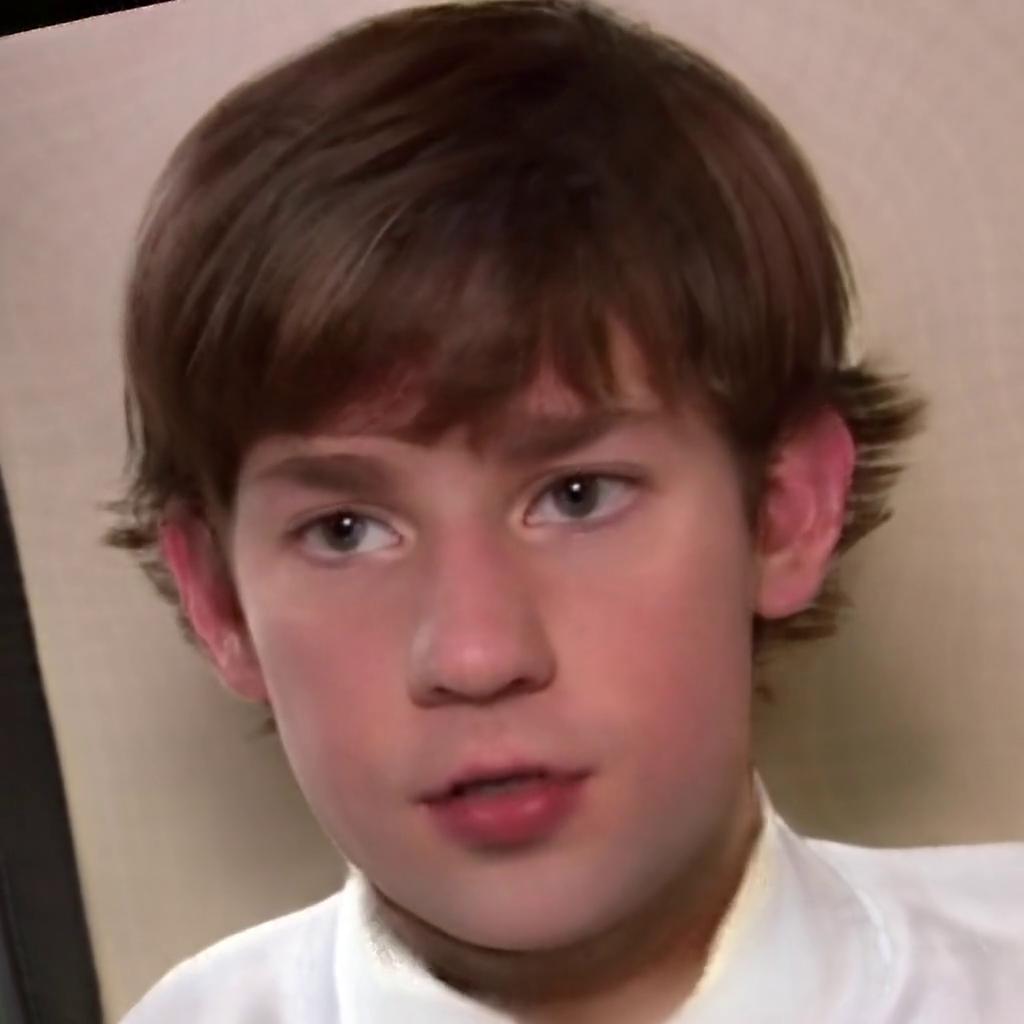} & 
        \includegraphics[width=0.215\columnwidth]{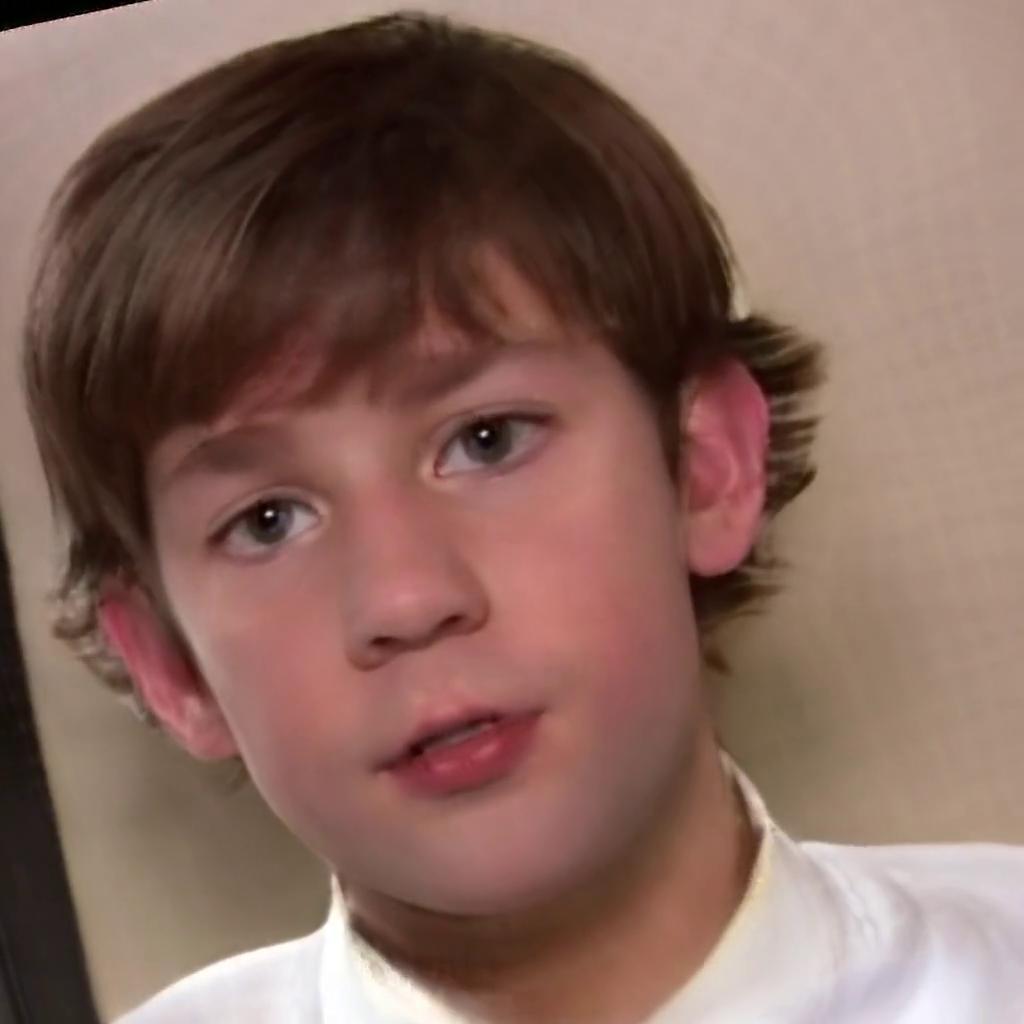} & 
        \includegraphics[width=0.215\columnwidth]{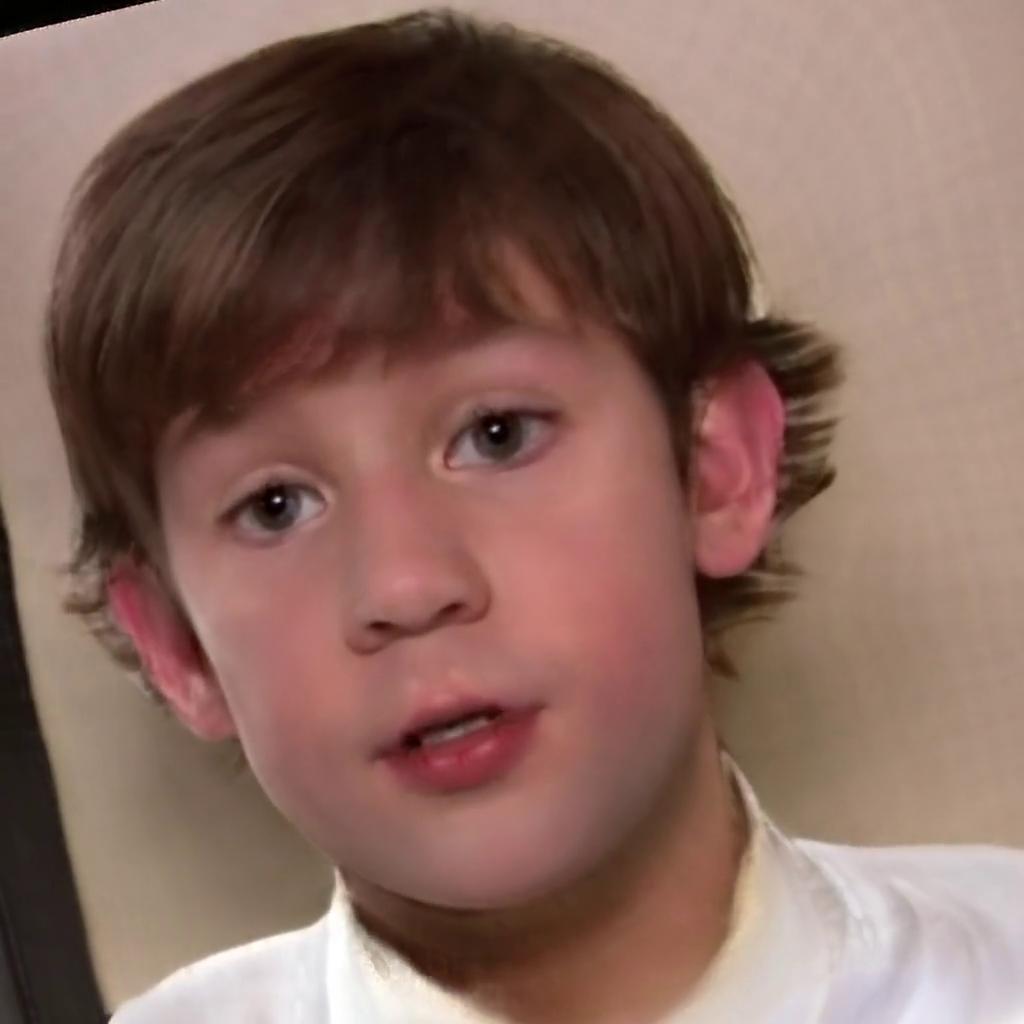} & 
        \includegraphics[width=0.215\columnwidth]{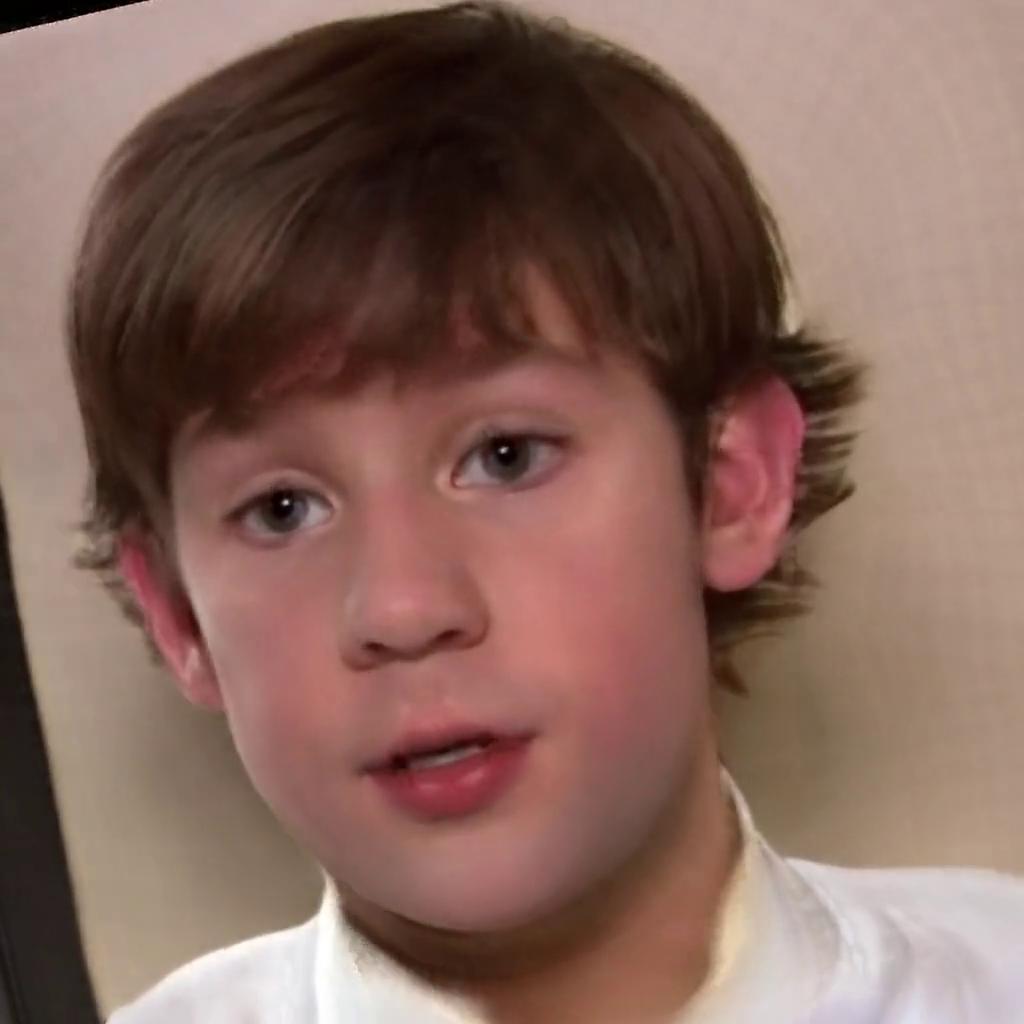} \\
        
		\raisebox{0.15in}{\rotatebox{90}{$+$ Sketch}} &
        \includegraphics[width=0.215\columnwidth]{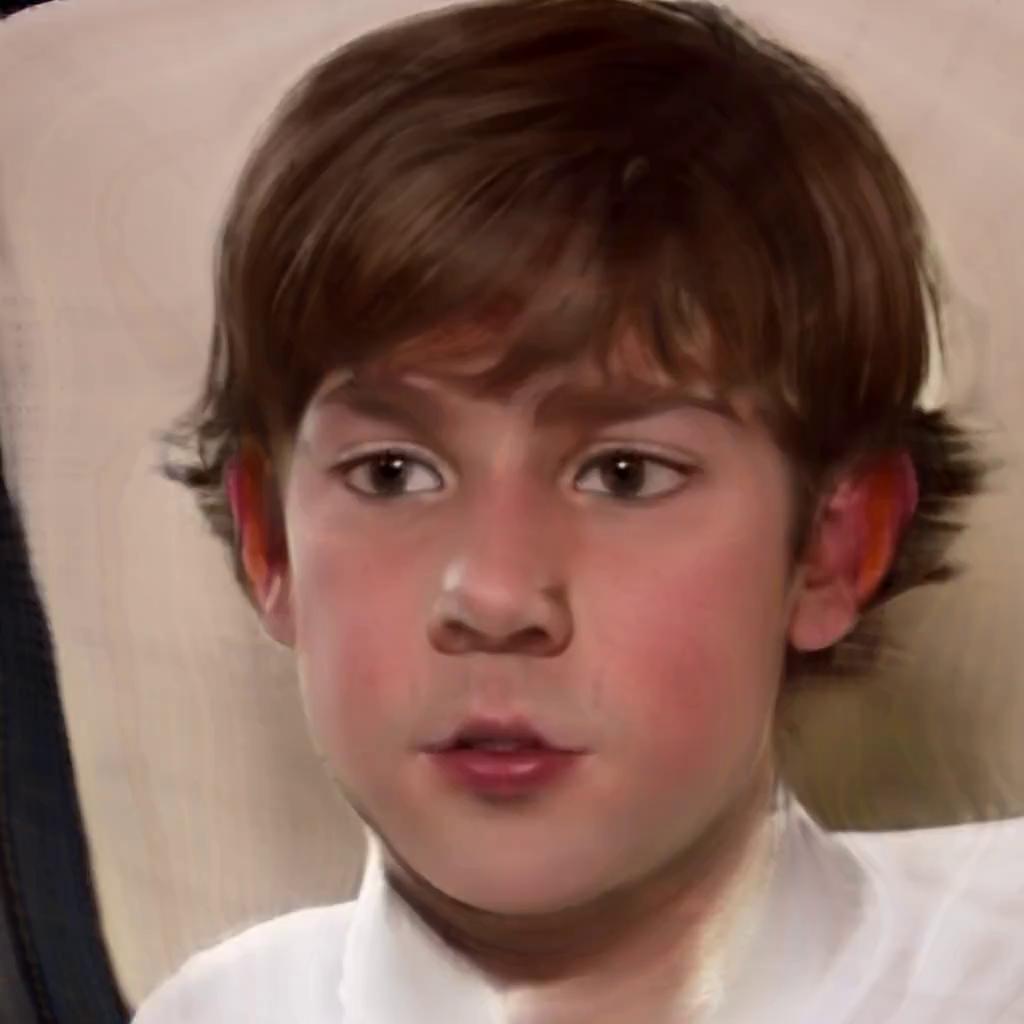} & 
        \includegraphics[width=0.215\columnwidth]{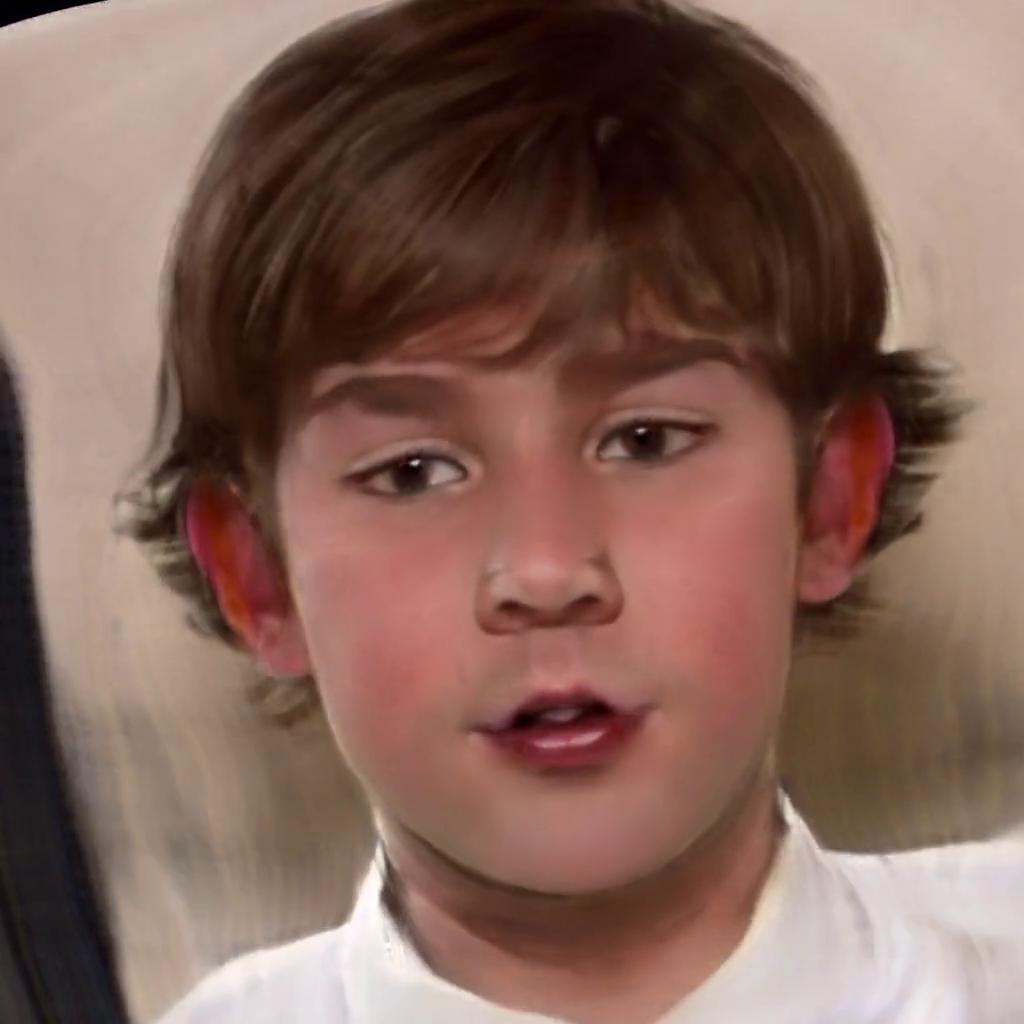} & 
        \includegraphics[width=0.215\columnwidth]{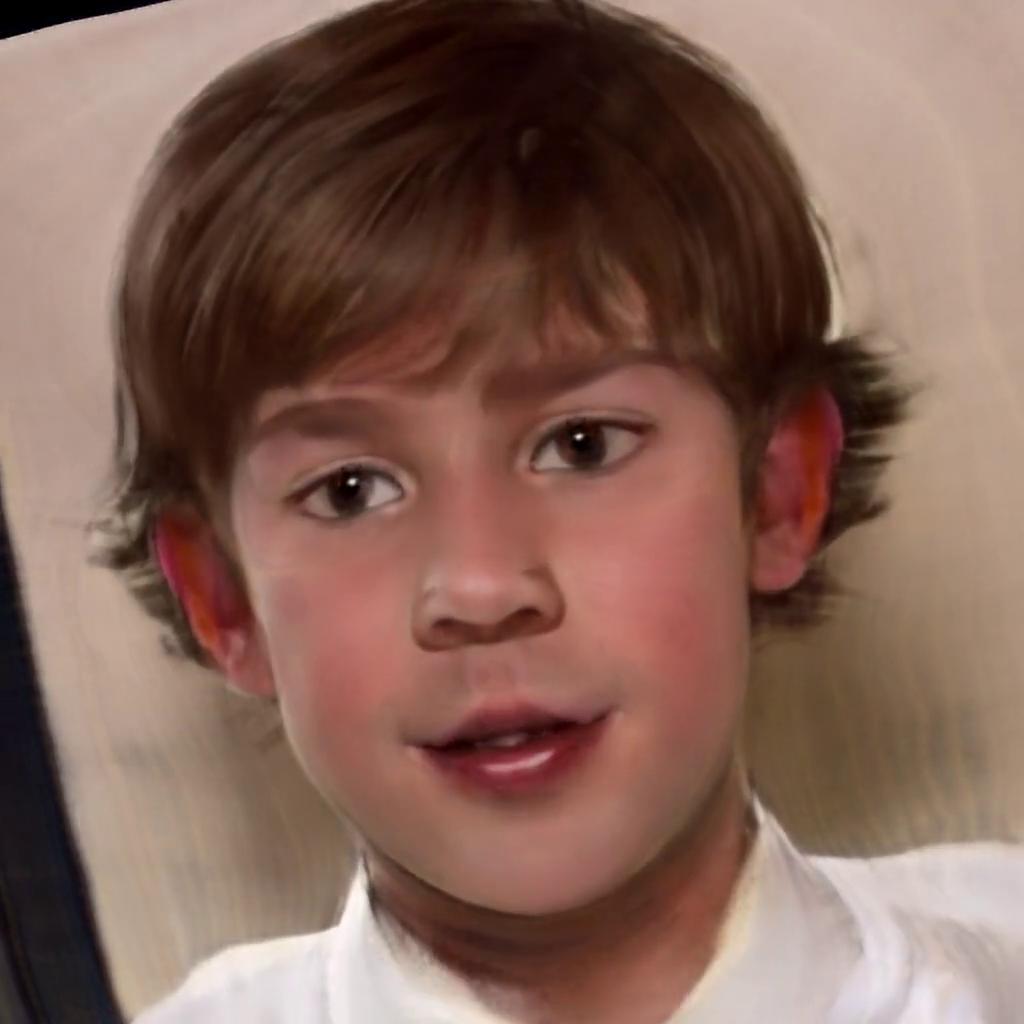} & 
        \includegraphics[width=0.215\columnwidth]{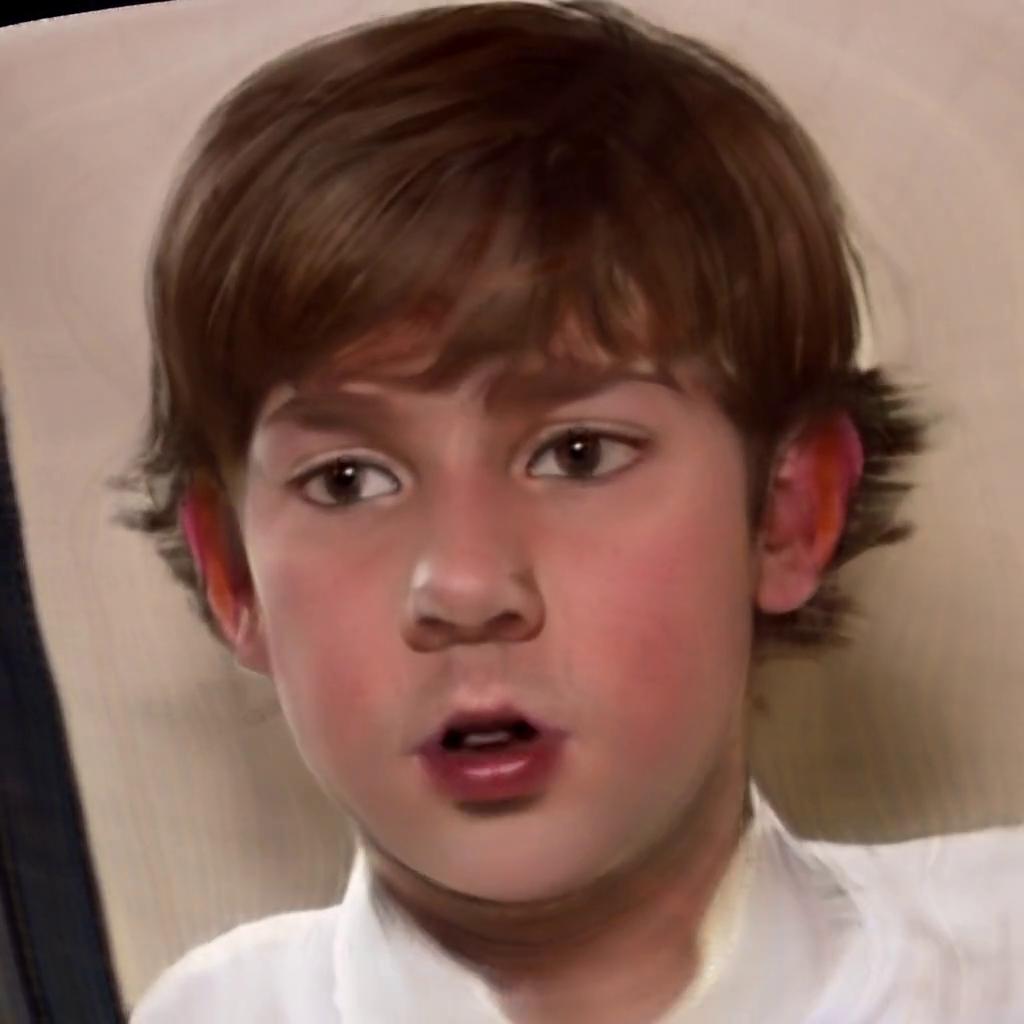} & 
        \includegraphics[width=0.215\columnwidth]{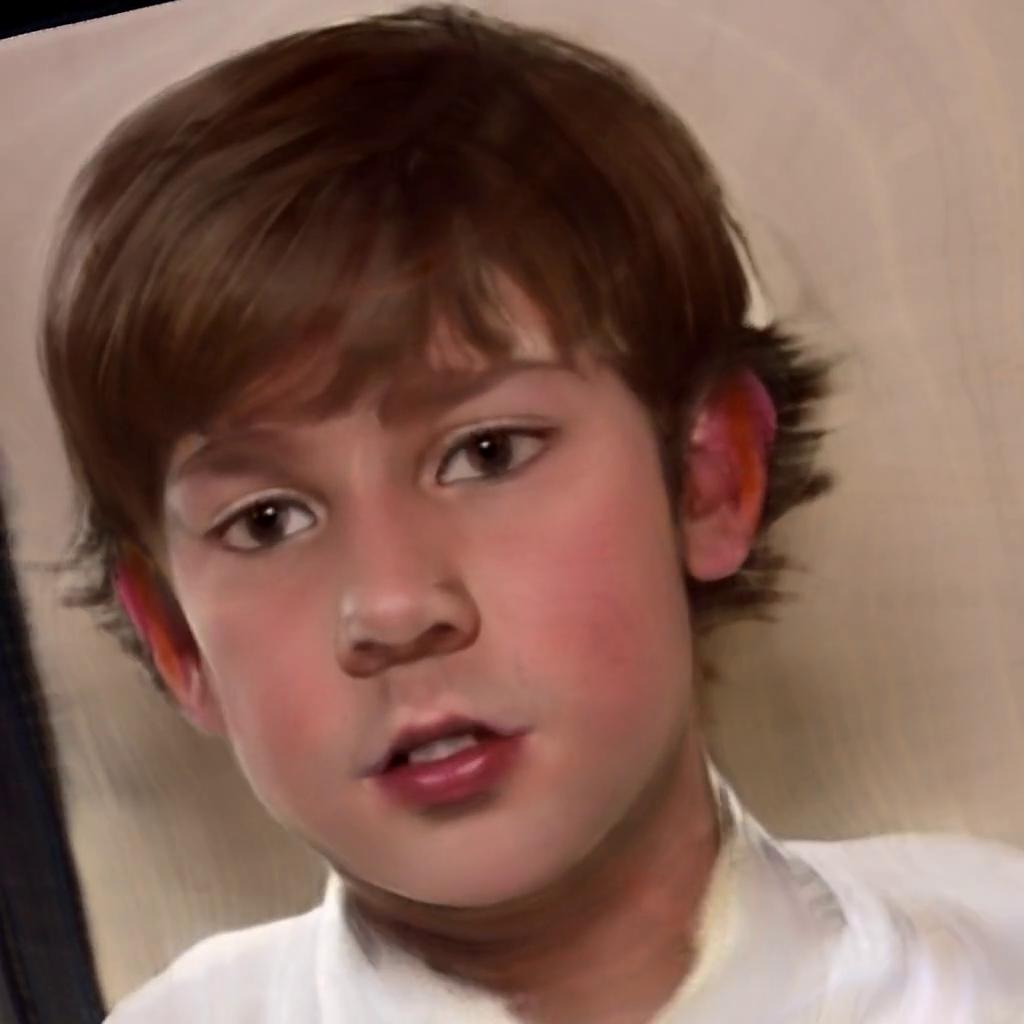} & 
        \includegraphics[width=0.215\columnwidth]{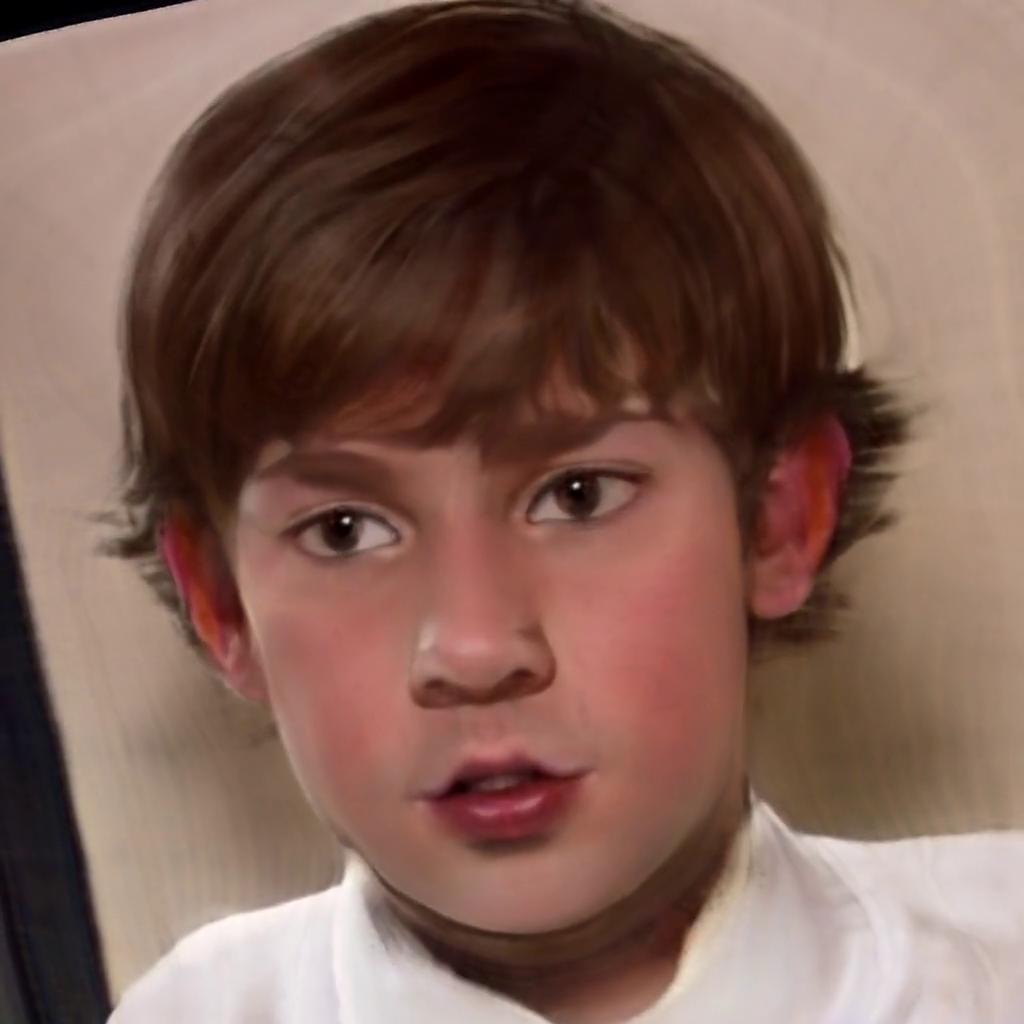} & 
        \includegraphics[width=0.215\columnwidth]{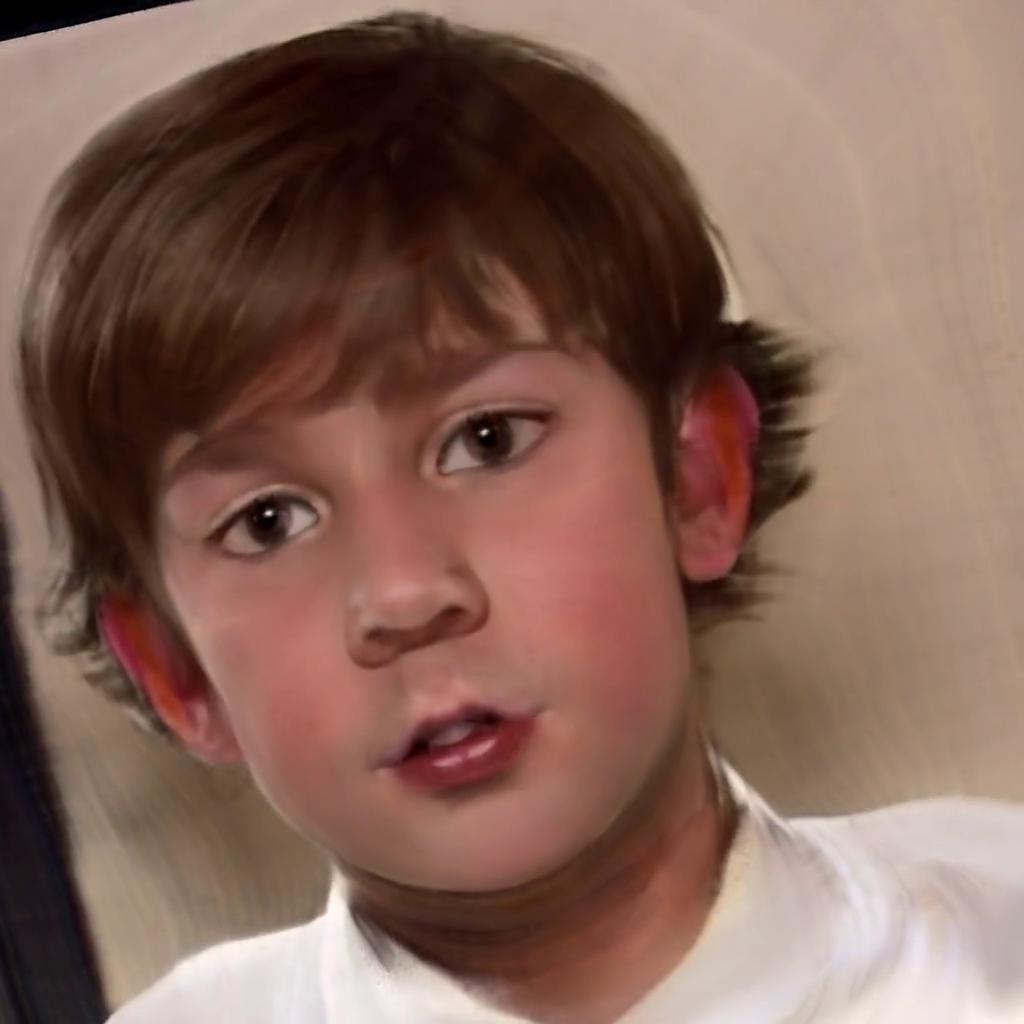} & 
        \includegraphics[width=0.215\columnwidth]{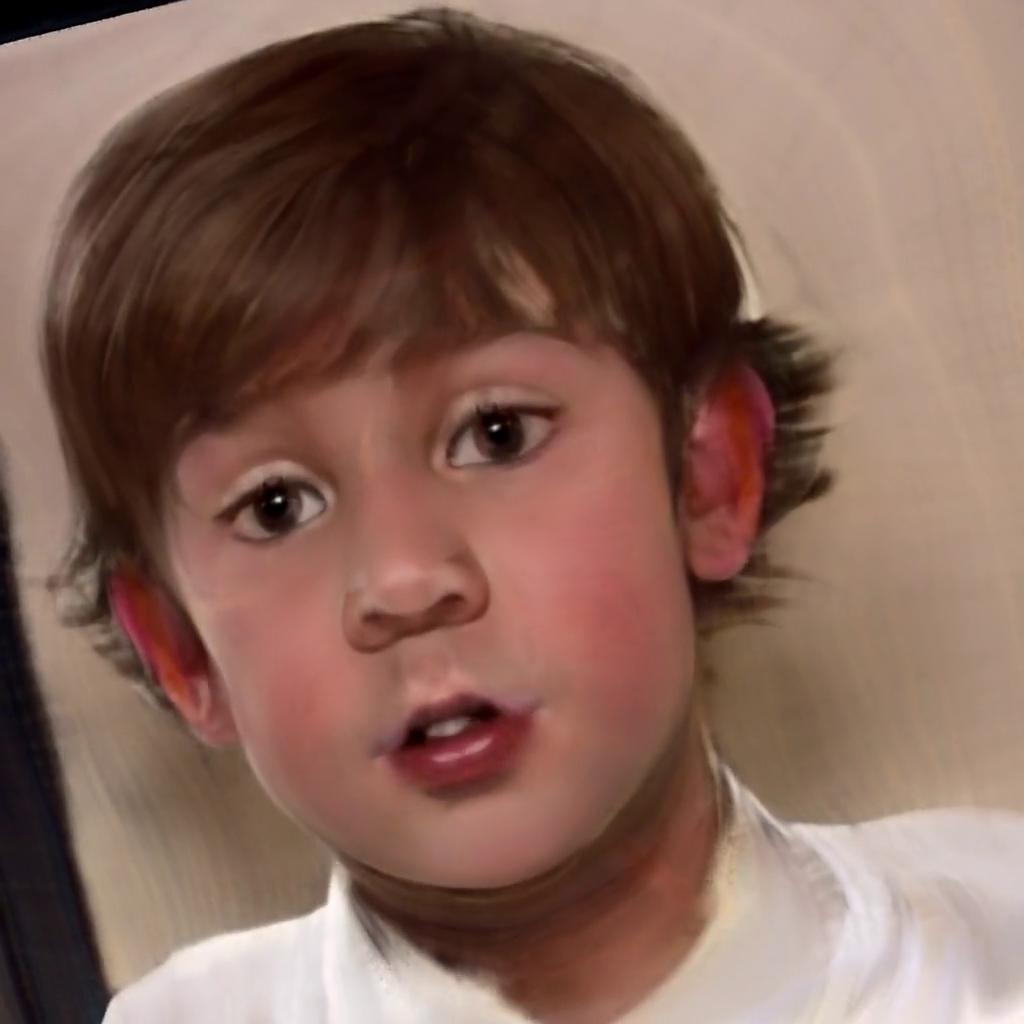} & 
        \includegraphics[width=0.215\columnwidth]{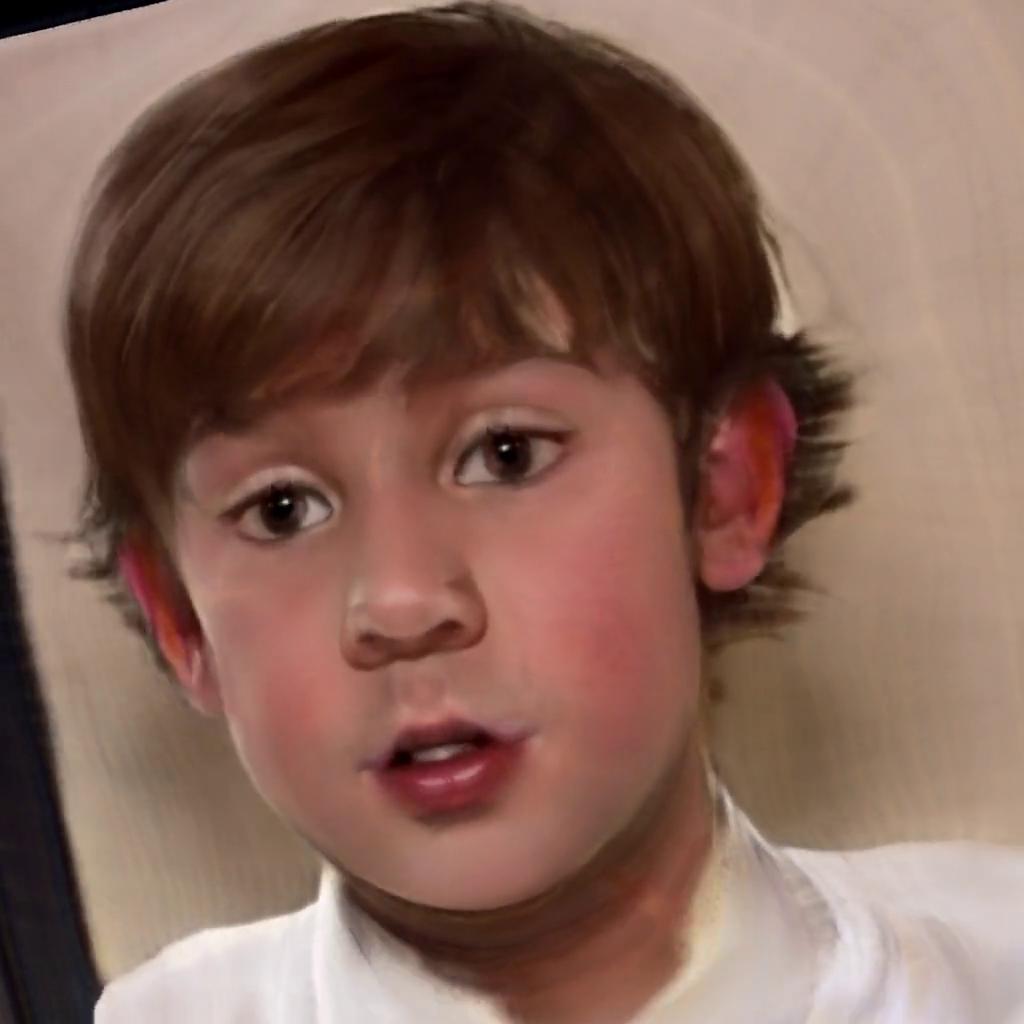} \\

	\end{tabular}
	}
	
	\vspace{-0.3cm}
	\caption{
	Sample results of our full video encoding and editing pipeline with StyleGAN3. The ``Afro'' edit is obtained with StyeCLIP's global direction technique~\cite{patashnik2021styleclip} while the ``Age'' edit is obtained using InterFaceGAN~\cite{shen2020interpreting}. The ``Pixar'' and ``Sketch'' styles are obtained with StyleGAN-NADA~\cite{gal2021stylegannada}. Note, the StyleGAN-NADA results are shown on the \textit{edited} video frames. Full videos are provided in the supplementary materials.}
	\label{fig:video_results_supplementary_1}
\end{figure*}

%% file: figures/supplementary/video_results_2.tex
\begin{figure*}[tb]
	\centering
	\setlength{\tabcolsep}{1pt}	
	{\footnotesize
	\begin{tabular}{c c c c c c c c c c}

        \\ 
        \\ 
        \\
        \\ 
        
		\raisebox{0.15in}{\rotatebox{90}{Original}} &
        \includegraphics[width=0.215\columnwidth]{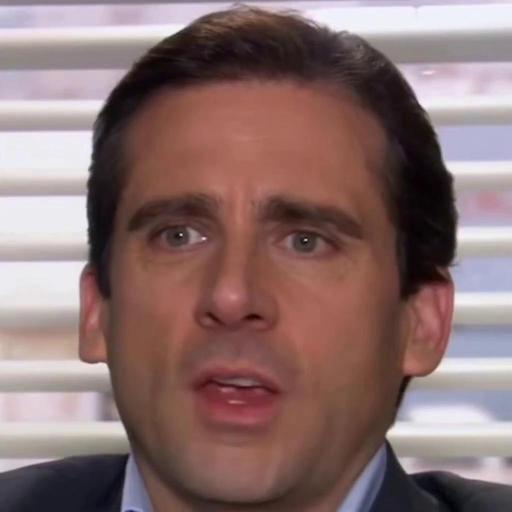} & 
        \includegraphics[width=0.215\columnwidth]{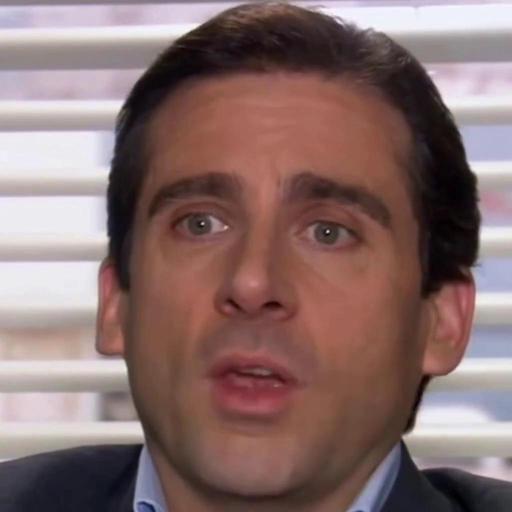} & 
        \includegraphics[width=0.215\columnwidth]{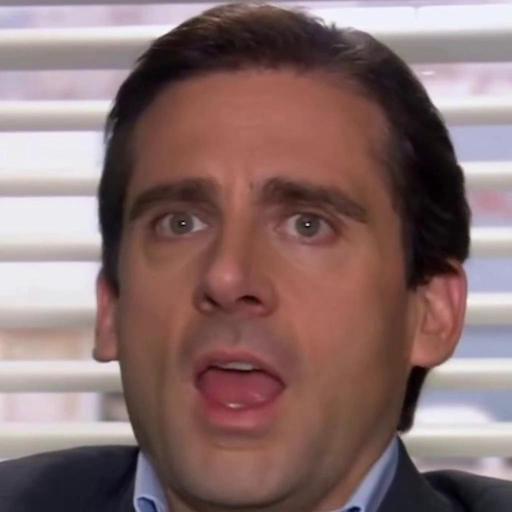} & 
        \includegraphics[width=0.215\columnwidth]{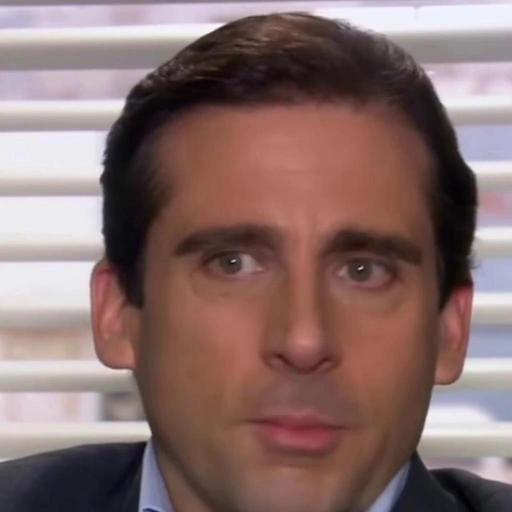} & 
        \includegraphics[width=0.215\columnwidth]{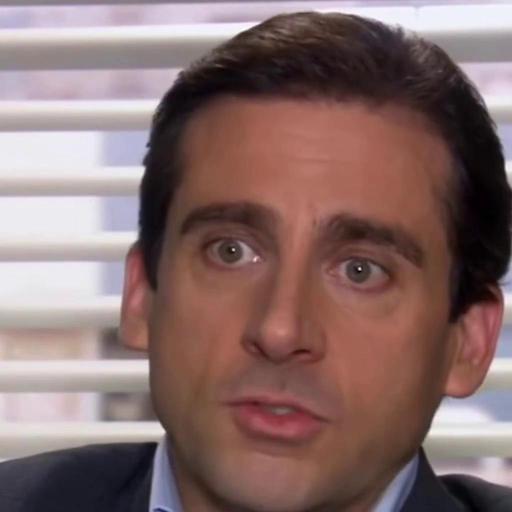} & 
        \includegraphics[width=0.215\columnwidth]{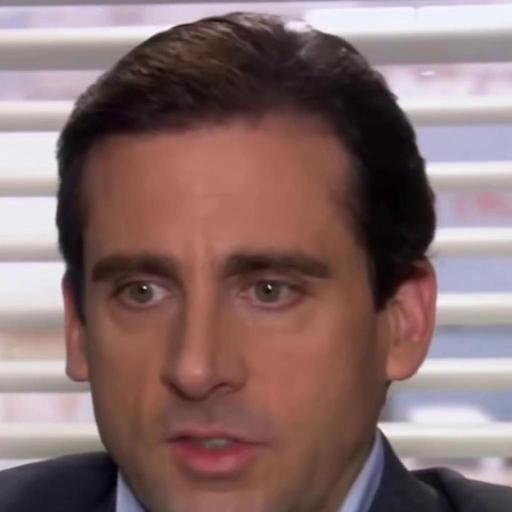} & 
        \includegraphics[width=0.215\columnwidth]{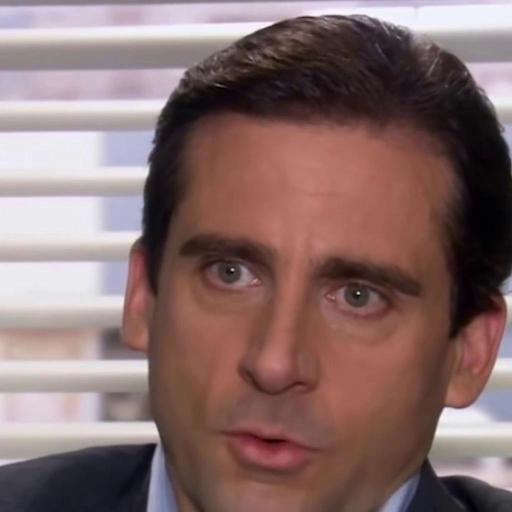} & 
        \includegraphics[width=0.215\columnwidth]{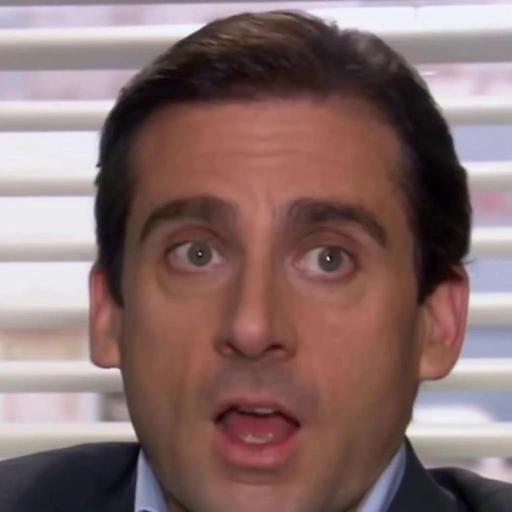} & 
        \includegraphics[width=0.215\columnwidth]{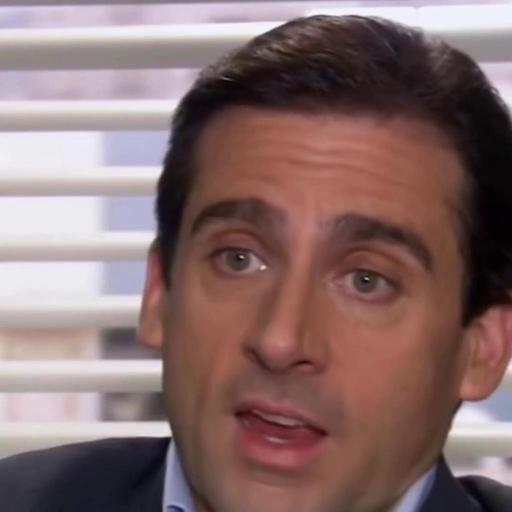} \\

		\raisebox{0.05in}{\rotatebox{90}{Reconstruction}} &
        \includegraphics[width=0.215\columnwidth]{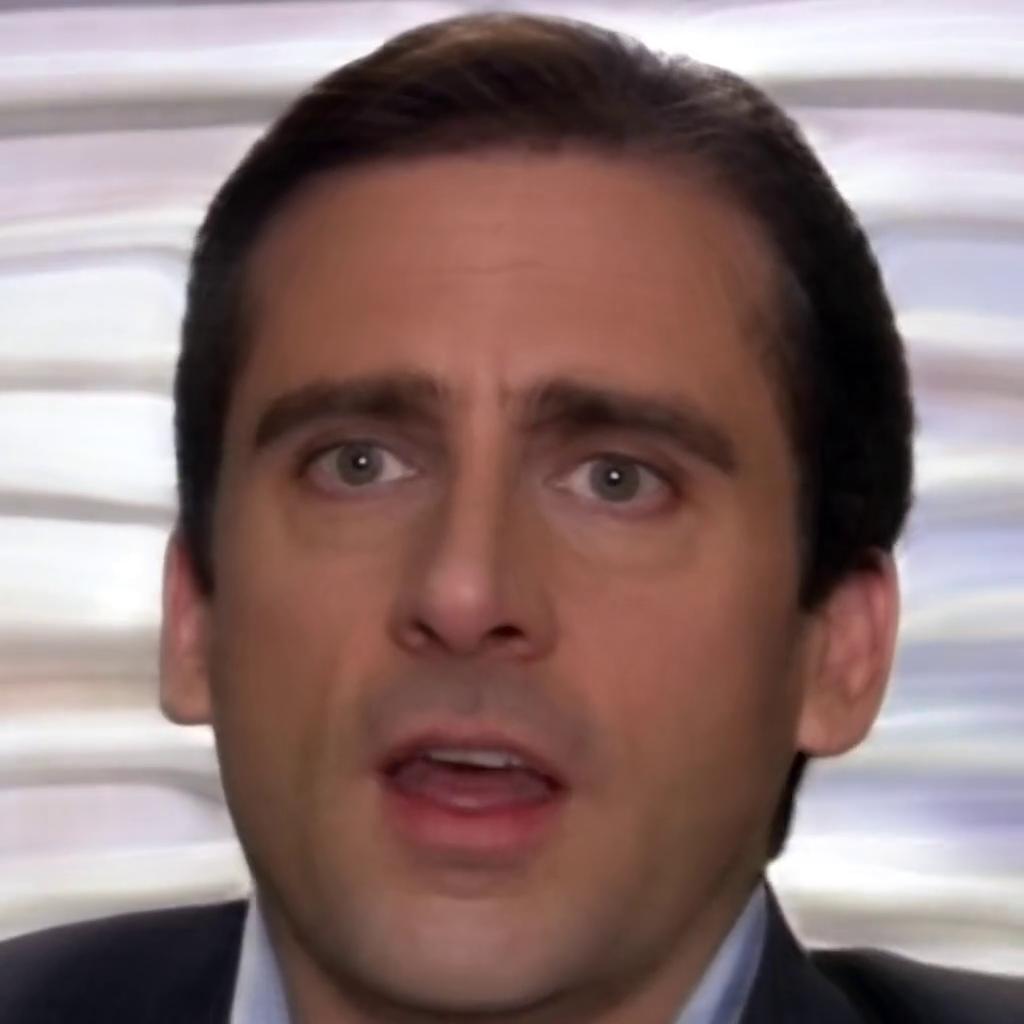} & 
        \includegraphics[width=0.215\columnwidth]{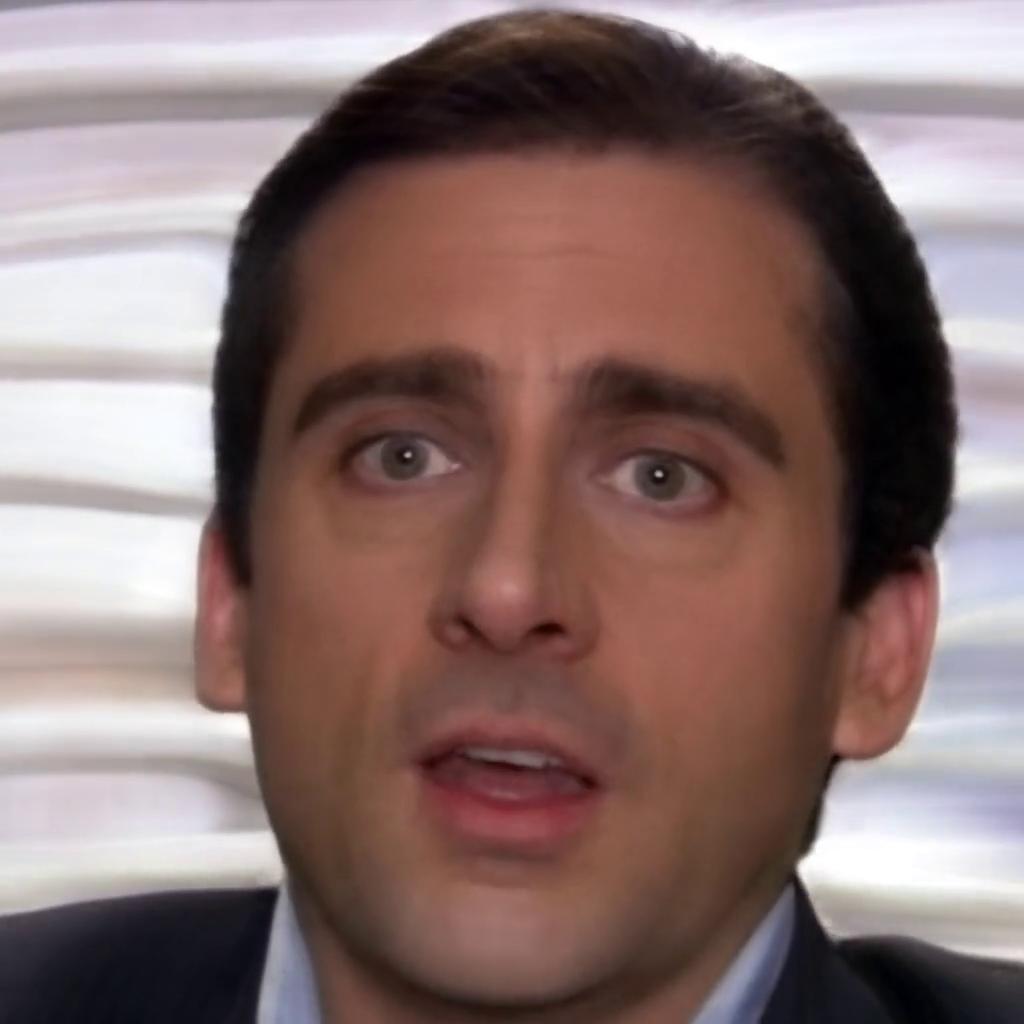} & 
        \includegraphics[width=0.215\columnwidth]{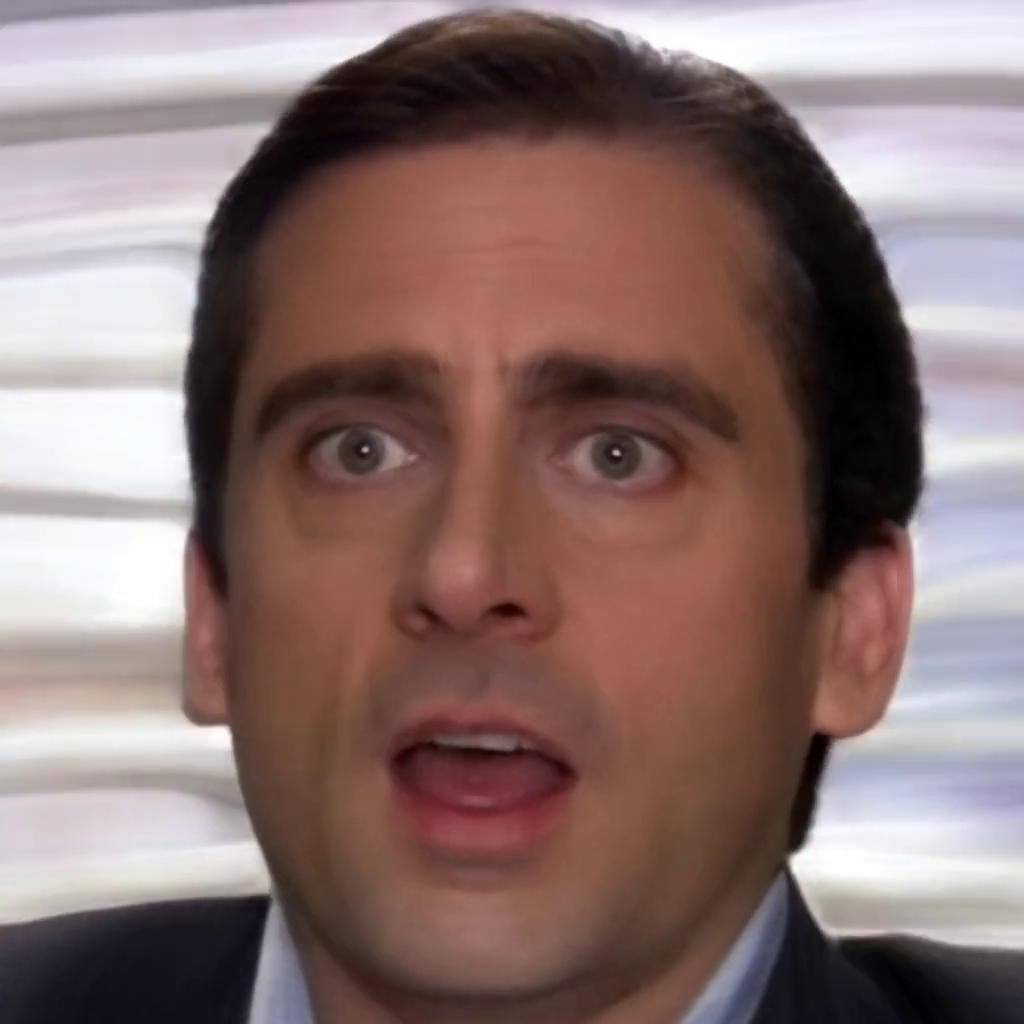} & 
        \includegraphics[width=0.215\columnwidth]{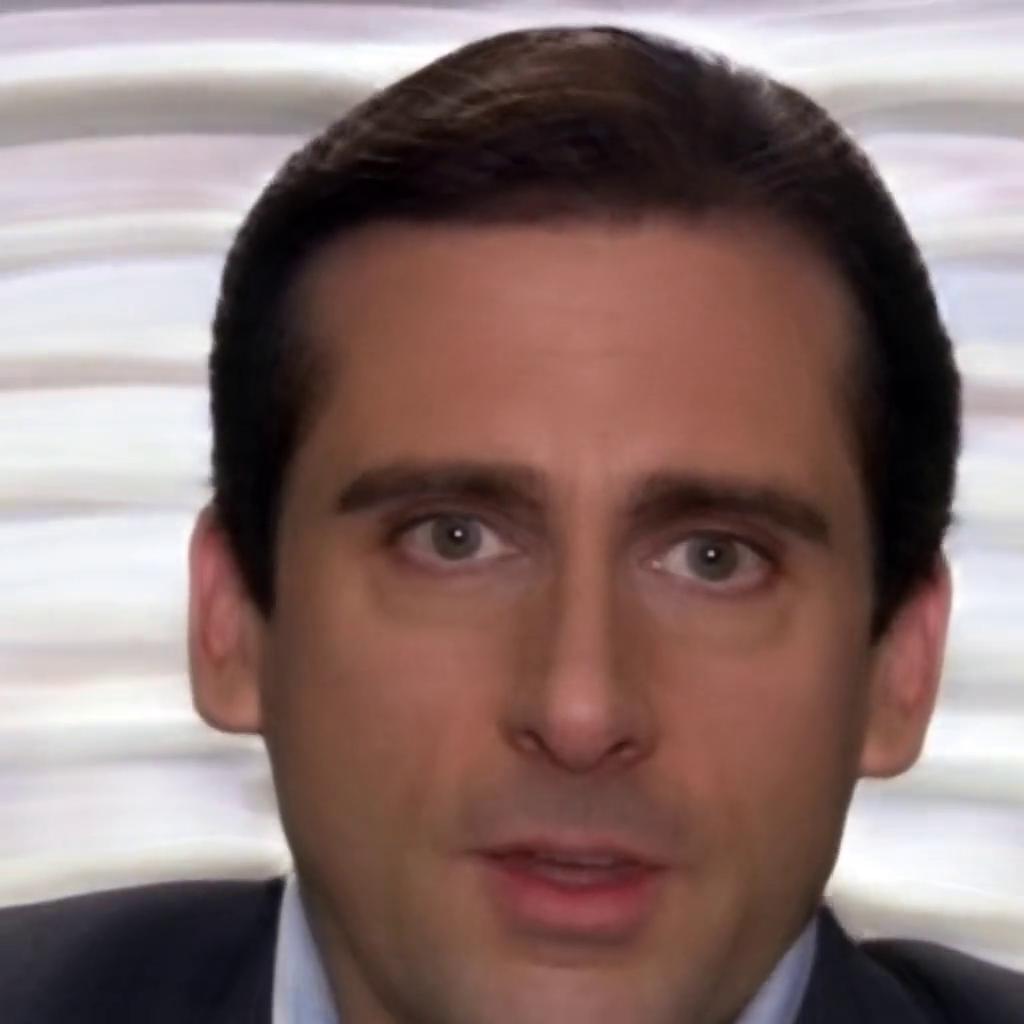} & 
        \includegraphics[width=0.215\columnwidth]{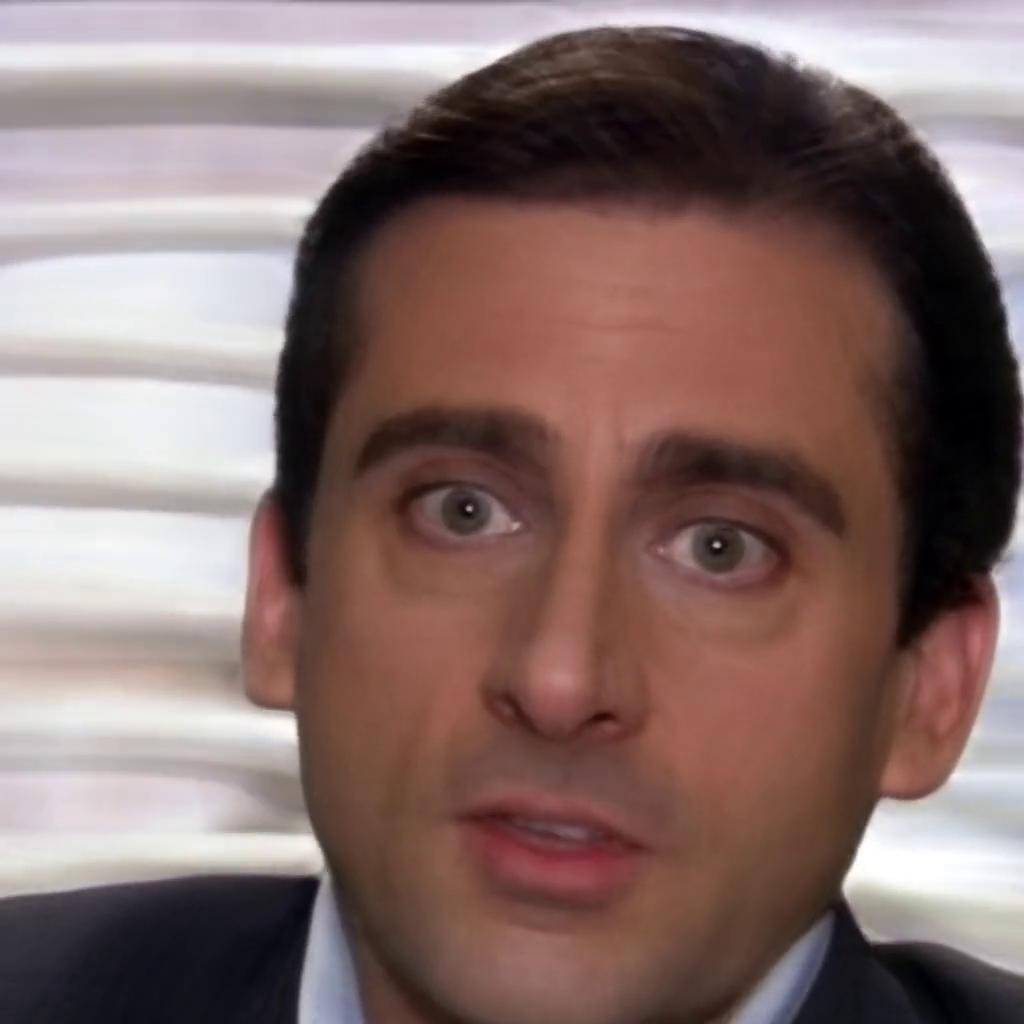} & 
        \includegraphics[width=0.215\columnwidth]{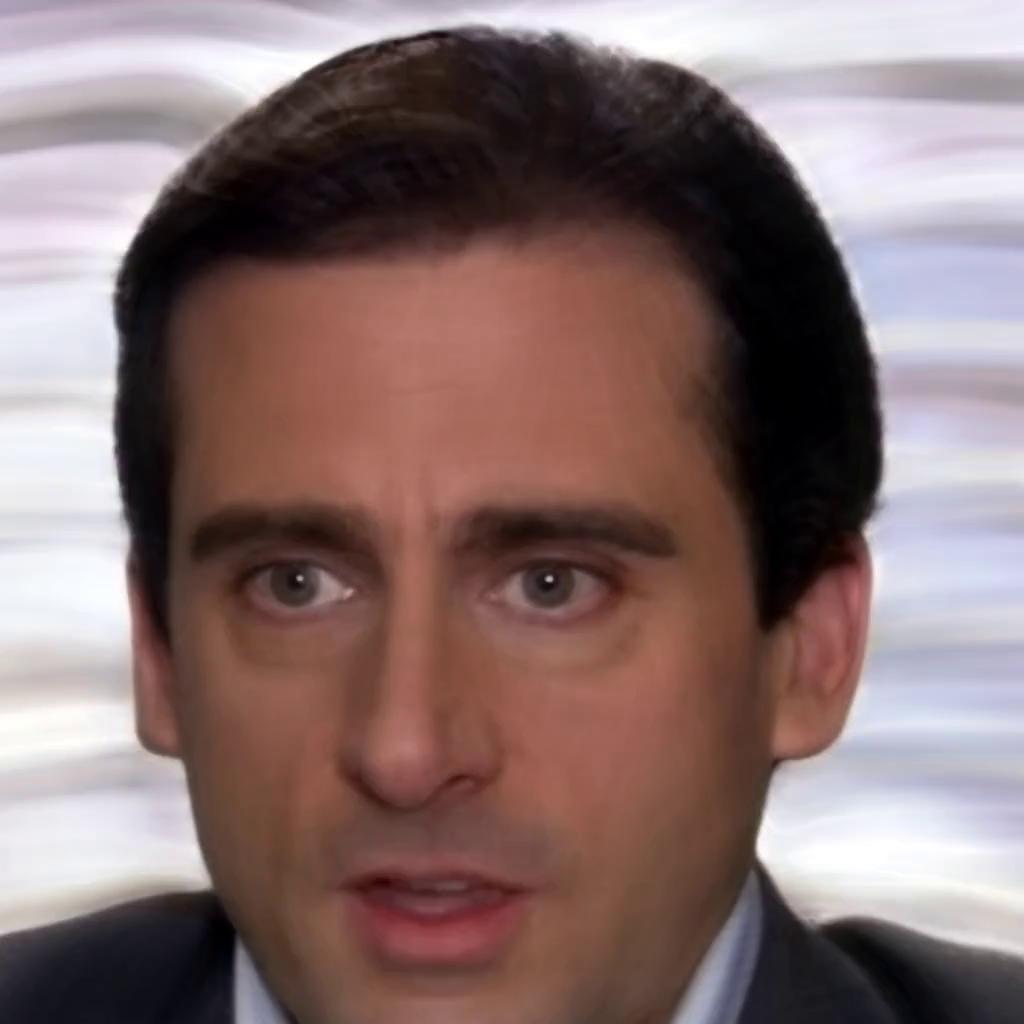} & 
        \includegraphics[width=0.215\columnwidth]{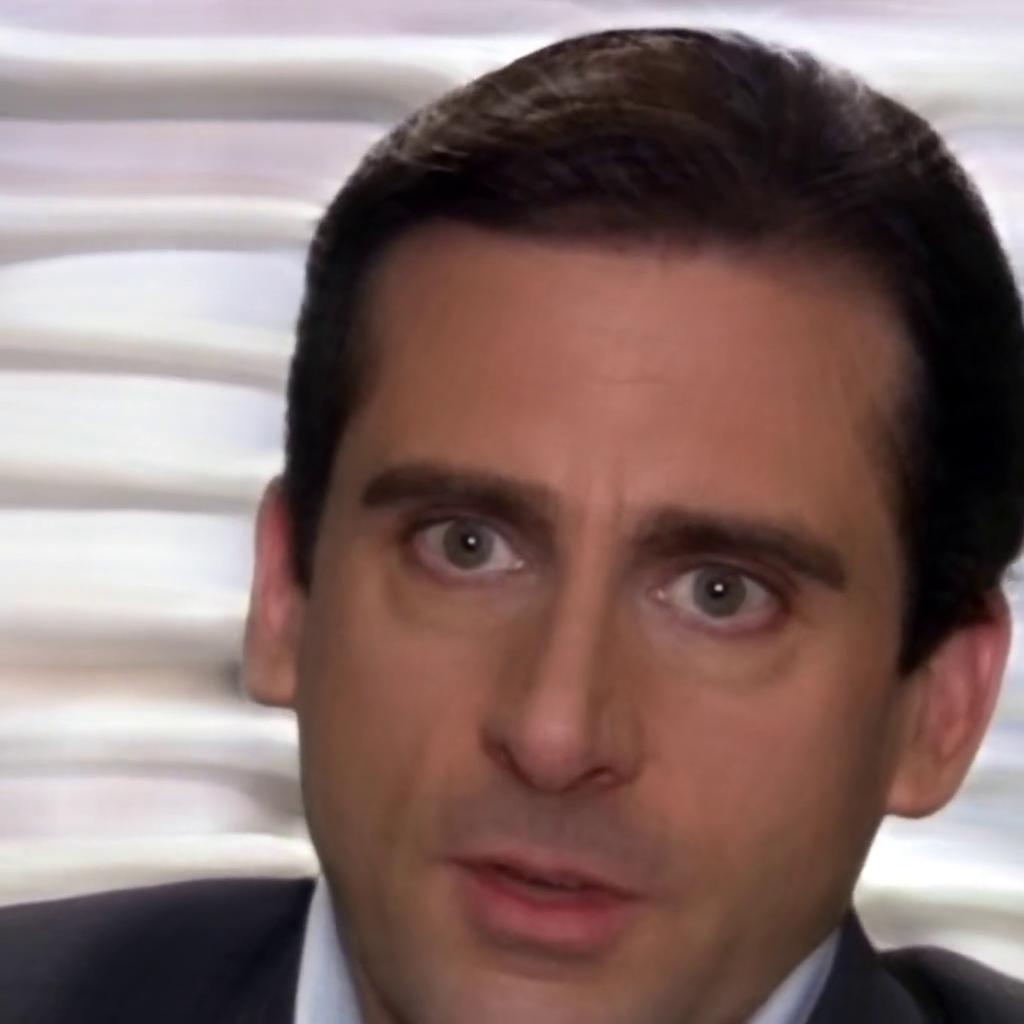} & 
        \includegraphics[width=0.215\columnwidth]{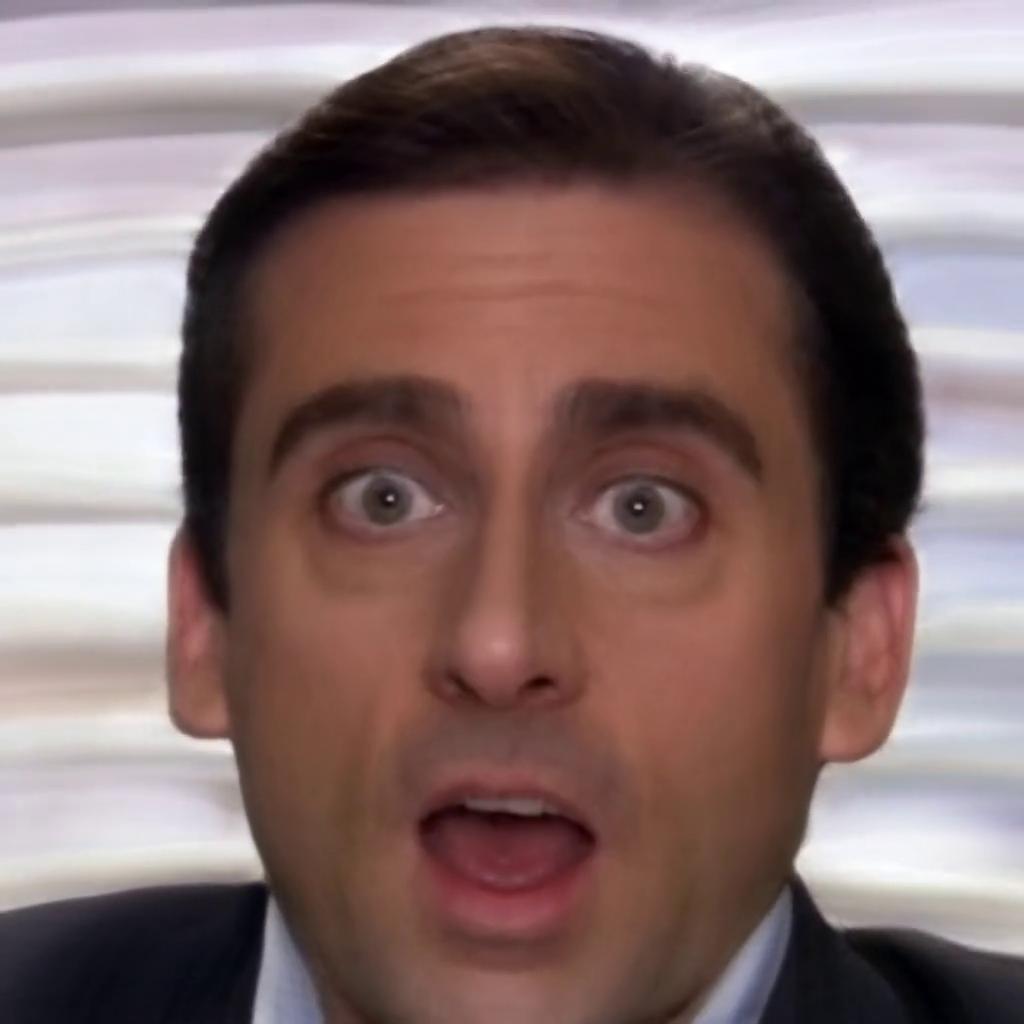} & 
        \includegraphics[width=0.215\columnwidth]{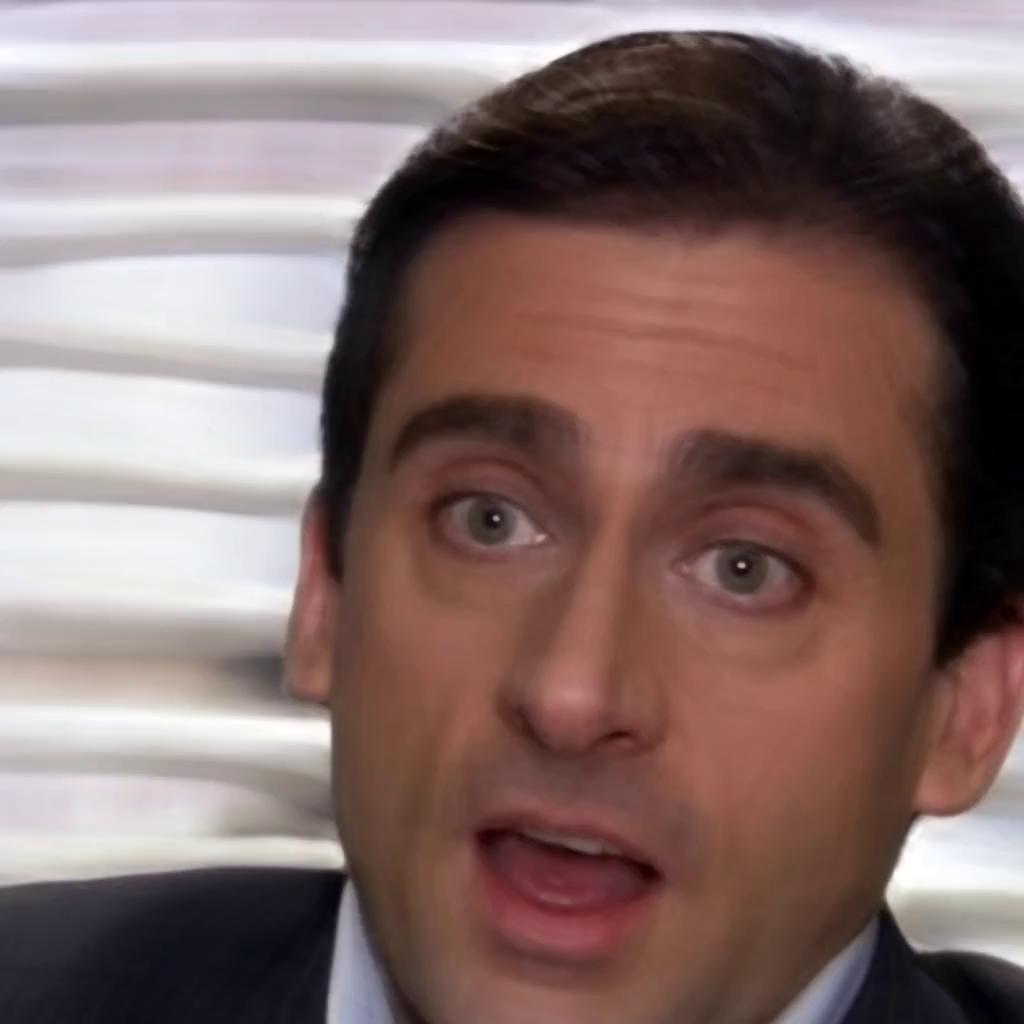} \\

		\raisebox{0.05in}{\rotatebox{90}{$+$ Double Chin}} &
        \includegraphics[width=0.215\columnwidth]{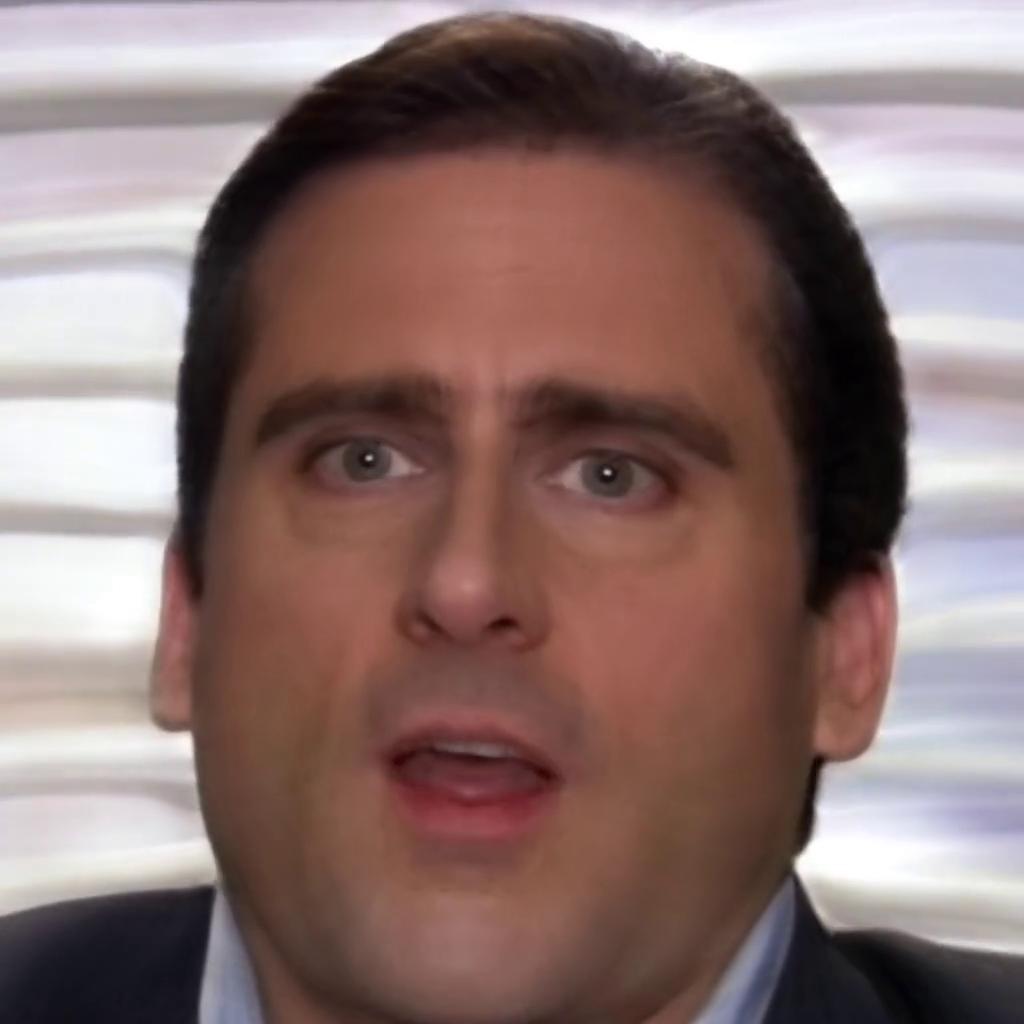} & 
        \includegraphics[width=0.215\columnwidth]{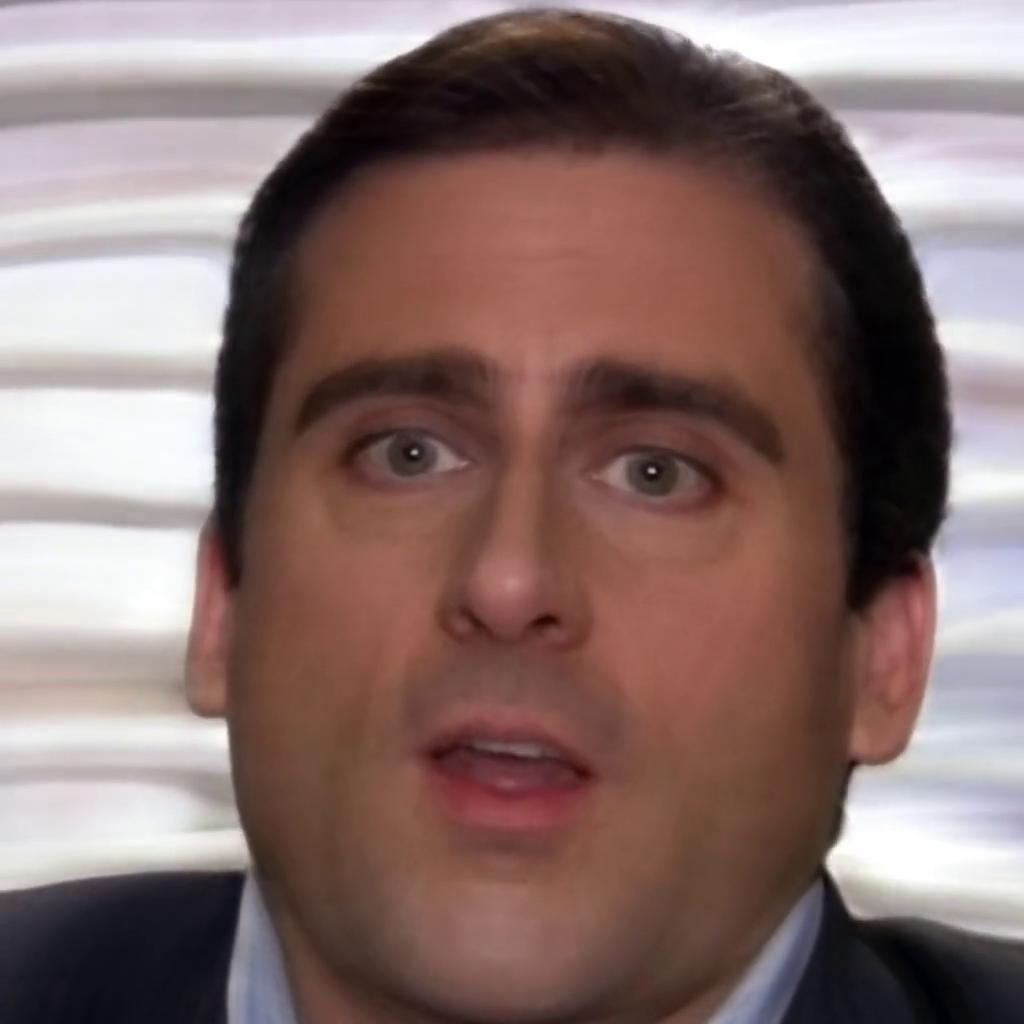} & 
        \includegraphics[width=0.215\columnwidth]{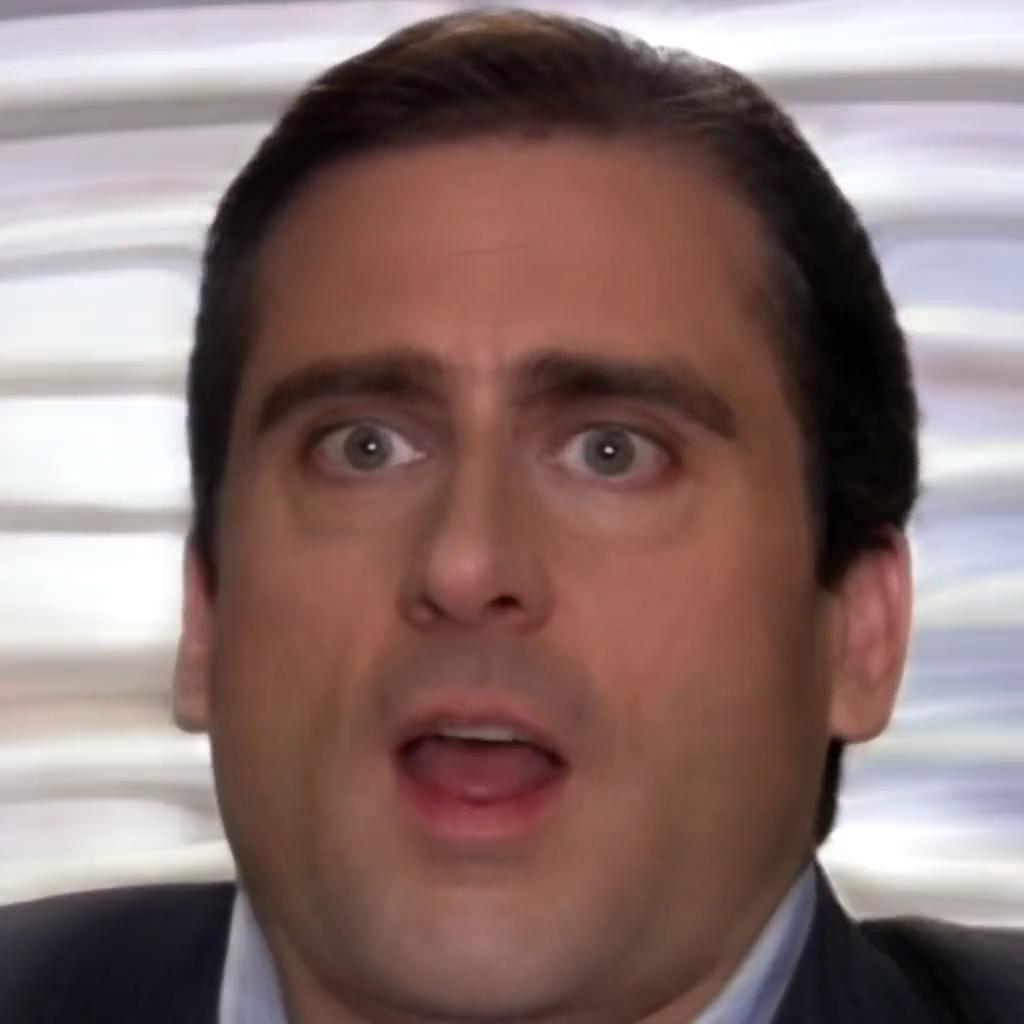} & 
        \includegraphics[width=0.215\columnwidth]{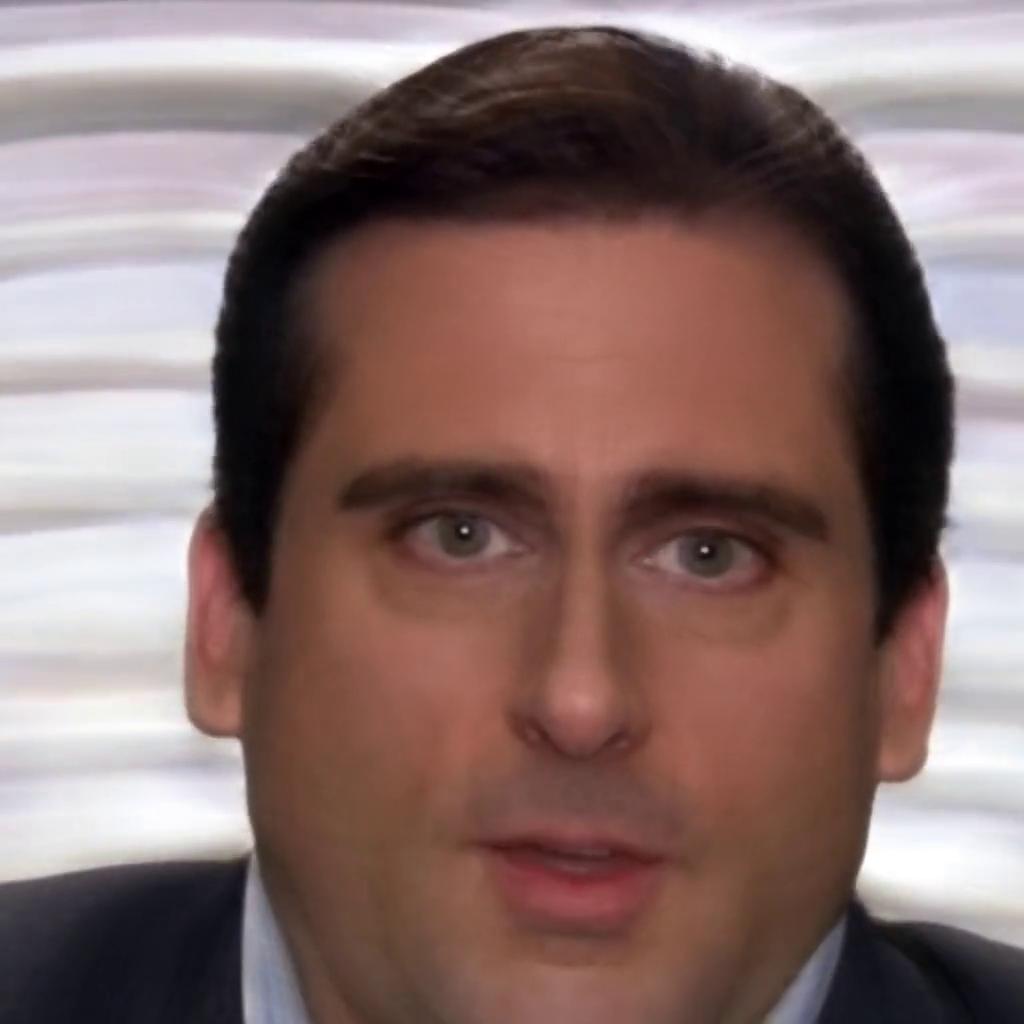} & 
        \includegraphics[width=0.215\columnwidth]{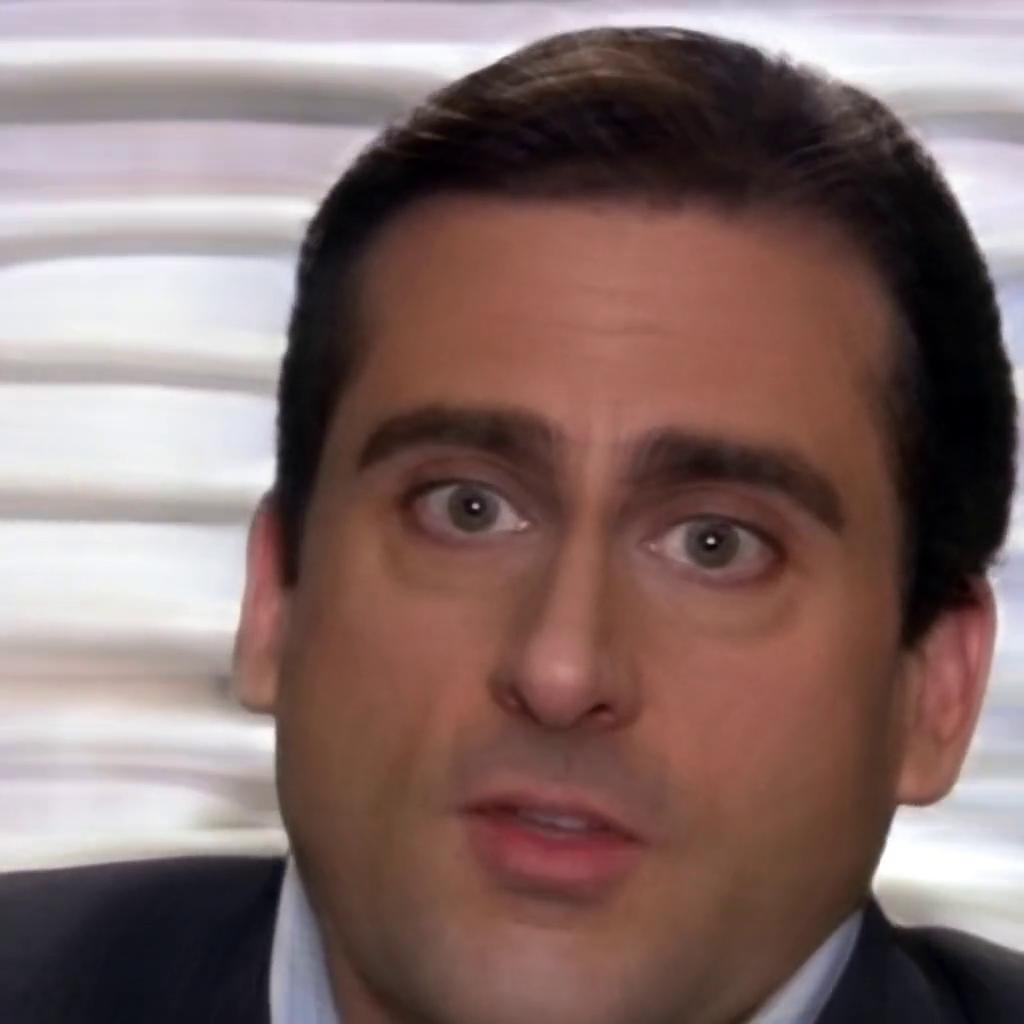} & 
        \includegraphics[width=0.215\columnwidth]{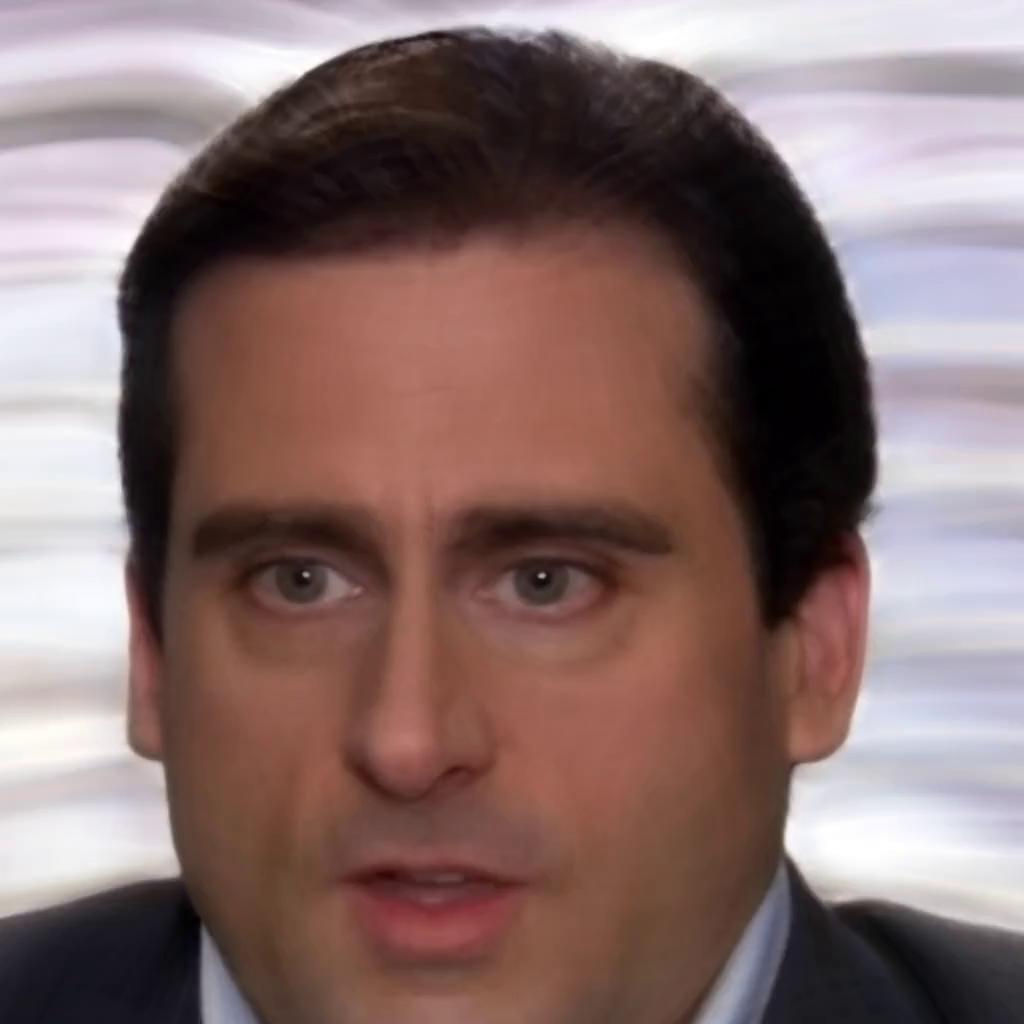} & 
        \includegraphics[width=0.215\columnwidth]{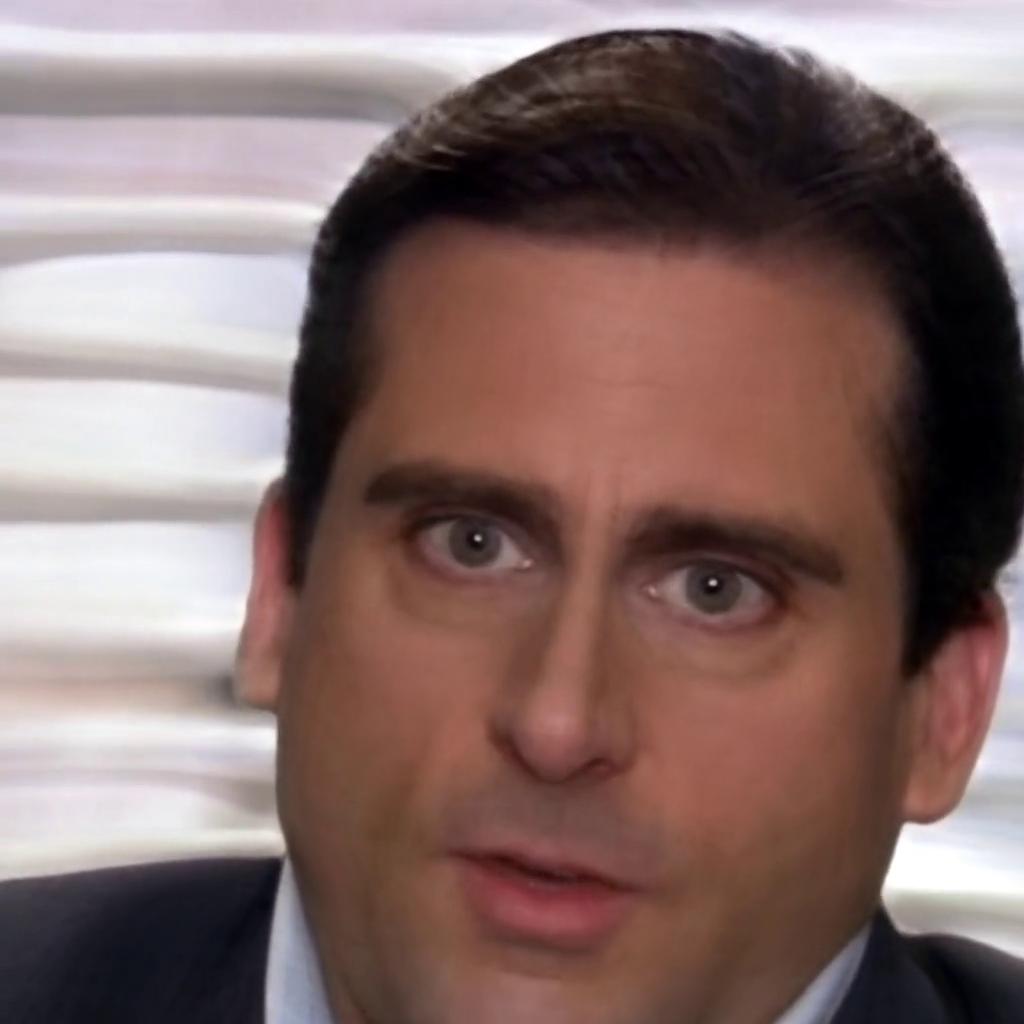} & 
        \includegraphics[width=0.215\columnwidth]{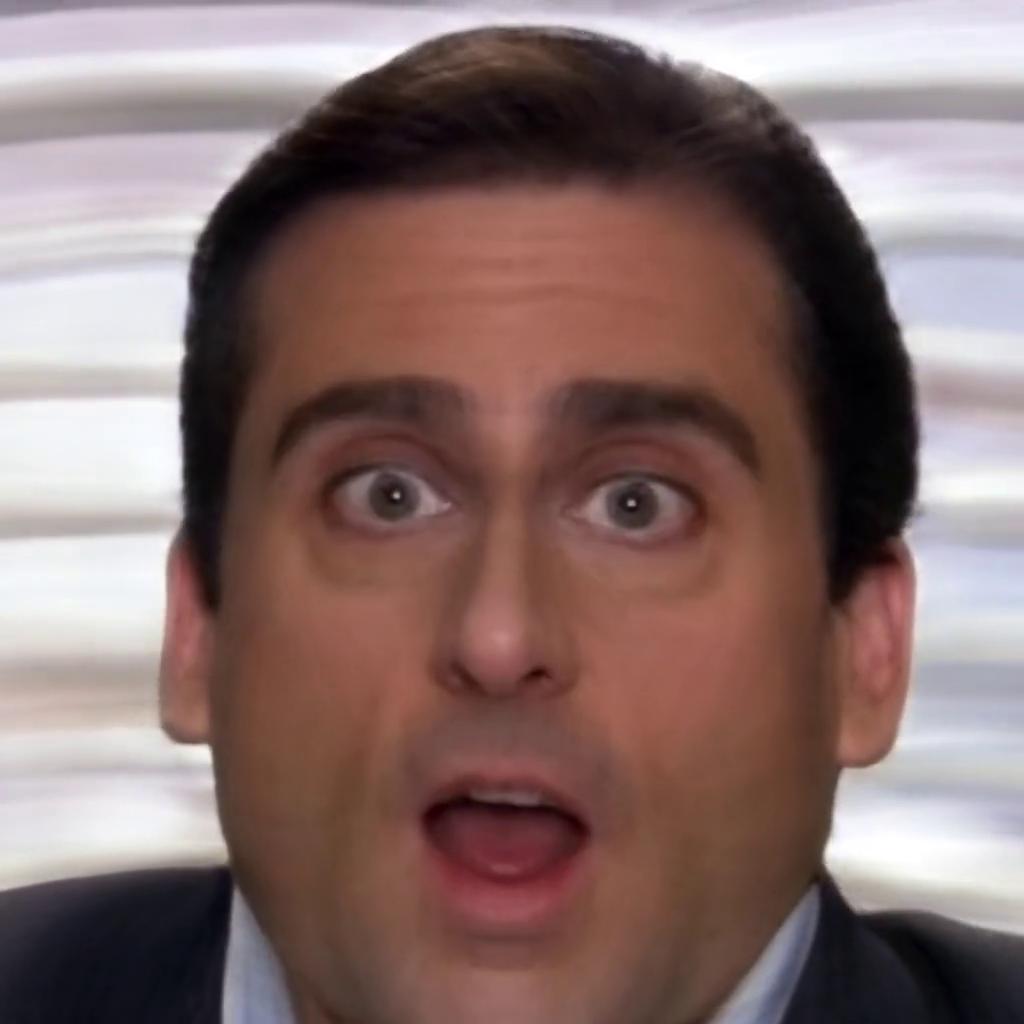} & 
        \includegraphics[width=0.215\columnwidth]{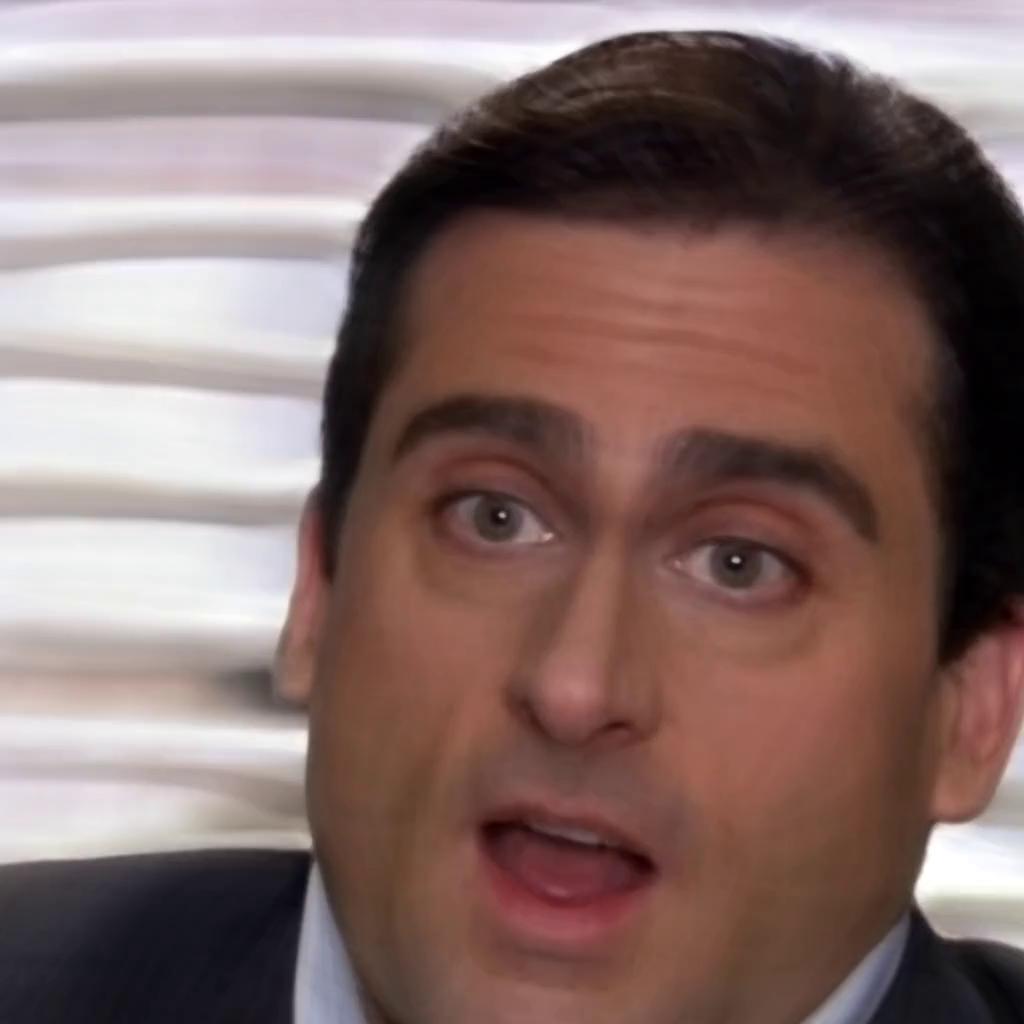} \\
        
		\raisebox{0.175in}{\rotatebox{90}{$+$ Pixar}} &
        \includegraphics[width=0.215\columnwidth]{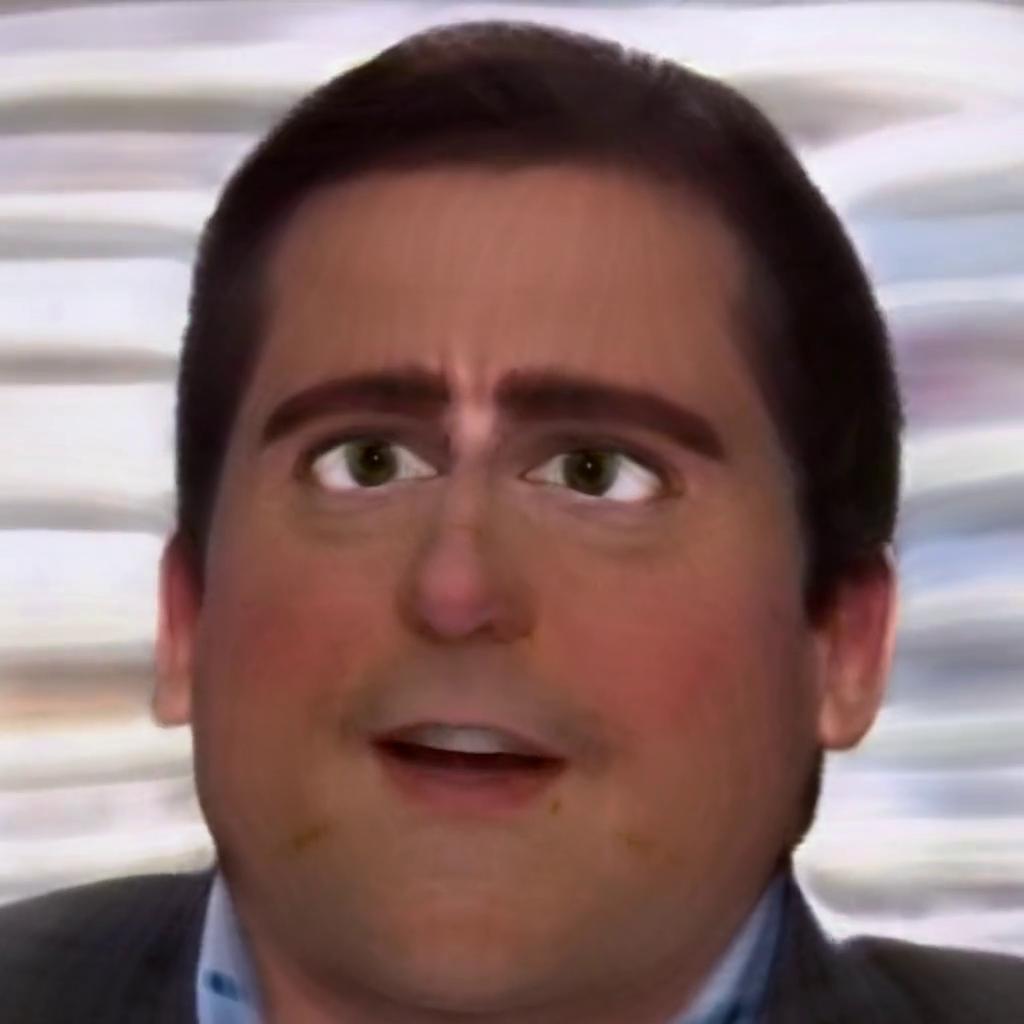} & 
        \includegraphics[width=0.215\columnwidth]{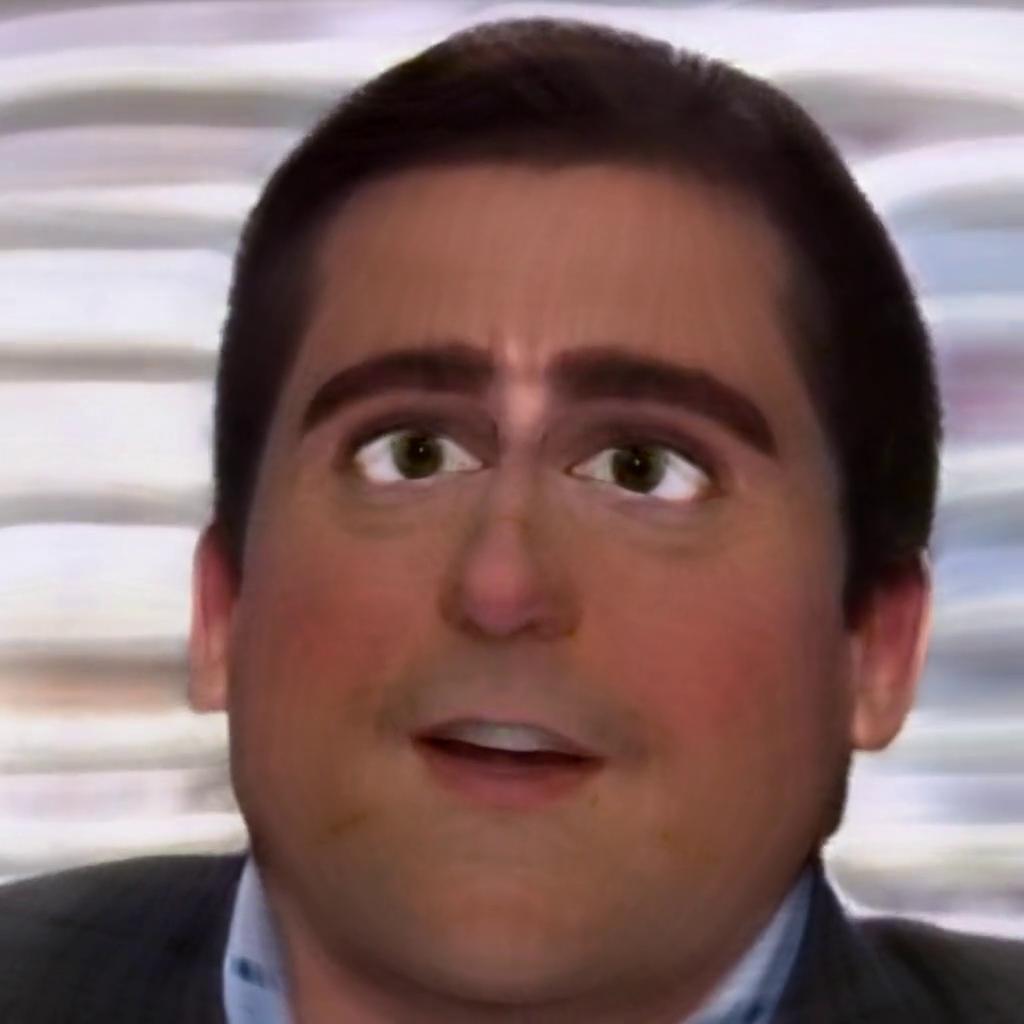} & 
        \includegraphics[width=0.215\columnwidth]{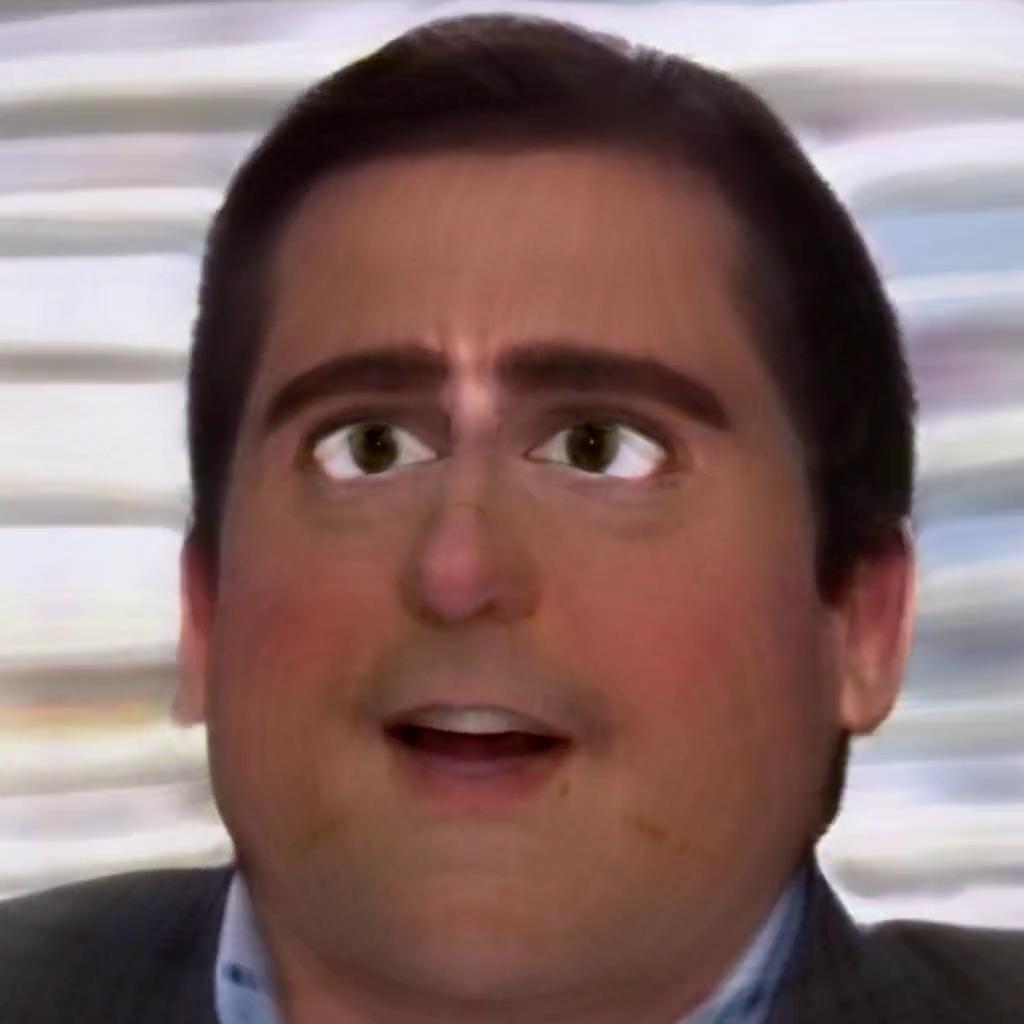} & 
        \includegraphics[width=0.215\columnwidth]{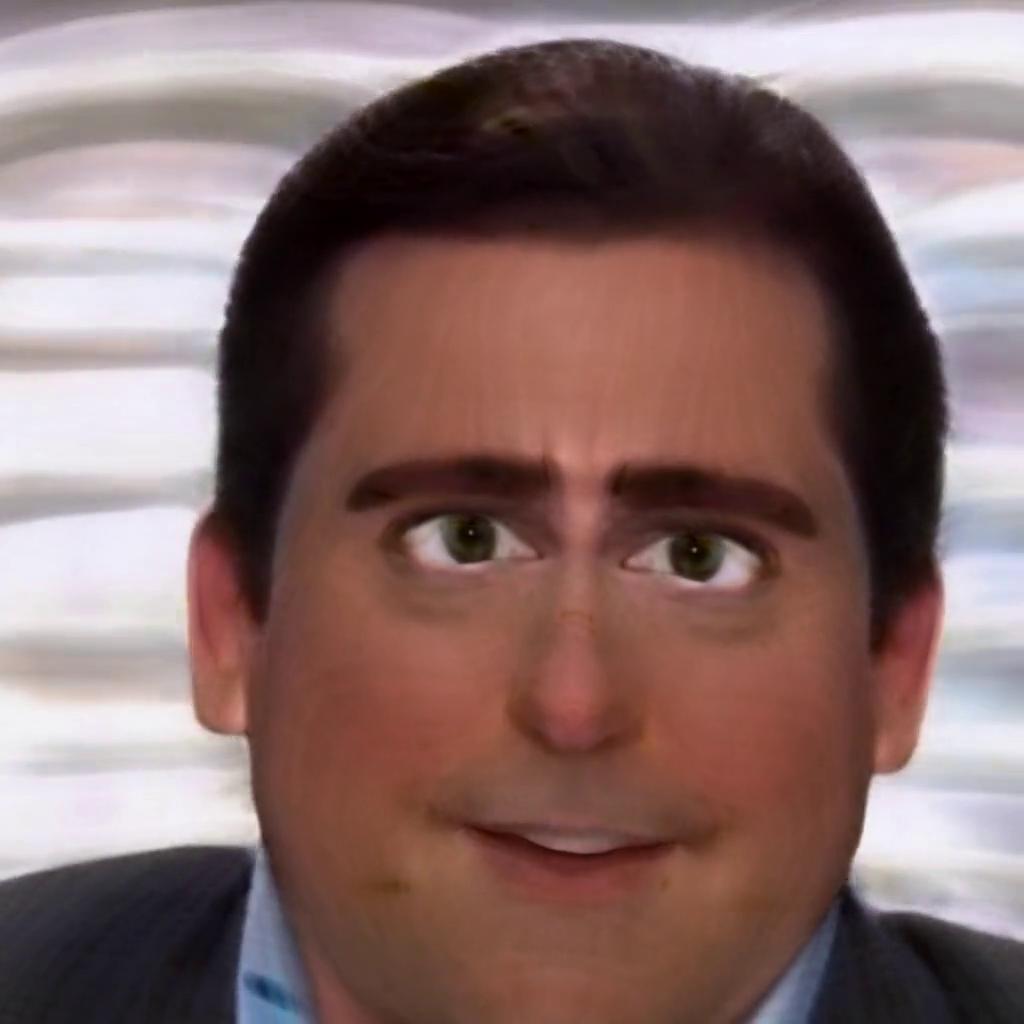} & 
        \includegraphics[width=0.215\columnwidth]{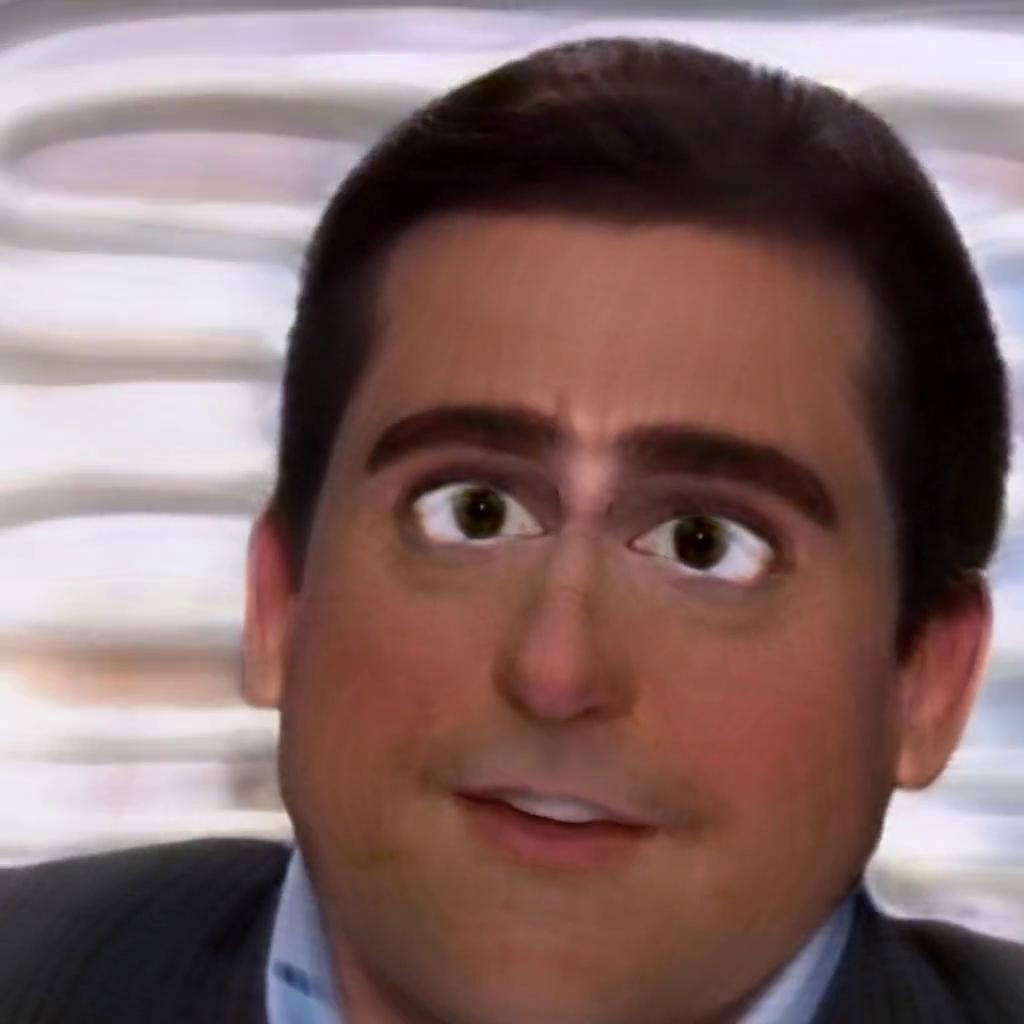} & 
        \includegraphics[width=0.215\columnwidth]{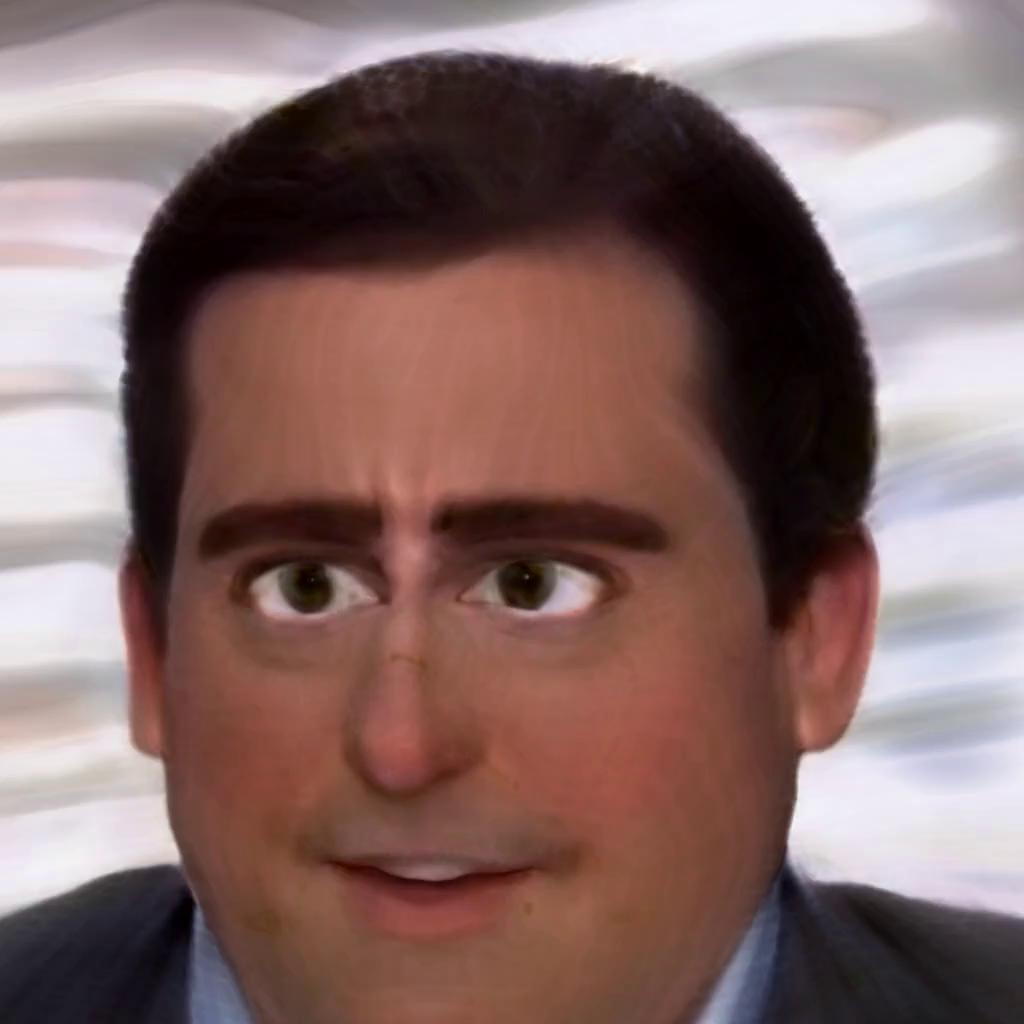} & 
        \includegraphics[width=0.215\columnwidth]{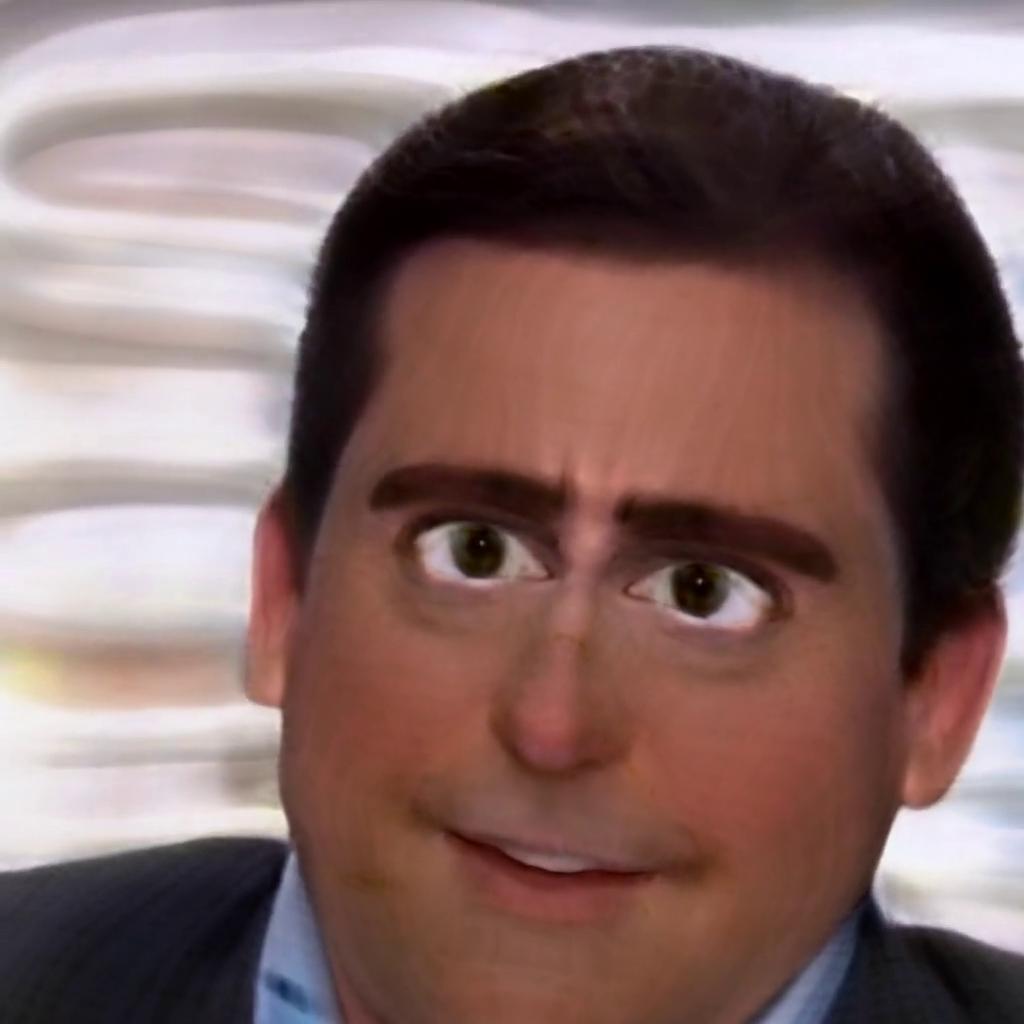} & 
        \includegraphics[width=0.215\columnwidth]{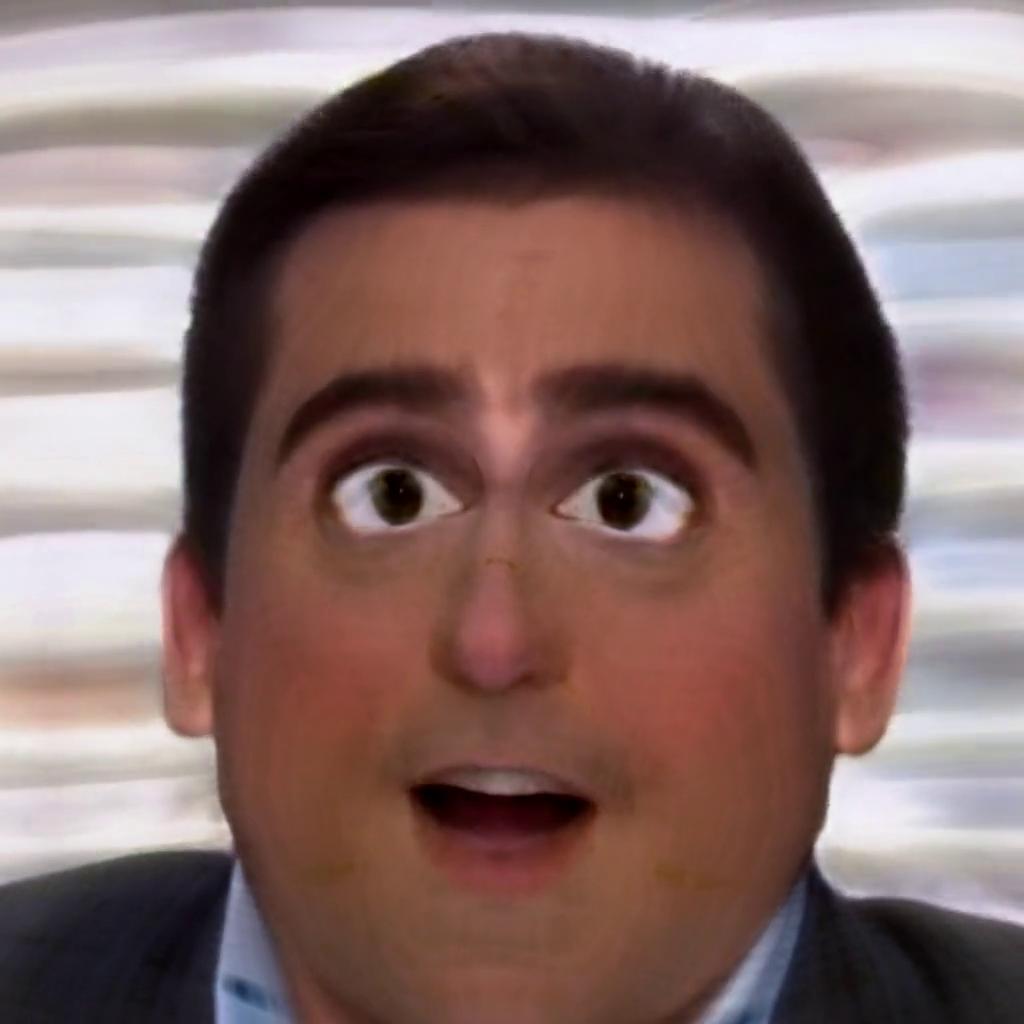} & 
        \includegraphics[width=0.215\columnwidth]{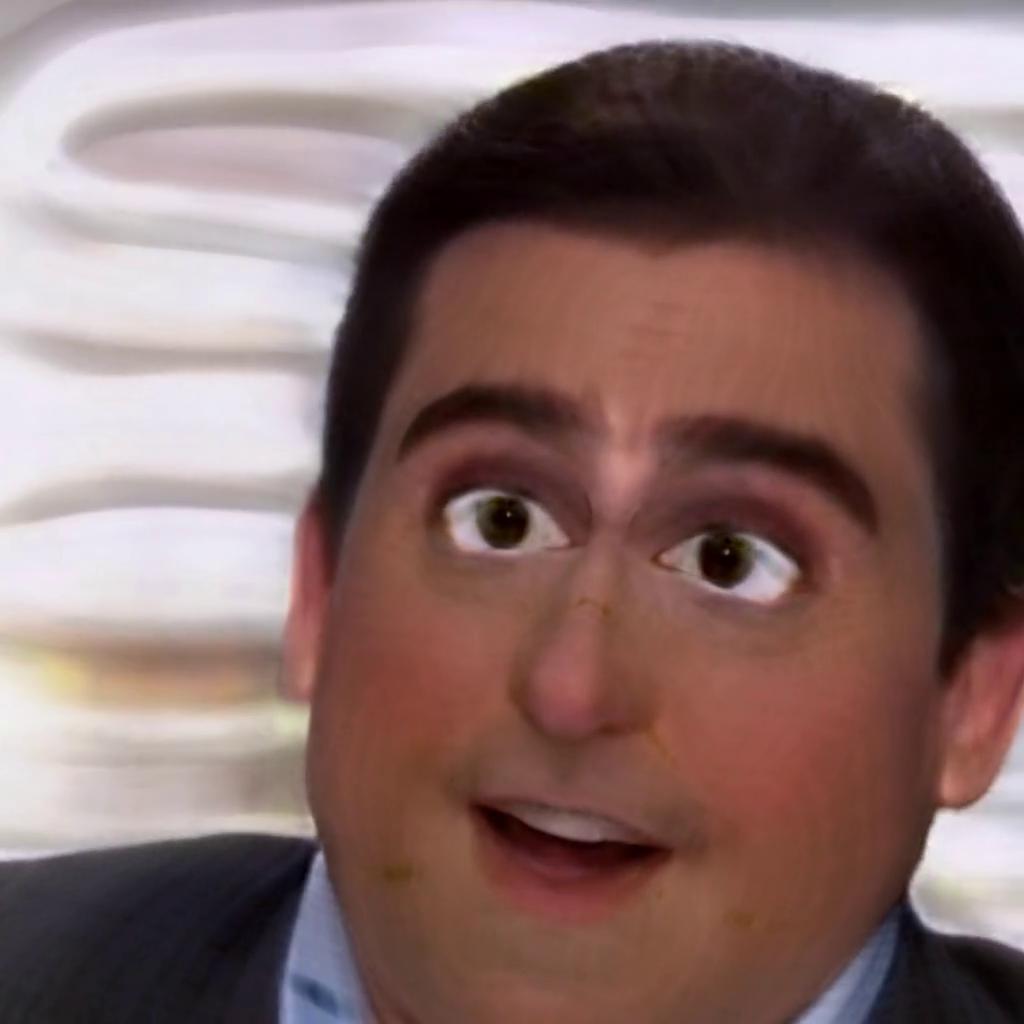} \\

	\end{tabular}
	}
	
	{\footnotesize
	\begin{tabular}{c c c c c c c c c c}
        
        \\ \\

		\raisebox{0.15in}{\rotatebox{90}{Original}} &
        \includegraphics[width=0.215\columnwidth]{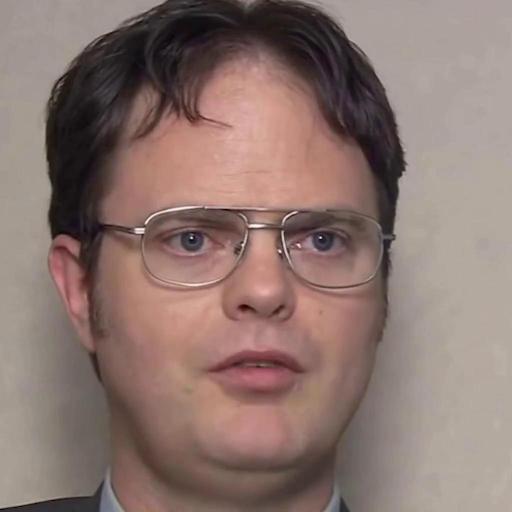} & 
        \includegraphics[width=0.215\columnwidth]{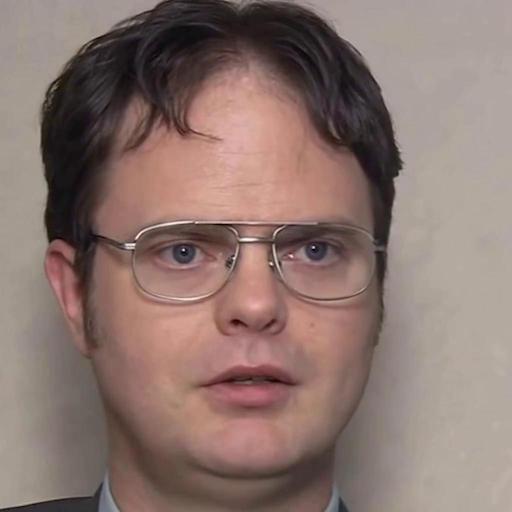} & 
        \includegraphics[width=0.215\columnwidth]{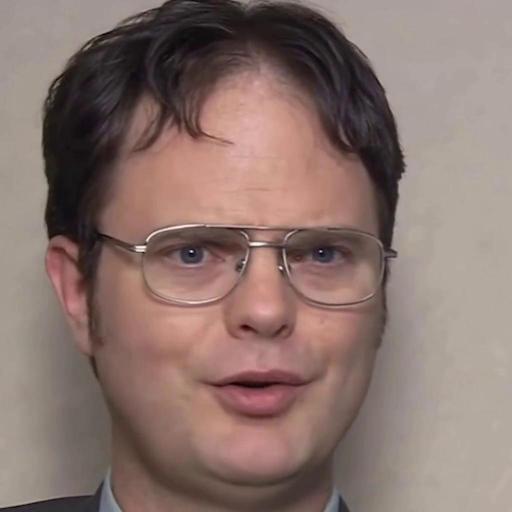} & 
        \includegraphics[width=0.215\columnwidth]{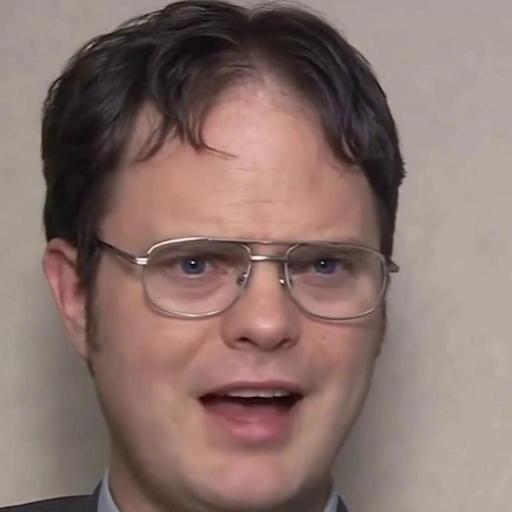} & 
        \includegraphics[width=0.215\columnwidth]{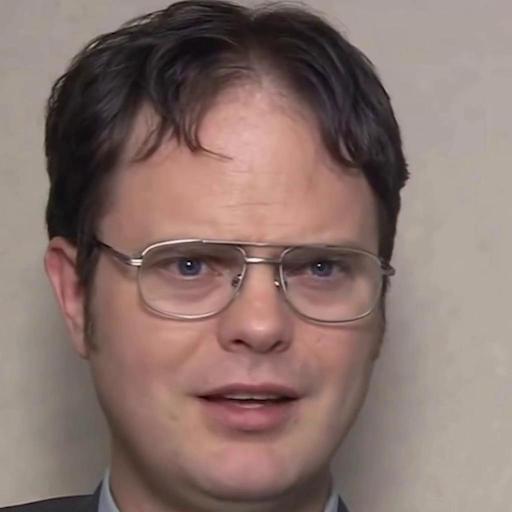} & 
        \includegraphics[width=0.215\columnwidth]{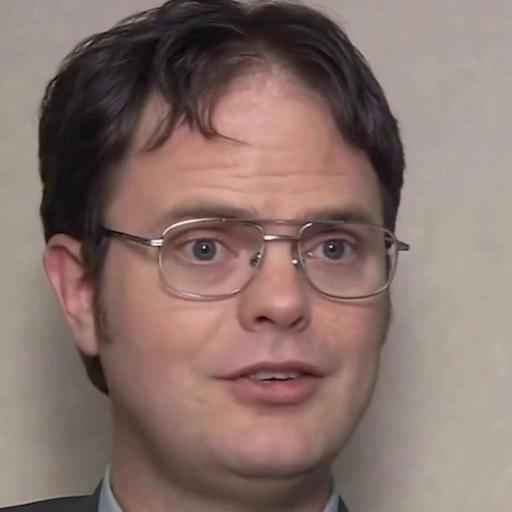} & 
        \includegraphics[width=0.215\columnwidth]{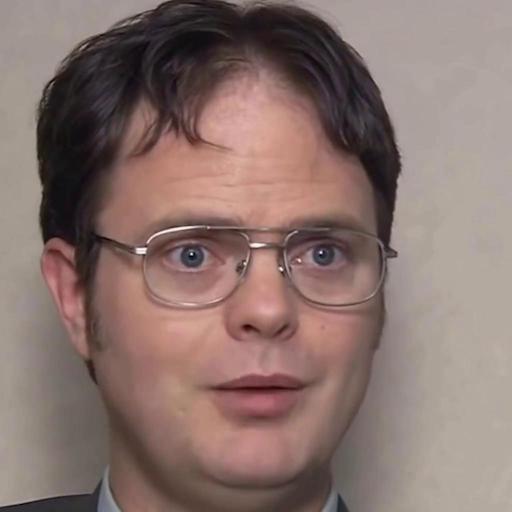} & 
        \includegraphics[width=0.215\columnwidth]{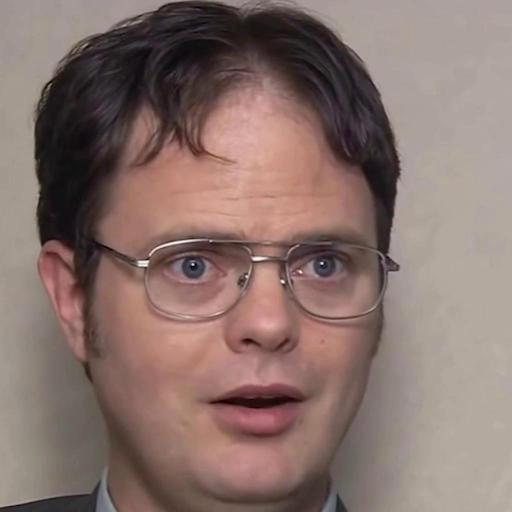} & 
        \includegraphics[width=0.215\columnwidth]{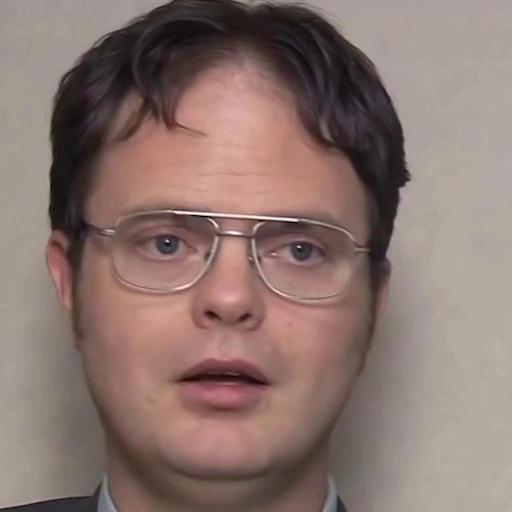} \\

		\raisebox{0.05in}{\rotatebox{90}{Reconstruction}} &
        \includegraphics[width=0.215\columnwidth]{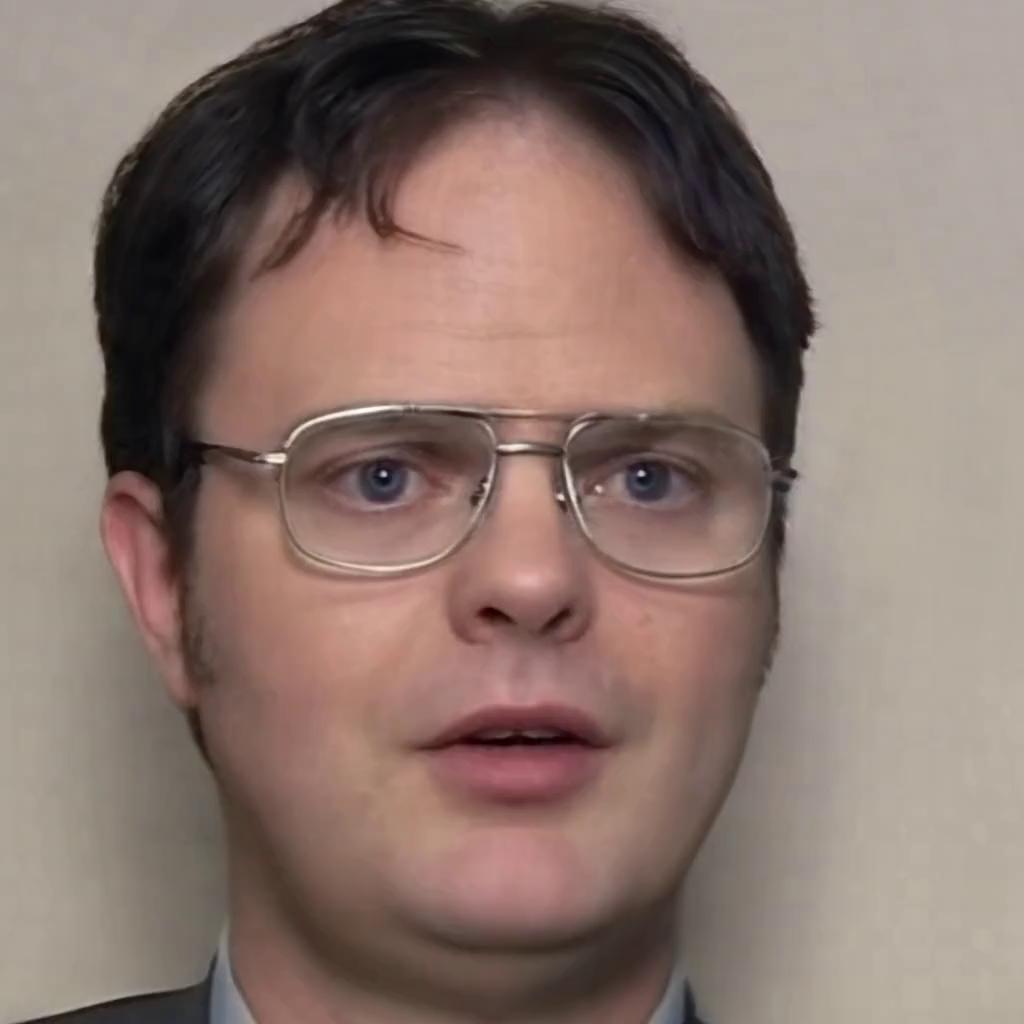} & 
        \includegraphics[width=0.215\columnwidth]{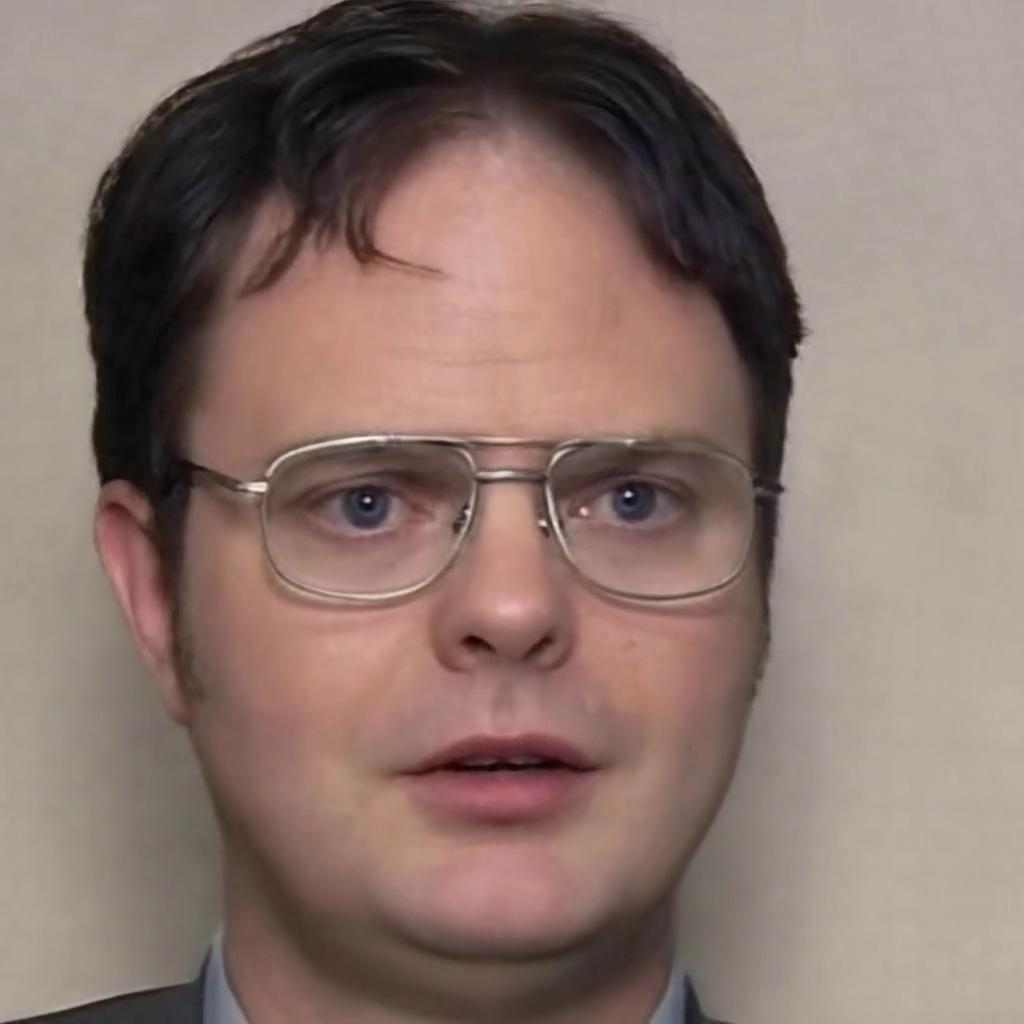} & 
        \includegraphics[width=0.215\columnwidth]{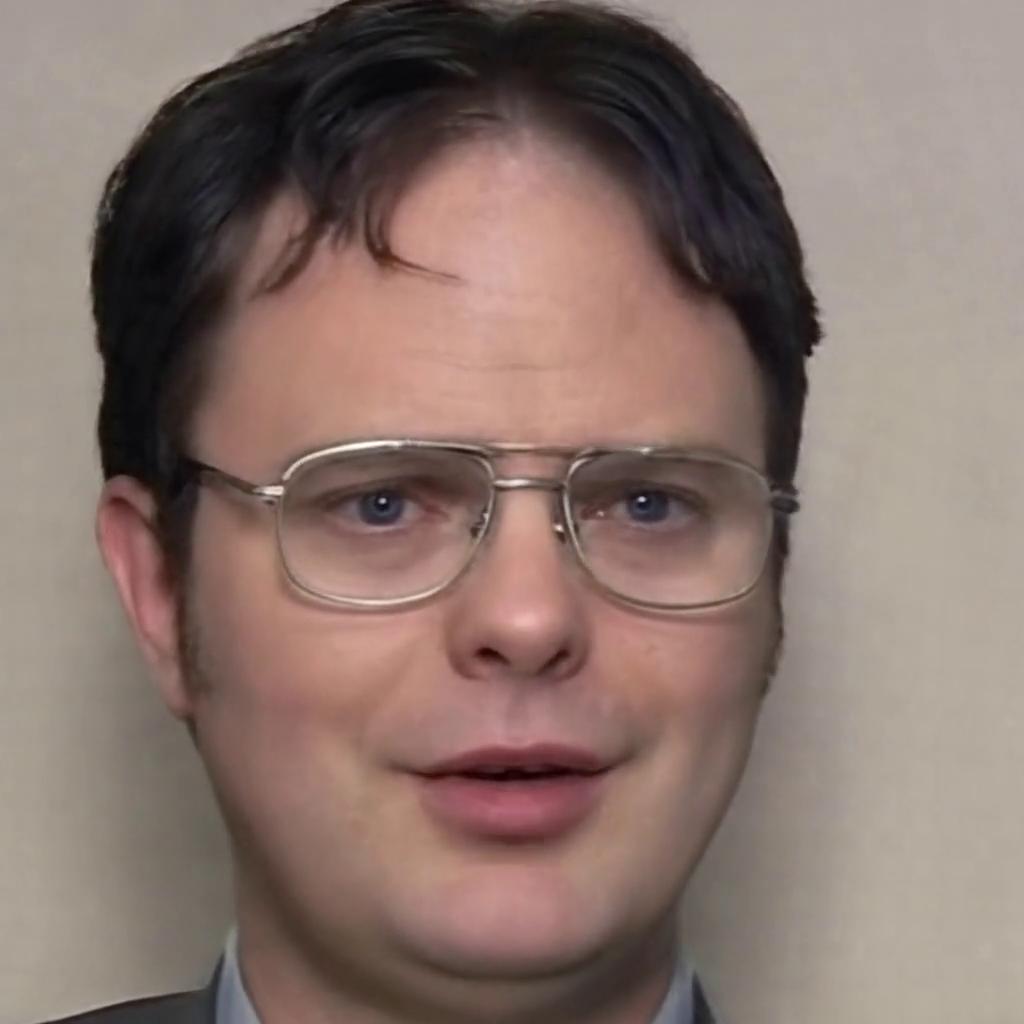} & 
        \includegraphics[width=0.215\columnwidth]{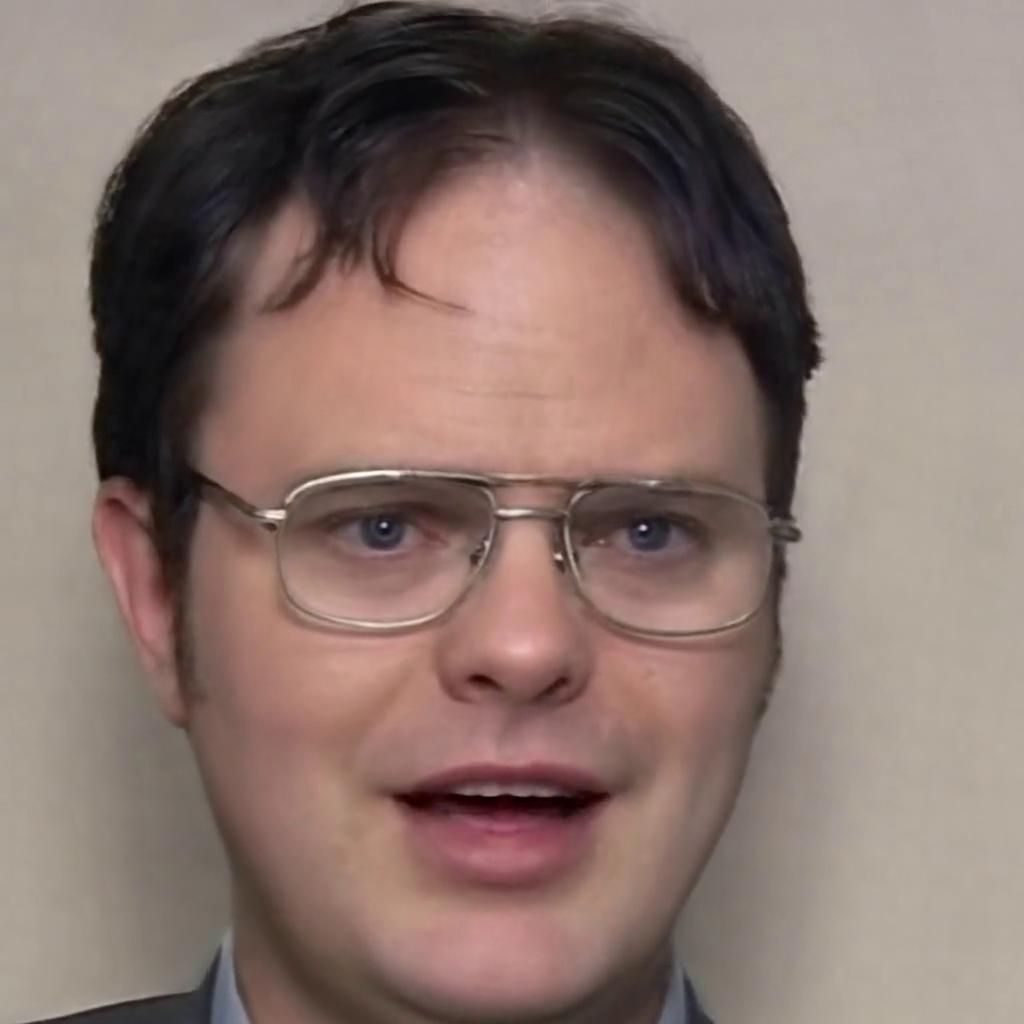} & 
        \includegraphics[width=0.215\columnwidth]{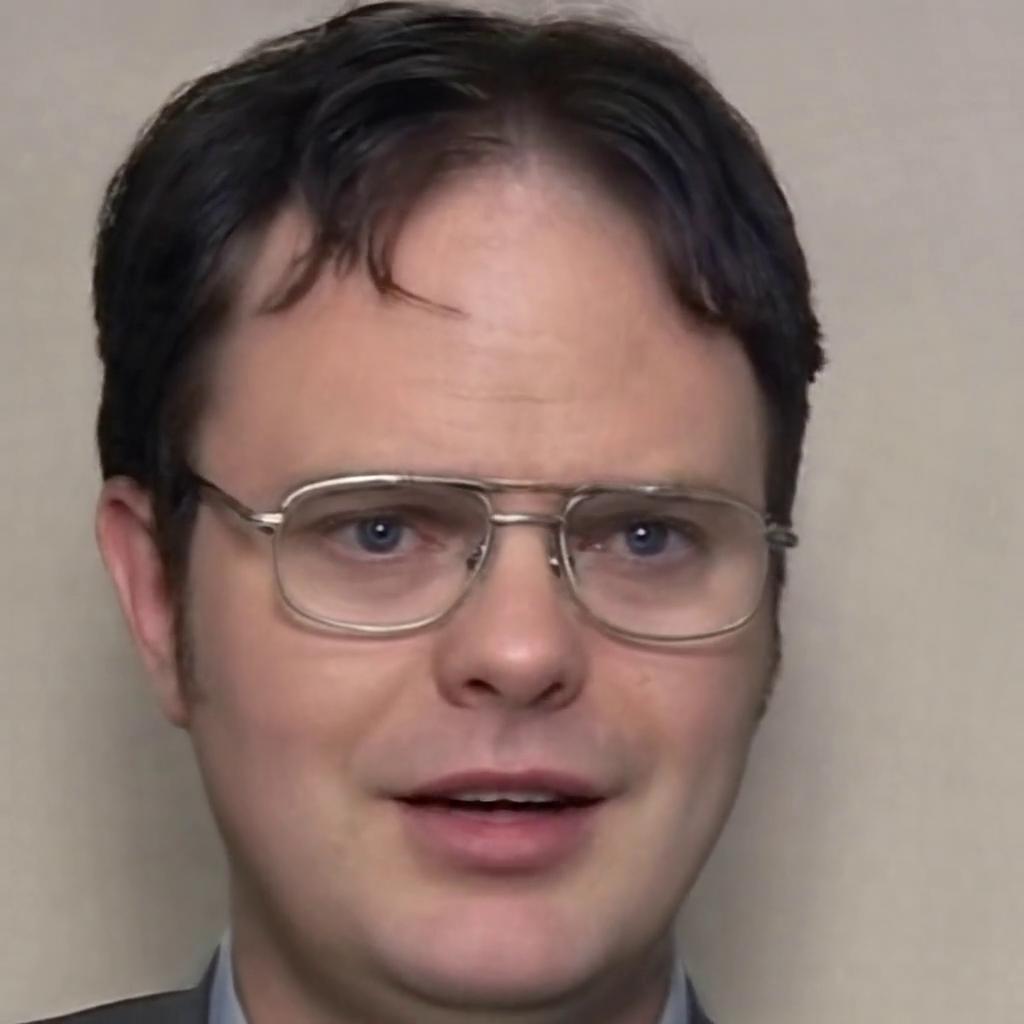} & 
        \includegraphics[width=0.215\columnwidth]{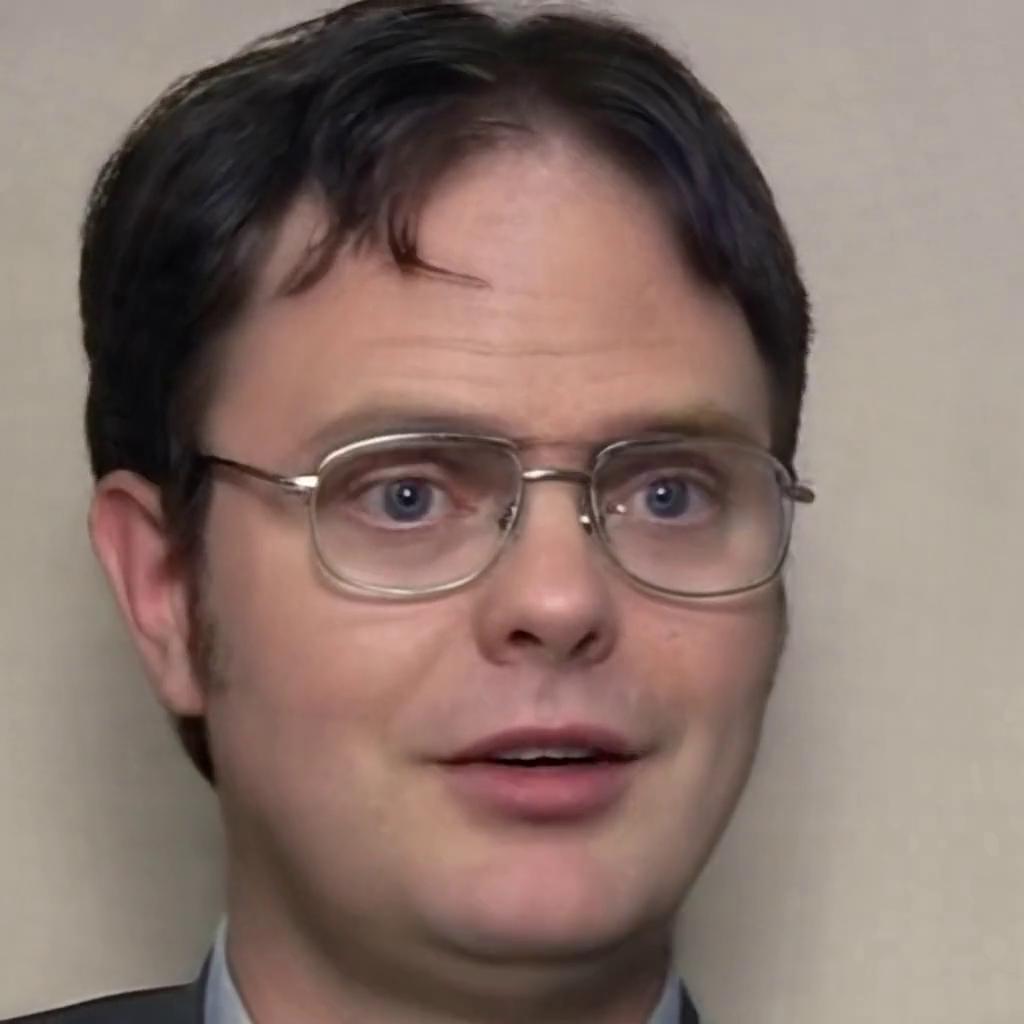} & 
        \includegraphics[width=0.215\columnwidth]{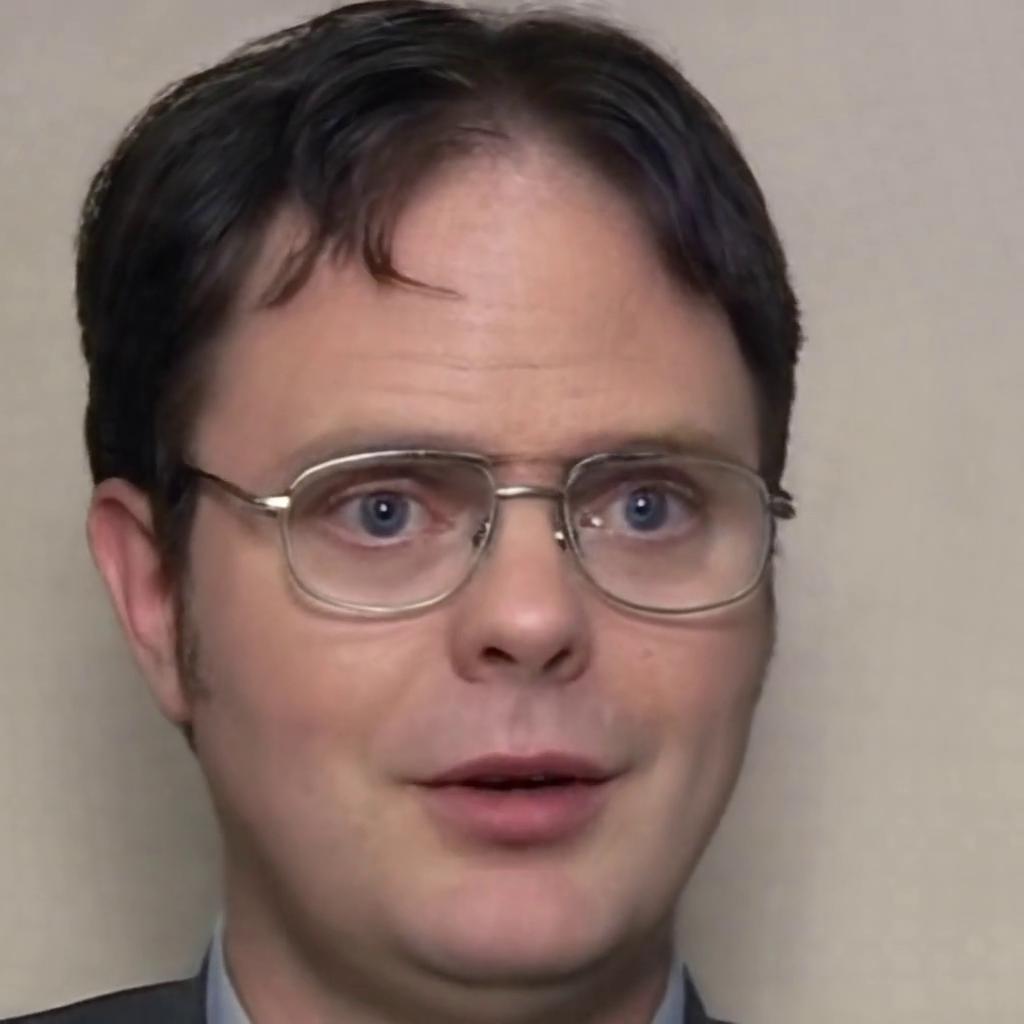} & 
        \includegraphics[width=0.215\columnwidth]{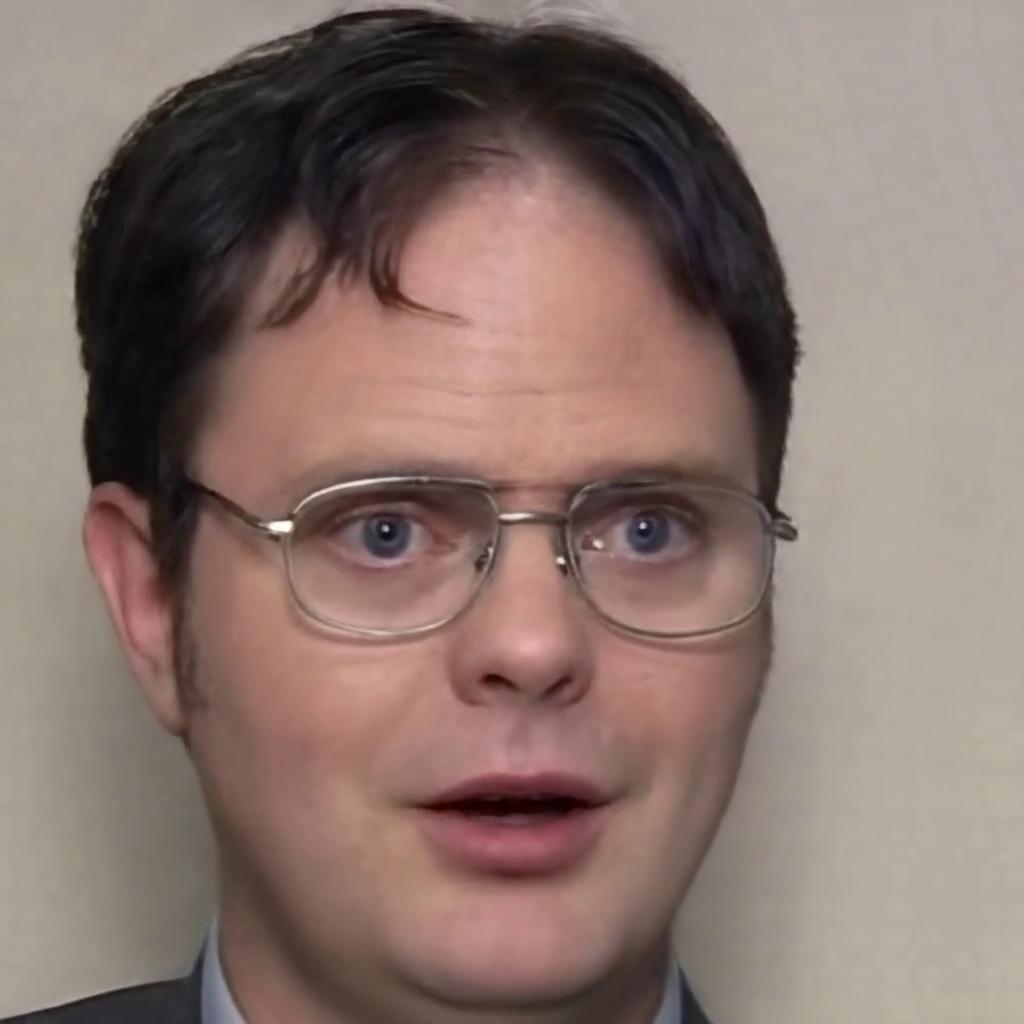} & 
        \includegraphics[width=0.215\columnwidth]{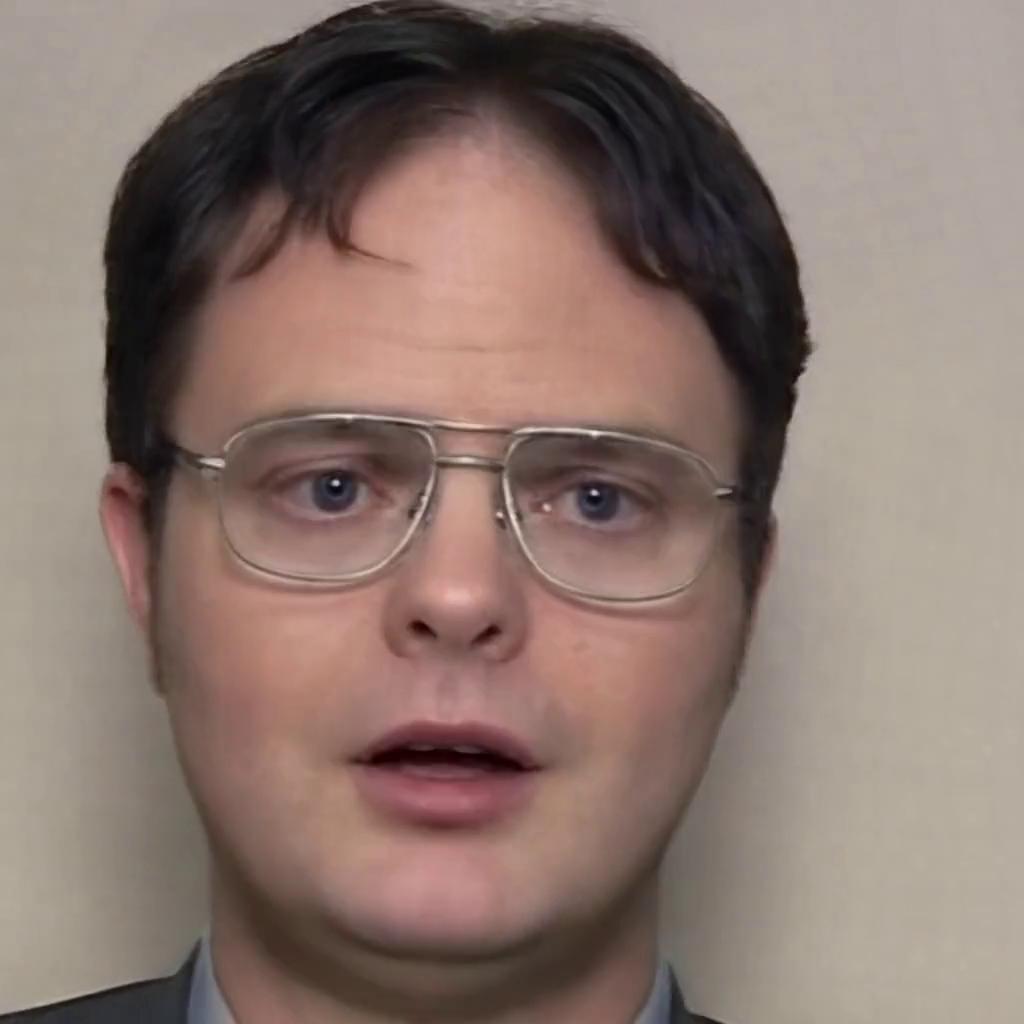} \\

		\raisebox{0.2in}{\rotatebox{90}{$-$ Age}} &
        \includegraphics[width=0.215\columnwidth]{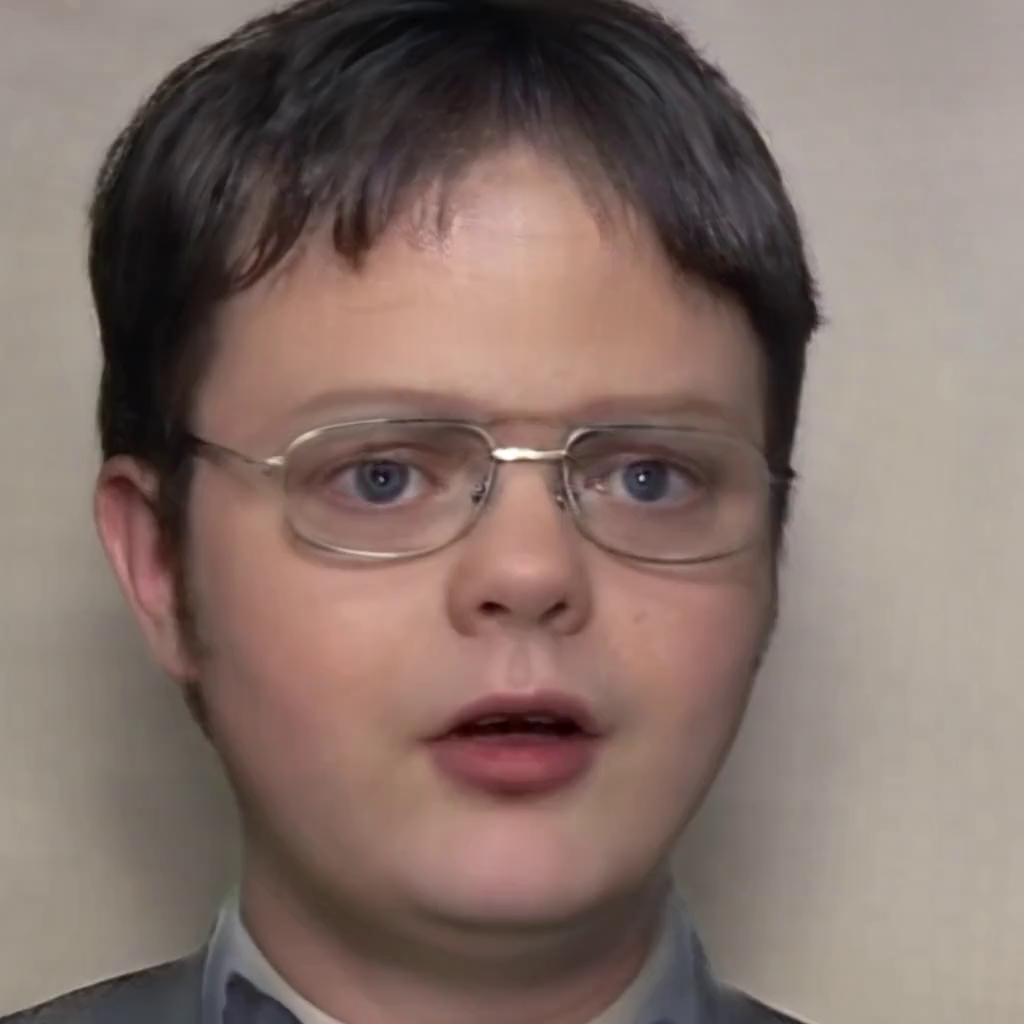} & 
        \includegraphics[width=0.215\columnwidth]{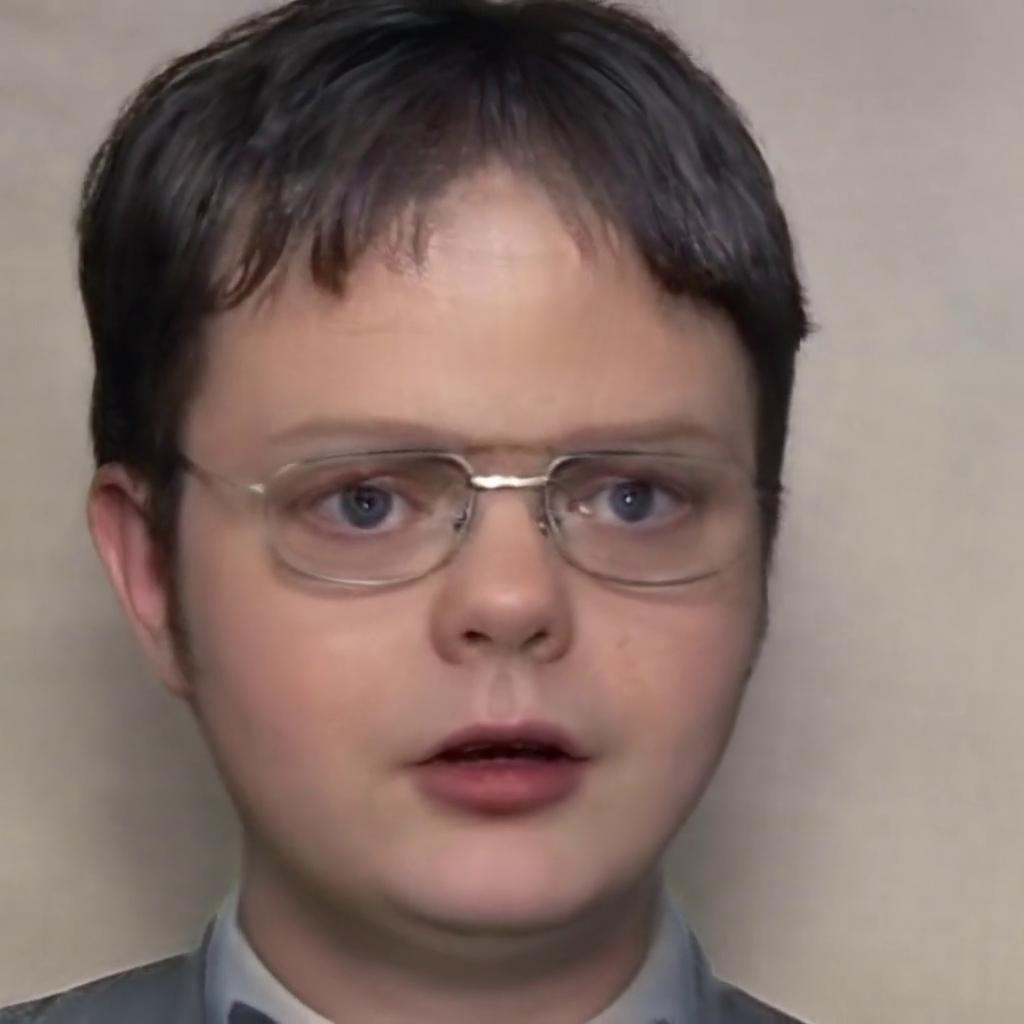} & 
        \includegraphics[width=0.215\columnwidth]{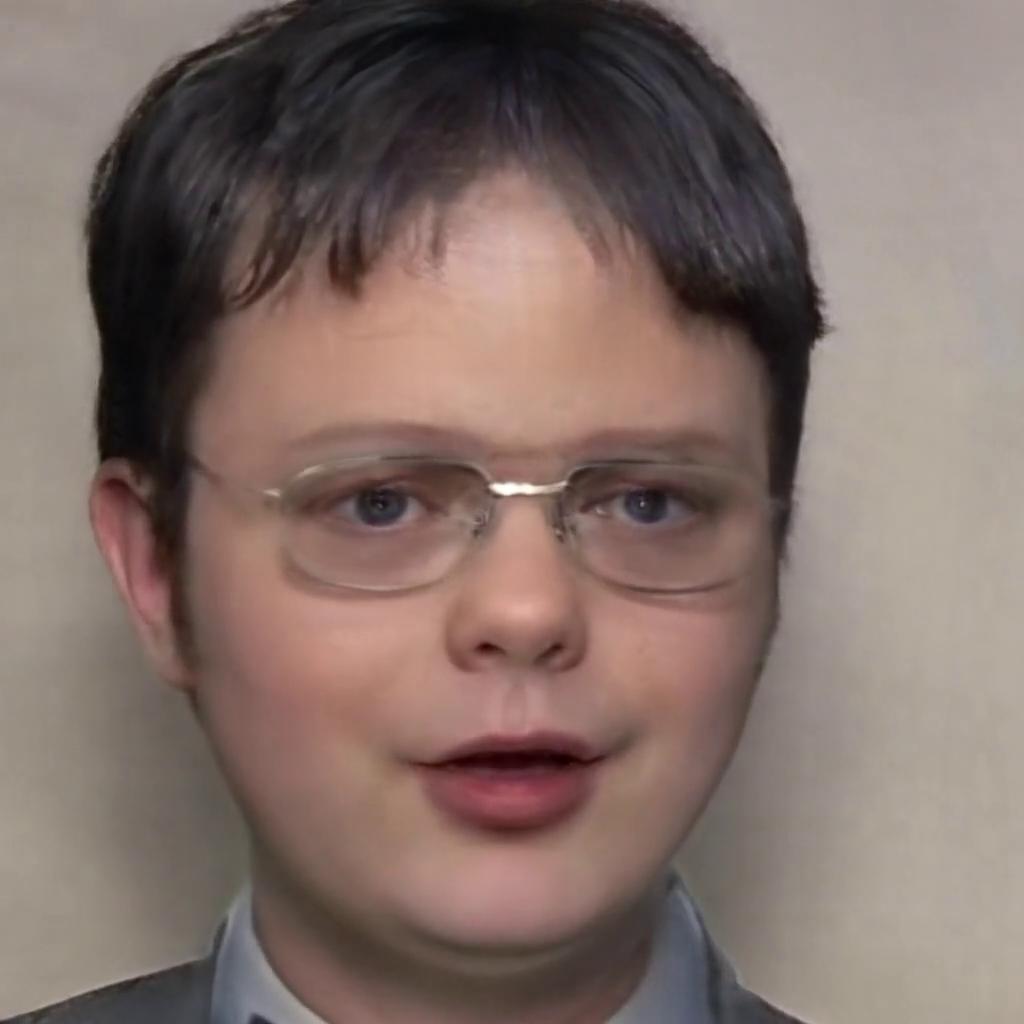} & 
        \includegraphics[width=0.215\columnwidth]{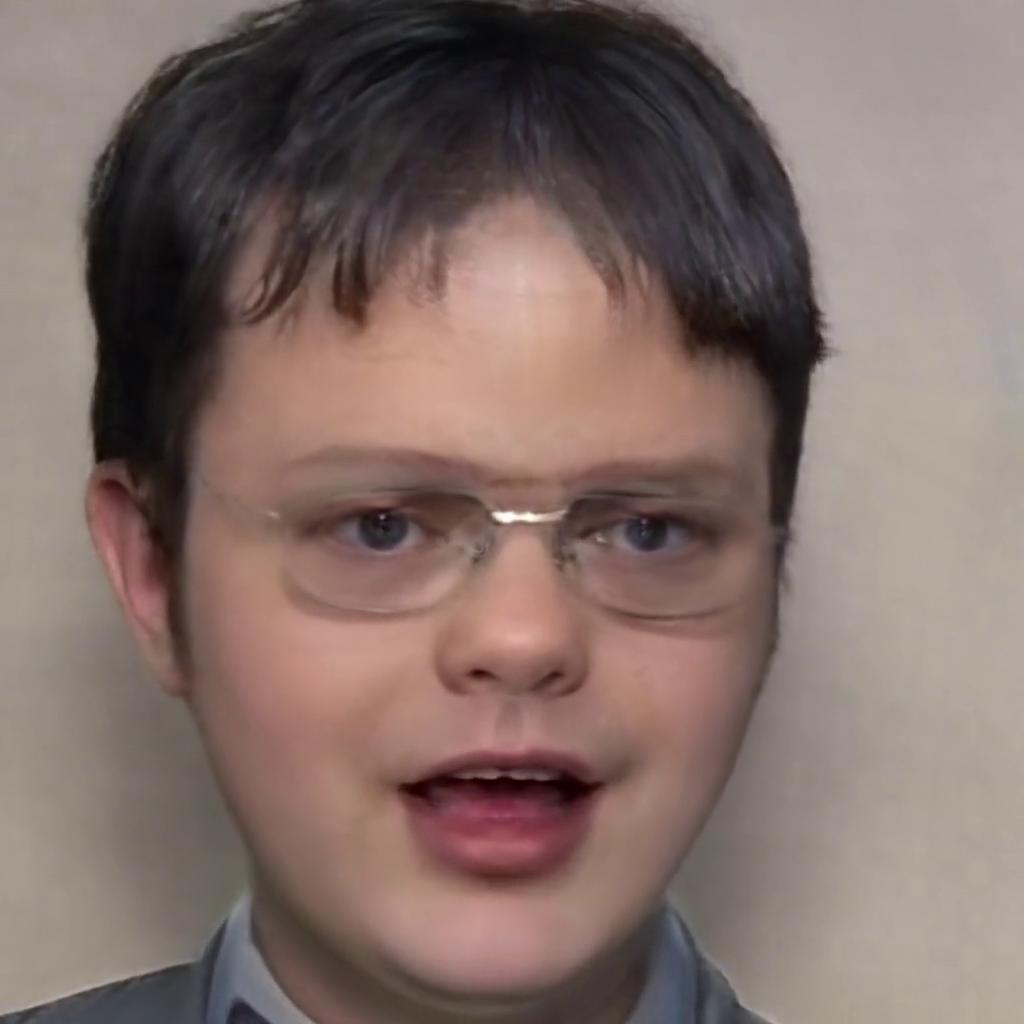} & 
        \includegraphics[width=0.215\columnwidth]{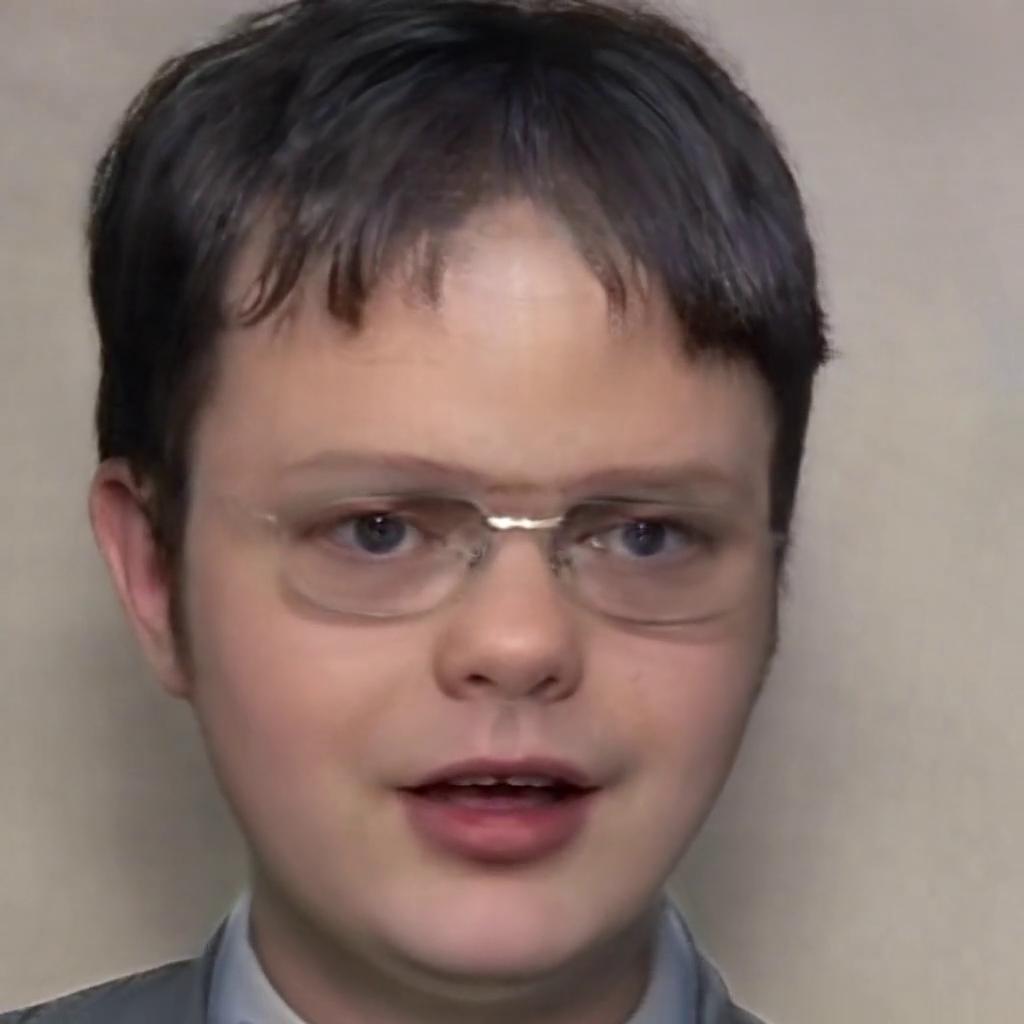} & 
        \includegraphics[width=0.215\columnwidth]{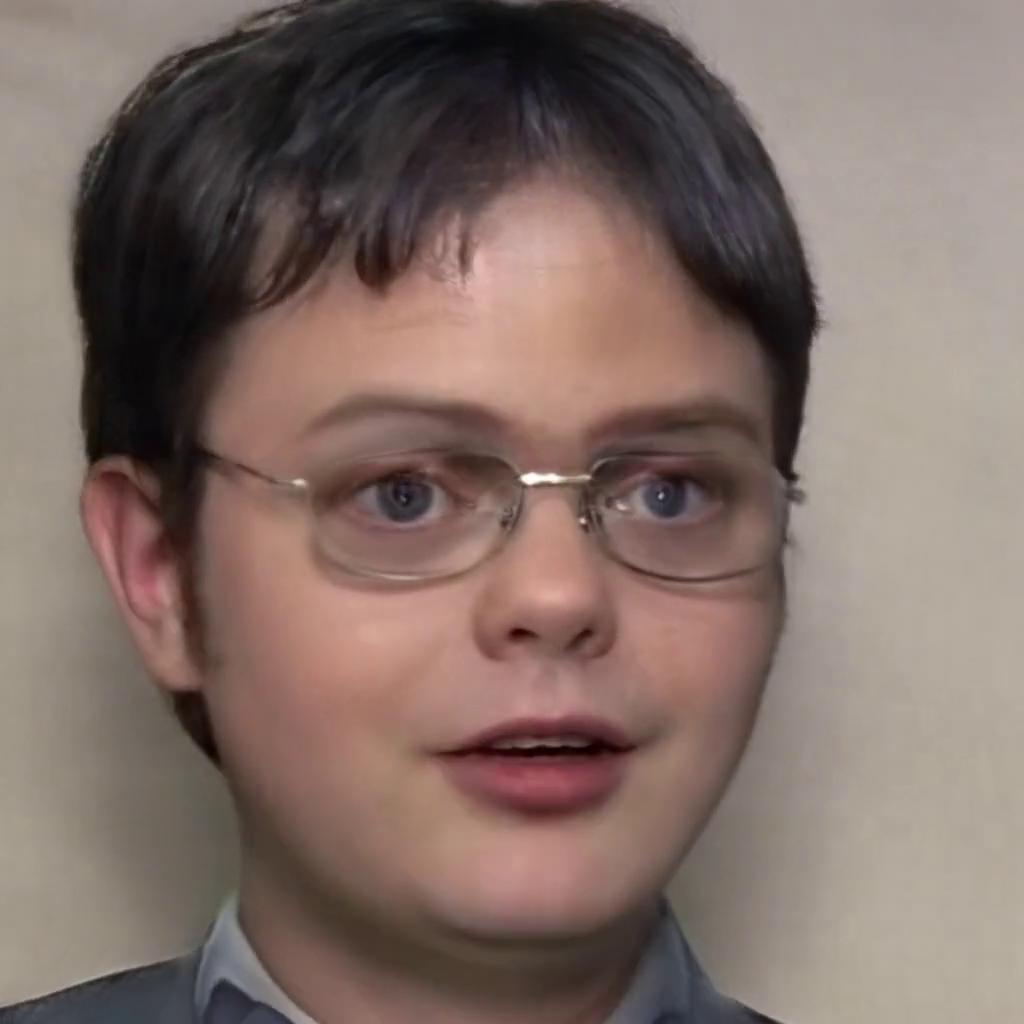} & 
        \includegraphics[width=0.215\columnwidth]{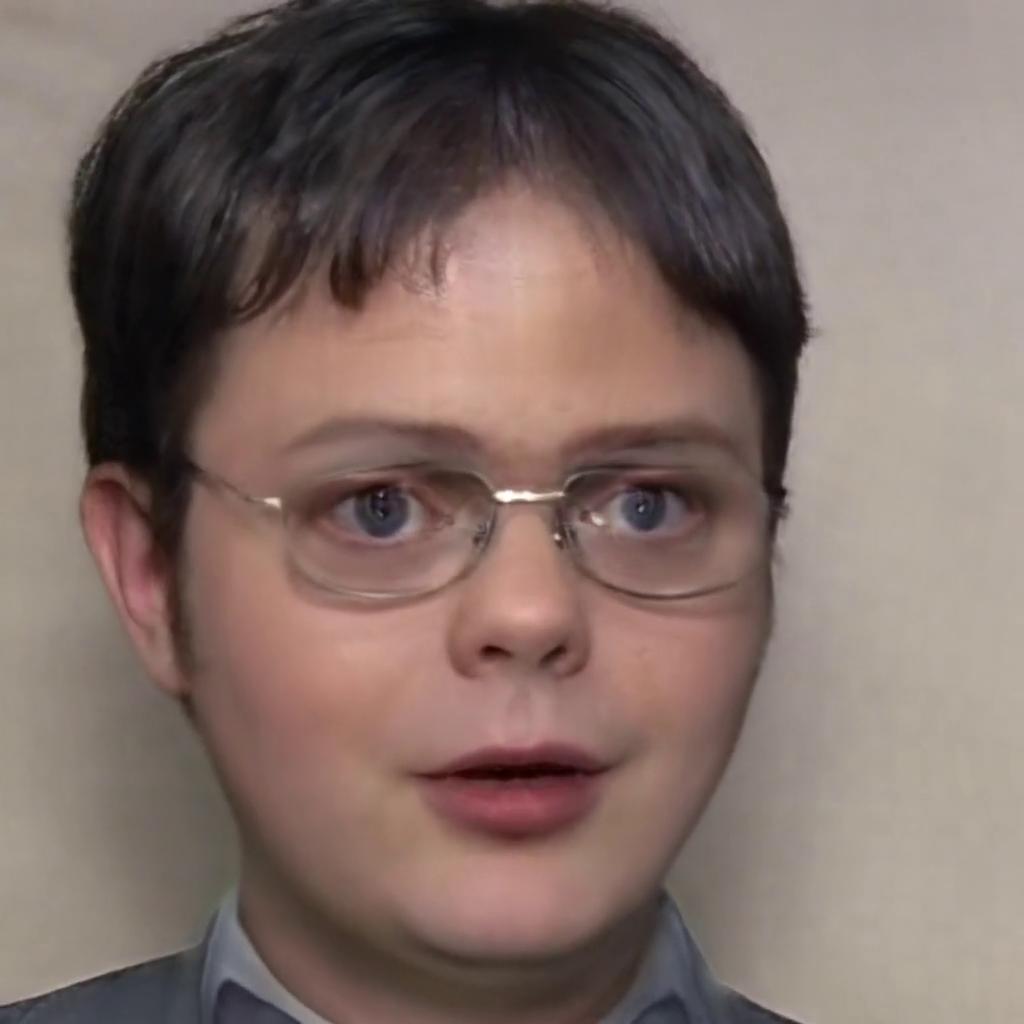} & 
        \includegraphics[width=0.215\columnwidth]{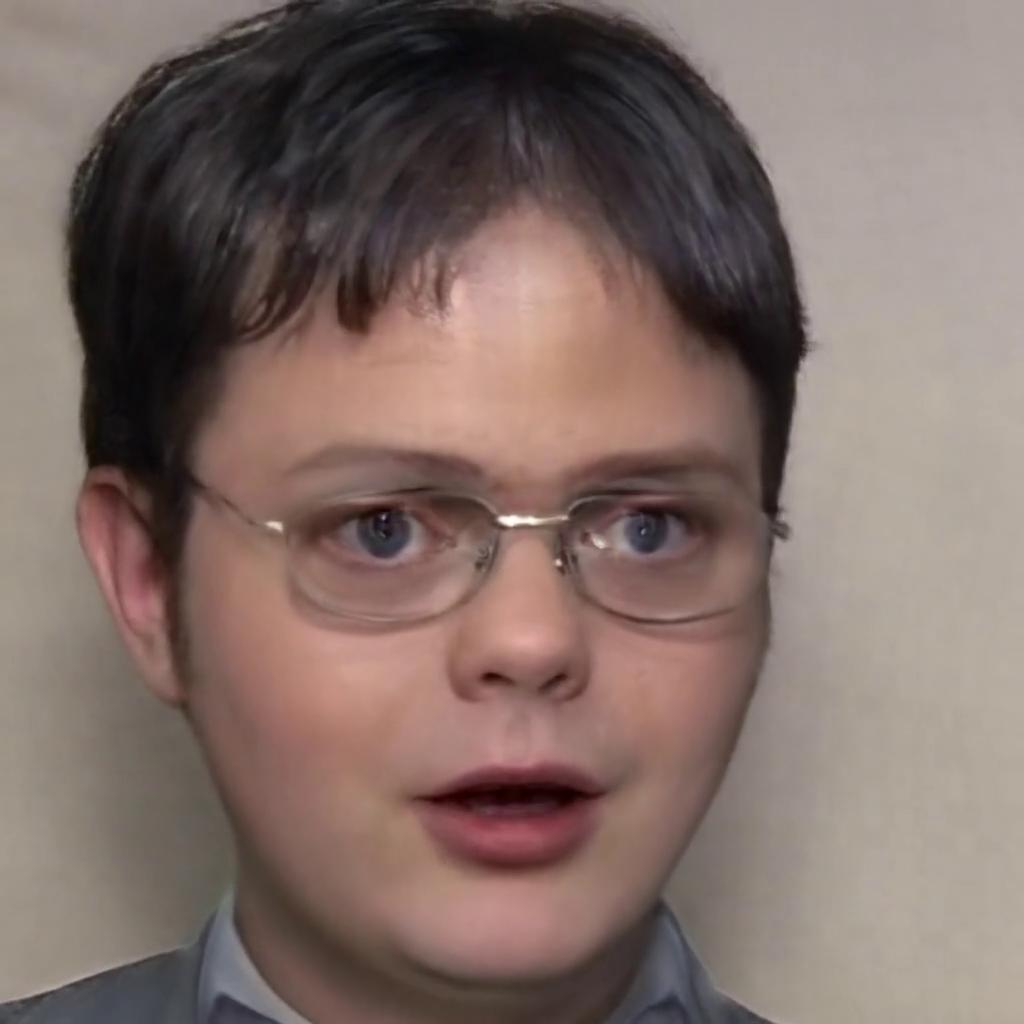} & 
        \includegraphics[width=0.215\columnwidth]{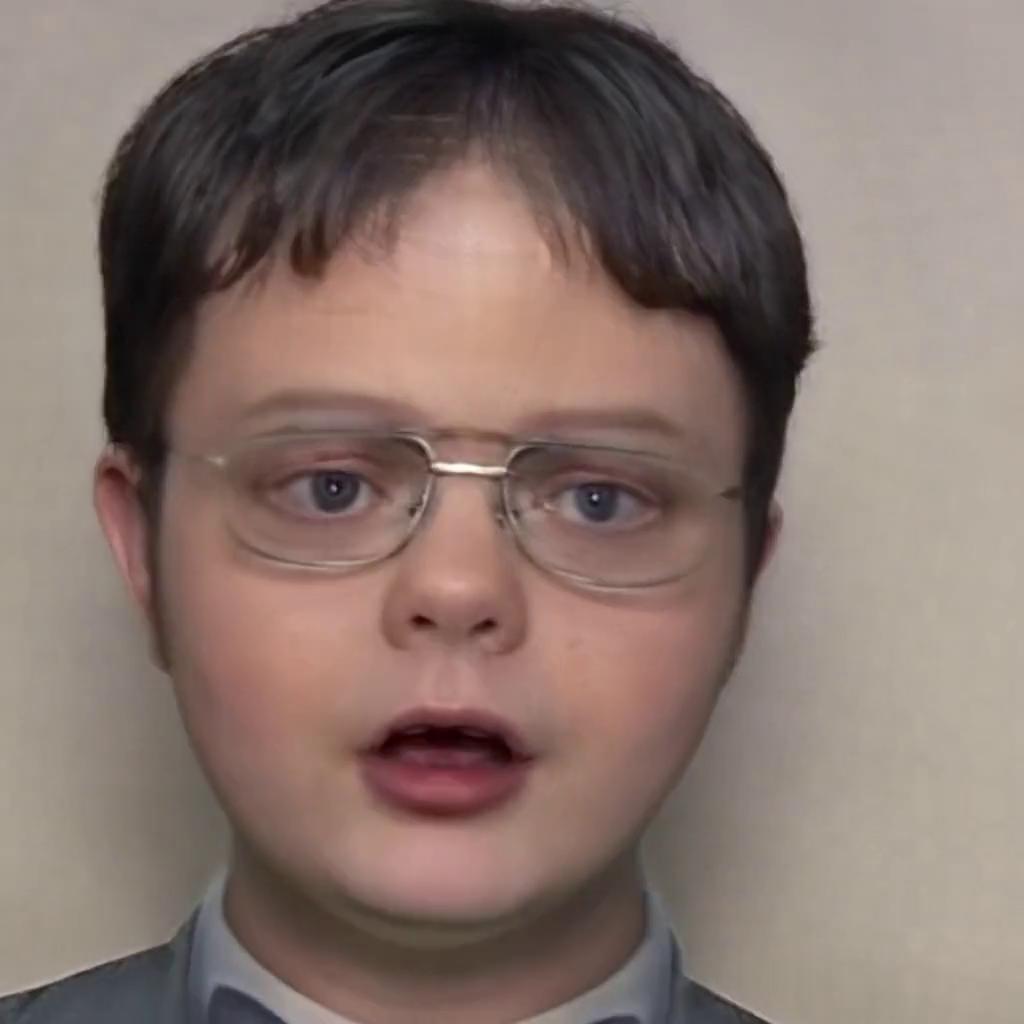} \\
        
		\raisebox{0.15in}{\rotatebox{90}{$+$ Sketch}} &
        \includegraphics[width=0.215\columnwidth]{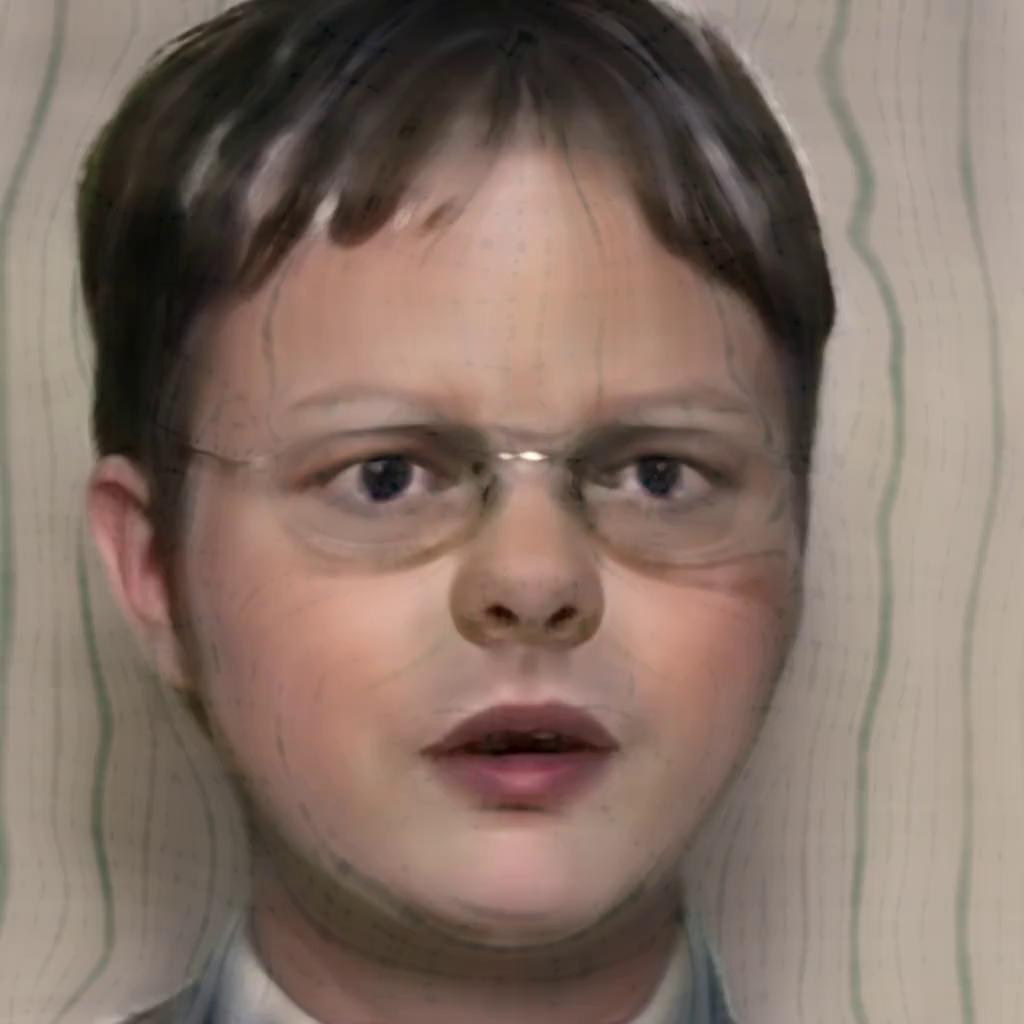} & 
        \includegraphics[width=0.215\columnwidth]{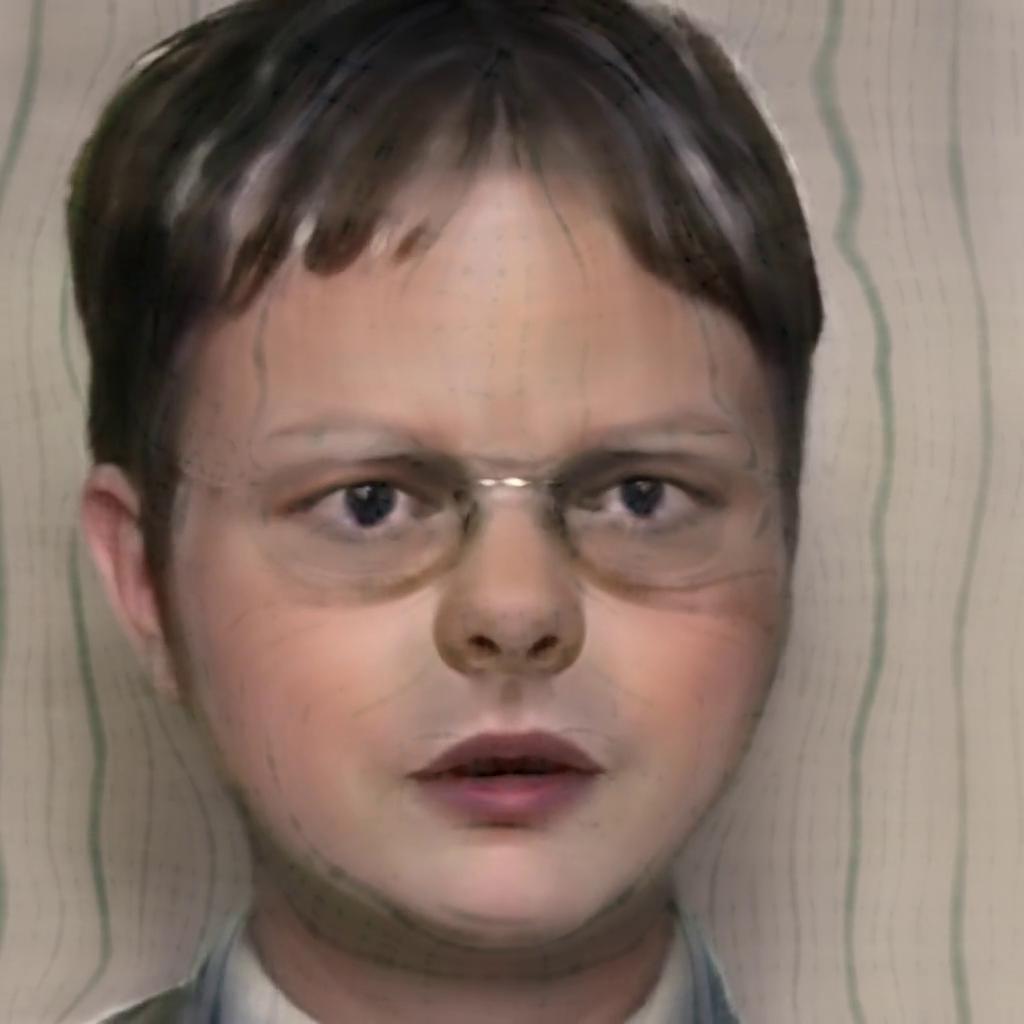} & 
        \includegraphics[width=0.215\columnwidth]{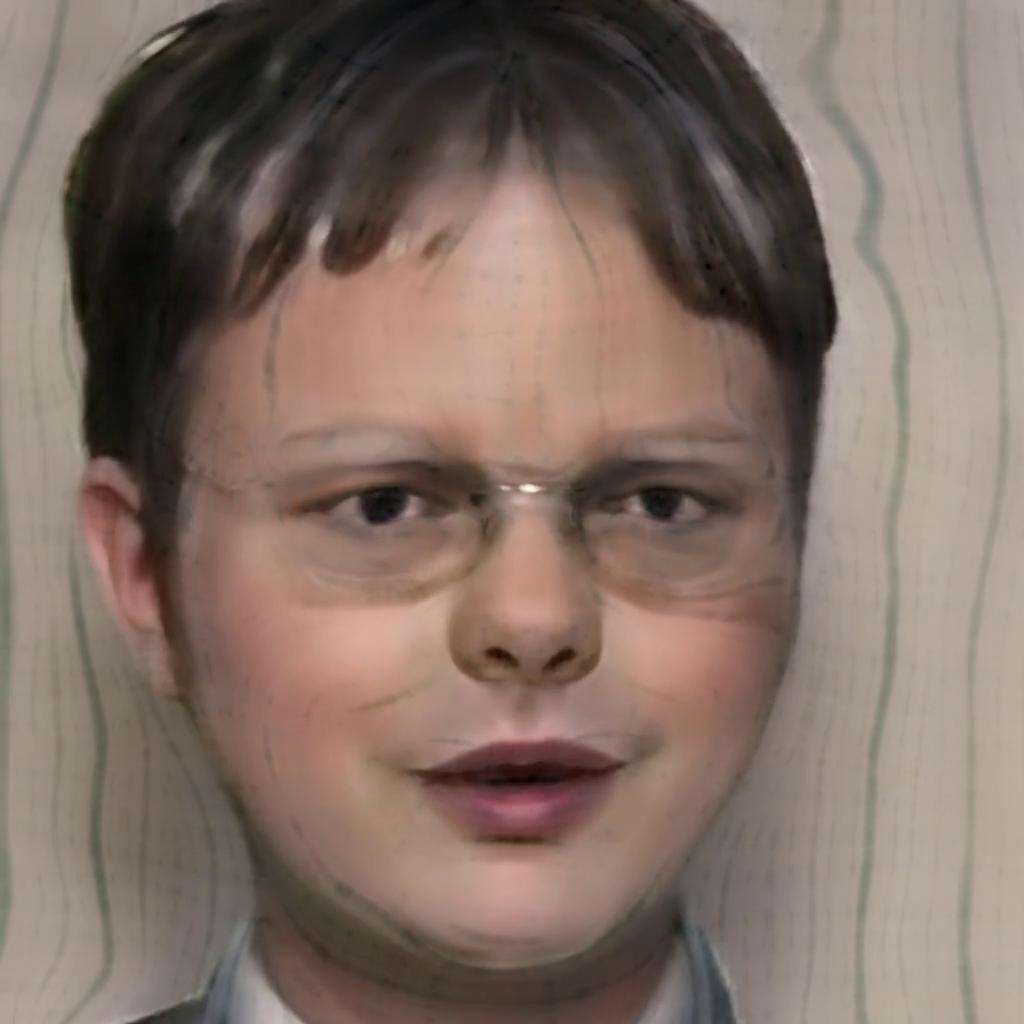} & 
        \includegraphics[width=0.215\columnwidth]{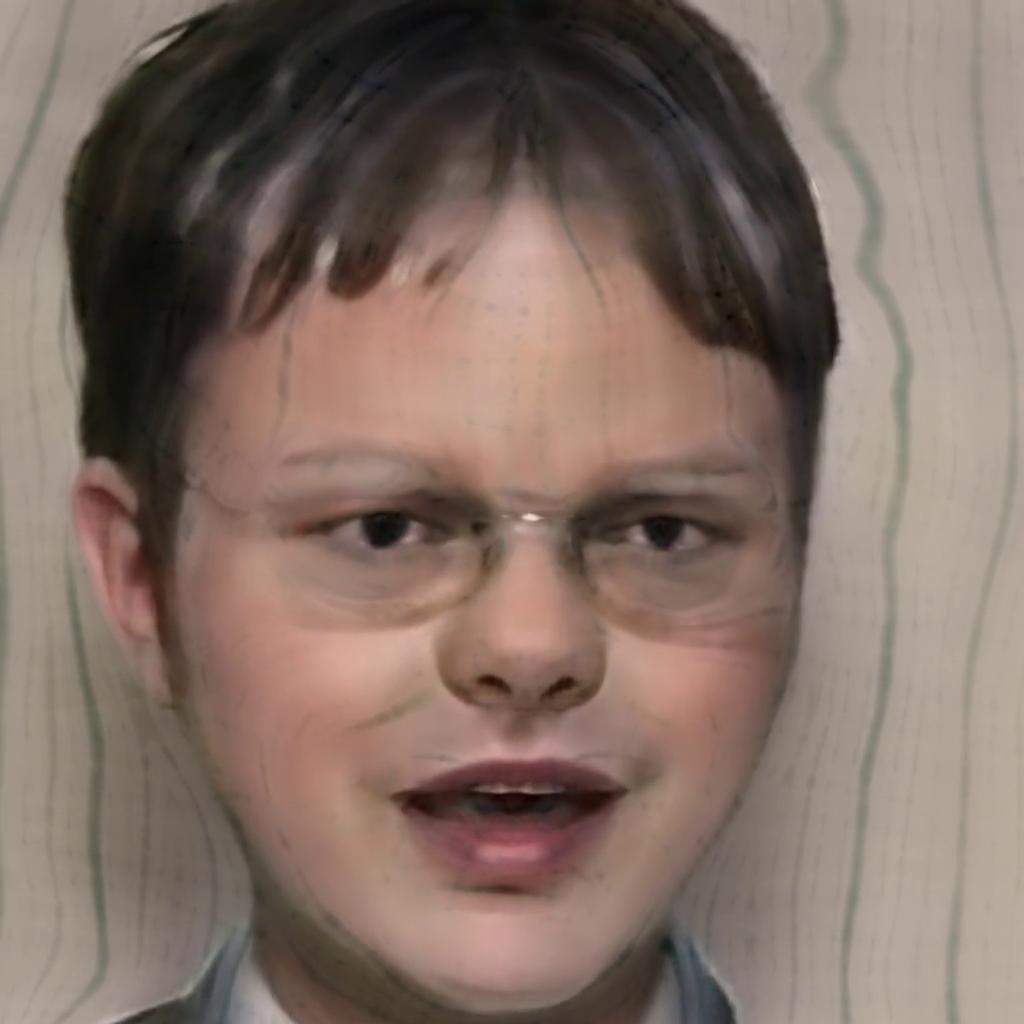} & 
        \includegraphics[width=0.215\columnwidth]{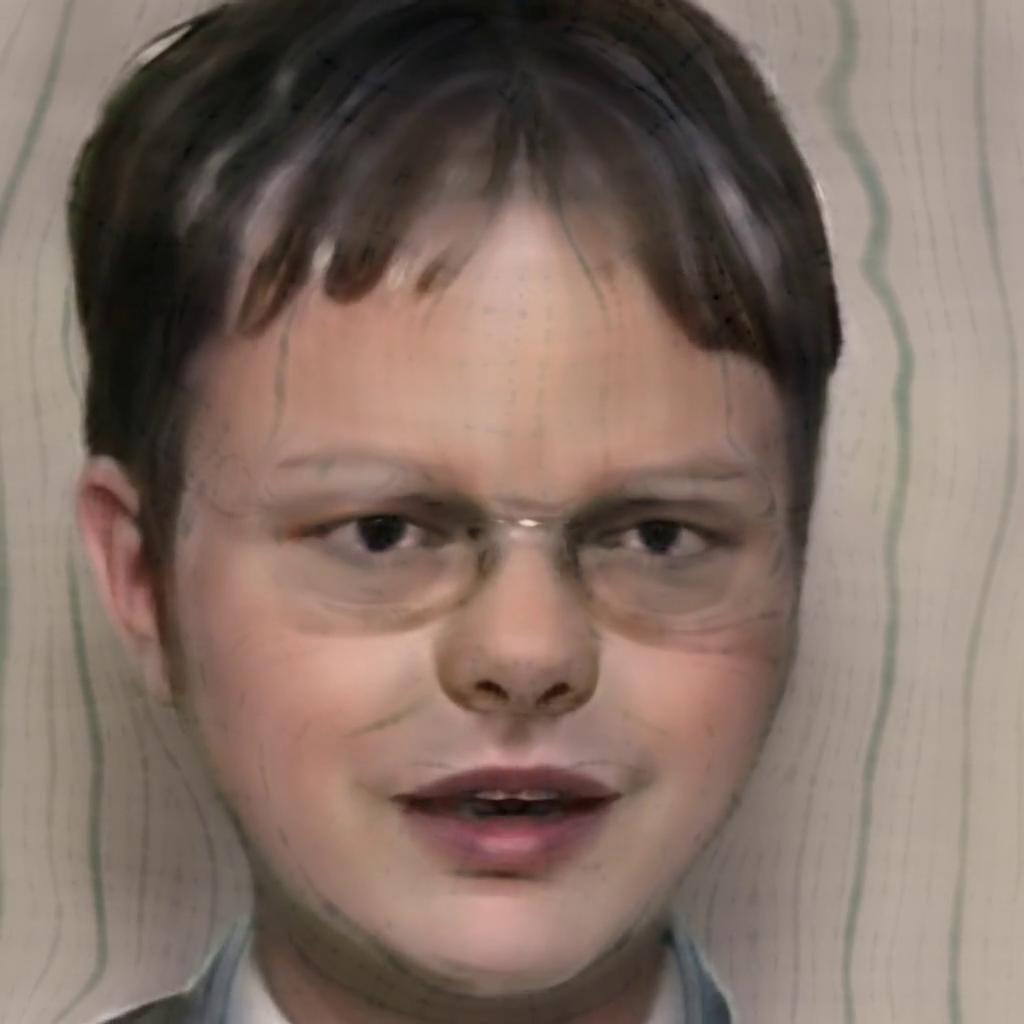} & 
        \includegraphics[width=0.215\columnwidth]{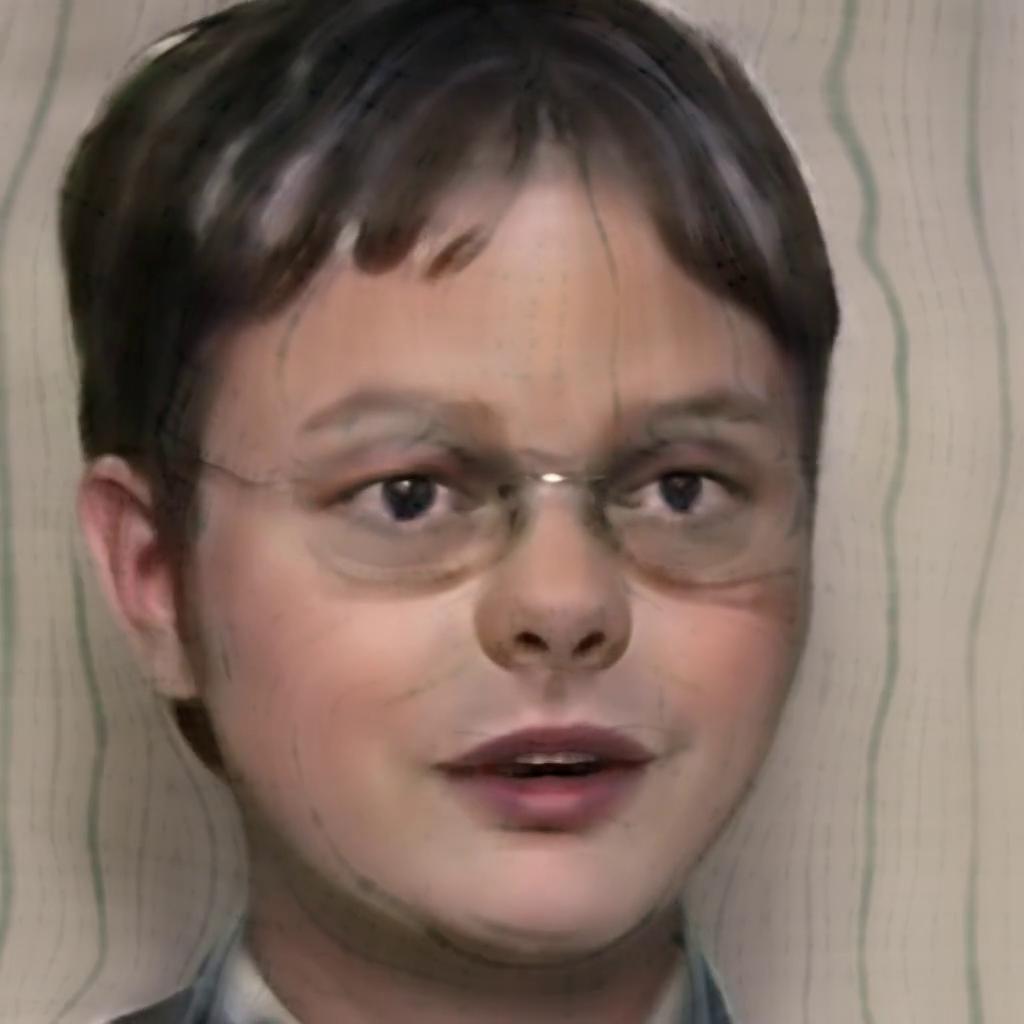} & 
        \includegraphics[width=0.215\columnwidth]{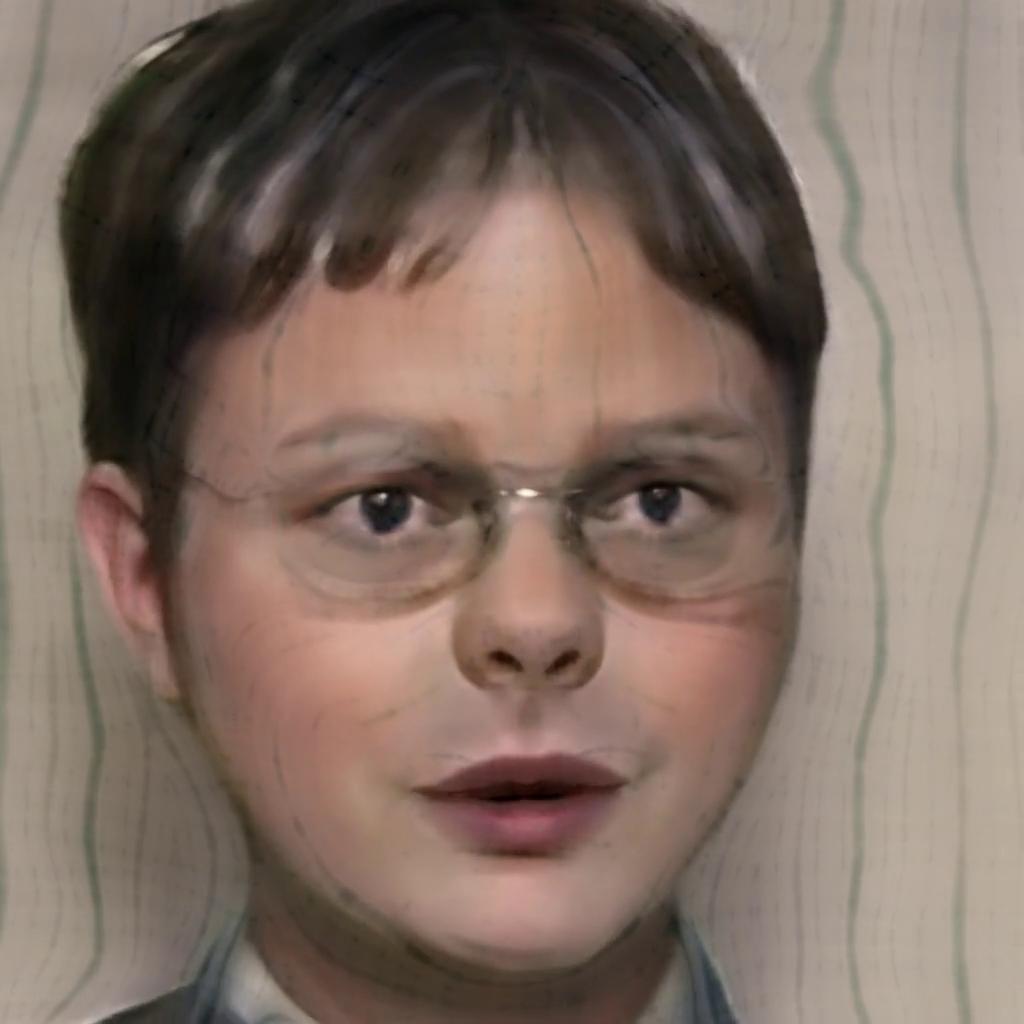} & 
        \includegraphics[width=0.215\columnwidth]{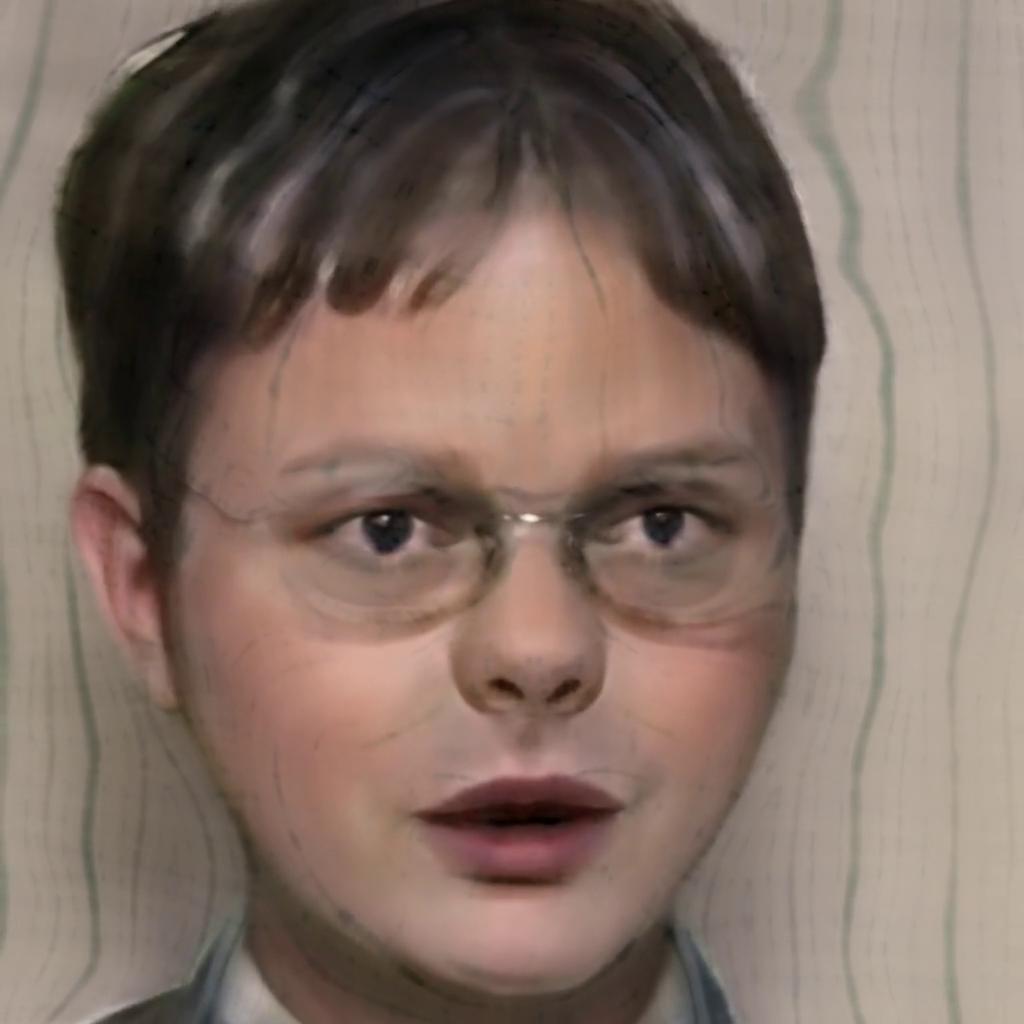} & 
        \includegraphics[width=0.215\columnwidth]{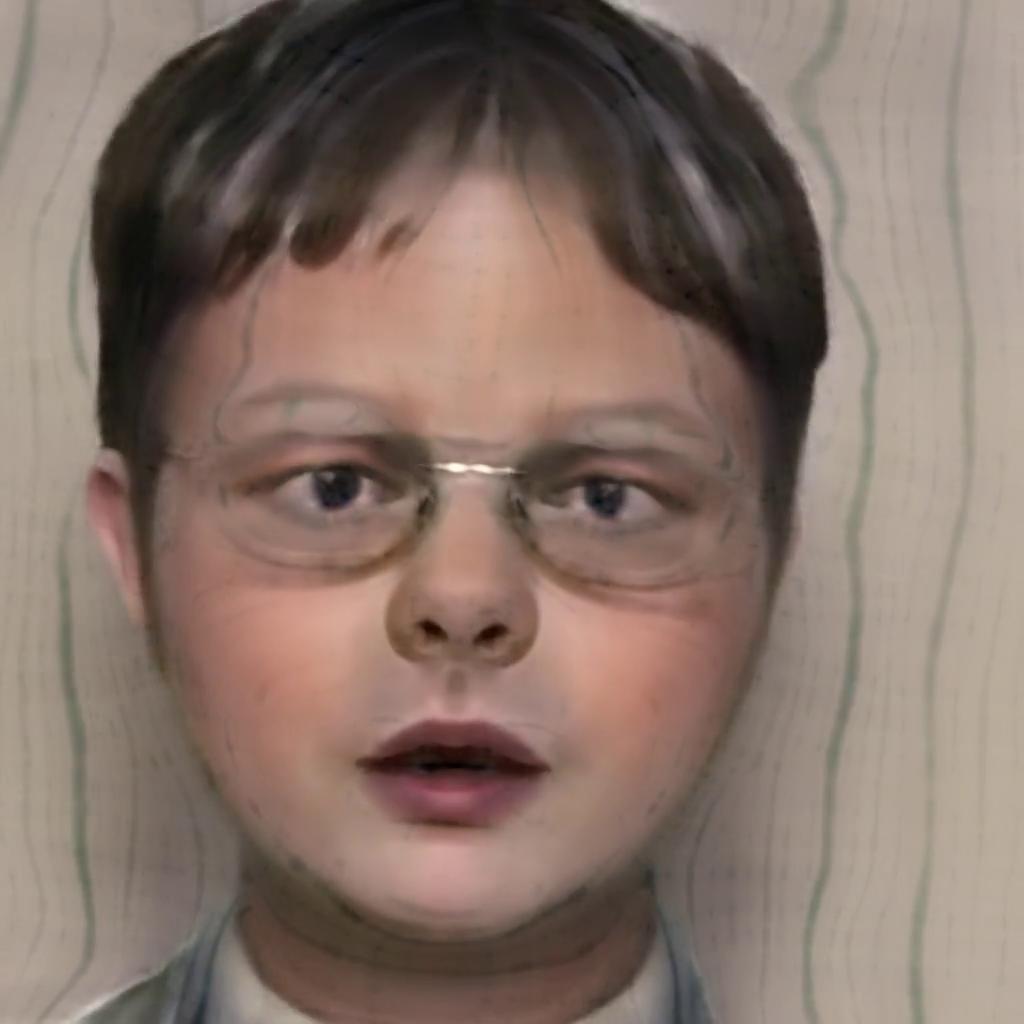} \\

	\end{tabular}
	}
	
	\vspace{-0.3cm}
	\caption{
	Sample results of our full video encoding and editing pipeline with StyleGAN3. The ``Double Chin'' edit is obtained with StyeCLIP's global direction technique~\cite{patashnik2021styleclip} while the ``Age'' edit is obtained using InterFaceGAN~\cite{shen2020interpreting}. The ``Pixar'' and ``Sketch'' styles are obtained with StyleGAN-NADA~\cite{gal2021stylegannada}. Note, the StyleGAN-NADA results are shown on the \textit{edited} video frames. Full videos are provided in the supplementary materials.
	}
	\label{fig:video_results_supplementary_2}
\end{figure*}

%% file: figures/supplementary/video_results_wide.tex
\begin{figure*}[tb]
	\centering
	\setlength{\tabcolsep}{1pt}	
	{\footnotesize
	\begin{tabular}{c c c c c c c c c c}

        \\ 
        \\ 
        \\
        \\ 

		\raisebox{0.15in}{\rotatebox{90}{Original}} &
        \includegraphics[width=0.215\columnwidth]{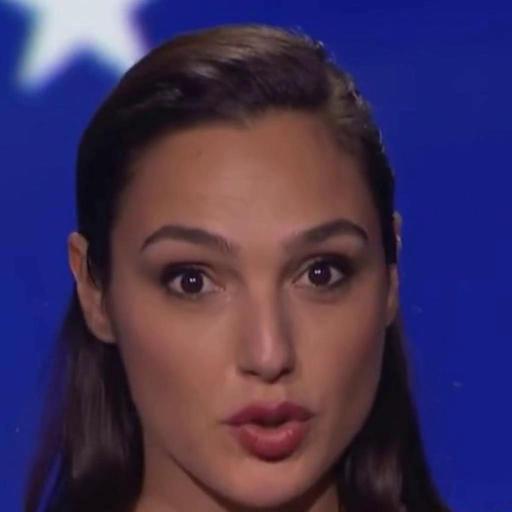} & 
        \includegraphics[width=0.215\columnwidth]{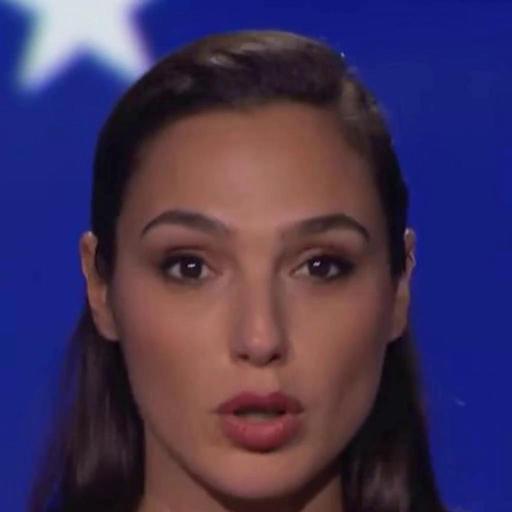} & 
        \includegraphics[width=0.215\columnwidth]{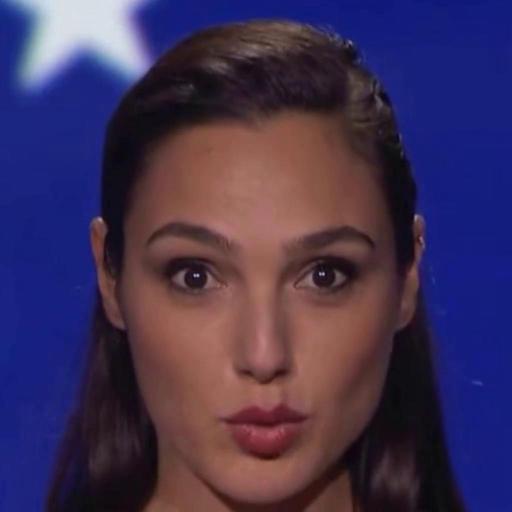} & 
        \includegraphics[width=0.215\columnwidth]{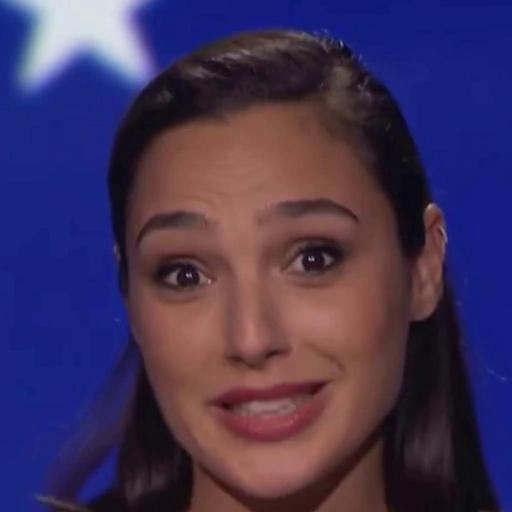} & 
        \includegraphics[width=0.215\columnwidth]{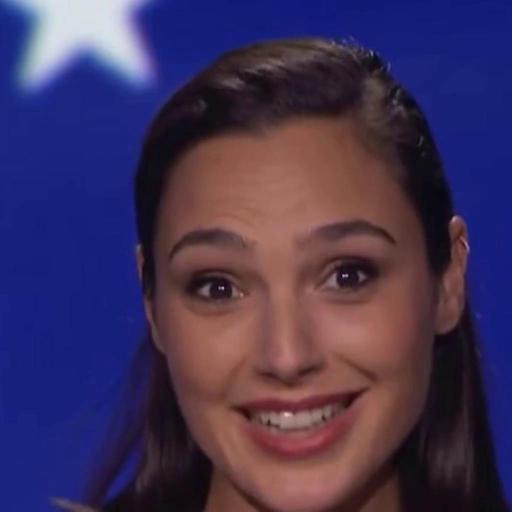} & 
        \includegraphics[width=0.215\columnwidth]{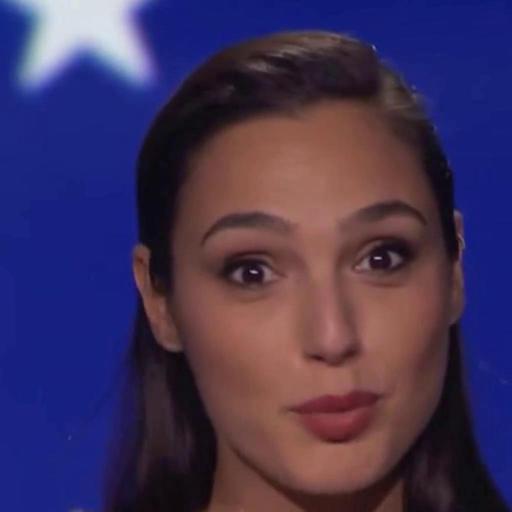} & 
        \includegraphics[width=0.215\columnwidth]{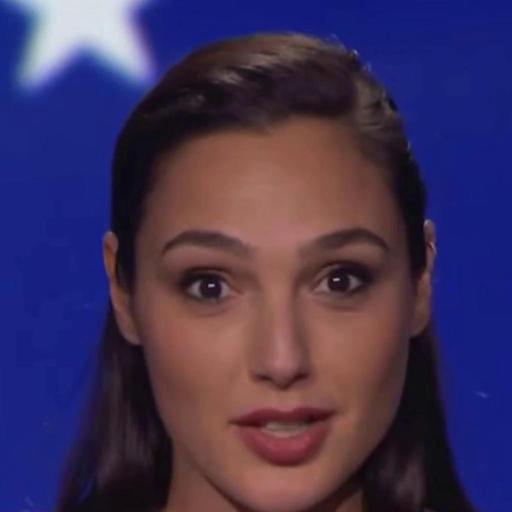} & 
        \includegraphics[width=0.215\columnwidth]{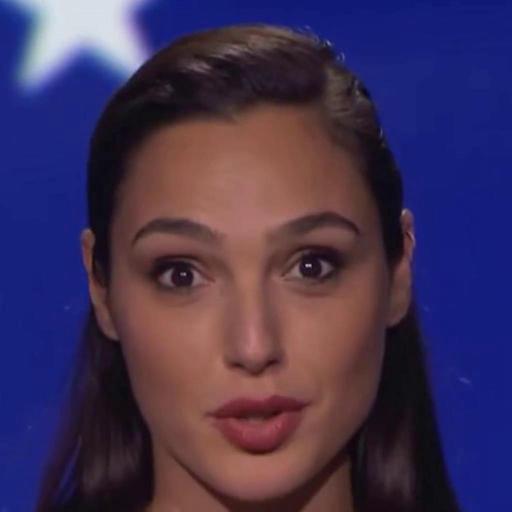} & 
        \includegraphics[width=0.215\columnwidth]{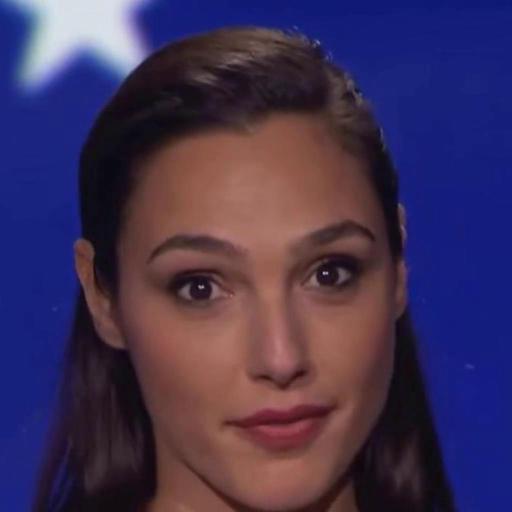} \\

		\raisebox{0.05in}{\rotatebox{90}{Reconstruction}} &
        \includegraphics[width=0.215\columnwidth]{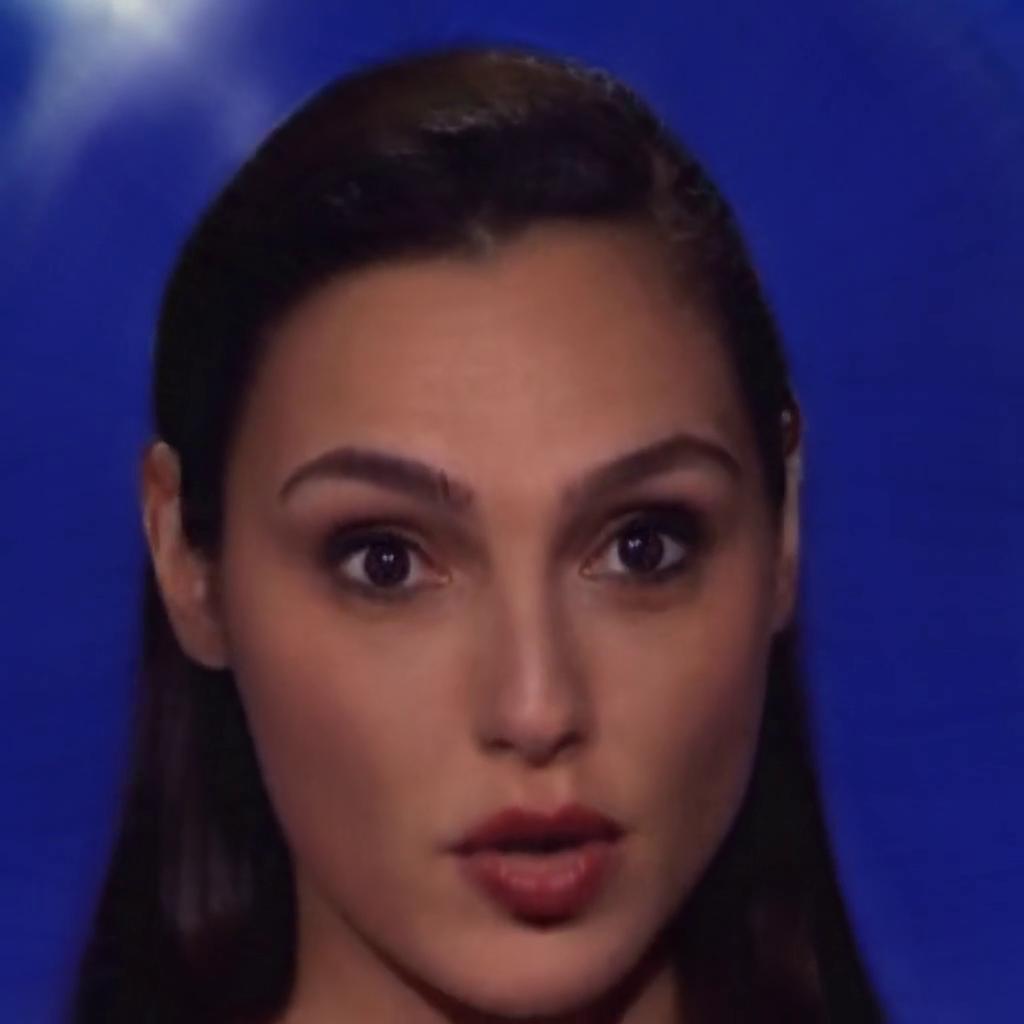} & 
        \includegraphics[width=0.215\columnwidth]{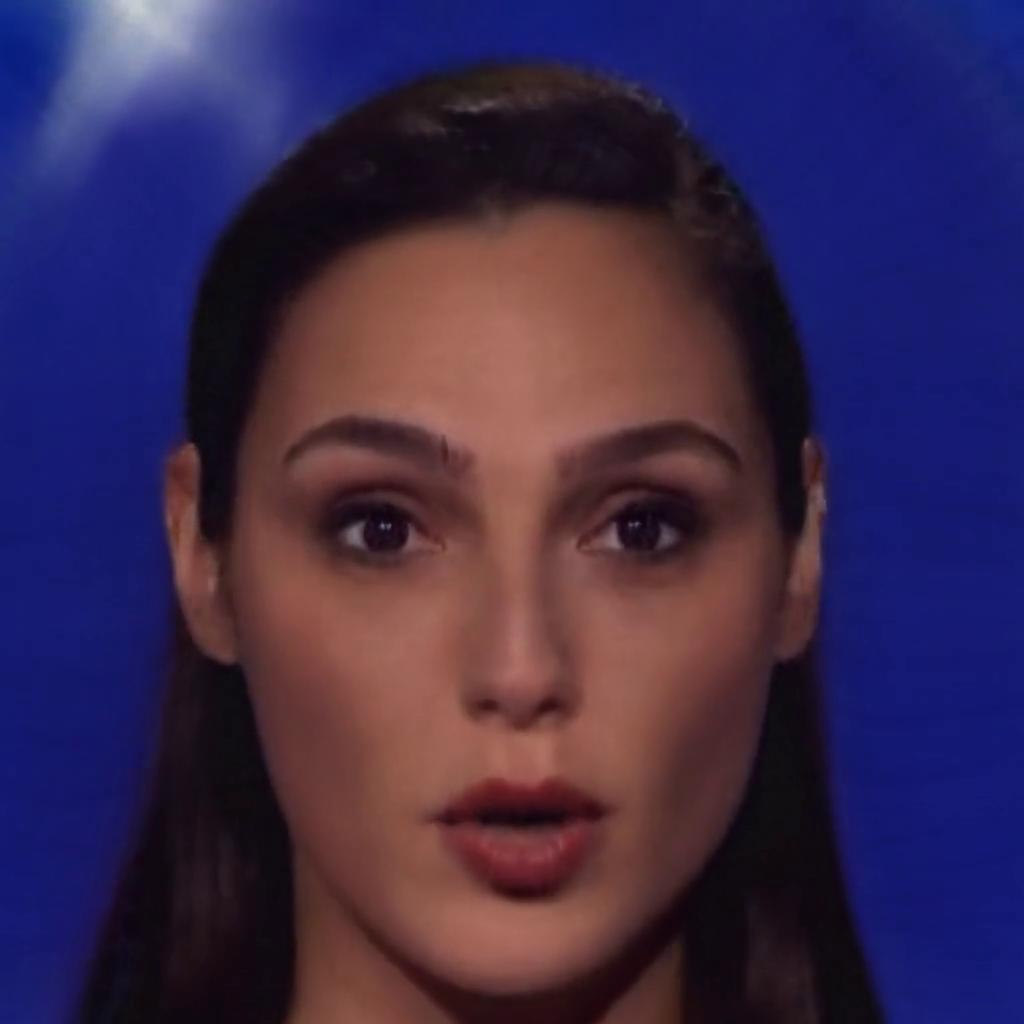} & 
        \includegraphics[width=0.215\columnwidth]{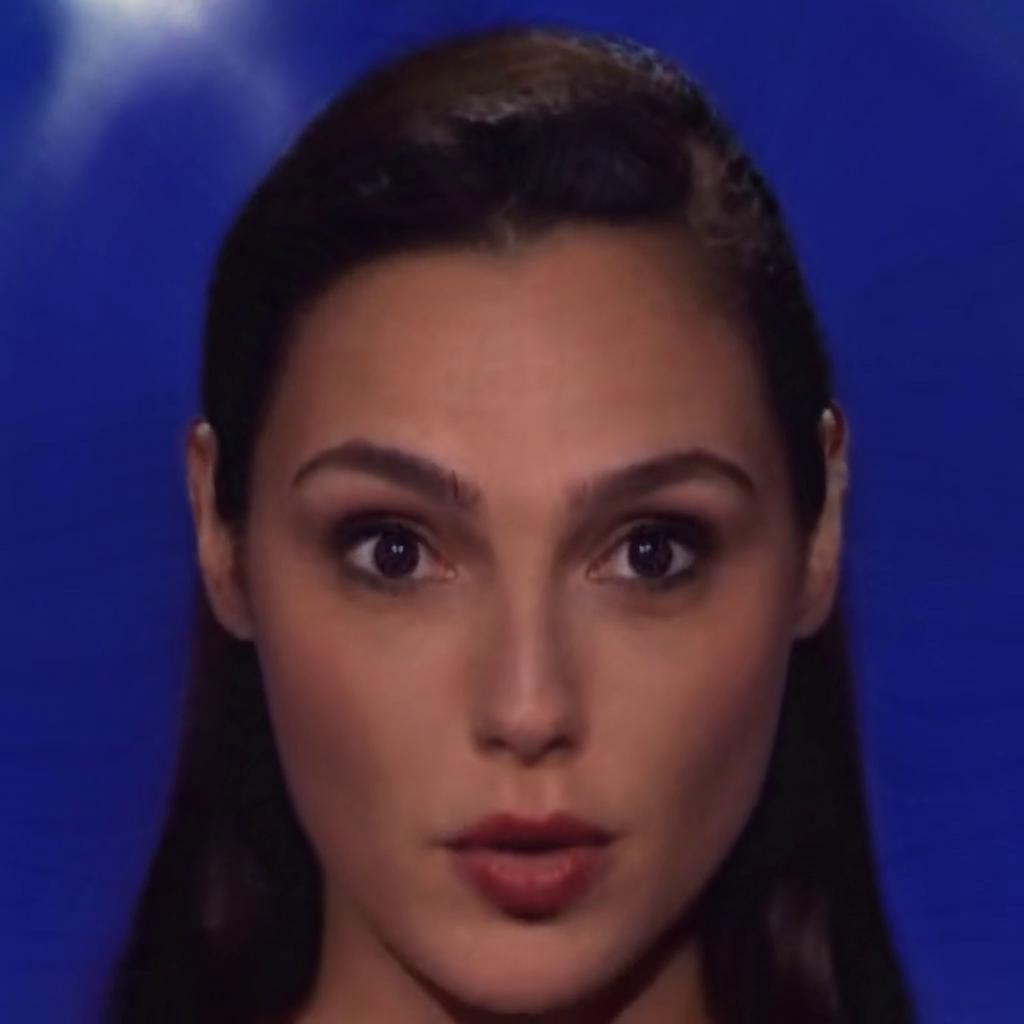} & 
        \includegraphics[width=0.215\columnwidth]{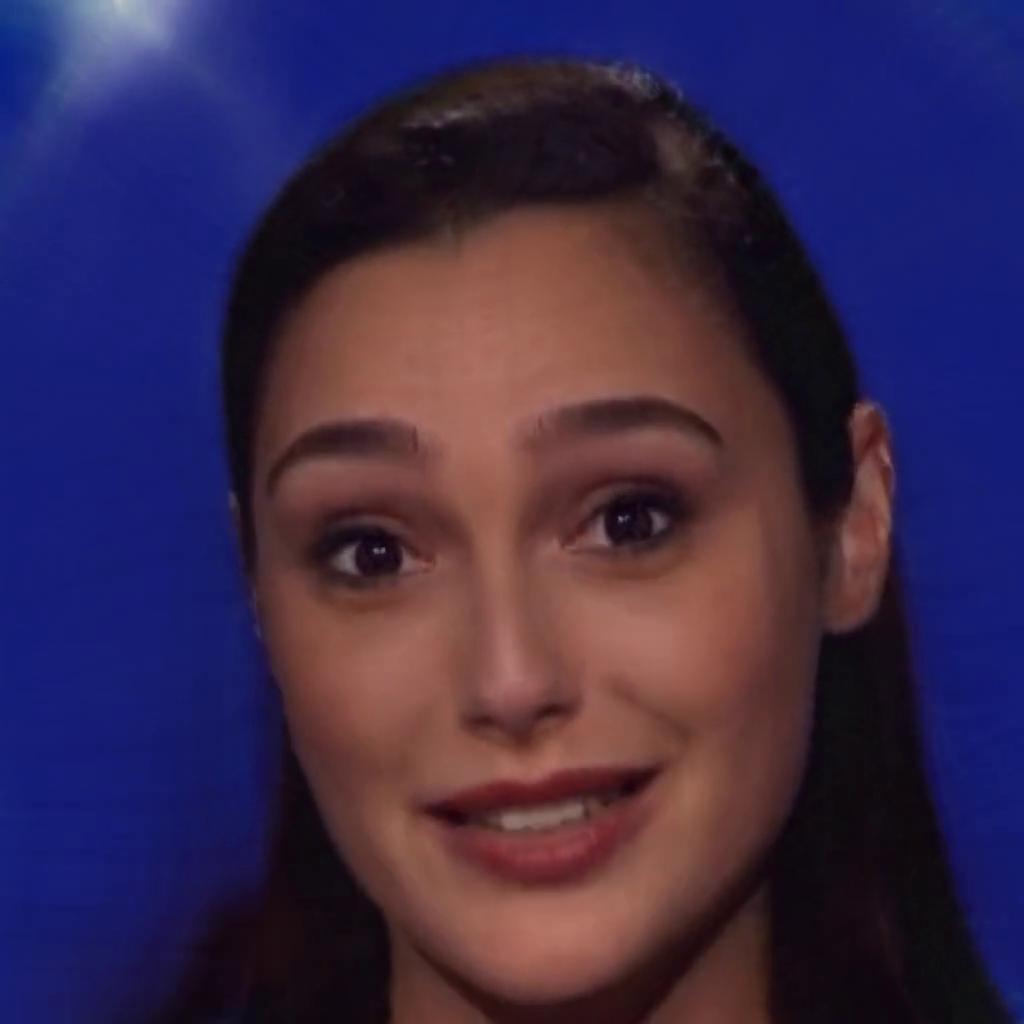} & 
        \includegraphics[width=0.215\columnwidth]{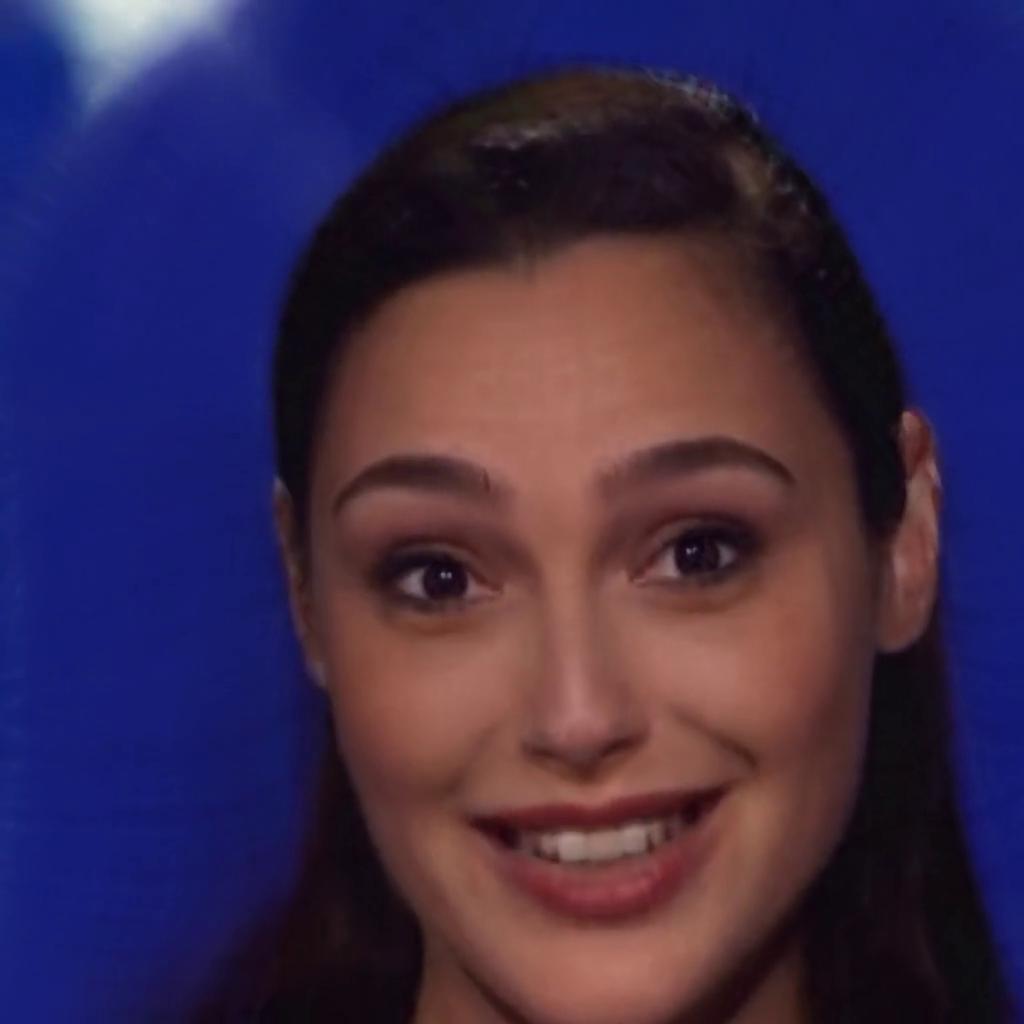} & 
        \includegraphics[width=0.215\columnwidth]{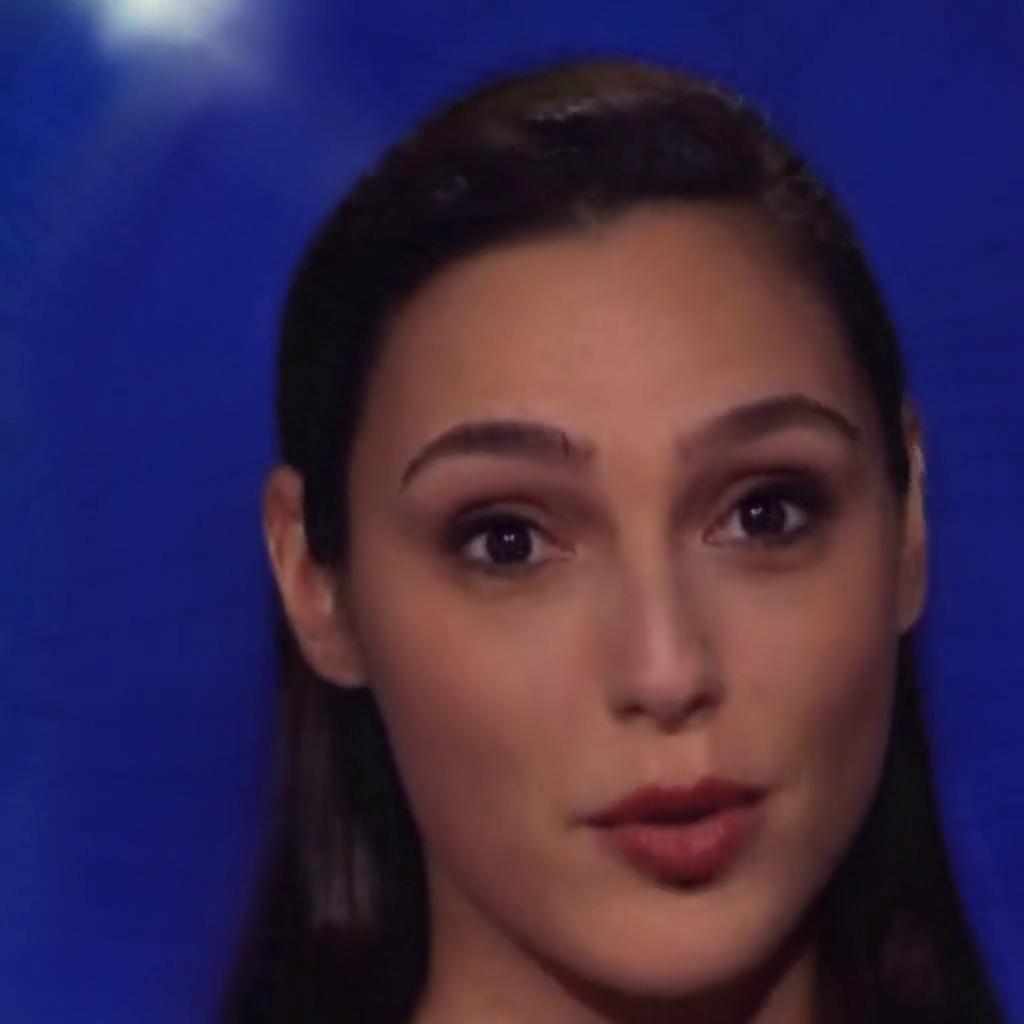} & 
        \includegraphics[width=0.215\columnwidth]{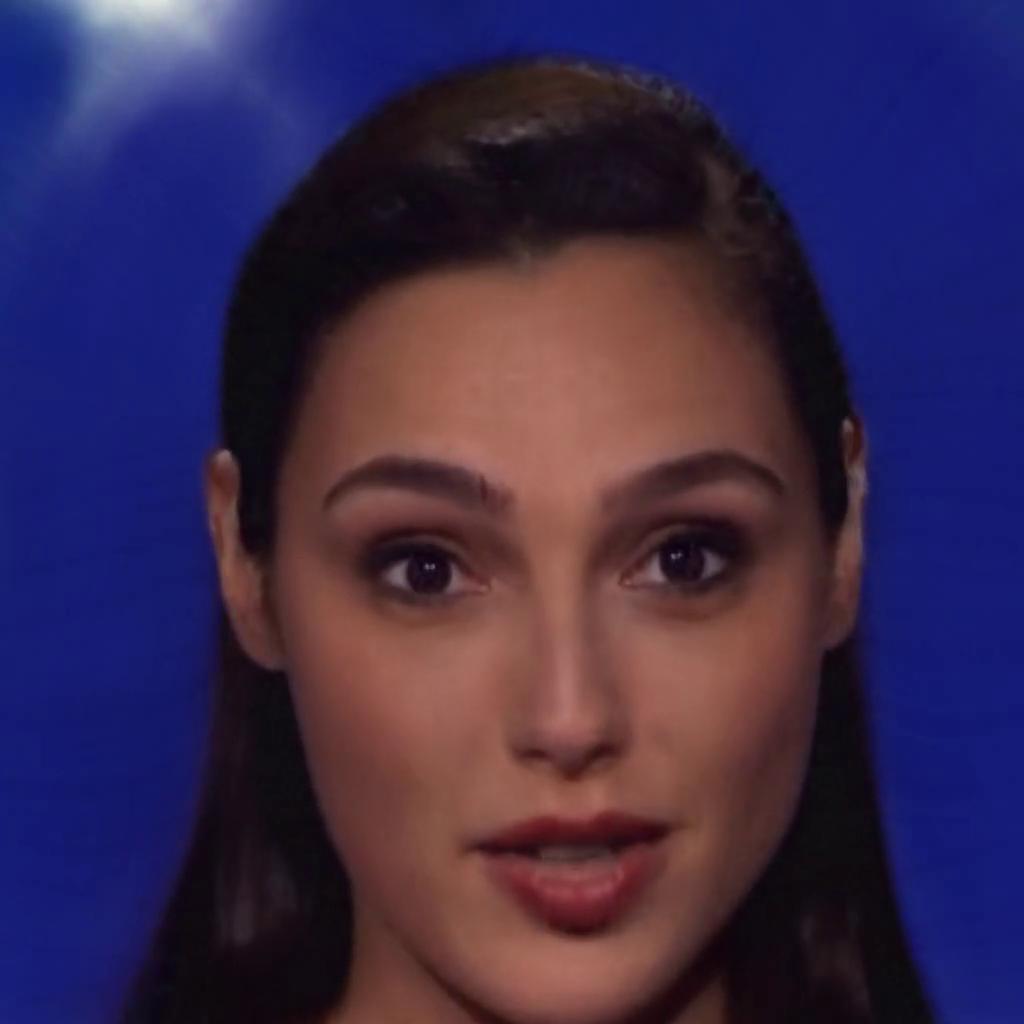} & 
        \includegraphics[width=0.215\columnwidth]{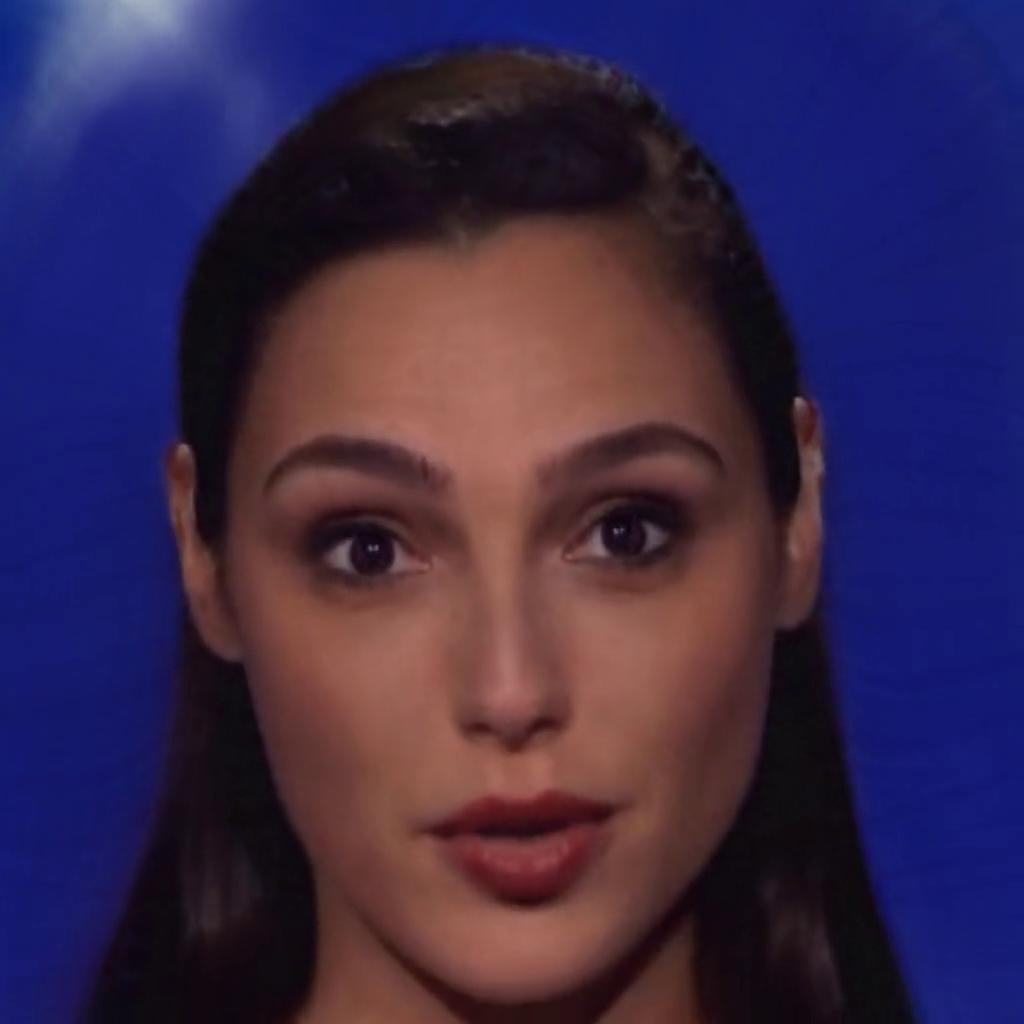} & 
        \includegraphics[width=0.215\columnwidth]{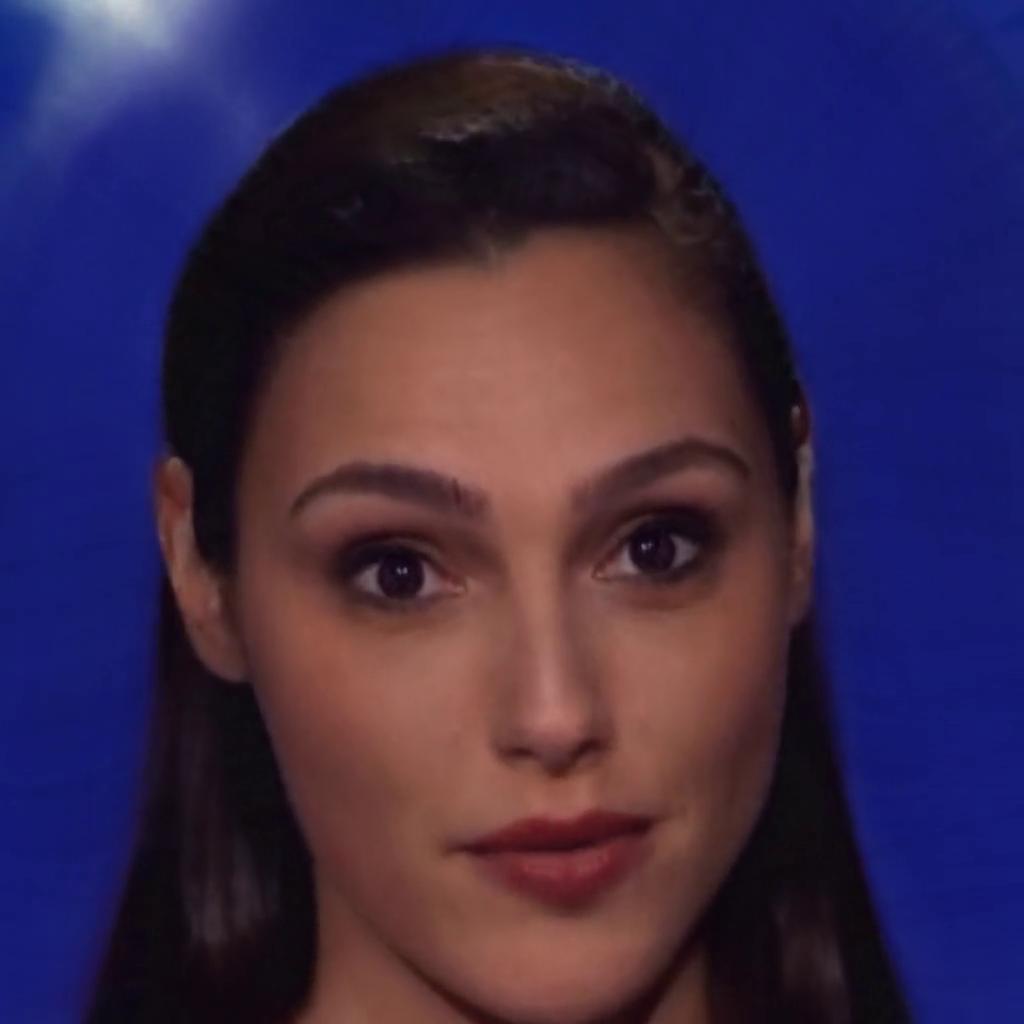} \\

		\raisebox{0.175in}{\rotatebox{90}{Wide View}} &
        \includegraphics[width=0.215\columnwidth]{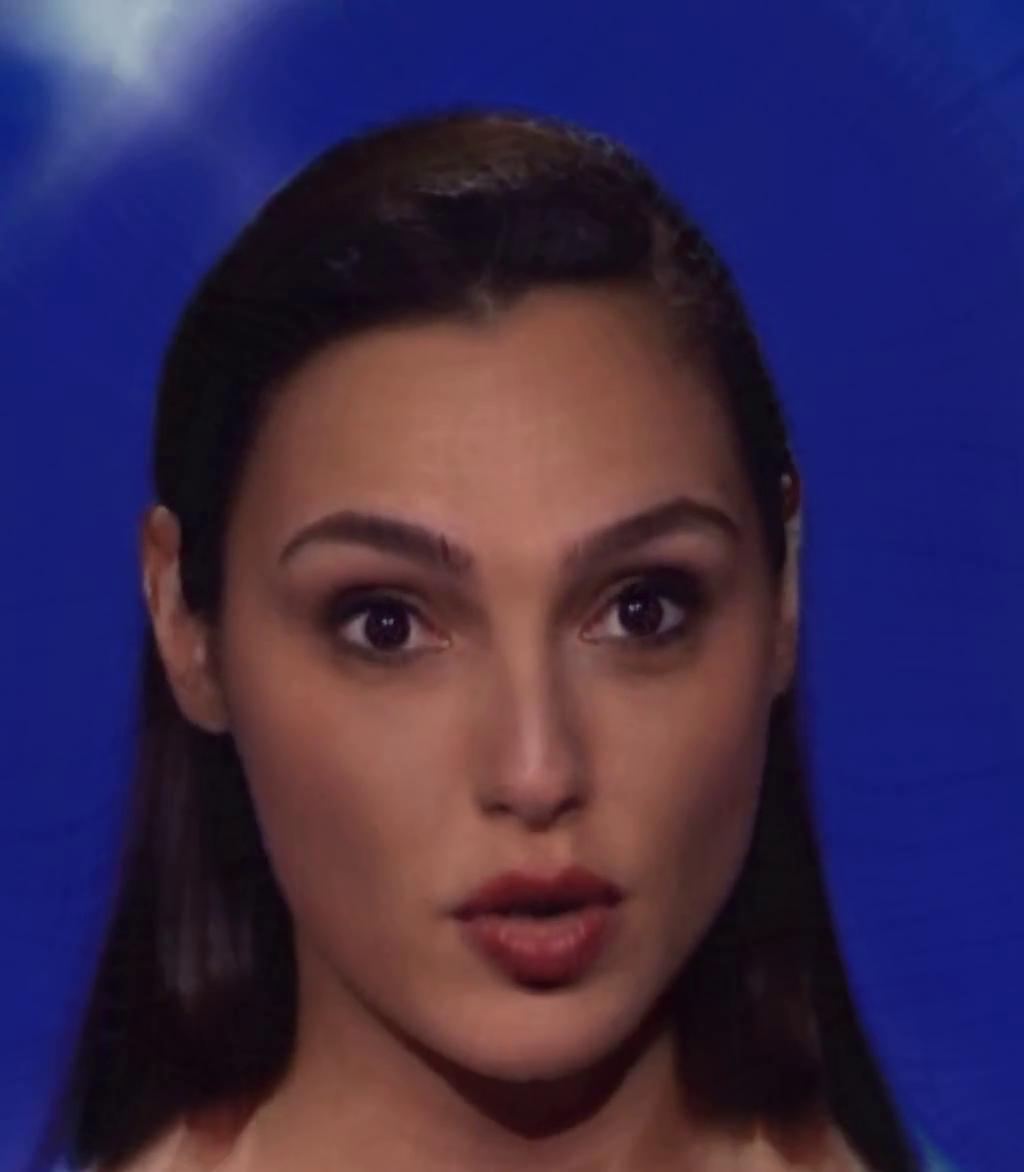} & 
        \includegraphics[width=0.215\columnwidth]{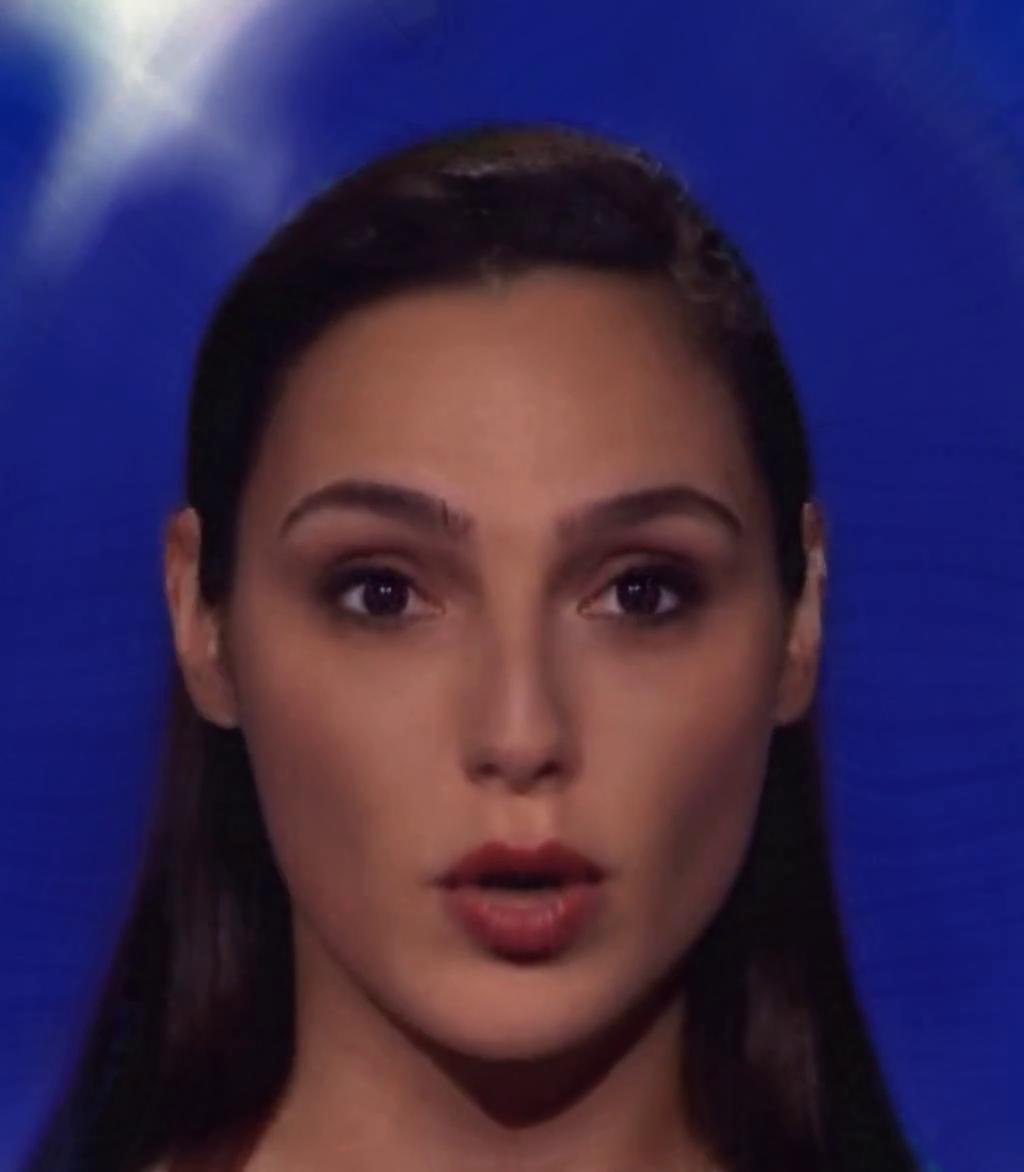} & 
        \includegraphics[width=0.215\columnwidth]{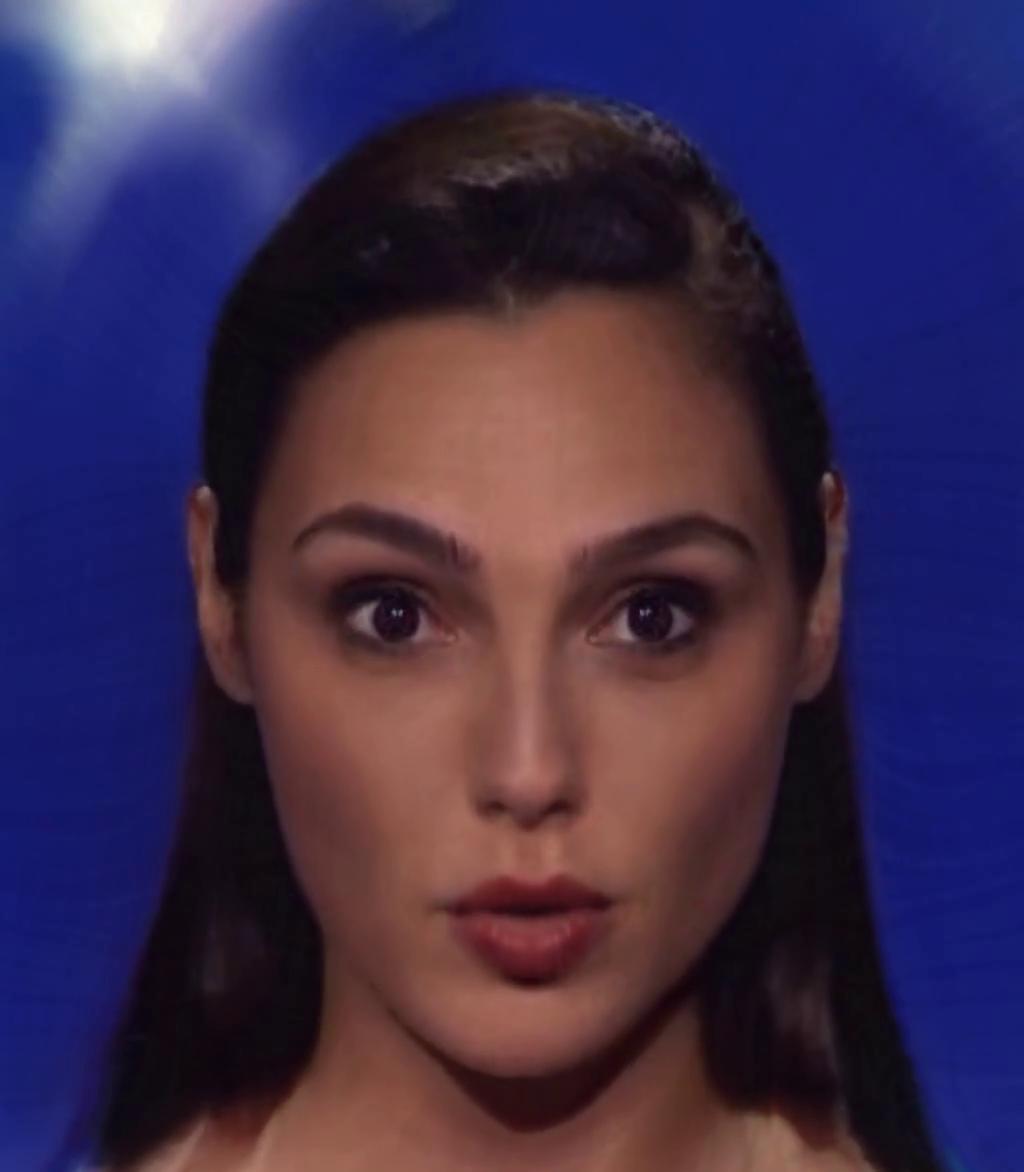} & 
        \includegraphics[width=0.215\columnwidth]{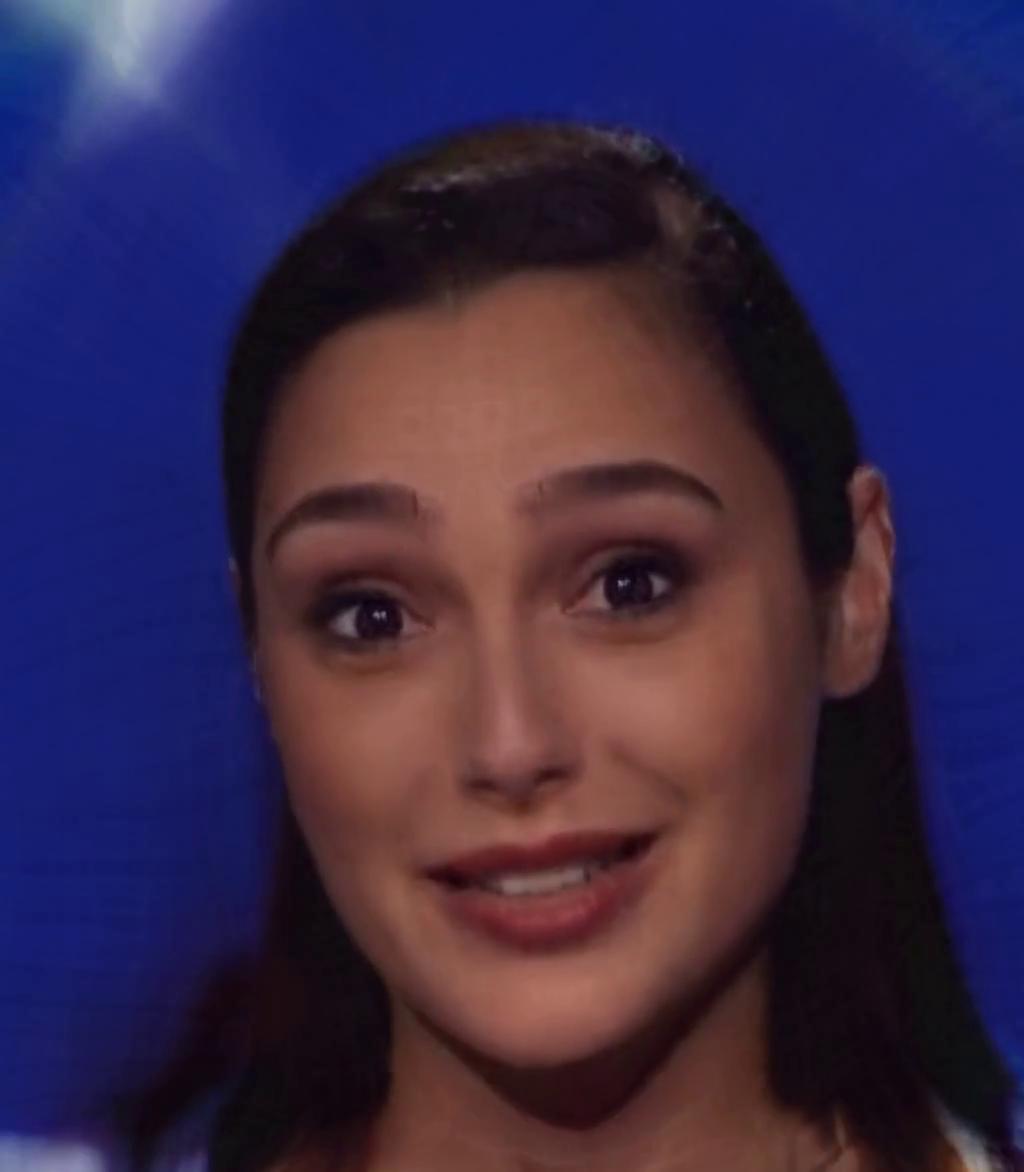} & 
        \includegraphics[width=0.215\columnwidth]{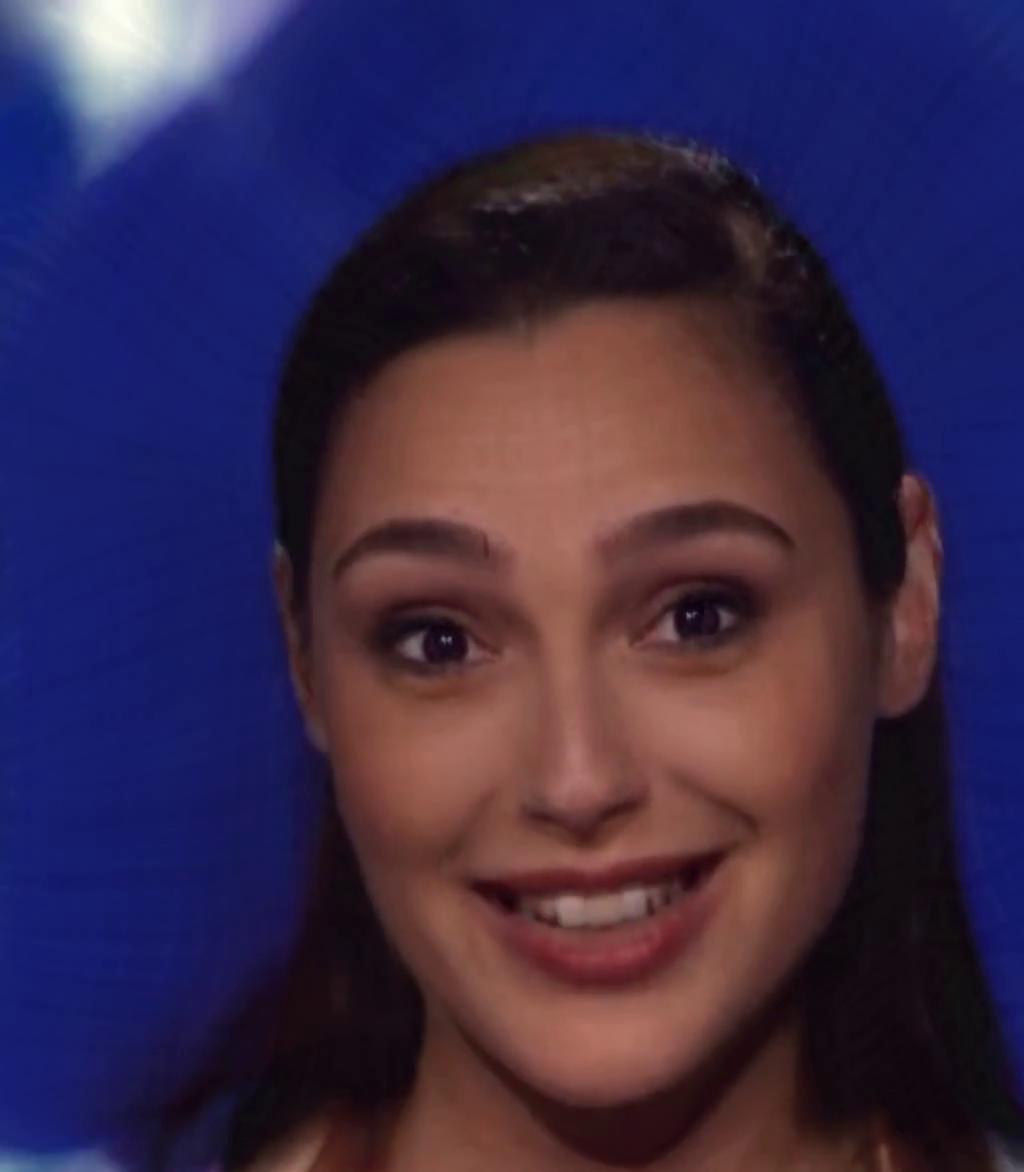} & 
        \includegraphics[width=0.215\columnwidth]{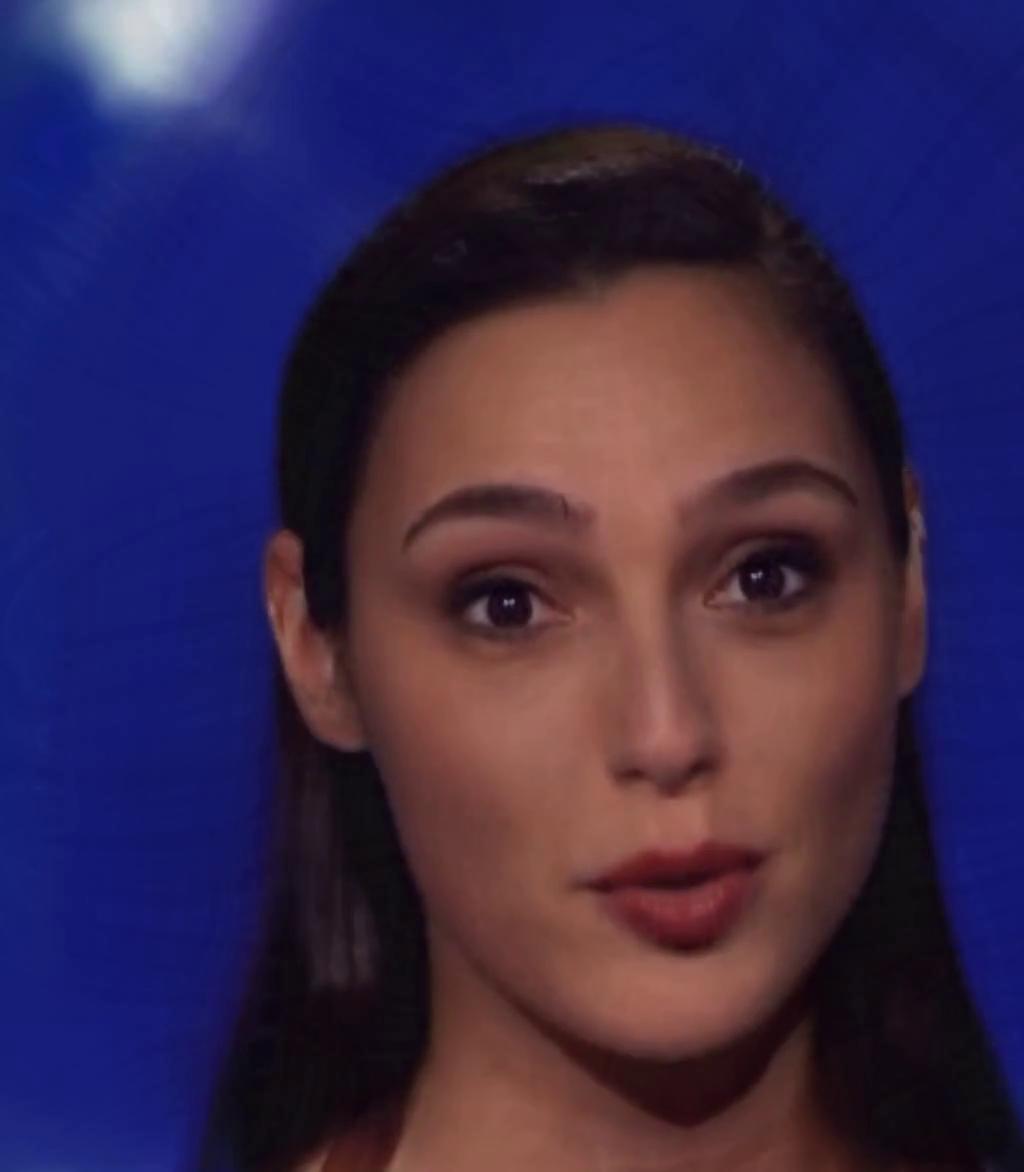} & 
        \includegraphics[width=0.215\columnwidth]{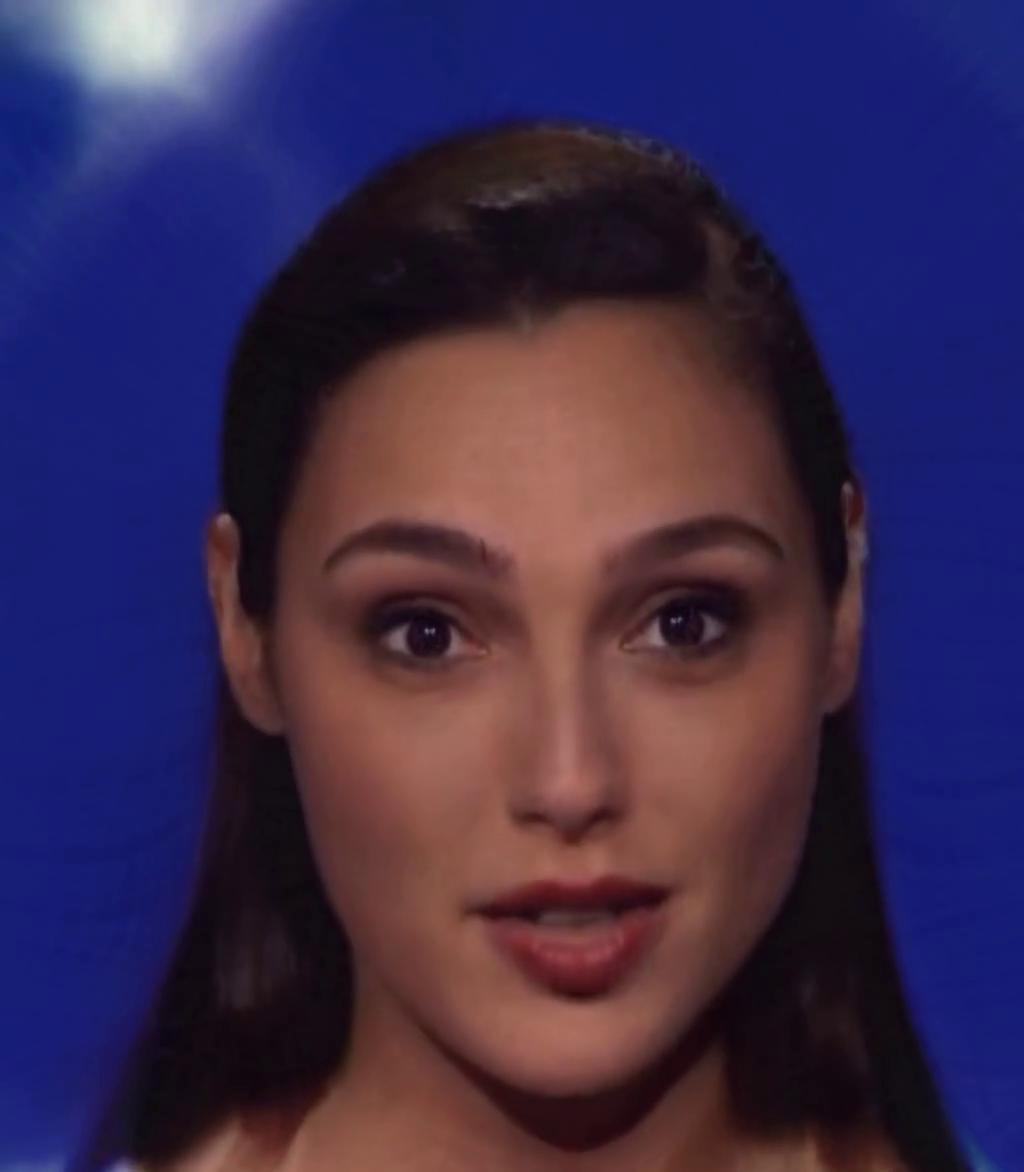} & 
        \includegraphics[width=0.215\columnwidth]{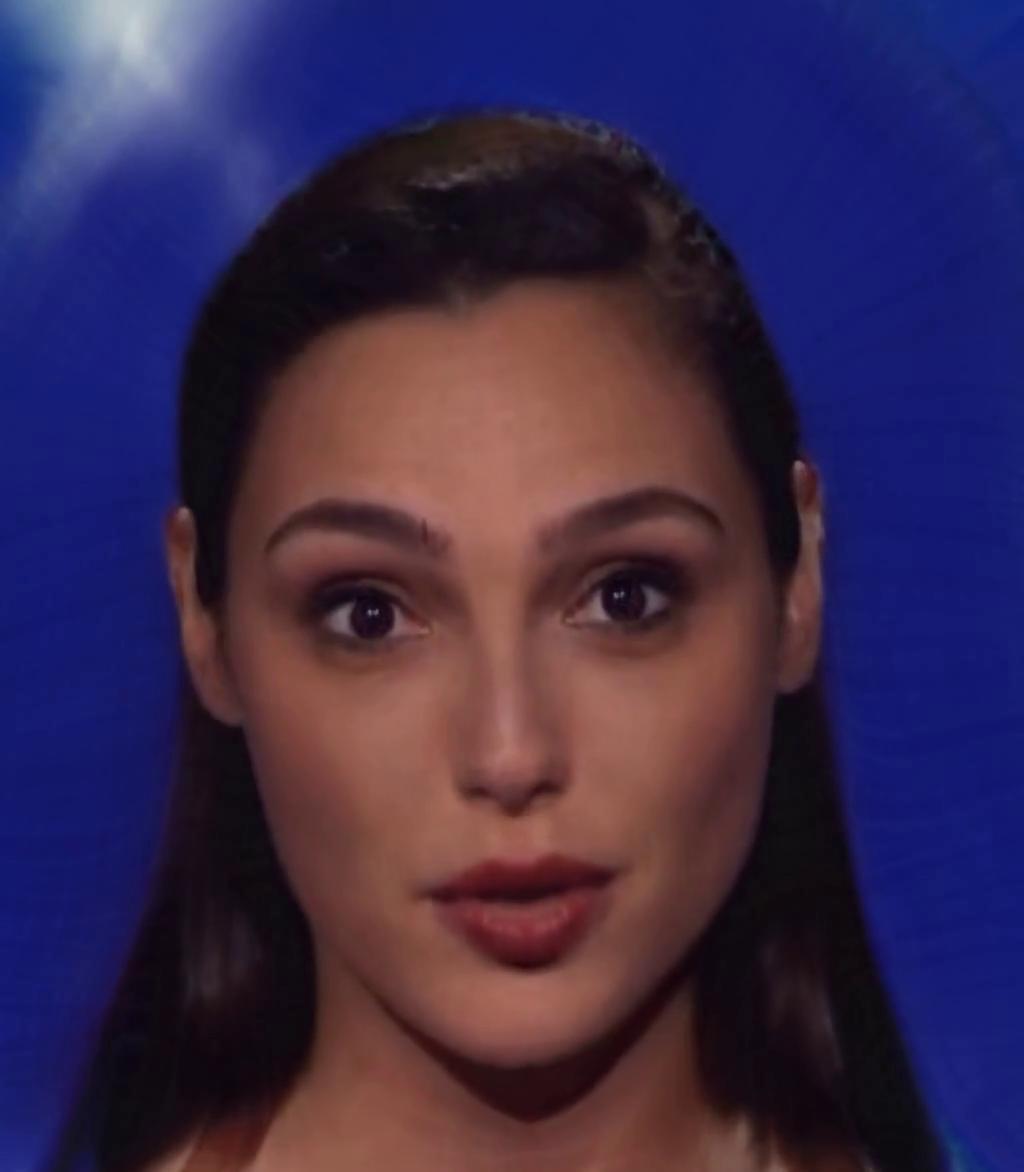} & 
        \includegraphics[width=0.215\columnwidth]{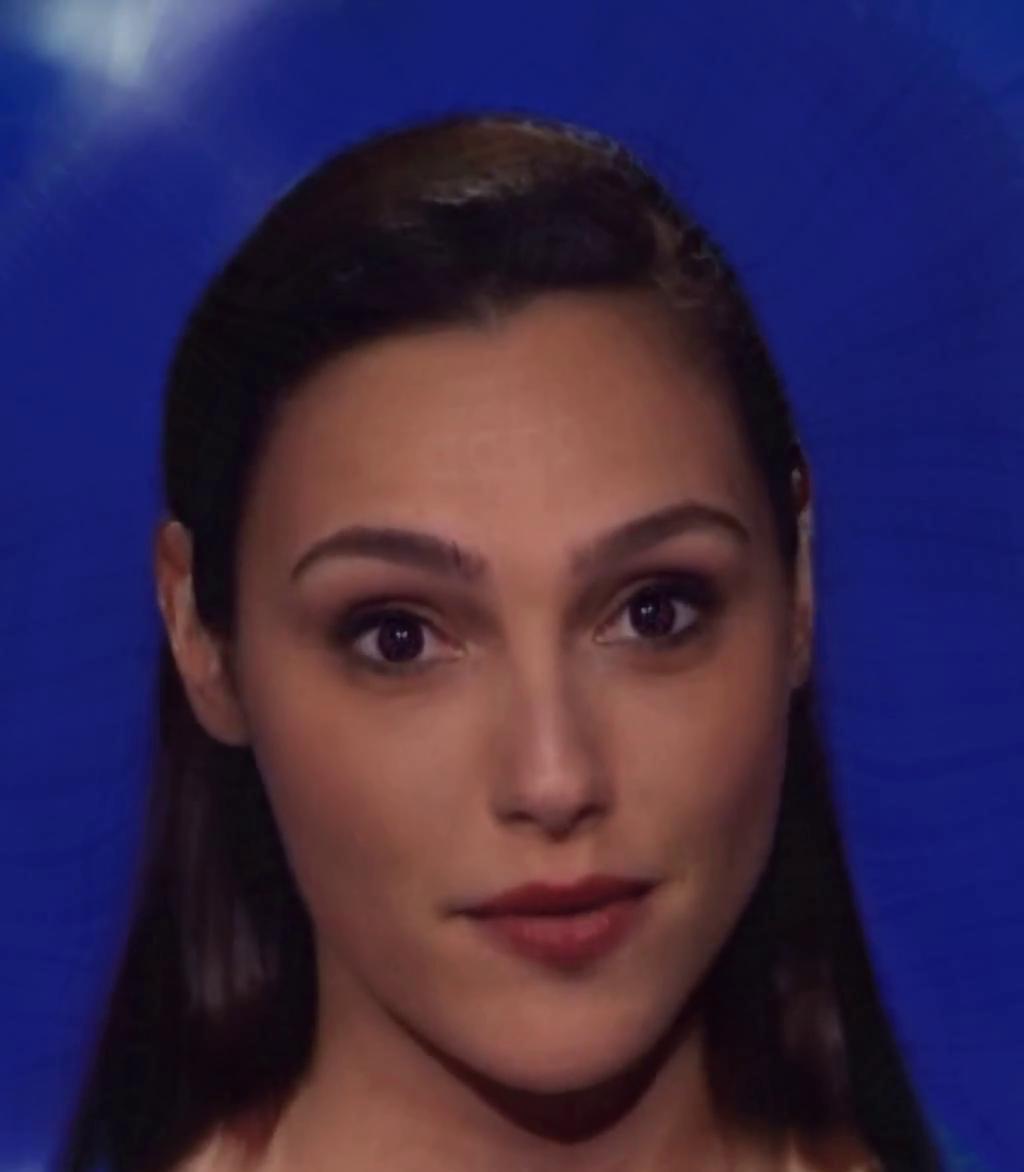} \\

	\end{tabular}
	}
	
	{\footnotesize
	\begin{tabular}{c c c c c c c c c c}

        \\ \\

		\raisebox{0.15in}{\rotatebox{90}{Original}} &
        \includegraphics[width=0.215\columnwidth]{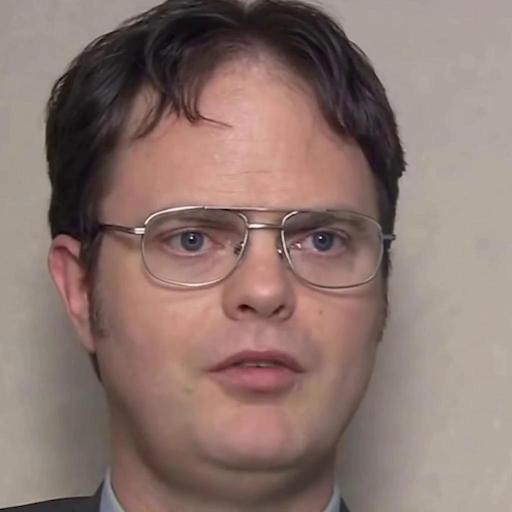} & 
        \includegraphics[width=0.215\columnwidth]{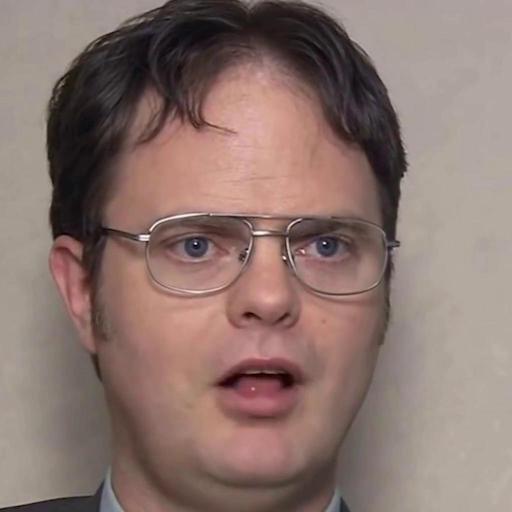} & 
        \includegraphics[width=0.215\columnwidth]{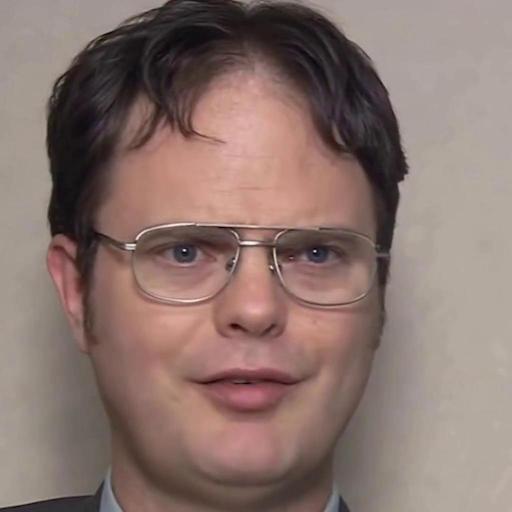} & 
        \includegraphics[width=0.215\columnwidth]{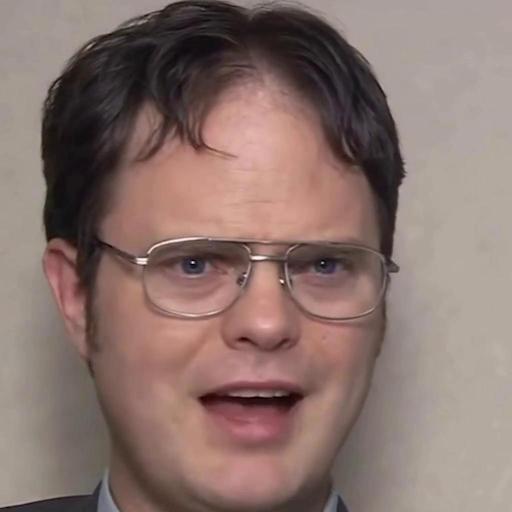} & 
        \includegraphics[width=0.215\columnwidth]{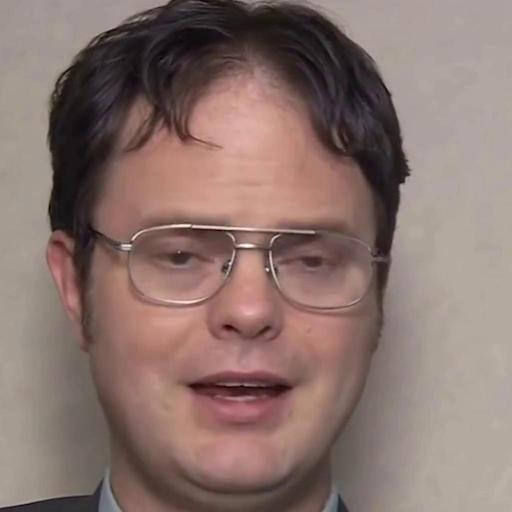} & 
        \includegraphics[width=0.215\columnwidth]{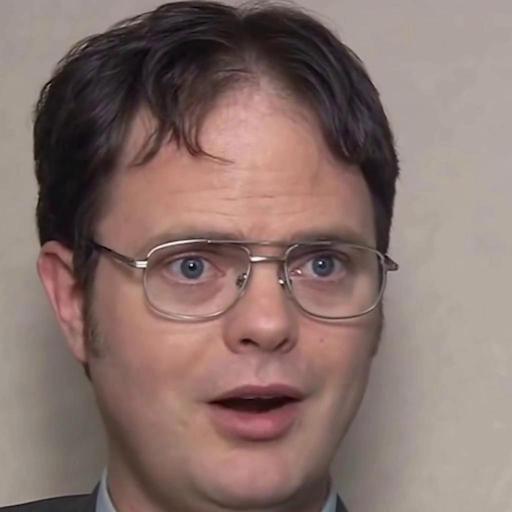} & 
        \includegraphics[width=0.215\columnwidth]{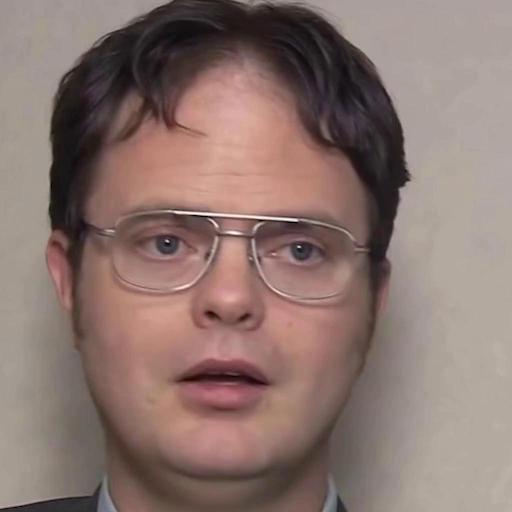} & 
        \includegraphics[width=0.215\columnwidth]{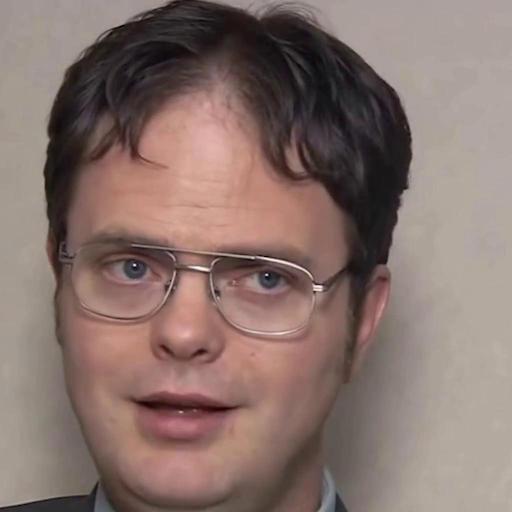} & 
        \includegraphics[width=0.215\columnwidth]{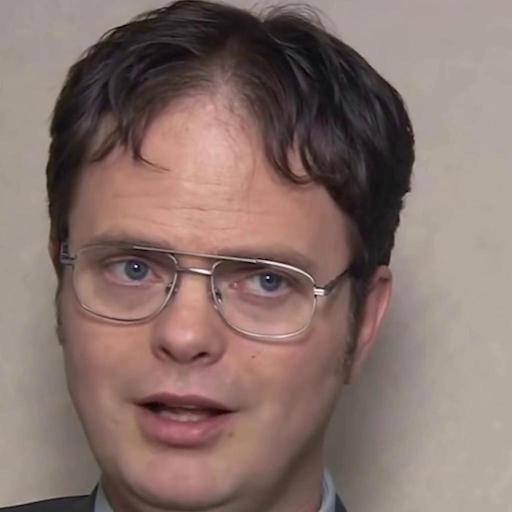} \\

		\raisebox{0.05in}{\rotatebox{90}{Reconstruction}} &
        \includegraphics[width=0.215\columnwidth]{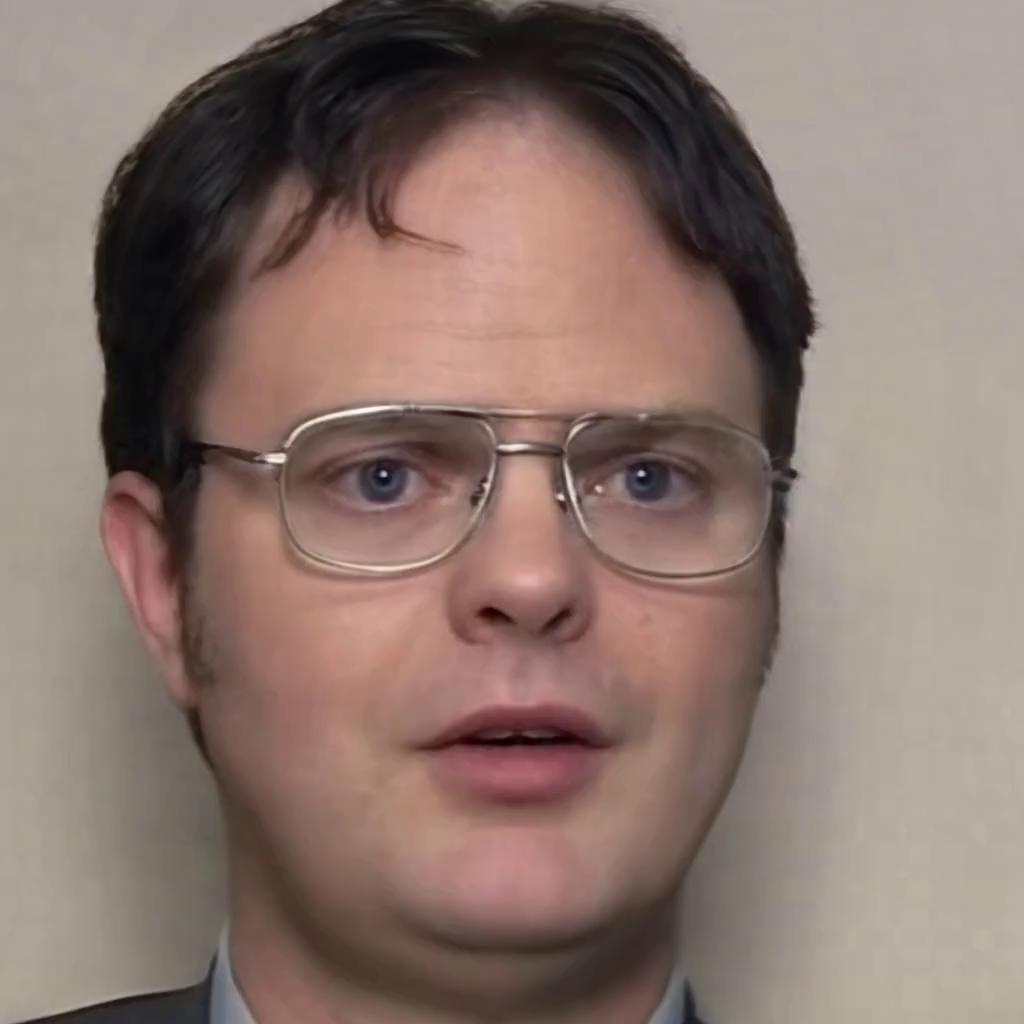} & 
        \includegraphics[width=0.215\columnwidth]{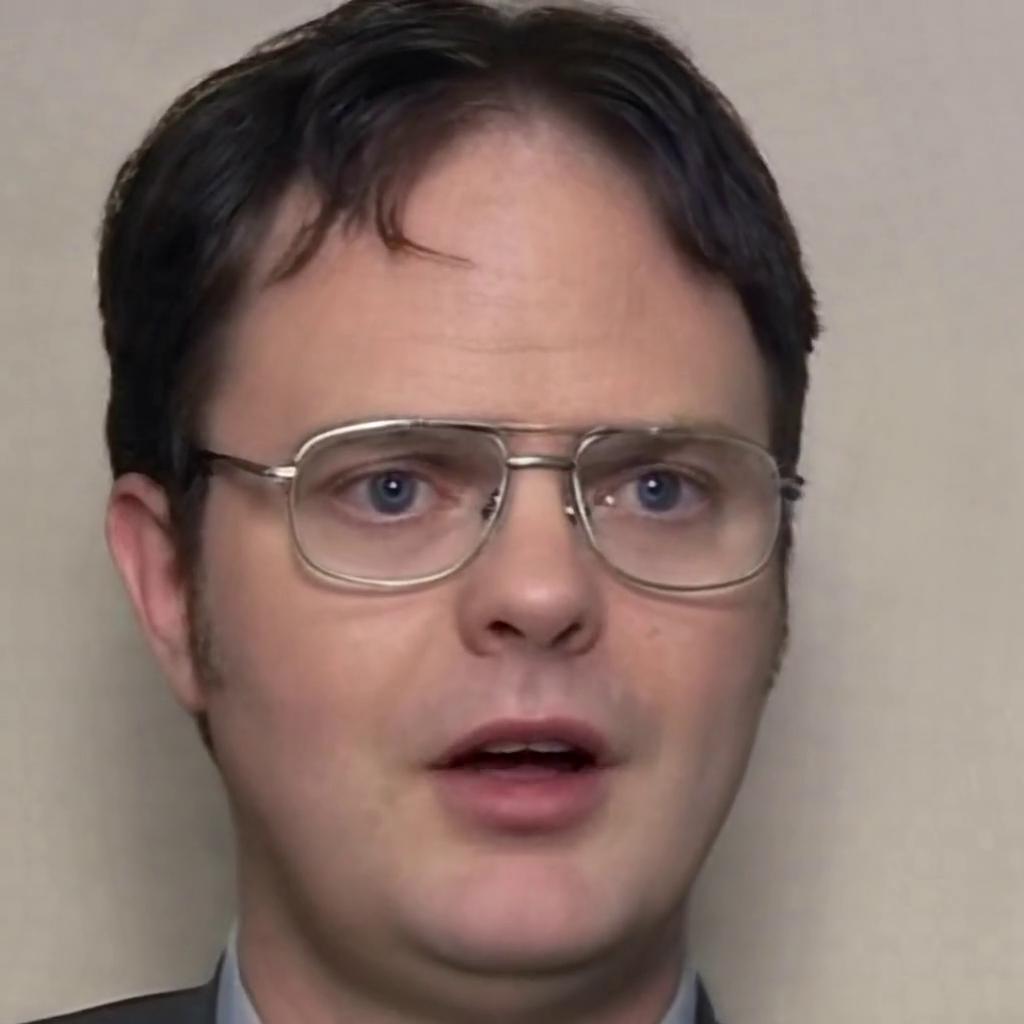} & 
        \includegraphics[width=0.215\columnwidth]{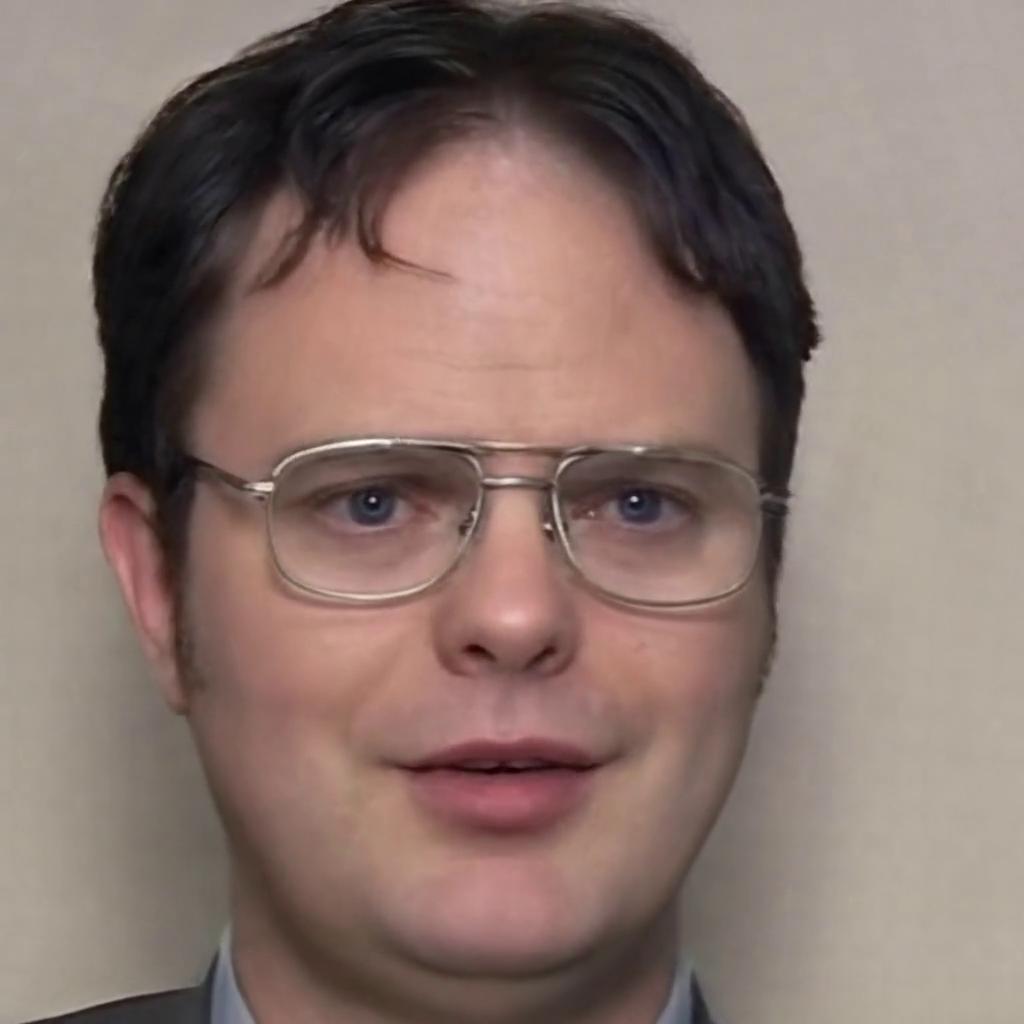} & 
        \includegraphics[width=0.215\columnwidth]{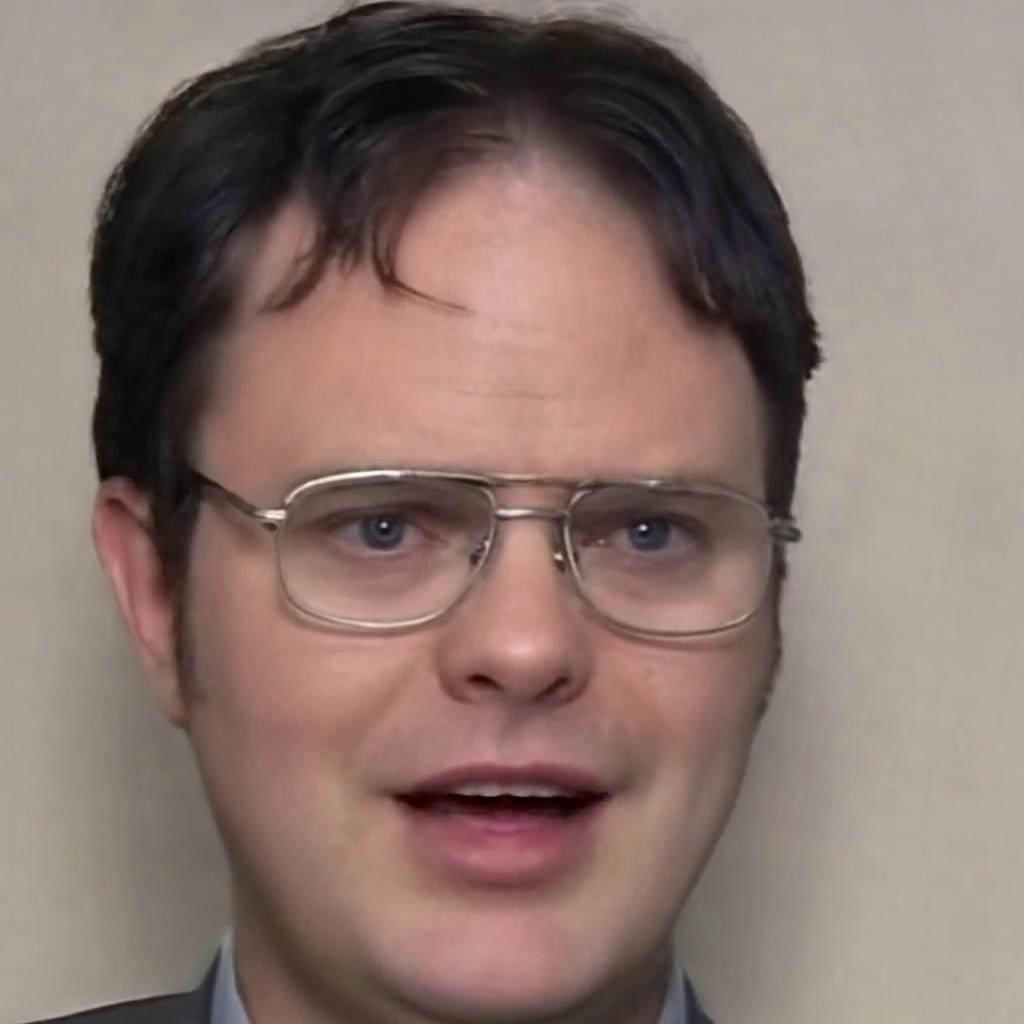} & 
        \includegraphics[width=0.215\columnwidth]{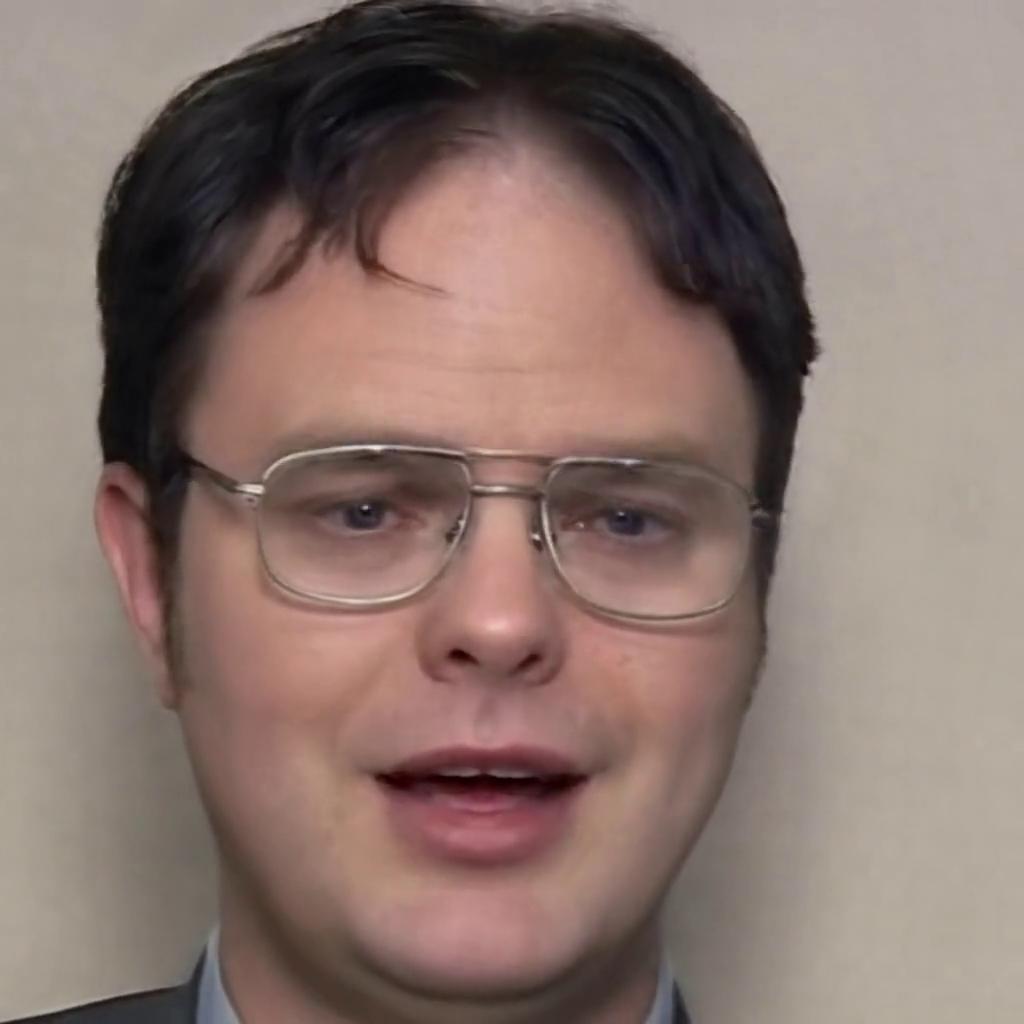} & 
        \includegraphics[width=0.215\columnwidth]{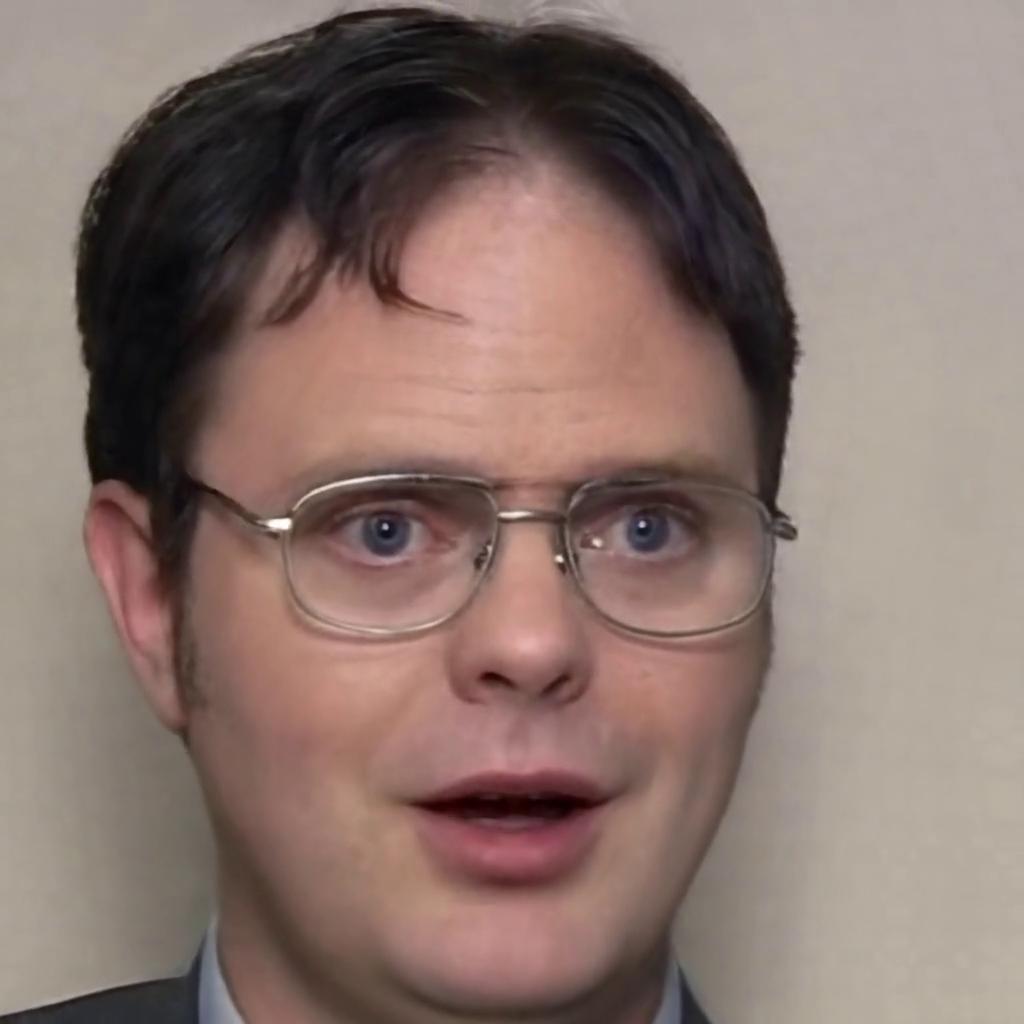} & 
        \includegraphics[width=0.215\columnwidth]{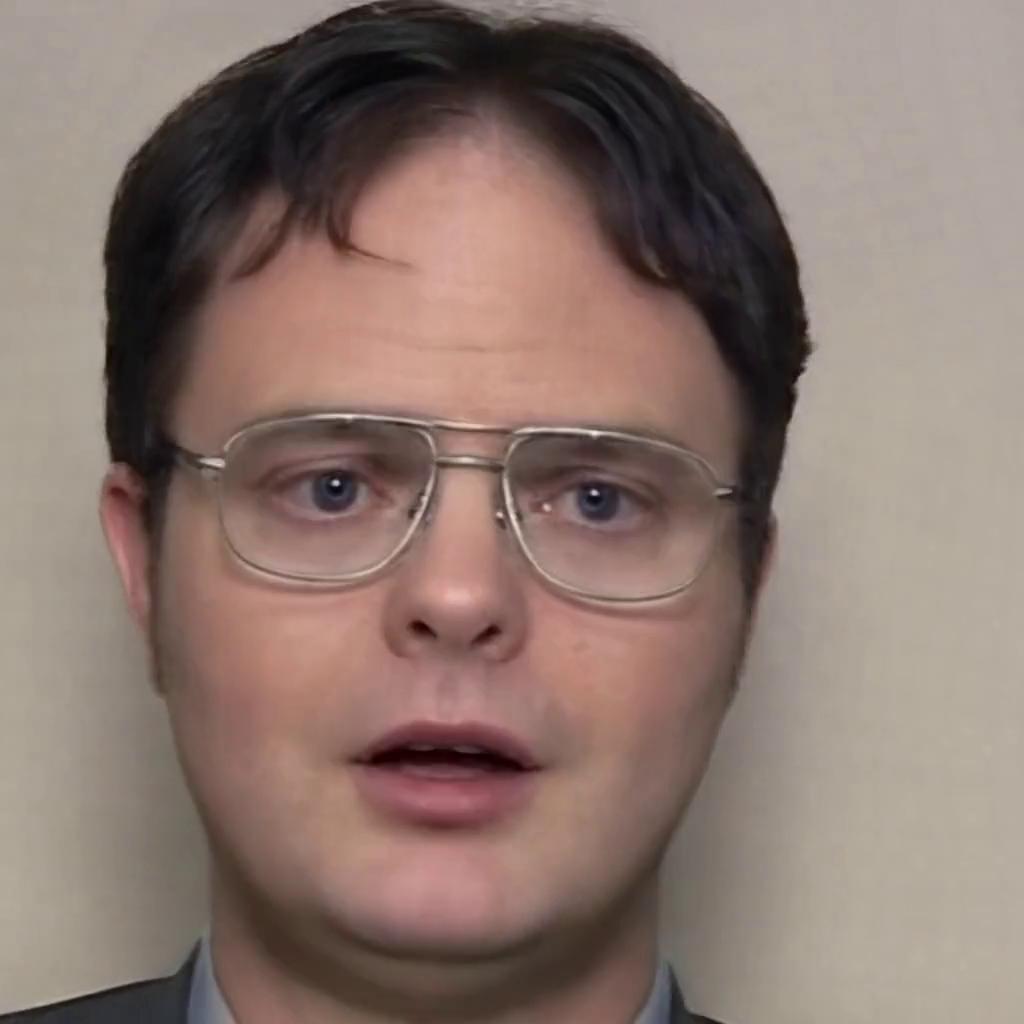} & 
        \includegraphics[width=0.215\columnwidth]{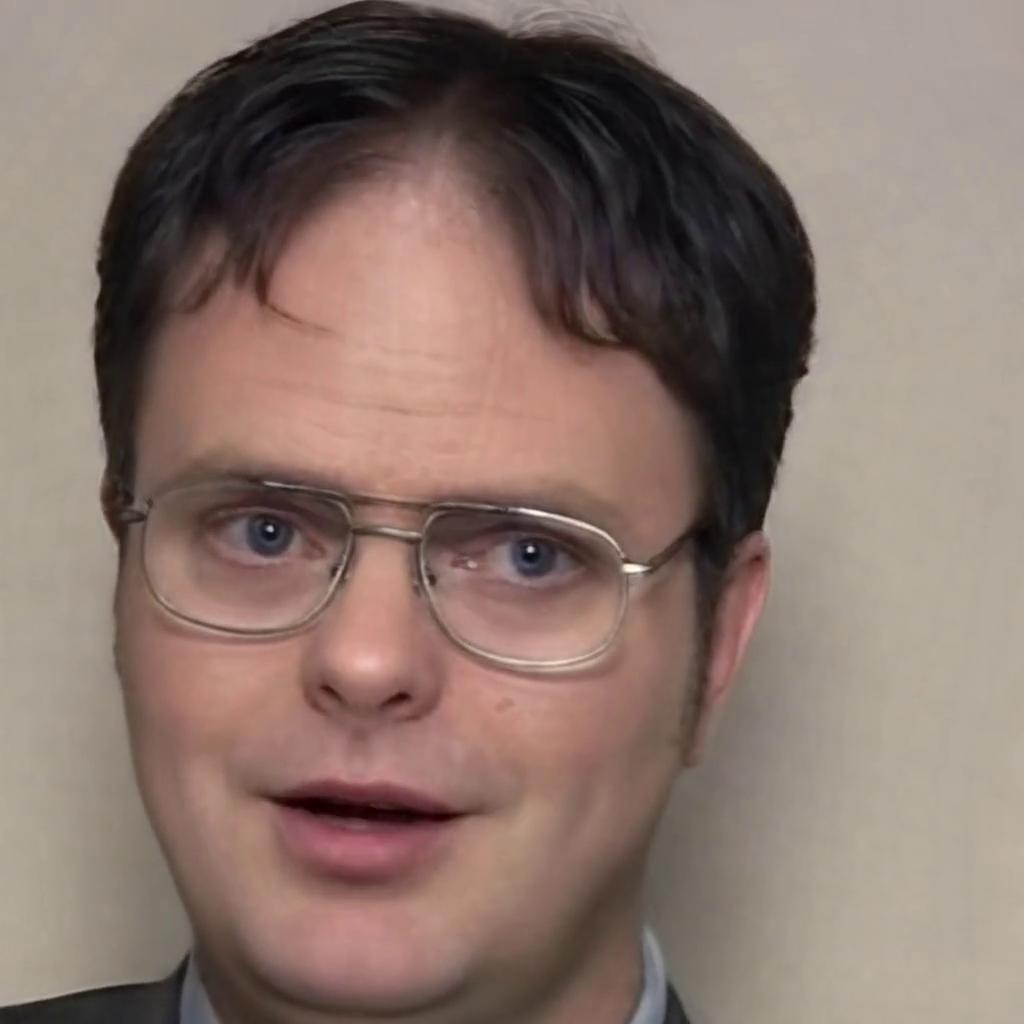} & 
        \includegraphics[width=0.215\columnwidth]{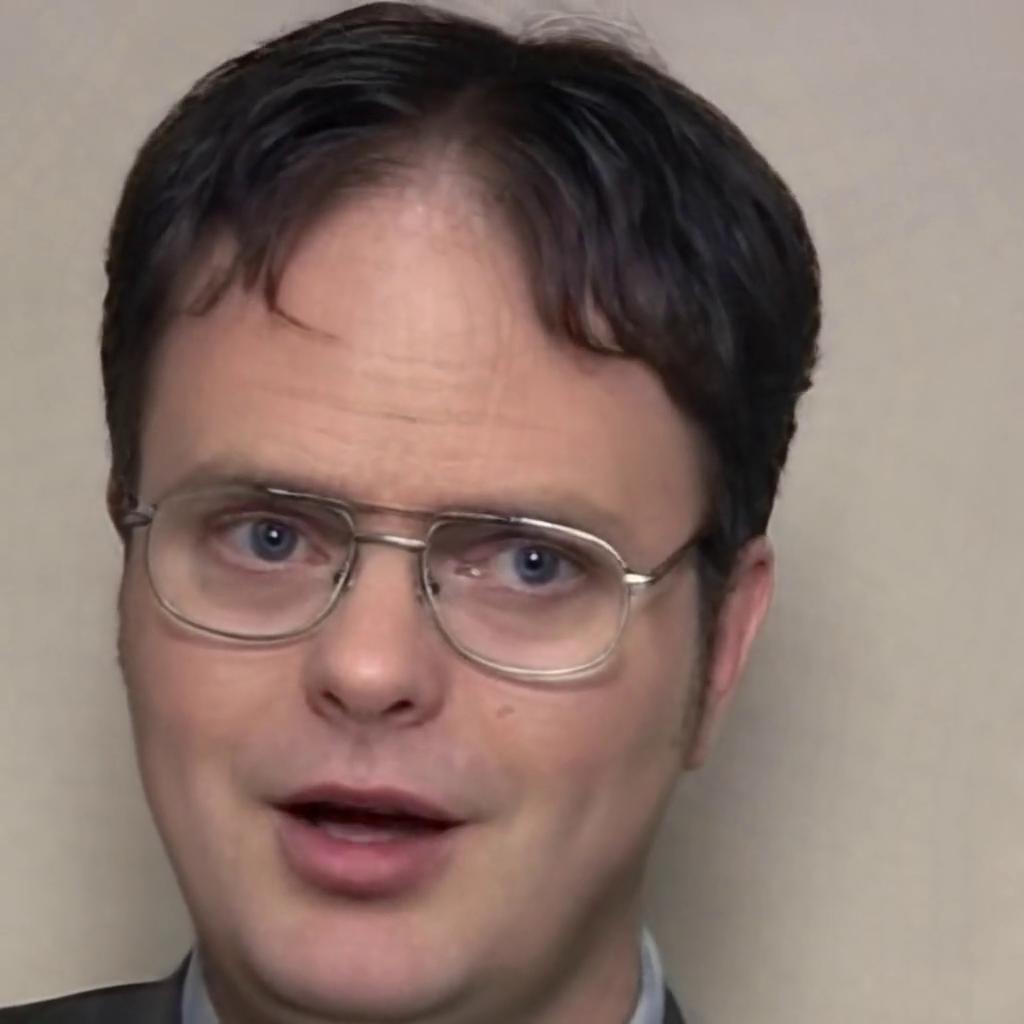} \\

		\raisebox{0.175in}{\rotatebox{90}{Wide View}} &
        \includegraphics[width=0.215\columnwidth]{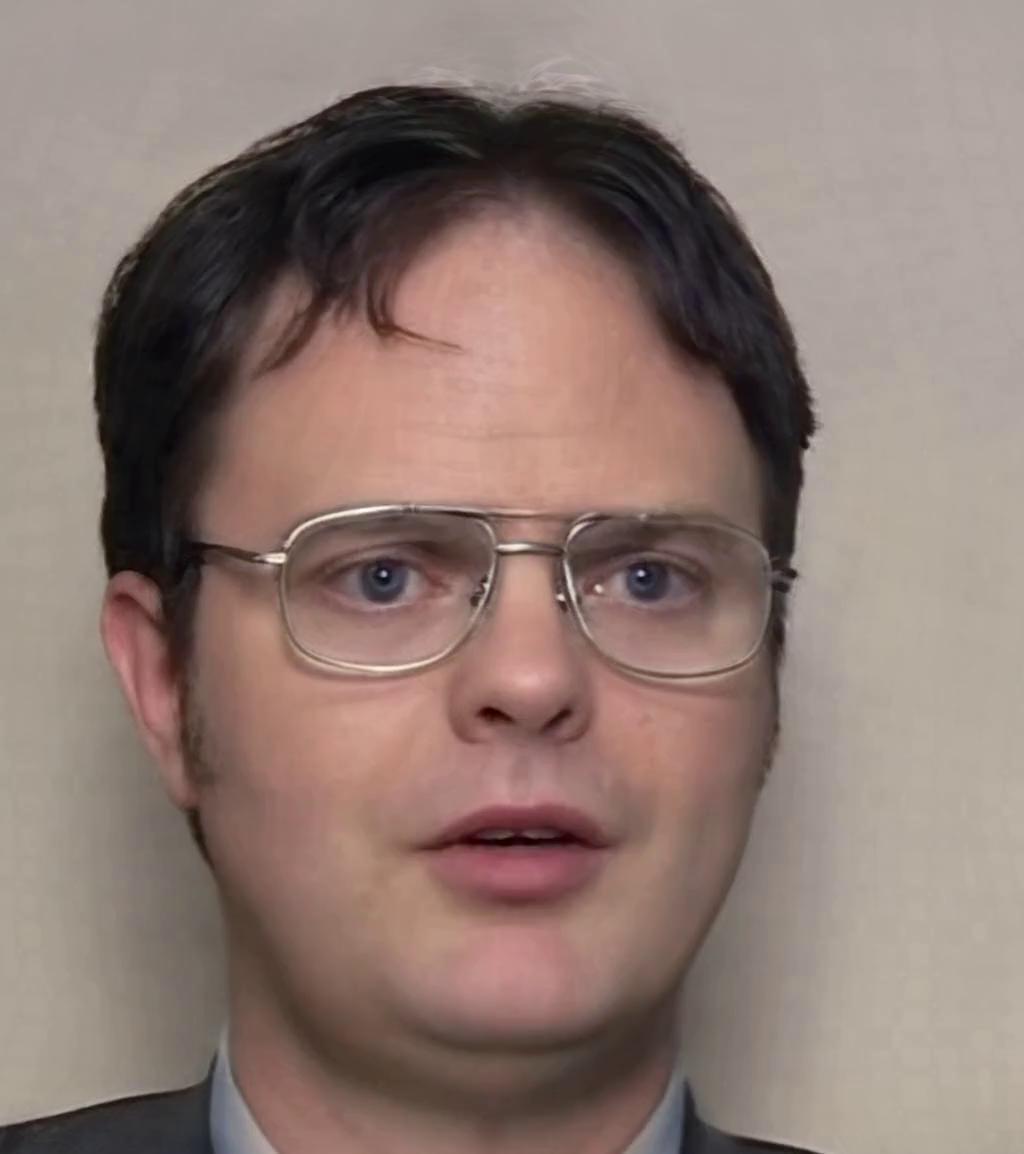} & 
        \includegraphics[width=0.215\columnwidth]{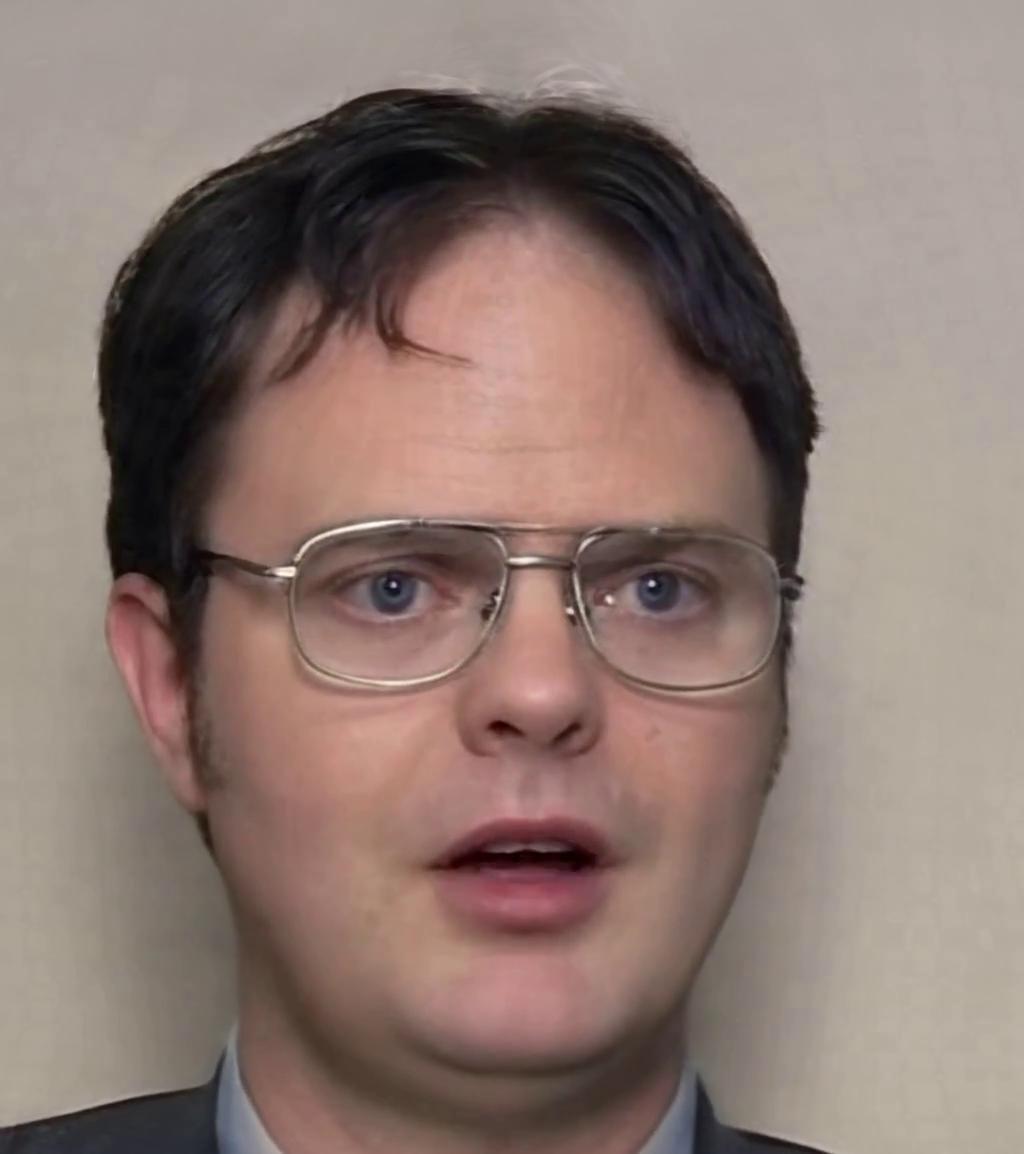} & 
        \includegraphics[width=0.215\columnwidth]{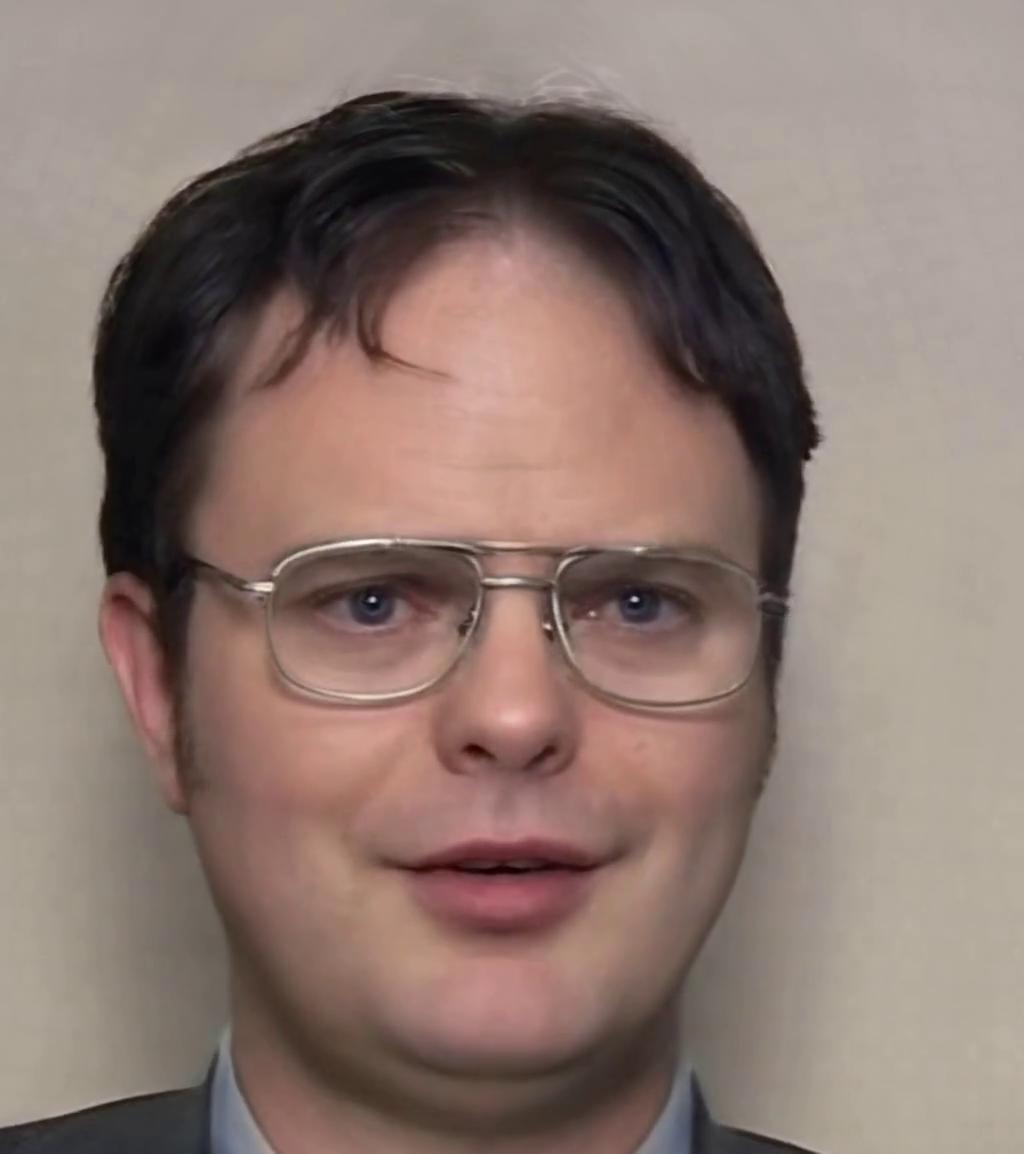} & 
        \includegraphics[width=0.215\columnwidth]{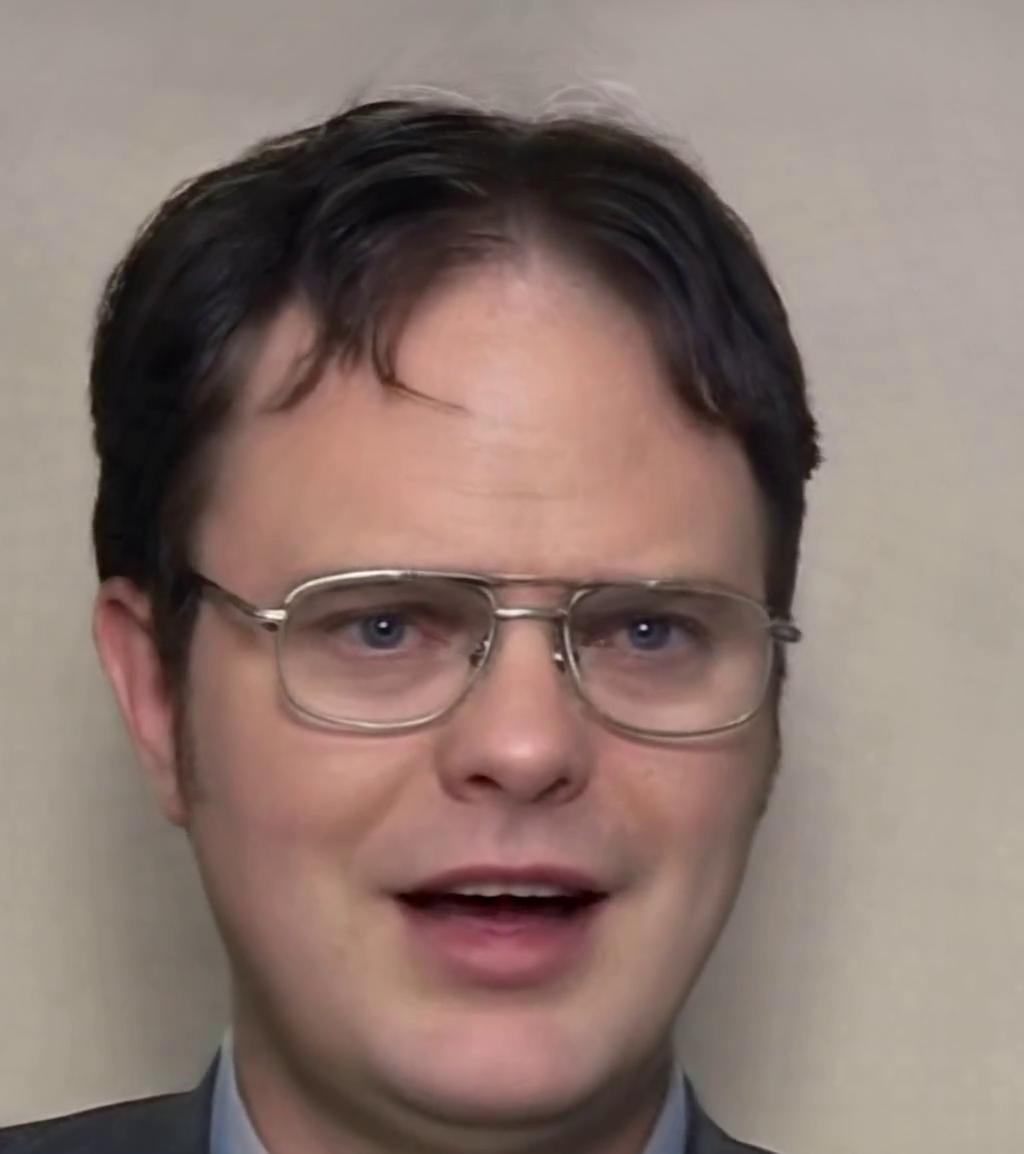} & 
        \includegraphics[width=0.215\columnwidth]{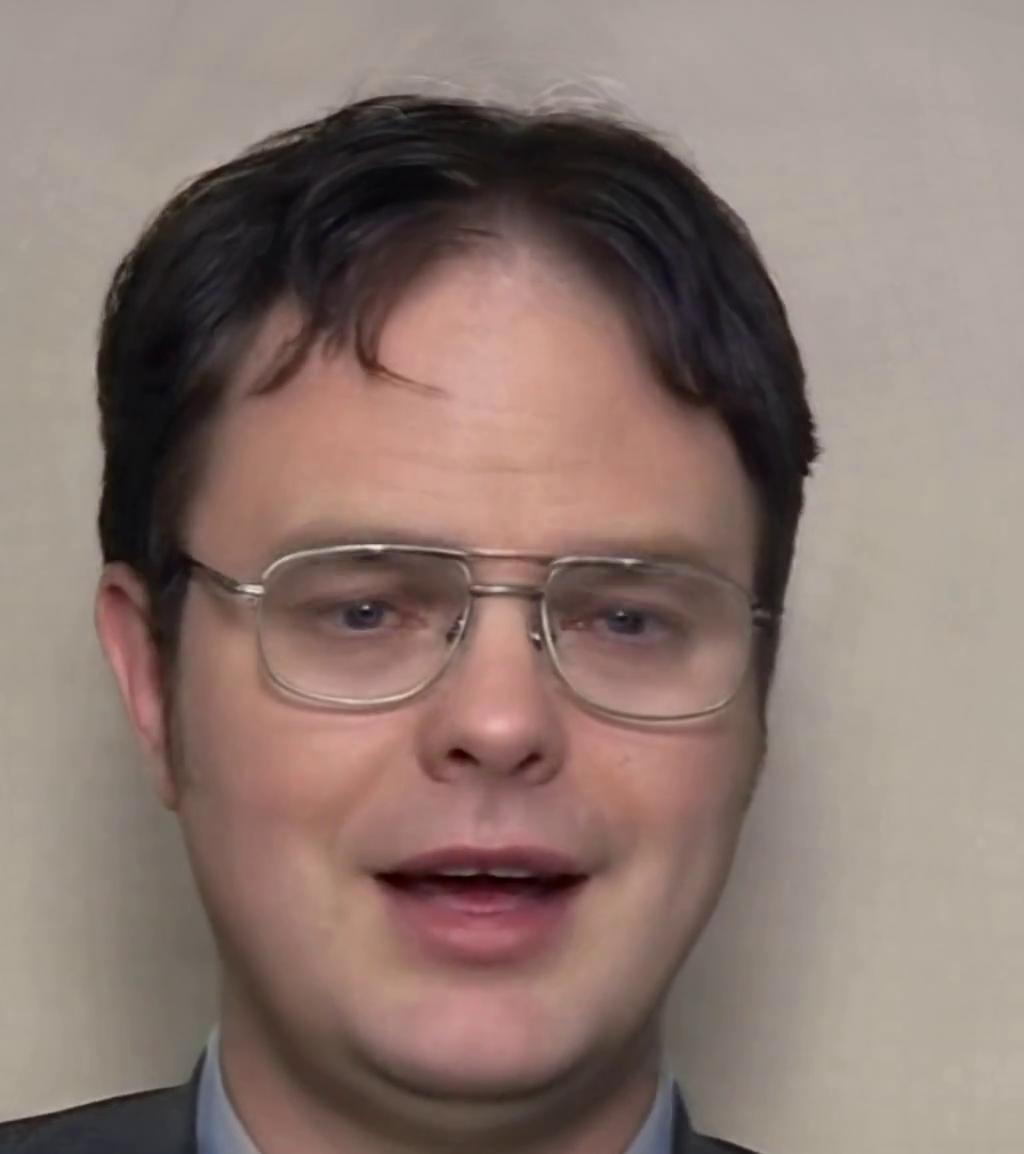} & 
        \includegraphics[width=0.215\columnwidth]{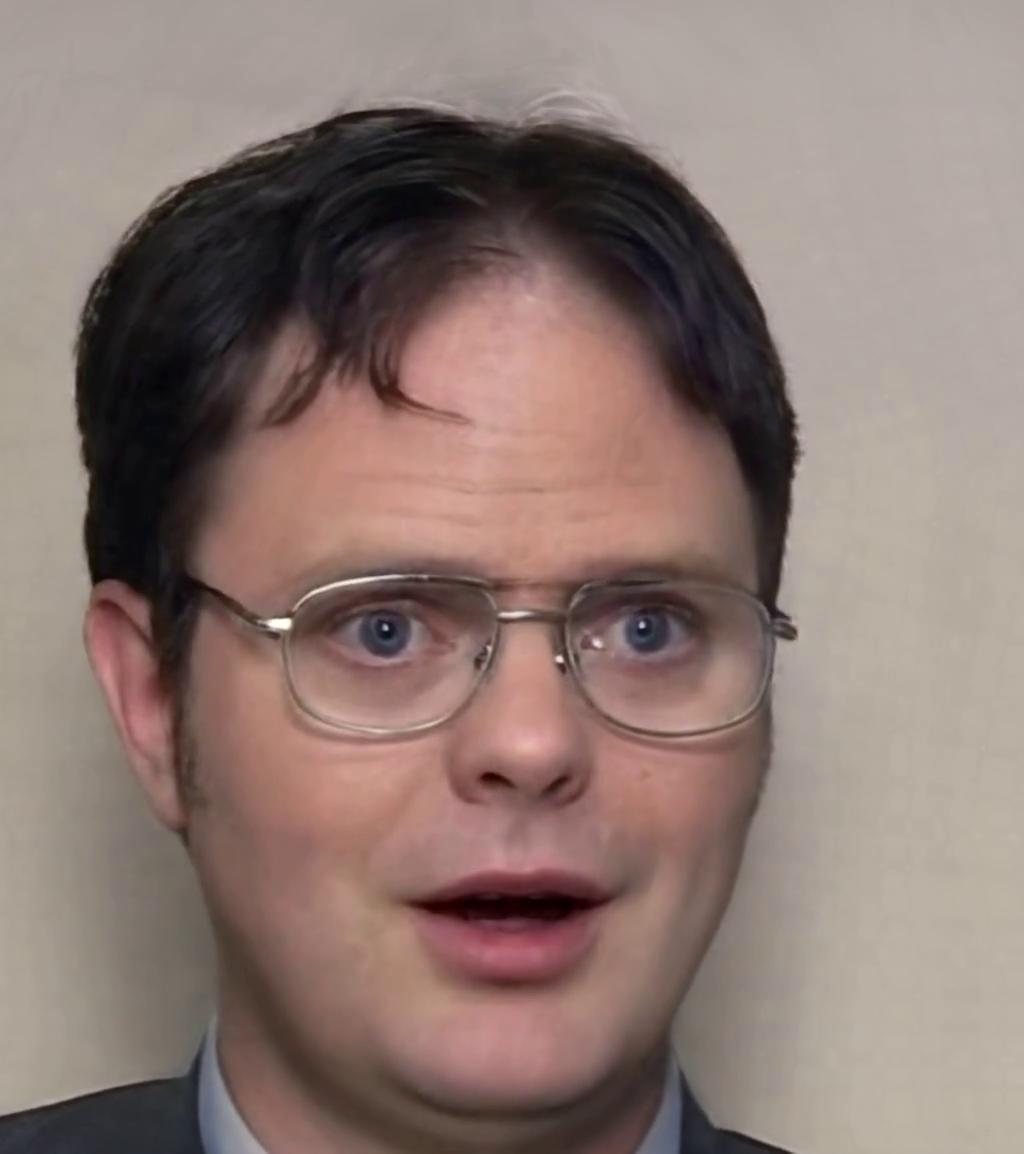} & 
        \includegraphics[width=0.215\columnwidth]{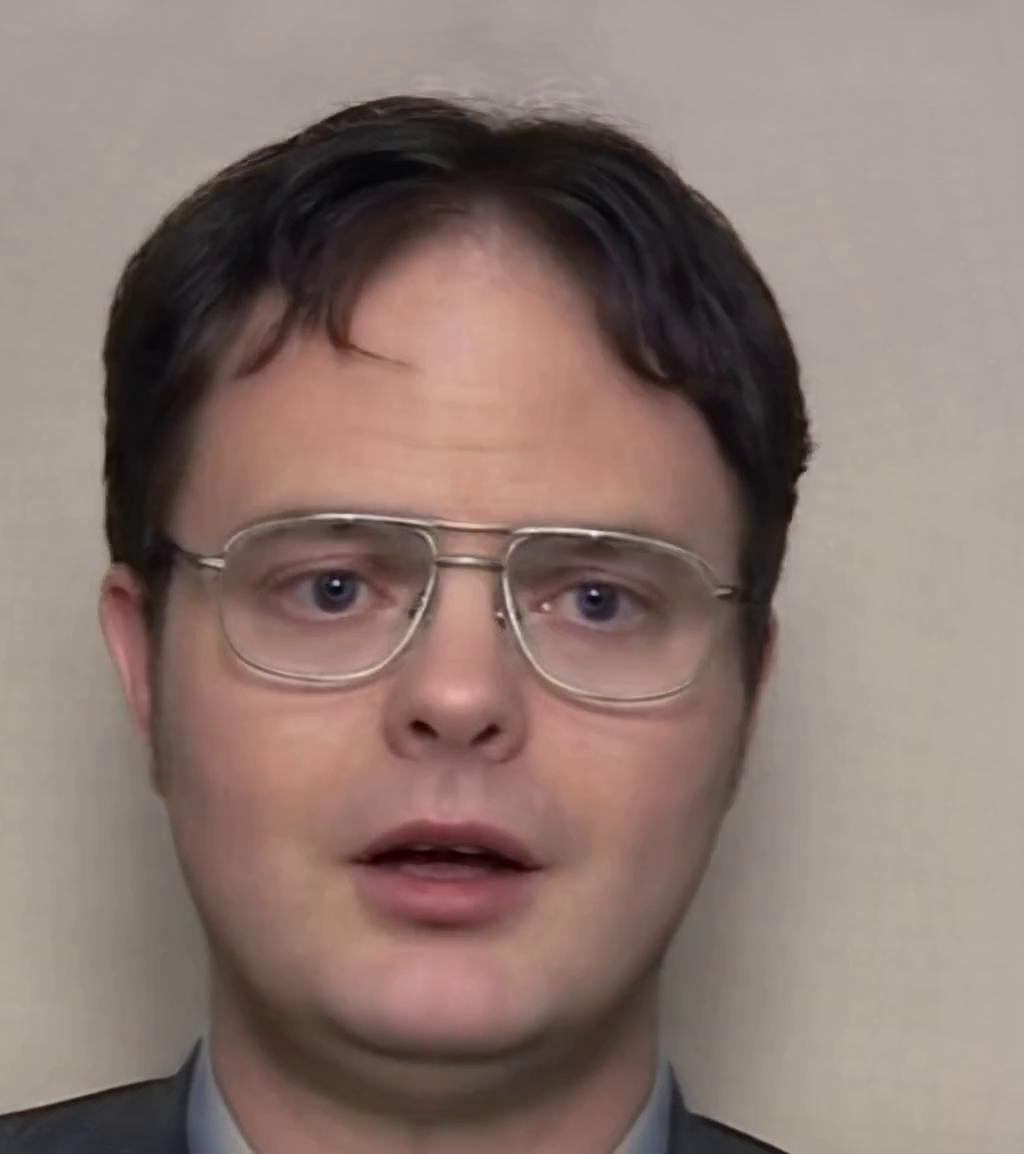} & 
        \includegraphics[width=0.215\columnwidth]{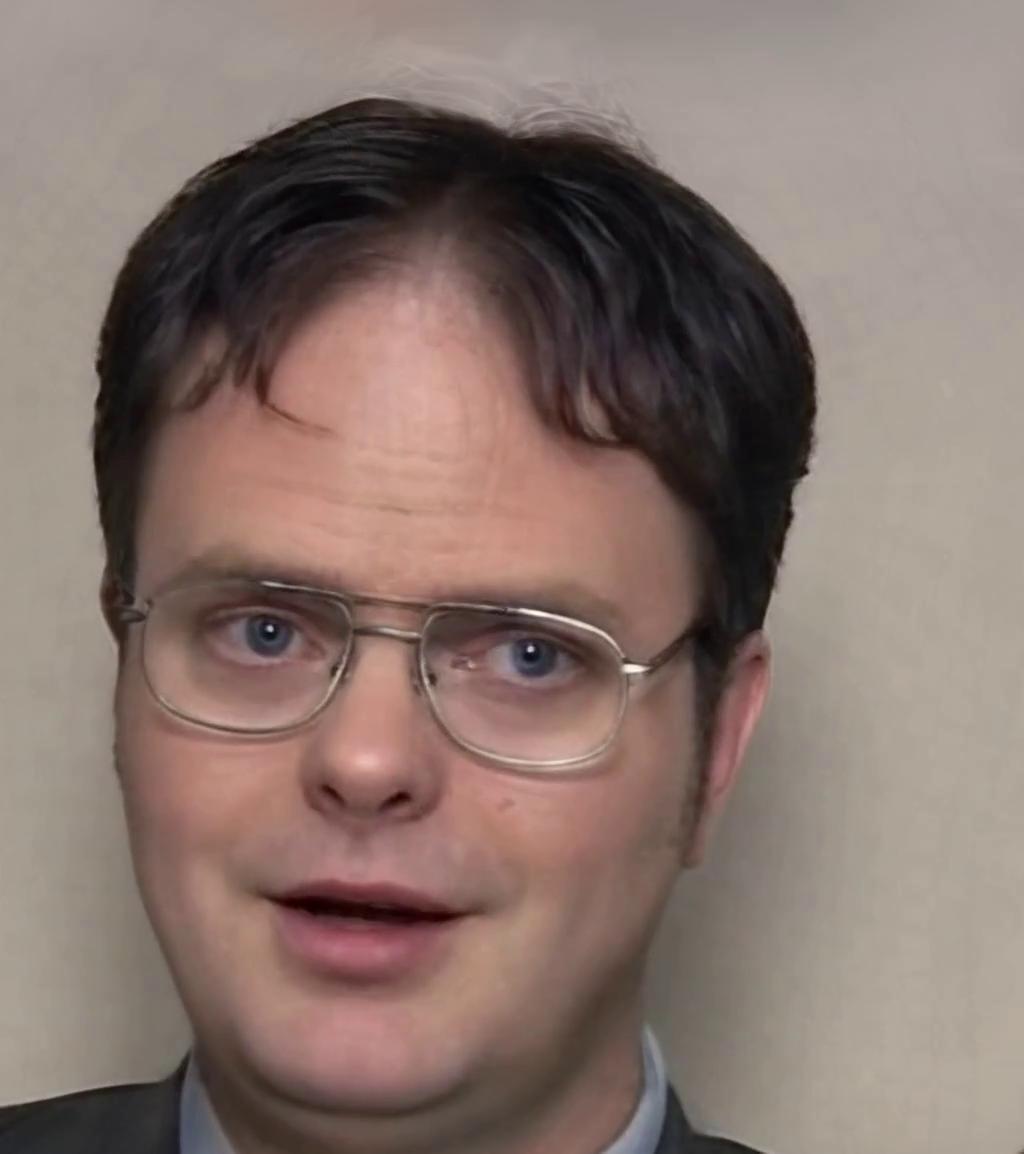} & 
        \includegraphics[width=0.215\columnwidth]{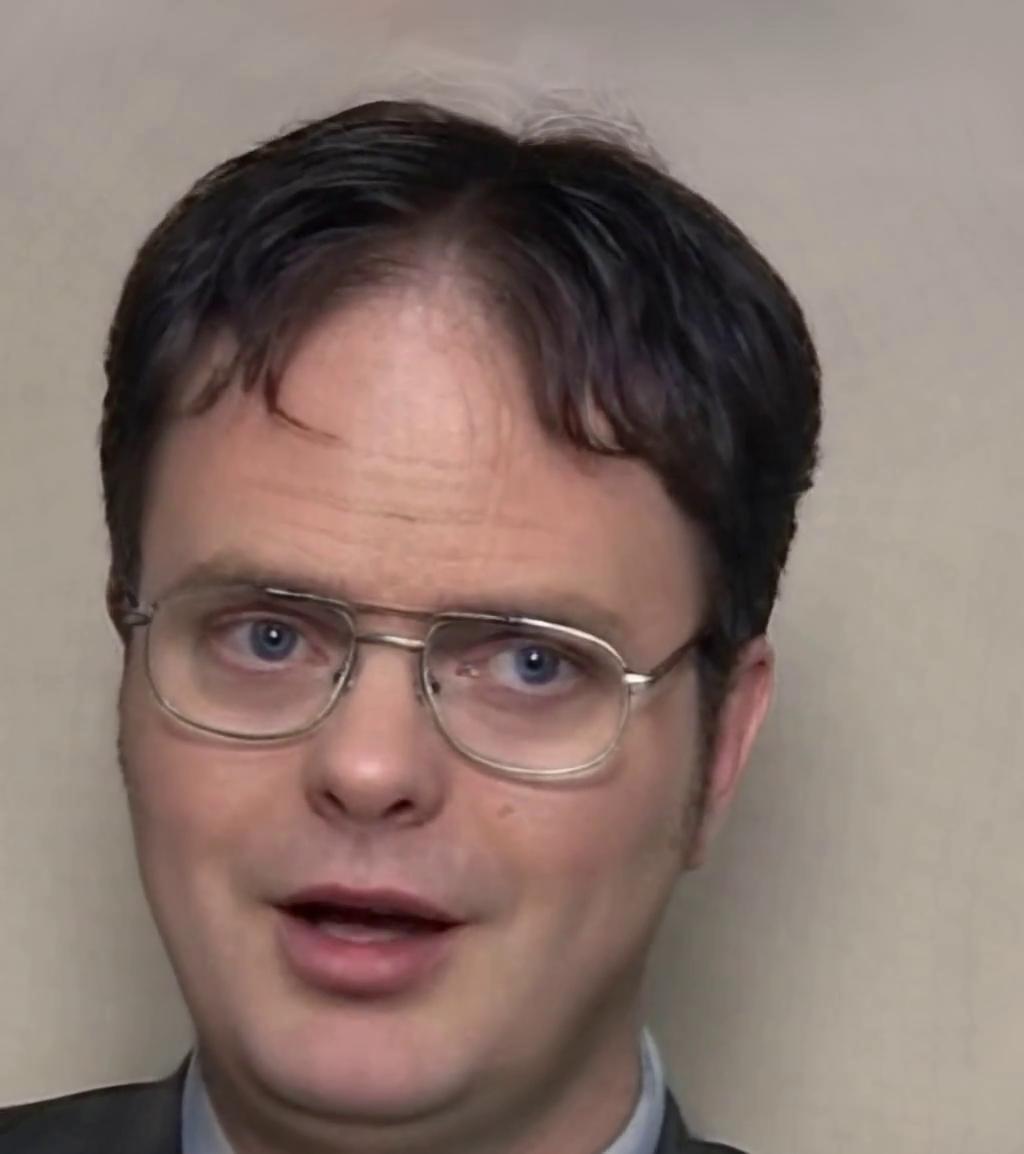} \\

	\end{tabular}
	}
	
	\vspace{-0.3cm}
	\caption{
	Sample results of our full video encoding pipeline with StyleGAN3 with our field of view expansion trick. Observe in the top video sequence how the expansion allows for completing ``cut-off'' regions of the original reconstruction (e.g., the chin of the woman or the hair of the man). Full videos are provided in the supplementary materials.
	}
	\label{fig:video_results_wide_supplementary}
\end{figure*}